\documentclass[lettersize,journal]{IEEEtran}
\usepackage{amsmath,amsfonts}
\usepackage{algorithmic}
\usepackage{array}
\usepackage[caption=false,font=normalsize,labelfont=sf,textfont=sf]{subfig}
\usepackage{textcomp}
\usepackage{stfloats}
\usepackage{url}
\usepackage{verbatim}
\usepackage{graphicx}
\def\BibTeX{{\rm B\kern-.05em{\sc i\kern-.025em b}\kern-.08em
    T\kern-.1667em\lower.7ex\hbox{E}\kern-.125emX}}
\usepackage{balance}
\usepackage{cite}

\usepackage{amssymb}
\usepackage[colorlinks,linkcolor=red,anchorcolor=blue,citecolor=green]{hyperref}
\usepackage{multirow}
\usepackage{tabularx}
\usepackage{url}
\usepackage{verbatim}
\usepackage{mathtools}
\usepackage{amsmath}
\usepackage{xparse}
\usepackage{float}

\newcommand{\tao}[1]{\textcolor{magenta}{#1}}
\newcommand{\emp}[1]{\textcolor{cyan}{#1}}

\newcommand{\revised}[1]{\textcolor{black}{#1}}

\newcommand{\ryn}[1]{\textcolor{black}{#1}}

\begin{document}
\title{Rain Removal from Light Field Images with 4D Convolution and Multi-scale Gaussian Process}
\author{Tao Yan*, \textit{Member, IEEE}, \ \ Mingyue Li, \ \ Bin Li, \ \ Yang Yang, \ \ Rynson W.H. Lau, \textit{Senior Member, IEEE}
\thanks{This work was supported by the National Natural Science Foundation of China (Grant No. 61902151) and the Natural Science Foundation of Jiangsu Province, China (Grant No. BK20170197), and two Strategic Research Grants from City University of Hong Kong (Ref.: 7005674 and 7005843).}
\thanks{Tao Yan, Mingyue Li and Bin Li are with the School of Artificial Intelligence and Computer Science, Jiangnan University, Wuxi, China.}
\thanks{Yang Yang is with the Department of Computer Science, Jiangsu University, Zhenjiang, China.}
\thanks{Rynson W.H. Lau is with the Department of Computer Science, City University of Hong Kong, Hong Kong.}
\thanks{Corresponding author is Tao Yan, E-mail: yantao.ustc@gmail.com.}
}

\markboth{Journal of \LaTeX\ Class Files,~Vol.~18, No.~9, September~2020}%
{How to Use the IEEEtran \LaTeX \ Templates}

\maketitle

\begin{abstract}
Existing deraining methods focus mainly on a single input image. However, with just a single input image, it is extremely difficult to accurately detect \ryn{and remove rain streaks, in order to restore a rain-free image}. In contrast, a light field image (LFI) embeds abundant 3D structure and texture information of the target scene by recording the direction and position of each incident ray via a plenoptic camera. LFIs are becoming popular in the computer vision and graphics communities. \revised{However, making full use of the abundant information available from LFIs, such as 2D array of sub-views and the disparity map of each sub-view, for effective rain removal is still a challenging problem.}
In this paper, we propose a novel method, 4D-MGP-SRRNet, for rain streak removal from LFIs. Our method takes as input all sub-views of a rainy LFI. To make full use of the LFI, it adopts 4D convolutional layers to simultaneously process all sub-views of the LFI. In the pipeline, the rain detection network, MGPDNet, with a novel Multi-scale Self-guided Gaussian Process (MSGP) module is proposed to detect \revised{high-resolution} rain streaks from all sub-views of the input LFI \revised{at multi-scales}. Semi-supervised learning is introduced \revised{for MSGP} to accurately detect rain streaks by training on both virtual-world rainy LFIs and real-world rainy LFIs at multi-scales via computing pseudo ground \revised{truths} for real-world rain streaks. \ryn{We then feed all sub-views subtracting the predicted rain streaks} into a 4D \revised{convolution-based Depth Estimation Residual Network (DERNet)} to estimate the depth maps, \revised{which are later converted into fog maps}. Finally, all sub-views concatenated with the corresponding rain streaks and fog maps are fed into a \revised{powerful} rainy LFI restoring model based on the adversarial recurrent neural network to progressively eliminate rain streaks and recover the rain-free LFI. Extensive quantitative and qualitative evaluations conducted on both synthetic LFIs and real-world LFIs demonstrate the effectiveness of our proposed method.
\end{abstract}


\section{Introduction}

\IEEEPARstart{A}{dverse} weather conditions, such as raining, snowing and fogging, can degrade the quality of outdoor captured images/videos and the performances of computer vision systems. Rain streaks are semi-transparent and have various sizes, directions and even appearances (e.g., strips and fog/mist) depending on their distances from the camera. Thus, to accurately detect rain streaks and eliminate them \revised{for} \ryn{a clean rain-free image} is a nontrivial task. To solve this problem, researchers have proposed many rain streak removal methods based on classical optimization methods or deep neural networks.

\begin{figure}[tb]
	\centering
	\includegraphics[width=0.46\textwidth]{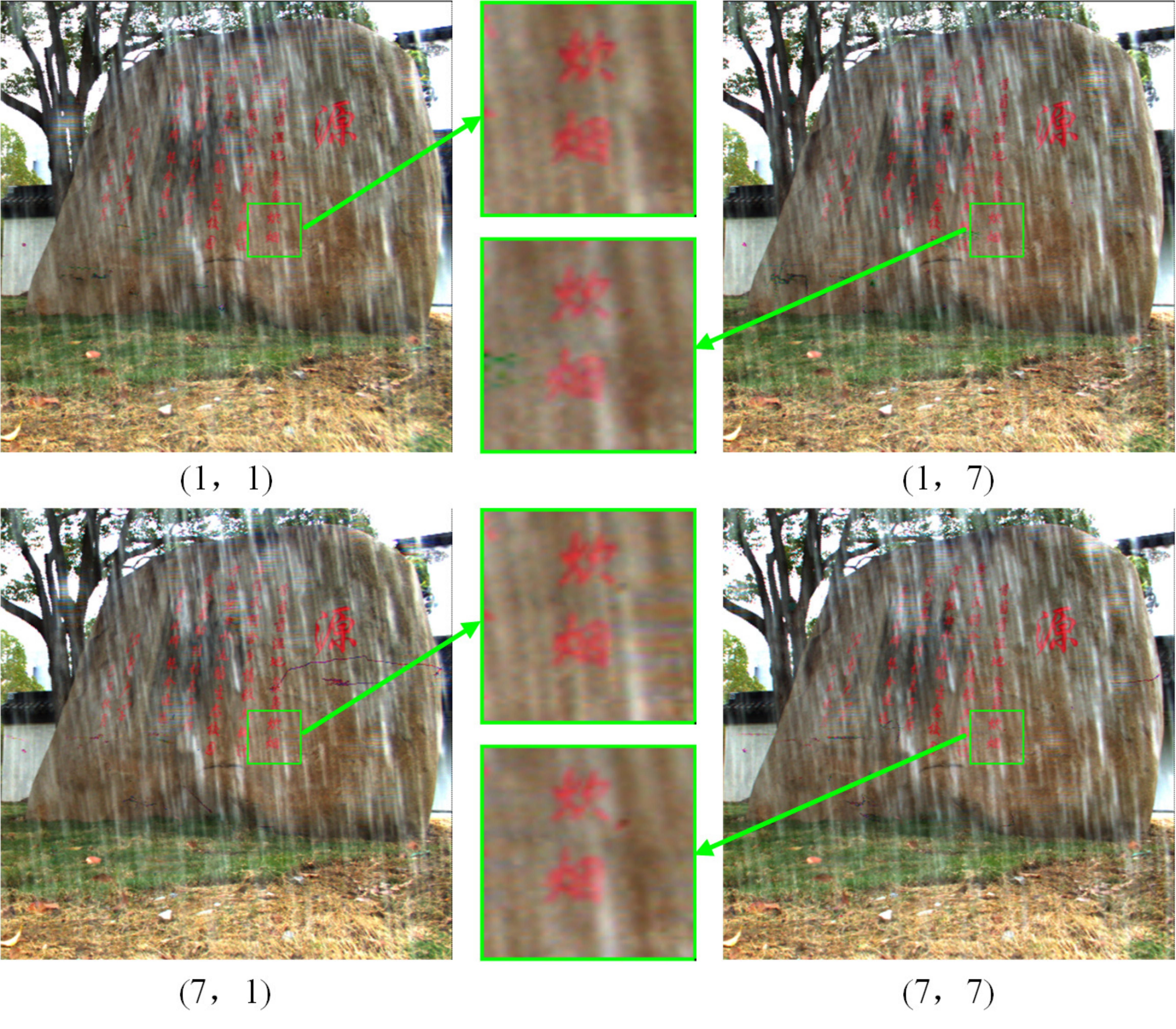}
    \vspace{-2mm}
	\caption{Sub-views of a real-world rainy LFI. We can see that the same rain streaks appear at different positions and occlude different regions of the background in different sub-views.}
	\label{fig:rainstreak}
\end{figure}

An LFI can be decoded into a set of sub-views of different perspectives, and accurate disparity maps can be inferred from it. \revised{Thus, we believe that rain removal on LFIs~\cite{ding21}, and stereo~\cite{zhang2022stereoderaining} or Dual-Pixel~\cite{li2022dual} images \ryn{may produce much higher quality de-rained results than those on single images,} by properly exploring and exploiting multi-views and depth maps.} Leveraging the advantages of plenoptic cameras, LFIs could also be used to improve the performances of a variety of computer vision applications, such as surveillance and autonomous navigation systems. %However, rain streaks can also degrade the visibility/quality of rainy LFIs as in 2D images, and impact the depth map prediction from LFIs.
Meanwhile, we \revised{observe} that the same rain streaks always appear in different areas of the background in different sub-views, which means that the rain-free LFI could be well restored from a rainy LFI by fully exploiting the abundant information from an LFI~\cite{ding21}, as shown in Fig.~\ref{fig:first_figure} and~\ref{fig:rainstreak}. In this paper, we propose a novel neural framework for removing rain streaks from a rainy LFI to recover a rain-free LFI.

\footnotetext[1]{Our dataset and models are available at \url{https://github.com/YT3DVision/4D-MGP-SRRNet}.}

\newcommand{\subwidth}{0.46}
\begin{figure}[t]
	\renewcommand{\tabcolsep}{1.0pt}
	\renewcommand\arraystretch{0.8}
	\begin{center}
		\begin{tabular}{cc}
			\includegraphics[width=\subwidth\linewidth]{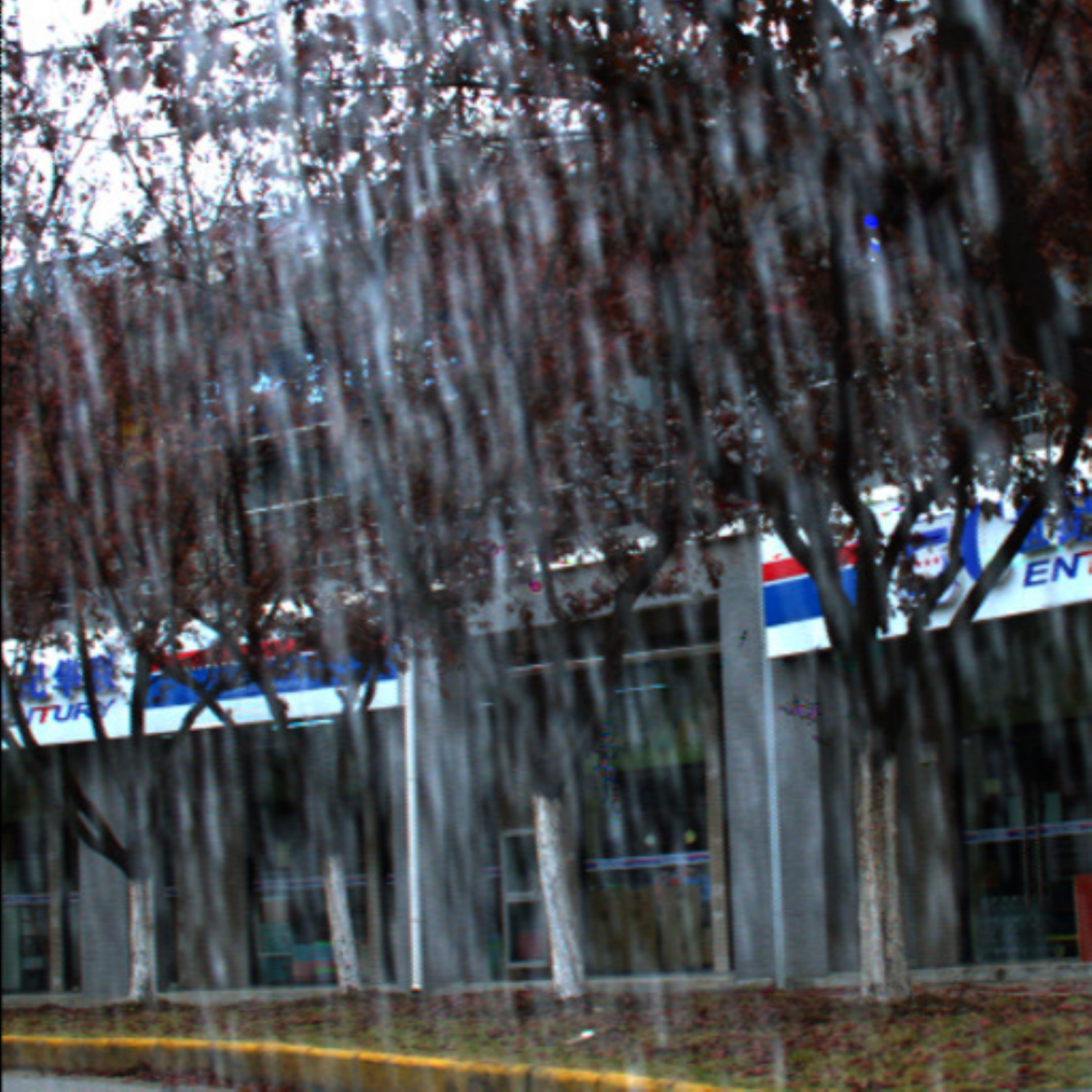} &
			\includegraphics[width=\subwidth\linewidth]{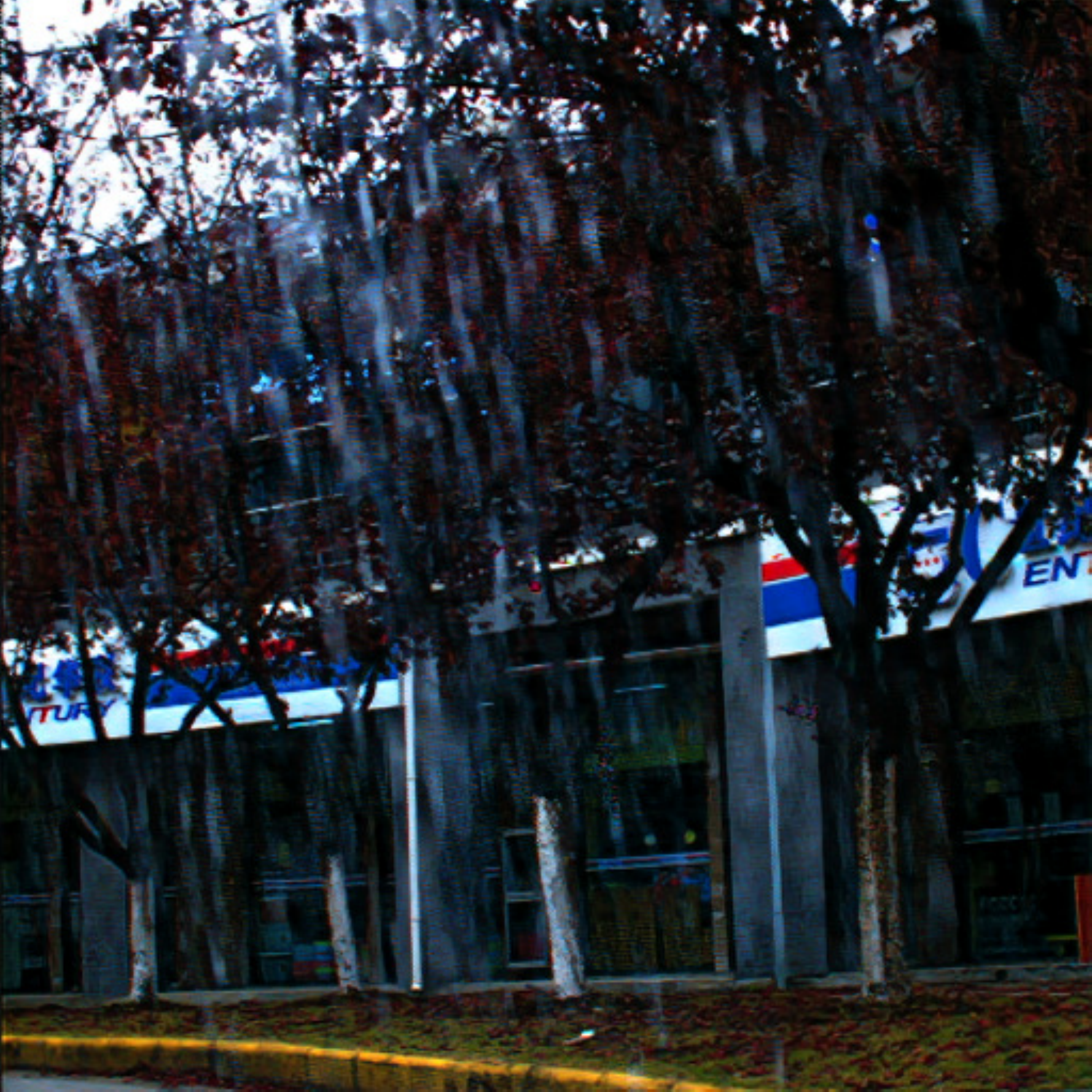}\\

    		\scriptsize{(a) Input LFI (central-view)} &
            \scriptsize{(b) Li et al.~\cite{Li19a}'s Result} \\
			
			\includegraphics[width = \subwidth\linewidth]{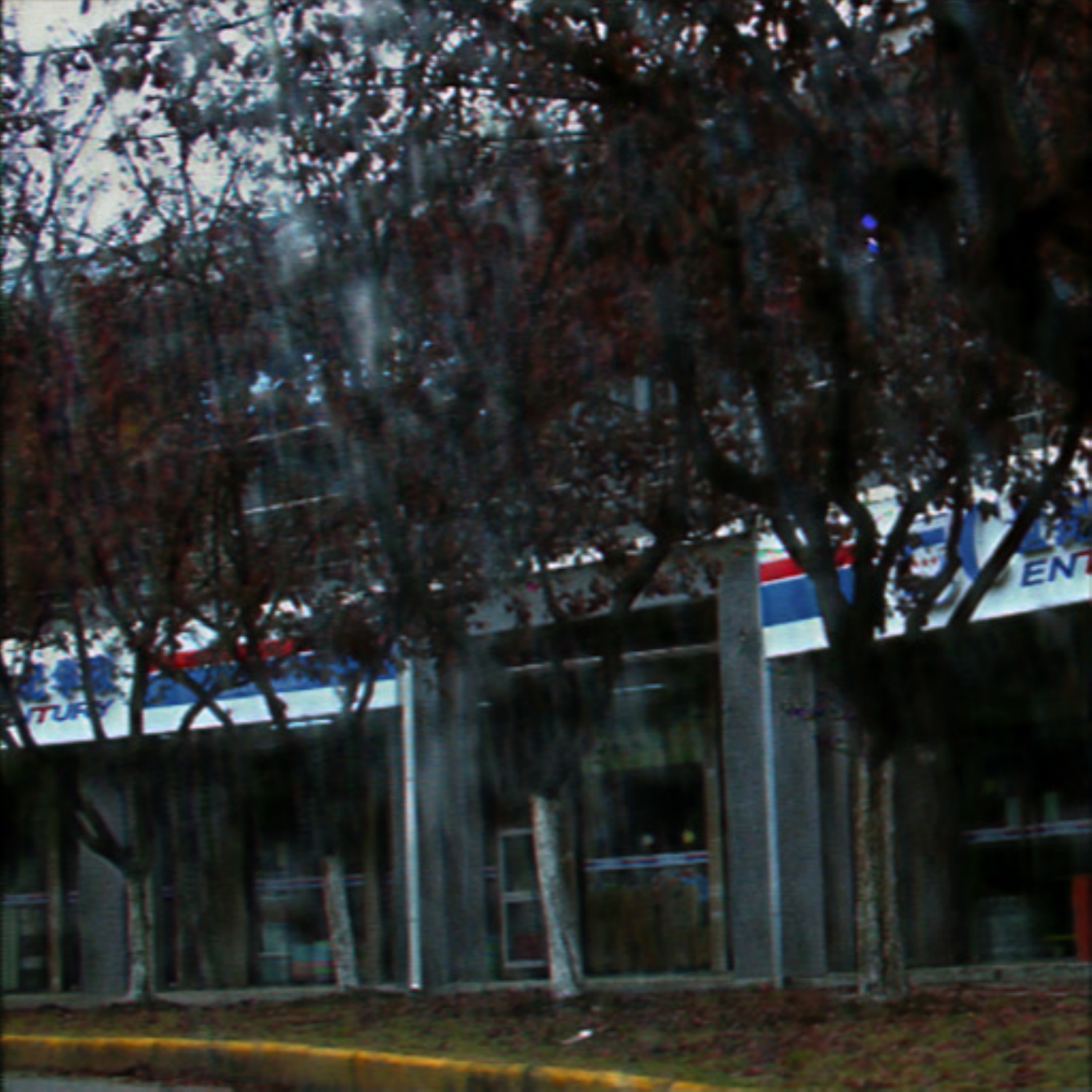}&
			\includegraphics[width = \subwidth\linewidth]{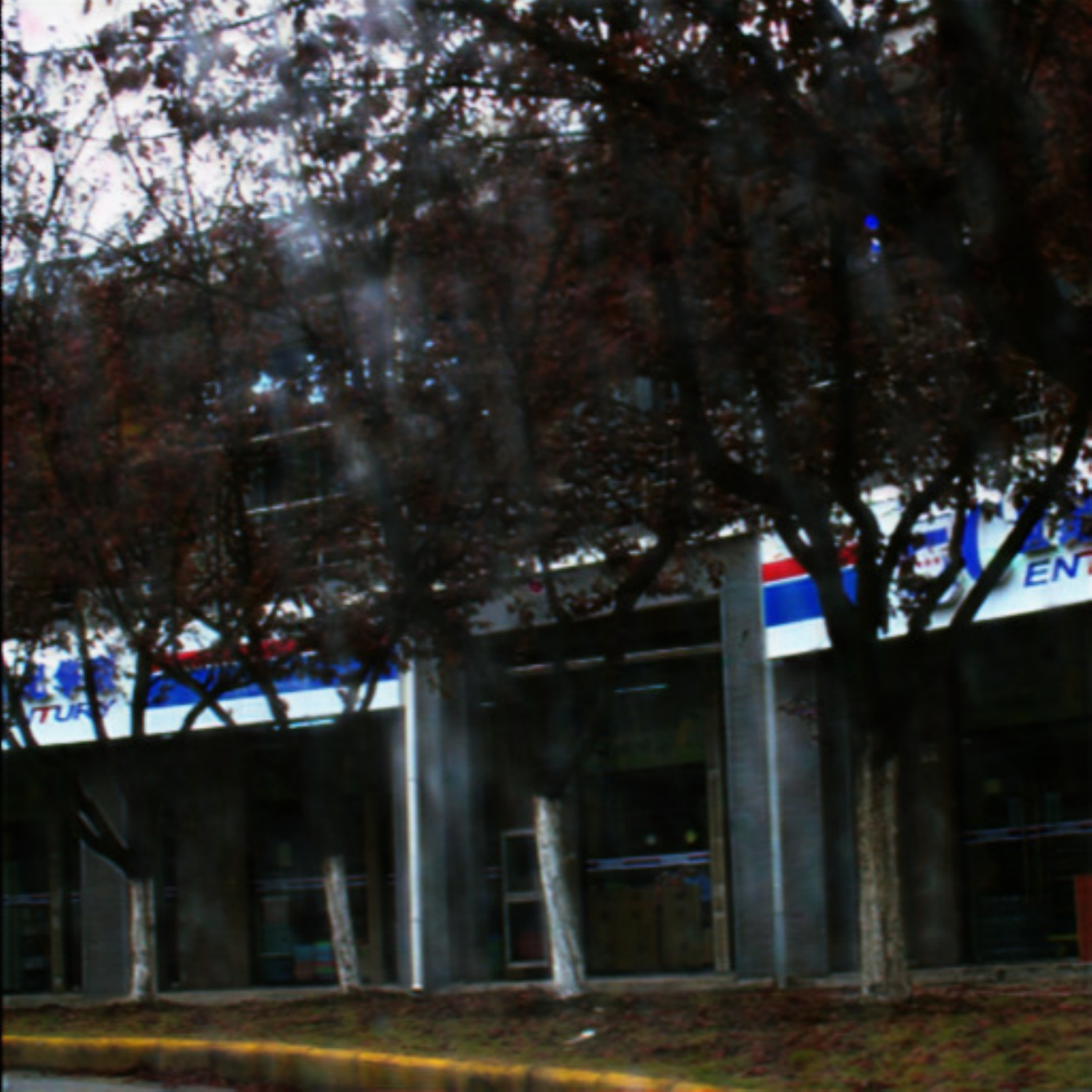}\\

            \scriptsize{(c) Yasarla et al.~\cite{yasarla20}'s Result}&
            \scriptsize{(d) Our Result}\\

		\end{tabular}
	\end{center}
    \vspace{-2mm}
	\caption{Recovered real-world de-rained LFIs (only center views are shown), obtained by two state-of-the-art methods~\cite{Li19a,yasarla20} and our method.}
    \label{fig:first_figure}
    \vspace{-4mm}
\end{figure}

%For the rain removal methods,
\ryn{Traditional} image-based rain streak removal methods~\cite{Chen13,Luo15,Yu16,li14} usually perform poorly under complex rain conditions. Recently, deep-learning-based methods~\cite{Li19a,yasarla20,Zhang18,Hu19,Zhu20,que20,ren20,Yang20b,Ahn2021,fu21,huang21,wei21,gou2020,wang2022uformer,vala2022transweather,zamir22,xiao22,li2022,zhang2022,wang2022deraing} have shown huge advantages. Although these methods can remove most of the rain streaks, there are always some tiny or large rain streaks that cannot be clearly removed. In general, these methods have three main limitations. First, none of them can accurately detect various types of real-world rain streaks~\cite{li2021comprehensive} that have different directions/shapes and varying degrees of opacity, although accurately detecting rain streaks is a crucial step in the rain streak removal task. In view of the fact that rain streaks are continuous and translucent stripes, which resemble the shape of small blood vessels, accurate rain streak detection is a challenging task. Second, distant dense rain streaks are more like fog/mist. \ryn{Existing methods still have difficulties in eliminating such dense rain streaks.} \ryn{Third}, due to the domain gap between synthetic and real-world rainy images, most algorithms trained on synthetic images do not work well on real scenes. Fig.~\ref{fig:first_figure} shows an example.

Recently, Wei et al.~\cite{wei19} and Ding et al.~\cite{ding21} propose using the Gaussian Mixture Model (GMM) to construct semi-supervised learning methods for modeling rain streaks in real-world images by approximating a feature vector of real-world rain streaks as the weighted average of a set of feature vectors of the rain streaks in synthetic virtual-world images. The consistency between the two kinds of rain streaks is ensured by minimizing the Kullback-Leibler (KL) divergence between their distributions. Both methods use real-world images to generalize the rain streak detection \ryn{ability from} the virtual-world domain to the real-world domain. However, in the early stages of training, the learned GMM parameters are usually inaccurate. Thus, it may not be appropriate to minimize the KL divergence in order to reduce the difference between the two distributions, as it would lead to sub-optimal performances. To overcome this problem, Yasarla et al.~\cite{yasarla20} propose using the Gaussian Process (GP) for rain streak detection, which is a non-parametric model to predict real-world rain streaks based on virtual-world rain streaks. It achieves better performances \revised{than~\cite{wei19}}. To improve the generalization of our network for deraining on real-world rainy LFIs, we also adopt GP to realize semi-supervised learning by supervising the feature vector of real-world rain streaks as a weighted average of the feature vectors of synthetic rain streaks. \revised{On the other hand, to overcome the problem of insufficient utilization of information available from an LFI (i.e., only utilizing an array of sub-views in the same row/column of an LFI, called a 3D EPI, each time) of the former method~\cite{ding21}, we introduce 4D convolution to make full use of all sub-views of an LFI for effective and efficient LFI deraining.}

\begin{figure*}%[h]
	\centering
	\includegraphics[width=1.0\textwidth]{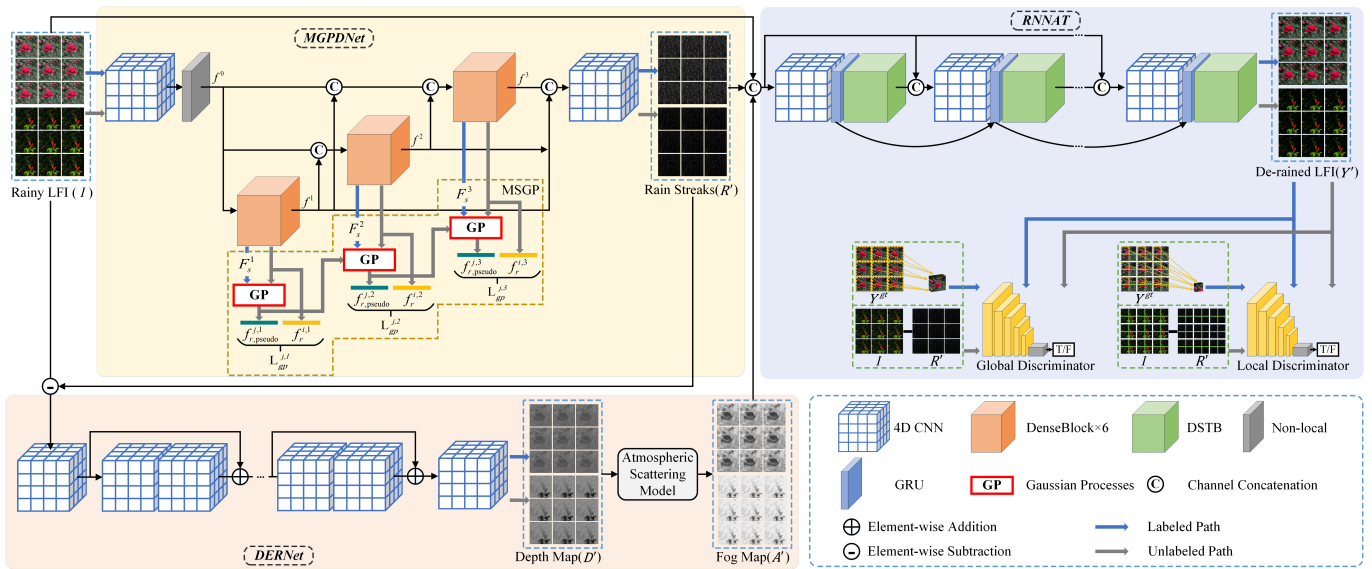}
    \vspace{-4mm}
	\caption{The architecture of our proposed 4D-MGP-SRRNet, \revised{which contains three parts (i.e., sub-networks)}. First, the \textit{Multi-scale Gaussian Process Dense Network (MGPDNet)} takes as input all sub-views of a rainy LFI, (i.e., a 5D data of cropped patch stacks), and works in a semi-supervised learning manner on multi-scale features to detect \revised{high-resolution} rain streaks. Second, estimated rain streaks \ryn{subtracted} from sub-views are fed into the 4D \textit{Depth Estimation Residual Network (DERNet)} to estimate the depth maps. An atmospheric scattering model is also adopted to convert the predicted depth maps into fog maps. Finally, all rainy sub-views concatenated with the corresponding rain streaks and fog maps estimated above are fed into the \textit{Recurrent Neural Network with Adversarial Training (RNNAT)} to progressively remove rain streaks and recover the rain-free LFI.}
	\label{fig:network_architecture}
\end{figure*}

In this paper, we propose a novel \ryn{CNN-based} method for rain streak removal from LFIs. The architecture of our method is shown in Fig.~\ref{fig:network_architecture}. Our network takes all sub-views of a rainy LFI as input. 4D convolutional layers that can simultaneously process all sub-views are \revised{introduced} for our network. In addition, semi-supervised learning with MSGP (Multi-scale Self-guided GP) module is also \revised{introduced} to improve the effectiveness and generalization of our network for rain streak detection and rain-free LFI recovery, by training on both synthetic and real-world LFIs.

Our proposed network consists of three \revised{sub-networks}. First, a Multi-scale Gaussian Process based Dense Network (MGPDNet) is constructed to extract \revised{high-resolution} rain streaks from all sub-views. A 4D Depth Estimation Residual Network (DERNet) is then introduced to estimate the depth maps for all sub-views subtracting the estimated rain streaks. The predicted depth maps are later converted into fog maps, which can be used to remove distant dense rain streaks like fog/mist. Finally, all input rainy sub-views concatenated with the corresponding rain streaks and fog maps are fed into a Recurrent Neural Network with Adversarial Training (RNNAT) to progressively remove the rain streaks and recover the rain-free sub-views. The main contributions of our work can be summarized as:
\begin{itemize}
	\item We propose a novel 4D Convolution and Multi-scale Gaussian Process based Semi-supervised Rain Removal Network (4D-MGP-SRRNet), which takes all sub-views of a rainy LFI as input to recover a rain-free LFI.
	\item We propose a Multi-scale Self-guided GP (MSGP) module based semi-supervised learning strategy for accurate rain streak detection and rain-free LFI recovery.
	\item We propose a new rainy LFI \revised{dataset}, Rainy LFI with Motion Blur (RLMB), which contains $400$ sets of synthetic rainy LFIs with ground truth rain-free LFIs and $200$ sets of real-world rainy LFIs. Motion blur is first carefully considered in realistic rain streak generation for rainy LFI synthesis and rain streak detection/removal from LFIs.
\end{itemize}

\section{Related Work}
In the past few years, with the popularity of deep learning, great progress has been achieved for rain removal from images~\cite{li2021comprehensive}. In this section, we review works related to rain streak removal from images, videos and LFIs.

\subsection{Rain Streak Removal From Images}
\label{subsec:rain_removal_image}
Traditional deraining methods usually separate a rainy image into rain-streak layer and rain-free layer based on image priors and classical optimization models such as low-rank model~\cite{Chen13}, sparse coding~\cite{Luo15}, Gaussian Mixture Model~\cite{Yu16}. These methods are time-consuming and their performances are unsatisfactory on complex real-world rainy images. Recently, deep-learning based rain removal methods~\cite{Yang17,Zhu17b,Zhang18,Li18,Li2018non,Hu19,Li19a,wang19b,wei19,zhu19b,jiang20,ren20,lin20,yasarla20,Zhu20,zamir2021,hu2021,wang2020b,Zhang20,Ahn2021,wang2021,Luo20,jiang2020,huang2021,wei21,fu21,chen2021,liu21}\revised{~\cite{gou2020,wang2022uformer,vala2022transweather,zamir22,xiao22,li2022,wang2022deraing}} have attracted much attention and shown impressive performances.

Zhang et al.~\cite{Zhang18} propose a residual-aware rain-density classifier to measure the density of rain streaks, and a multi-stream densely connected deraining network to extract features of rain streaks at different scales and shapes.
Jiang et al.~\cite{jiang20} propose a multi-scale progressive fusion network for collaborative representing rain streaks. %, which utilizes recurrent calculations to capture the global texture for similar rain streaks at different positions and multi-scale pyramid structure to guide the fine fusion of rain streaks across different scales.
Although these two methods that rely on multi-scale networks to remove rain streaks can restore the details of the rain-free background excellently, they often fail to completely remove all rain streaks.
Recently, %the Conditional GAN, physical rainy image model, and motion blur kernel based rain streak removal methods~\cite{Zhang20,wang2020b,wang2021} are proposed.
Zhang et al.~\cite{Zhang20} propose a conditional GAN-based framework for image deraining. %The densely connected generator network was constructed to enable steady gradient flow, and a multi-scale discriminator was introduced to utilize features from multi-scales in order to authenticate the de-rained image.
Wang et al.~\cite{wang2020b} propose a new physical rainy image model to explicitly model the degradation of the haze-like effect on rainy images. Wang et al.~\cite{wang2021} propose a kernel-guided convolutional neural network to explore motion blur and line pattern appearances of rain streaks.

RNN-based methods have also been proposed to progressively remove rain streaks by decomposing rain streaks into multiple rain streak layers with different sizes, directions and densities.
Li et al.~\cite{Li18} propose the first deep recurrent network for image deraining, which uses contextual dilated  networks to acquire large receptive field and recurrent neural networks to decompose the rain removal task into multiple stages.
Yang et al.~\cite{Yang17,Yang20b} propose a recurrent multi-task deep learning network to learn the binary rain streak map, the appearance of rain streaks, and the rain-free background in each recurrence. %They first obviously modeled and handled the rain streaks accumulation, termed fog/mist.
Zamir et al.~\cite{zamir2021} propose a multi-stage architecture for progressive image restoration, which balances spatial details and high-level contextualized information. %It extracts the contextualized features using encoder-decoder architectures in earlier stages and later in last stage combines them to generate accurate results at the original resolution.

Memory neural networks and attention blocks are also used to remove rain streaks from images~\cite{ren20,Zhu20,lin20,Li2018non,jiang2020,Ahn2021}.
Ren et al.~\cite{ren20} propose the bilateral recurrent network, in which two recurrent networks having recurrent layers (convolutional LSTM) in each stage are coupled to simultaneously obtain rain streak layer and clean background layer. % separately by each SRN.
Bilateral LSTMs are proposed to propagate deep features across stages and bring interplay between the two networks.
Zhu et al.~\cite{Zhu20} propose a gated non-local deep residual learning framework to remove rain from images. % by learning a novel channel-wise gated prediction model to adaptively adjust the amount of global residual passed for generating the final de-rained image.
Similarly, Lin et al.~\cite{lin20} propose a network based on sequential dual attention blocks for rain streak removal.
Li et al.~\cite{Li2018non} propose a non-locally enhanced encoder-decoder network for image deraining, which consists of a series of non-locally enhanced dense blocks and adopts pooling indices guided decoding scheme.
Jiang et al.~\cite{jiang2020} propose an attention-guided deraining network to model multiple rain streak layers under the multiple network stages.
%An improved non-local block efficiently exploits the self-similarity of rain streaks at the begining, and a mixed attention mechanism is proposed to guide the fusion of estimated rain layers.
Ahn et al.~\cite{Ahn2021} propose a rain streak removal network using the maximum color channel selection. %It consists of proposed elementwise attentive gating blocks to selectively passes the desired components from the input feature maps by applying different weights to each element.
Most recently, Fu et al.~\cite{fu21} propose the first GNN-based model for iteratively removing rain streaks.
%They designed two graphs that one models the global spatial relationship between pixels in the feature, while the other models the interrelationship across the channels.
%However, none of these methods obviously take into account the far away dense rain streaks similar to fog/mist and clearly remove them.

Depth map estimation for rain streak removal is also studied~\cite{Hu19,hu2021,Li19a}.
Hu et al.~\cite{Hu19} investigate visual effects of rain streaks subjected to scene depth, and propose a rainy image generation model that is composed of rain streaks and fog. A deep neural network is proposed to estimate depth-attentional features by leveraging the depth-guided attention mechanism and regressing on a residual map to restore the de-rained image. However, since the depth maps predicted from rainy images are unreliable, it may seriously affect the quality of de-rained results.
Recently, Hu et al.~\cite{hu2021} improve the network speed for real-time deraining.
Li et al.~\cite{Li19a} propose a neural network with a physics-based backbone to separate the entangled rain streaks and rain accumulation, and  a depth-guided GAN refinement step to eliminate heavy rain steaks and fog. However, the network fails to eliminate distant rain streaks, especially under heavy rain conditions. %Overall, rain streaks could not be sufficiently removed from rainy images by these methods.

Recently, several works~\cite{wang19b,wei19,yasarla20,zhu19b,Luo20,wei21}\revised{\cite{wang2022deraing}} improve the deraining performances on real-world scenes by adopting semi-/un-supervised learning and/or real-world rainy images.
Wang et al.~\cite{wang19b} propose a semi-automatic method to incorporate temporal priors and human supervision to generate high-quality de-rained images from each input sequence of real rainy images. %By utilizing the method, a large set of rain/rain-free image pairs covering a wide range of natural rain scenes is constructed. Then, Spatial Attentive Network is proposed to remove rain streaks in a local-to-global manner.
Wei et al.~\cite{wei19} propose a deraining network to solve the domain adaption problem of transferring from synthetic images to real-world images by designing a semi-supervised learning strategy. % using both supervised data and unsupervised data for training.
Yasarla et al.~\cite{yasarla20} propose a semi-supervised learning framework based on the Gaussian Process, which enables the network to use synthetic datasets for learning and unlabeled real-world images for fine-tuning.
Zhu et al.~\cite{zhu19b} propose the first unsupervised rain streak removal method, which alleviates the popular paired training constraints by introducing a physical model that explicitly learns a recovered image and corresponding rain streaks from the differentiable programming perspective. %The proposed end-to-end adversarial model, RainRemoval-GAN, consists of a novel multi-scale attention memory generator and a novel multi-scale deeply supervised discriminator.
CycleGAN~\cite{Zhu17b} is also a popular network for unsupervised raindrop removal~\cite{Luo20} and rain streak removal~\cite{wei21}. Luo et al.~\cite{Luo20} propose a weakly supervised learning based raindrop removal network, which can be trained with both pairwise and unpaired samples. Wei et al.~\cite{wei21} explore the unsupervised single-image rain removal problem using unpaired data, and propose an unsupervised network called DerainCycleGAN for rain streak removal and generation. %Particularly, an unsupervised rain attentive detector was designed to progressively detect rain streaks on both rainy and rain-free images.
\revised{Huang et al.~\cite{huang2021} propose an encoder-decoder network augmented with a self-supervised memory module, which enables the network to exploit the properties of rain streaks from both synthetic and real data.}
\revised{Wang et al.~\cite{wang2022deraing} propose a task transfer learning mechanism to learn favorable representations from real data through task transfer to improve deraining generalization to real scenes.}
\revised{Notably, Li et al.~\cite{li2021} propose an Unsupervised and Untrained Single-Image Dehazing Neural Network (YOLY), which performs image dehazing by only using the information contained in the observed single hazy image and avoids training itself on an image set with ground-truth.}

While these semi-supervised/unsupervised-based \revised{single-image deraining} methods may help improve the performances on various real-world images, the de-rained images generated by these methods still contain some large/tiny rain streaks, especially for input images of challenging rain scenarios.

%\needconfirm{~\cite{gou2020,li2021,li2022}} \\

\revised{More recently, with the Transformer model mitigating the shortcomings of CNNs, such as limited receptive field and inadaptability
to input content, several state-of-the-art networks~\cite{chen2021,liu21,wang2022uformer,vala2022transweather,zamir22,xiao22,li2022} are proposed for image restoration (deraing, denoising, super-resolution, motion deblurring, defocus deblurring). Wang et al.~\cite{wang2022uformer} propose Uformer with a U-shaped structure formed by locally-enhanced window (LeWin) Transformer blocks that perform non-overlapping window-based self-attention instead of global self-attention. However, Uformer can only enable communications between adjacent windows.
Liu et al.~\cite{liu21} propose the first Swin Transformer-based network called SwinIR for image super-resolution and image denoising. The Residual Swin Transformer block (RSTB) composing of several Swin Transformer layers together with a residual connection is proposed for deep feature extraction in this network. Valanarasu et al.~\cite{vala2022transweather} propose TransWeather, which is a Transformer-based end-to-end model with just a single encoder and a decoder for restoring images degraded by any weather \ryn{conditions}. Specifically, intra-patch transformer blocks within Transformer encoder are proposed to enhance attention inside the patches, and Transformer decoder with learnable weather type embeddings is introduced to learn the weather degradation type and uses this information to restore clean images. The latest works~\cite{zamir22,xiao22} are all Swin Transformer-based networks.}

\revised{Zamir et al.~\cite{zamir22} propose an efficient hierarchical architecture called Restormer for high-resolution image restoration by making several key designs in building blocks, such as multi-head attention and feed-forward network (FFN). Xiao et al.~\cite{xiao22} propose an effective and efficient Transformer-based encoder-decoder architecture for single image deraining, with each encoder/decoder block constructed by several window-based transformer modules that capture local relationships within the local window and a spatial transformer module that complements the locality by modeling cross-window dependencies. In addition, Li et al.~\cite{li2022} propose a contrastive-learning-based All-In-One image restoration network, AirNet, which could recover images from unknown corruption types and levels. Unlike the networks~\cite{chen2021,li2020all} that treat multiple degradations as a multi-task learning problem with multiple input and output heads, all these networks~\cite{liu21,wang2022uformer,vala2022transweather,zamir22,xiao22,li2022} are single pass networks that do not differentiate between different corruption types and ratios.}

%\label{sec:Real-world-like-data}

\subsection{Rain Streak Removal From Frame Sequences}
\label{subsec:rain_removal_video}
Video deraining methods~\cite{Li18v,Liu18v,Yang19v,Yang20v,li21,zhang2022,yang2022} exploit spatial and temporal redundancies in frame sequences. They are important in practical applications.

Li et al.~\cite{Li18v} propose a multi-scale convolutional sparse coding model, which uses multiple convolution filters convolved on the sparse feature map, to deliver repetitive local patterns of rain streaks, and further uses multi-scale filters to represent rain stripes of different scales. Liu et al.~\cite{Liu18v} build a joint recurrent deraining network. The network seamlessly integrates rain degradation classification, spatial texture appearances-based rain streaks removal, and  temporal coherence-based background detail reconstruction.
Yang et al.~\cite{Yang19v} propose a two-stage recurrent framework built with dual-level flow regularizations to perform the inverse recovery process of the rain synthesis model for video deraining.
Yang et al.~\cite{Yang20v} propose a two-stage self-learned framework for deraining based on both temporal correlation and consistency.
Li et al.~\cite{li21} propose an online multi-scale convolutional sparse coding model constructed for encoding dynamic rain/snow and background motions with temporal variations.

\revised{Zhang et al.~\cite{zhang2022} propose an end-to-end video deraining framework, called ESTINet, which takes advantages of deep residual networks and convolutional long short-term memory. It can capture the spatial features and temporal correlations among successive frames at low computational costs.}
\revised{Yang et al.~\cite{yang2022} proposed an augmented Self-Learned Deraining Network called SLDNet+ to remove both rain streaks and rain accumulation by utilizing temporal correlation, consistency, and rain-related priors.}
\revised{Yan et al.~\cite{yan2022} propose a two-stage (i.e., the single image module and the multiple frame module) video-based raindrop removal network, which is based on temporal correlation.}
%\needconfirm{Recently, more excellent video deraining/restoring methods~\cite{yan2022,yang2022} have been proposed.}

Video rain removal methods usually assume that the locations of rain steaks across neighboring frames are uncorrelated, which are not like the motion of background layers. Although rain streak removal from videos has been carefully studied, these existing methods still fail to completely remove rain and the performance evaluated on each single frame is worse than those of image-based rain streak removal methods.

\subsection{Rain Streak Removal From LFIs}
\label{subsec:rain_removal_LFI}
Rain removal methods for LFIs are still in their infancy. The earlier method~\cite{Tan17} proposes to first align all sub-views with the central sub-view. Robust principal component analysis is then applied to decompose each image of a set of deformed sub-views into low-rank data and sparse data. Finally, a dark view image is introduced to estimate non-rainfall disparity edges from sparse data whose remaining part is considered as rain, and the non-rainfall disparity edges are restored back to a low-rank image to generate the rain-free LFI. Ding et al.~\cite{ding21} propose a GAN-based architecture for removing rain streaks from 3D EPI of a rainy LFI. It first estimates depth maps, and then simultaneously detects rain streaks and restores rain-free sub-views by exploiting the correlation between rain streaks and the clean background layer. However, it can only exploit the sub-views in the same row/column of a rainy LFI for rain streak removal at each time.

In conclusion, since these methods have difficulties in accurately predicting various rain streaks from challenging real-world LFIs, they may not be able to recover high-quality rain-free LFIs from various rainy LFIs. To address this limitation, we propose a novel and more effective deep-learning-based method to eliminate rain streaks from rainy LFIs.

\section{Rain Imaging Model and Rainy LFI Dataset}
\label{sec:rain_model_data_generation}
The widely used rain imaging model~\cite{Zhang18}~\cite{Qian18} regards a rainy image as a linear combination of the rain-free background and the rain streak/drop layer. Yang et al.~\cite{Yang17} note that in heavy rain, rain streaks of various shapes and directions would overlap with each other. They thus propose a rain imaging model consisting of multiple layers of rain streaks and a global atmospheric light caused by atmospheric veiling effect as well as blur. Later, Hu et al.~\cite{Hu19} note that the visual effects of rain depend on scene depth and proposed another rainy image generation model, i.e., a rainy image is a composition of a rain-free image, a rain streak layer and a fog layer, in which both the rain streak and fog transmissions depend on the depth.

In complex real-world scenes, high-speed falling rain always produces motion blur in the captured images. This phenomenon can seriously hinder the rain streak removal task. Wang et al.~\cite{wang2021} propose a rain streak observation model based on the motion blur of rain streaks, and propose a network to learn the angle and the length of the motion blur kernel from the detail layer of a rainy image. \revised{The motion blur kernel} is then stretched into a degradation map for rain streak \revised{prediction}. %, their rain streak observation model did not consider the fog and simply adds motion blur to the matrix to generate rain streaks map.
However, none of the existing rainy image generation models or datasets have taken motion blur of rain streaks into account.

\begin{figure}[h]
	\centering
	\includegraphics[width=0.46\textwidth]{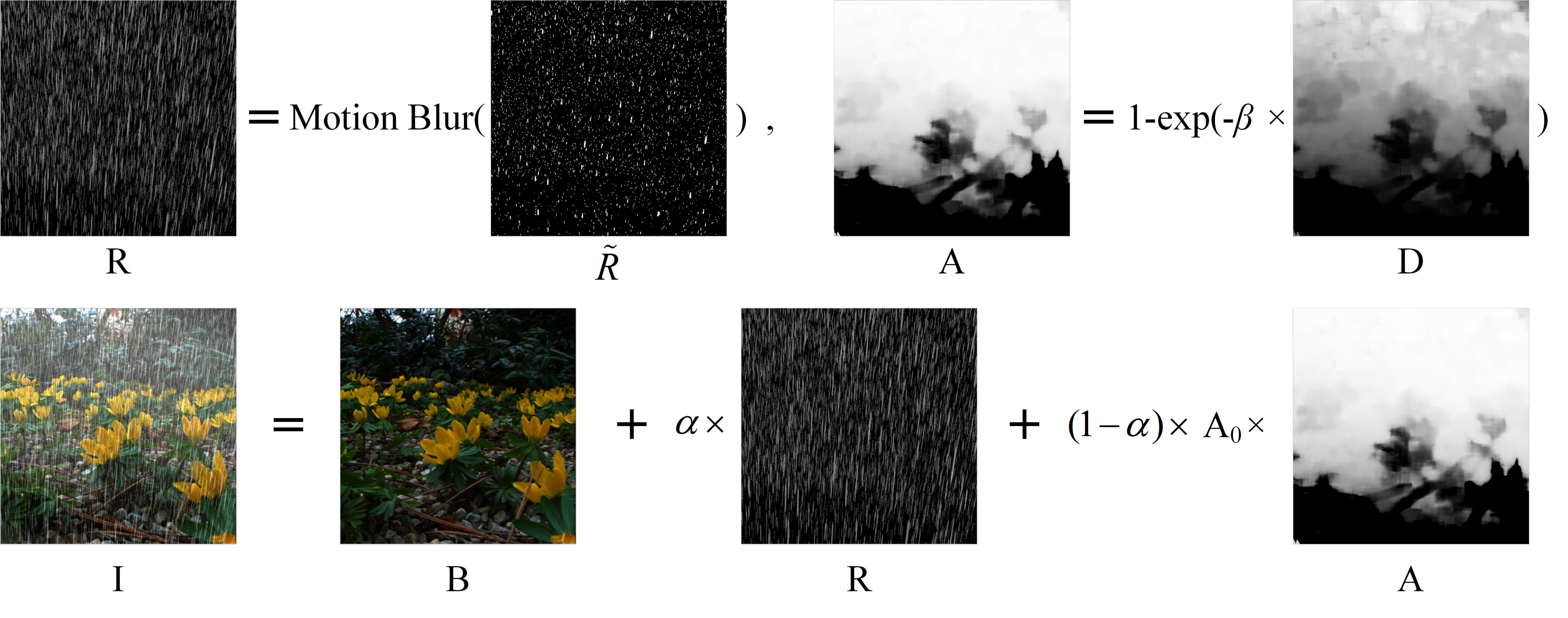}
    \vspace{-4mm}
	\caption{Our rainy LFI generation process. $\tilde{R}$ represents the static rain streaks without the motion blur effect. $R$ denotes the rain streak layer after adding the motion blur effect. $I$ is the rainy image, and $B$ is the rain-free layer. $D$ is the depth map of $B$, and $A$ is the fog/mist layer. }
	\label{fig:rainimage}
	\vspace{-0.016\textwidth}
\end{figure}

We carefully take into account the motion blur of rain streaks and propose a novel rainy image generation model, which is much closer to real situation, as shown in Fig.~\ref{fig:rainimage}. \revised{Comparing with~\cite{wang2021},} our rainy image generation model is \revised{an improved additive degradation model with consideration of the fog/mist effect}, and is defined as:
\renewcommand{\arraystretch}{1.5}
\begin{equation}
I = B + \alpha \mathrm{M_b}(R) + (1-\alpha)A_0A,
\label{eq:rainy_image}
\end{equation}
where $I$ denotes the rainy image. $B$ is the rain-free image coming from the real-world LFI dataset~\cite{LILFIs} captured by a Lytro Illum camera. $R$ is the rain streak layer. $\mathrm{M_b}(\cdot)$ represents a motion blur operation. $A_0$ represents the global atmospheric light, whose value is assumed to be a constant~\cite{Hu19}. $A$ denotes the fog/mist layer. $\alpha$ is a constant parameter used to adjust the relative intensity between the rain-streak layer and fog layer, \revised{empirically} set as $0.6$ \revised{in our experiment}.

The particle system in Blender~\cite{Blender} is adopted to simulate 3D rain streaks for our rainy LFI generation. A large number of rain streaks with various shapes, directions and densities are generated. At the same time, the simulated rain streaks are added with a motion blur effect~\cite{potmesil1983}, which is supplied by the Motion Blur function of the Render module in Blender.

The visual intensity of fog increases exponentially with the scene depth~\cite{Hu19}, i.e., degrades linearly with the transmission map $T$~\cite{he2010}. Specifically, the fog layer can be modeled as:
\begin{equation}
A = 1-T = 1-e^{-\beta D},
\label{ep:d2f}
\end{equation}
where $D$ refers to the depth map. $\beta$ is \revised{empirically} set as $1.8$ to control the thickness of the fog. Larger $\beta$ means thicker fog.

Since the LFI dataset~\cite{LILFIs} does not provide ground-truth disparity maps \revised{for} each LFI, the depth estimation method~\cite{jeon15} is adopted to estimate acceptable depth maps for our rainy \revised{LFI} generation. Finally, the rain-free LFIs, rain-streak layers and fog layers are combined as defined in Eq.~\ref{eq:rainy_image} to generate the vivid rainy LFIs. A total number of $400$ synthetic rainy LFIs and their ground truth rain-free LFIs are generated. In addition, we have captured $200$ real-world LFIs with a Lytro ILLUM camera to construct a real-world rainy LFI subset for learning the proposed \revised{LFI} rain removal network.

\section{LFI Rain Detection and Removal Network}
\label{sec:method_description}

%In our rainy LFI generation model, a rainy sub-view consists a rain-free background layer, a rain streak layer and a fog layer.
We propose our method, 4D-MGP-SRRNet, to detect rain streaks and recover the rain-free LFI from a rainy LFI, as shown in Fig.~\ref{fig:network_architecture}. The network consists of three parts: the rain streak detection network MGPDNet, the depth estimation sub-network DERNet, and the recurrent rain streak removal sub-network RNNAT. To make full use of all sub-views of a rainy LFI, 4D convolutions~\cite{zhang20204d} implemented using 3D convolutions are chosen to construct our rain removal network 4D-MGP-SRRNet. Specifically, 4D convolutional layers are used to exploit all sub-views of an LFI in a more meaningful 5D feature space by learning complicated interactions of spatial and angular representations \revised{in our network}.

\begin{figure}[tb]
	\centering
	\includegraphics[width=0.49\textwidth]{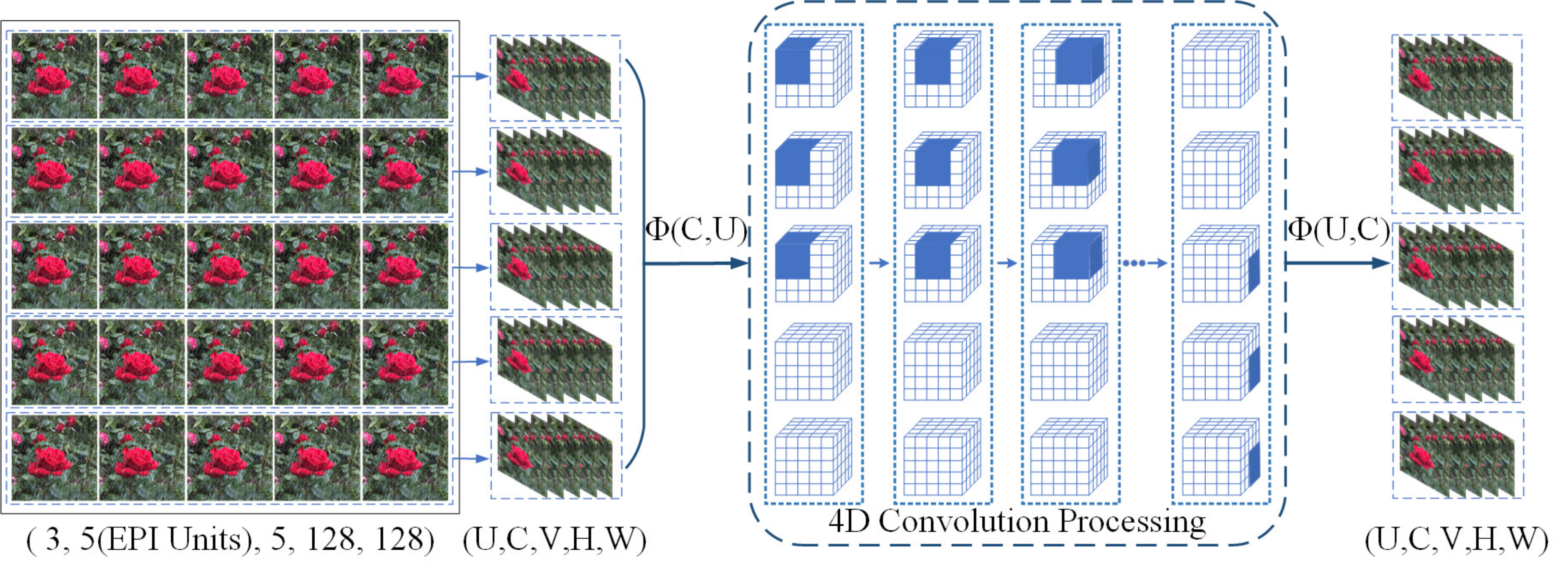}
	\vspace{-4mm}
	\caption{4D Convolution operates on an LFI. \revised{$\Phi()$ means permuting two specified dimensions of the input data.}}
	\label{fig:4DConv}
	\vspace{-0.016\textwidth}
\end{figure}

\revised{4D convolution can surpass 3D convolution and 2D convolution in many video processing~\cite{zhang20204d} and LFI processing tasks. 4D convolution~\cite{zhang20204d} is originally proposed to model both short-term and long-term spatio-temporal representations simultaneously, and overcome short-term spatio-temporal representations modeled by 3D convolution from videos of RGB frames. We find that such 4D convolution can be used to properly and effectively model the interaction of sub-views intra and inter \textit{3D-EPIs} of an LFI, \revised{as shown in Fig.~\ref{fig:4DConv}}, by just considering each \textit{3D-EPI} as an \textit{action unit} of video-level representation for action recognition~\cite{zhang20204d}. For LFI processing, 3D convolution can only model and exploit the interaction of sub-views within a \textit{3D-EPI}, and 2D convolution cannot properly model the structural relationship between all sub-views of an LFI for exploiting as much meaningful information as possible. Therefore, by utilizing 4D convolutional layers to build our network, our network can effectively explore and exploit all sub-views of an LFI for rain streak detection and removal. ReLU activation~\cite{nair2010} is used after each 4D convolutional layer in our network.}

In our network, for an input rainy LFI with a spatial resolution of $[512, 512]$ and angular resolution of $[5, 5]$, we crop each sub-view to small patches of resolution $[128, 128]$. \revised{The corresponding patches cropped from the same position of sub-views in the same row are then arranged} as a \textit{3D-EPI unit}~\cite{zhang20204d}. In this way, a total of $5$ \textit{3D-EPI units} is obtained to form 5D data with the size of $\revised{5(\textit{3D-EPI\ units})\times 3}\times 5\times128\times128$, where $3$ refers to the channels of sub-views. \revised{As shown in Fig.~\ref{fig:4DConv}, the} channel dimension permutes with the \textit{3D-EPI unit} dimension before the 5D data is fed into a 4D convolutional layer, and then the output is permuted reversely to 3D form for subsequent 3D convolutional layers~\cite{zhang20204d}. \revised{At the beginning of MGPDNet shown in Fig.~\ref{fig:network_architecture},} a non-local block~\cite{bai21} \revised{followed the 4D convolutional layer} is used to exploit self-attention for \revised{rough intra- and inter-sub-views rain streak features}.

\subsection{Rain Streak Detection}
\label{subsec:rain_streak_detection}

Leveraging the densely-connected block~\cite{huang17,Yu16}, we construct \revised{rain streak detection sub-network} MGPDNet, as shown in Fig.~\ref{fig:network_architecture}, to detect rain streaks over different scales of features via three different branches (from bottom to top) that consist of several 4D dense blocks with convolution kernels of $3\times3\times3\times3$, $5\times5\times5\times5$, and $7\times7\times7\times7$. MGPDNet uses novel progressive fusion instead of direct fusion to integrate the extracted features. The rain streak detection process can be expressed as:
\renewcommand{\arraystretch}{1.5}
\begin{equation}
\begin{array}{l}
f^1=\mathrm{Dense_1}(f^0),\\
f^2=\mathrm{Dense_2}([f^0,f^1]),\\
f^3=\mathrm{Dense_3}([f^0,f^1,f^2]),\\
R'=\mathrm{\mathcal{F}_{conv}}([f^1,f^2,f^3]),
\end{array}
\end{equation}
where $f^0$ denotes the rain streak features extracted by a 4D convolution block from the input LFI $I$ and \revised{then processed} by a non-local block~\cite{bai21} to enhance the features of rain streaks. Specifically, the non-local block is used to make self-attention for rain \revised{streak features} by exploiting their correlation \revised{intra- and inter-sub-views}. \revised{$\mathrm{Dense}(\cdot)$} represents the convolution operation of dense blocks. $f^1$, $f^2$ and $f^3$ are extracted features from different branches. \revised{$\mathrm{\mathcal{F}_{conv}}(\cdot)$} is introduced to produce the final rain streaks by merging the rain \revised{streak features} detected at multi-scales, and $R'$ refers to the predicted rain streaks.

In our MGPDNet, we use GP models to jointly model the distribution of synthetic data features and real-world data features on multi-scale dense block branches for supervising unlabeled real-world rain streak detection and improving the generalization of our method.

%\revised{\subsection{Multi-scale Gaussian Processes Model}}
%\label{subsec:Gaussian processes}

\textbf{Gaussian Process (GP):}
It is a collection of random variables, any finite number of which have consistent joint Gaussian distributions~\cite{Rasmussen2003gaussian,williams2006gaussian,murray2001gaussian}. A GP \revised{$\mathrm{g}(\cdot)$} is fully specified by its mean function $\revised{\mathrm{m}}(\cdot)$ and covariance function $\revised{\mathrm{K}}(\cdot)$ as:
\begin{equation}
  \revised{\mathrm{g}}(x)\sim \mathcal{GP}(\revised{\mathrm{m}}(x), \revised{\mathrm{K}}(x,x')),
\end{equation}
where $x$ and $x'$ are the possible inputs that index the GP, and
\begin{eqnarray}
% \nonumber % Remove numbering (before each equation)
  \revised{\mathrm{m(x)}} &=& \mathbb{E}[\revised{\mathrm{g}}(x)],  \\
  \revised{\mathrm{K}}(x,x') &=& \mathbb{E}[(\revised{\mathrm{g}}(x)-\revised{\mathrm{m}}(x))(\revised{\mathrm{g}}(x')-\revised{\mathrm{m}}(x'))].
\end{eqnarray}

Assume that the training sample $\revised{\mathrm{g}}(x^i)$ and the unseen test sample $\revised{\mathrm{g}}(x_*^j)$, where $i=\{1,...,n\}$ and $j=\{1,...,r\}$, conform to the Gaussian distribution.  Therefore, the formula for conditioning a joint Gaussian distribution is defined as:
\begin{eqnarray}\label{eq:joint_distribution}
\left[ \begin{array}{c}
\revised{\mathrm{g}}(X)\\
\revised{\mathrm{g}}({{X}_{*}})\\
\end{array}
\right ]
\sim \mathbb{N} \left (
\left[ \begin{array}{c}
\mu \\
\mu_{*}
\end{array}
\right ],
\left[ \begin{array}{cc}
\Sigma & \Sigma_* \\
\Sigma^T_* & \Sigma_{**}\\
\end{array}
\right ]
\right ),
\end{eqnarray}
where $\mu=\revised{\mathrm{m}}(X)$. $\mu_{*}=\revised{\mathrm{m}}(X_*)$. $\Sigma=\revised{\mathrm{K}}(X,X)$, $\Sigma_*=\revised{\mathrm{K}}(X,X_*)$ and $\Sigma_{**}=\revised{\mathrm{K}}(X_*,X_*)$, which denotes the covariances evaluated at all pairs of training and/or testing points.
%Then, the conditional distribution of $g_*\revised{=g(X_*)}$ given $g\revised{=g(X)}$ can be expressed as
%\begin{equation}\label{eq:posterior_distribution}
%g_*|g \sim \mathcal{N} (\mu_*+\Sigma^{T}_{*}\Sigma^{-1}(g-\mu), \Sigma_{**}-\Sigma^{T}_{*}\Sigma^{-1}\Sigma_{*}),
%\end{equation}

When modeling real situations (e.g., real-world rainy LFIs), we always access noisy samples, i.e., $\revised{\mathrm{y}}(x^i)=\revised{\mathrm{g}}(x^i)+\epsilon^i$ and $\revised{\mathrm{y}}(x_*^{j})=\revised{\mathrm{g}}(x_{*}^{j})+\epsilon^j$, where $\epsilon^i$ and $\epsilon^j$ are independent Gaussian noise $\mathcal{N}(0, \sigma_{\epsilon}^2)$~\cite{murray2001gaussian}. %Thus, the Eq.~\ref{eq:joint_distribution} can be rewritten as
The conditional distribution of $g_*=\revised{\mathrm{g}}(X_*)$ given $g=\revised{\mathrm{g}}(X)$ can then be expressed as:
\begin{comment}
\revised{Thus, the formula for conditioning a joint gaussian distribution (Eq.~\ref{eq:joint_distribution}) can be rewritten as }
\begin{eqnarray} \label{eq:joint_distribution_noise}
\left[ \begin{array}{c}
y(X)\\
y({{X}_{*}})\\
\end{array}
\right ]
\!\sim\! \mathbb{N} \left (
\left[ \begin{array}{c}
\mu  \\
\mu_{*}
\end{array}
\!\right ]\!\!,\!
\left[ \begin{array}{cc}
\Sigma \!+\! \sigma_{\epsilon}^2I_n \!&\! \Sigma_* \\
\Sigma^T_* \!&\! \Sigma_{**} \!+\! \sigma_{\epsilon}^2I_r\\
\end{array}
\!\right ]\!
\right ),
\end{eqnarray}
where $I_n$ and $I_r$ are the identity matrixes with size $n$ and $r$, respectively. %The Eq.~\ref{eq:posterior_distribution} can also be rewritten as
\revised{Then, the conditional distribution of $y_*\revised{=y(X_*)}$ given $y\revised{=y(X)}$ can be expressed as}
\end{comment}
\begin{equation}
\begin{aligned} \label{eq:posterior_distribution_noise}
y_*|y \sim \mathcal{N} (& \mu_*+\Sigma^{T}_{*}(\Sigma+\sigma_{\epsilon}^2I_n)^{-1}(y-\mu), (\Sigma_{**}+\sigma_{\epsilon}^2I_r) \\
  &-\Sigma^{T}_{*}(\Sigma+\sigma_{\epsilon}^2I_n)^{-1}\Sigma_{*}),
\end{aligned}
\end{equation}
%where $y=y(X)$.
where $I_n$ and $I_r$ are the identity matrices with sizes $n$ and $r$, respectively.

For the rest of this section, we simplify the notation $X_*$ to ${x_*'}$ to indicate that we try to predict the resulting features $\revised{\mathrm{y}}(x_{*}')$ for feature vector $x_*'$, given the training data set including inputs $X$ and corresponding ground-truth/output $\revised{\mathrm{g}}(X)$. In data preprocessing of GP, the mean value is always subtracted from the training/testing samples, which means $\mu=0$ and $\mu_*=0$.
%in Eq.~\ref{eq:posterior_distribution_noise}.

%\limingyue{
%\begin{equation}
%\begin{array}{l}
%\bar{f}_*\triangleq E[f_*|X,y,X_*] = K(X_*,X)[K(X,X)+\sigma_n^2]^{-1}y ,\\
%cov(f_*)=K(X_*,X_*)-K(X_*,X)[K(X,X)+\sigma_n^2]^{-1}K(X,X_*) ,
%\label{eq:mean}
%\end{array}
%\end{equation}}

\begin{figure}[tb]
	\centering
	\includegraphics[width=0.46\textwidth]{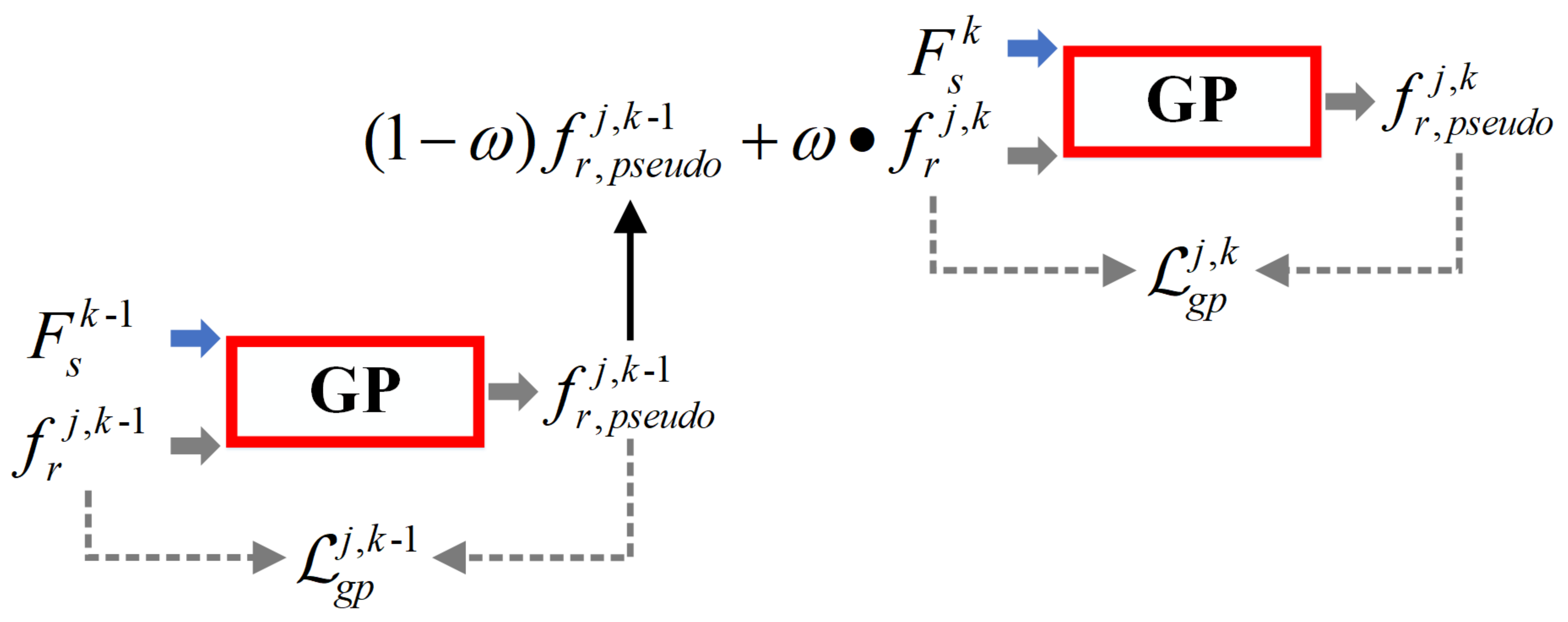}
	\vspace{-0.01\textwidth}
	\caption{The working process of the $k^{th}$ GP of the Multi-scale Self-guided GP (MSGP).}
	\label{fig:MSGP}
	\vspace{-0.01\textwidth}
\end{figure}

\textbf{Multi-scale Self-guided Gaussian Process for Rain Streak Detection:}
Leveraging the GP, we design the semi-supervised learning process for our MGPDNet. The seminal work~\cite{yasarla20} introduced GP to model the latent features extracted by the encoder. Our semi-supervised process models synthetic LFI features and real-world LFI features extracted by 4D dense blocks \revised{of} multi-scale branches with different kernel sizes, as shown in Fig.~\ref{fig:network_architecture} and~\ref{fig:MSGP}. The calculated mean of the conditional distribution is treated as the pseudo ground truth \revised{of the} predicted latent feature vector of a real-world LFI \revised{for supervising} the feature extraction of \revised{the} dense blocks. In our multi-scale GP, the extracted latent feature vectors of the synthetic LFI form matrix $F_{s}^k=\{f_s^{i,k}\}_{i=1}^{N_l}$ for the $k^{th}$ scale/branch, where $N_l$ refers to the total number of training patches cropped from our synthetic rainy LFIs.

\begin{comment}
\revised{The feature vector $f_r^i$ of a real-world LFI obtained by the multi-scale dense block can be expressed as
\begin{equation}\label{eq:v_er}
  f_{r}^{k}=\sum_{i=1}^{N_l}\alpha_{i}f_{s}^{i}+\epsilon.
\end{equation}}
\end{comment}

Since the inversion of a $N\times N$ matrix is computationally nontrivial while $N >1000$, the GP approach is suitable for small and medium-size data sets~\cite{murray2001gaussian}. Therefore, for efficient calculation, we adopt the cosine similarity measure to select $N_n$ nearest labeled latent vectors for one unlabeled vector of the corresponding real-world LFI, instead of utilizing all vectors in $F_s^{k}$. $N_n$ is set as $16$, which is the same as the channel size of the input features extracted for every pixel of the rainy LFI sub-views. The input feature size for the GP model in each branch is $1\times 49152$ ($3\times 128\times 128$) according to the spatial size $[128, 128]$ for every sub-view.

It is worth emphasizing that 4D convolutions exploiting features and correlations across all sub-views of \revised{the input 5D LFI patch} can obtain the predicted features with abundant information for rain streak detection. The \revised{4D convolutions of MGPDNet} not only retain the spatial sizes of the predicted features for \revised{all} sub-views \revised{of} the input 5D data, but also maintain the number of channels. In this way, large predicted features for each sub-view of the input patch are fed into the following GP in each scale/branch. It should be noted that the distribution of rain streak features in different sub-views of the LFI is similar. Therefore, for computational efficiency, GPs in our MSGP only \revised{process} the rain streak features of the central sub-view of an input patch.

In our MSGP, to model features of rain streaks, ground truth for the features $f_s^{i,k}$ extracted from the $i^{th}$ synthetic LFI (patch) at scale $k$ is itself, $f_s^{i,k}$. The pseudo ground truth for the $j^{th}$ testing features $f_r^{j,k}$ in the $k^{th}$ dense block branch \revised{is} defined as $f_{r,pseudo}^{j,k}=\Sigma^{T}_{*}(\Sigma+\sigma_{\epsilon}^2I_n)^{-1}F_{s}^k$ \revised{(according to Eq.~\ref{eq:posterior_distribution_noise})}, which also has a size $1\times 49152$.

\begin{figure}[b]
	\centering
	\includegraphics[width=0.42\textwidth]{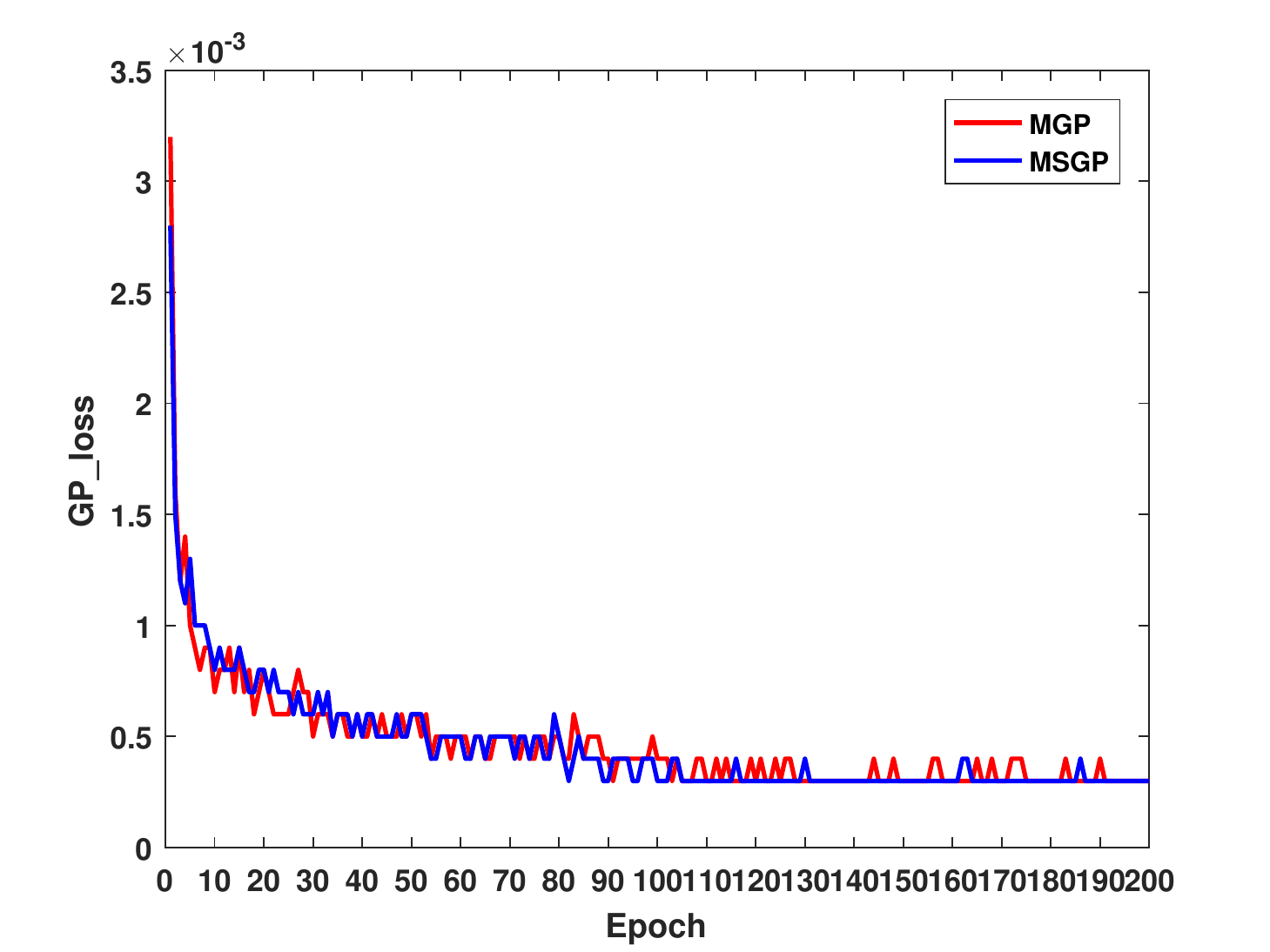}
	\vspace{-3mm}
	\caption{Convergence process comparison for the MSGP with and without the self-guided scheme. Loss refers to $\mathcal{L}_{gp}$.}
	\label{fig:MSGP_converge}
    \vspace{-0.016\textwidth}
\end{figure}

In order to fuse the predicted features across different scales and speed up the convergence of multi-scale GP models, we propose a self-guided scheme, as shown in Fig.~\ref{fig:MSGP}. During training, the weighted average of the pseudo ground truth features obtained by the GP in the previous ($(k-1)^{th}$) branch and the features produced by the dense blocks in the current ($k^{th}$) branch are taken as the input of the current GP.
This means that the pseudo ground truth obtained by the GP in the previous scale (or called branch) is used to guide the modeling of the GP in the subsequent scale (branch). The weighted average of these features can fuse the features across different scales for more accurate detection of rain streaks for the patches cropped from a real-world rainy LFI.
Obviously, this multi-scale GP structure is conducive to the rapid convergence of the network. The MSGP module of the \revised{rain streak detection sub-network} MGPDNet models more regular rain streaks with fewer parameters. Therefore, it is feasible to detect rain streaks more effectively than other state-of-the-art methods. Fig.~\ref{fig:MSGP_converge} demonstrates that our MSGP with a self-guided scheme converges more quickly and stably than one without the self-guided scheme.

Specifically, the training process of MGPDNet \revised{that leverages} a multi-scale GP module contains two stages. The first stage is a fully-supervised learning process, in which the ground-truth rain streaks of synthetic rainy LFIs are used to supervise MGPDNet to extract rain streak features. The second stage is an unsupervised process, in which pseudo ground-truth rain streaks for each real-world rainy LFI are produced by the pre-trained MGPDNet via the multi-scale GP module. Real-world rainy LFIs taken for training can update the parameters of MGPDNet and improve its generalization \revised{to} real-world rainy \revised{LFIs}.

%\limingyue{In order to obtain the posterior distribution on the function, the joint prior distribution is limited to include only those functions that are consistent with the observed data points. In fact, Eq.~\ref{eq:mean} is also the average function and covariance function of the posterior process.}

\textbf{Synthetic rainy LFIs training phase:} In this phase, MGPDNet learns network parameters by utilizing synthetic LFIs to extract the rain streaks. $L_{1}$ (i.e., MAE) loss and $L_2$ (i.e., MSE) loss are common losses for training networks. However, the MSE loss is sensitive to abnormal points, which are not \revised{conducive} to network training. MAE loss is less sensitive to abnormal points, but it is not differentiable at the target point. \revised{Smooth $L_{1}$ loss}~\cite{girshick15} perfectly avoids the defects of MSE and MAE, since its gradient value remains to be $1$ even though the absolute error is larger than $1$, which is defined as:
\begin{equation}
\revised{\mathrm{Smooth_{L_1}}}(x) =
\left\{
\begin{array}{lcr}
\begin{aligned}
0.5x^2  && \revised{\mathrm{if}} \ |x|<1 \\
|x|-0.5 && \revised{\mathrm{otherwise}},
\end{aligned}
\end{array}
\right. .
\end{equation}

In the early stage of training, \revised{smooth $L_1$ loss} uses $L_1$ loss to stabilize the gradient and converge quickly. In the later stage of training, \revised{smooth $L_1$ loss adopts $L_{2}$ loss} to make the network gradually converge to the optimal solution.

The \textit{smooth $L_{1}$ loss} and perceptual loss are both adopted in our supervised loss function.
\begin{equation}
\mathcal{L}_{s}= \mathrm{Smooth_{L_1}}(R'_{s}-R_{s}^{gt}) + \lambda_{p}\|\revised{\mathrm{VGG}}(R'_{s}) - \revised{\mathrm{VGG}}(R_{s}^{gt})\|_{2}^{2},
\label{eq:Ls}
\end{equation}
where $\lambda_{p}$ is a constant weighting parameter. $R'_{s}$ is the predicted rain streaks for a synthetic rainy LFI, and $R_{s}^{gt}$ is the ground truth. $\revised{\mathrm{VGG}}(\cdot)$ represents the pre-trained VGG-16~\cite{Simon15}.

Besides minimizing the above loss function, all the intermediate feature vectors $f_s^{i,k}$ for \revised{the central} sub-view of the $i^{th}$ 5D LFI patch cropped from the synthetic rainy LFI in the $k^{th}$ dense block branch are stored in the matrix $F_{s}^k={\{f_{s}^{i,k}\}}_{i=1}^{N_{l}}$ during the training process.%, where $N_{l}$ is the total number of synthetic rainy LFIs.

\textbf{Real-world rainy LFIs training phase:} After the synthetic rainy LFI training phase, our collected real-world rainy LFIs are used to update the network weights of MGPDNet, which can improve the generalization ability of MGPDNet for rain streak detection on real-world LFIs. %Specifically, we use GP to jointly model the distribution of synthetic data features and real-world data features obtained \revised{in multi-scale dense block branches.}

\begin{comment}
\limingyue{The conditional joint distribution is defined as follow:
\begin{equation}
P(f_{r}^{i'}|f_{r}^{i},F_{fs}[i])=\mathcal{N}(\mu_{r}^{i},\Sigma_{r}^{i}),
\end{equation}
where $f_{r}^{i}$ is the features of real-world rain streaks extracted by the network. $f_{r}^{i'}$ represents the unknown features between real data and synthetic data, which is defined as:
\begin{equation}
f_{r}^{i'}=\sum\limits_{n = 1}^{N_{s}}\alpha_{n}f_{s}^{n} + \epsilon,
\end{equation}
where $\alpha_{n}$ are the coefficients, and $\epsilon$ is noise. $f_{r,pseudo}^{i}=\mu_{r}^{i}$ is defined as the pseudo ground truth of real-word data $f_{r}^{i}$. Then, in the next iteration, the input f of GP will obtain a pseudo ground truth $f_{r,pseudo}^{i}$.}
\end{comment}

\begin{comment}
\limingyue{In our multi-scale GP model, former GP utilizing the features extracted from fine-scale \textit{4D dense blocks} models the pseudo ground truth of real data features.}
\end{comment}

Since it is hard to utilize all synthetic data features to model the pseudo ground truth for a real sample~\cite{murray2001gaussian}, only $N_n$ nearest features from the synthetic data are chosen to form $F_{s,n}^k$. However, it is still non-trivial to accurately detect rain streaks, and utilizing synthetic data to approximate real data may predict false pseudo ground truth. Hence, we minimize the variance $\Sigma_{r,n}^{j,k}$ calculated by $f_r^{j,k}$ and $F_{s,n}^k$, and maximize the variance $\Sigma_{r,f}^{j,k}$ calculated by $f_r^{j,k}$ and $N_f$ farthest synthetic features $F_{s,f}^k$, in order to ensure that $F_{s,f}^k$ are dissimilar to the unlabeled $f_r^{j,k}$ and it does not affect the prediction of GP~\cite{yasarla20}. \revised{Specifically,} according to Eq.~\ref{eq:posterior_distribution_noise}, $\Sigma_{r,n}^{j,k}$ can be written as:
\begin{equation}
\begin{aligned}
\Sigma_{r,n}^{j,k}=&(\revised{\mathrm{K}}(f_r^{j,k},f_r^{j,k})+\sigma _\epsilon^2)-\revised{\mathrm{K}}(f_r^{j,k},F_{s,n}^k)[\revised{\mathrm{K}}(F_{s,n}^k,F_{s,n}^k) \\
&+\sigma_\epsilon^2I_n]^{-1} \revised{\mathrm{K}}(F_{s,n}^k, f_r^{j,k}).
\end{aligned}
\end{equation}
The definition of $\Sigma_{r,f}^{i,k}$ is analogous to $\Sigma_{r,n}^{i,k}$. Similarly, the pseudo ground truth for the real data features $f_{r}^{j,k}$ is:
\begin{equation}\label{eq:final_mean}
f_{r,pseudo}^{j,k}=\mu_{r}^{j,k}=\revised{\mathrm{K}}(F_{s,n}^k,f_r^{j,k})^T(\revised{\mathrm{K}}(F_{s,n}^{k},F_{s,n}^{k})+\sigma_{\epsilon}^2)^{-1}F_{s,n}^{k},
\end{equation}
which is used to guide the feature extraction in each 4D dense block branch.

%The loss introduced by the multi-scale GP for training our MGPDNet is:
%\begin{equation}
%\mathcal{L}_{GP}=\frac{1}{N_uN_b}\sum_{j=1}^{N_u}\sum_{k=1}^{N_b}\mathcal{L}_{gp}^{j,k},
%\end{equation}
%where $N_b$ is the number of 4D dense block branches, and $N_u$ refers to the total number of patches cropped from the real-world rainy LFIs of our RLMB dataset. $\mathcal{L}_{gp}^{j,k}$ is defined as:
%\begin{equation}
%\mathcal{L}_{gp}^{j,k}=\|f_{r}^{j,k}-f_{r,pseudo}^{j,k}\|_2 + \log\Sigma_{r,n}^{j,k}+\log(1-\Sigma _{r,f}^{j,k}),
%\end{equation}
%where $j$ and $k$ are used to indicate the $j^{th}$ real-world LFI patch, and its features $f_{r,pred}^{j,k}$ extracted from the $k^{th}$ dense block branch.
%
%
%The GP module converts features into vectors for computation convenience, which may result in a lack of global supervision. To ensure the matching of global features, we also adopt perceptual loss as:
%\begin{equation}
%\mathcal{L}_{p,real}=\frac{1}{N_uN_b}\sum_{j=1}^{N_u}\sum_{k=1}^{N_b}\|VGG(f_{r,pred}^{j,k})-VGG(f_{r,pseudo}^{j,k})\|_2^2.
%\end{equation}
%
%Ultimately, the loss function during the unsupervised training of our MGPDNet is defined as:
%\begin{equation}
%\mathcal{L}_{r}=\lambda_{GP}\mathcal{L}_{GP}+\lambda_{p,real}\mathcal{L}_{p,real},
%\label{eq:Lr}
%\end{equation}
%where $\lambda_{GP}$ and $\lambda_{p,real}$ are constant weighting parameters.

The loss function during the unsupervised training of our MGPDNet is defined as:
\begin{equation}
\mathcal{L}_{r}=\frac{1}{N_uN_b}\sum_{j=1}^{N_u}\sum_{k=1}^{N_b}\mathcal{L}_{gp}^{j,k},
\end{equation}
where $N_b$ is the number of 4D dense block branches, and $N_u$ refers to the total number of patches cropped from the real-world rainy LFIs of our RLMB dataset. GPs in our MSGP convert features into vectors for computation convenience, which may result in a lack of global supervision. To ensure the matching of global features, we also introduce perceptual loss to the original GP loss. $\mathcal{L}_{gp}^{j,k}$ is defined as:
\begin{equation}
\begin{aligned}
\mathcal{L}_{gp}^{j,k}=&\lambda_{GP}(\|f_{r}^{j,k}-f_{r,pseudo}^{j,k}\|_2+\log\Sigma_{r,n}^{j,k}+\log(1-\Sigma _{r,f}^{j,k}))\\
&+\lambda_{p,real}(\|\mathrm{VGG}(f_{r}^{j,k})-\mathrm{VGG}(f_{r,pseudo}^{j,k})\|_2^2),
\end{aligned}
\label{eq:Lr}
\end{equation}
where $j$ and $k$ are used to indicate the $j^{th}$ real-world LFI patch and its features $f_{r}^{j,k}$ extracted from the $k^{th}$ dense block branch, \revised{respectively}.

Thus, the overall loss function used to train MGPDNet is:
\begin{equation}
\mathcal{L}_{rain}= \mathcal{L}_{s}+ \mathcal{L}_{r}.
\end{equation}

\subsection{Depth Estimation}
\label{subsec:depth_estimation}

Due to the occlusion of rain streaks with various shapes and densities, it is difficult to obtain depth maps for the input rainy LFI. In order to alleviate this problem, we subtract the predicted rain streaks from the rainy LFI, and then feed the resulting sub-views to DERNet to estimate accurate depth maps. The depth estimation function for synthetic/real-world rainy LFIs is defined as:
\renewcommand{\arraystretch}{1.5}
\begin{equation}
D' = \revised{\mathrm{\mathcal{F}_{DERNet}}}(I - R'),
\end{equation}
where $D'$ denotes the estimated depth maps, and $R'$ is the predicted rain streaks. $\mathrm{\mathcal{F}_{DERNet}}(\cdot)$ represents the convolution operation of DERNet.

DERNet is trained on the synthetic LFI with ground truth depth maps coming from the synthetic LFI dataset of~\cite{ding21} and a part of the real-world LFI dataset~\cite{LILFIs} with estimated depth maps produced by~\cite{jeon15}. %Concretely, the DERNet is first pre-trained solely on the above dataset, and then trained together with the MGPDNet and RNNAT.}
% \limingyue{DERNet directly loads the pre-training parameters during the training process to estimate the depth map.***???***}
After obtaining the depth maps, we apply Eq.~\ref{ep:d2f} to convert the estimated depth maps to fog maps, $A'$, in order to assist the subsequent RNNAT to recover rain-free LFIs with consideration of the fog/mist effect.

\subsection{Rain-free Image Recovery}
\label{subsec:image_recovery}

%\limingyue{In this subsection,  rainy sub-views concatenated with the estimated rain streaks and fog map are fed into the rain removal network RNNAT to progressively restore the rain-free sub-views. }
	
\begin{figure}[tb]
	\centering
	\includegraphics[width=0.46\textwidth]{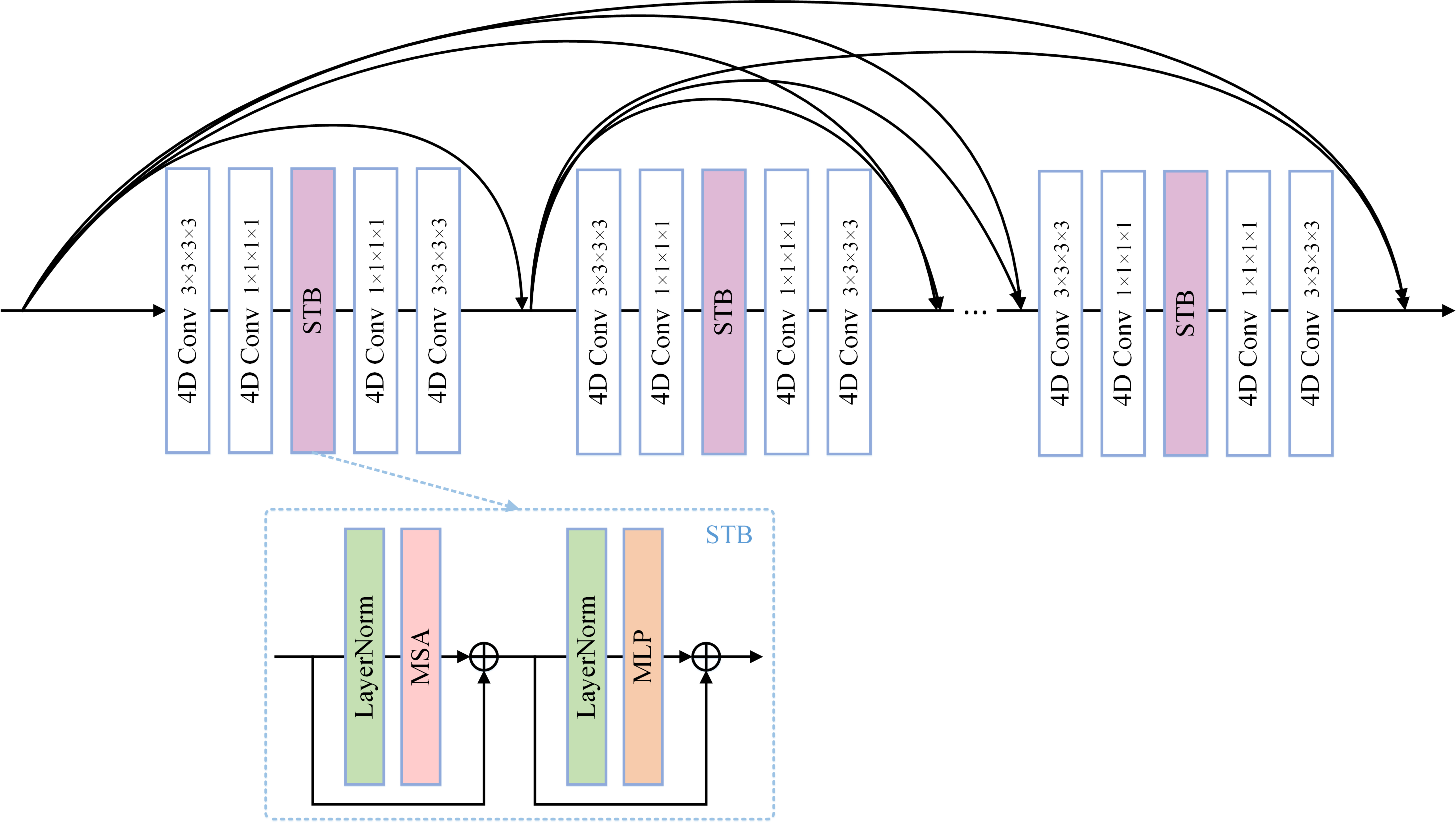}
	\vspace{-3mm}
	\caption{Architecture of the Dense Swin Transformer Block (DSTB).}
	\label{fig:DSTB}
\end{figure}

The generator of our RNNAT adopts a recurrent neural network structure to remove rain streaks. It first uses 4D convolution to extract features of all sub-views of an LFI. Then, the Gate Recurrent Unit (GRU) retains and selectively transfers features to the subsequent DSTB shown in Fig.~\ref{fig:DSTB}. A DSTB is composed of several repeated blocks, each consisting of a Swin Transformer Block (STB)~\cite{liu21,liang21} and several 4D Convolution layers in a densely connected manner. The DSTB combines the advantages of CNN and Transformer. The self-attention mechanism \revised{within} STB can expand the receptive field and capture local and global dependences, and the dense connections can concatenate different levels of features for recovering the rain-free LFI.

The rain removal process of RNNAT is expressed as:
\begin{equation}
Y' = \revised{\mathrm{\mathcal{F}_{RNNAT}}}([I, R', A']),
\end{equation}
where $Y'$ denotes the de-rained sub-views produced by RNNAT. $A'$ denotes the fog maps predicted by DERNet. $\mathrm{\mathcal{F}}_{RNNAT}(\cdot)$ represents the function of RNNAT.

In order to strengthen the supervision for the network, both the $L_1$ loss and the perceptual loss are adopted in the generator:
\begin{equation}
\begin{aligned}
\mathcal{L}_{g}=&\|Y'-Y^{gt}\|_{1}+\lambda_{p,g}\|\mathrm{VGG}(Y')-\mathrm{VGG}(Y^{gt})\|_2^2,
\label{eq:Lg}
\end{aligned}
\end{equation}
where $Y'$ and $Y^{gt}$ indicate the de-rained sub-views and the ground-truth rain-free sub-views, respectively.

Global-local discriminator is adopted to guide the generator to generate much realistic de-rained sub-views and ensure consistency between the \revised{de-rained LFIs} and the ground truths for the synthetic LFIs (or pseudo ground truths obtained by subtracting the estimated rain streaks from the input rainy LFIs for real-world rainy scenes). The loss function for the global discriminator is defined as:
\begin{equation}
\mathcal{L}_{g}^{gan}=-\log(\revised{\mathrm{D_{g}}}(Y^{gt}))-\log(1-\revised{\mathrm{D_{g}}}(Y')),
\end{equation}
where $\revised{\mathrm{D_{g}}}(\cdot)$ represents the convolution operation of the global discriminator for rain streak removal.

The local discriminator is introduced to overcome the situation in some scenes, where rain streaks can be clearly removed in some regions but not in other local regions, e.g., rain streaks in distant regions. Indeed, our local discriminator, $\revised{\mathrm{D_{l}}}(\cdot)$, is an extended PatchGAN~\cite{isola2017} working together with the global GAN, much like~\cite{demir2018}. Its input is stacked local patches (3D patches) cropped from the same position of all sub-views at a time. The loss function for the local discriminator is defined in the same way as the global discriminator:
\begin{equation}
\mathcal{L}_{l}^{gan}=\frac{1}{N_p}\sum_{p=1}^{N_p}(-\log(\revised{\mathrm{D_{l}}}(Y_p^{gt}))-\log(1-\revised{\mathrm{D_{l}}}(Y_p'))),
\end{equation}
where $Y_p^{gt}$ and $Y_p'$ refer to the 3D local patches with spatial resolution of $[64, 64]$ cropped from $Y^{gt}$ and $Y'$, respectively. $N_p$ refers to the number of 3D local patches and is set to $4$.

Thus, the loss function for training RNNAT is:
\begin{equation}
\mathcal{L}_{derain}=\mathcal{L}_{g} +\lambda_{gan}(\mathcal{L}_{g}^{gan}+\mathcal{L}_{l}^{gan}),
\label{eq:Ld}
\end{equation}
where $\lambda_{gan}$ is a constant weighting parameter.

Concerning real-world rainy LFIs without ground truths, we subtract the estimated rain streaks from the rainy sub-views as pseudo ground truth for rain-free LFI recovery. Thus, both synthetic LFIs and real-world LFIs can be used to train our RNNAT. In this way, both the performance and generalization of our proposed method can be improved.

The overall loss function for training 4D-MGP-SRRNet is:
\begin{equation}
\mathcal{L}_{total}= \mathcal{L}_{rain}+ \mathcal{L}_{derain}.
\end{equation}

\renewcommand{\subwidth}{0.116}
\newcommand{\ssubwidth}{0.058}
\begin{figure*}[t]
	\renewcommand{\tabcolsep}{0.8pt}
	\renewcommand\arraystretch{0.8}
	\begin{center}
		\begin{tabular}{cccccccccccccccc}
            \multicolumn{2}{c}{\includegraphics[width=\subwidth\linewidth]{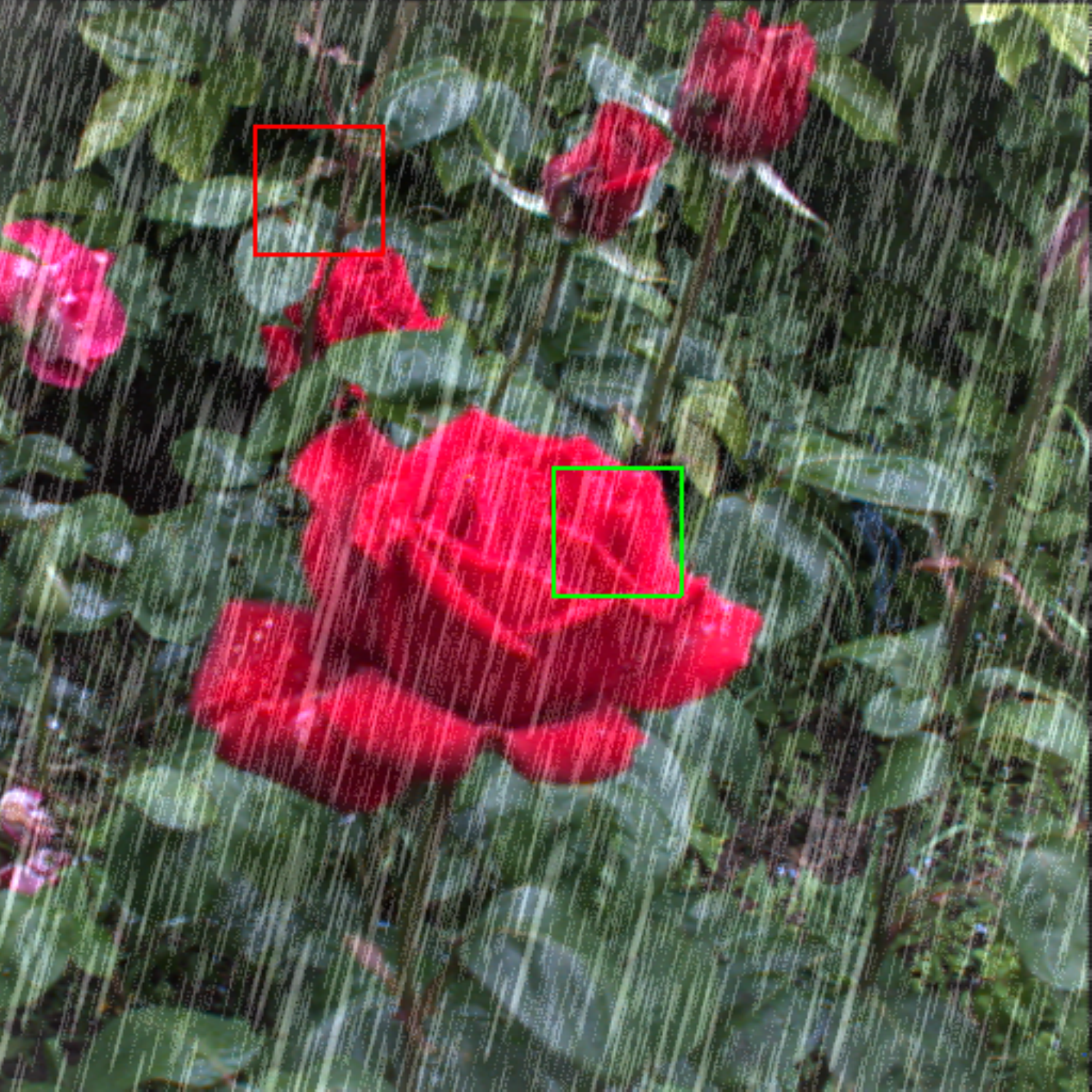}} &
            \multicolumn{2}{c}{\includegraphics[width=\subwidth\linewidth]{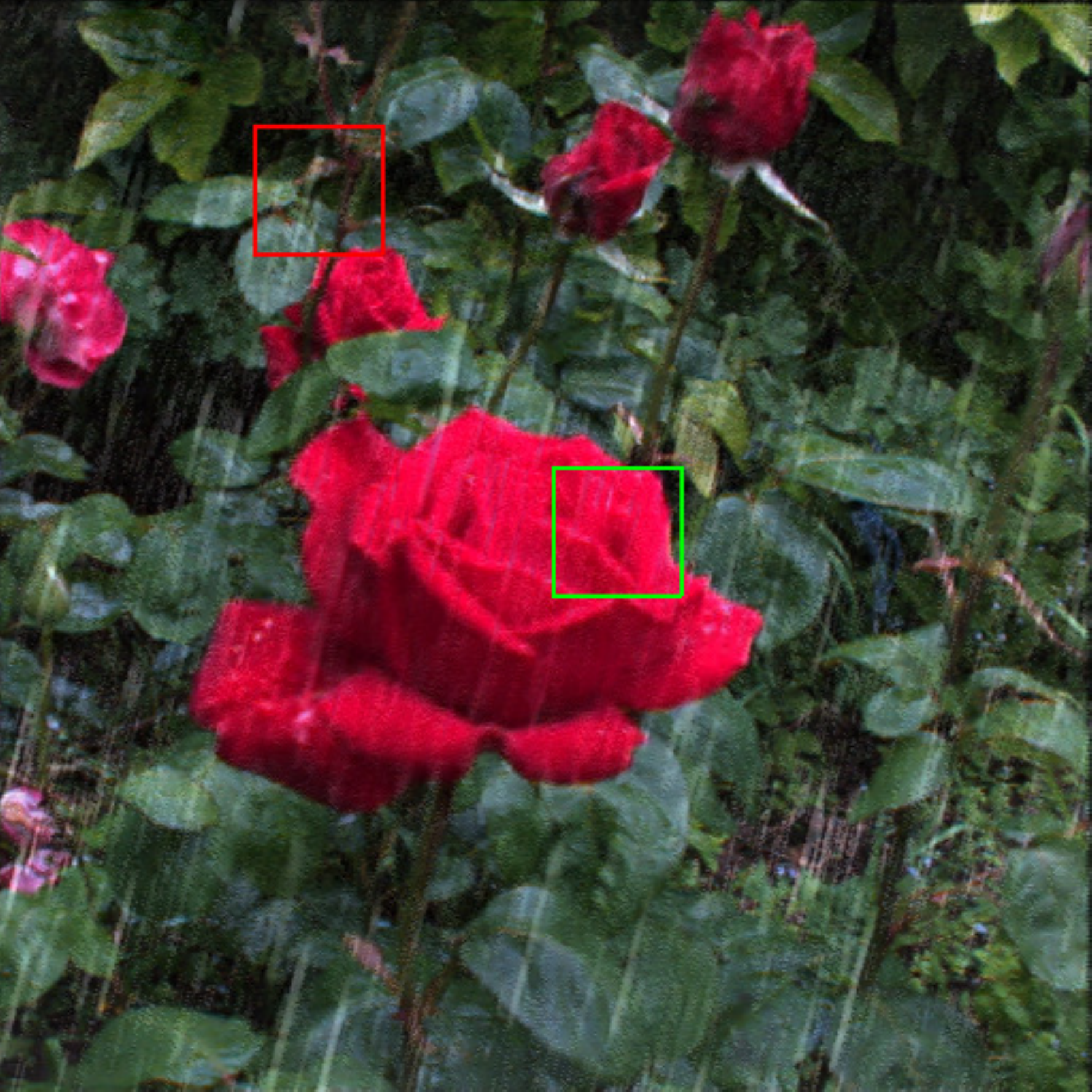}} &
            \multicolumn{2}{c}{\includegraphics[width=\subwidth\linewidth]{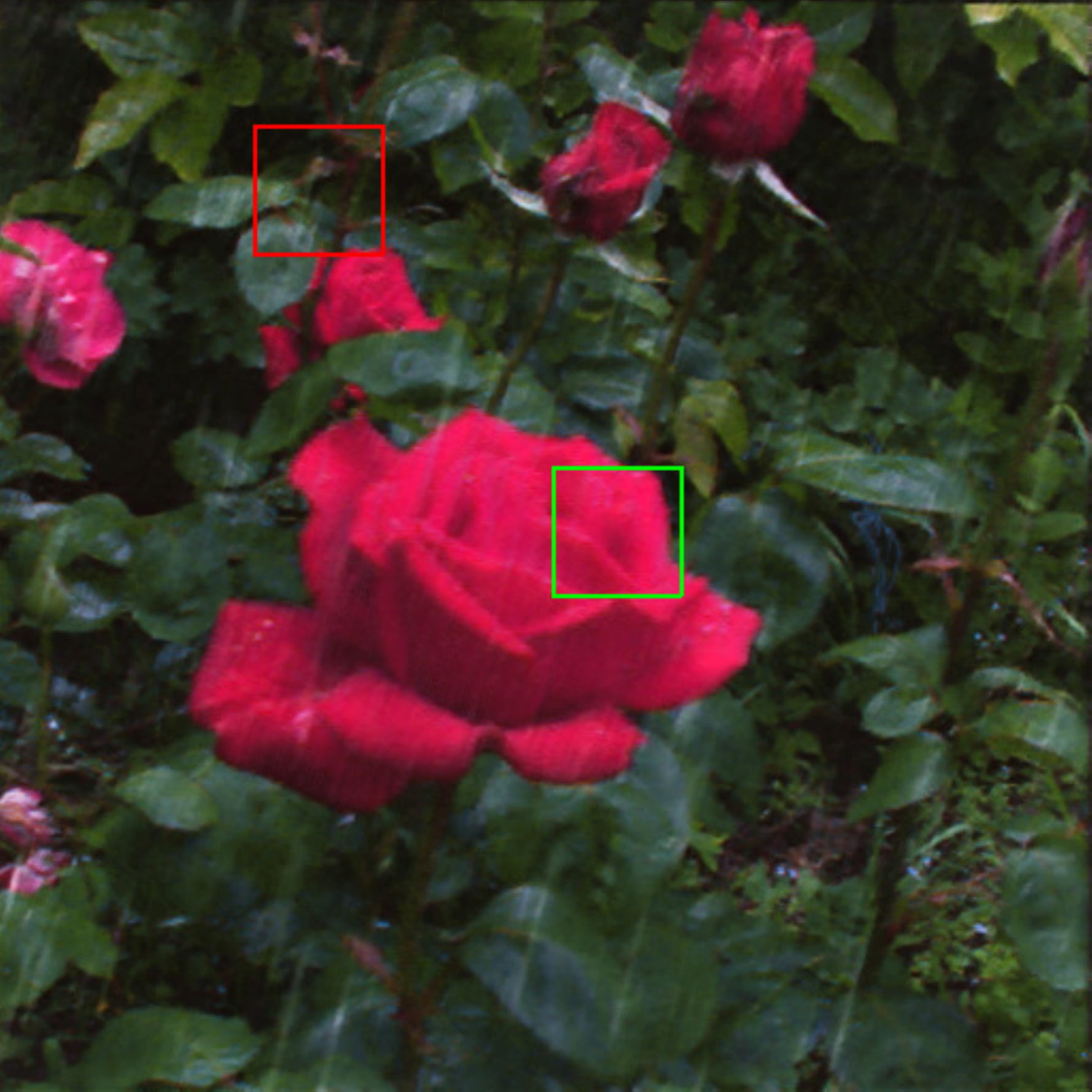}} &
            \multicolumn{2}{c}{\includegraphics[width=\subwidth\linewidth]{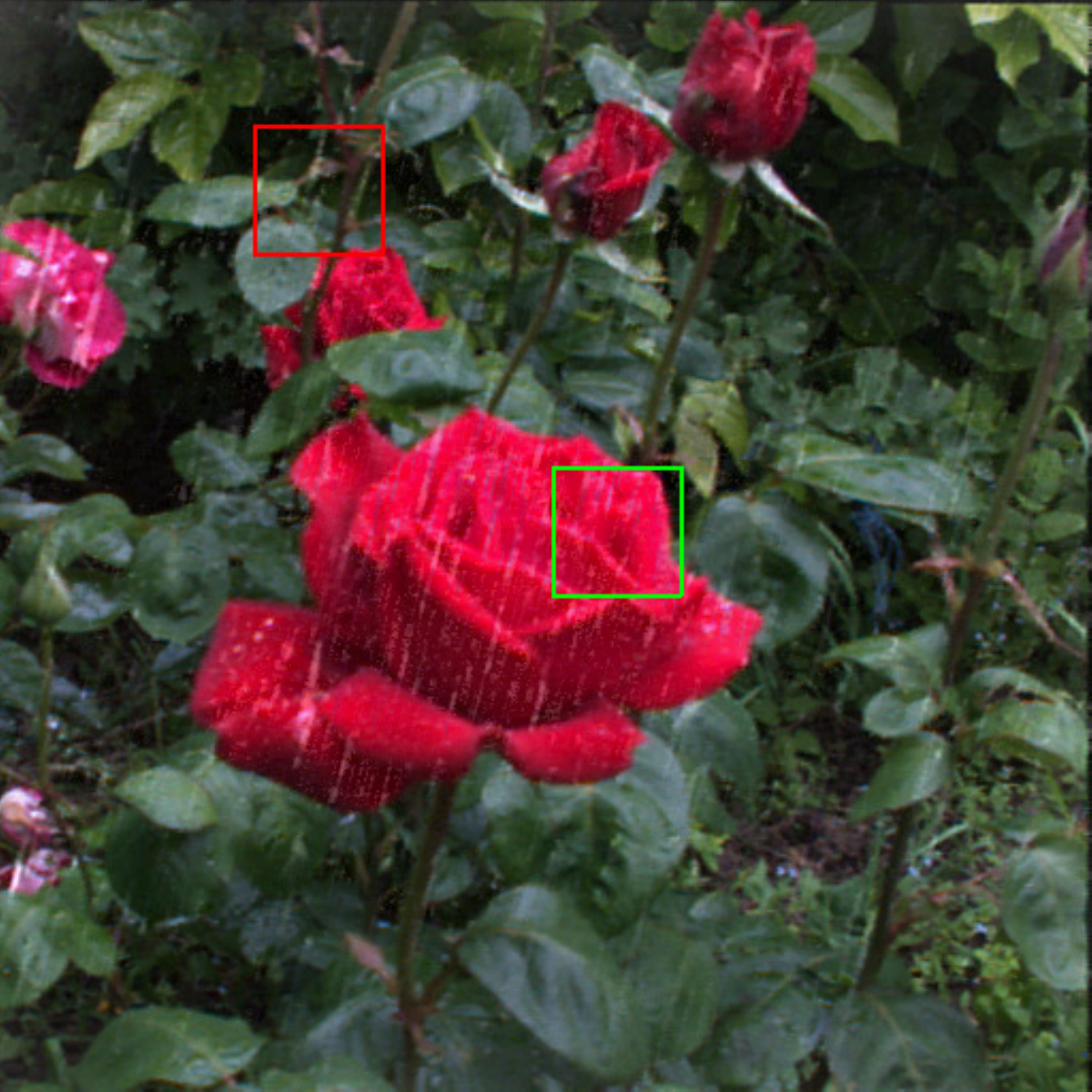}} &
            \multicolumn{2}{c}{\includegraphics[width=\subwidth\linewidth]{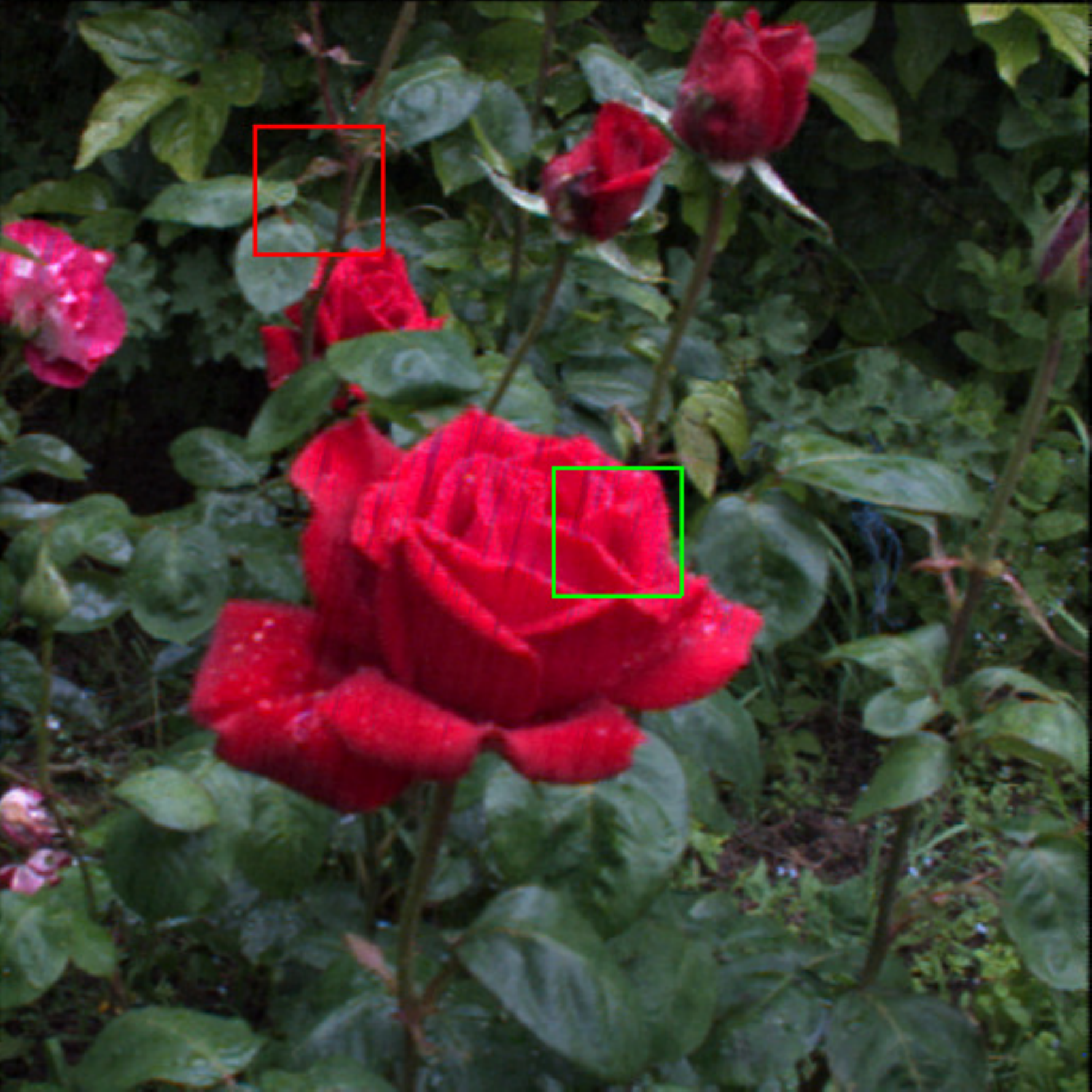}} &
            \multicolumn{2}{c}{\includegraphics[width=\subwidth\linewidth]{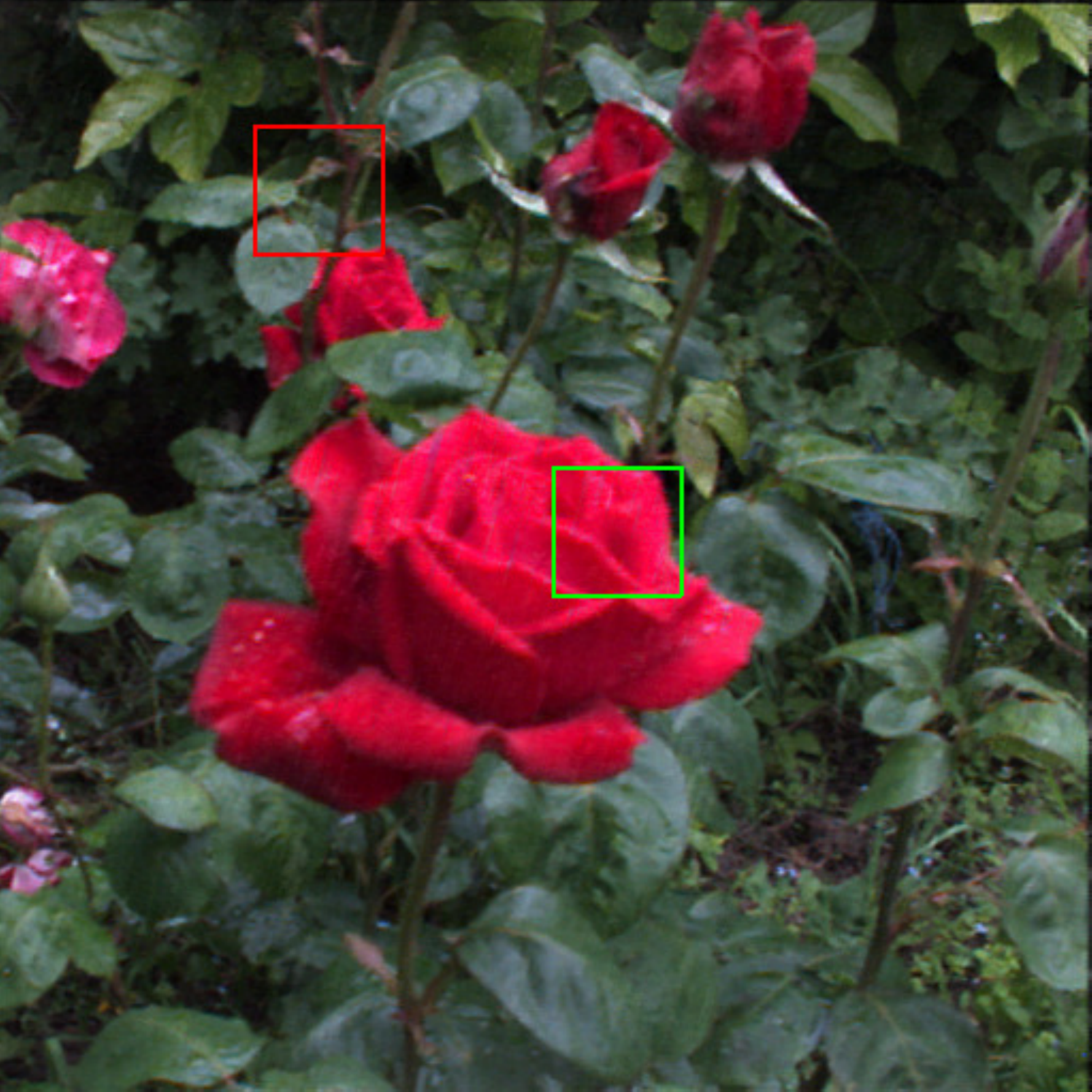}} &
            \multicolumn{2}{c}{\includegraphics[width=\subwidth\linewidth]{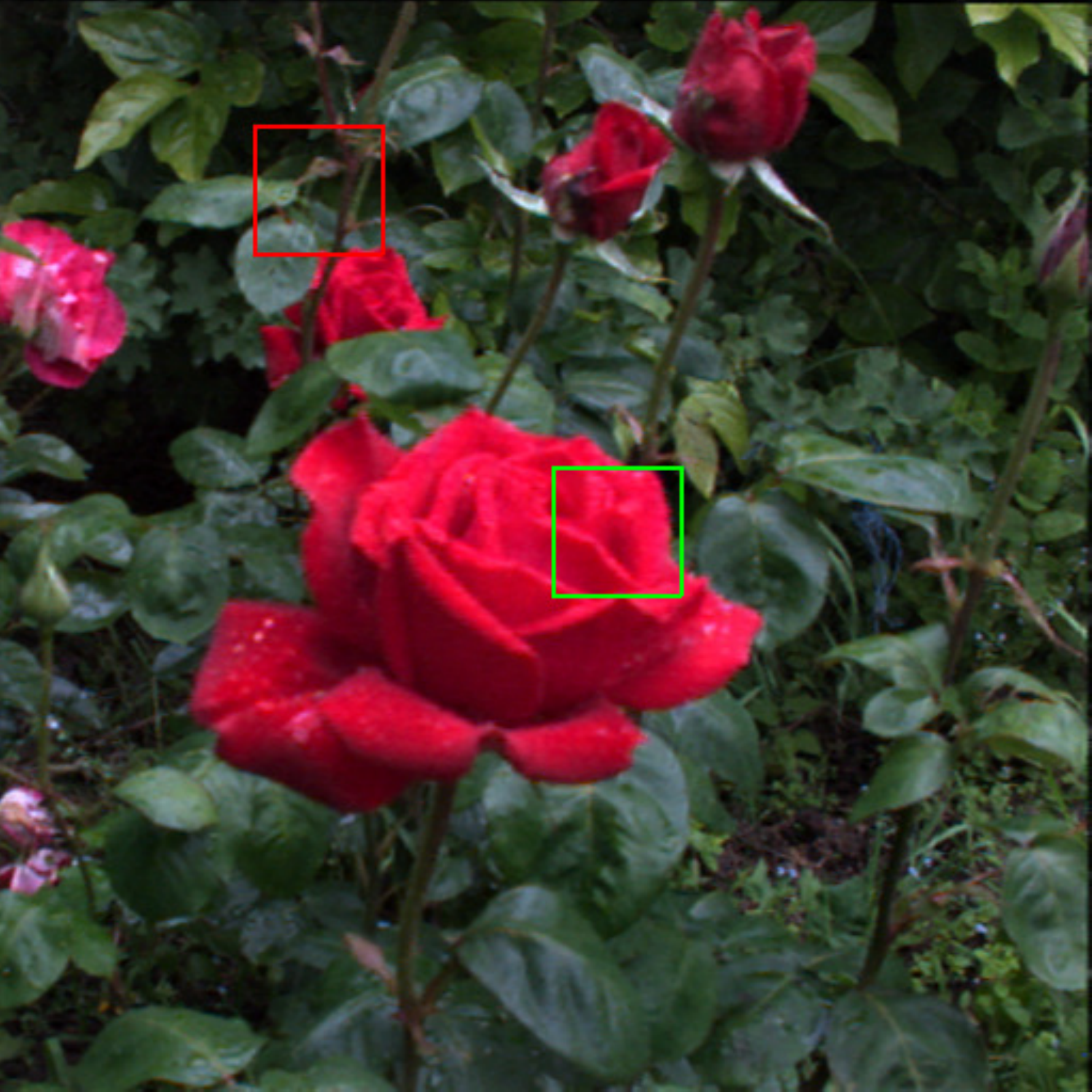}}&
            \multicolumn{2}{c}{\includegraphics[width=\subwidth\linewidth]{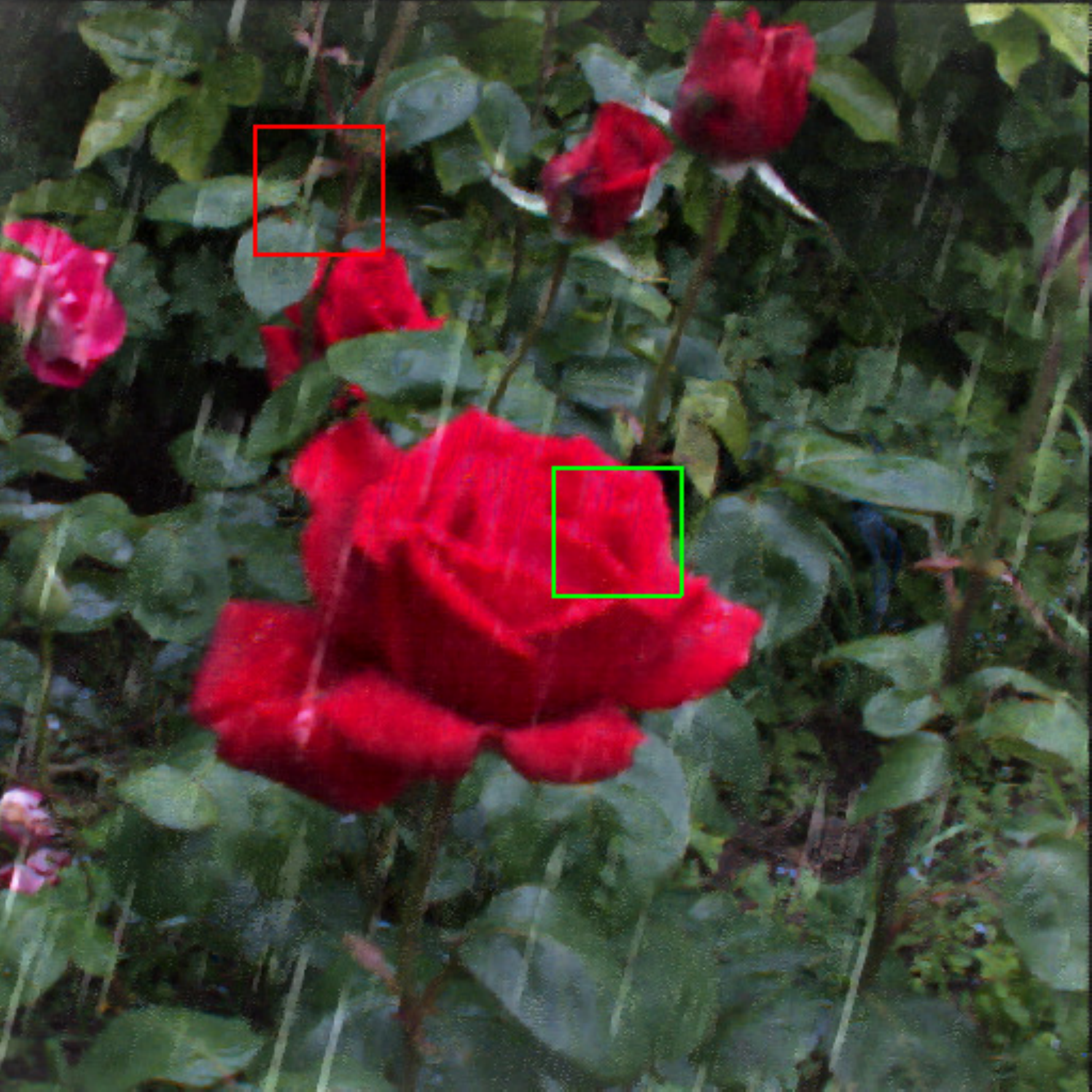}}  \\

            \includegraphics[width=\ssubwidth\linewidth]{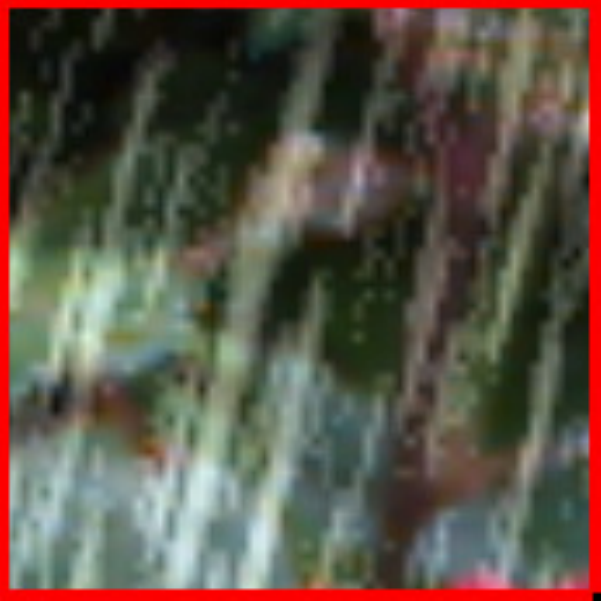} &
            \includegraphics[width=\ssubwidth\linewidth]{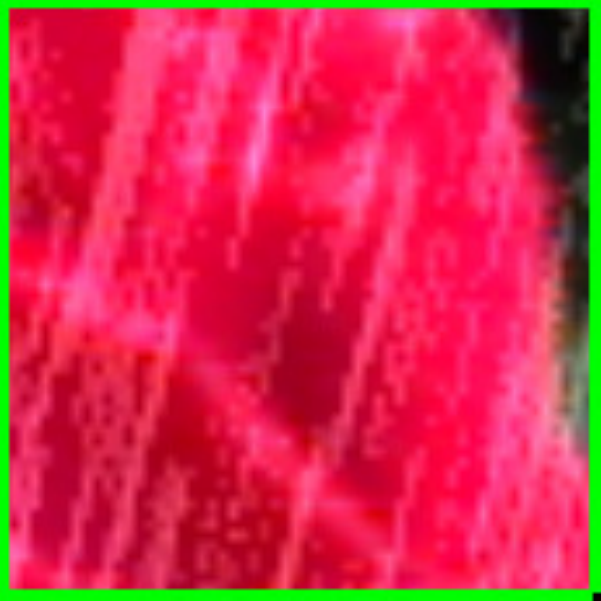} &
            \includegraphics[width=\ssubwidth\linewidth]{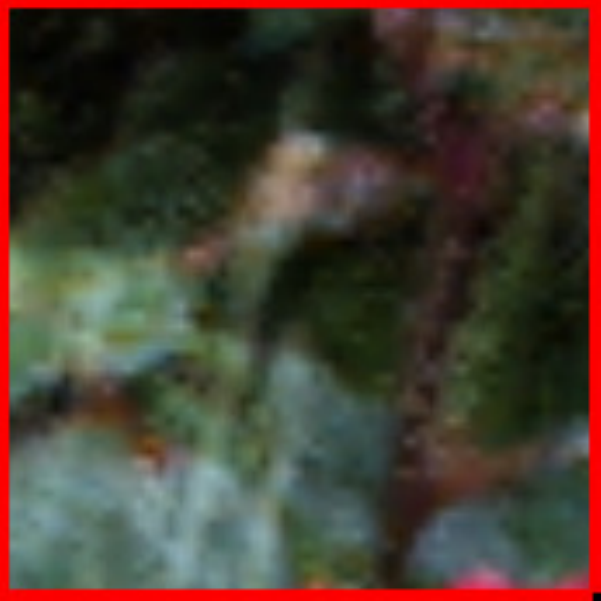} &
            \includegraphics[width=\ssubwidth\linewidth]{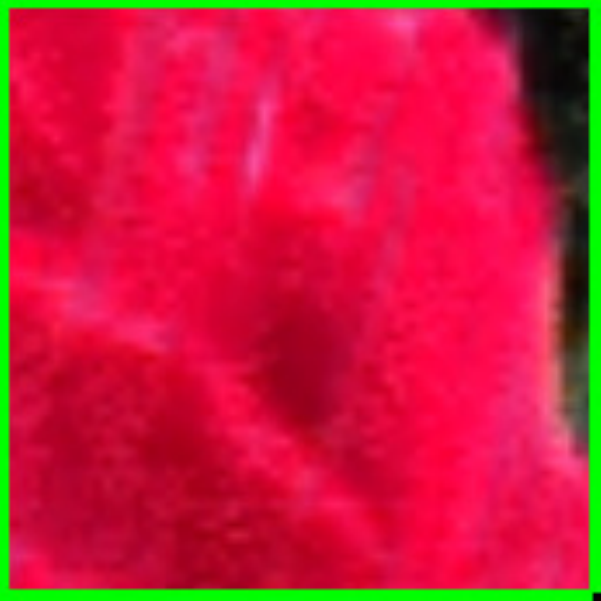} &
            \includegraphics[width=\ssubwidth\linewidth]{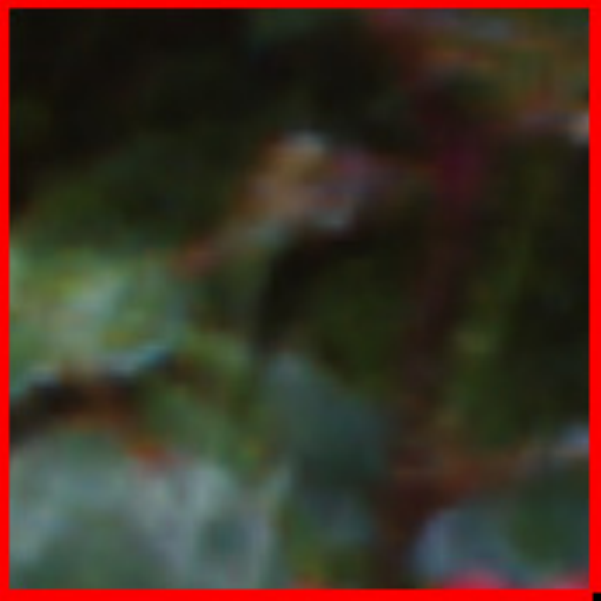} &
            \includegraphics[width=\ssubwidth\linewidth]{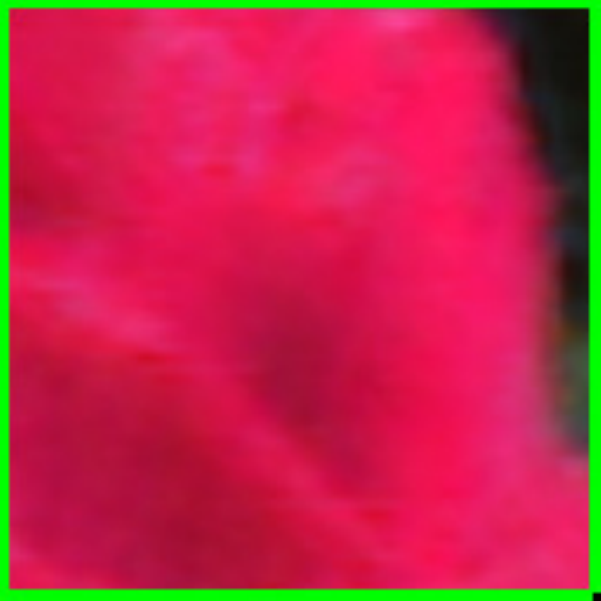} &
            \includegraphics[width=\ssubwidth\linewidth]{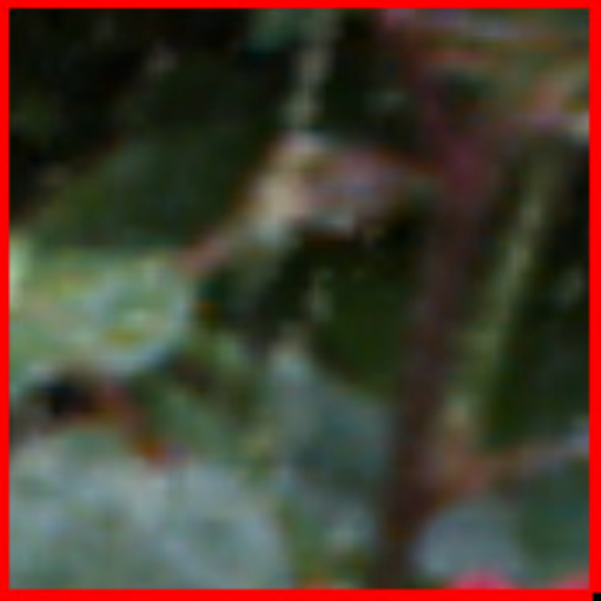} &
            \includegraphics[width=\ssubwidth\linewidth]{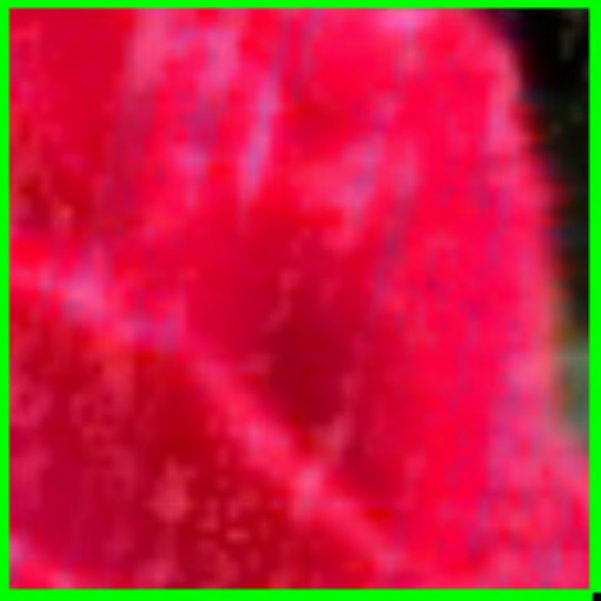} &
            \includegraphics[width=\ssubwidth\linewidth]{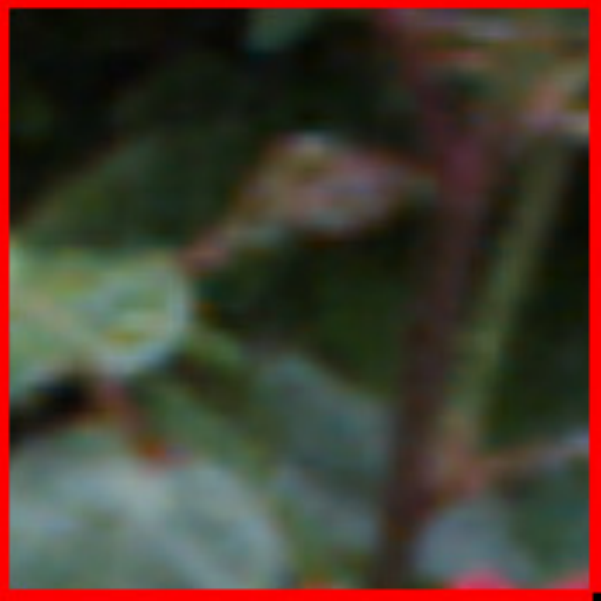} &
            \includegraphics[width=\ssubwidth\linewidth]{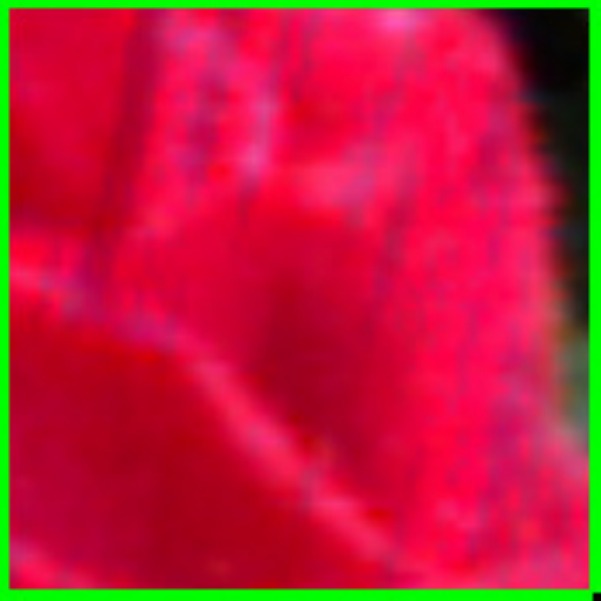} &
            \includegraphics[width=\ssubwidth\linewidth]{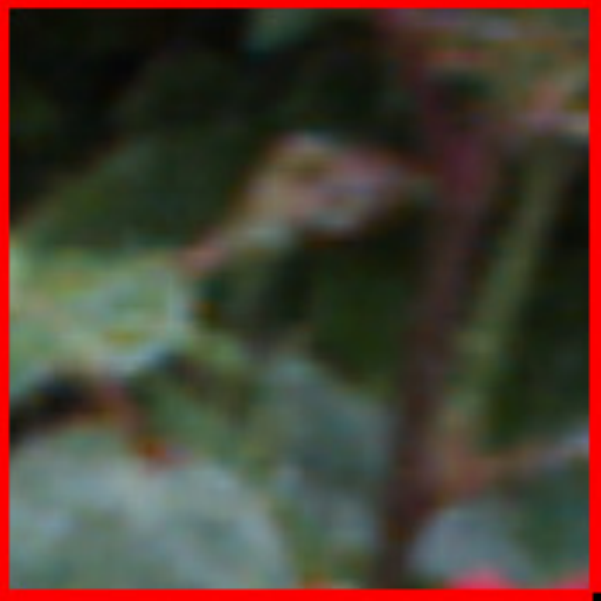} &
            \includegraphics[width=\ssubwidth\linewidth]{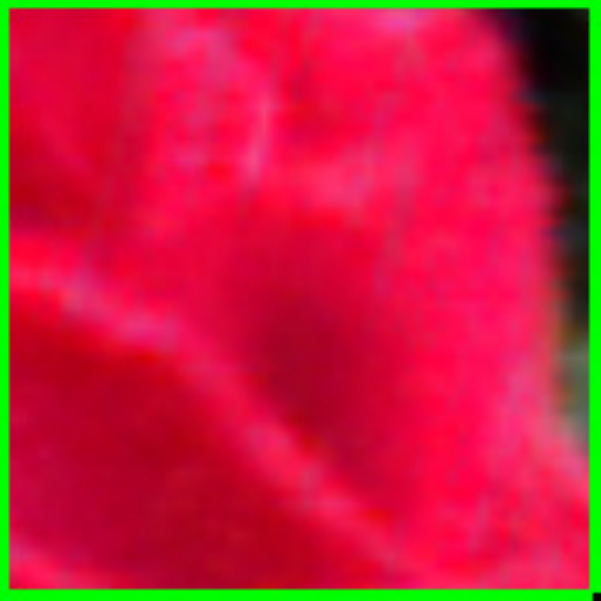} &
            \includegraphics[width=\ssubwidth\linewidth]{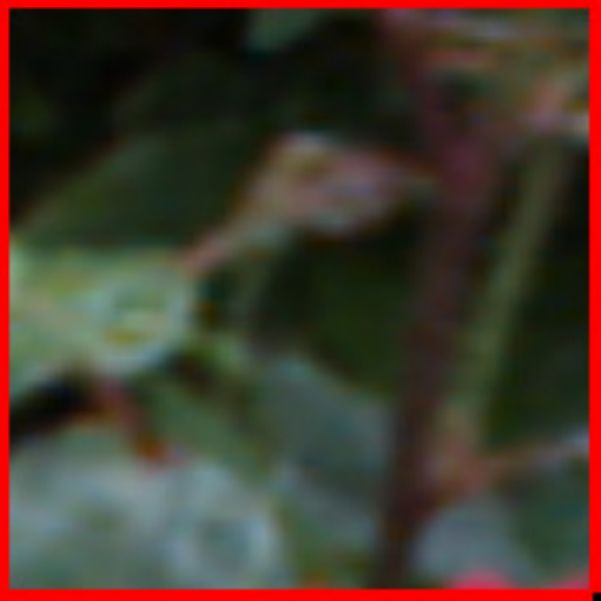} &
            \includegraphics[width=\ssubwidth\linewidth]{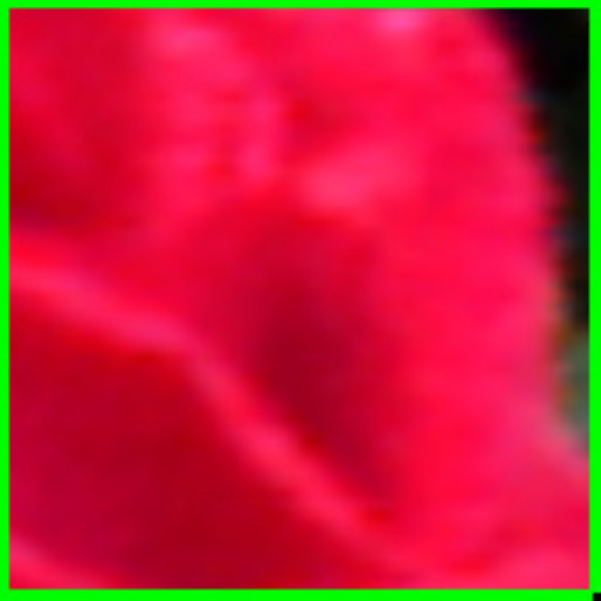} &
            \includegraphics[width=\ssubwidth\linewidth]{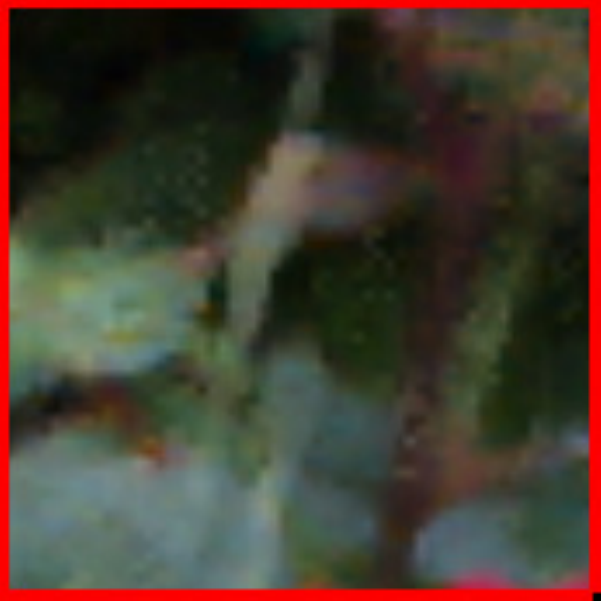} &
            \includegraphics[width=\ssubwidth\linewidth]{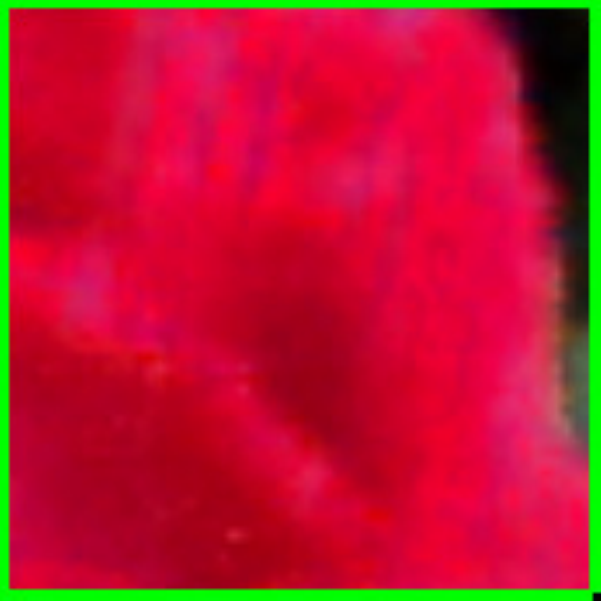} \\

            \multicolumn{2}{c}{\scriptsize{Input, 16.79/0.512}}&
            \multicolumn{2}{c}{\scriptsize{~\cite{wang19b}, 24.90/0.797}} &
            \multicolumn{2}{c}{\scriptsize{~\cite{Li19a}, 22.70/0.834}} &
            \multicolumn{2}{c}{\scriptsize{~\cite{wei19}, 27.38/0.927}} &
            \multicolumn{2}{c}{\scriptsize{~\cite{jiang2020}, 32.85/0.975}} &
            \multicolumn{2}{c}{\scriptsize{~\cite{Yang20b}, 30.60/0.958}} &
            \multicolumn{2}{c}{\scriptsize{~\cite{ren20}, 29.45/0.970}}&
            \multicolumn{2}{c}{\scriptsize{~\cite{jiang20}, 24.65/0.834}}\\

            \multicolumn{2}{c}{\includegraphics[width=\subwidth\linewidth]{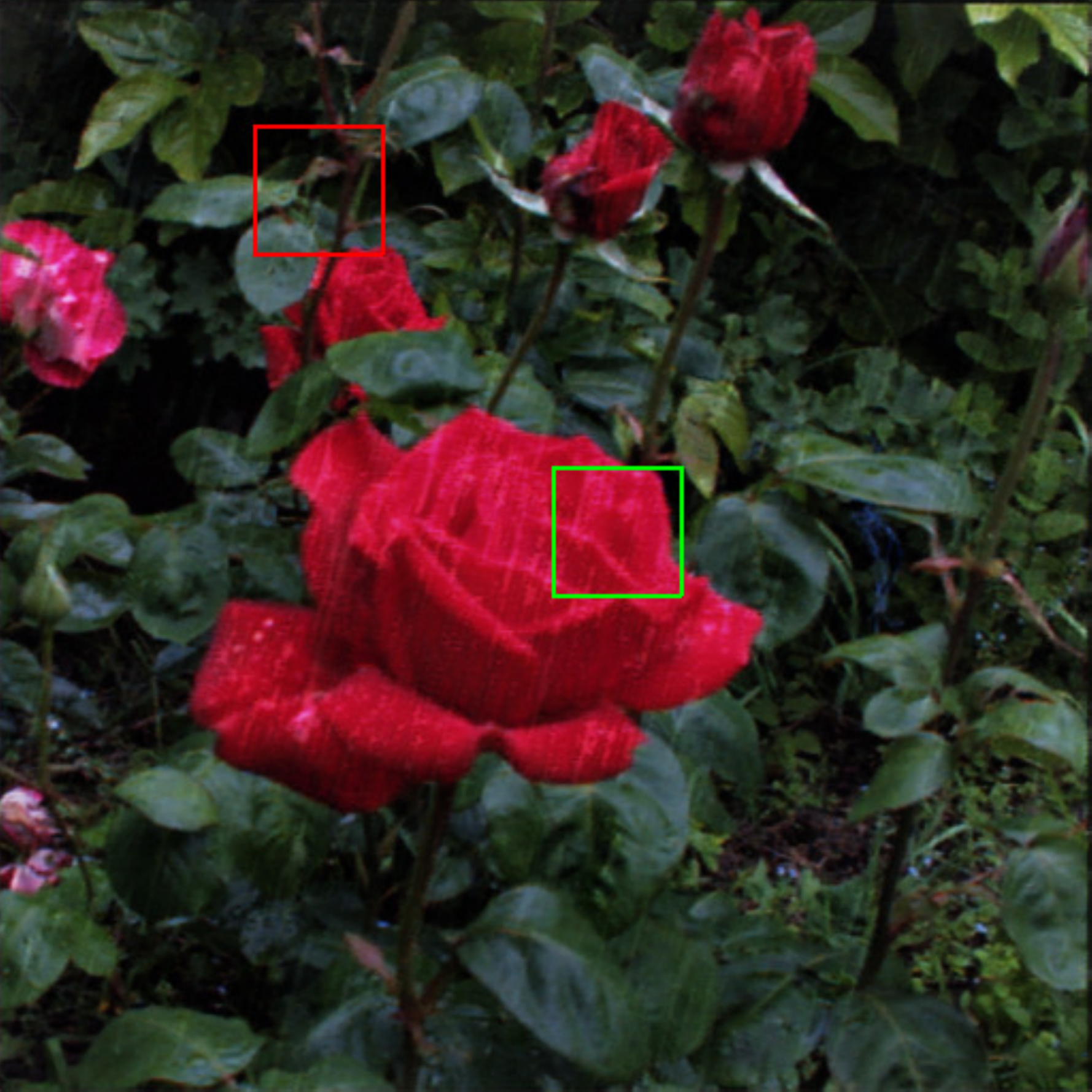}} &
            \multicolumn{2}{c}{\includegraphics[width=\subwidth\linewidth]{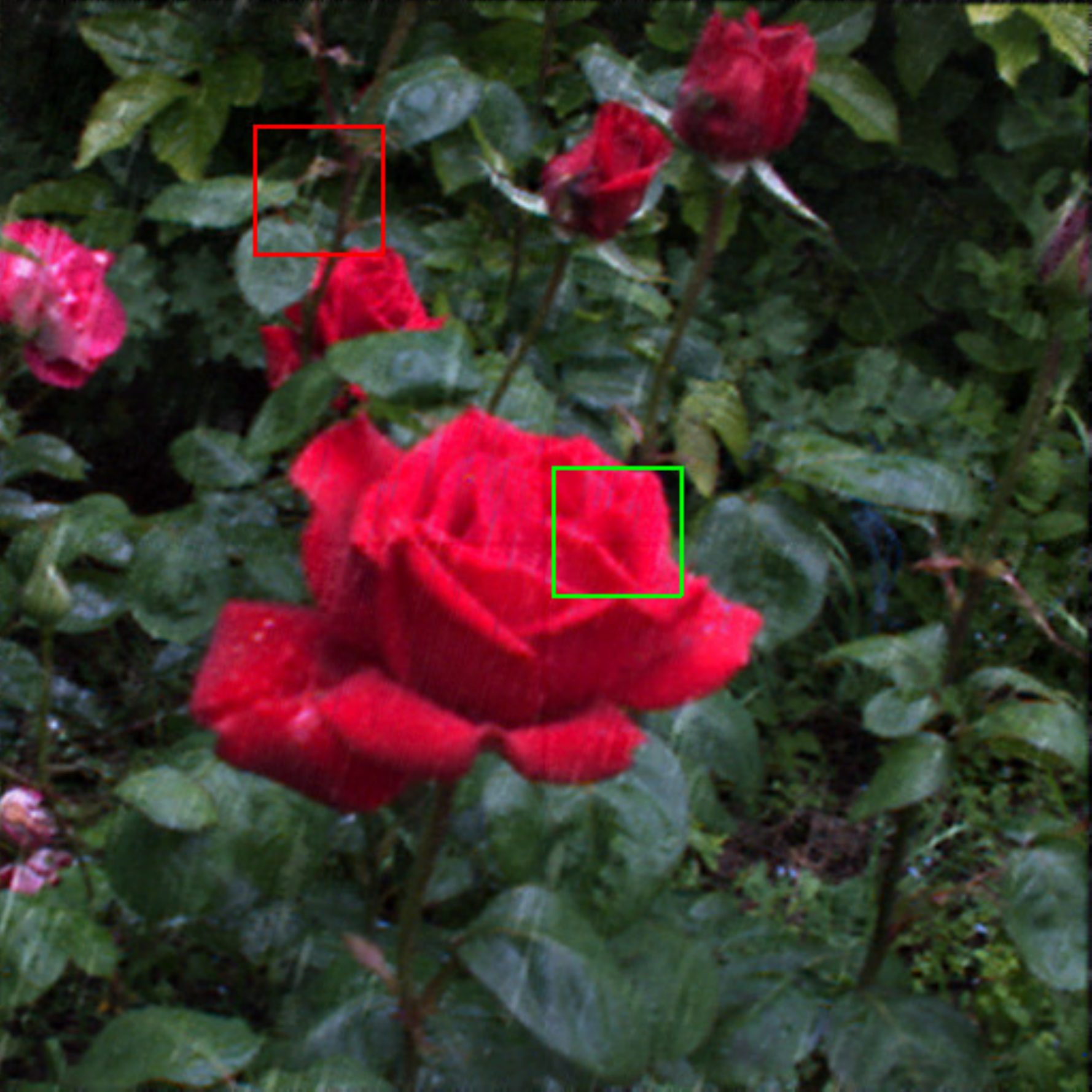}} &
            \multicolumn{2}{c}{\includegraphics[width=\subwidth\linewidth]{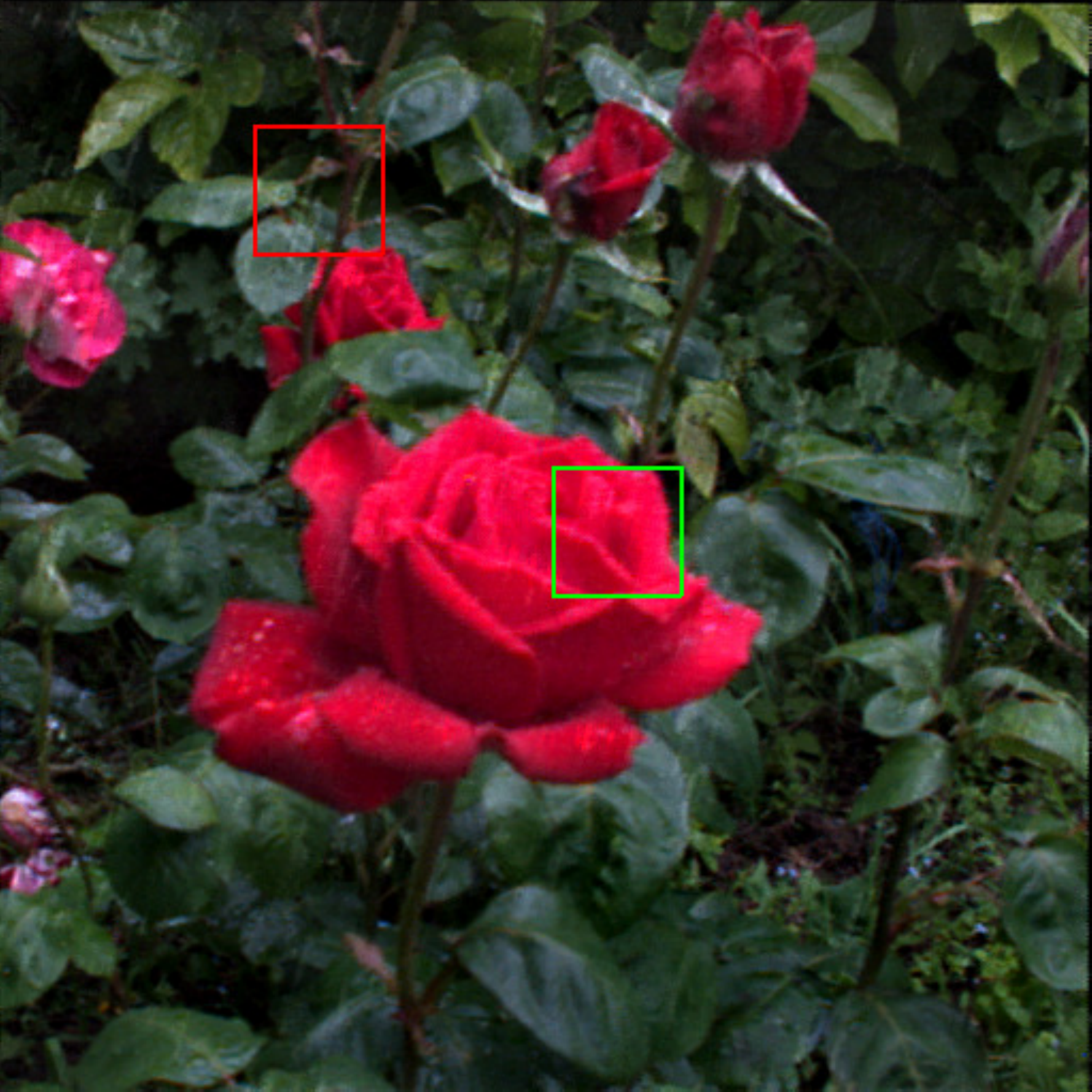}} &
            \multicolumn{2}{c}{\includegraphics[width=\subwidth\linewidth]{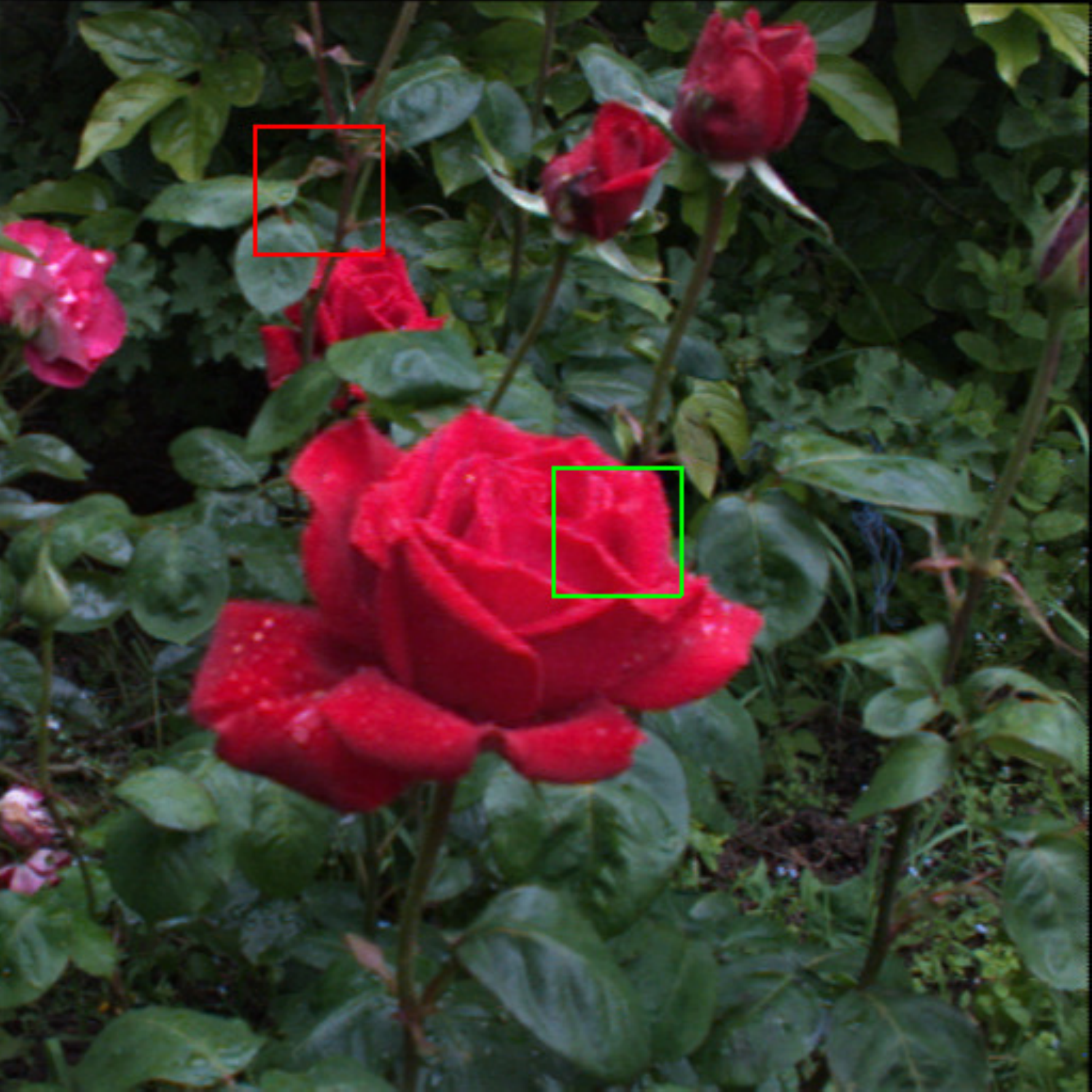}} &
            \multicolumn{2}{c}{\includegraphics[width=\subwidth\linewidth]{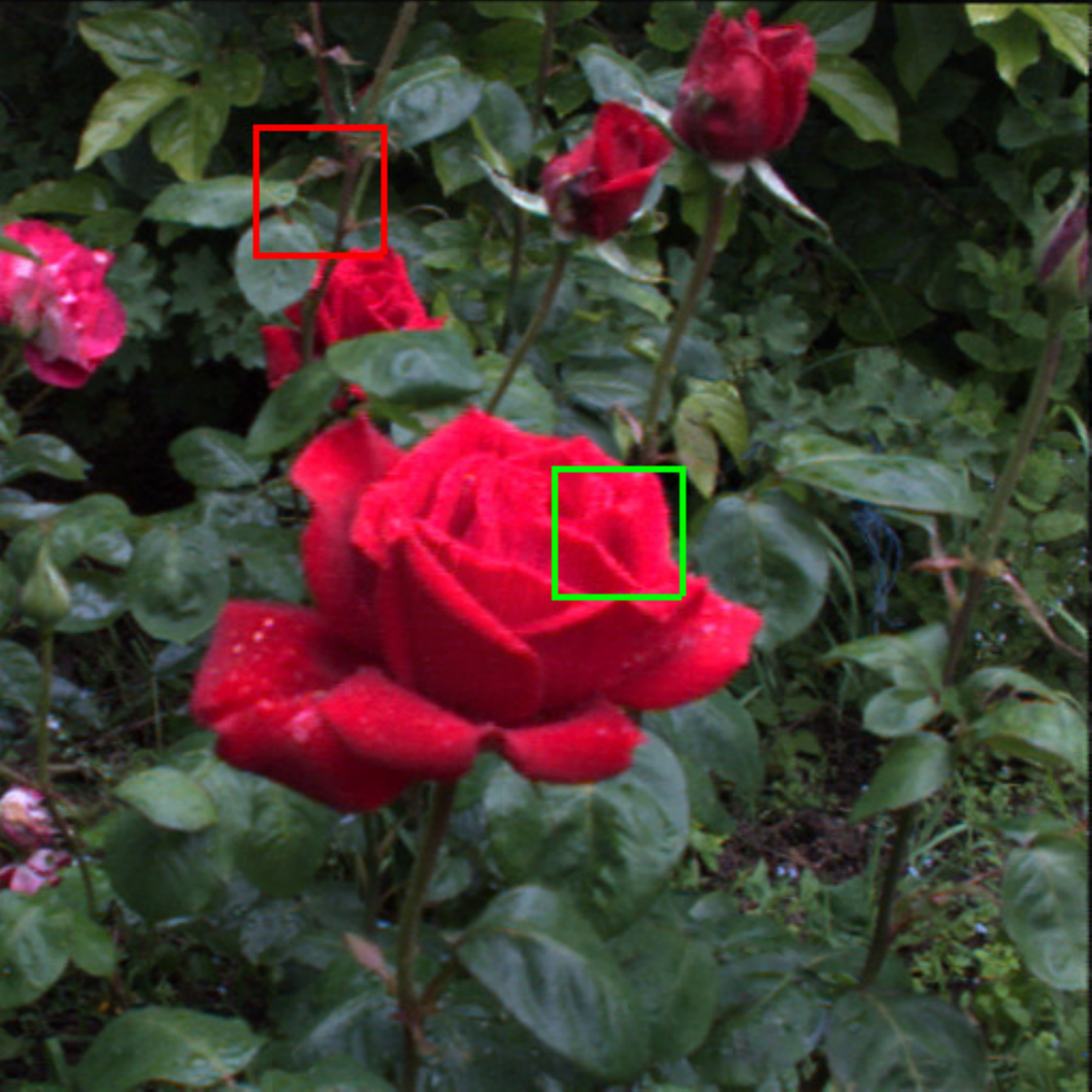}} &
            \multicolumn{2}{c}{\includegraphics[width=\subwidth\linewidth]{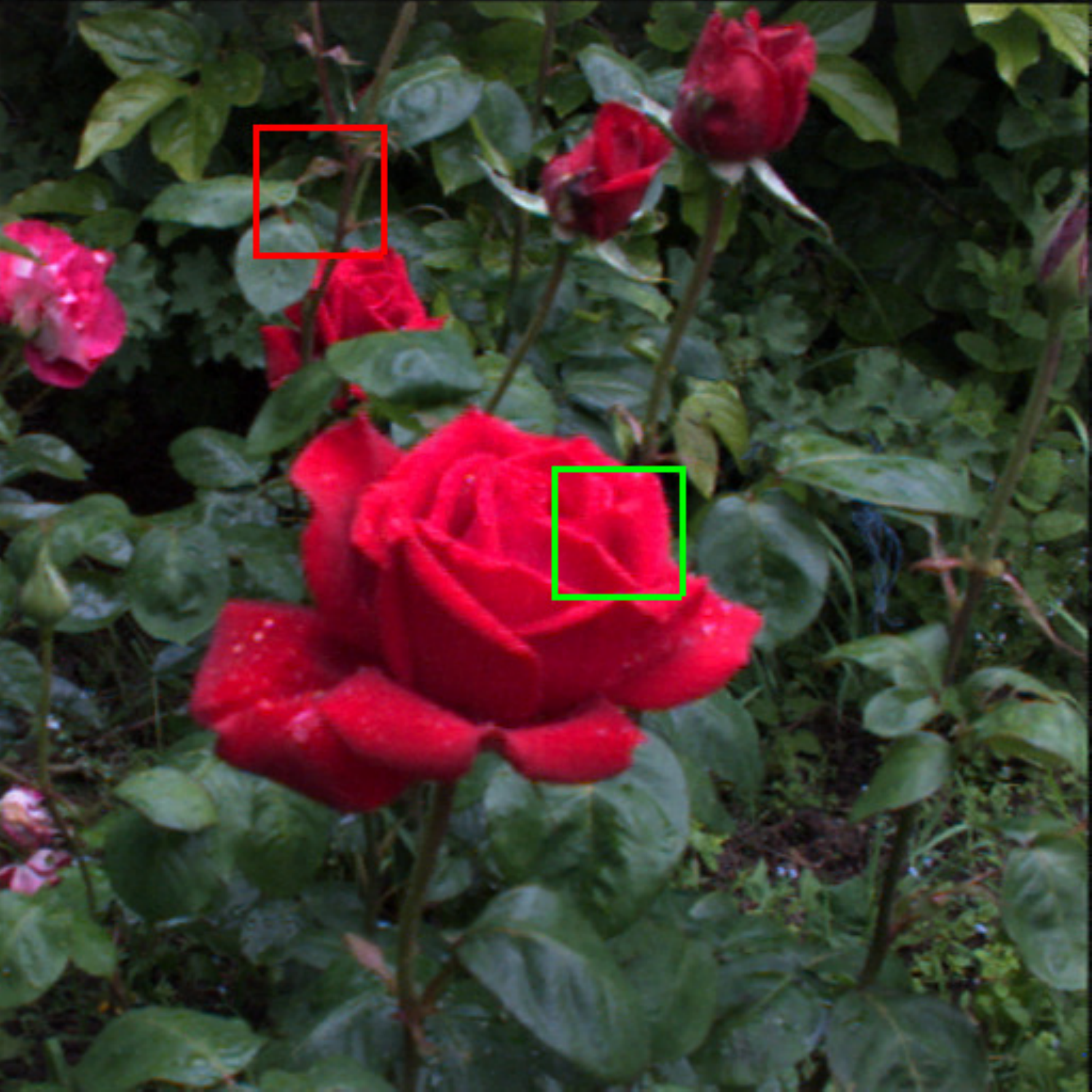}} &
            \multicolumn{2}{c}{\includegraphics[width=\subwidth\linewidth]{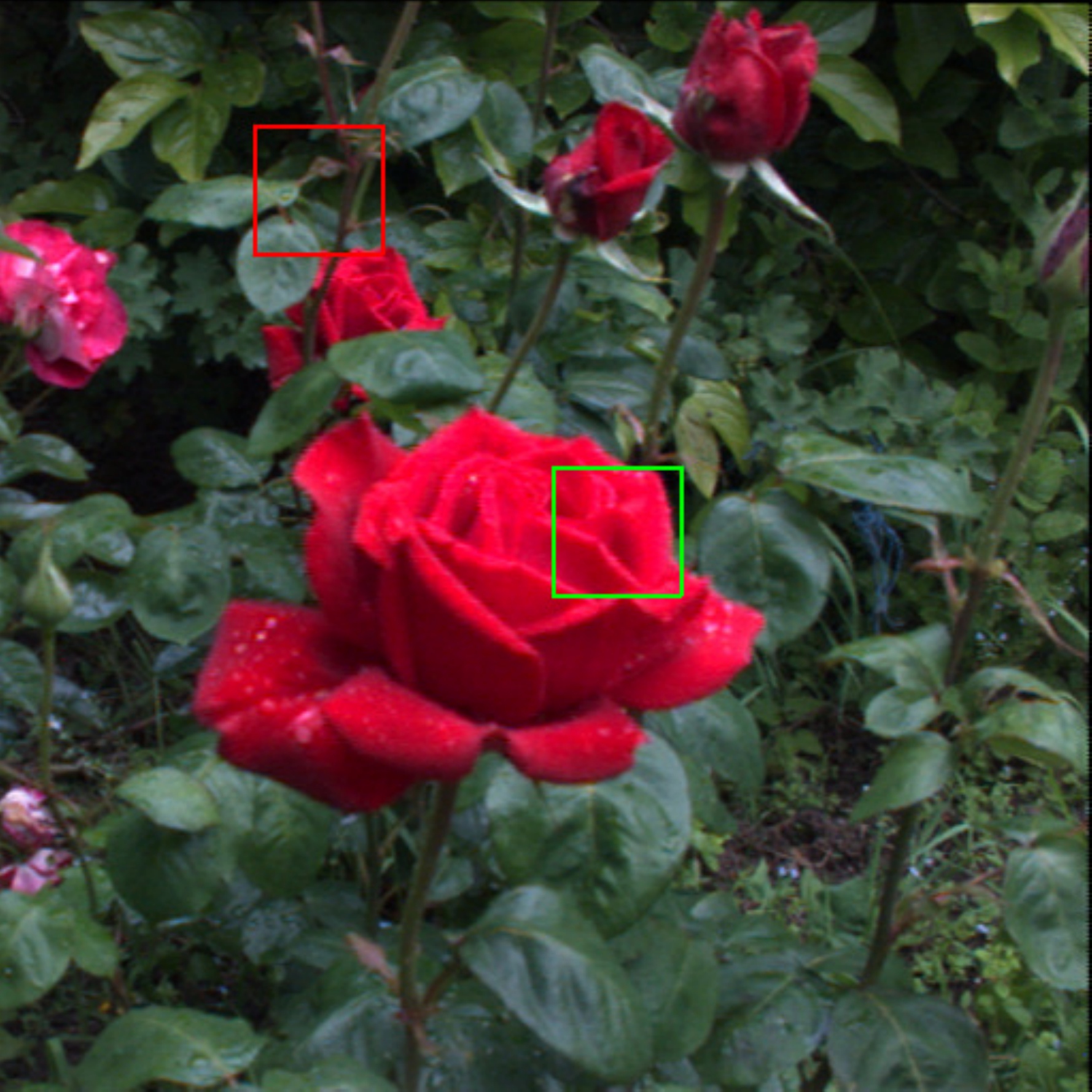}} &
            \multicolumn{2}{c}{\includegraphics[width=\subwidth\linewidth]{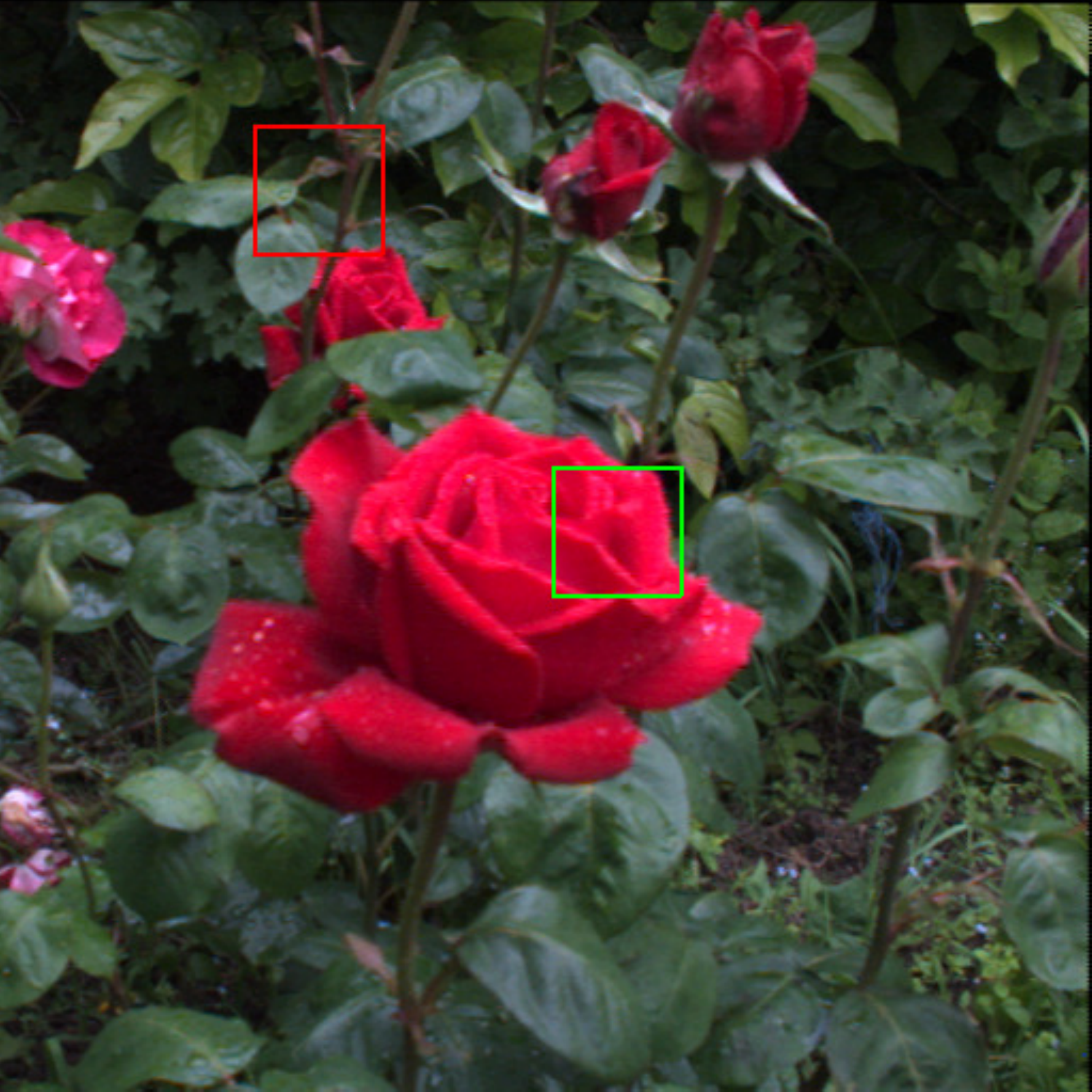}} \\

            \includegraphics[width=\ssubwidth\linewidth]{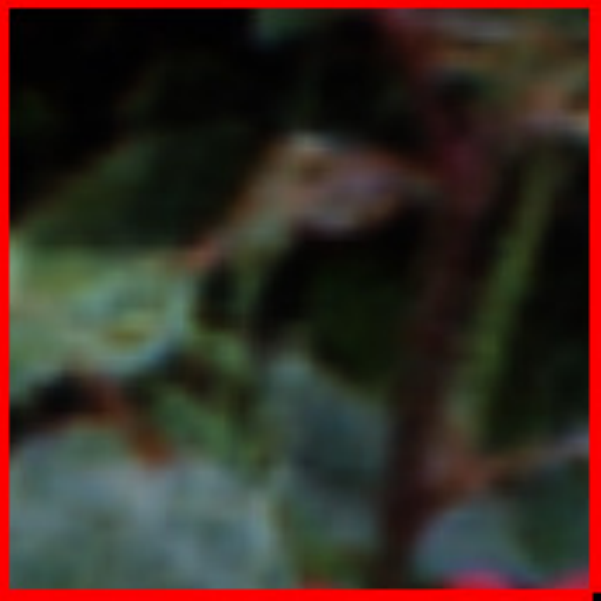} &
            \includegraphics[width=\ssubwidth\linewidth]{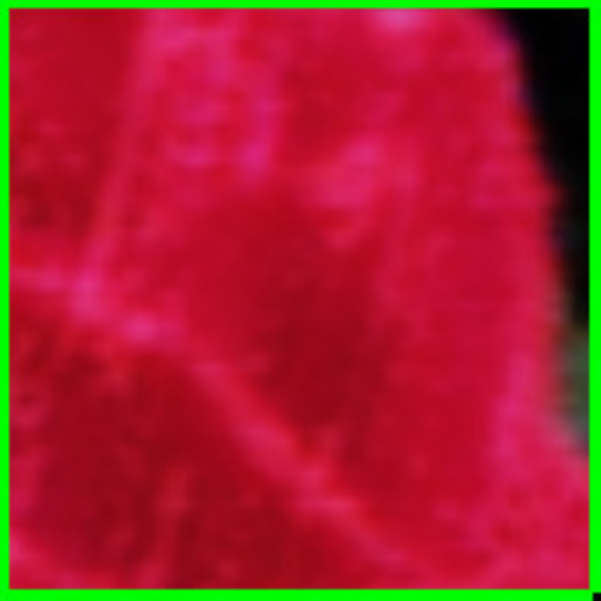} &
            \includegraphics[width=\ssubwidth\linewidth]{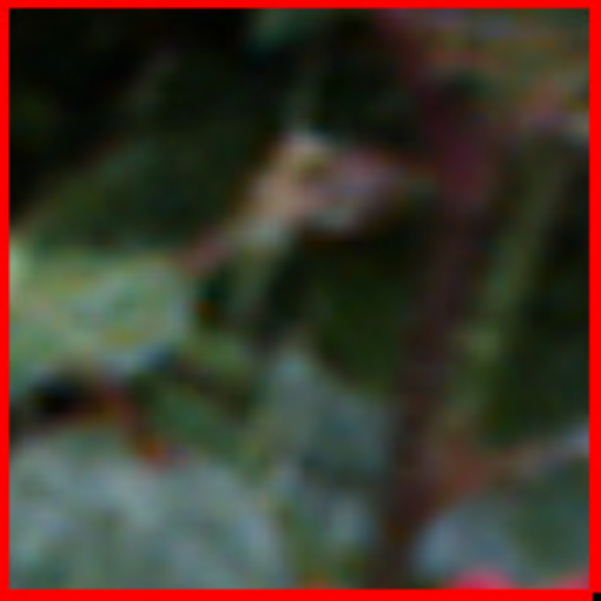} &
            \includegraphics[width=\ssubwidth\linewidth]{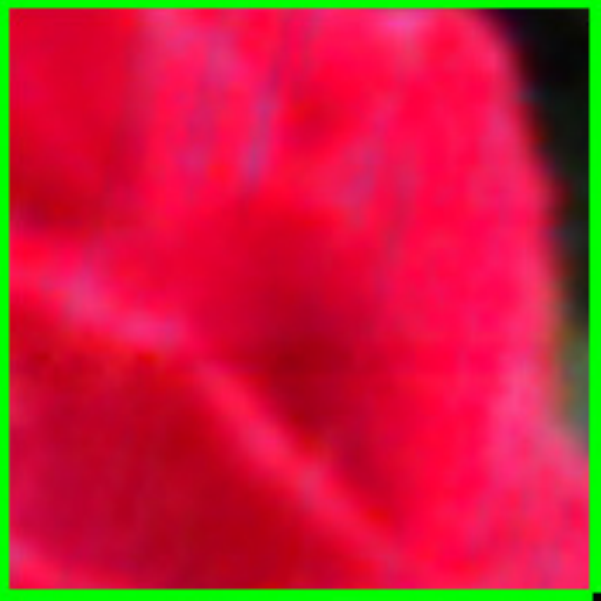} &
            \includegraphics[width=\ssubwidth\linewidth]{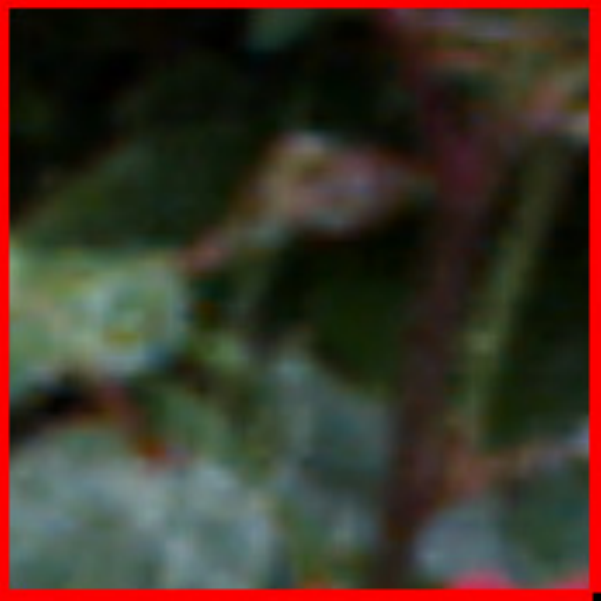} &
            \includegraphics[width=\ssubwidth\linewidth]{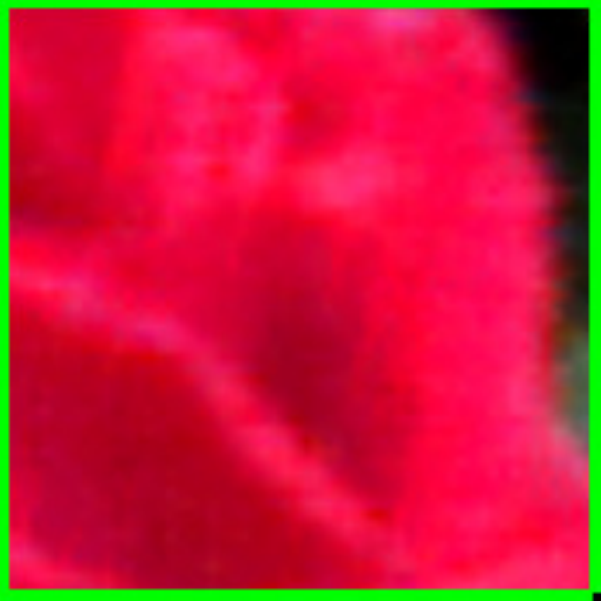} &
            \includegraphics[width=\ssubwidth\linewidth]{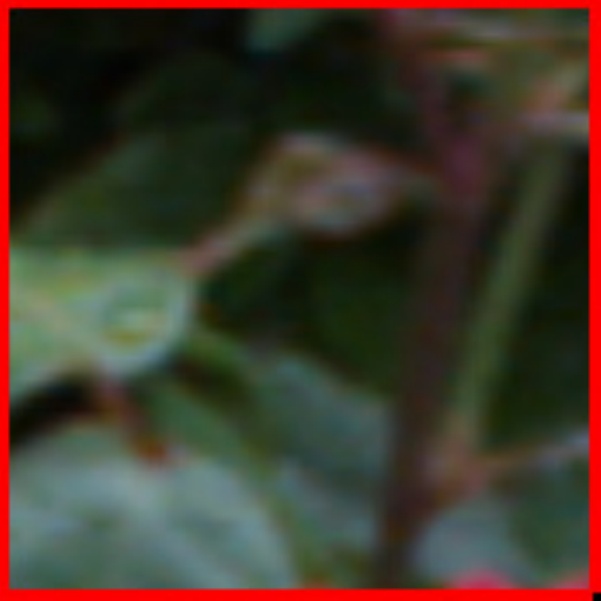} &
            \includegraphics[width=\ssubwidth\linewidth]{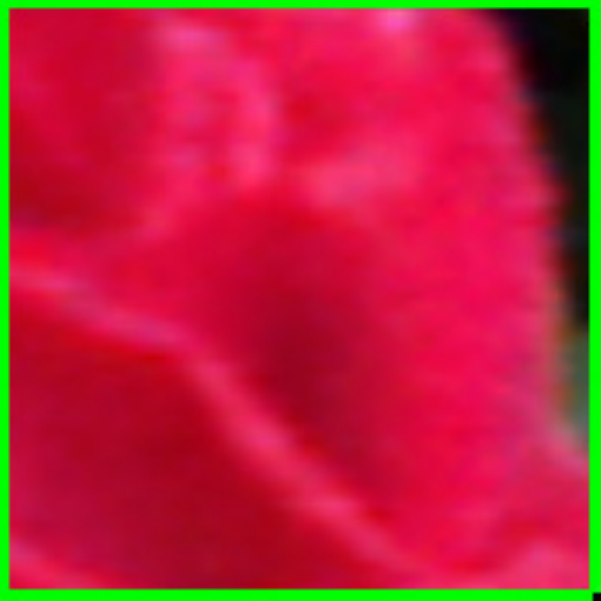} &
            \includegraphics[width=\ssubwidth\linewidth]{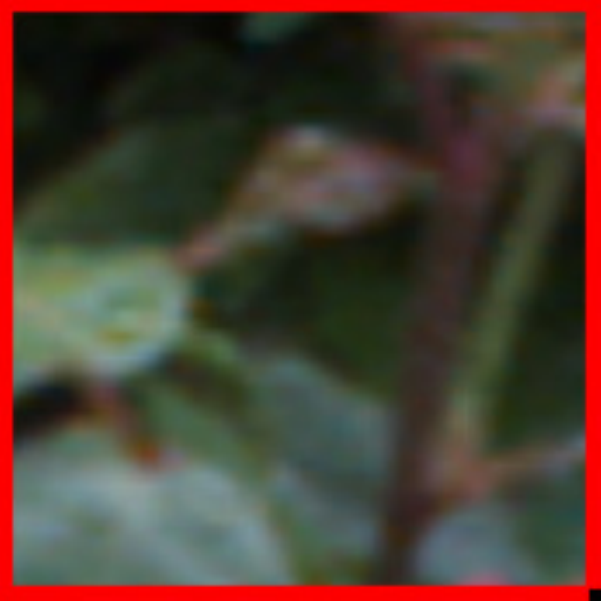} &
            \includegraphics[width=\ssubwidth\linewidth]{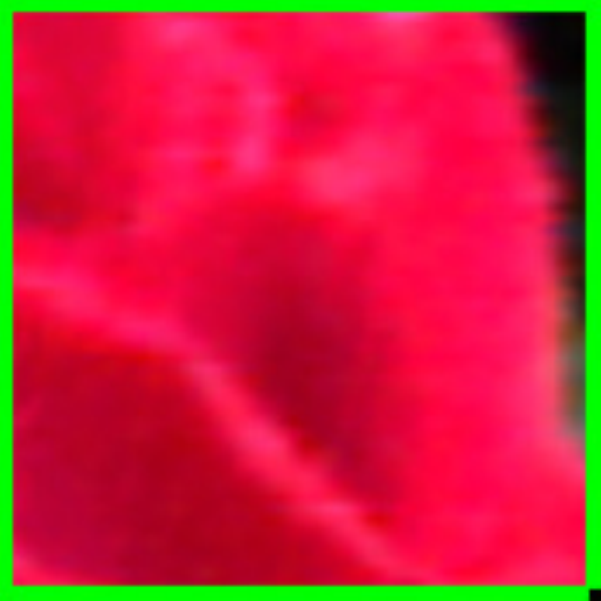} &
            \includegraphics[width=\ssubwidth\linewidth]{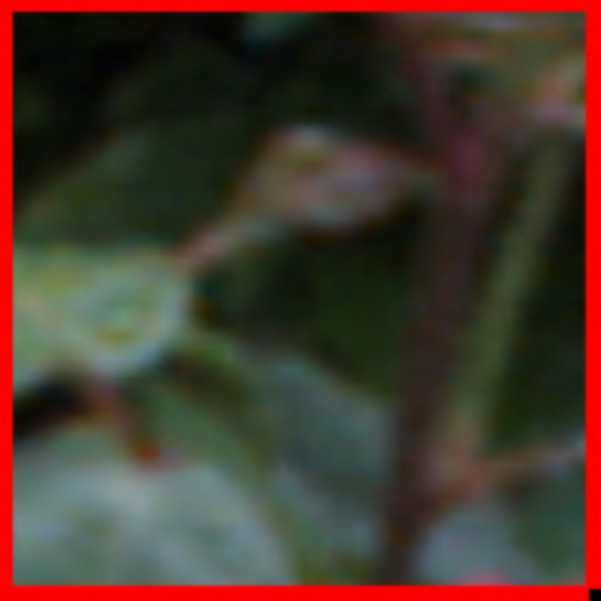} &
            \includegraphics[width=\ssubwidth\linewidth]{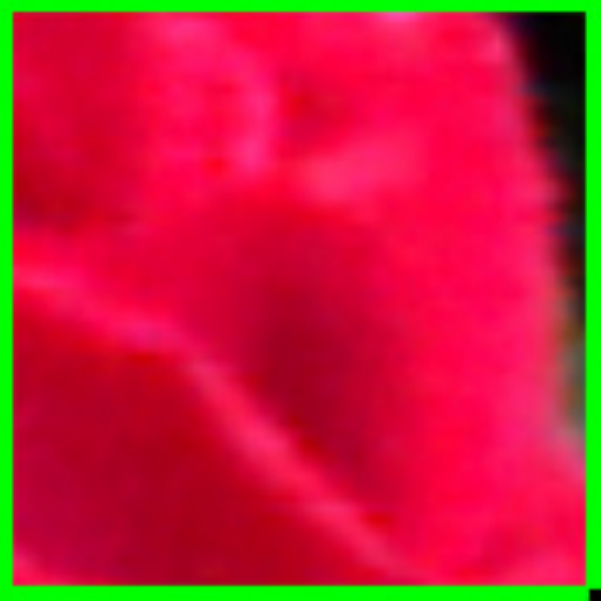} &
            \includegraphics[width=\ssubwidth\linewidth]{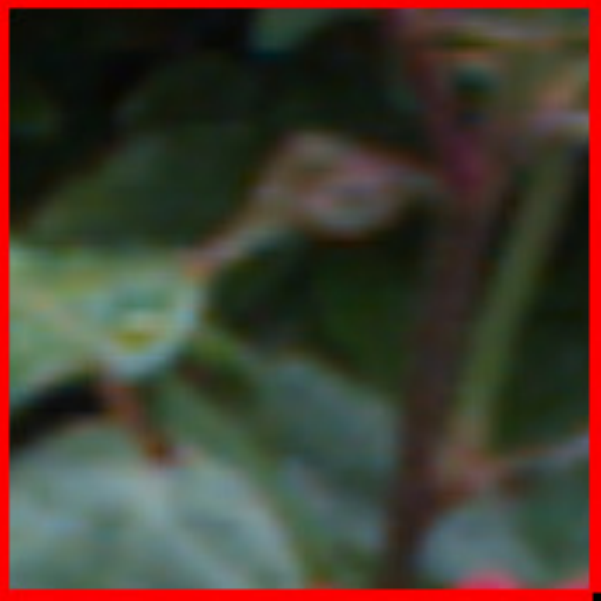} &
            \includegraphics[width=\ssubwidth\linewidth]{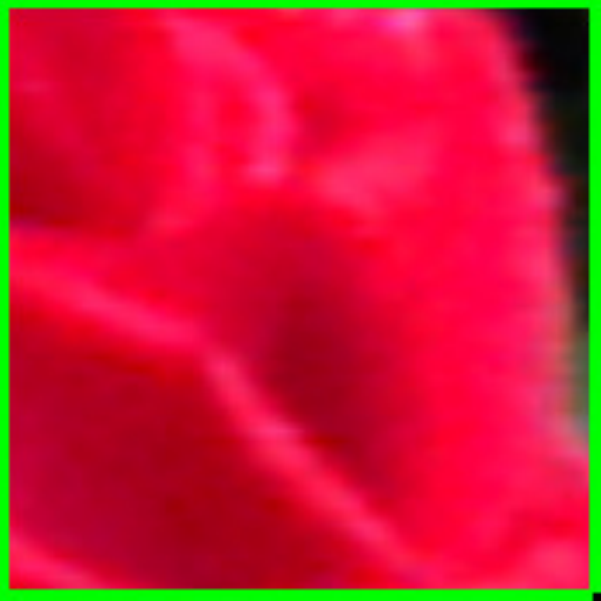} &
            \includegraphics[width=\ssubwidth\linewidth]{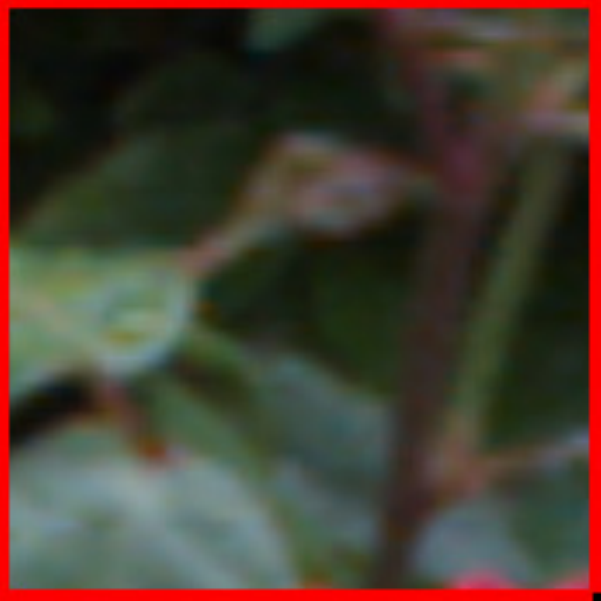} &
            \includegraphics[width=\ssubwidth\linewidth]{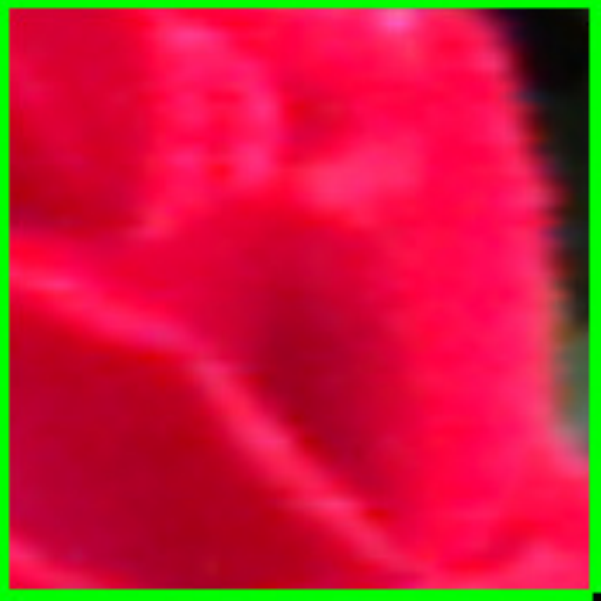}  \\

            \multicolumn{2}{c}{\scriptsize{~\cite{yasarla20}, 21.97/0.839}}&
            \multicolumn{2}{c}{\scriptsize{~\cite{zamir2021}, 27.05/0.914}}&
            \multicolumn{2}{c}{\scriptsize{~\cite{hu2021}, 23.65/0.883}}&
            \multicolumn{2}{c}{\scriptsize{~\cite{ding21}, 30.91/\emp{0.979}}}&
            \multicolumn{2}{c}{\scriptsize{\revised{~\cite{xiao22}},\emp{33.28}/0.964}}&
            \multicolumn{2}{c}{\scriptsize{\revised{~\cite{zamir22}}, 32.41/0.961}}&
            \multicolumn{2}{c}{\scriptsize{Ours, \textbf{33.69}/\textbf{0.980}}}&
            \multicolumn{2}{c}{\scriptsize{GT}}\\
		\end{tabular}
	\end{center}
	\vspace{-0.016\textwidth}
	\caption{Comparison of deraining methods on synthetic LFIs. The de-rained center sub-view generated by each method is evaluated on PSNR/SSIM. The best value is highlighted in \textbf{bold}, and the second-best value is colored in \emp{cyan}.}
	\label{fig:synthetic result2}
\end{figure*}

\renewcommand{\subwidth}{0.116}
\renewcommand{\ssubwidth}{0.058}
\begin{figure*}[t]
	\renewcommand{\tabcolsep}{0.8pt}
	\renewcommand\arraystretch{0.8}
	\begin{center}
		\begin{tabular}{cccccccccccccccc}
            \multicolumn{2}{c}{\includegraphics[width=\subwidth\linewidth]{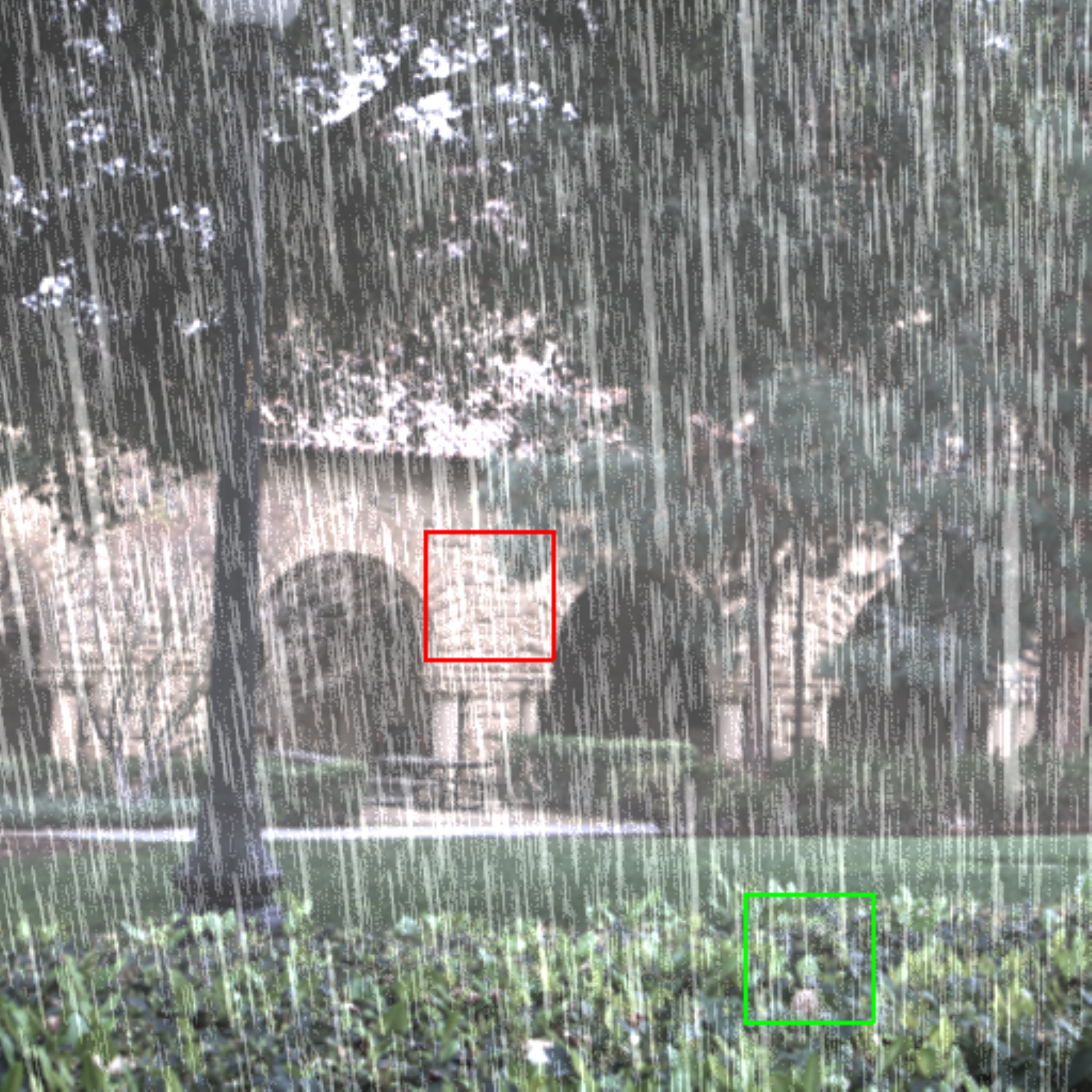}} &
            \multicolumn{2}{c}{\includegraphics[width=\subwidth\linewidth]{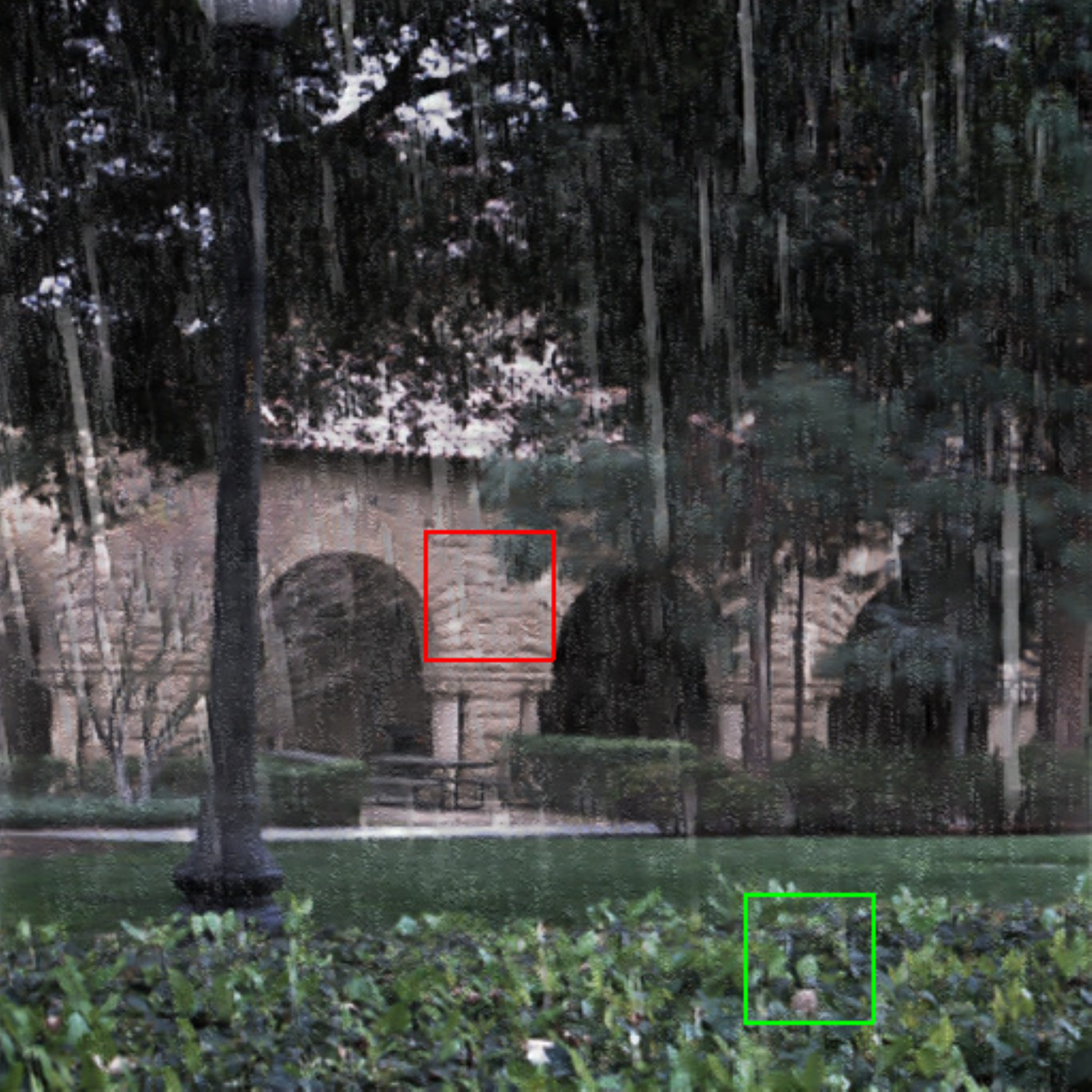}} &
            \multicolumn{2}{c}{\includegraphics[width=\subwidth\linewidth]{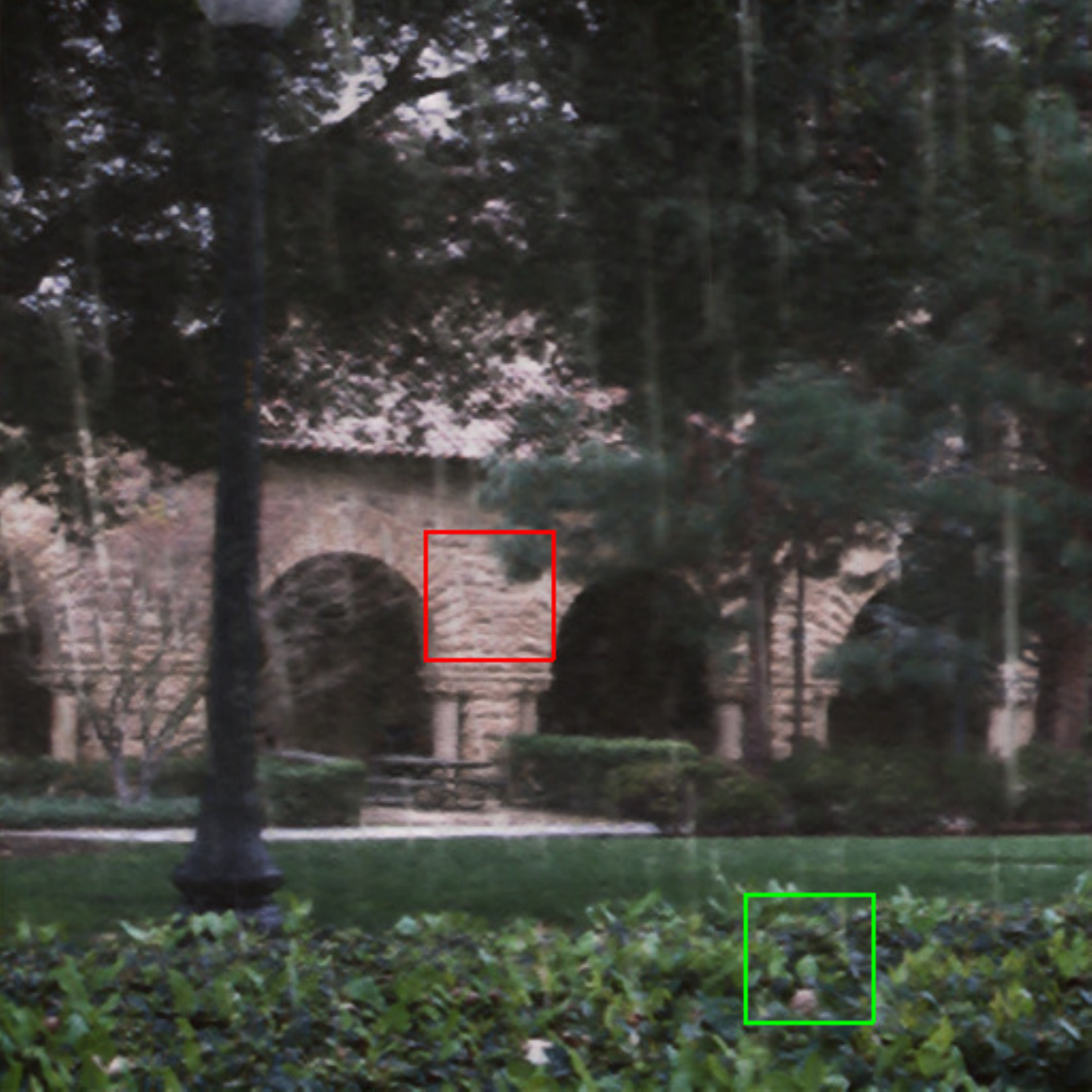}} &
            \multicolumn{2}{c}{\includegraphics[width=\subwidth\linewidth]{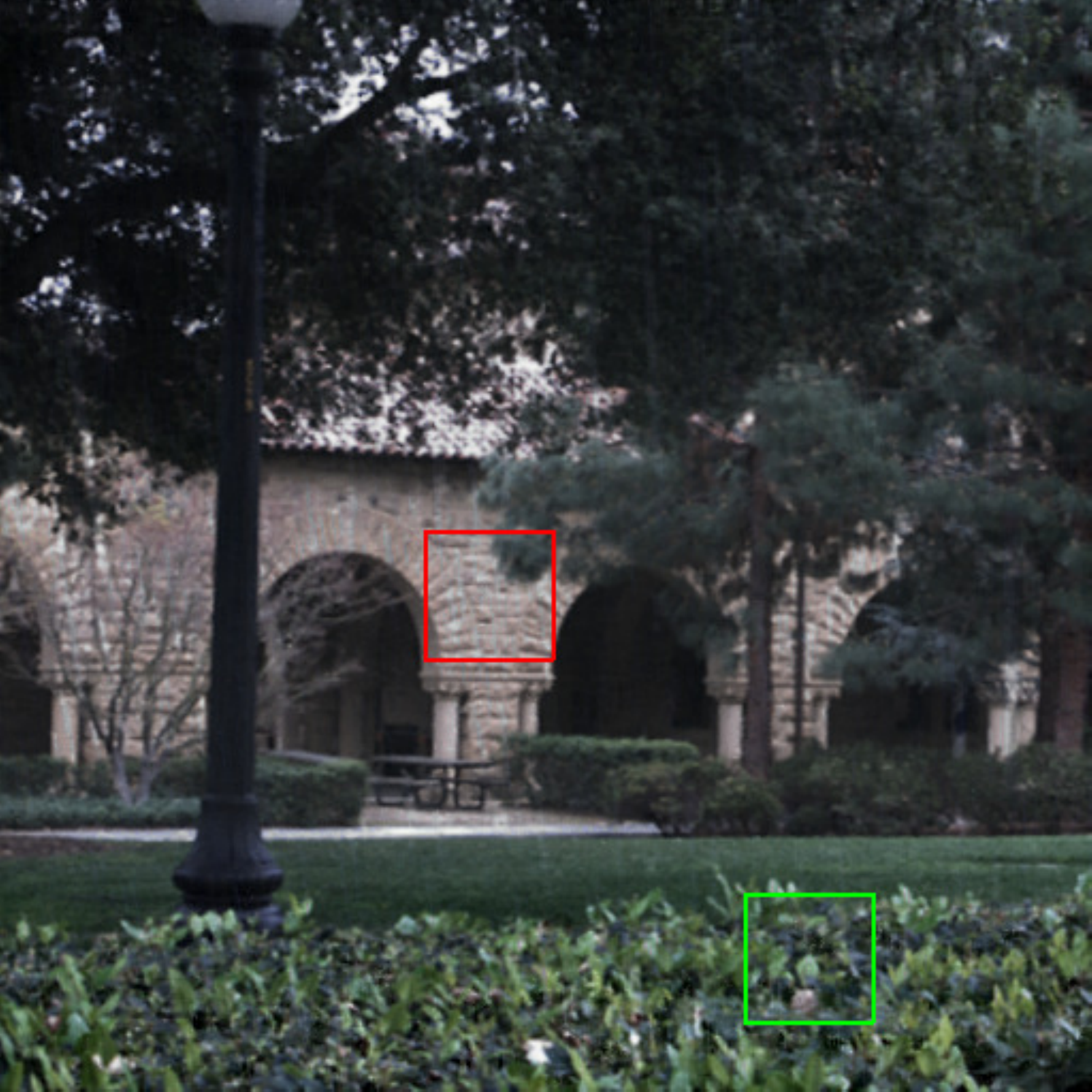}} &
            \multicolumn{2}{c}{\includegraphics[width=\subwidth\linewidth]{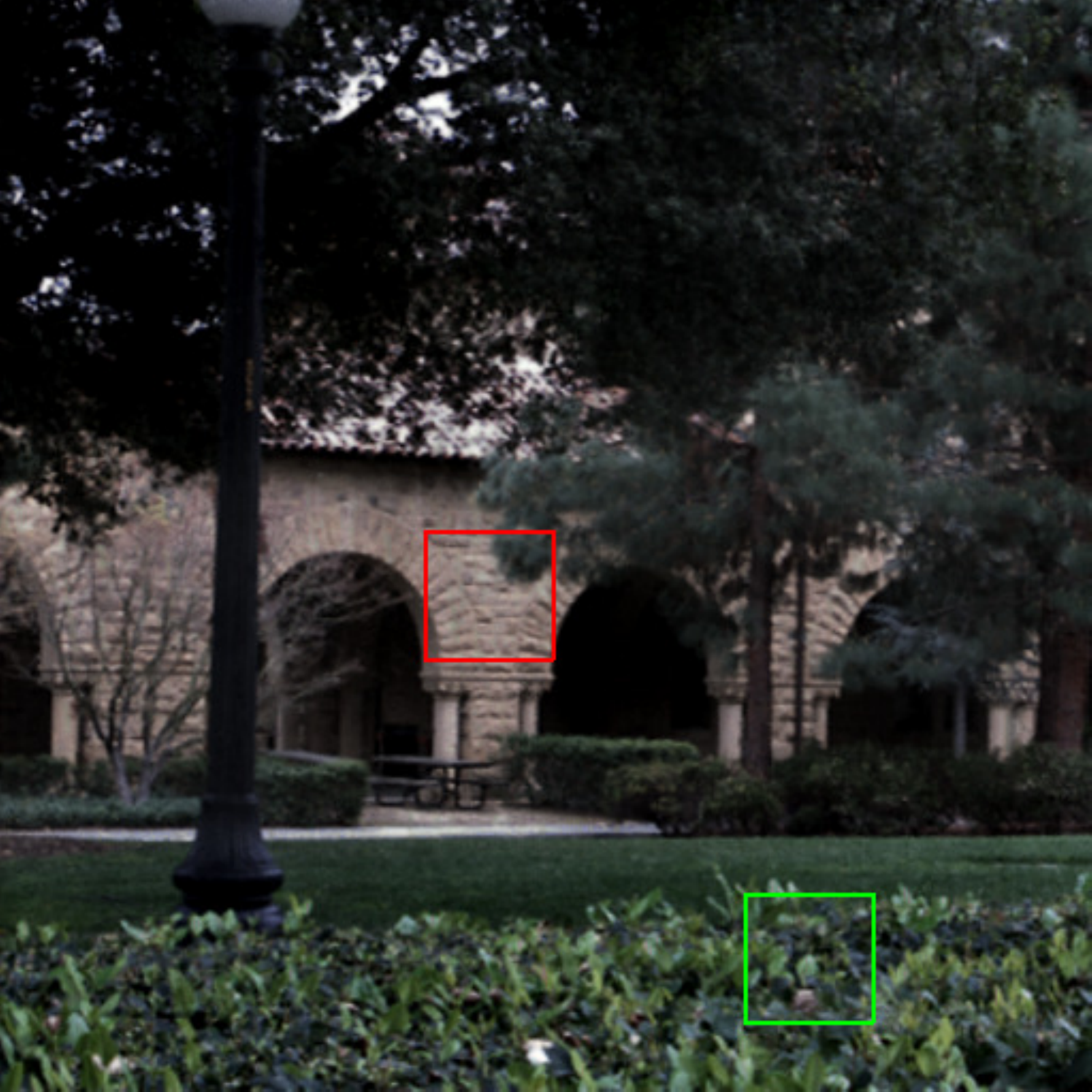}} &
            \multicolumn{2}{c}{\includegraphics[width=\subwidth\linewidth]{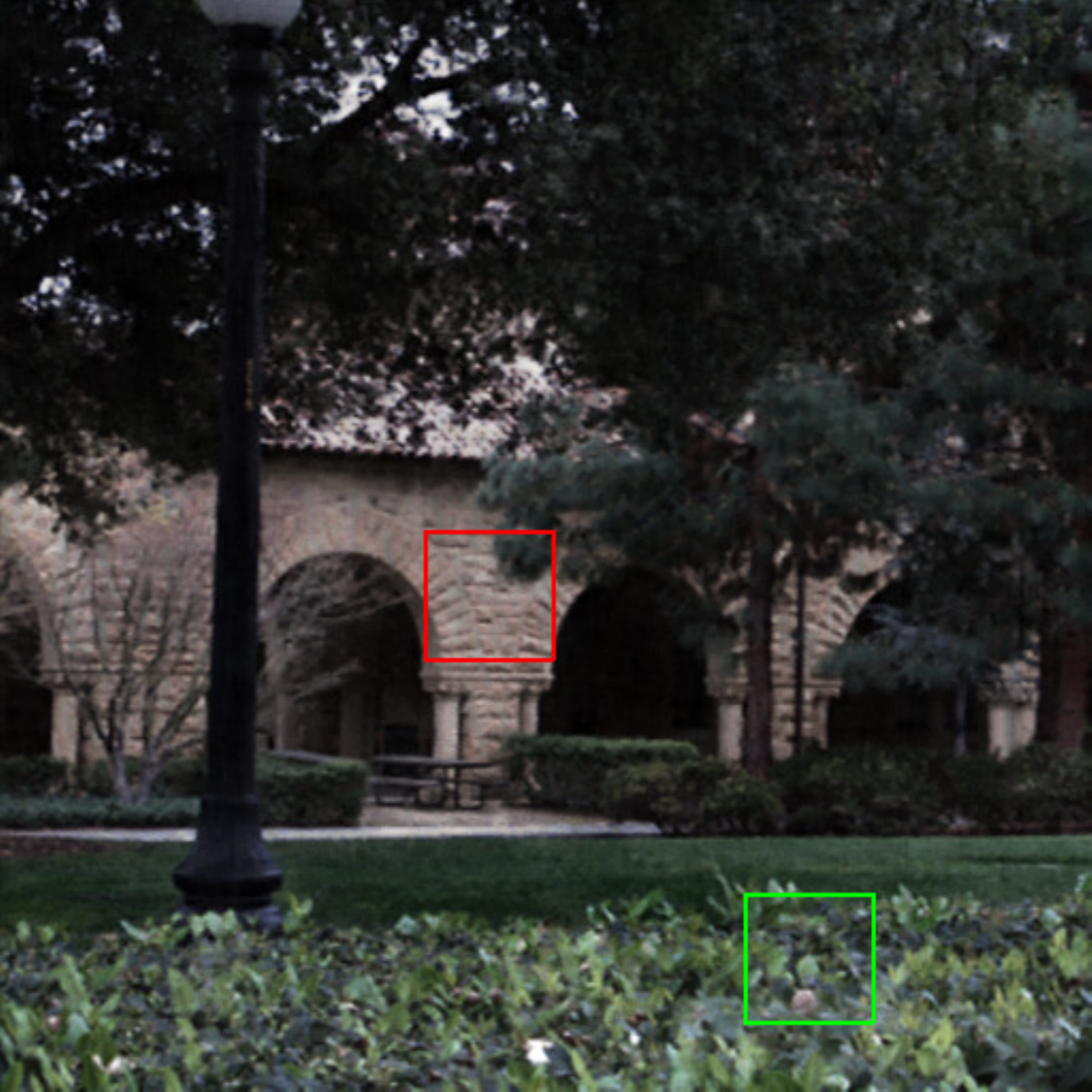}} &
            \multicolumn{2}{c}{\includegraphics[width=\subwidth\linewidth]{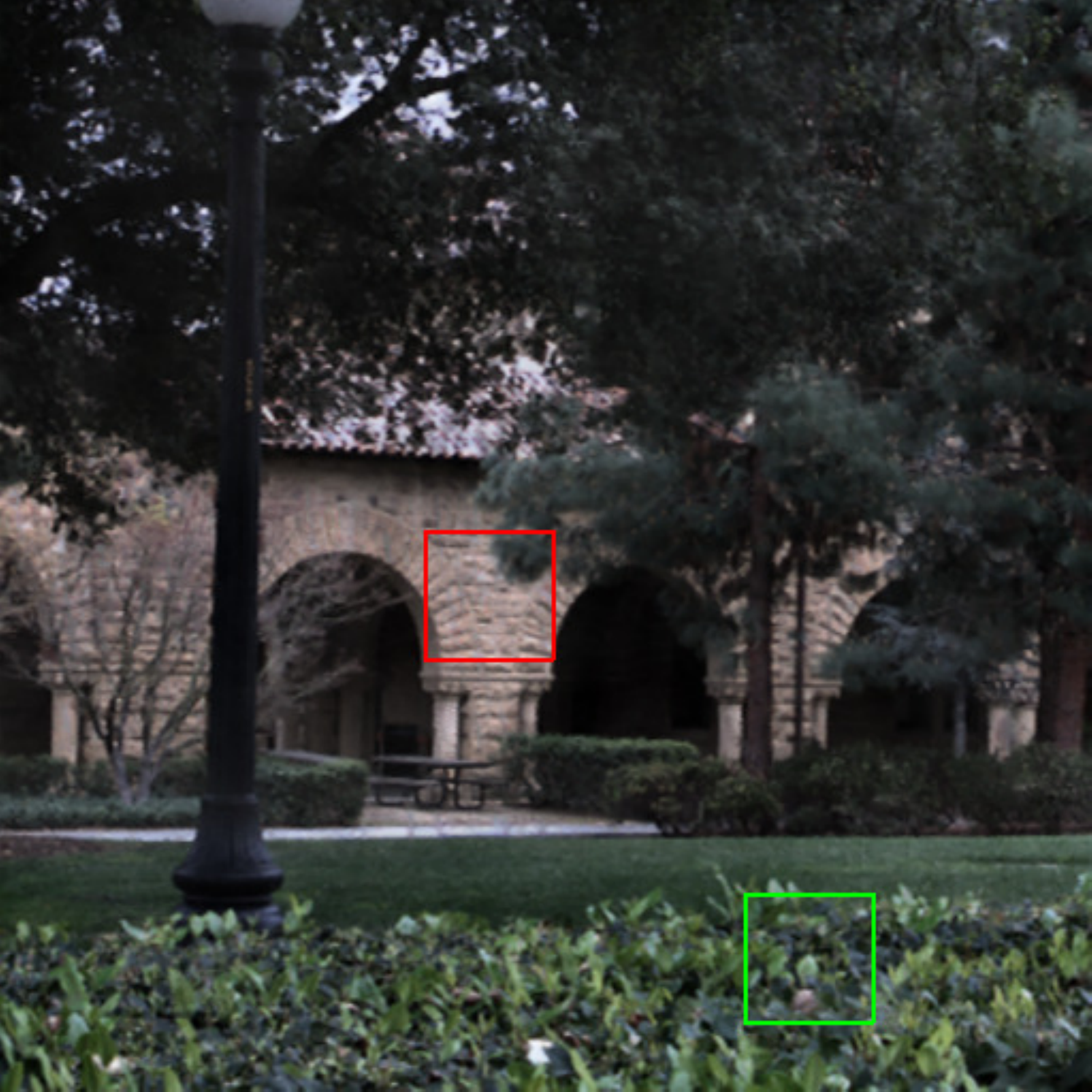}}&
            \multicolumn{2}{c}{\includegraphics[width=\subwidth\linewidth]{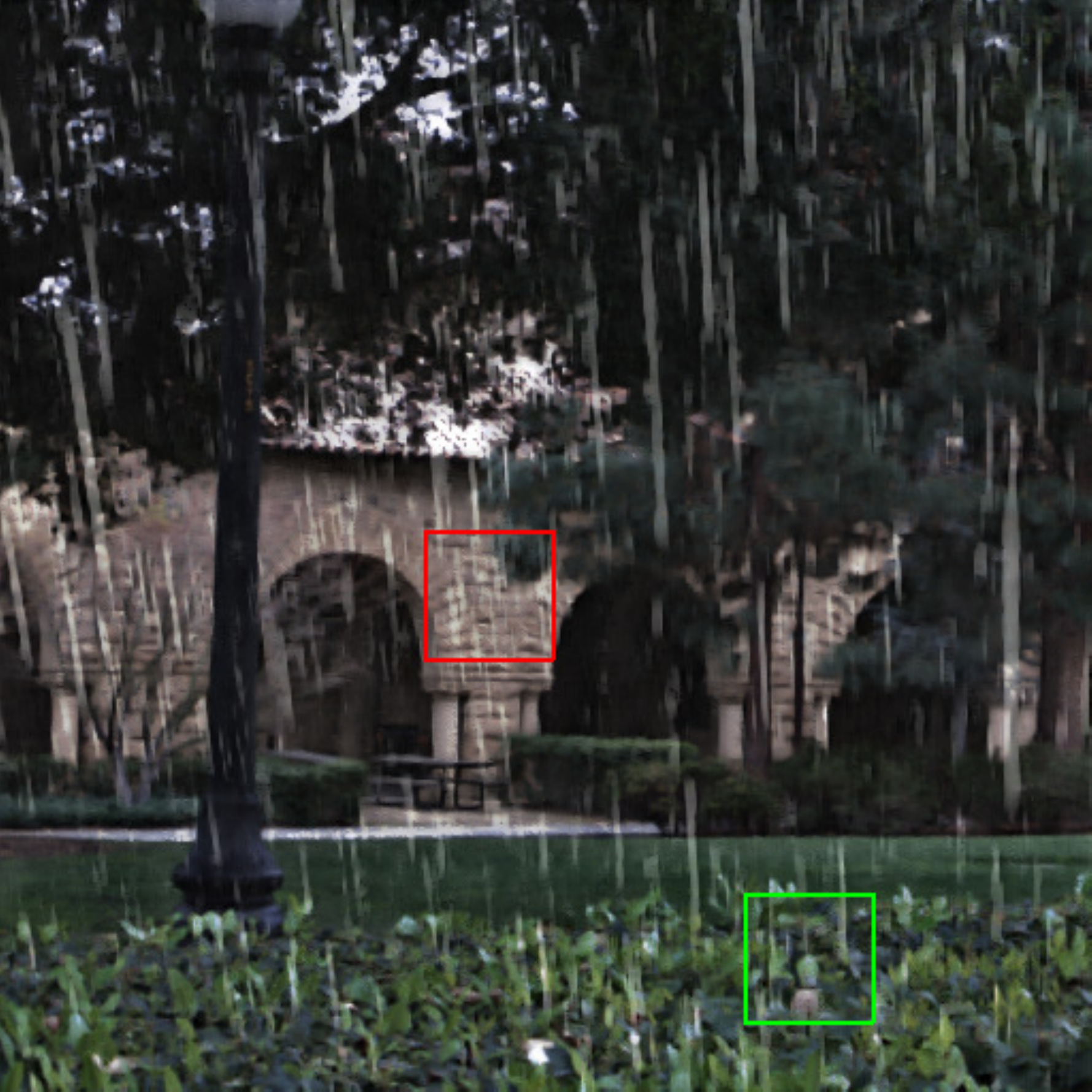}}  \\

            \includegraphics[width=\ssubwidth\linewidth]{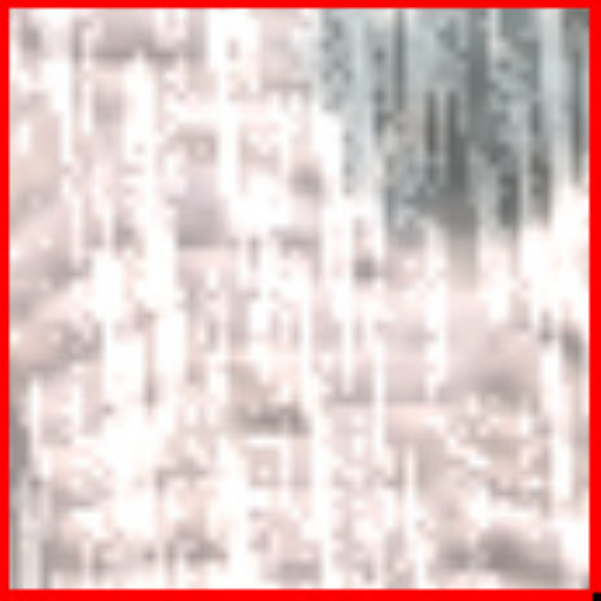} &
            \includegraphics[width=\ssubwidth\linewidth]{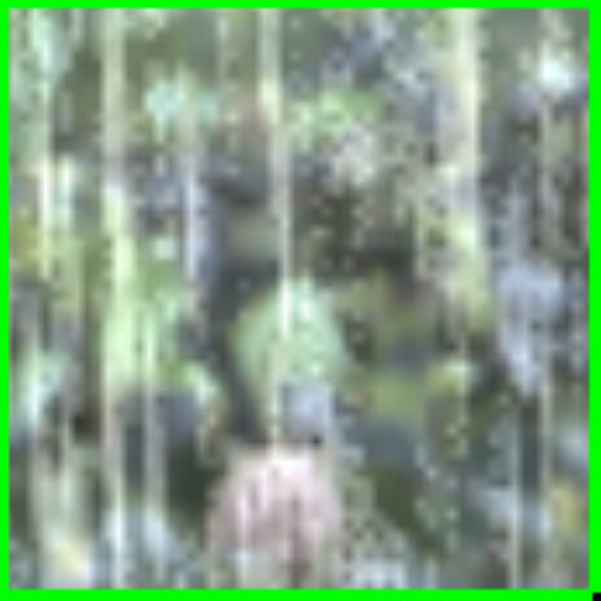} &
            \includegraphics[width=\ssubwidth\linewidth]{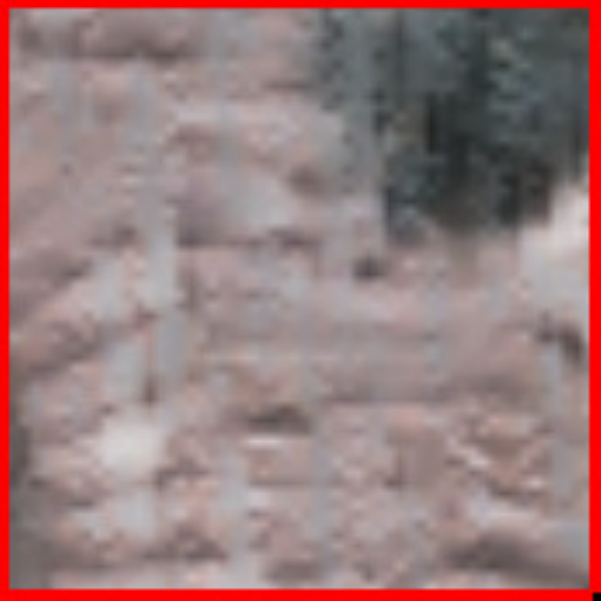} &
            \includegraphics[width=\ssubwidth\linewidth]{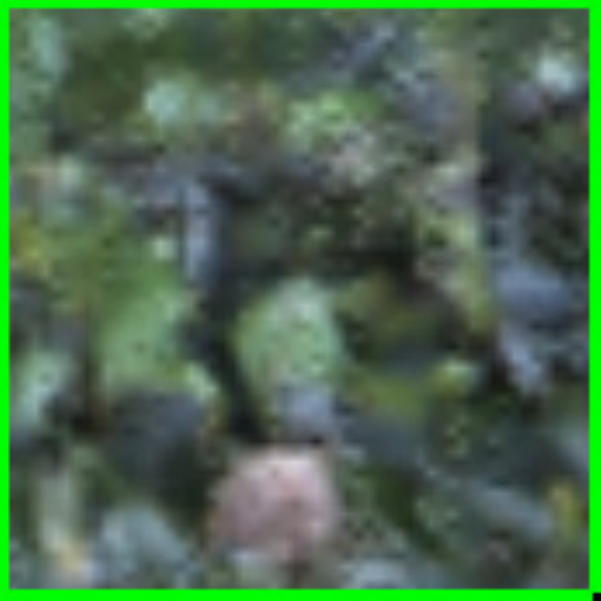} &
            \includegraphics[width=\ssubwidth\linewidth]{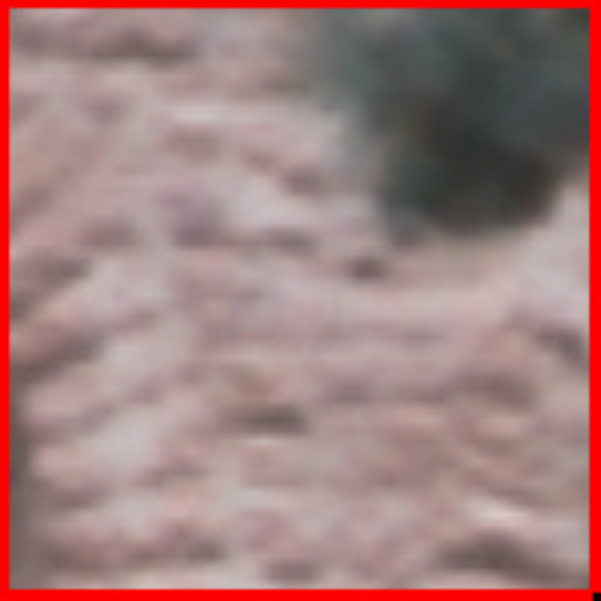} &
            \includegraphics[width=\ssubwidth\linewidth]{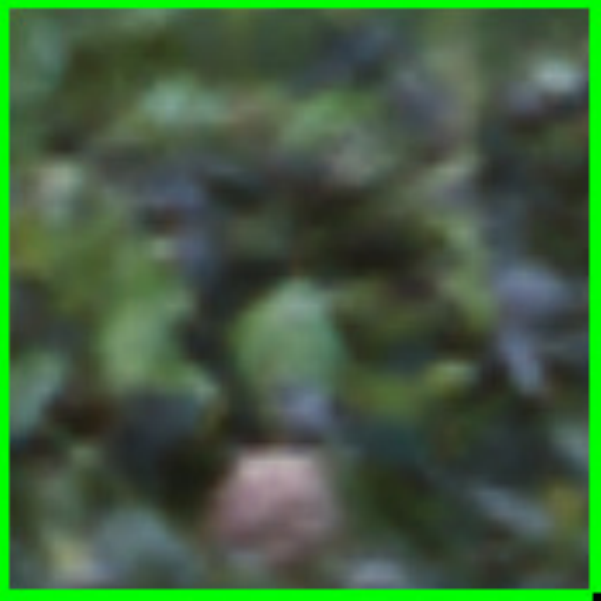} &
            \includegraphics[width=\ssubwidth\linewidth]{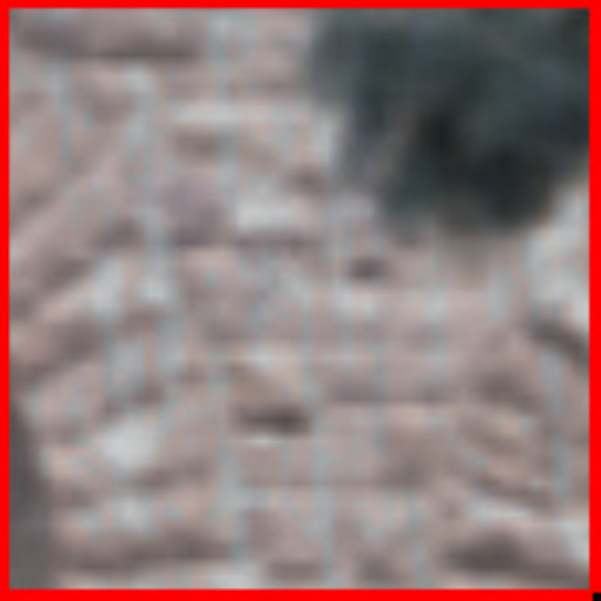} &
            \includegraphics[width=\ssubwidth\linewidth]{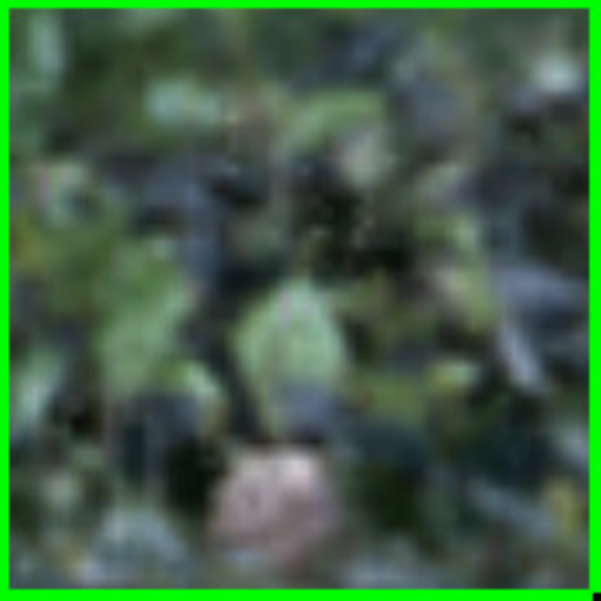} &
            \includegraphics[width=\ssubwidth\linewidth]{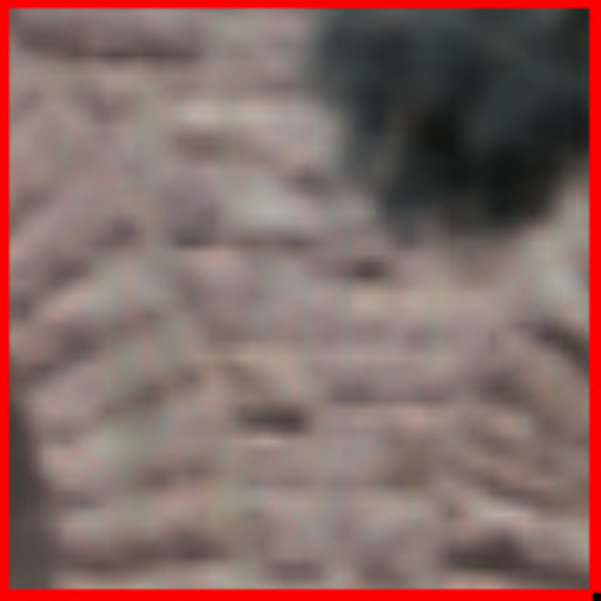} &
            \includegraphics[width=\ssubwidth\linewidth]{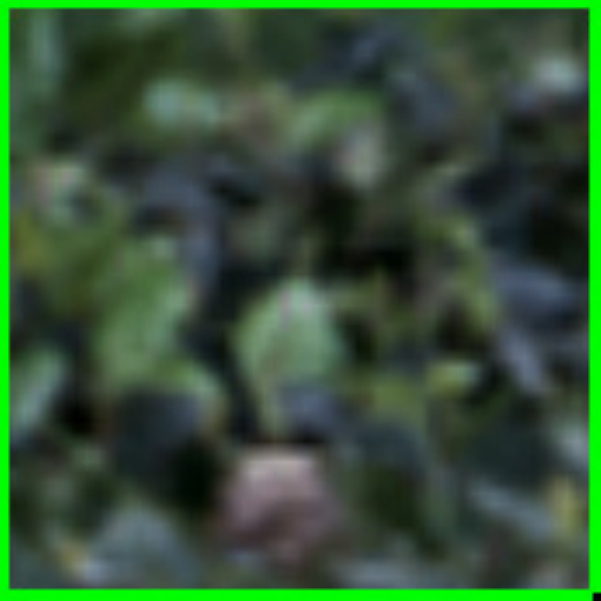} &
            \includegraphics[width=\ssubwidth\linewidth]{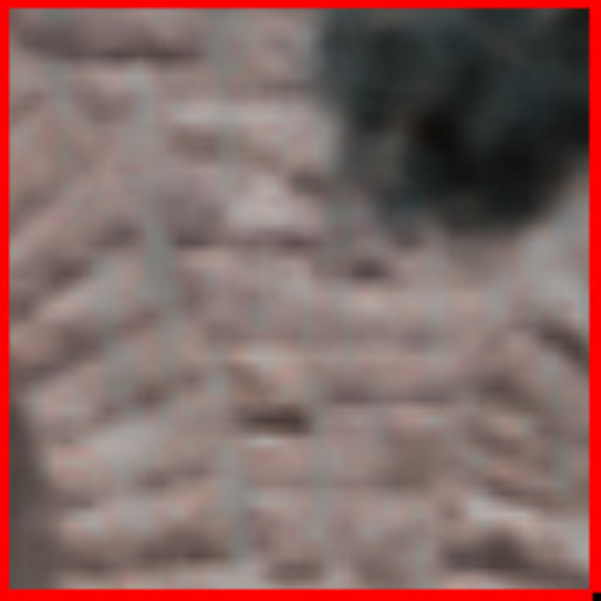} &            \includegraphics[width=\ssubwidth\linewidth]{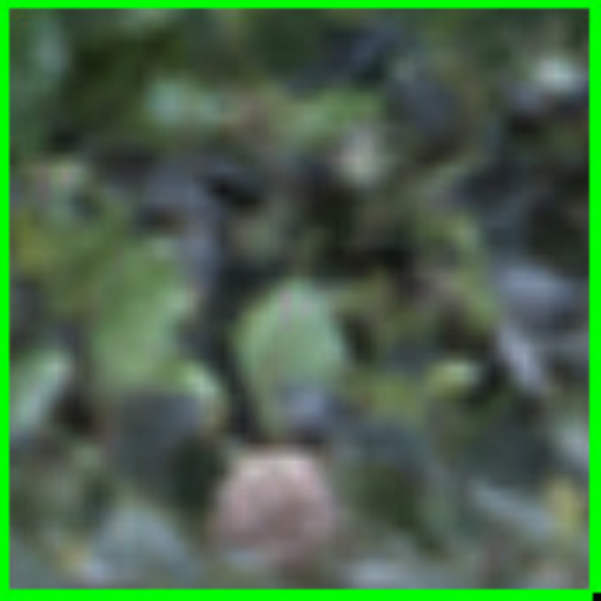} &
            \includegraphics[width=\ssubwidth\linewidth]{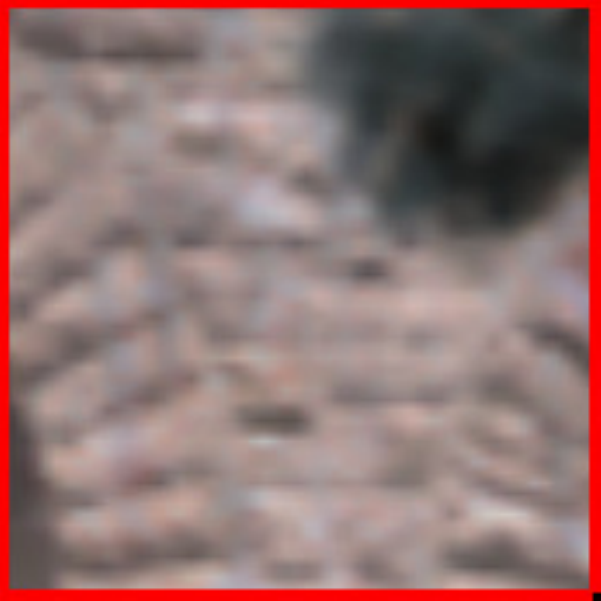} &
            \includegraphics[width=\ssubwidth\linewidth]{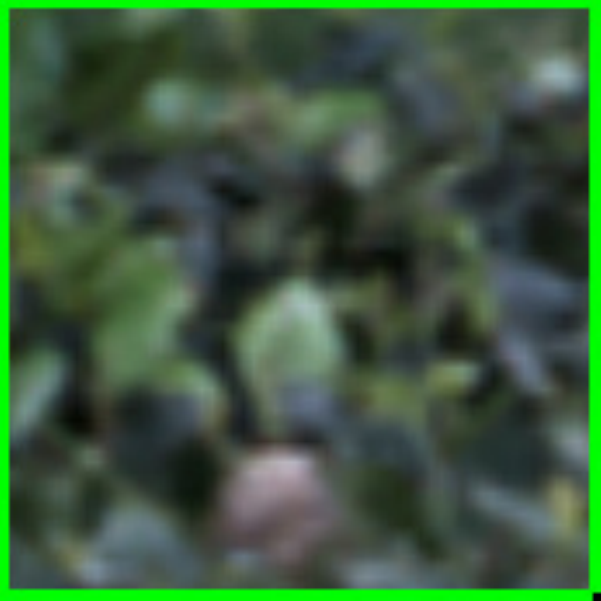}&
            \includegraphics[width=\ssubwidth\linewidth]{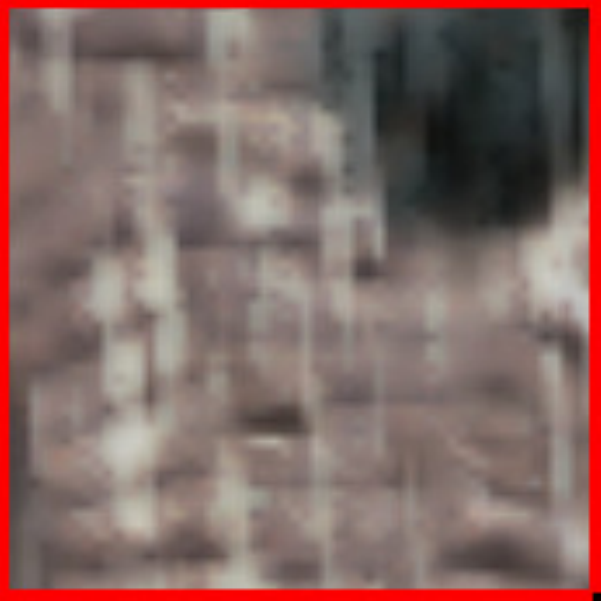} &
            \includegraphics[width=\ssubwidth\linewidth]{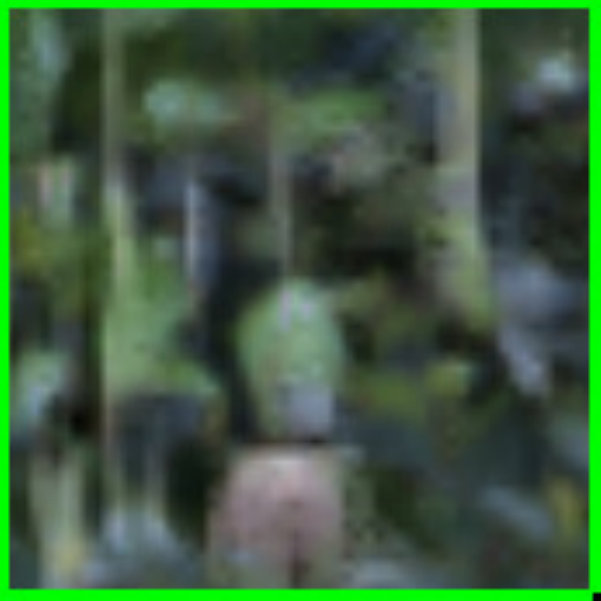}  \\

            \multicolumn{2}{c}{\scriptsize{ Input, 8.86/0.287}}&
            \multicolumn{2}{c}{\scriptsize{~\cite{wang19b}, 20.12/0.611}}&
            \multicolumn{2}{c}{\scriptsize{~\cite{Li19a}, 21.00/0.715}} &
            \multicolumn{2}{c}{\scriptsize{~\cite{wei19}, 24.94/0.874}} &
            \multicolumn{2}{c}{\scriptsize{~\cite{jiang2020}, 24.92/0.822}} &
            \multicolumn{2}{c}{\scriptsize{~\cite{Yang20b}, 27.64/0.915}} &
            \multicolumn{2}{c}{\scriptsize{~\cite{ren20}, 27.70/0.927}}&
            \multicolumn{2}{c}{\scriptsize{~\cite{jiang20}, 21.89/0.700}}\\

            \multicolumn{2}{c}{\includegraphics[width=\subwidth\linewidth]{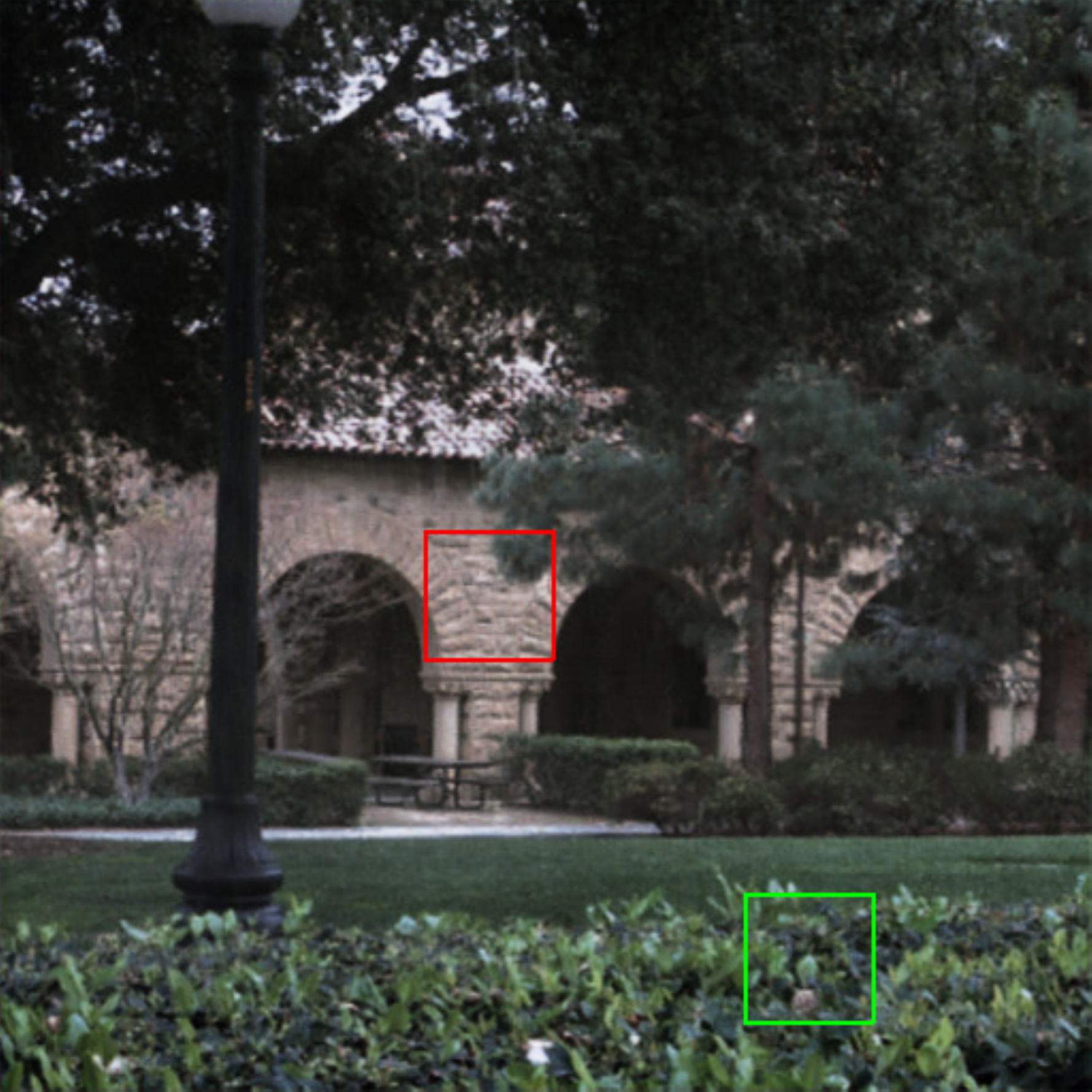}} &
            \multicolumn{2}{c}{\includegraphics[width=\subwidth\linewidth]{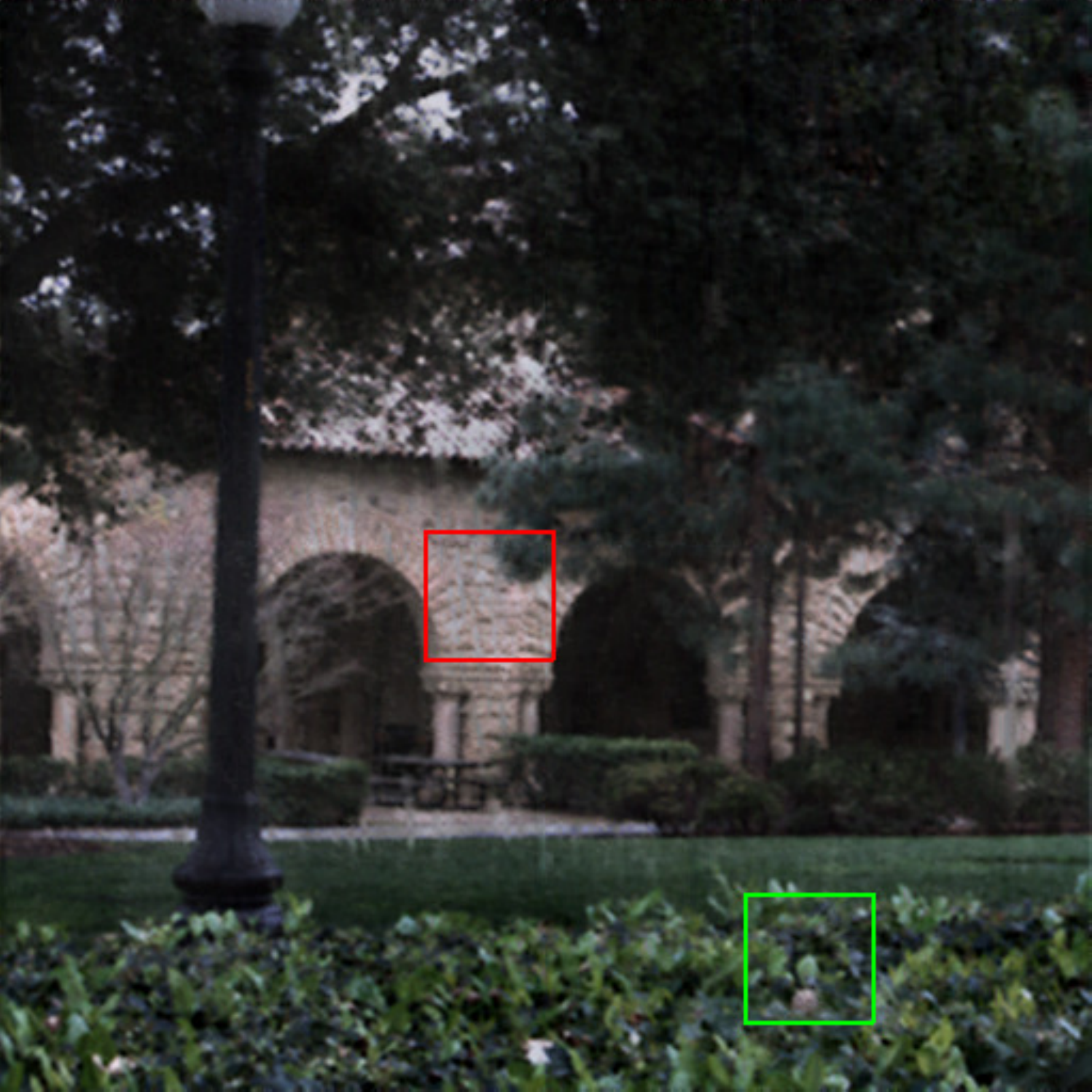}} &
            \multicolumn{2}{c}{\includegraphics[width=\subwidth\linewidth]{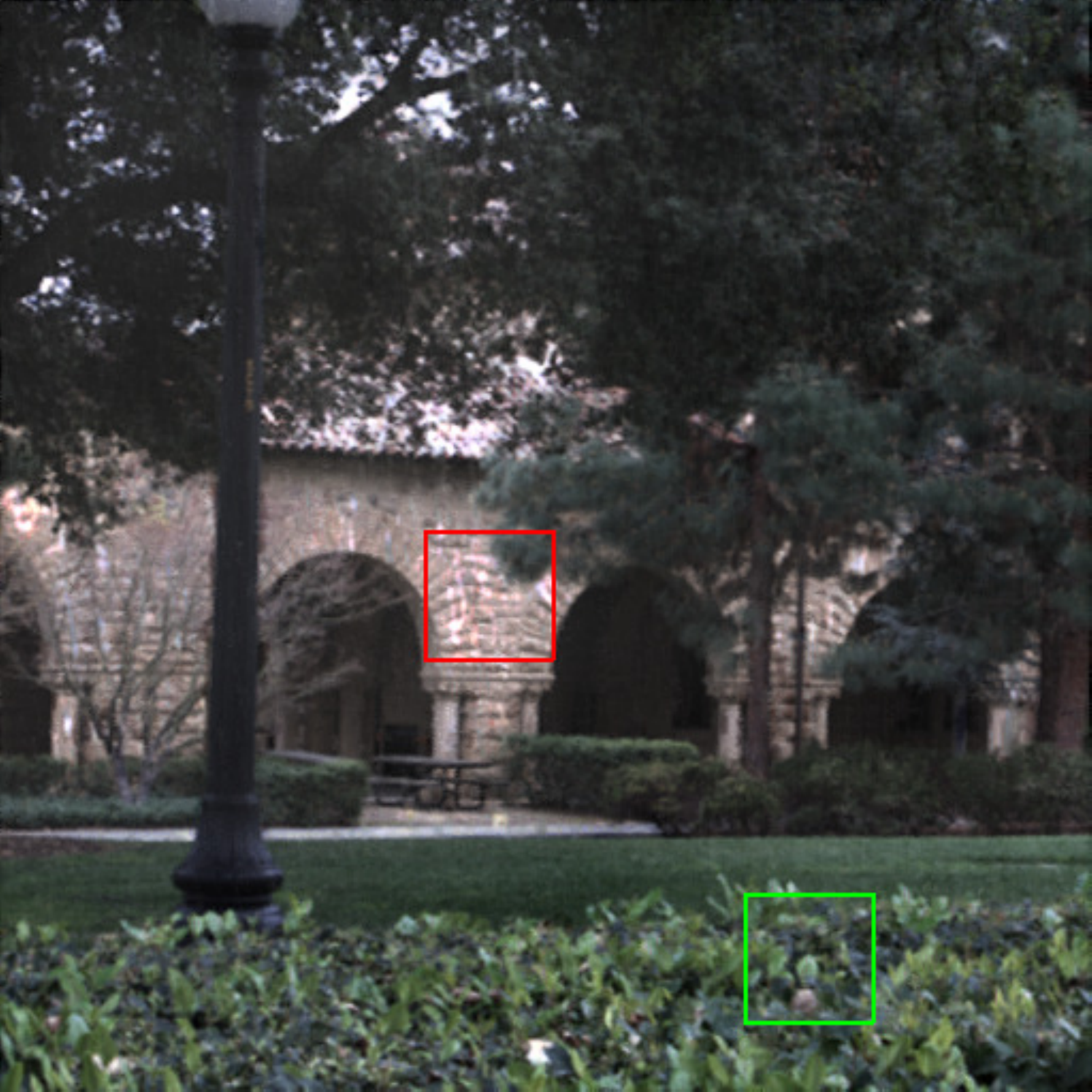}} &
            \multicolumn{2}{c}{\includegraphics[width=\subwidth\linewidth]{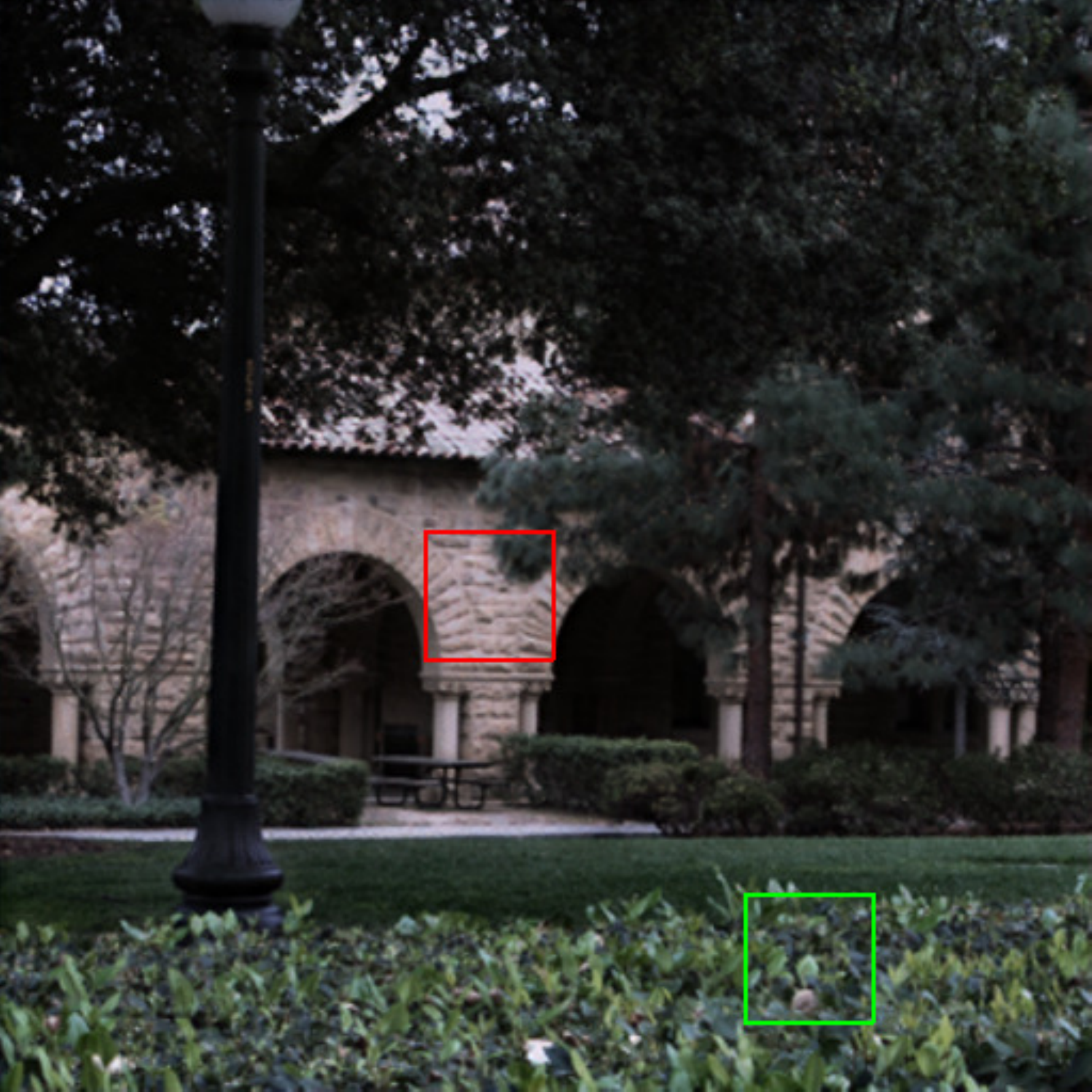}} &
            \multicolumn{2}{c}{\includegraphics[width=\subwidth\linewidth]{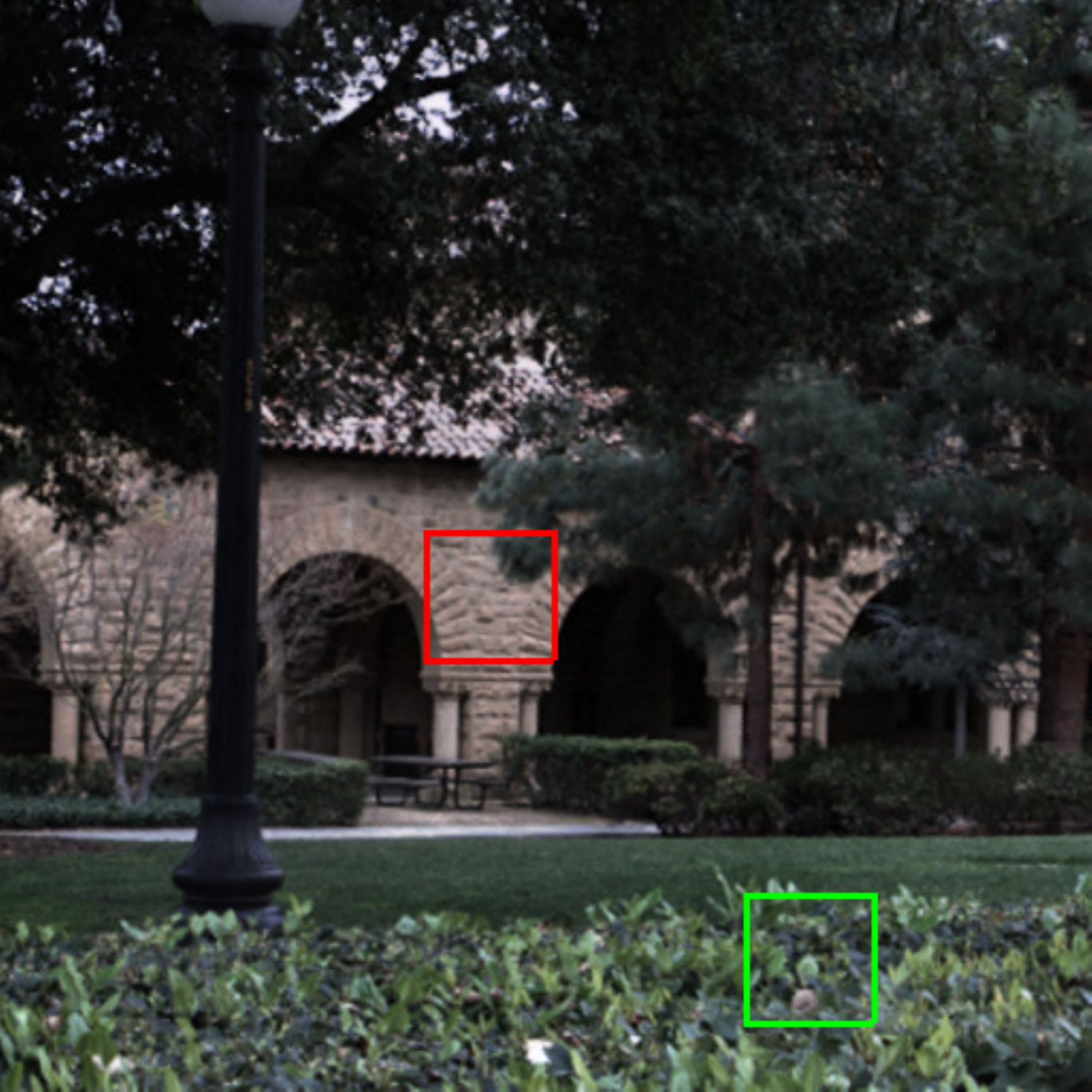}} &
            \multicolumn{2}{c}{\includegraphics[width=\subwidth\linewidth]{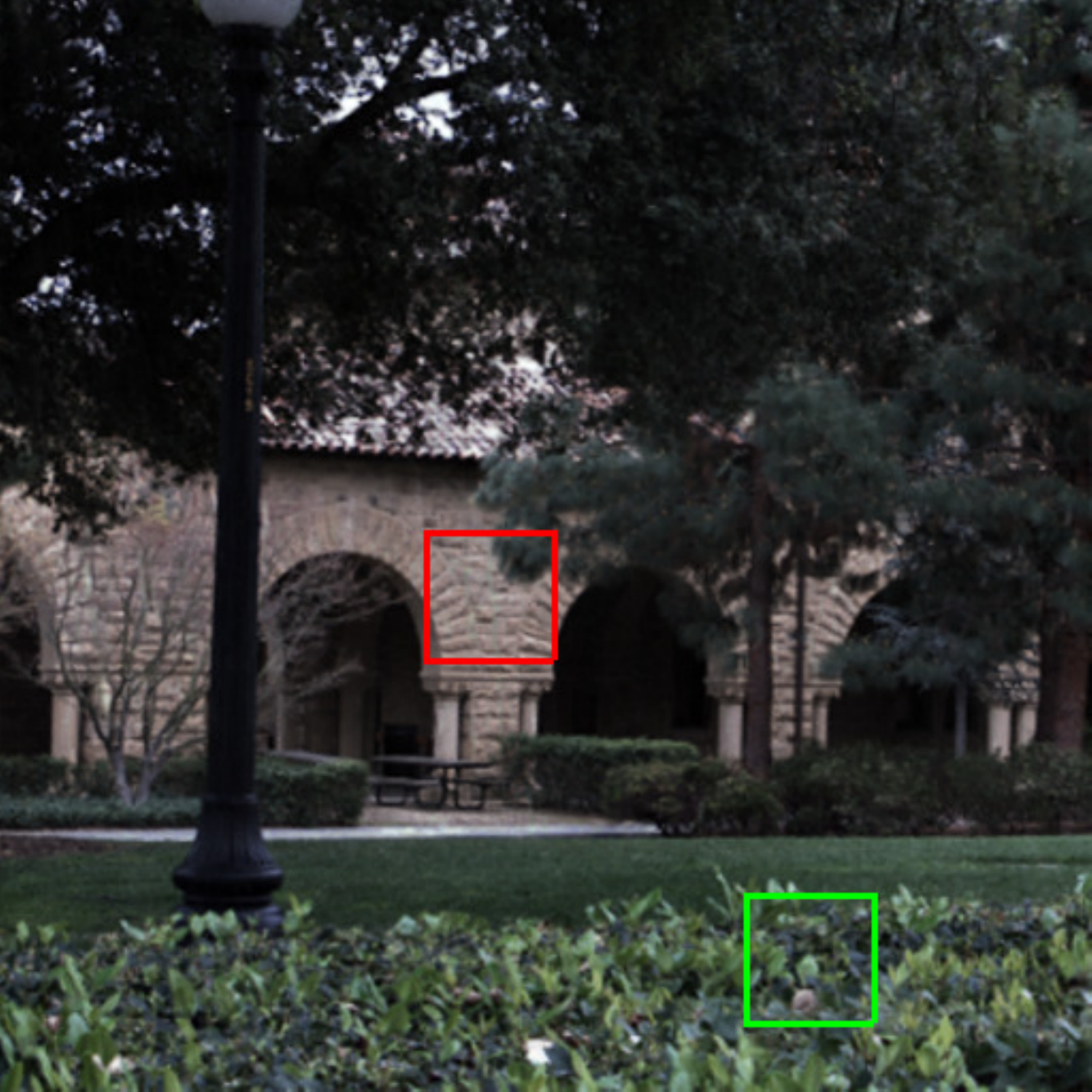}} &
            \multicolumn{2}{c}{\includegraphics[width=\subwidth\linewidth]{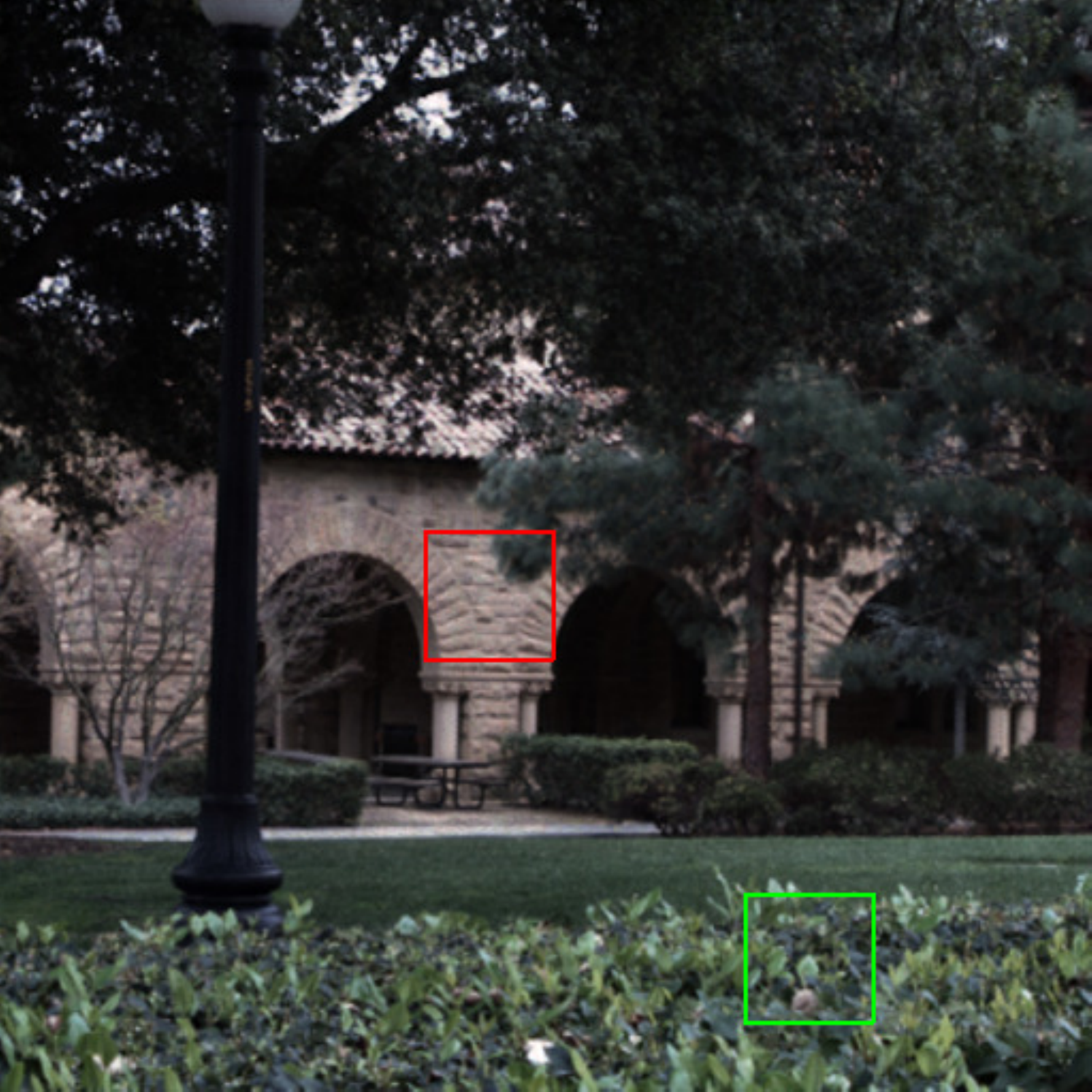}} &
            \multicolumn{2}{c}{\includegraphics[width=\subwidth\linewidth]{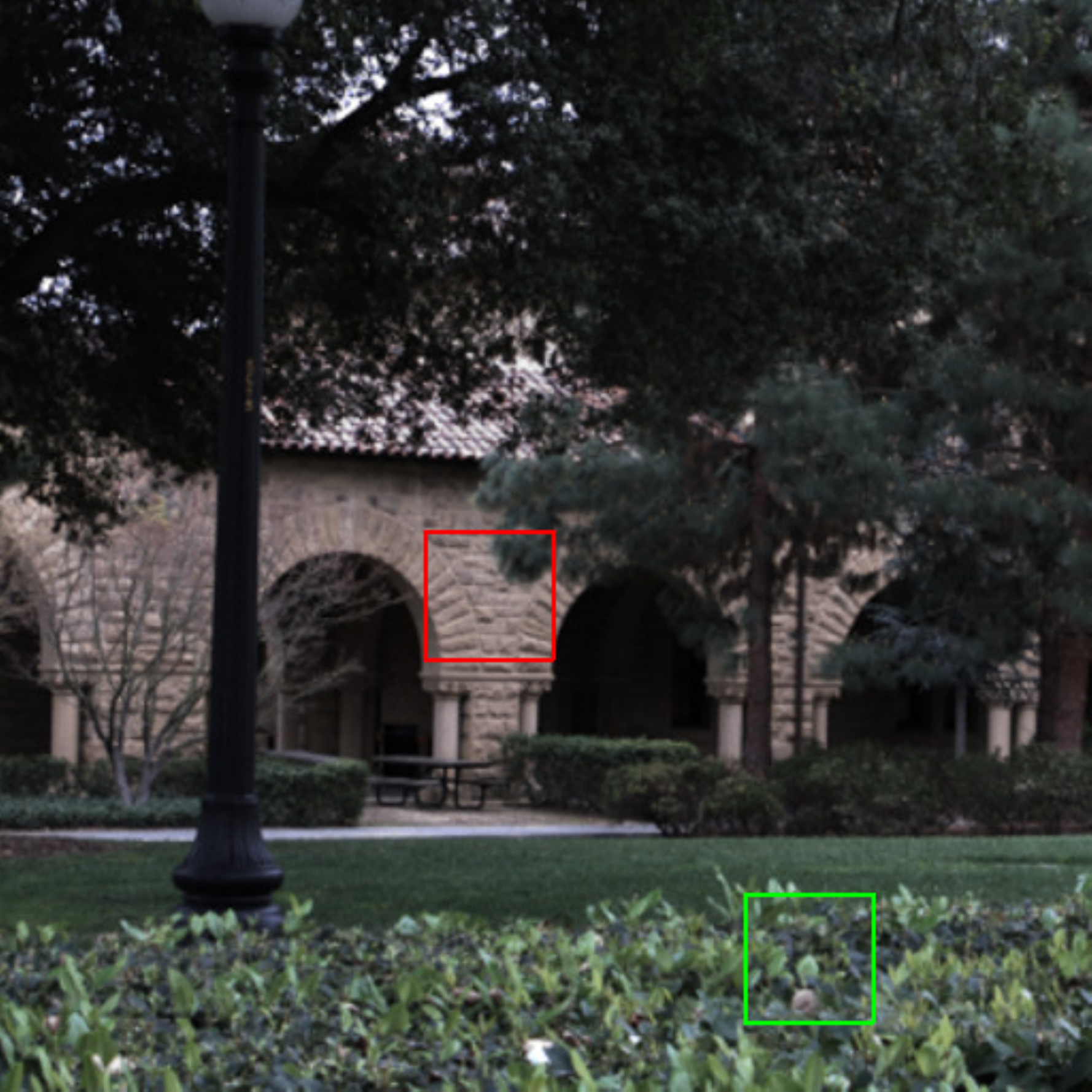}} \\

            \includegraphics[width=\ssubwidth\linewidth]{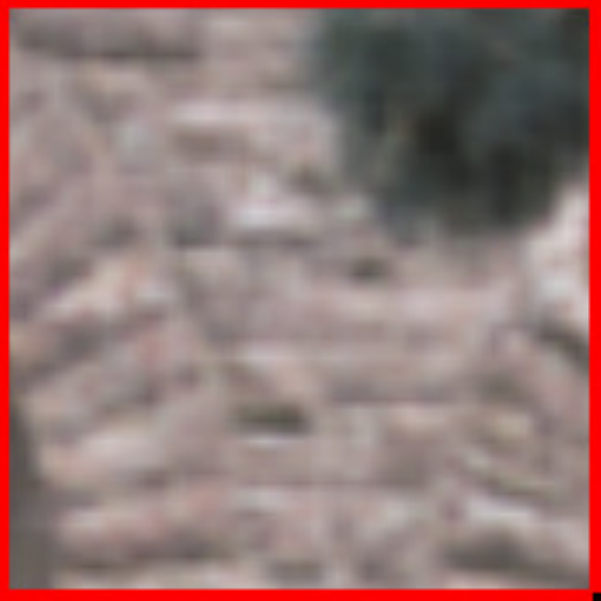} &
            \includegraphics[width=\ssubwidth\linewidth]{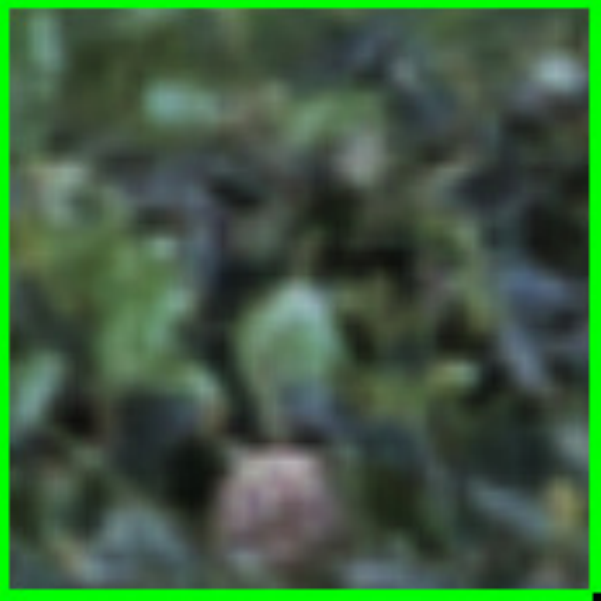} &
            \includegraphics[width=\ssubwidth\linewidth]{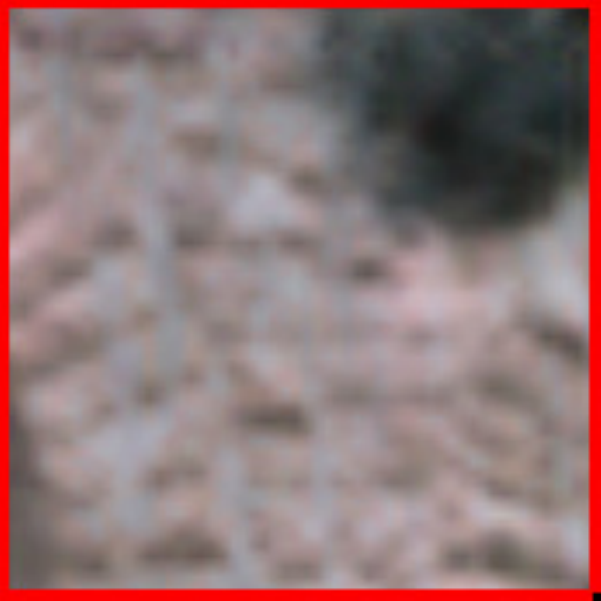} &
            \includegraphics[width=\ssubwidth\linewidth]{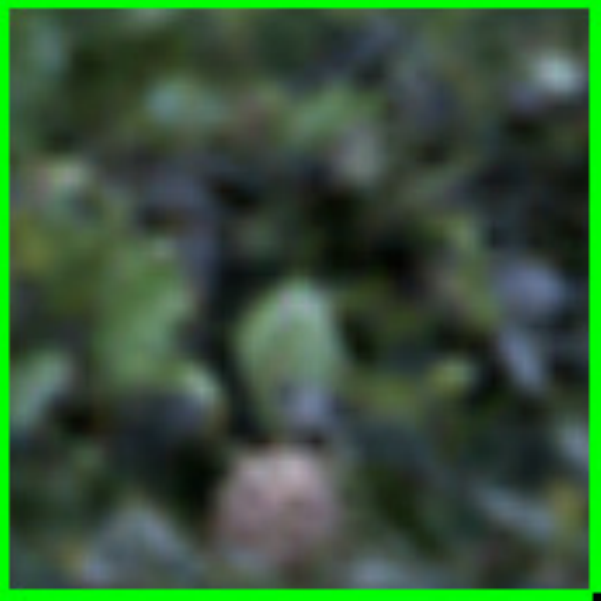} &
            \includegraphics[width=\ssubwidth\linewidth]{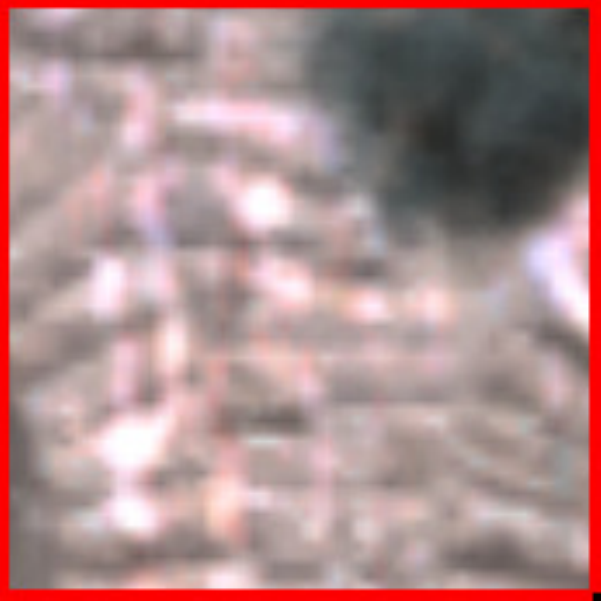} &
            \includegraphics[width=\ssubwidth\linewidth]{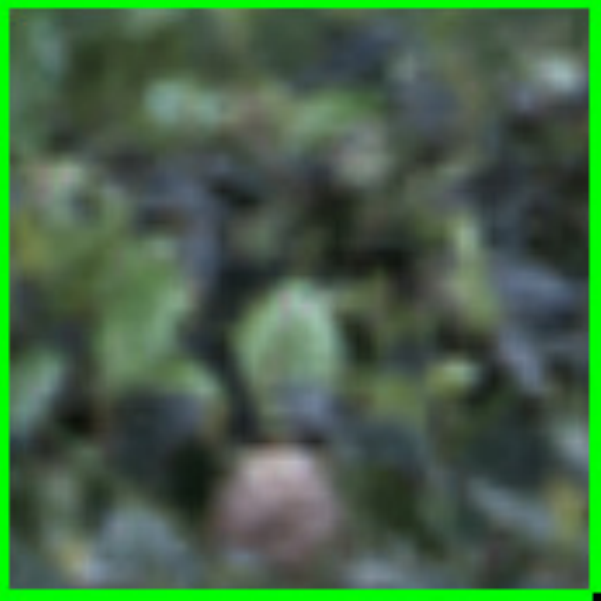} &
            \includegraphics[width=\ssubwidth\linewidth]{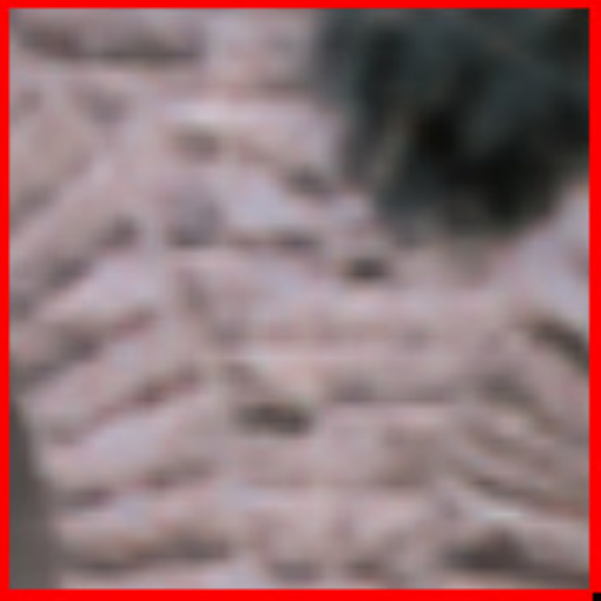} &
            \includegraphics[width=\ssubwidth\linewidth]{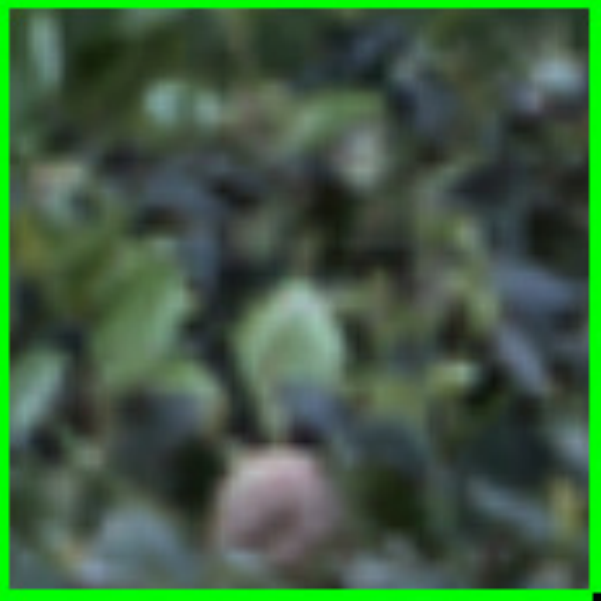} &
            \includegraphics[width=\ssubwidth\linewidth]{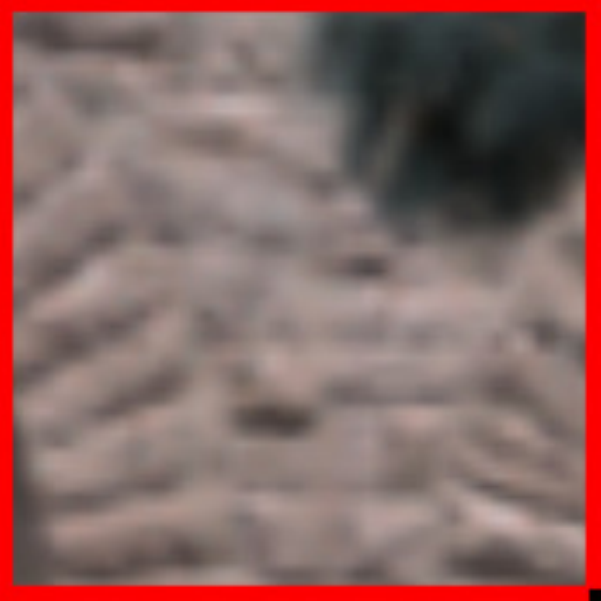} &
            \includegraphics[width=\ssubwidth\linewidth]{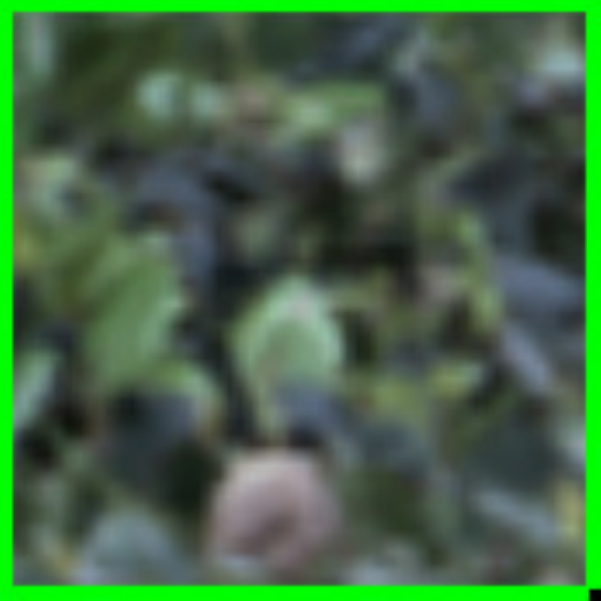} &
            \includegraphics[width=\ssubwidth\linewidth]{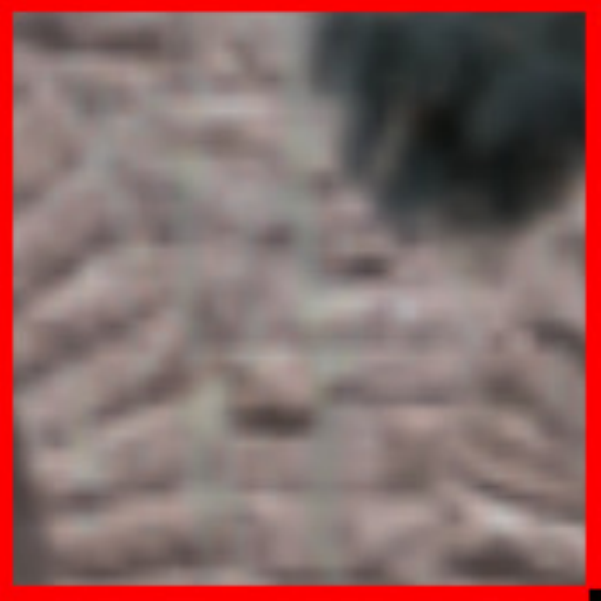} &
            \includegraphics[width=\ssubwidth\linewidth]{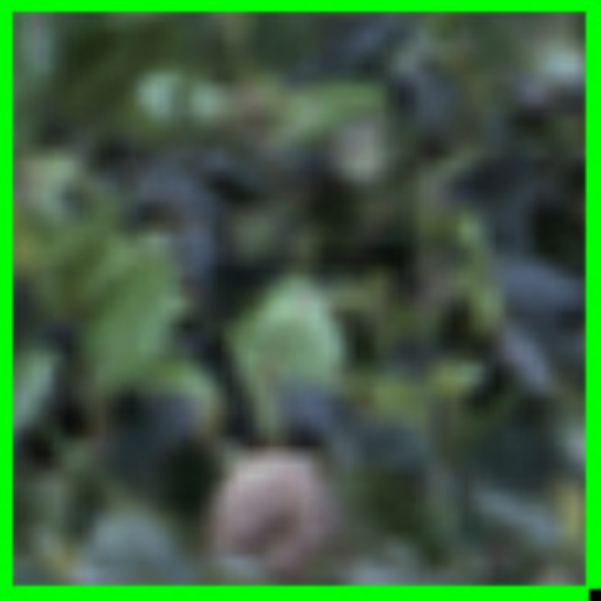} &
            \includegraphics[width=\ssubwidth\linewidth]{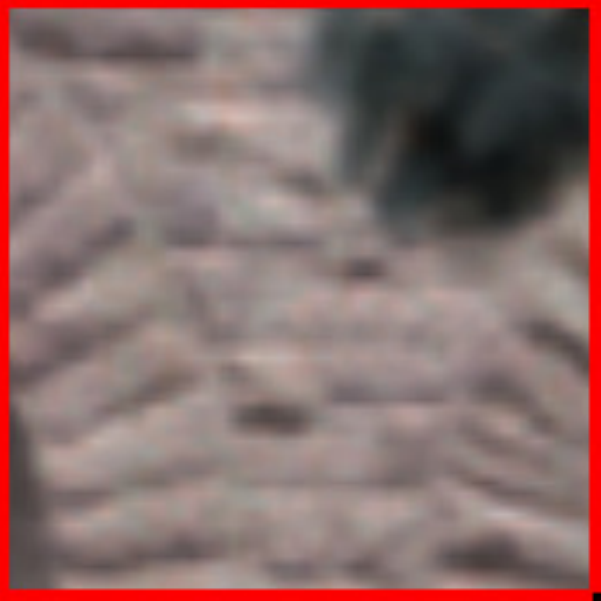} &
            \includegraphics[width=\ssubwidth\linewidth]{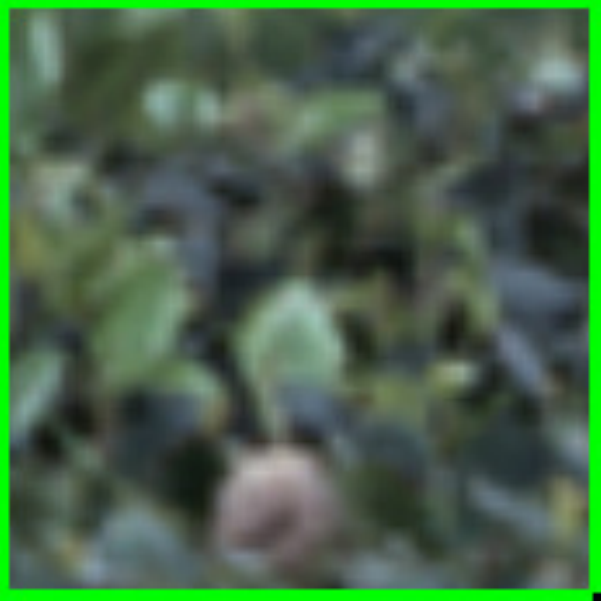} &
            \includegraphics[width=\ssubwidth\linewidth]{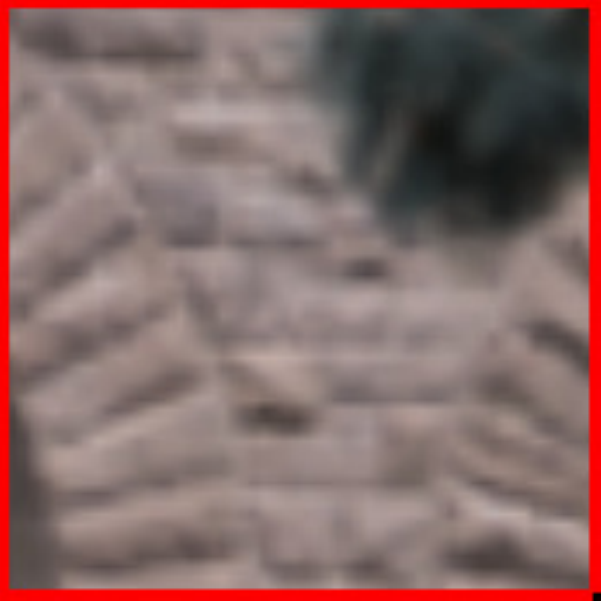} &
            \includegraphics[width=\ssubwidth\linewidth]{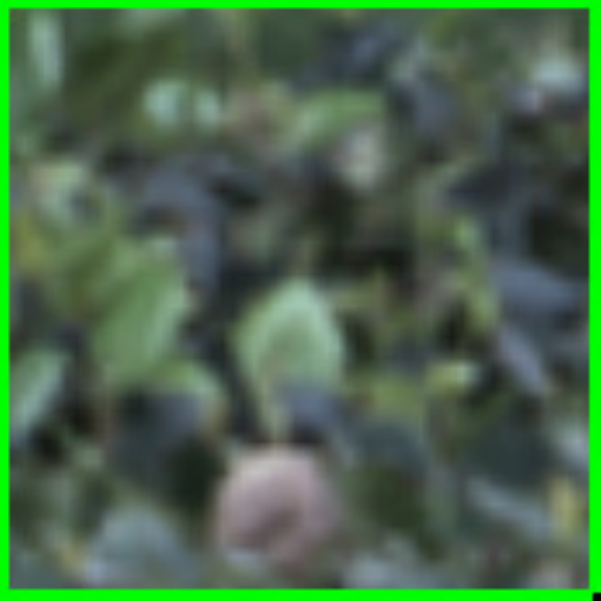}  \\

            \multicolumn{2}{c}{\scriptsize{~\cite{yasarla20},  26.07/0.905}}&
            \multicolumn{2}{c}{\scriptsize{~\cite{zamir2021}, 22.89/0.812}}&
            \multicolumn{2}{c}{\scriptsize{~\cite{hu2021}, 22.20/0.823}}&
            \multicolumn{2}{c}{\scriptsize{~\cite{ding21}, 28.89/\emp{0.961}}}&
            \multicolumn{2}{c}{\scriptsize{\revised{~\cite{xiao22}},\textbf{32.46}/0.960}}&
            \multicolumn{2}{c}{\scriptsize{\revised{~\cite{zamir22}},30.35/0.956}}&
            \multicolumn{2}{c}{\scriptsize{Ours, \emp{30.41}/\textbf{0.975}}}&
            \multicolumn{2}{c}{\scriptsize{GT}}\\
		\end{tabular}
	\end{center}
	\vspace{-0.016\textwidth}
	\caption{Comparison of deraining methods on synthetic LFIs. The de-rained center sub-view generated by each method is evaluated on PSNR/SSIM. The best value is highlighted in \textbf{bold}, and the second-best value is colored in \emp{cyan}.}
	\label{fig:synthetic result8}
	\vspace{-0.016\textwidth}
\end{figure*}

\setlength\tabcolsep{1.0pt}
\begin{table*}[h]
	\centering
	\caption{Mean PSNR/SSIM comparison of deraining methods evaluated on the testing set of the synthetic rainy LFIs. The last two columns show the results obtained by our network with the global discriminator and global-local discriminator, respectively.}
	\label{tab:result}
	\vspace{-0.01\textwidth}
	\begin{tabular*}{\linewidth}{c|ccccccccccccccc}
		\hline
		Methods & Wang~\cite{wang19b}&Li~\cite{Li19a}&Wei~\cite{wei19}&Jiang~\cite{jiang2020}&Yang~\cite{Yang20b} &Ren~\cite{ren20}&Jiang~\cite{jiang20}&Yasarla~\cite{yasarla20}
&Zamir~\cite{zamir2021}&Hu~\cite{hu2021}&Ding~\cite{ding21}& \revised{Xiao~\cite{xiao22}} & \revised{Zamir~\cite{zamir22}} &Ours-GD & Ours\\
		\hline
		PSNR &24.59 & 26.63 &  24.53& 26.40 &  27.93  & 28.59 &21.38 &  24.69&  24.27 & 28.18 &28.46& \textbf{31.52}  & \emp{30.50}  & 29.54  &  29.89\\
		SSIM&0.826 & 0.935 &  0.821& 0.904 &  0.901  & 0.938 & 0.734 &  0.860& 0.848 & 0.905 & 0.940& \emp{0.956}  & 0.945  & 0.951  &  \textbf{0.959}\\
		\hline
	\end{tabular*}
\end{table*}

\renewcommand{\subwidth}{0.16}
\begin{figure*}[t]
	\renewcommand{\tabcolsep}{1.0pt}
	\renewcommand\arraystretch{0.6}
	\begin{center}
		\begin{tabular}{cccccc}

			\includegraphics[width=\subwidth\linewidth]{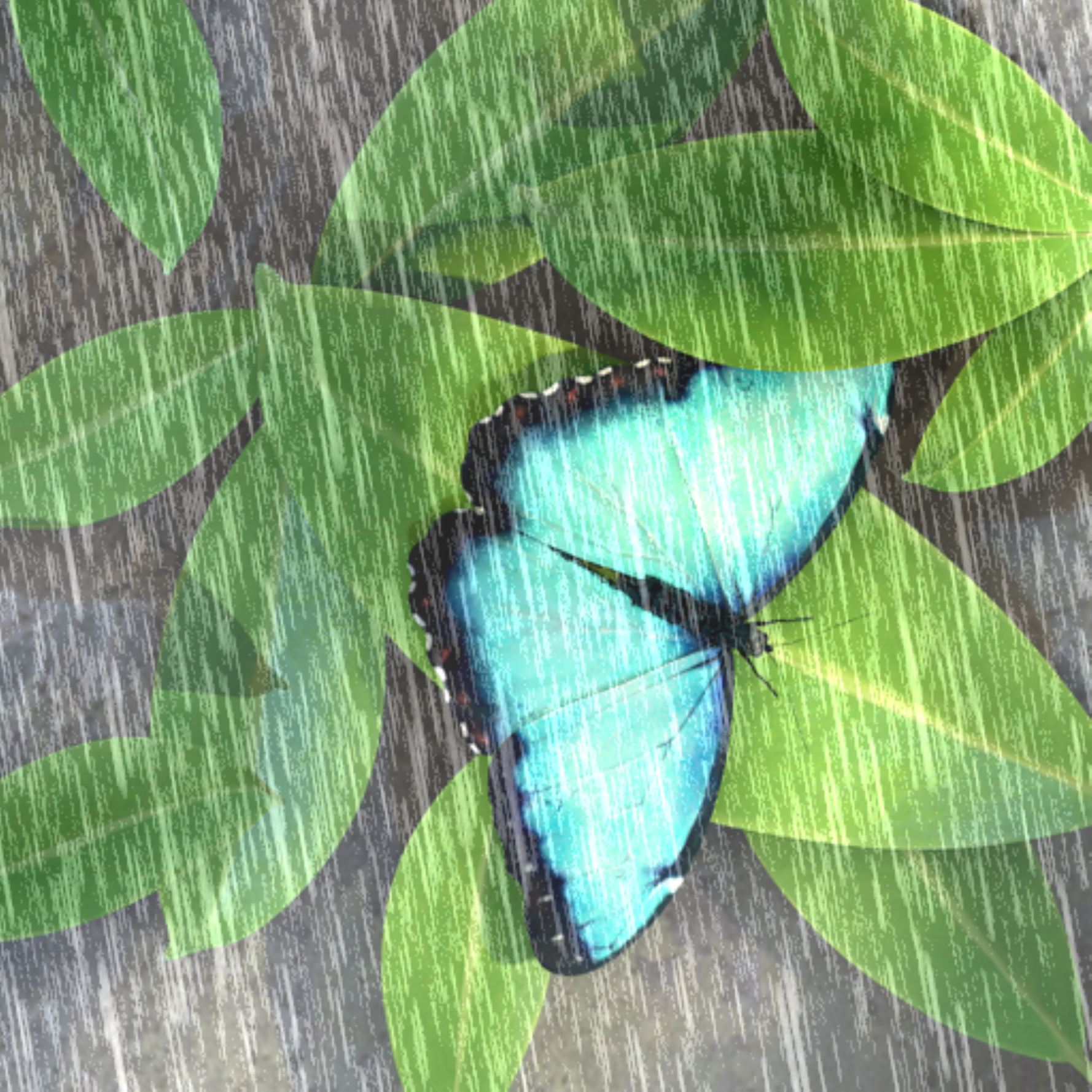} &
			\includegraphics[width=\subwidth\linewidth]{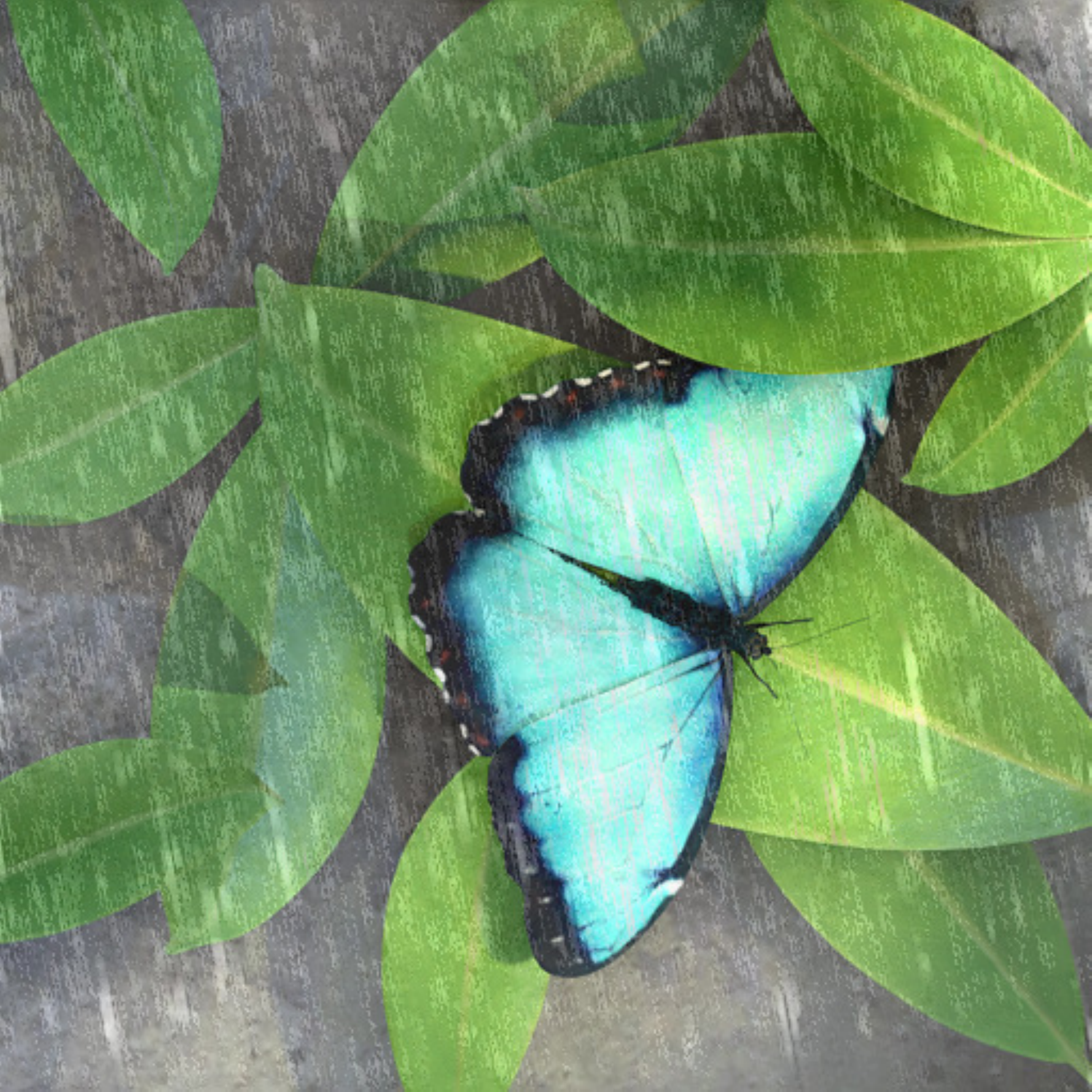} &
			\includegraphics[width=\subwidth\linewidth]{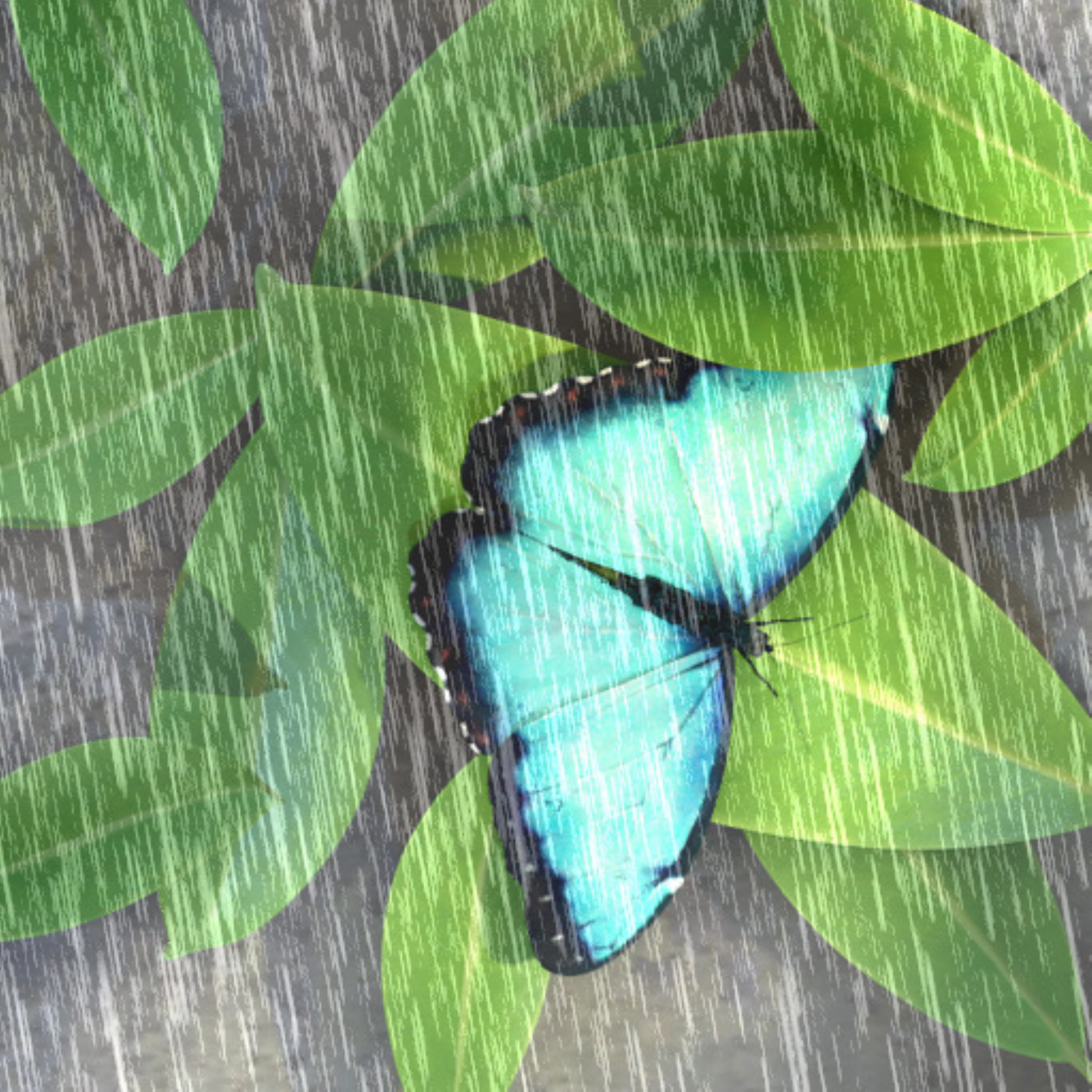} &
			\includegraphics[width=\subwidth\linewidth]{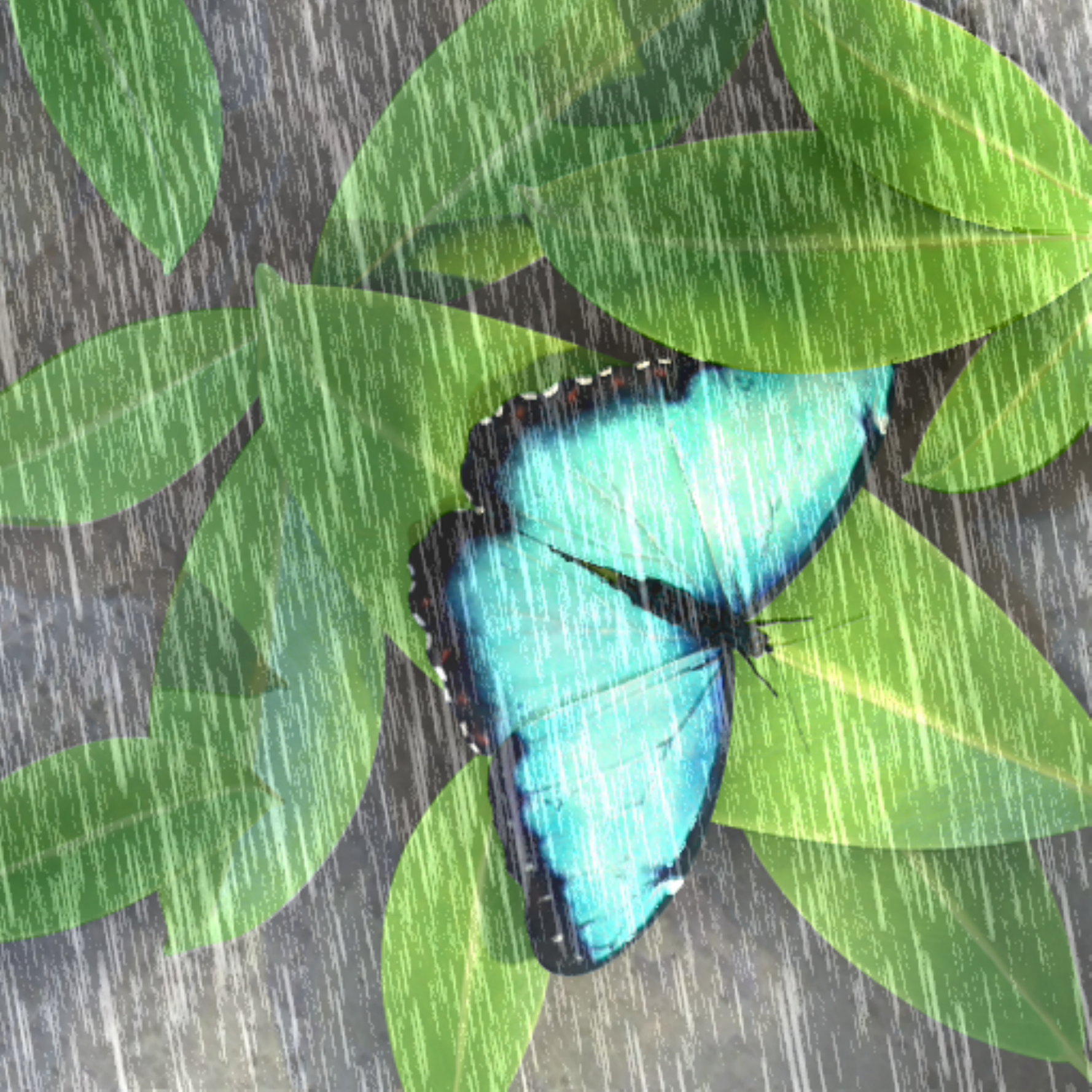} &
			\includegraphics[width=\subwidth\linewidth]{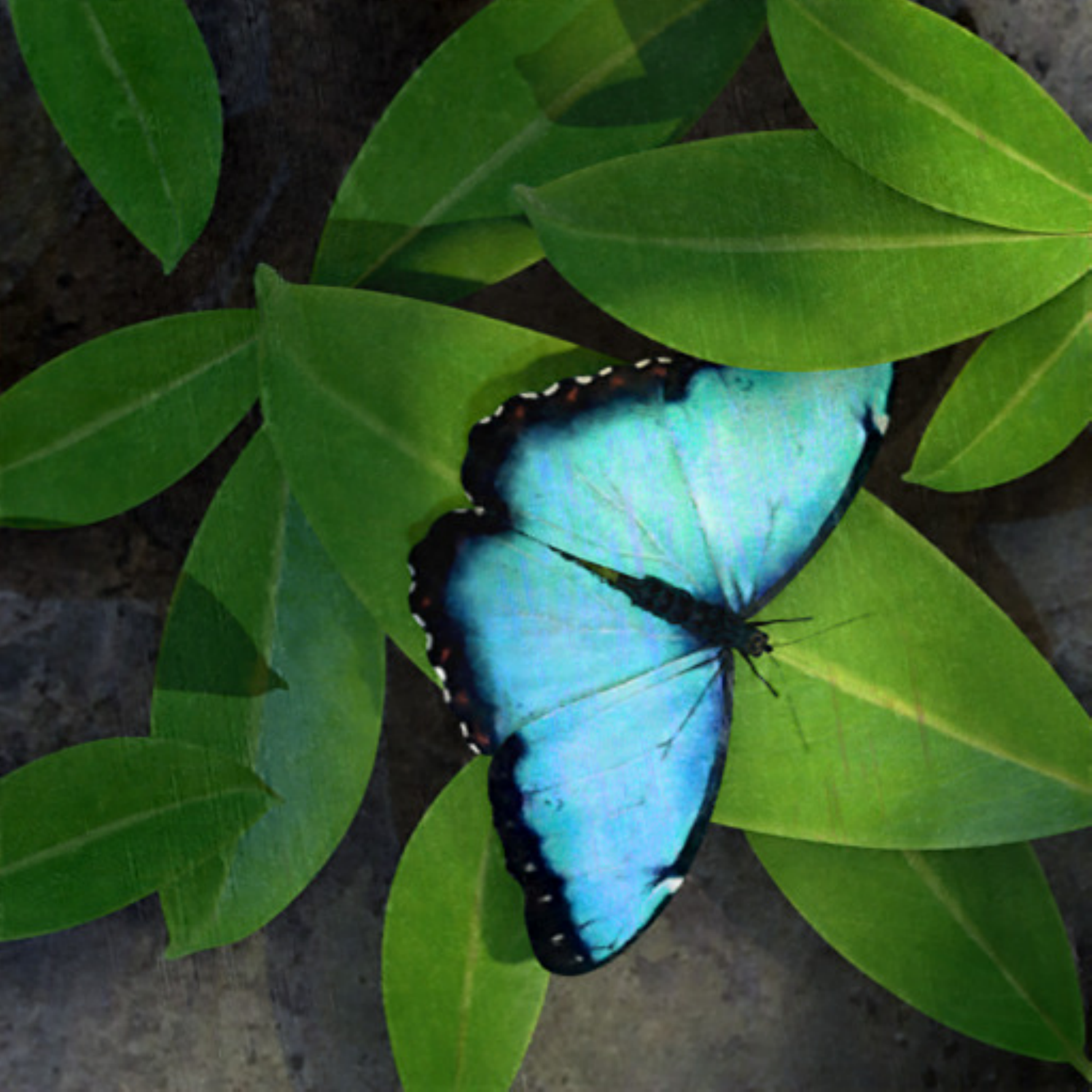}&
			\includegraphics[width=\subwidth\linewidth]{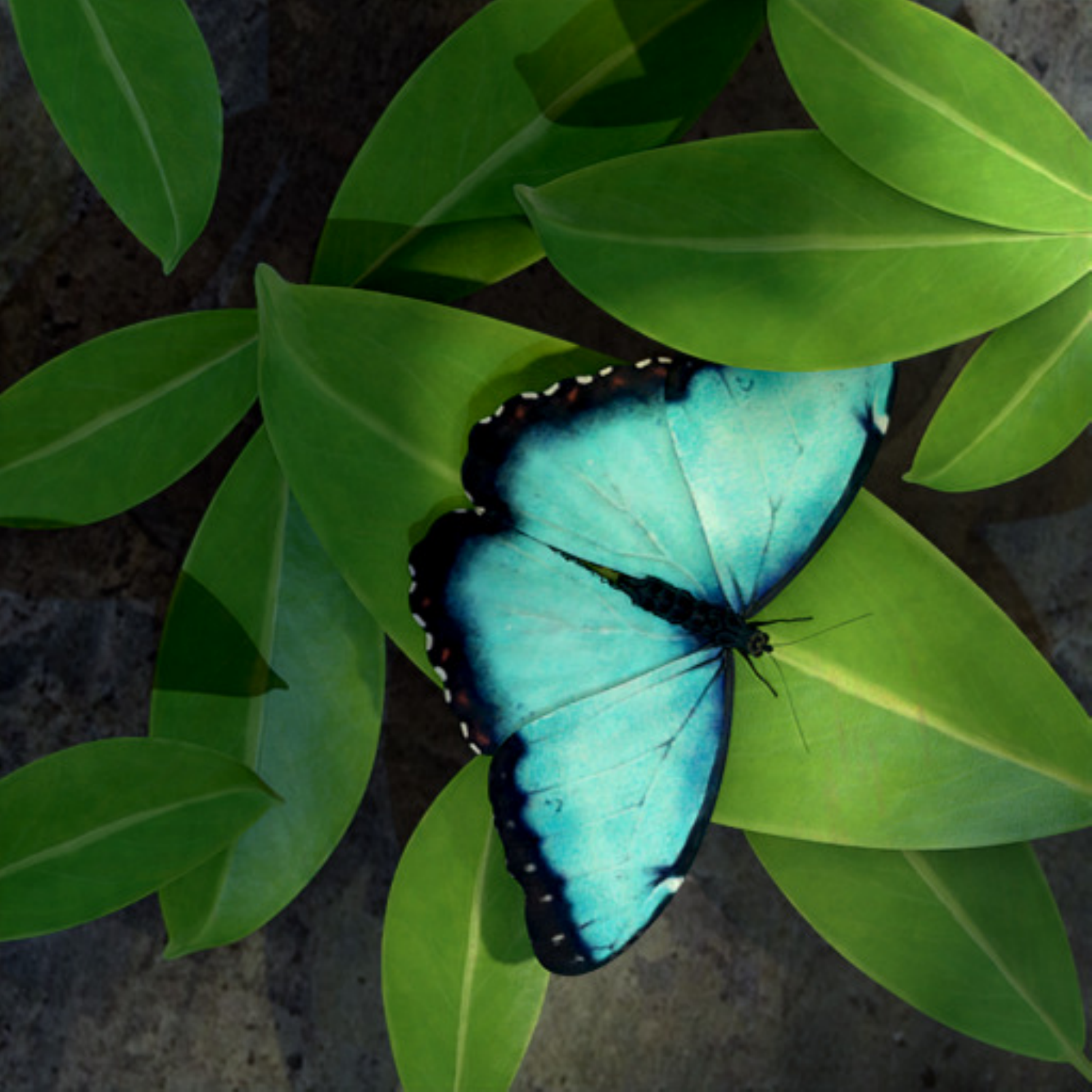} \\

			\normalsize{Rainy LFI}&			
			\normalsize{~\cite{Li18v}, $11.83/0.56$}&
			\normalsize{~\cite{Liu18v}, $10.40/0.48$}&
			\normalsize{~\cite{li21}, $10.40/0.46$}&
			\normalsize{~\cite{zhang2022},$25.13/0.83$}&
			\normalsize{Ours, $\textbf{29.81}/\textbf{0.96}$}\\

			\includegraphics[width=\subwidth\linewidth]{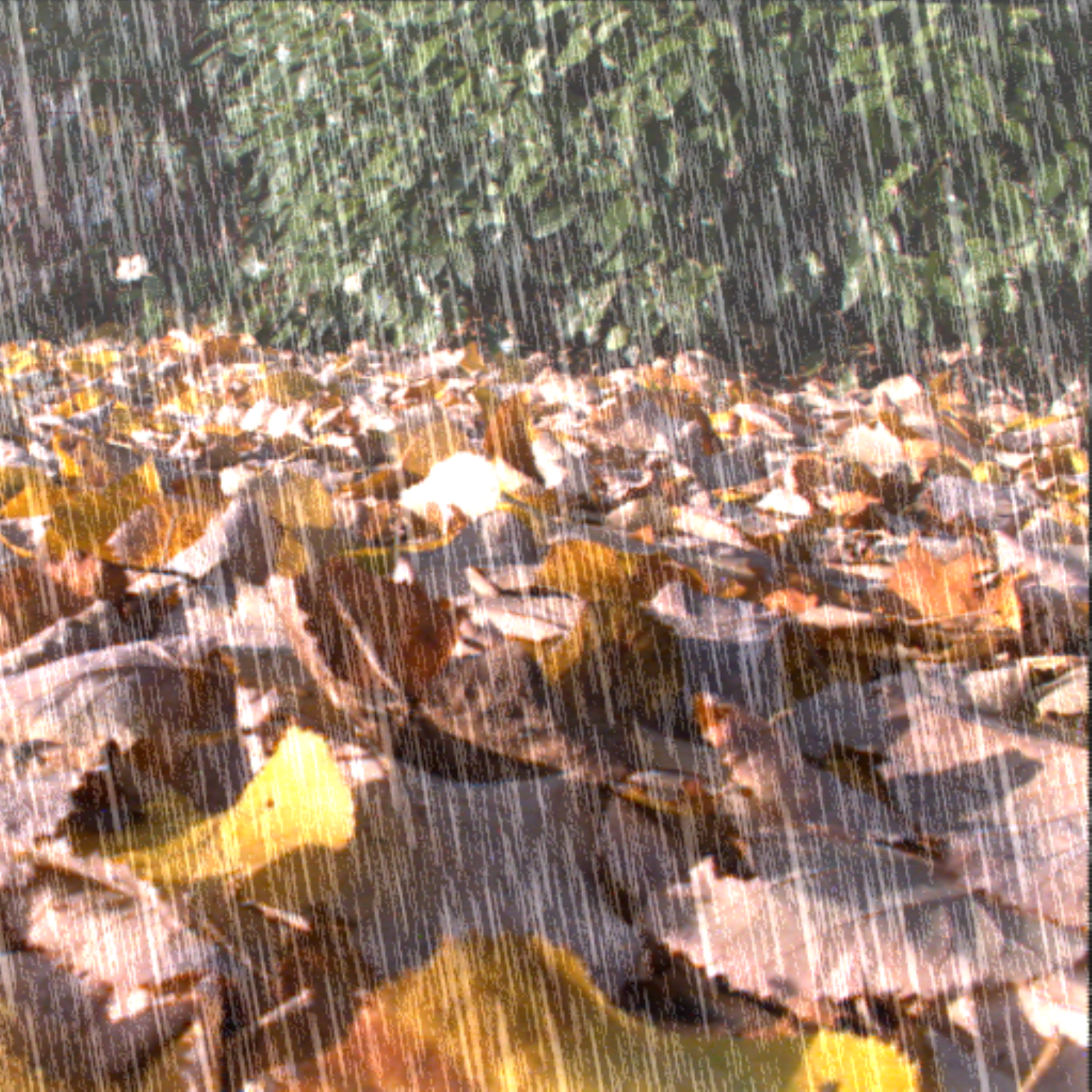} &
			\includegraphics[width=\subwidth\linewidth]{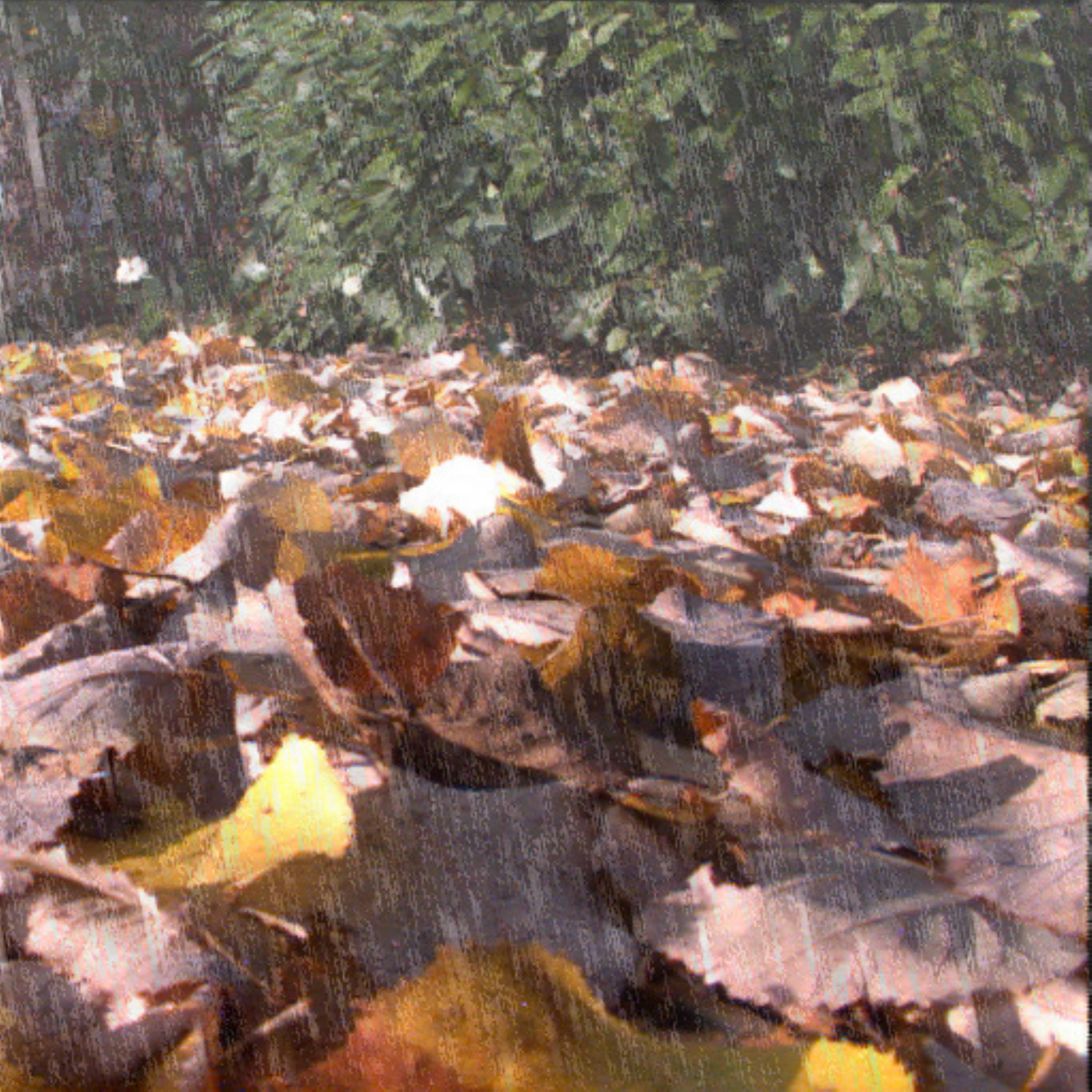} &
			\includegraphics[width=\subwidth\linewidth]{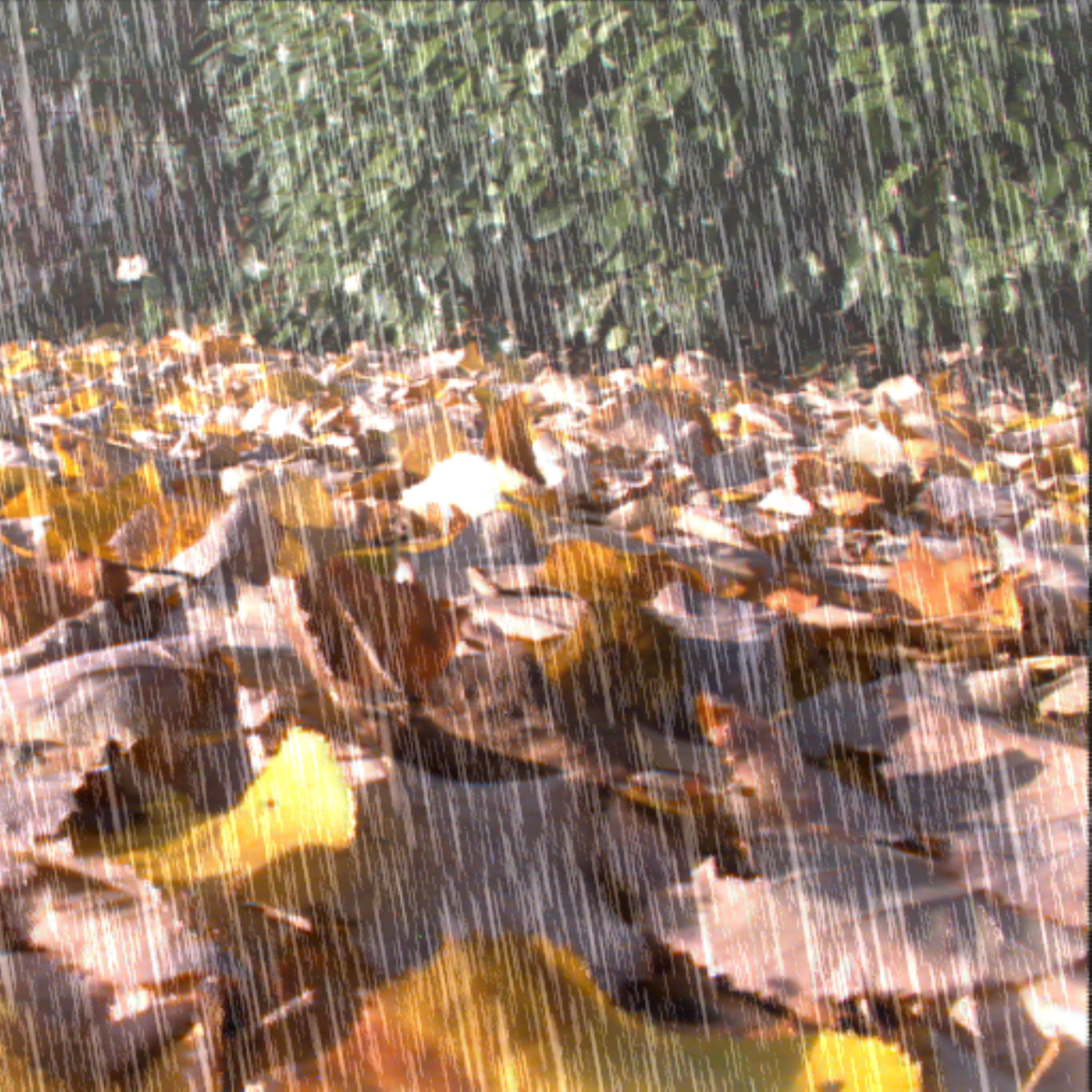} &
			\includegraphics[width=\subwidth\linewidth]{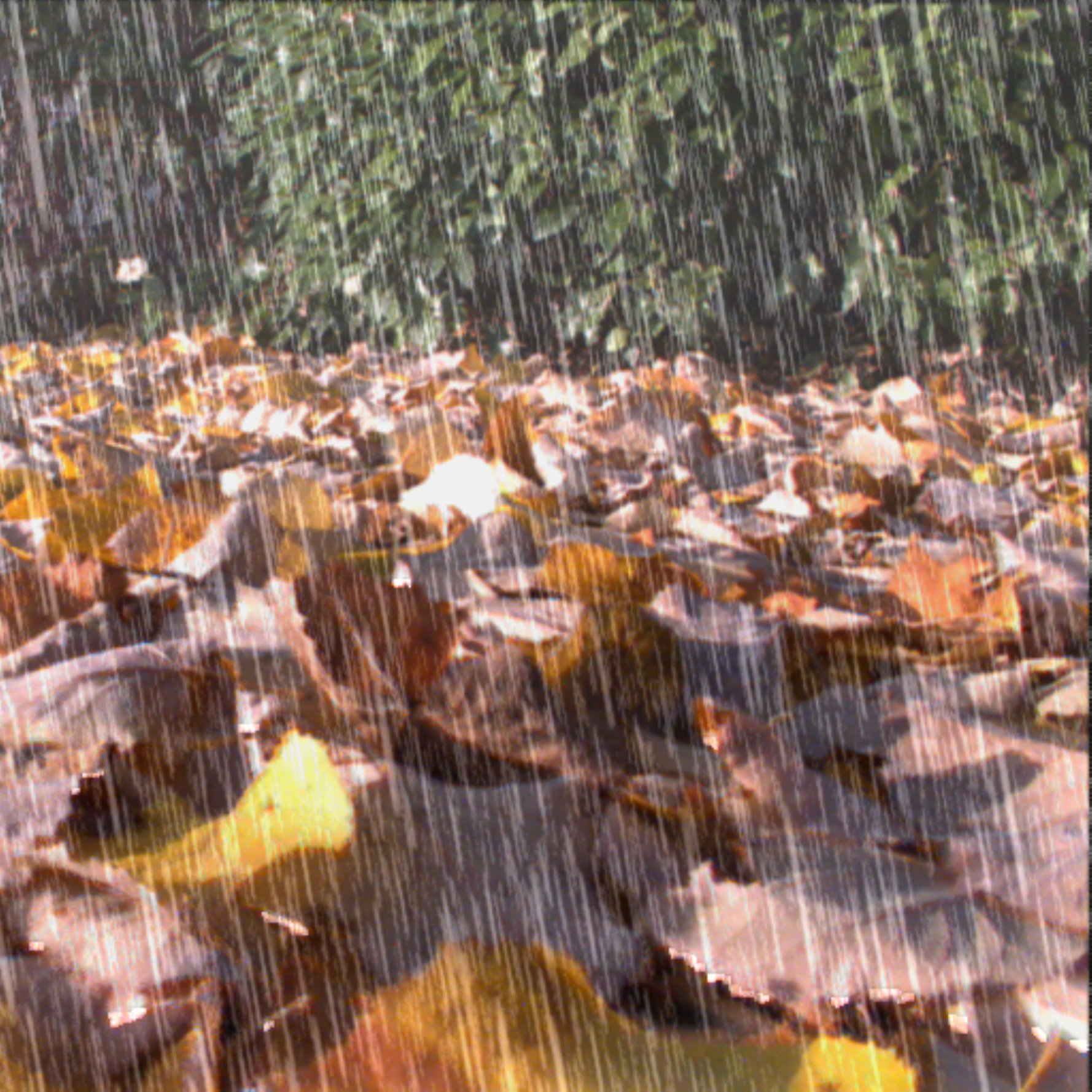} 	&		\includegraphics[width=\subwidth\linewidth]{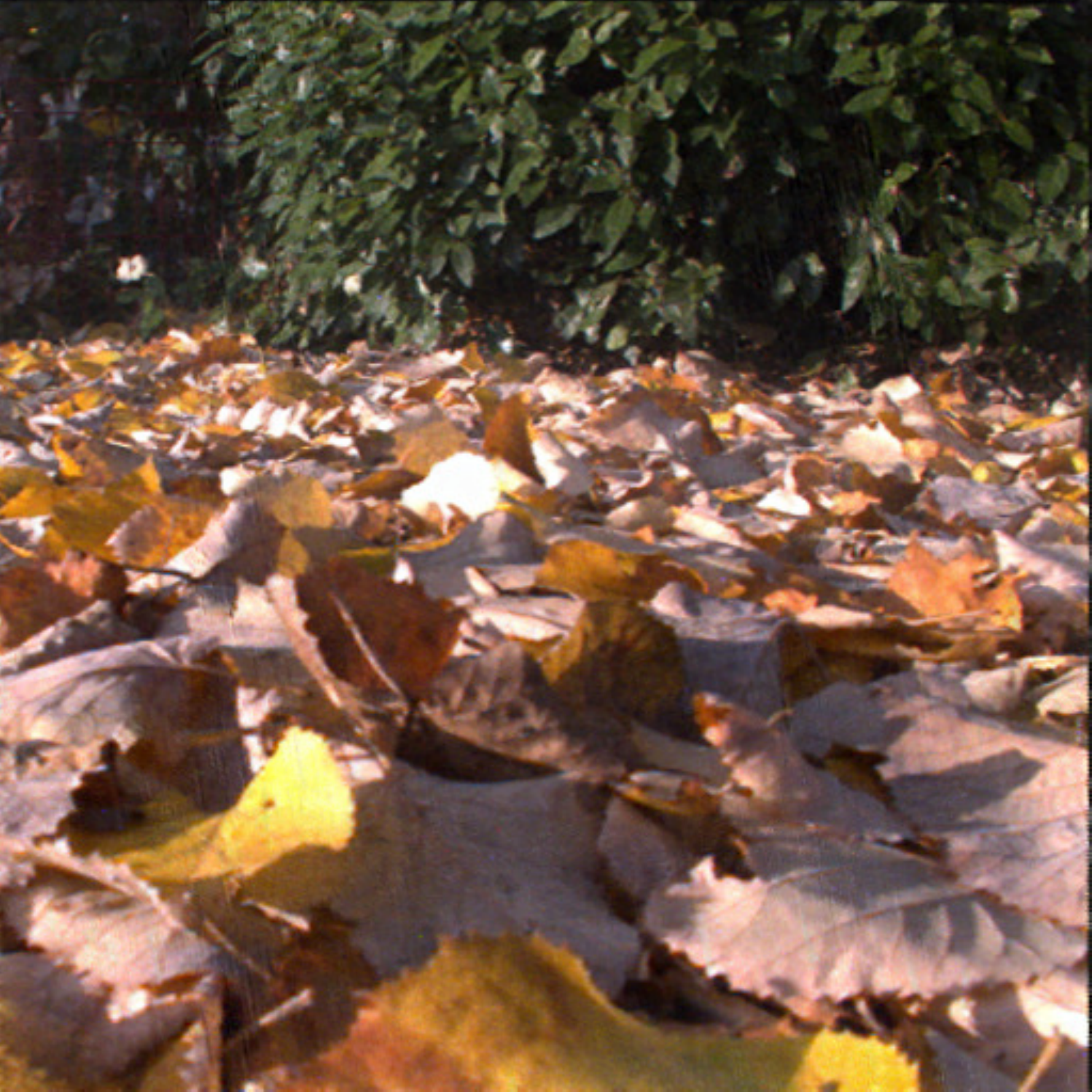}&
			\includegraphics[width=\subwidth\linewidth]{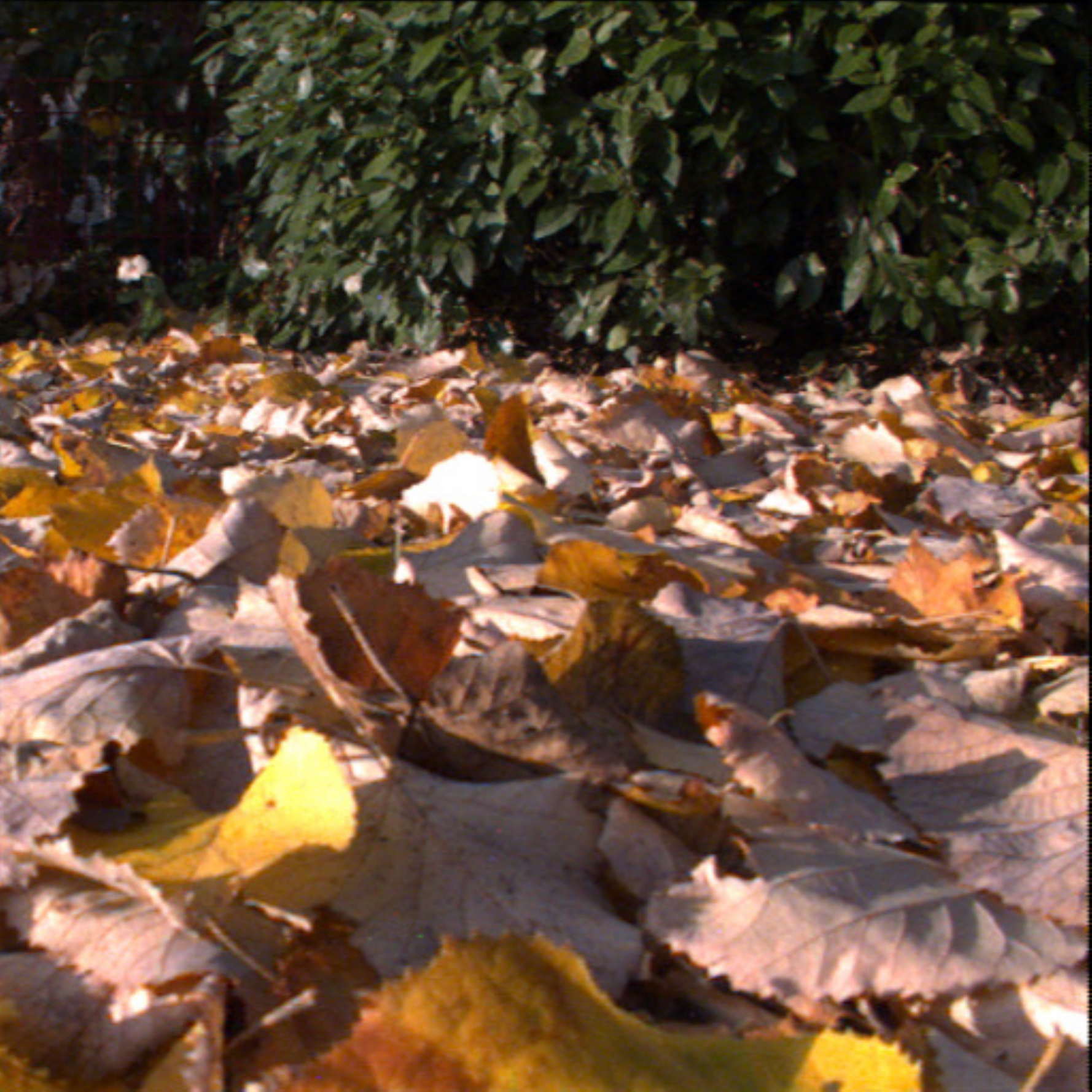} \\
			
			\normalsize{Rainy LFI}&			
			\normalsize{~\cite{Li18v}, $13.98/0.51$}&
			\normalsize{~\cite{Liu18v}, $12.47/0.56$}&
			\normalsize{~\cite{li21}, $14.15/0.59$}&
			\normalsize{~\cite{zhang2022},$25.53/0.90$}&
			\normalsize{Ours, $\textbf{27.42}/\textbf{0.95}$}\\
			
			\includegraphics[width=\subwidth\linewidth]{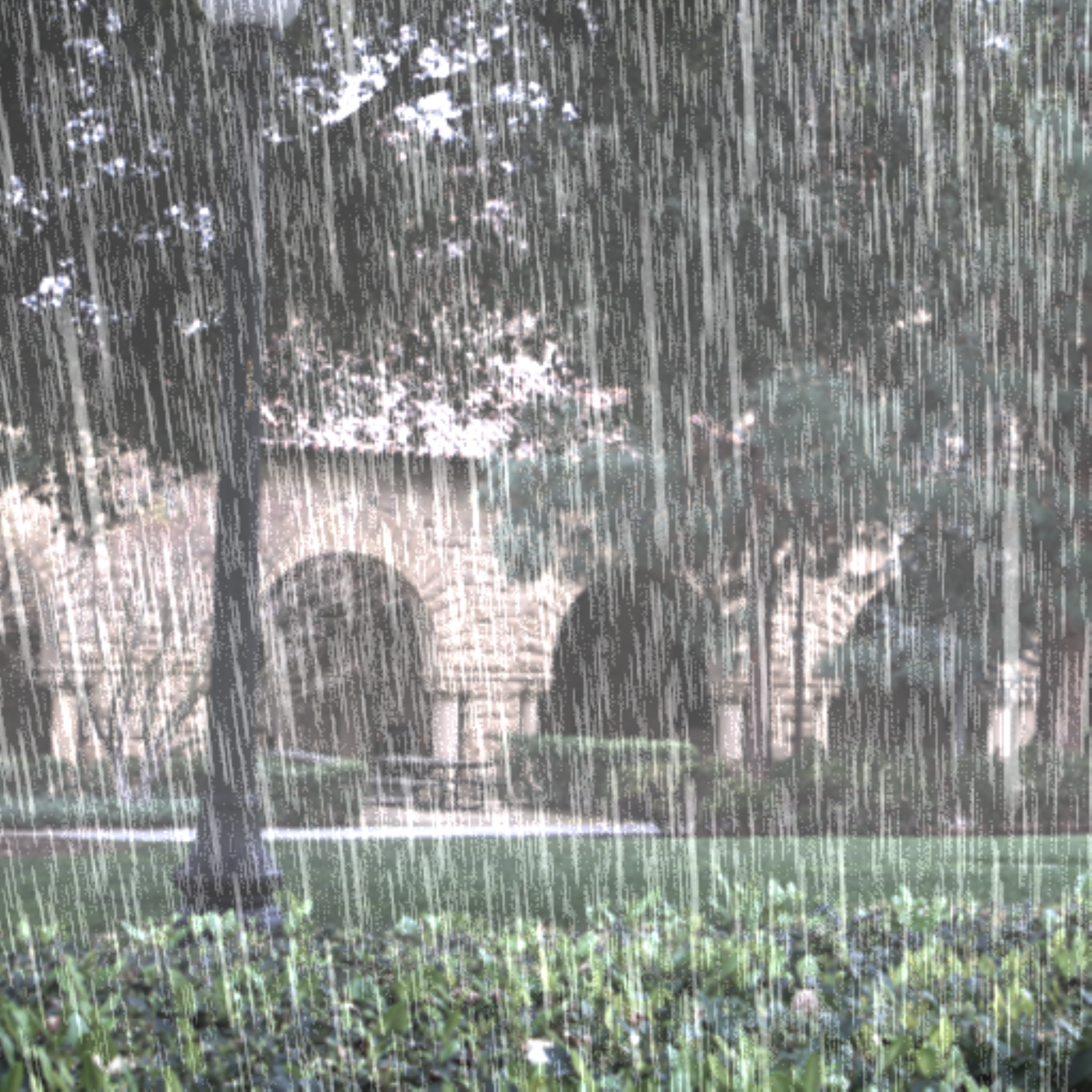} &
			\includegraphics[width=\subwidth\linewidth]{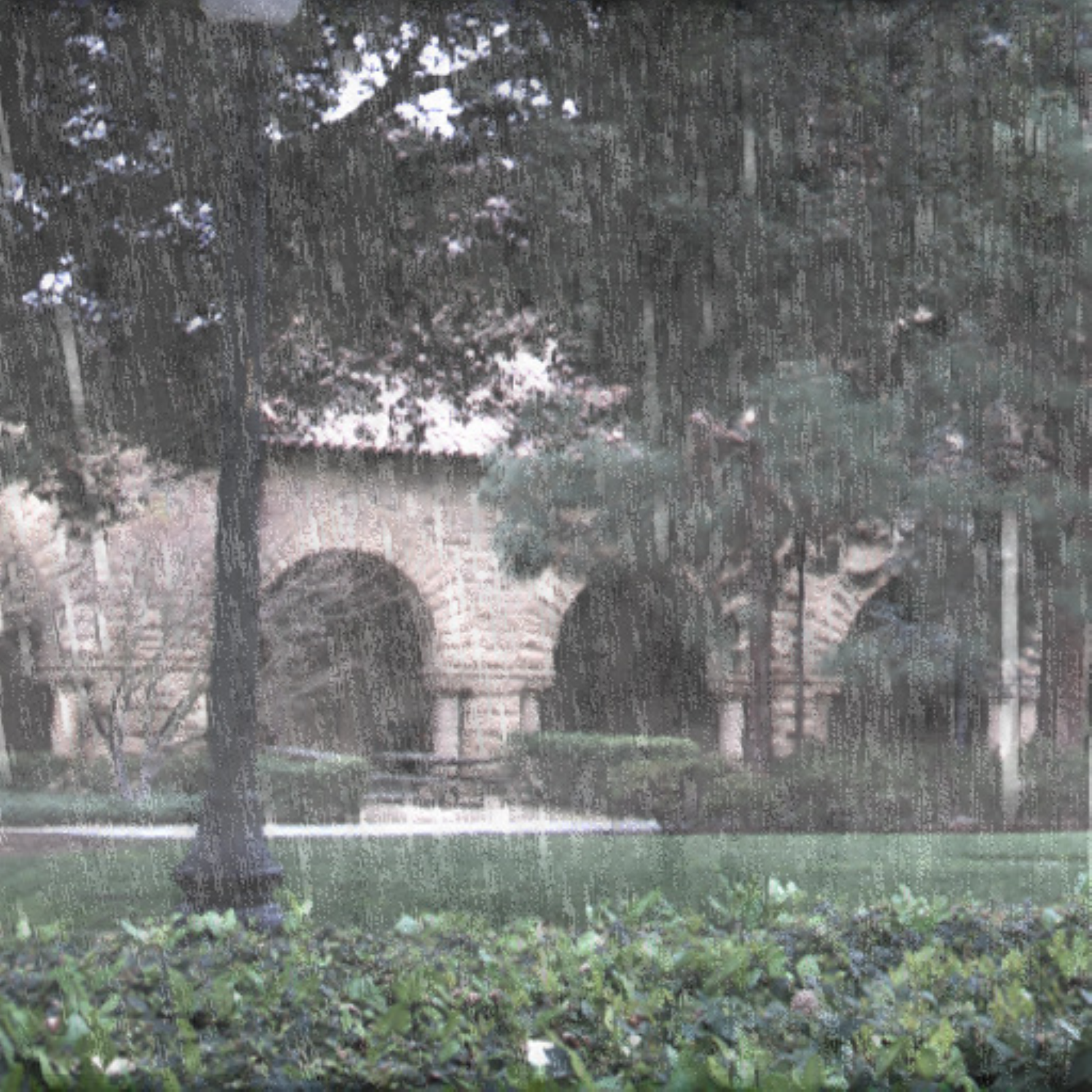} &
			\includegraphics[width=\subwidth\linewidth]{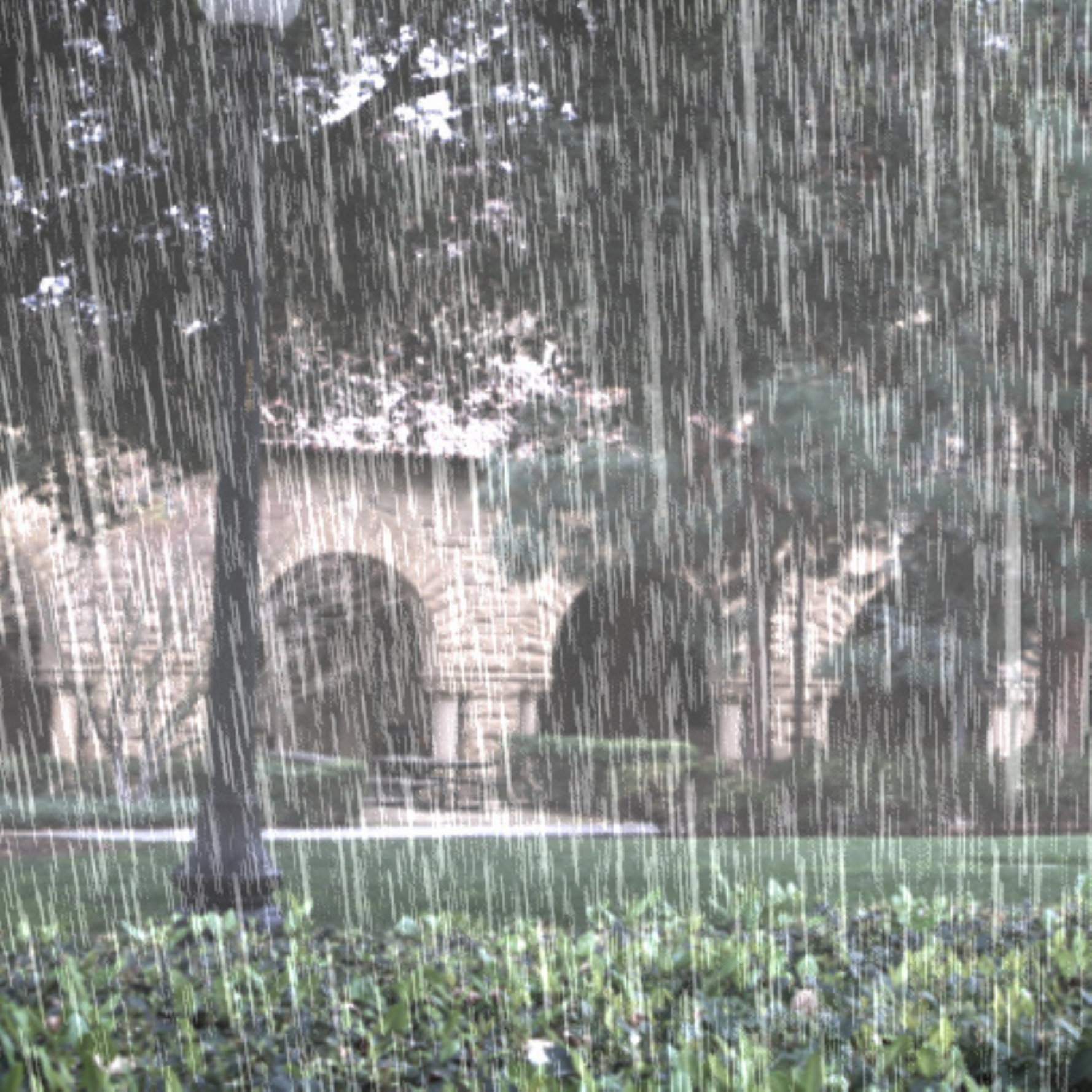} &
			\includegraphics[width=\subwidth\linewidth]{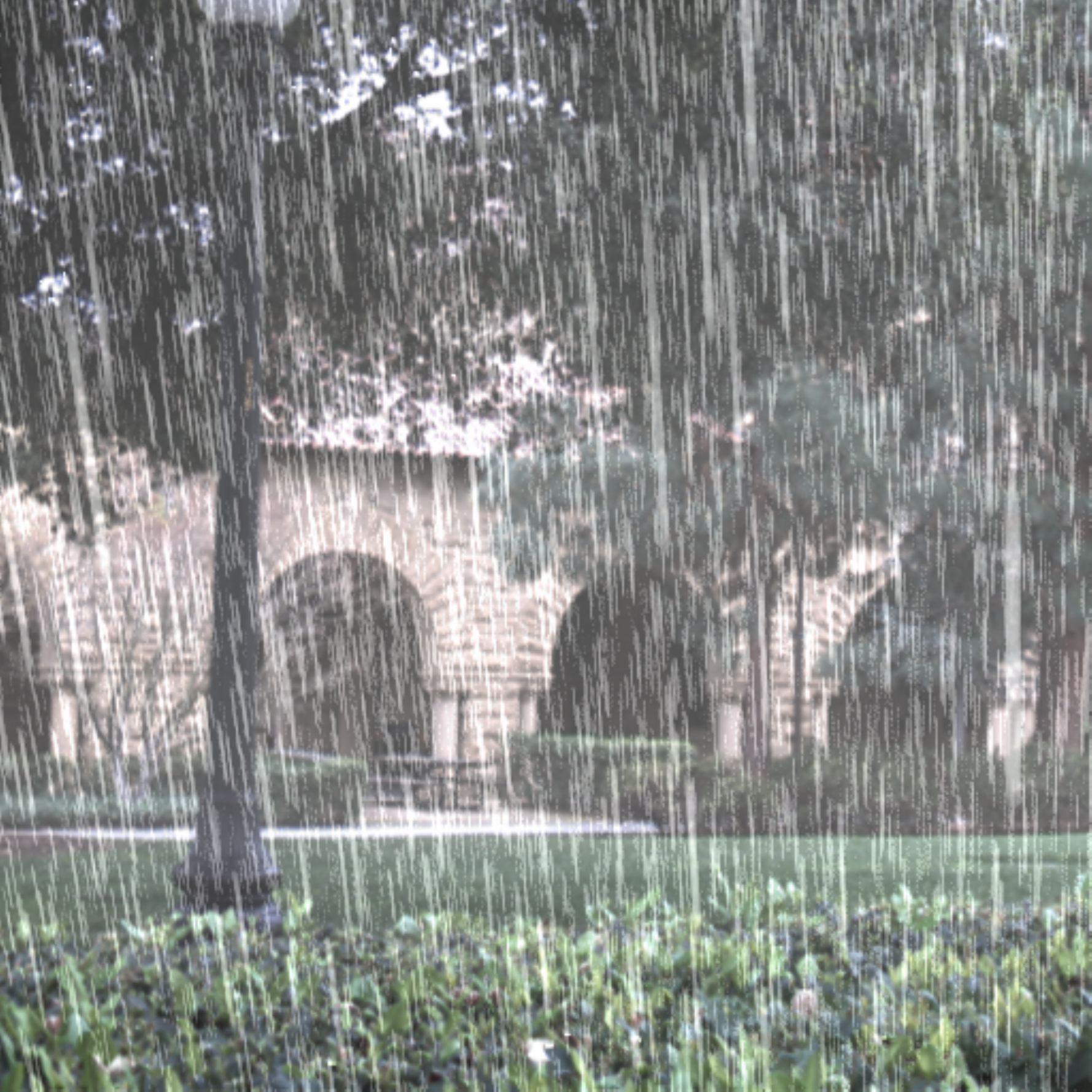} 	&		\includegraphics[width=\subwidth\linewidth]{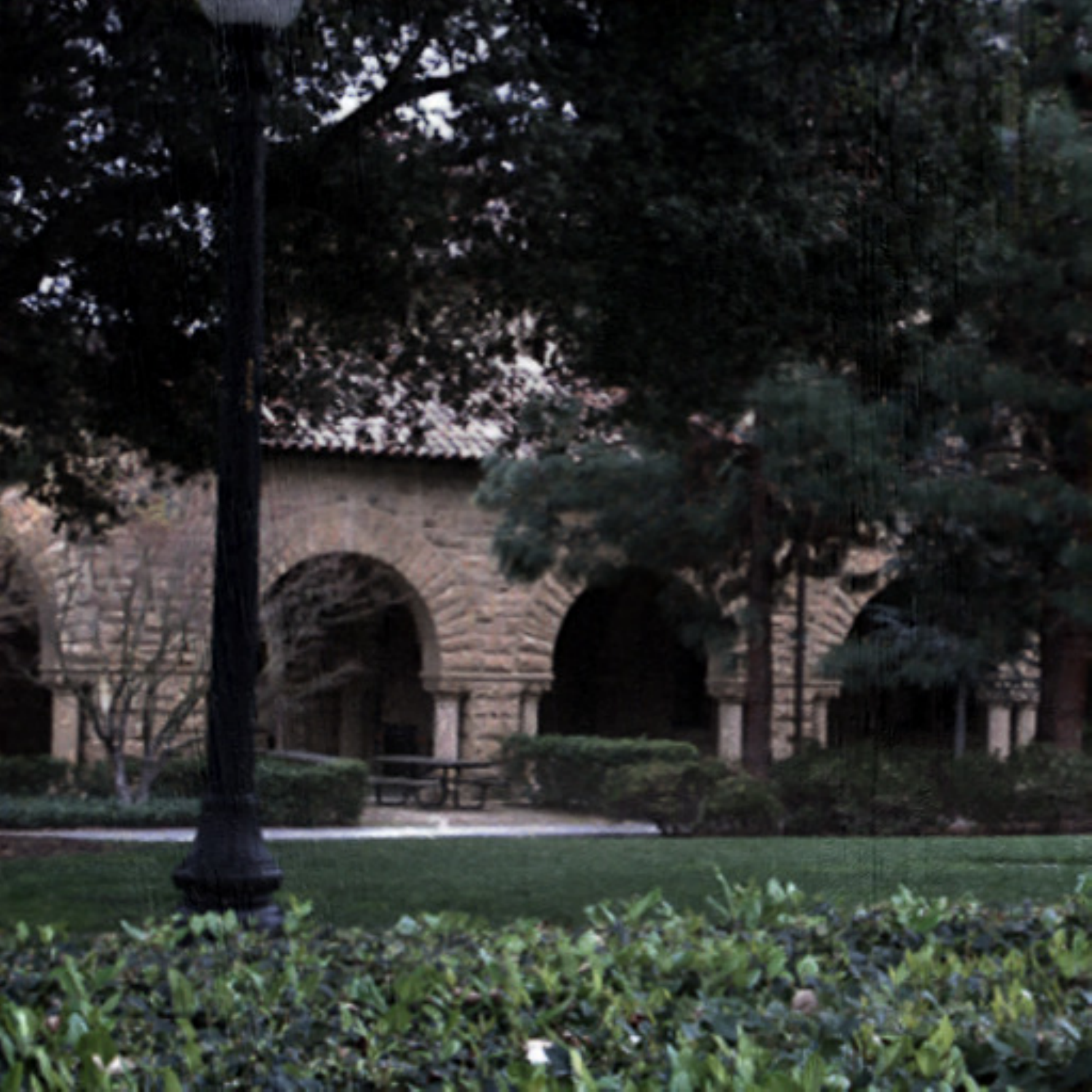}&
			\includegraphics[width=\subwidth\linewidth]{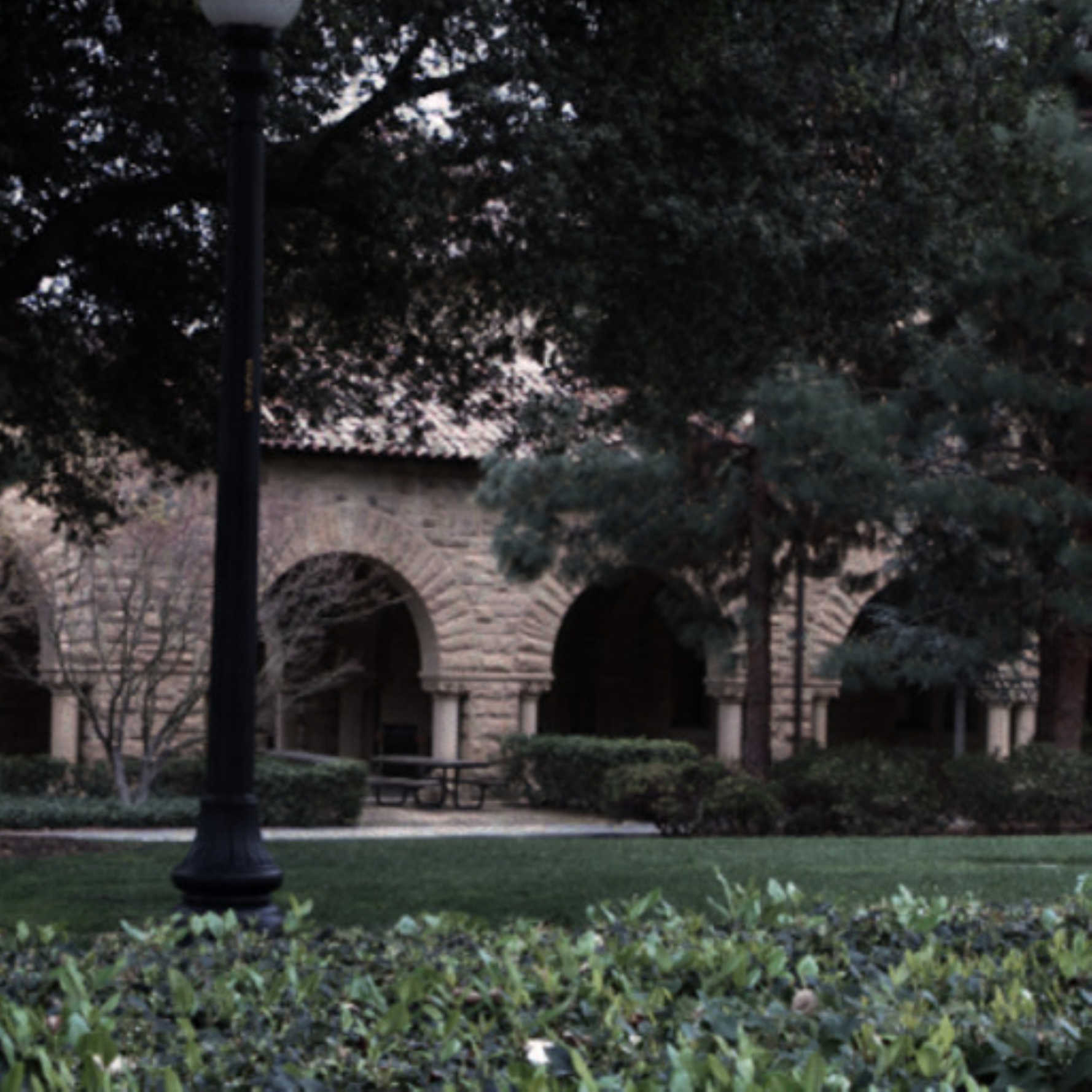} \\
			
			\normalsize{Rainy LFI}&			
			\normalsize{~\cite{Li18v}, $11.22/0.33$}&
			\normalsize{~\cite{Liu18v}, $9.12/0.30$}&
			\normalsize{~\cite{li21}, $8.99/0.29$}&
			\normalsize{~\cite{zhang2022},$27.80/0.87$}&
			\normalsize{Ours, $\textbf{30.41}/\textbf{0.97}$}\\
			
			\includegraphics[width=\subwidth\linewidth]{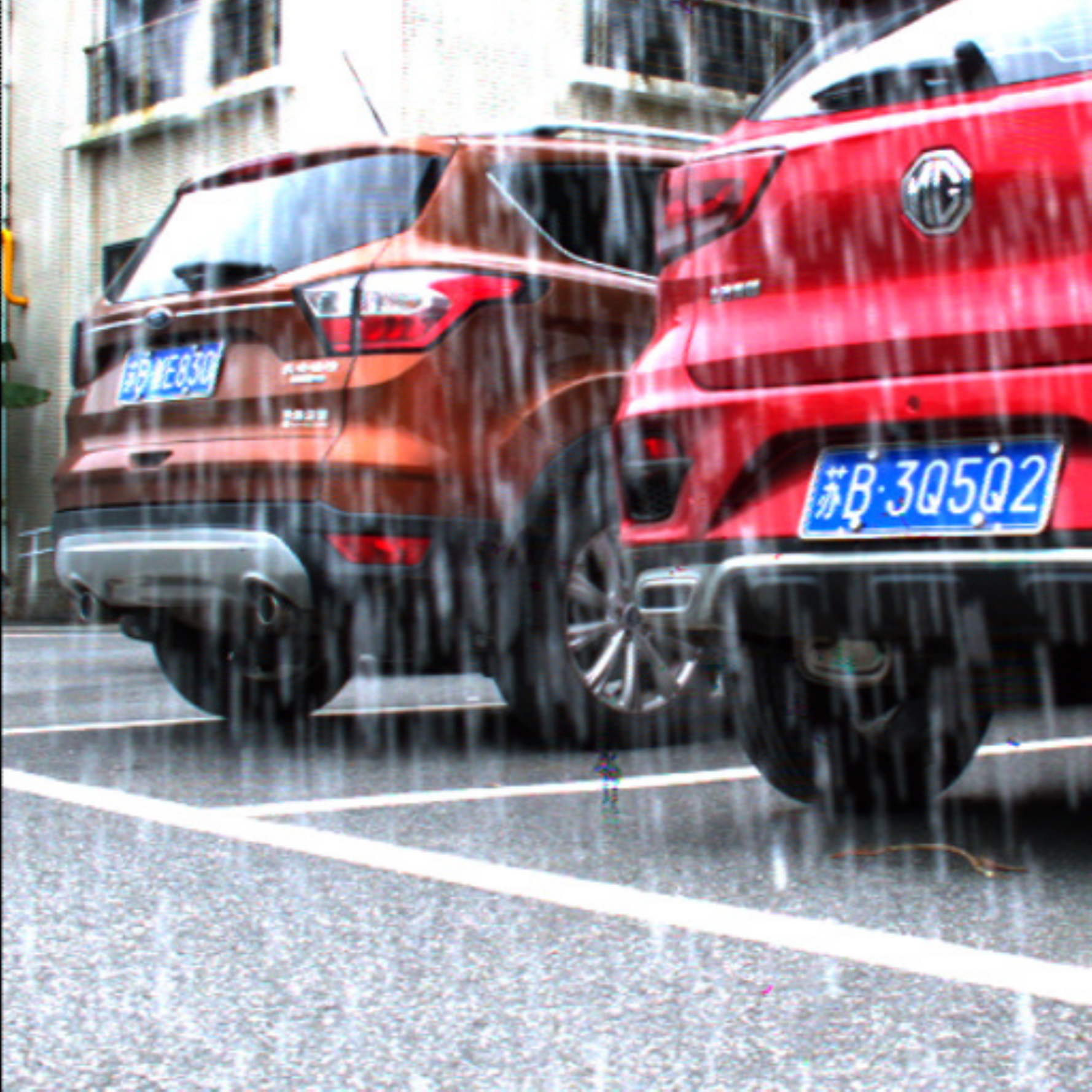} &
			\includegraphics[width=\subwidth\linewidth]{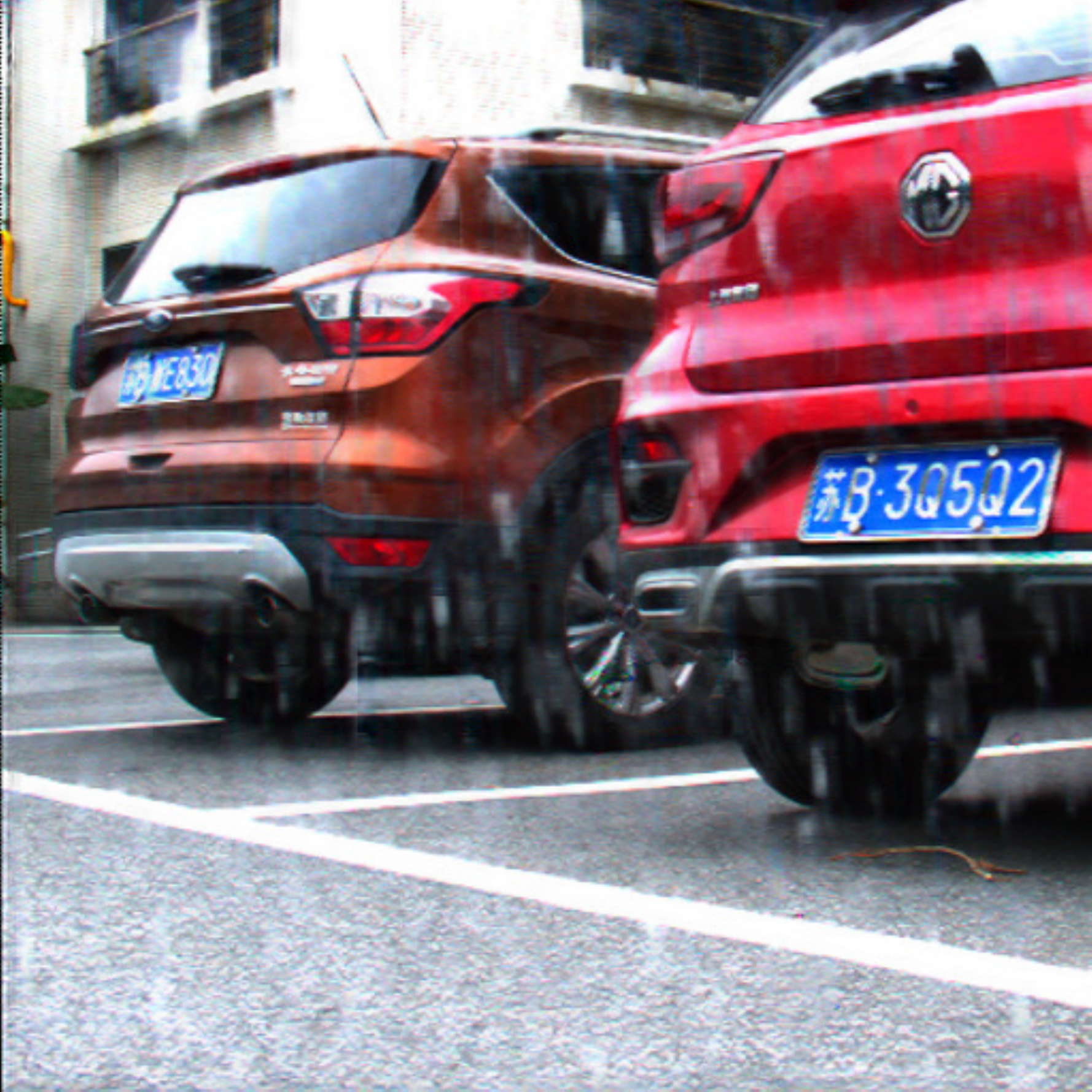} &
			\includegraphics[width=\subwidth\linewidth]{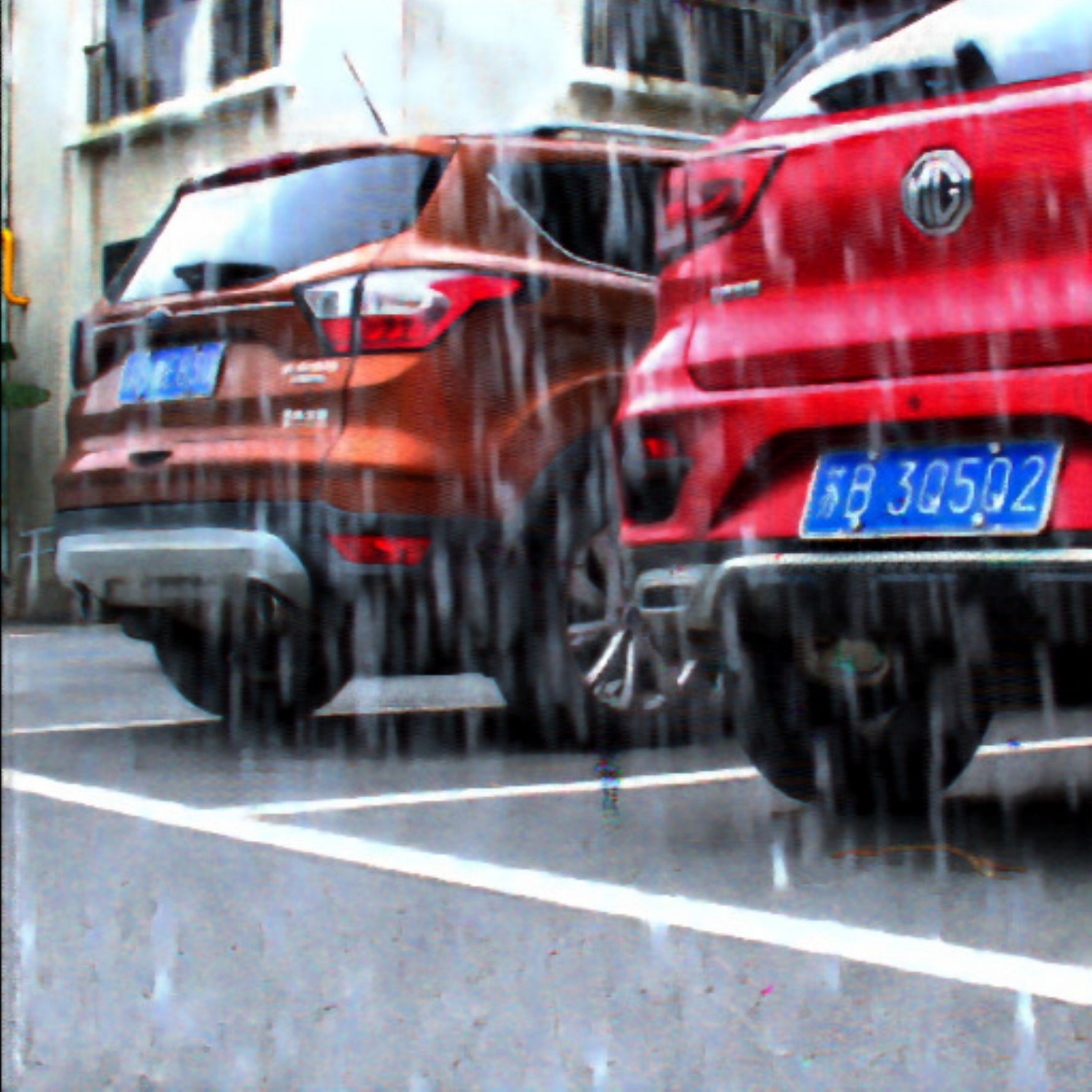} &
			\includegraphics[width=\subwidth\linewidth]{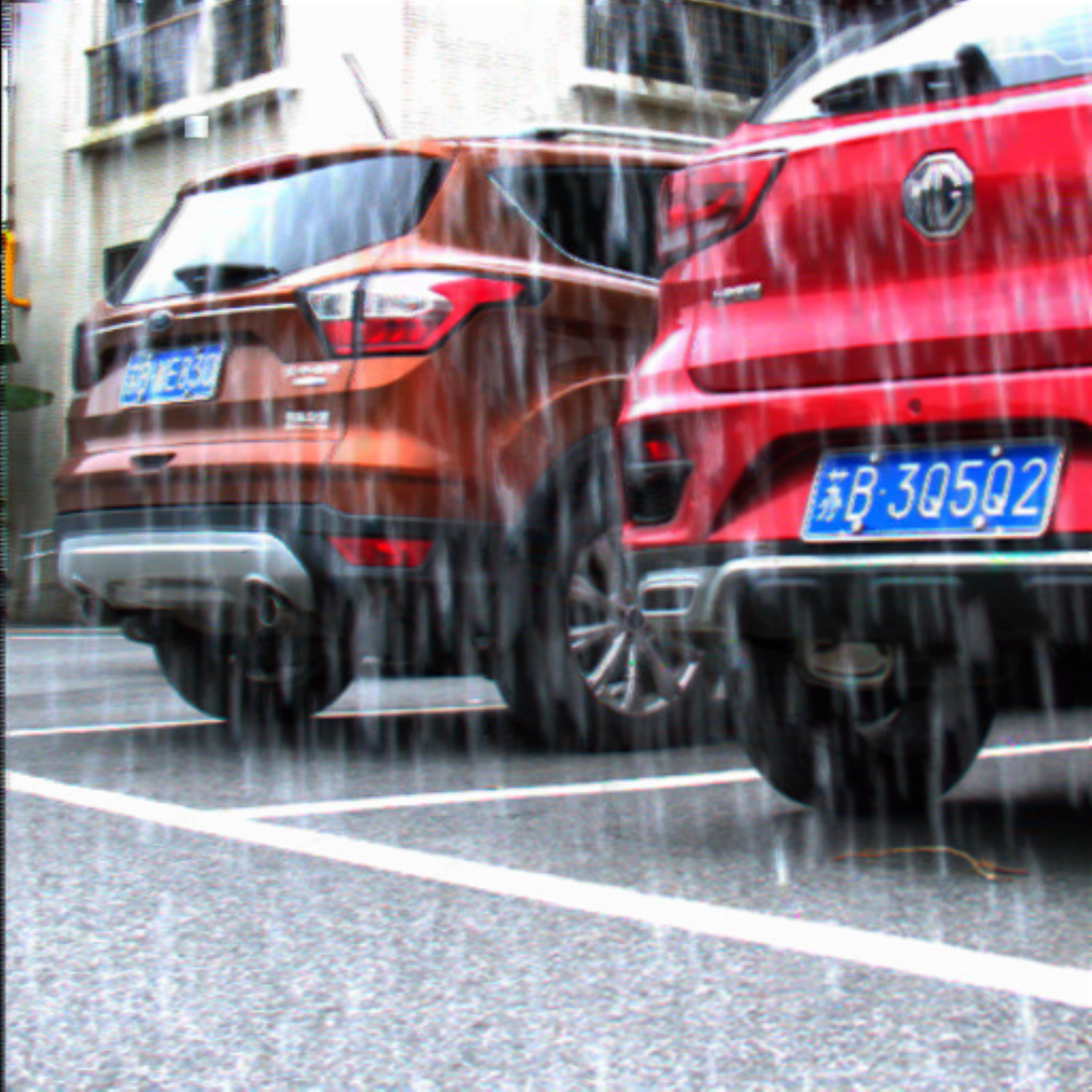} 	&		\includegraphics[width=\subwidth\linewidth]{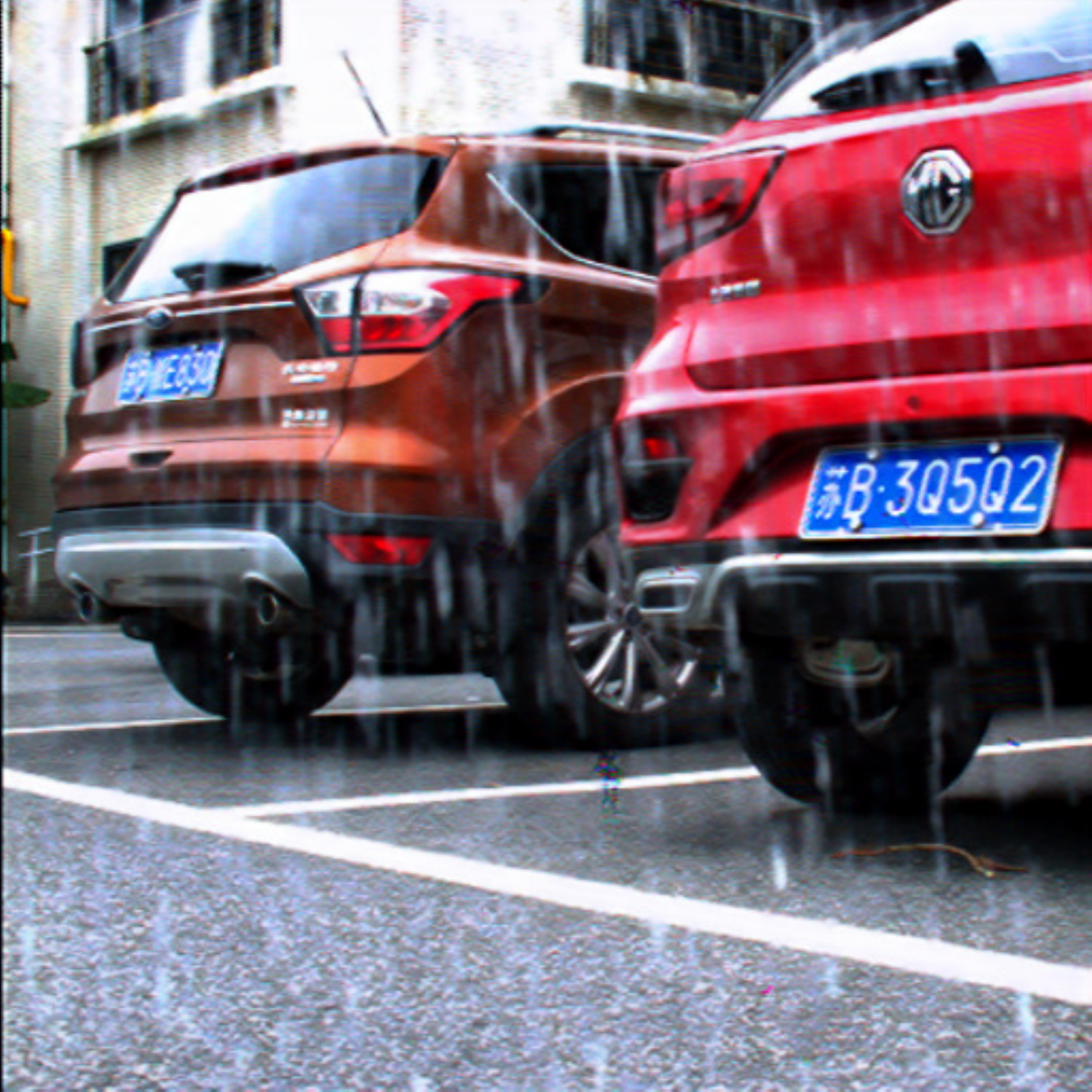}&
			\includegraphics[width=\subwidth\linewidth]{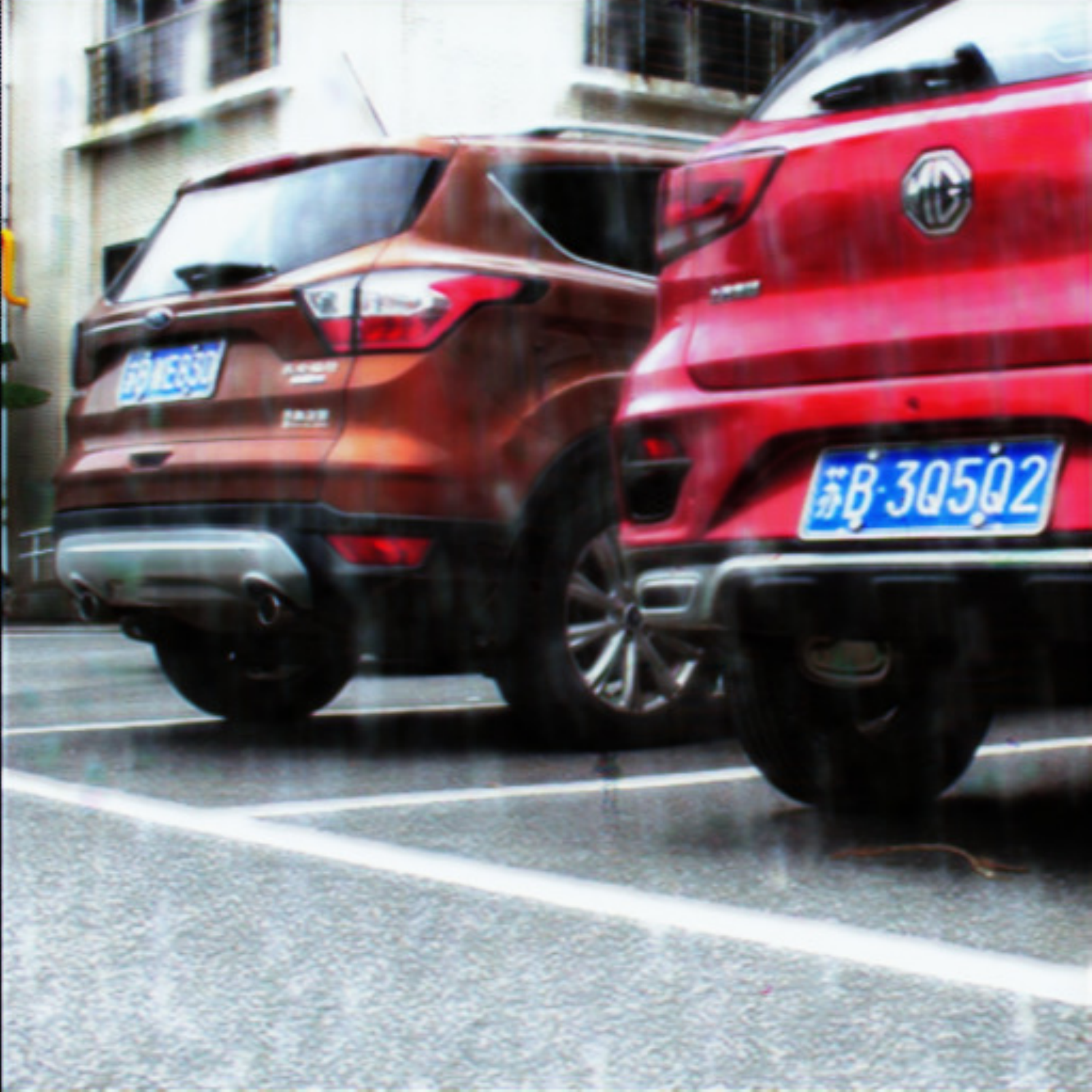} \\
			
			\normalsize{Rainy LFI}&			
			\normalsize{~\cite{Li18v}}&
			\normalsize{~\cite{Liu18v}}&
			\normalsize{~\cite{li21}}&
			\normalsize{~\cite{zhang2022}}&
			\normalsize{Ours}\\
			
			\includegraphics[width=\subwidth\linewidth]{fig2/test/input/1008} &
			\includegraphics[width=\subwidth\linewidth]{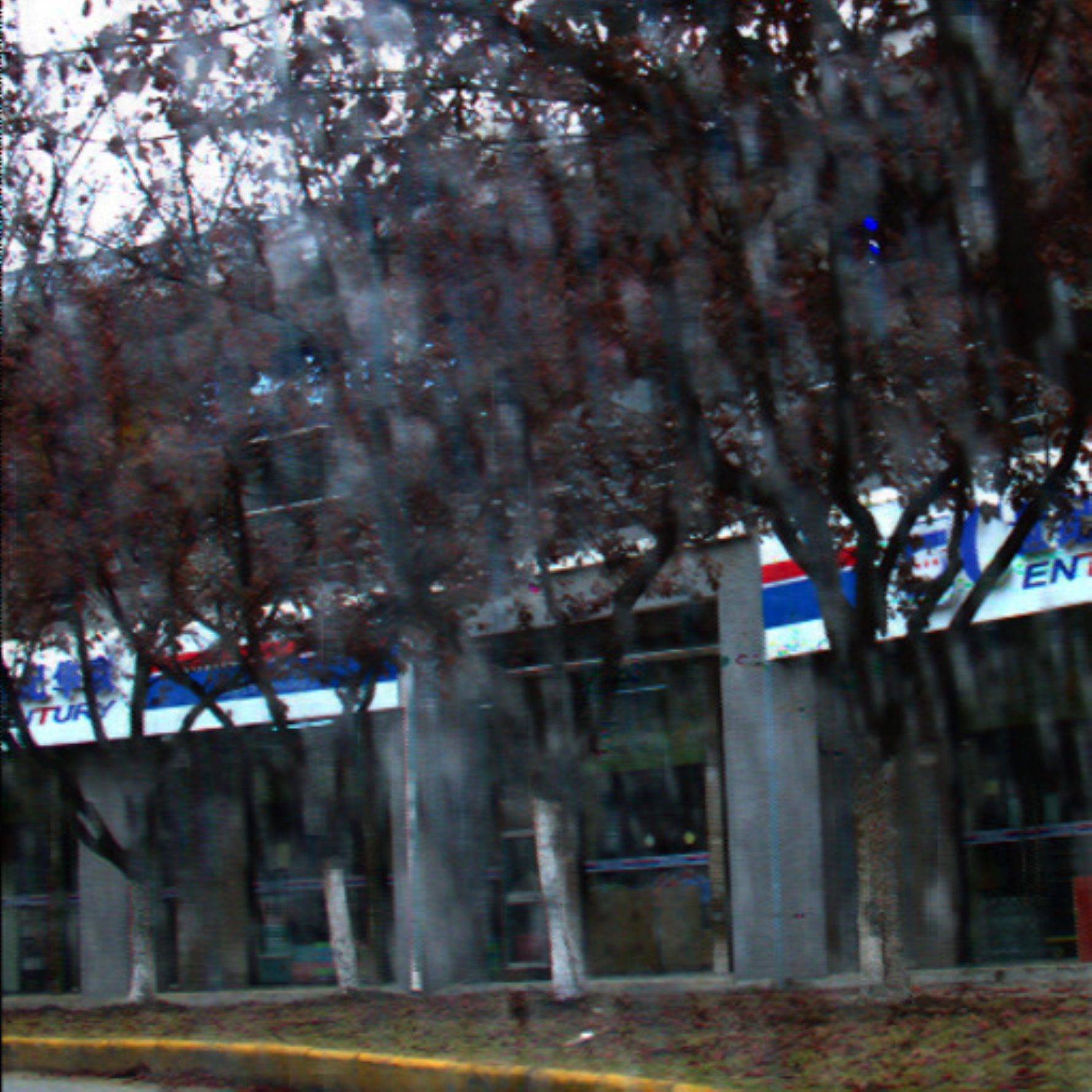} &
			\includegraphics[width=\subwidth\linewidth]{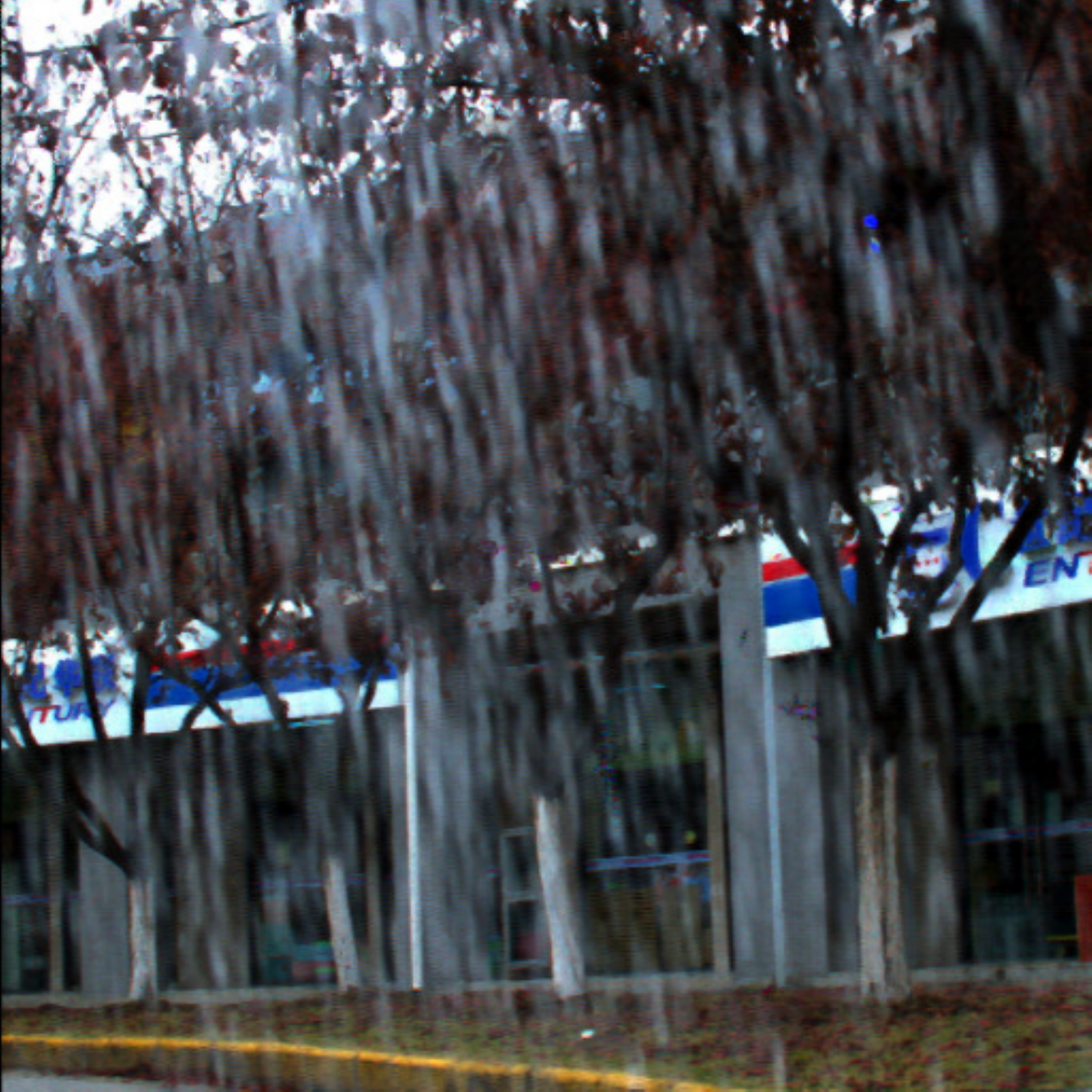} &
			\includegraphics[width=\subwidth\linewidth]{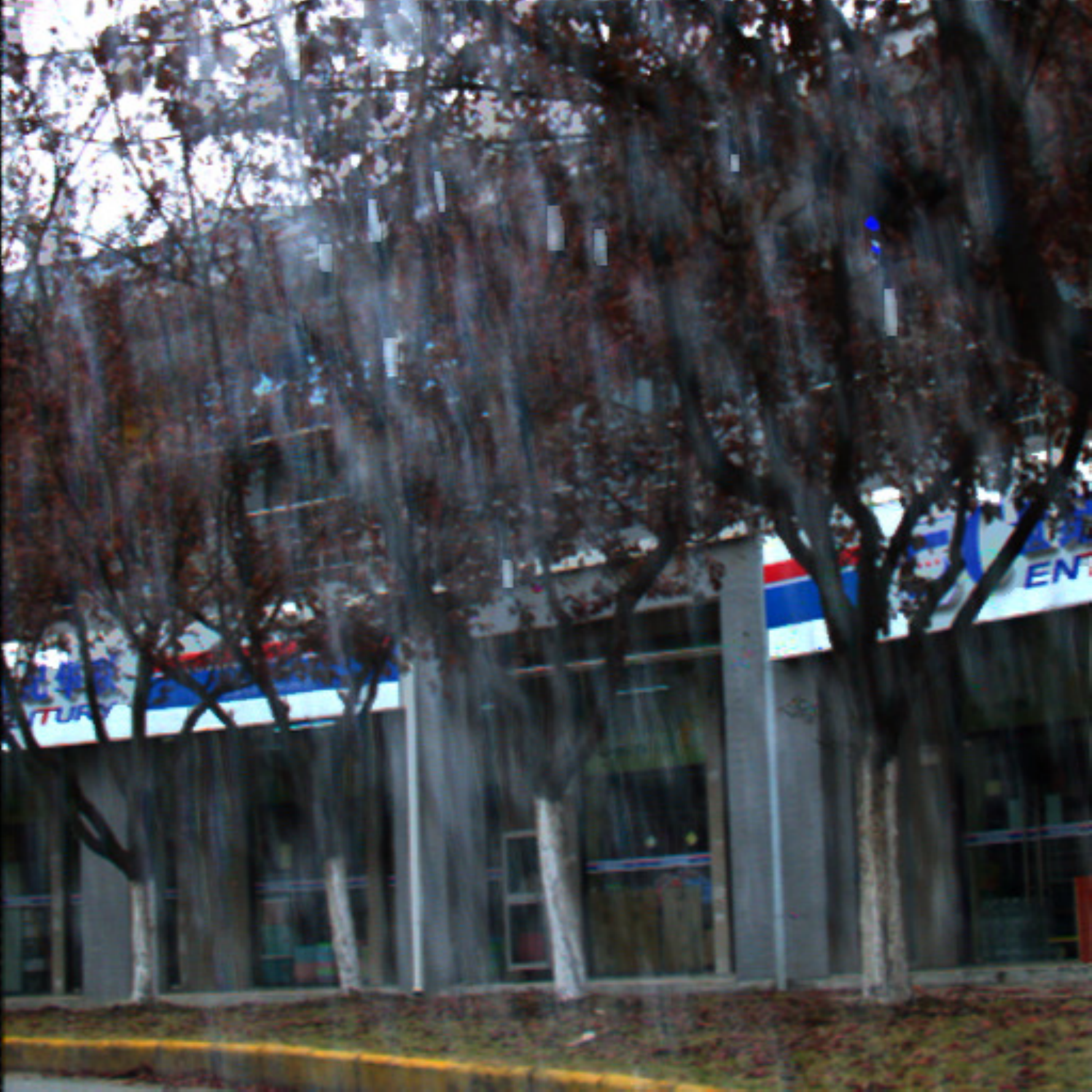} 	&		\includegraphics[width=\subwidth\linewidth]{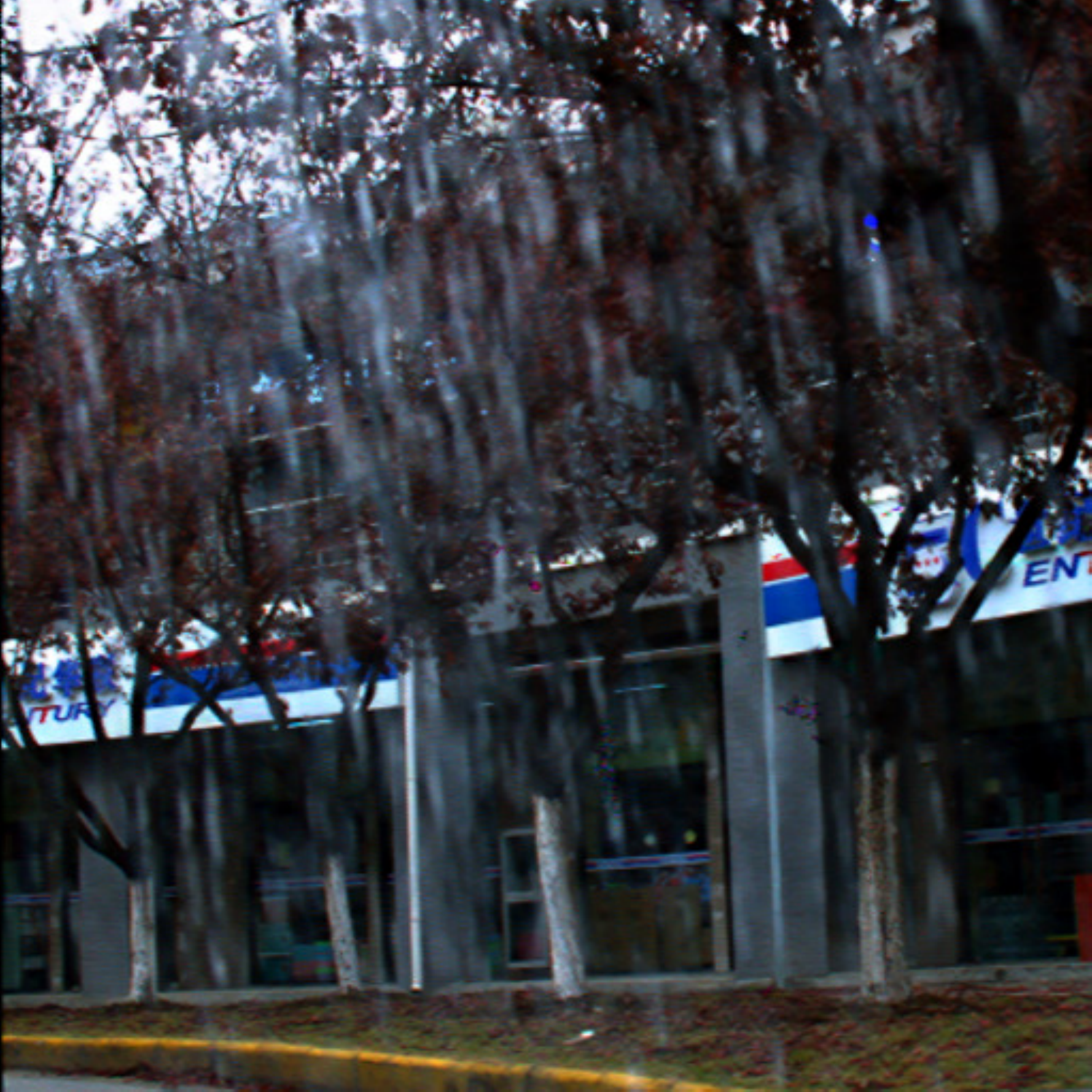}&
			\includegraphics[width=\subwidth\linewidth]{fig2/ours/1008} \\
			
			\normalsize{Rainy LFI}&			
			\normalsize{~\cite{Li18v}}&
			\normalsize{~\cite{Liu18v}}&
			\normalsize{~\cite{li21}}&
			\normalsize{\revised{~\cite{zhang2022}}}&
			\normalsize{Ours}\\
			
		\end{tabular}
	\end{center}
	\vspace{-0.01\textwidth}
	\caption{Comparison of our method with state-of-the-art video rain streak removal methods~\cite{Li18v,Liu18v,li21}\revised{~\cite{zhang2022}}. \revised{All these four methods take all sub-views of an LFI as input (frame sequence) for rain streak removal.} For rows from top to bottom, the top three rows show three synthetic rainy scenes/LFIs, and the last two rows exhibit two real-world rainy scenes/LFIs. For columns from left to right, rainy LFI and de-rained images (central-view) obtained by the methods~\cite{Li18v,Liu18v,li21}\revised{~\cite{zhang2022}} and our method, respectively, are shown.}
	\label{fig:Video}
	\vspace{-2mm}
\end{figure*}

\renewcommand{\subwidth}{0.0964}
\renewcommand{\ssubwidth}{0.047}
\begin{figure*}[th!]
	\renewcommand{\tabcolsep}{0.8pt}
	\renewcommand\arraystretch{0.8}
	\begin{center}
		\begin{tabular}{cccccccccccccccccccc}
            \multicolumn{2}{c}{\includegraphics[width=\subwidth\linewidth]{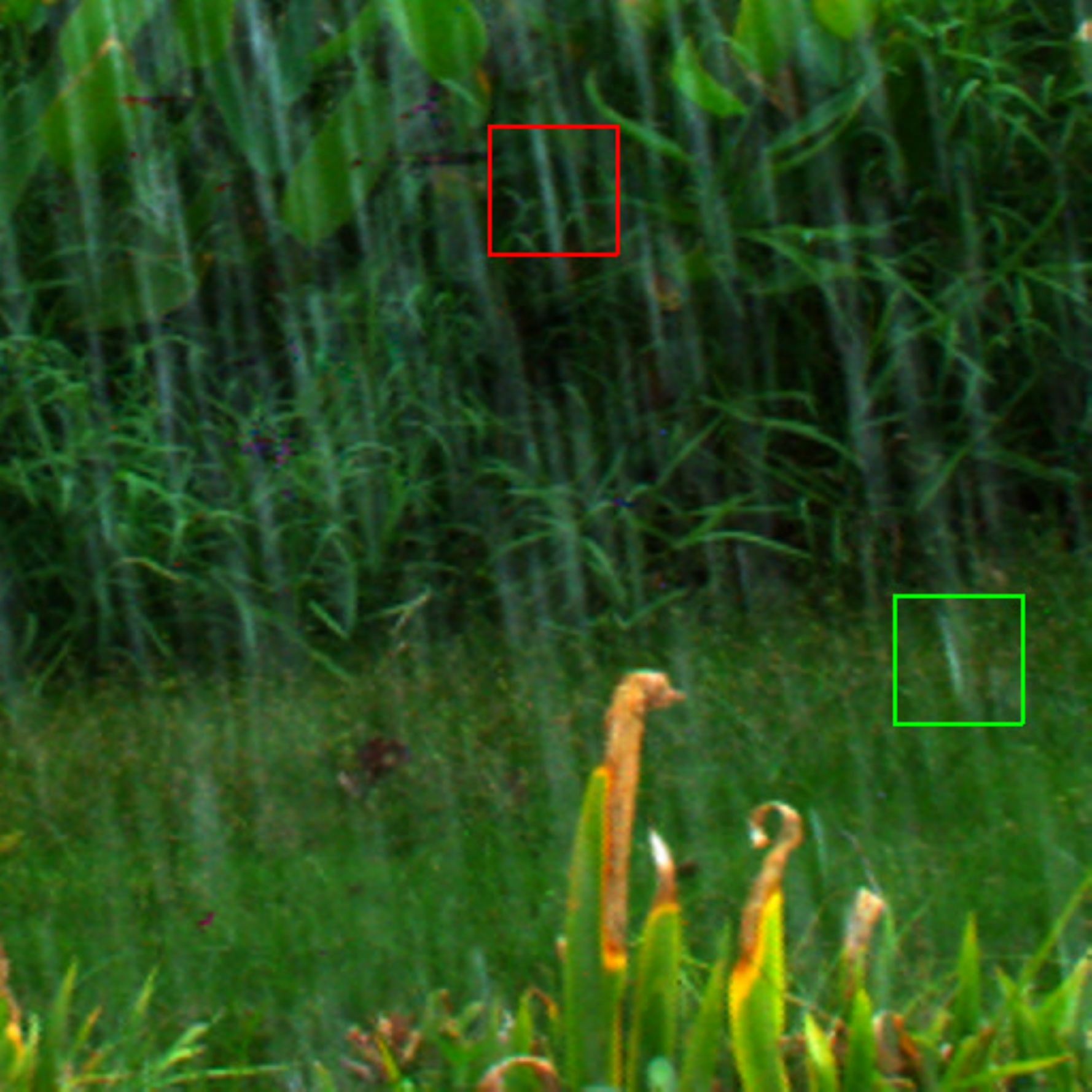}} &
            \multicolumn{2}{c}{\includegraphics[width=\subwidth\linewidth]{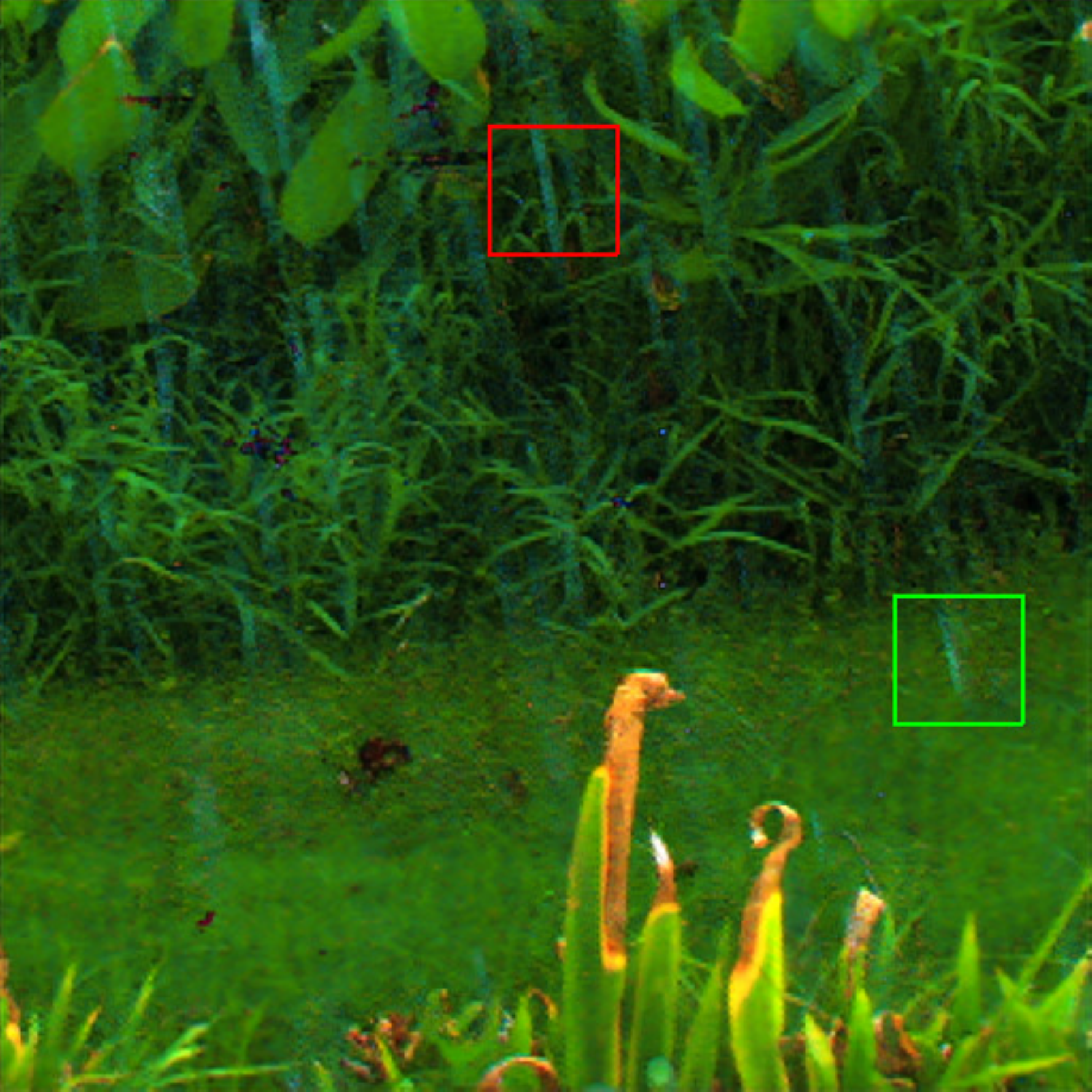}} &
            \multicolumn{2}{c}{\includegraphics[width=\subwidth\linewidth]{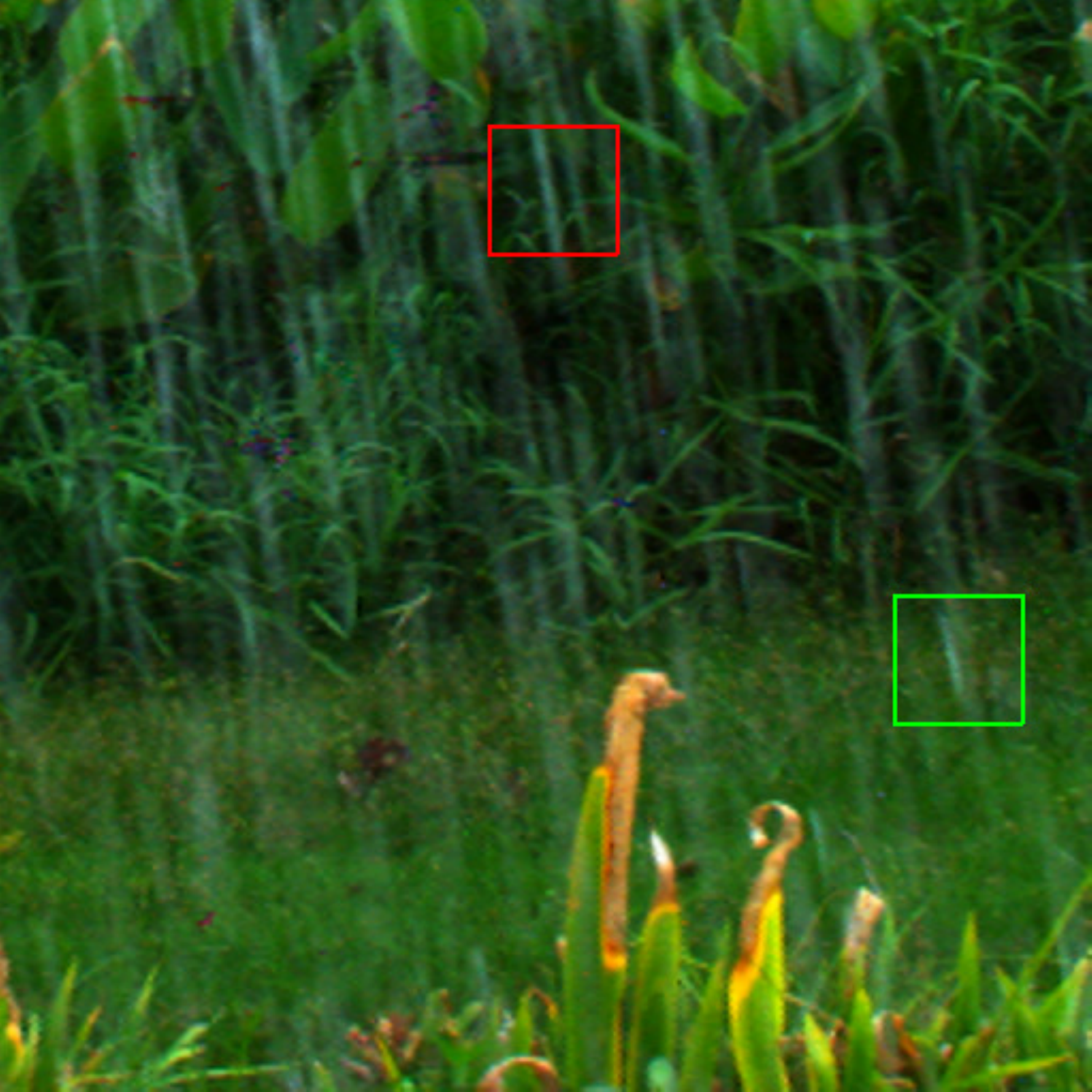}} &
            \multicolumn{2}{c}{\includegraphics[width=\subwidth\linewidth]{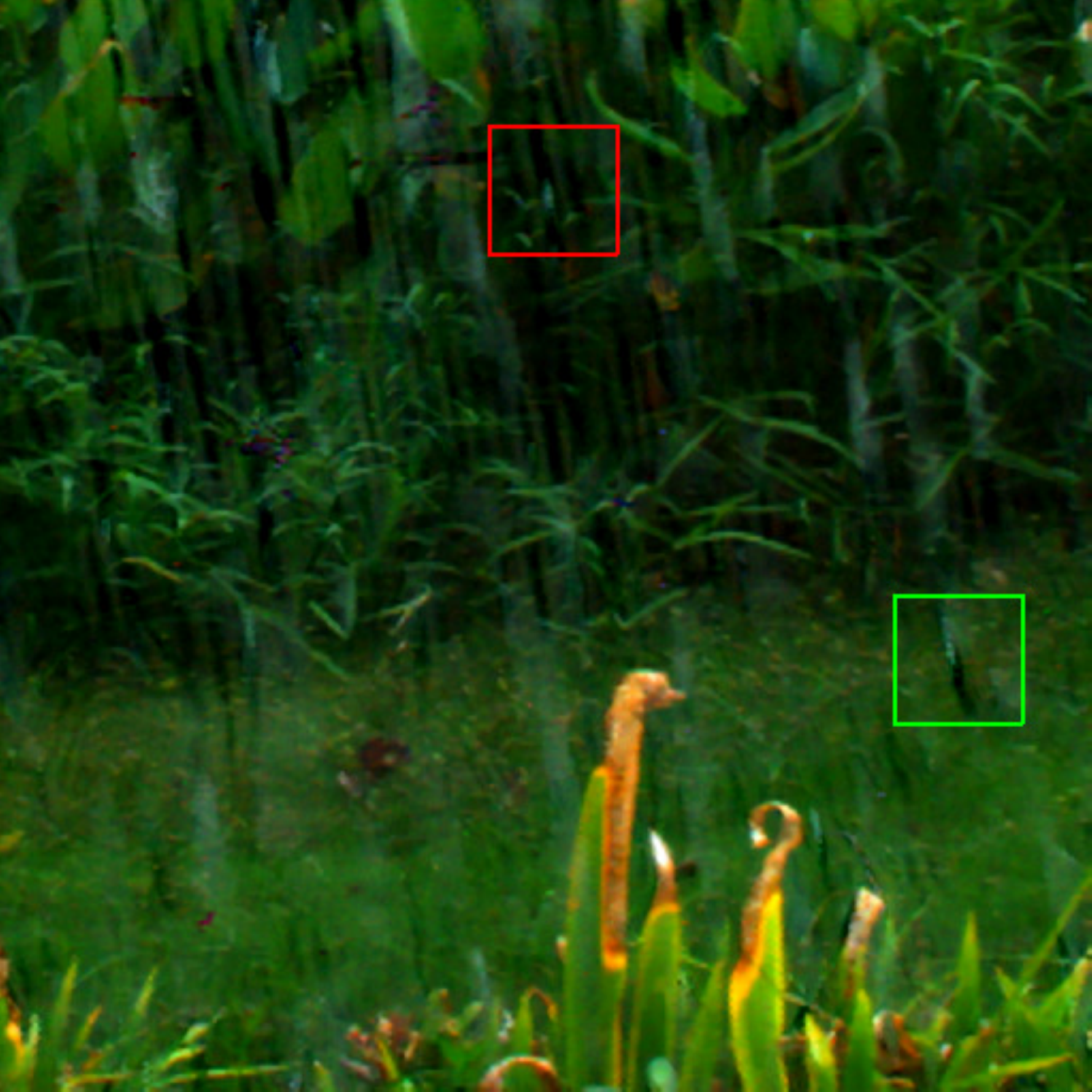}} &
            \multicolumn{2}{c}{\includegraphics[width=\subwidth\linewidth]{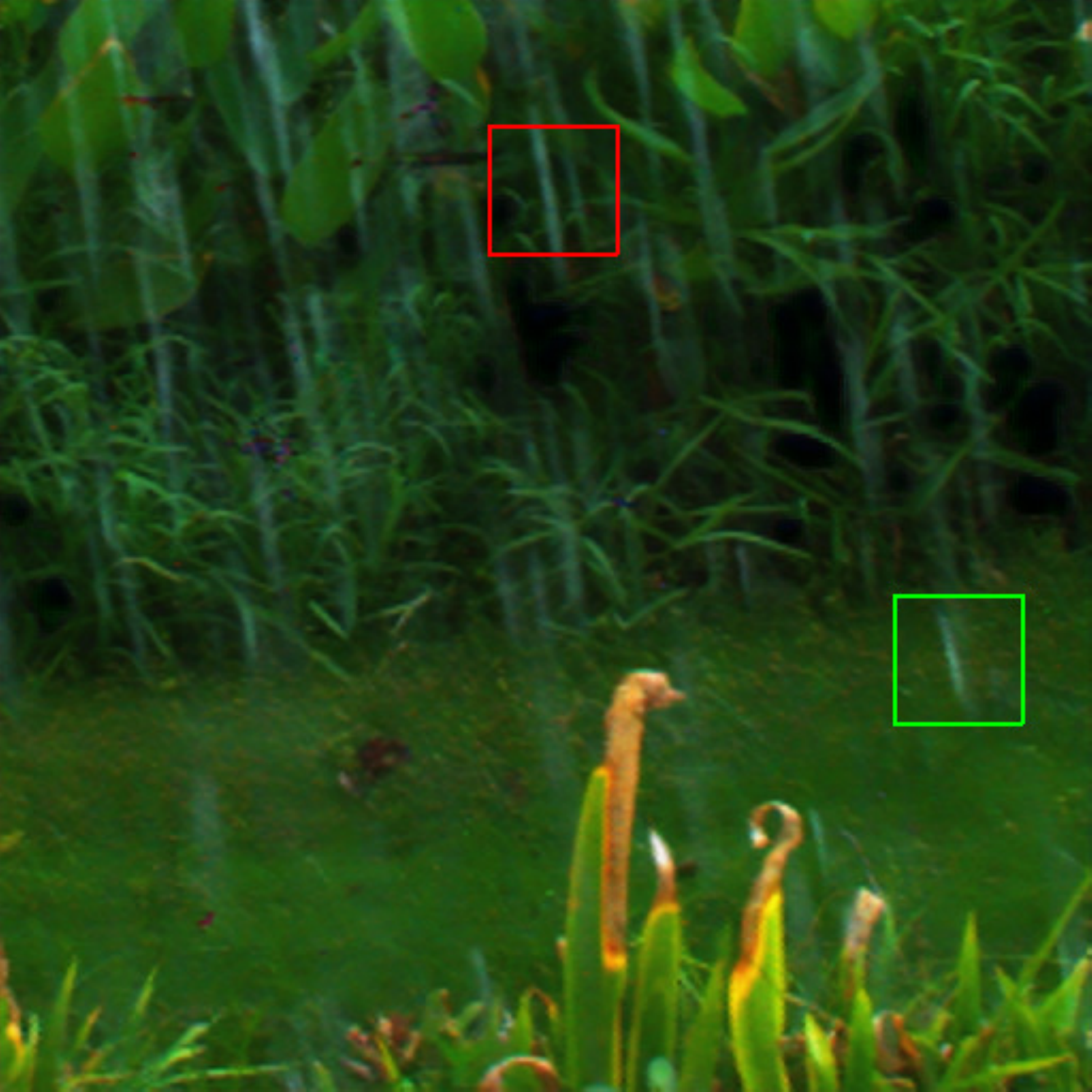}} &
            \multicolumn{2}{c}{\includegraphics[width=\subwidth\linewidth]{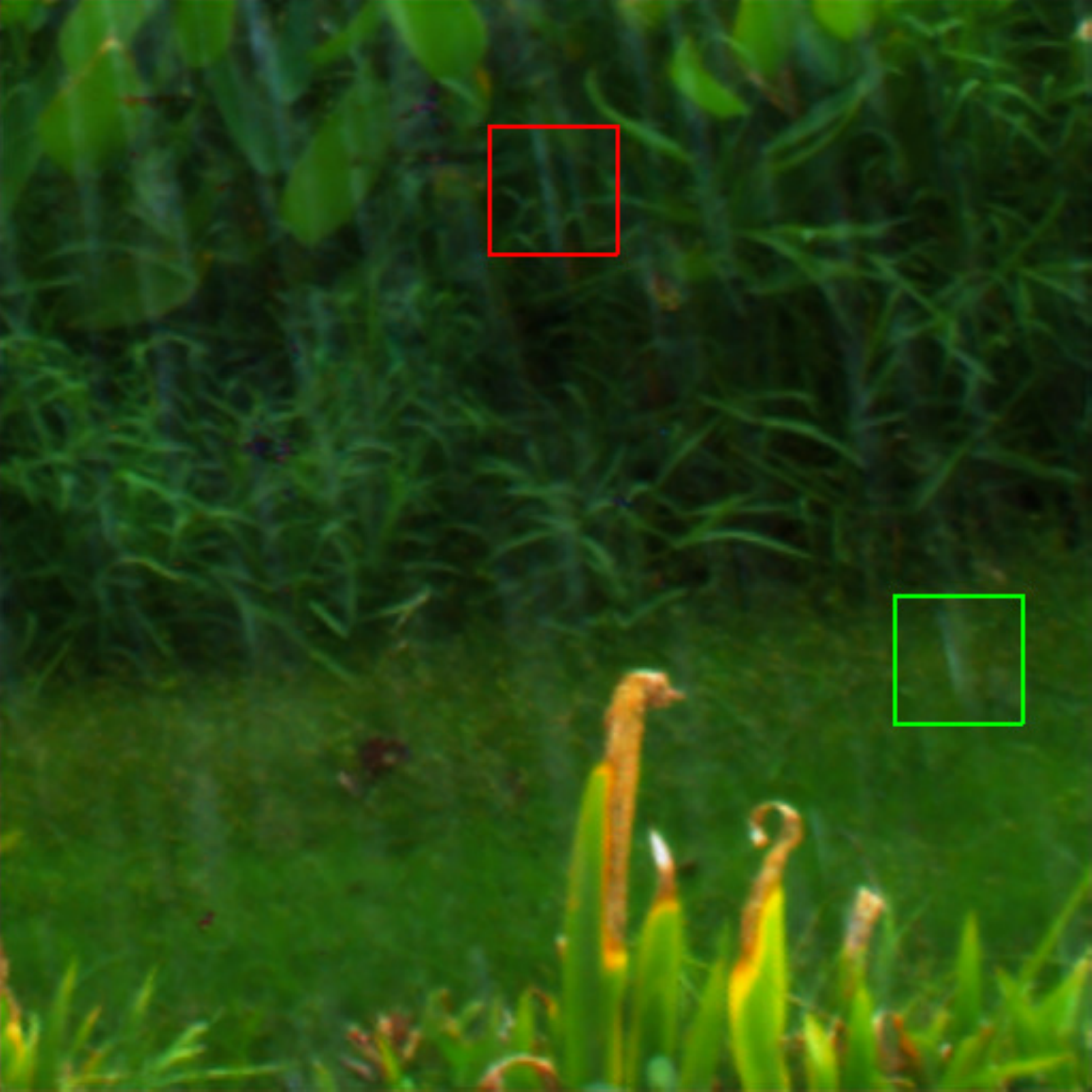}} &
            \multicolumn{2}{c}{\includegraphics[width=\subwidth\linewidth]{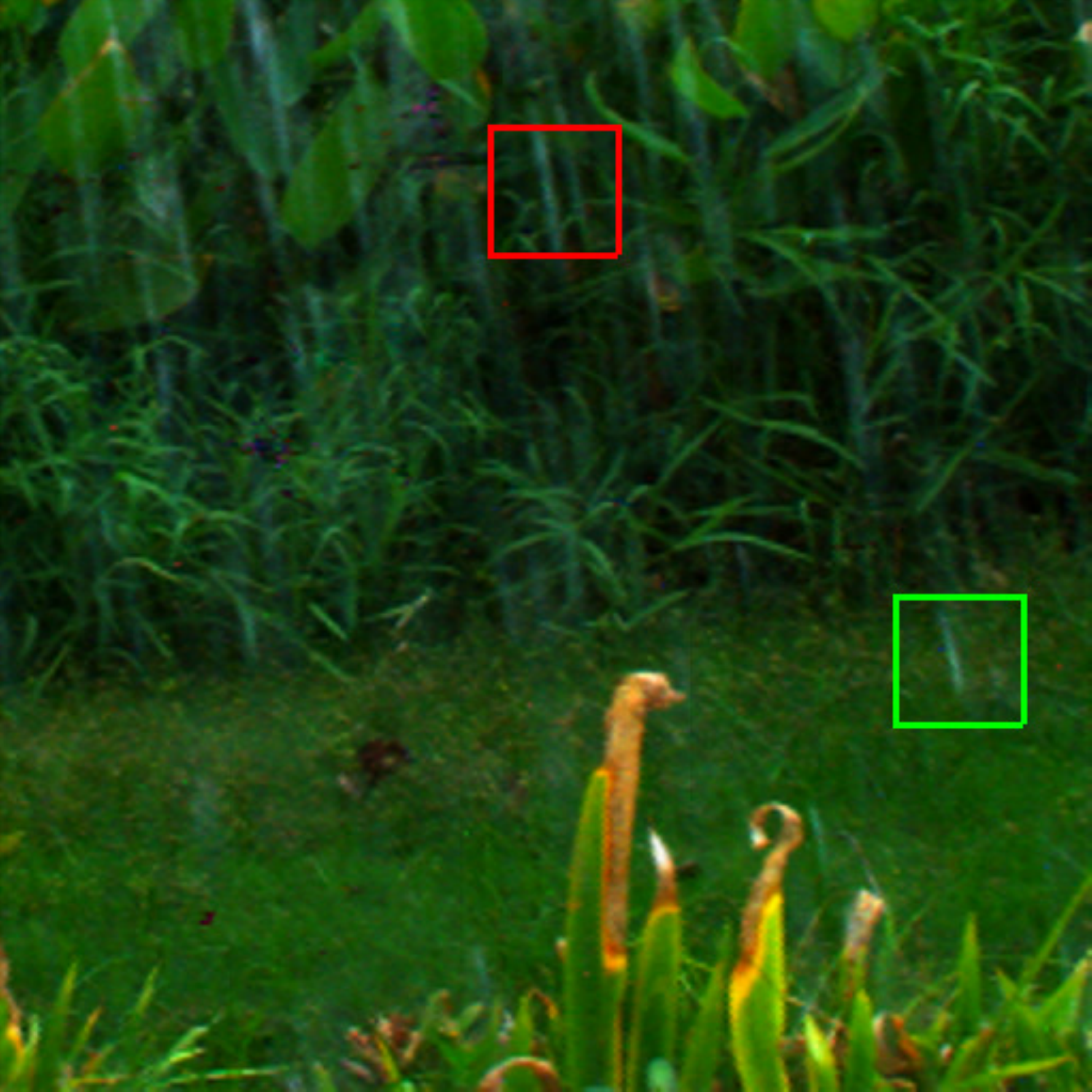}} &
            \multicolumn{2}{c}{\includegraphics[width=\subwidth\linewidth]{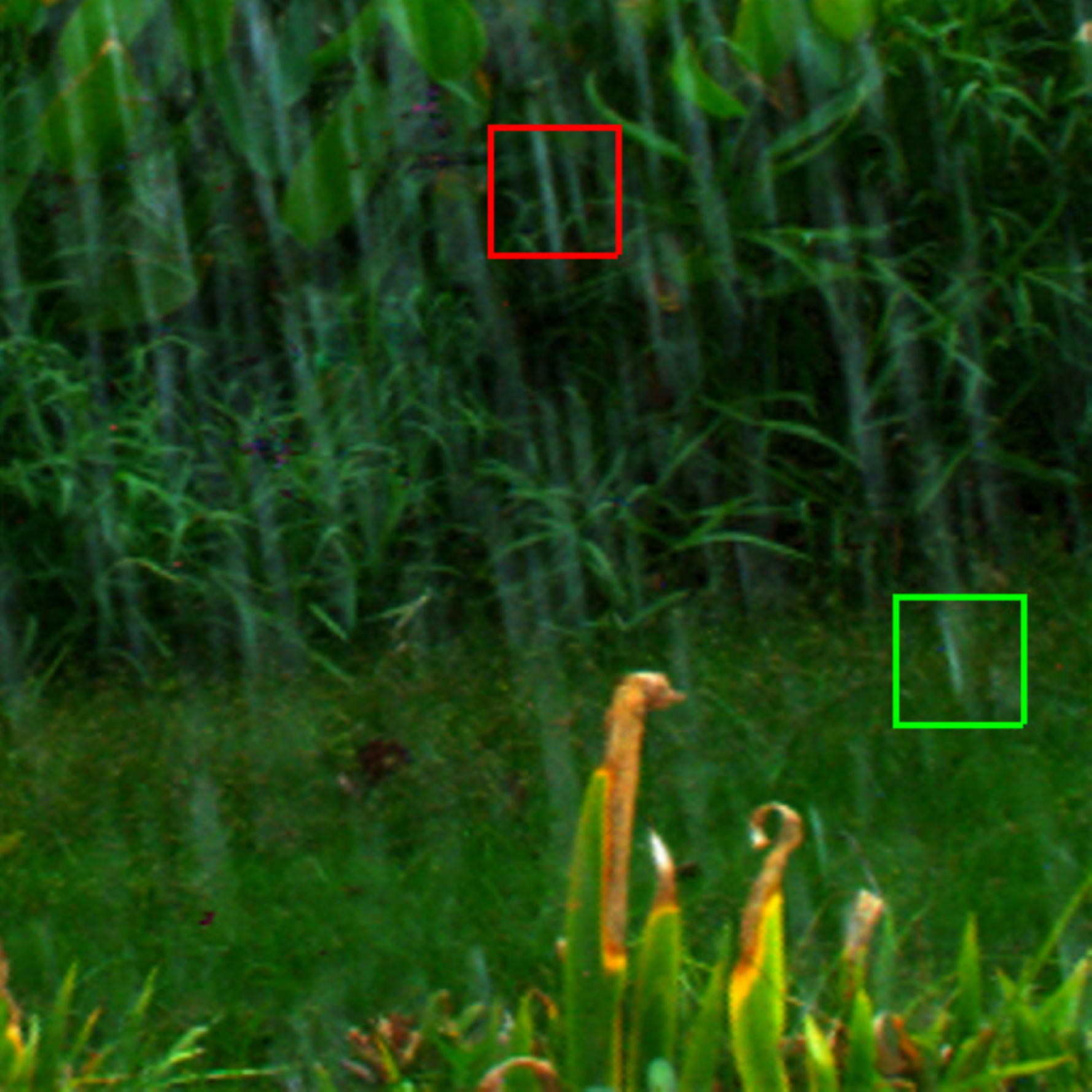}} &
            \multicolumn{2}{c}{\includegraphics[width=\subwidth\linewidth]{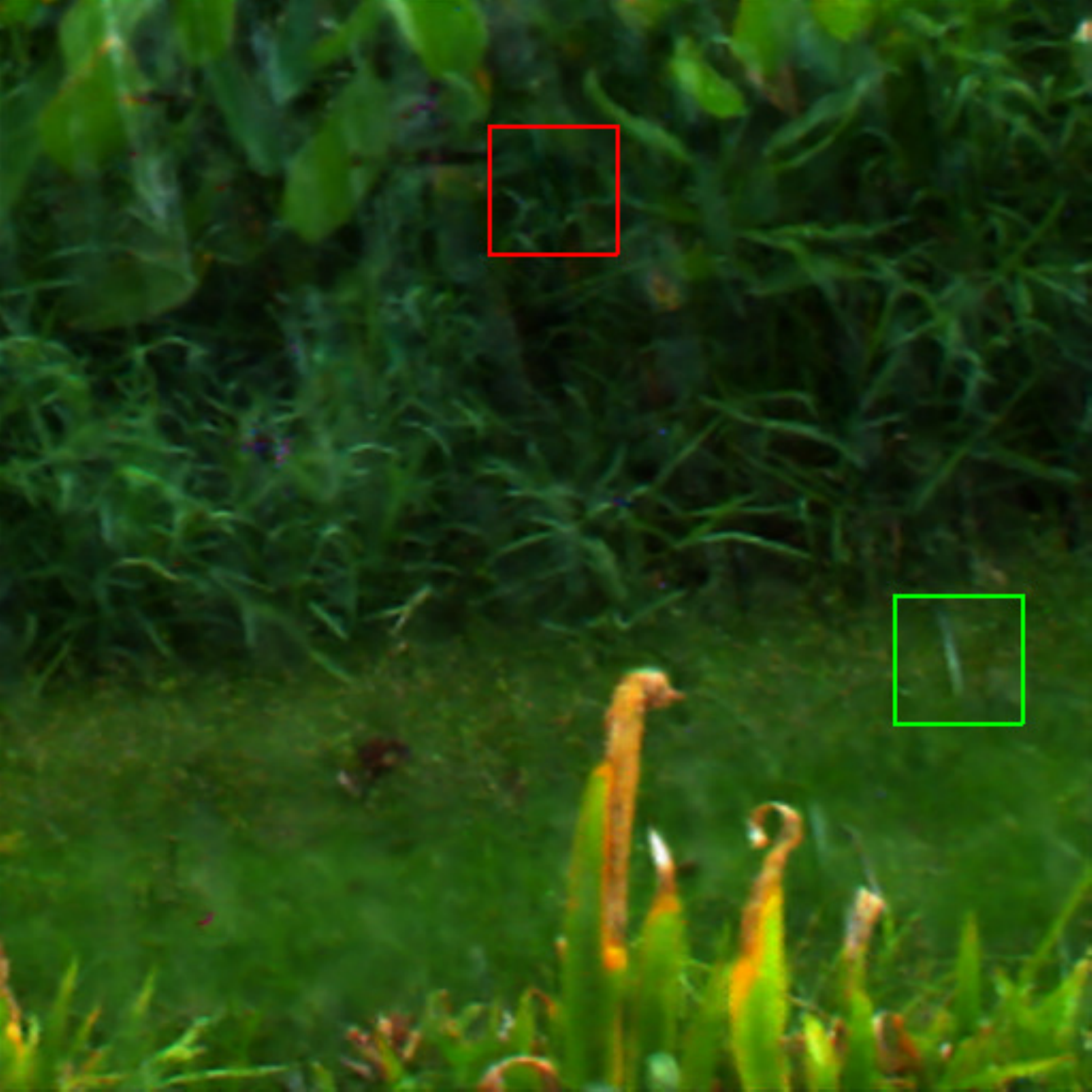}} &
            \multicolumn{2}{c}{\includegraphics[width=\subwidth\linewidth]{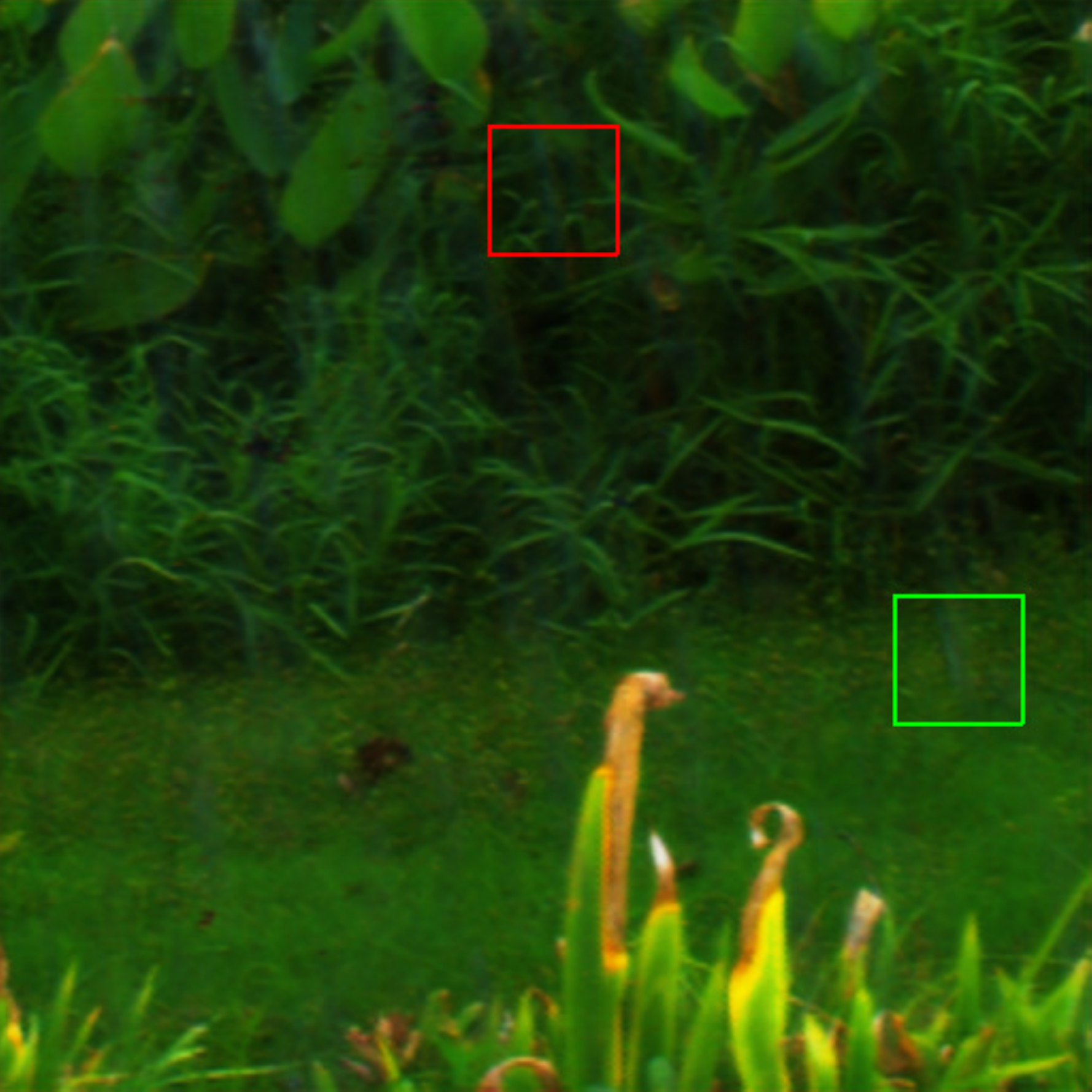}} \\

            \includegraphics[width=\ssubwidth\linewidth]{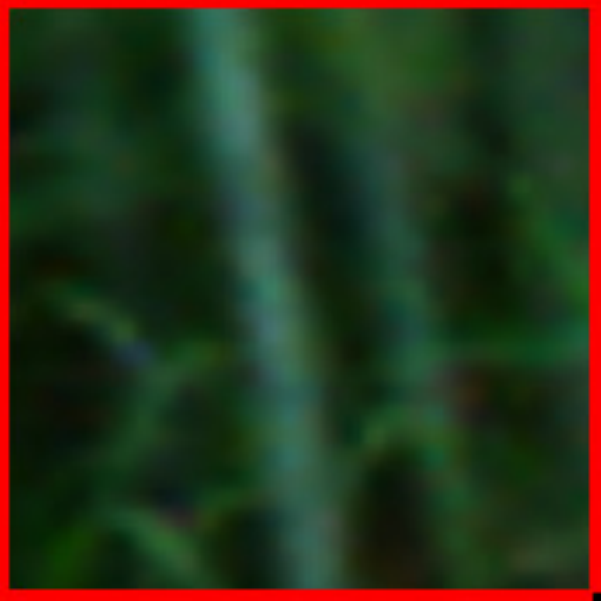} &
            \includegraphics[width=\ssubwidth\linewidth]{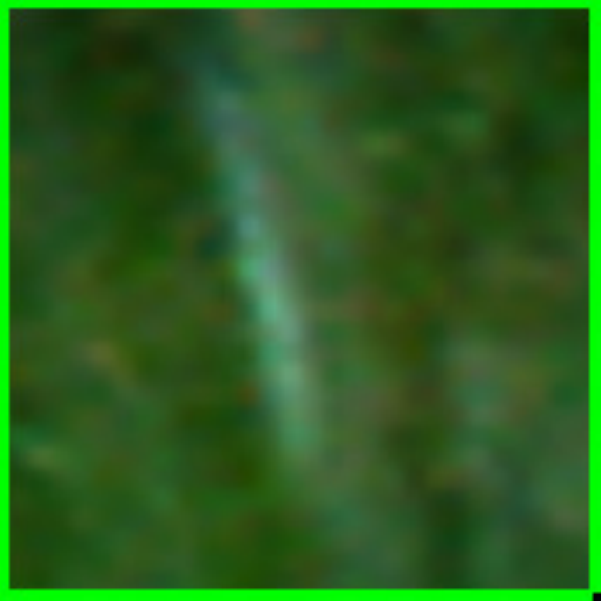} &
            \includegraphics[width=\ssubwidth\linewidth]{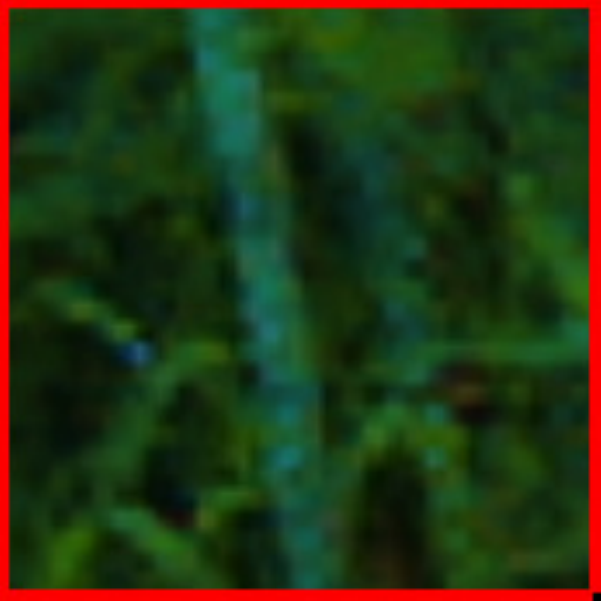} &
            \includegraphics[width=\ssubwidth\linewidth]{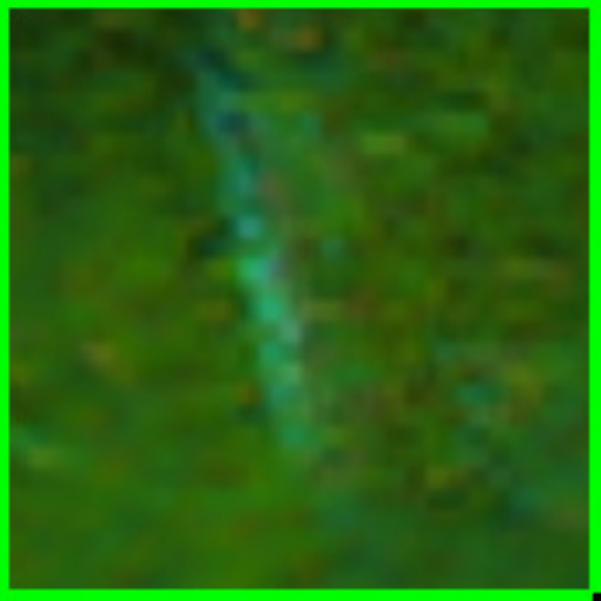} &
            \includegraphics[width=\ssubwidth\linewidth]{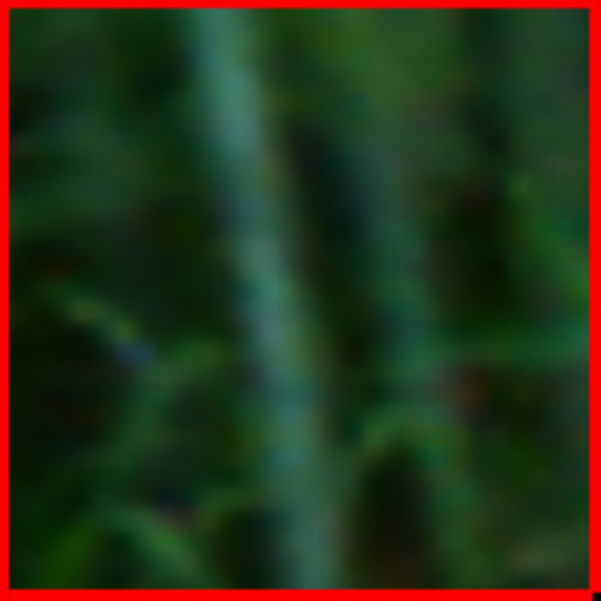} &
            \includegraphics[width=\ssubwidth\linewidth]{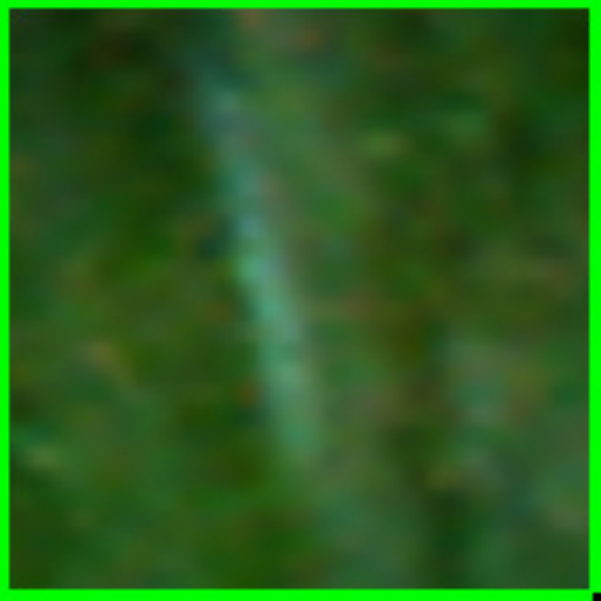} &
            \includegraphics[width=\ssubwidth\linewidth]{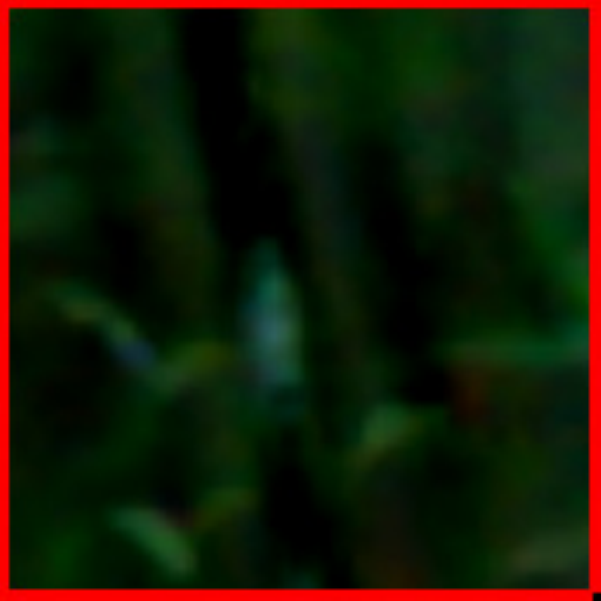} &
            \includegraphics[width=\ssubwidth\linewidth]{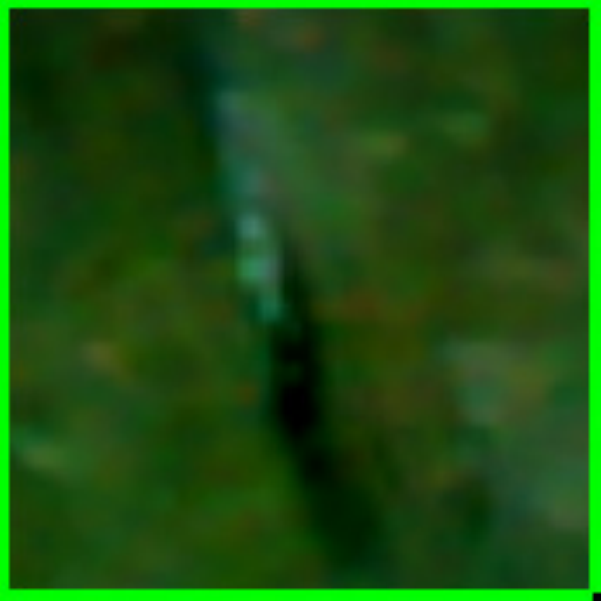} &
            \includegraphics[width=\ssubwidth\linewidth]{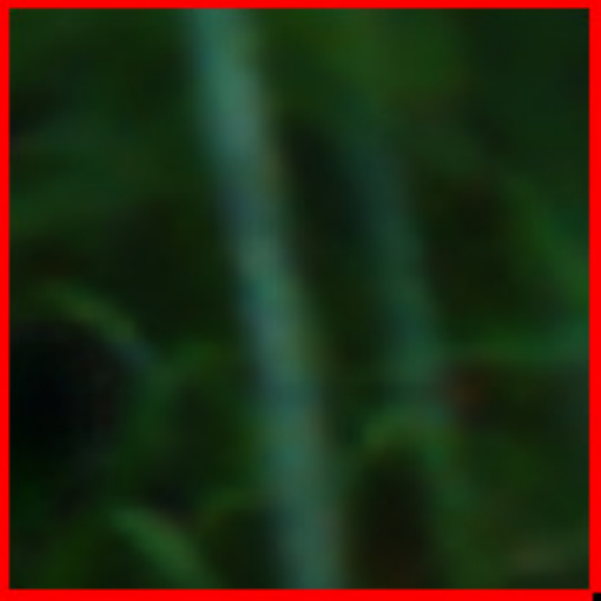} &
            \includegraphics[width=\ssubwidth\linewidth]{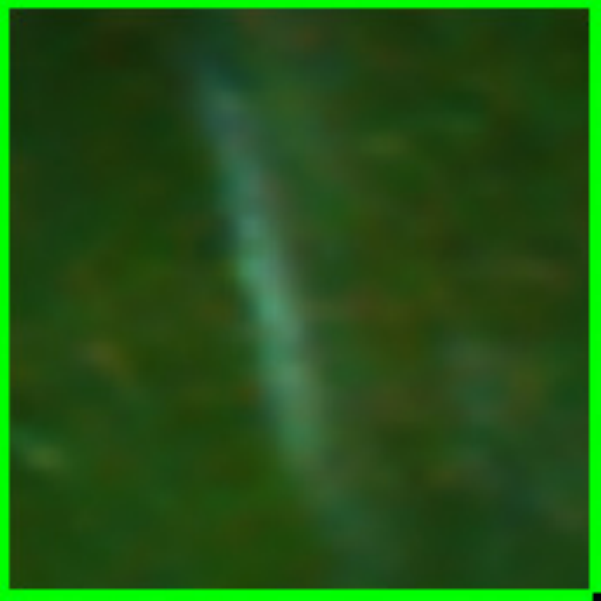} &

            \includegraphics[width=\ssubwidth\linewidth]{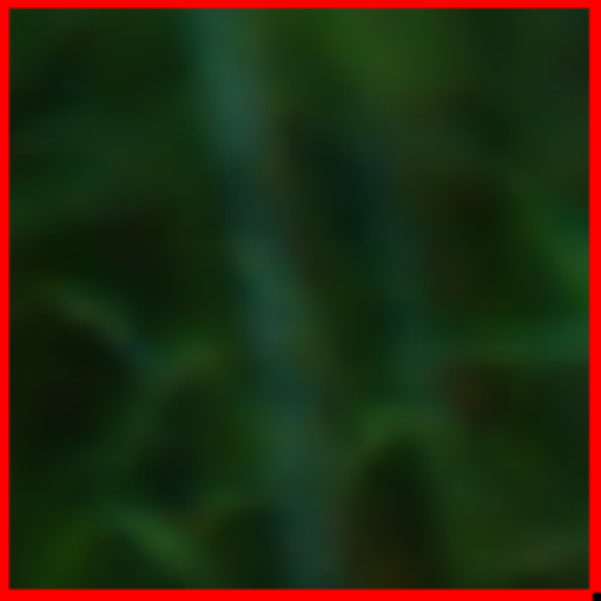} &
            \includegraphics[width=\ssubwidth\linewidth]{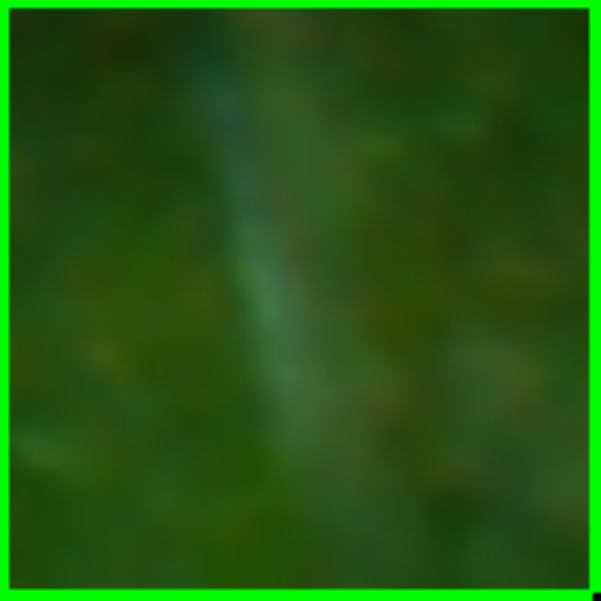} &
            \includegraphics[width=\ssubwidth\linewidth]{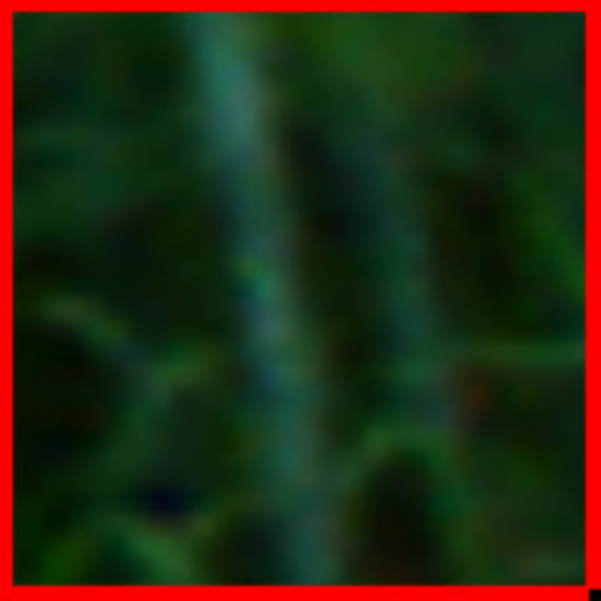} &
            \includegraphics[width=\ssubwidth\linewidth]{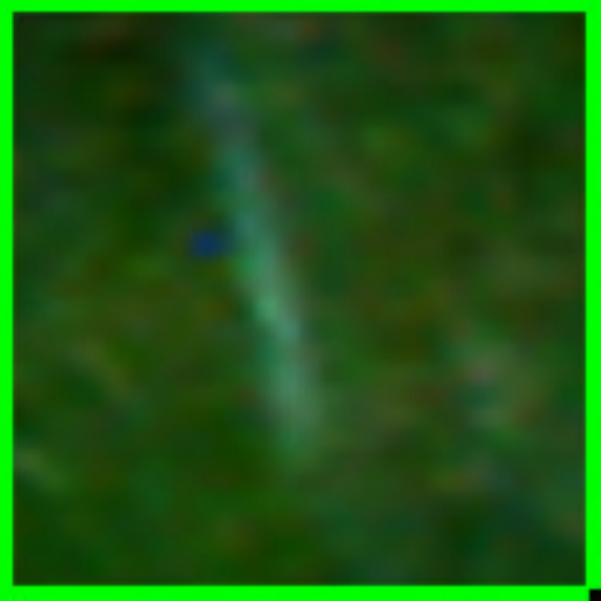} &
            \includegraphics[width=\ssubwidth\linewidth]{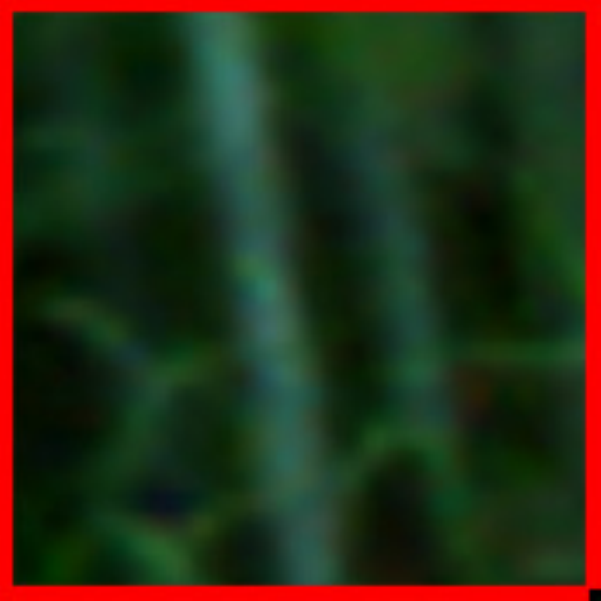} &
            \includegraphics[width=\ssubwidth\linewidth]{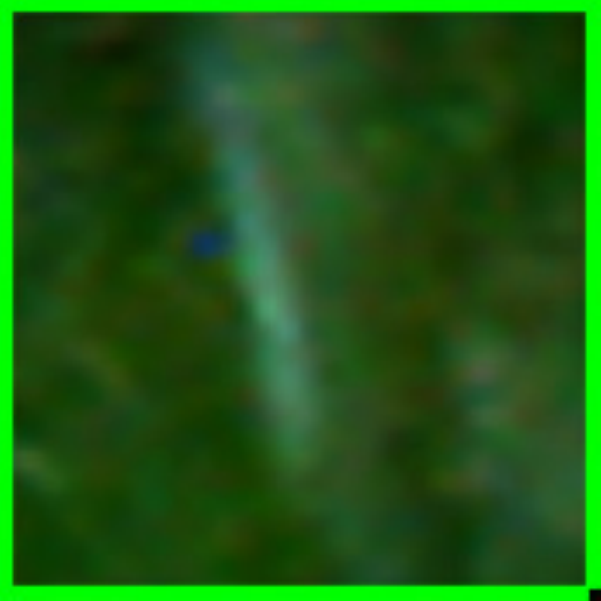} &
            \includegraphics[width=\ssubwidth\linewidth]{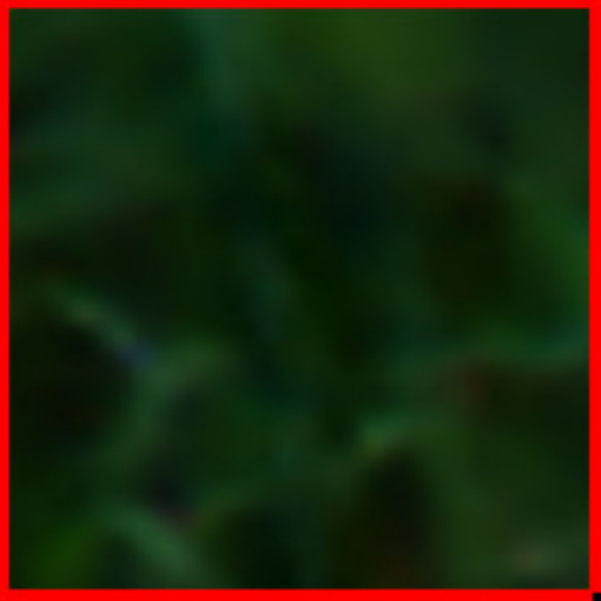} &
            \includegraphics[width=\ssubwidth\linewidth]{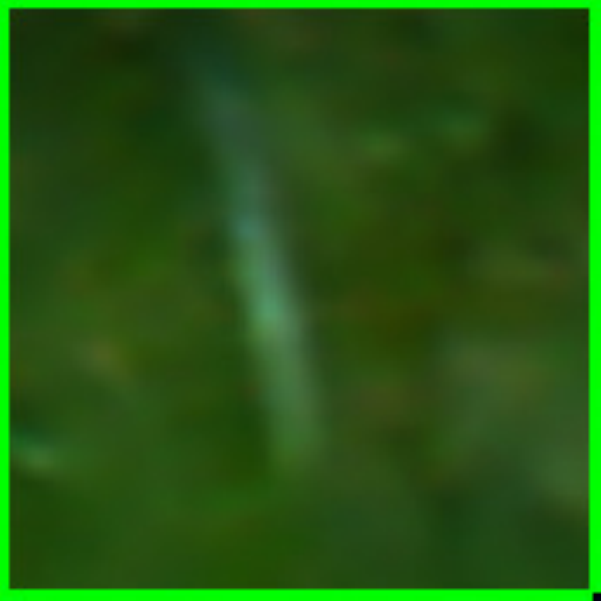} &
            \includegraphics[width=\ssubwidth\linewidth]{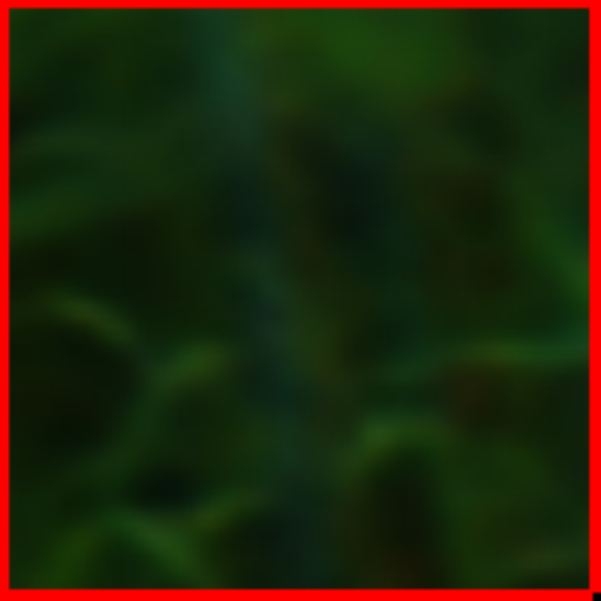} &
            \includegraphics[width=\ssubwidth\linewidth]{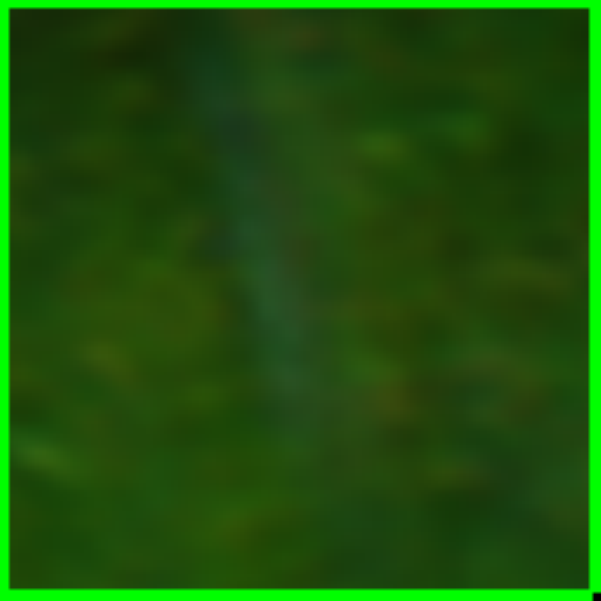} \\

            \multicolumn{2}{c}{\includegraphics[width=\subwidth\linewidth]{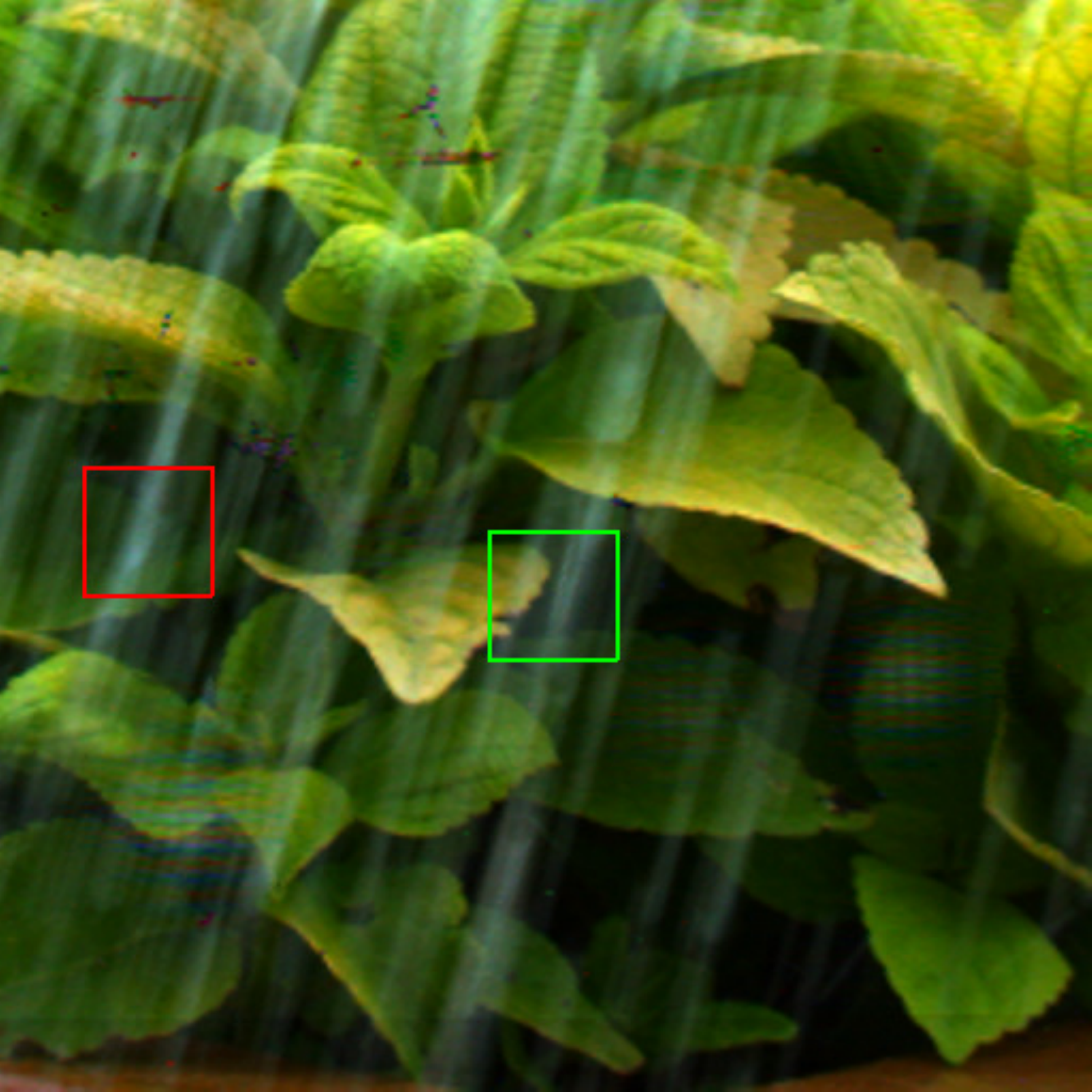}} &
            \multicolumn{2}{c}{\includegraphics[width=\subwidth\linewidth]{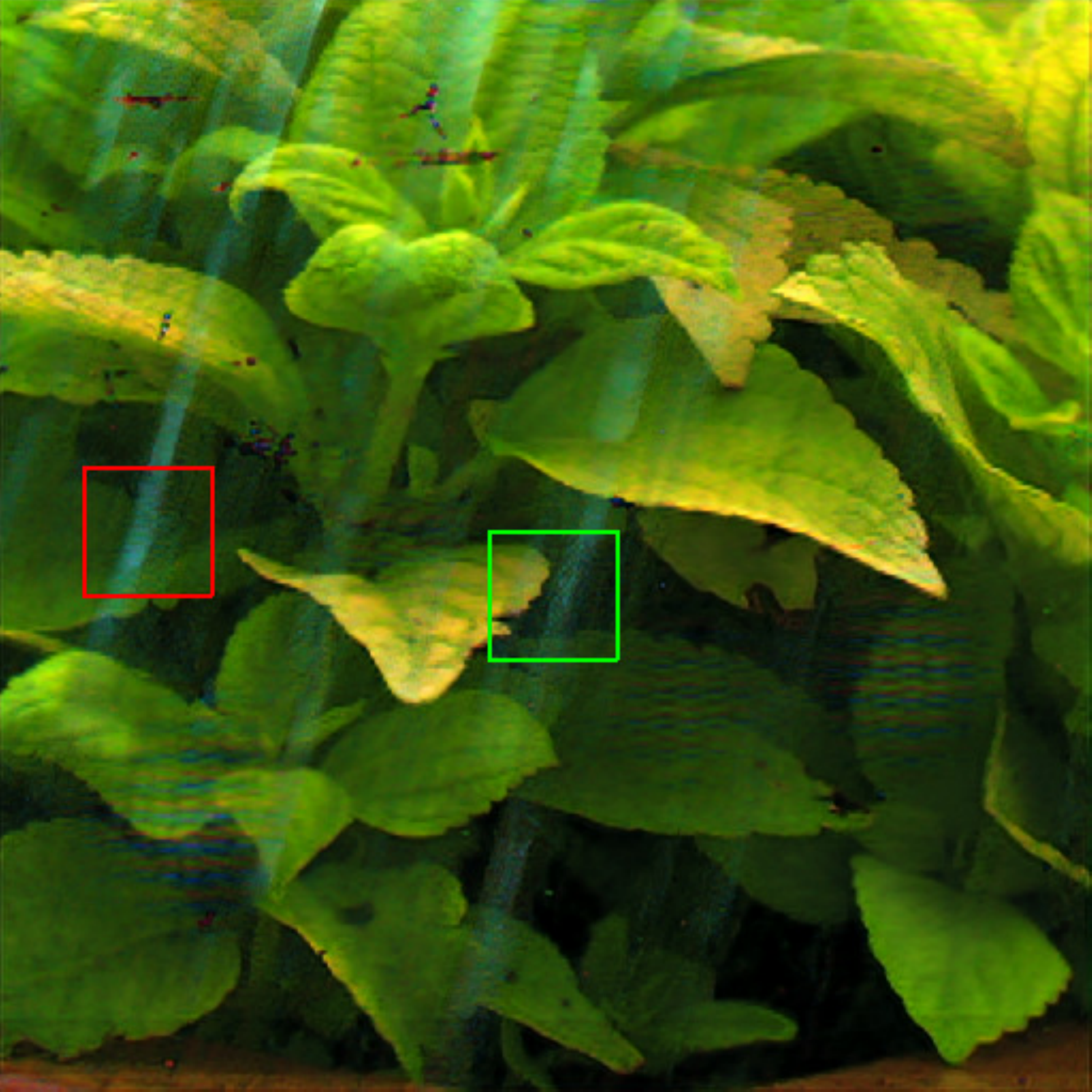}} &
            \multicolumn{2}{c}{\includegraphics[width=\subwidth\linewidth]{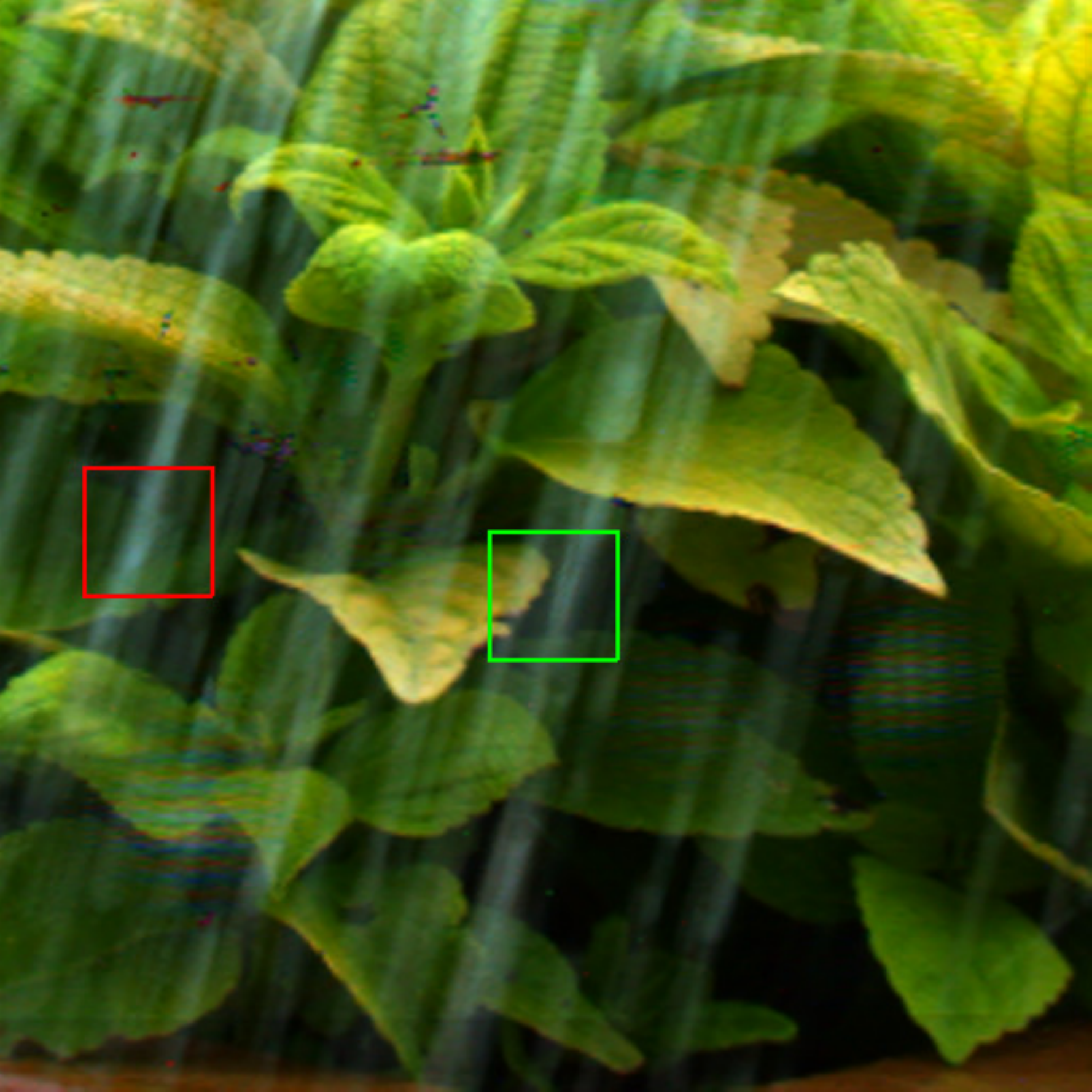}} &
            \multicolumn{2}{c}{\includegraphics[width=\subwidth\linewidth]{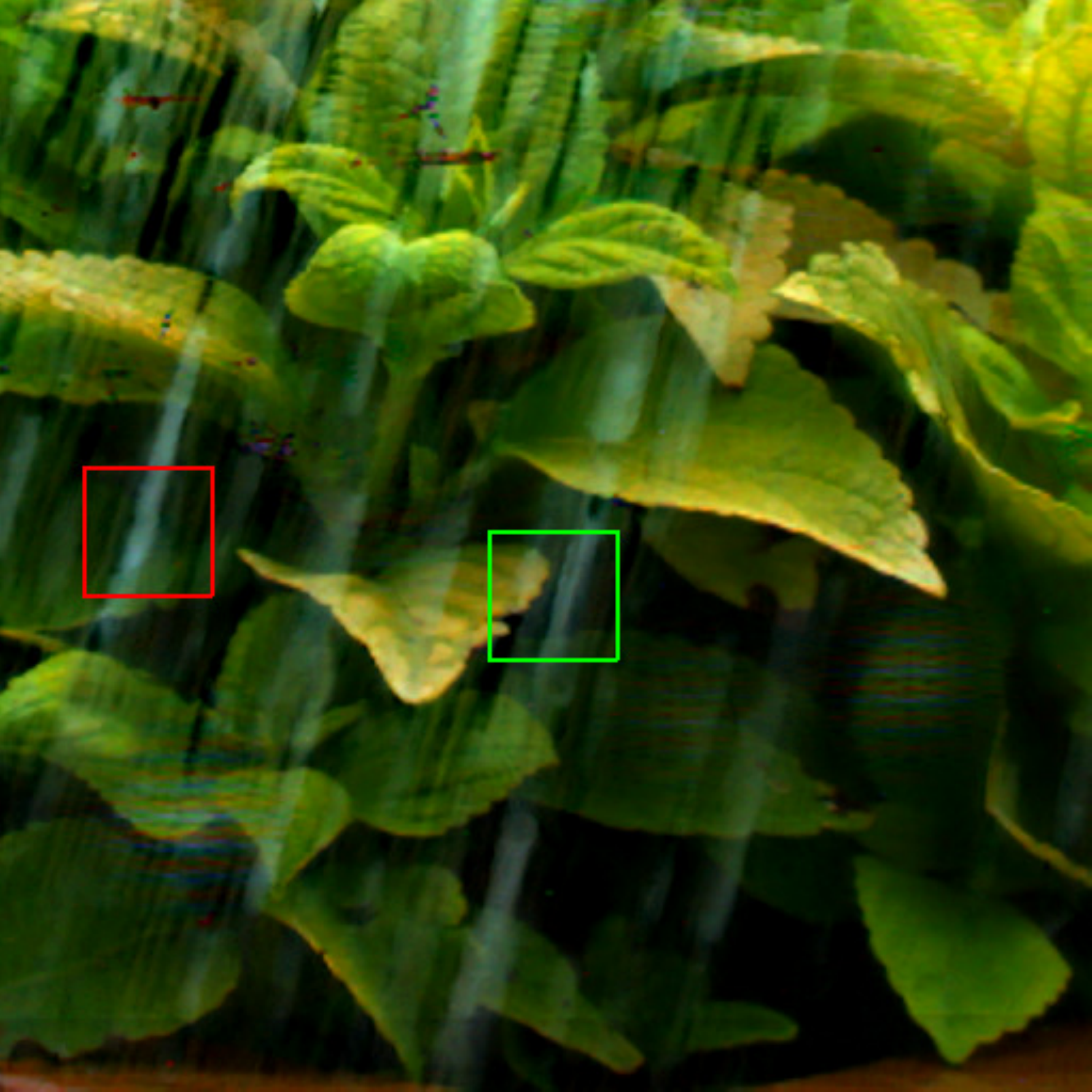}} &
            \multicolumn{2}{c}{\includegraphics[width=\subwidth\linewidth]{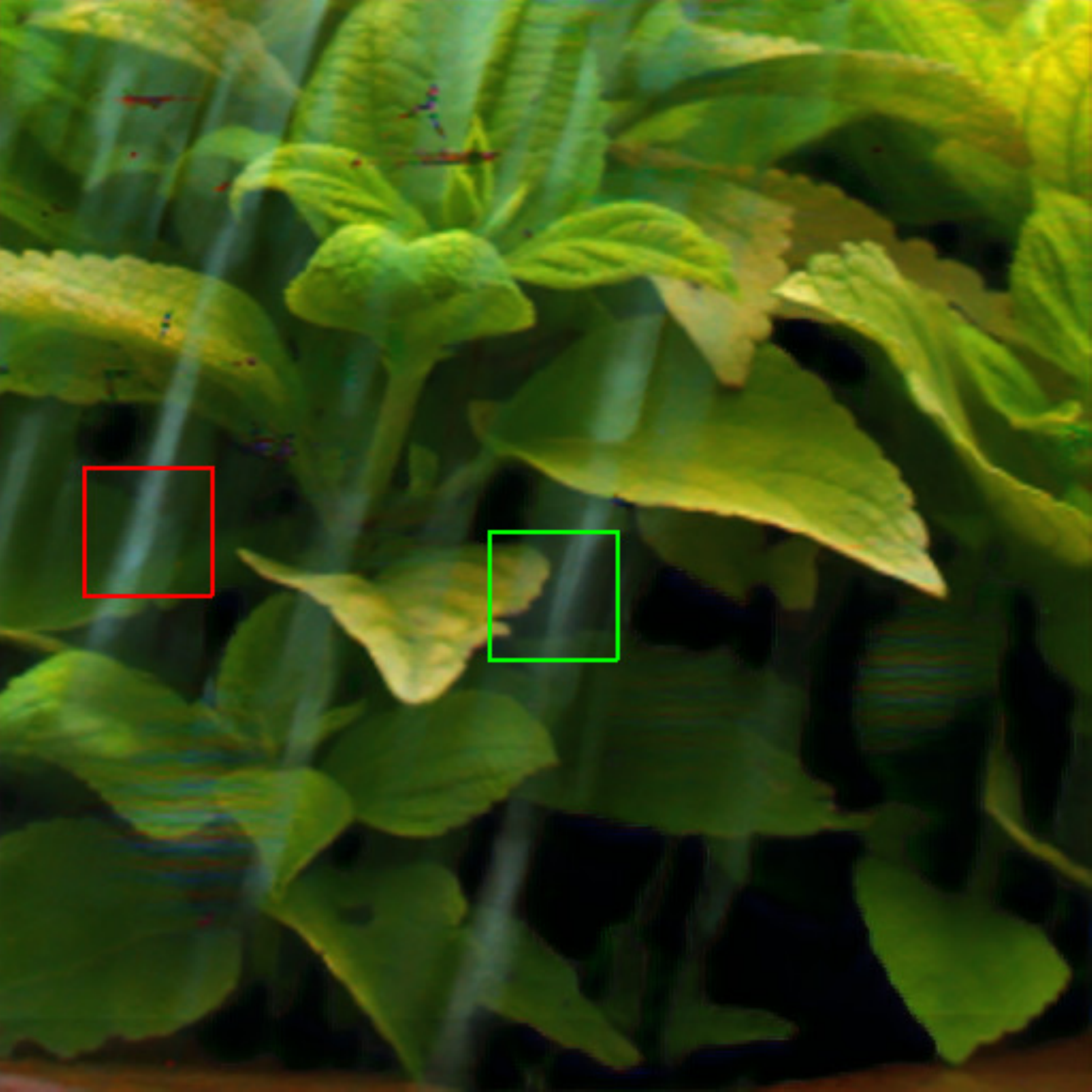}} &

            \multicolumn{2}{c}{\includegraphics[width=\subwidth\linewidth]{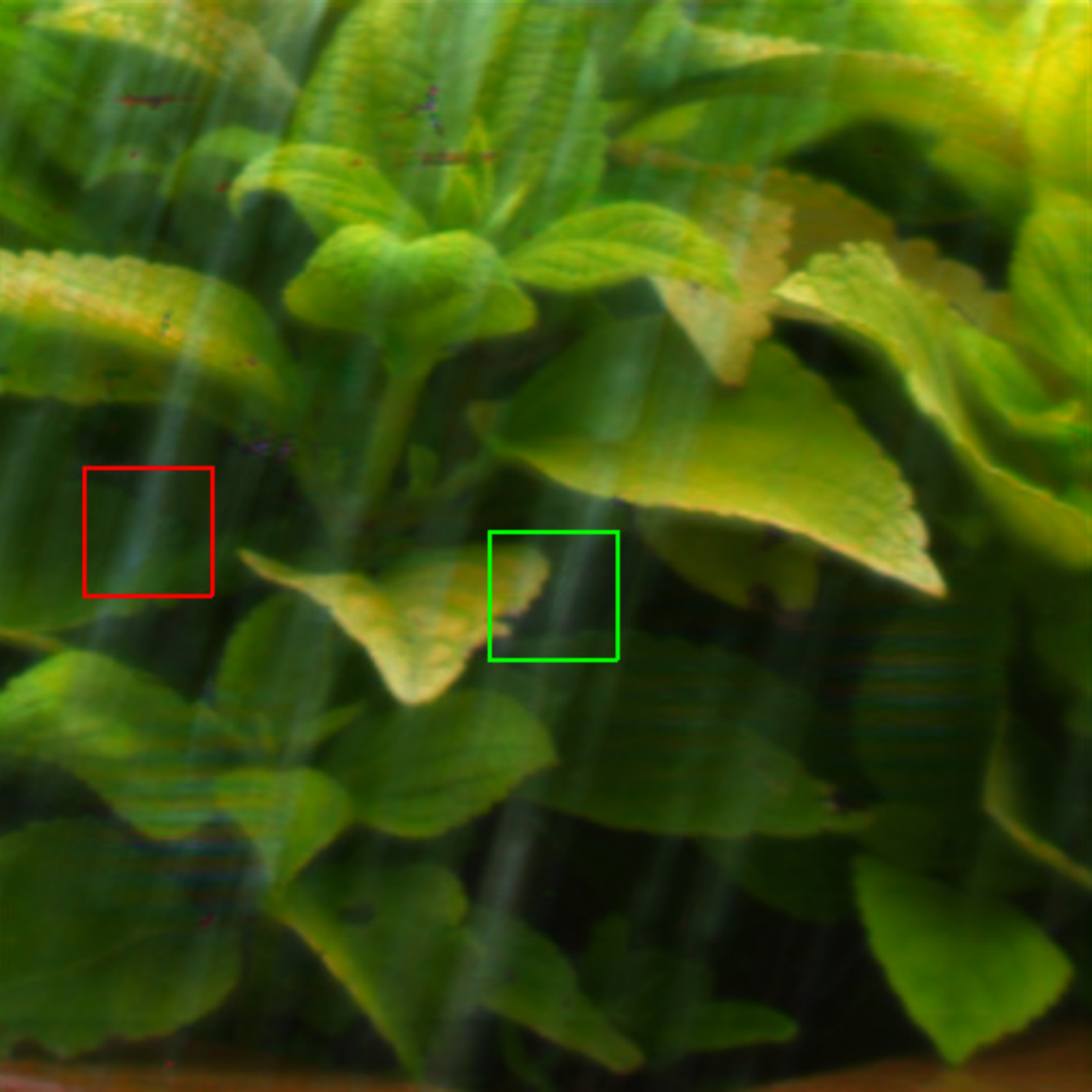}} &
            \multicolumn{2}{c}{\includegraphics[width=\subwidth\linewidth]{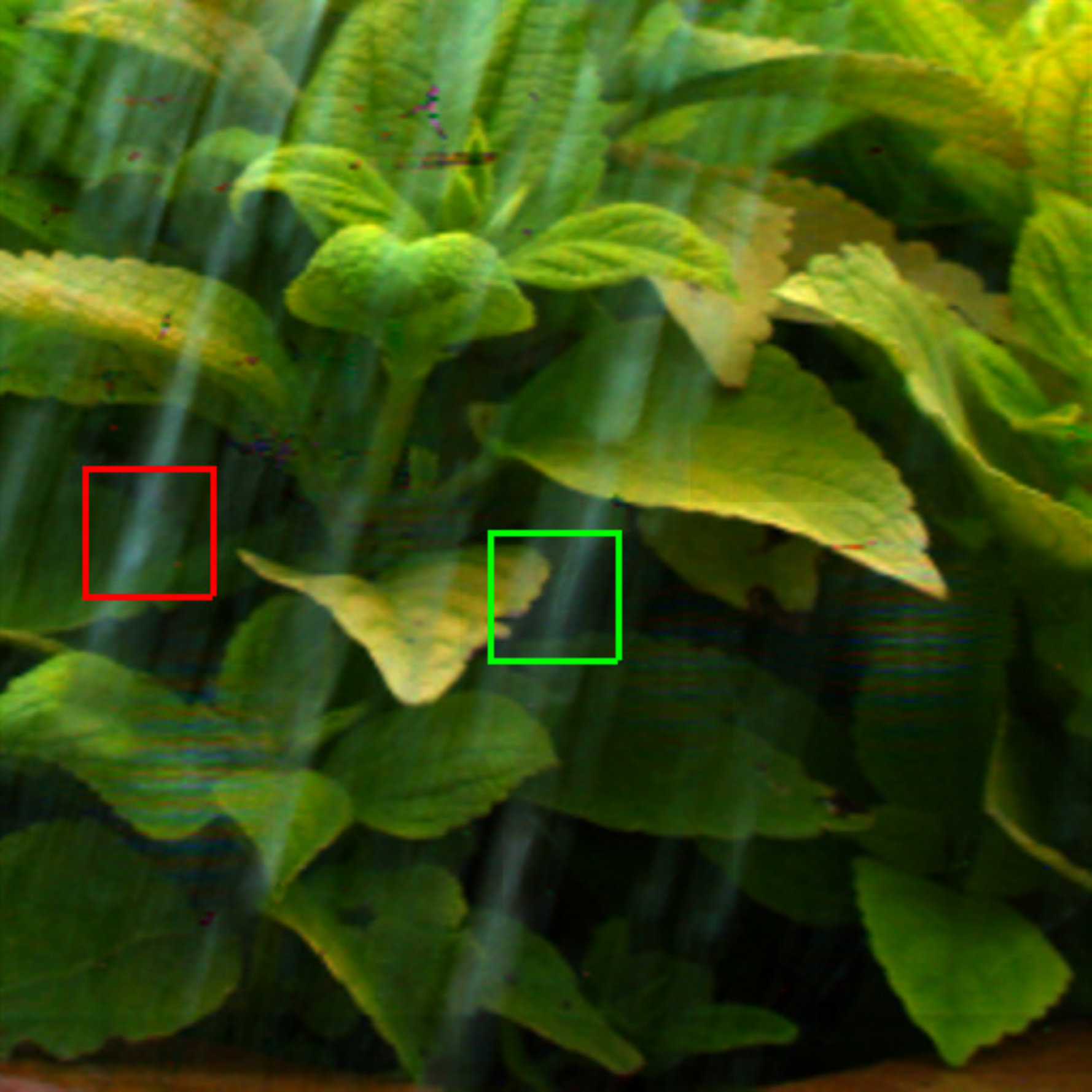}} &
            \multicolumn{2}{c}{\includegraphics[width=\subwidth\linewidth]{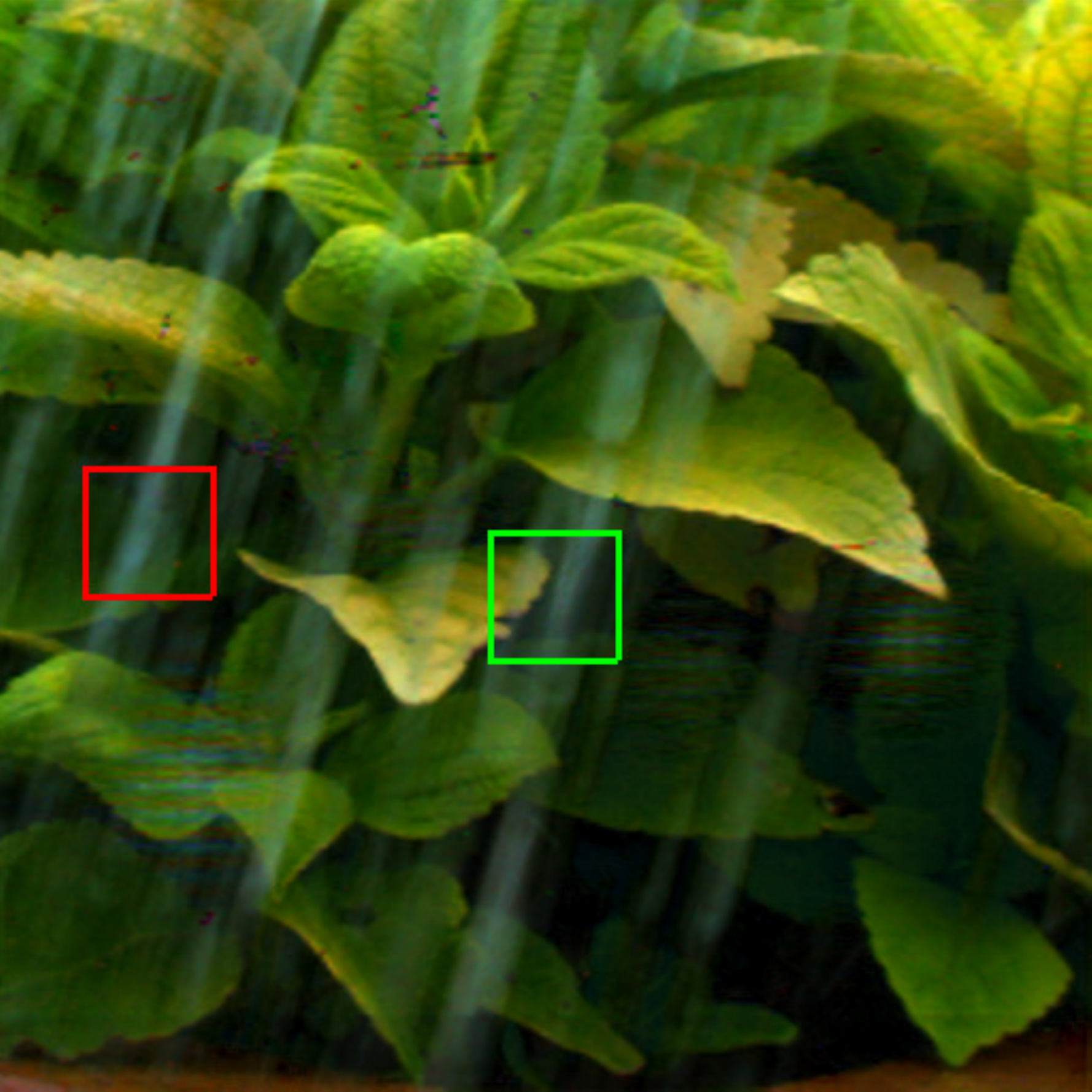}} &
            \multicolumn{2}{c}{\includegraphics[width=\subwidth\linewidth]{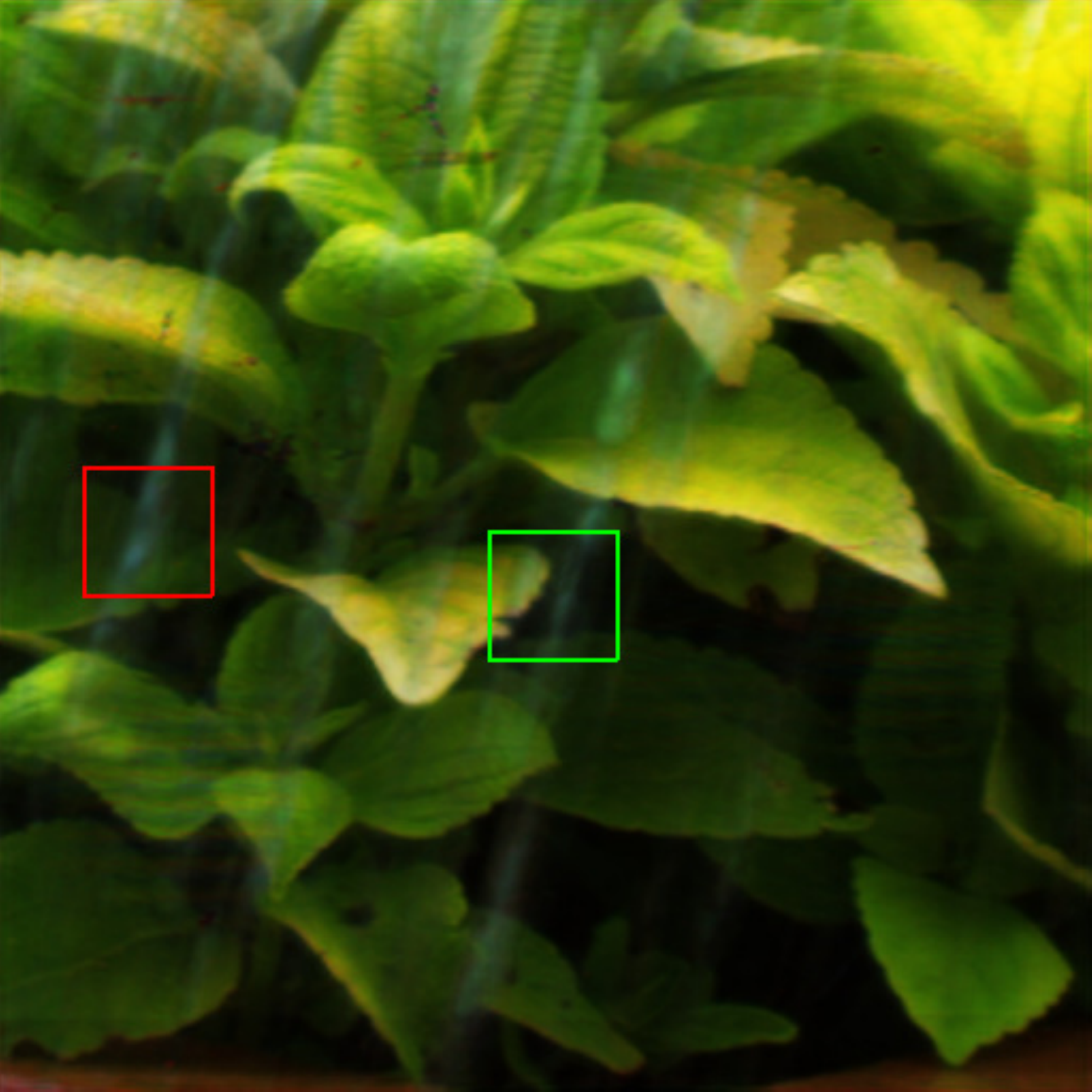}} &
            \multicolumn{2}{c}{\includegraphics[width=\subwidth\linewidth]{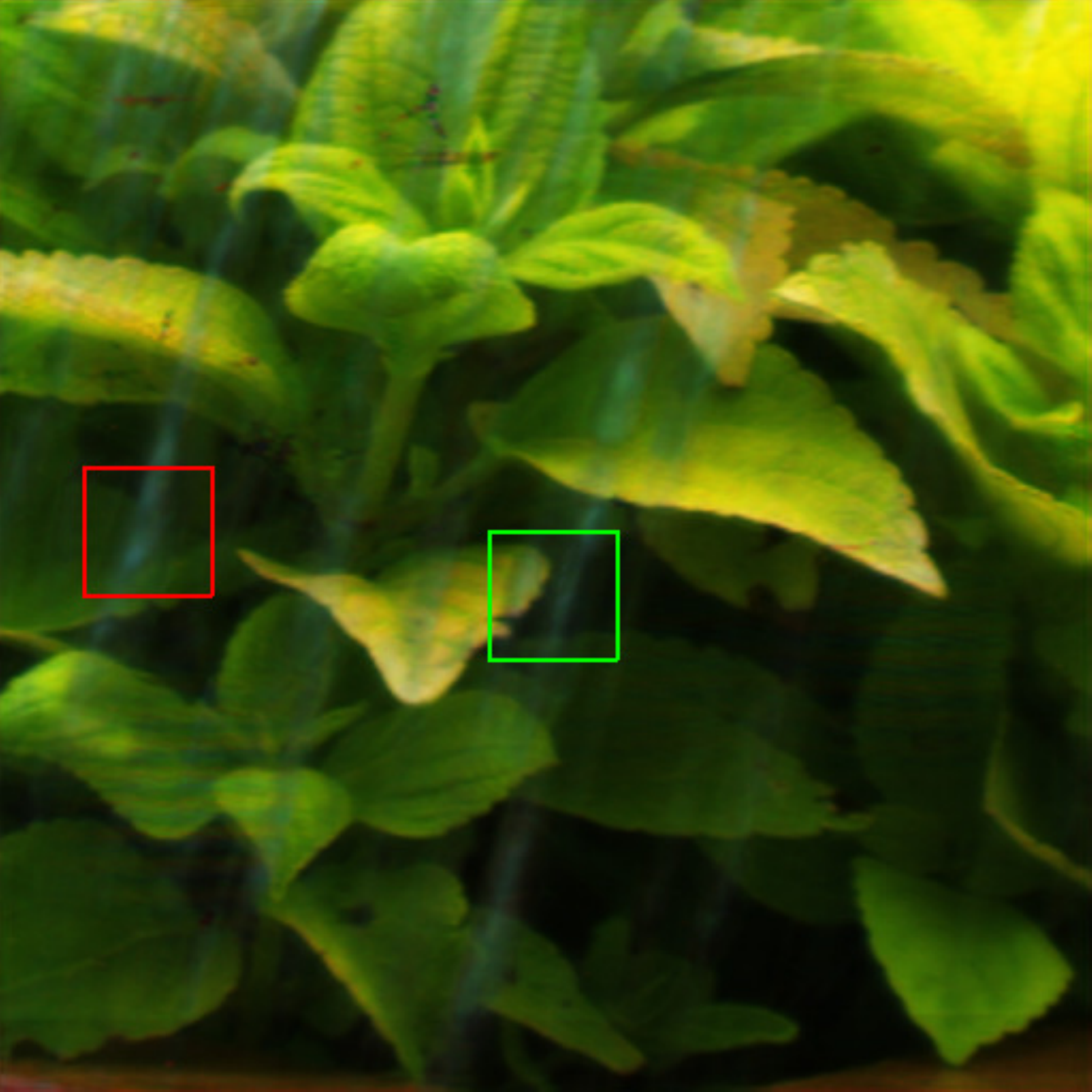}} \\

            \includegraphics[width=\ssubwidth\linewidth]{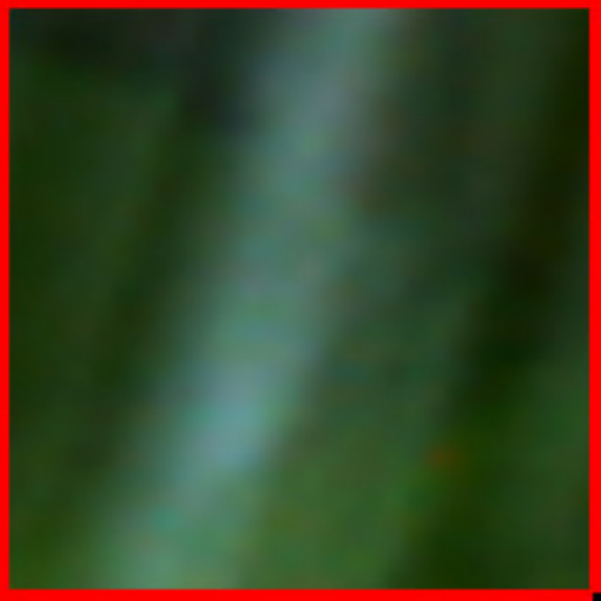} &
            \includegraphics[width=\ssubwidth\linewidth]{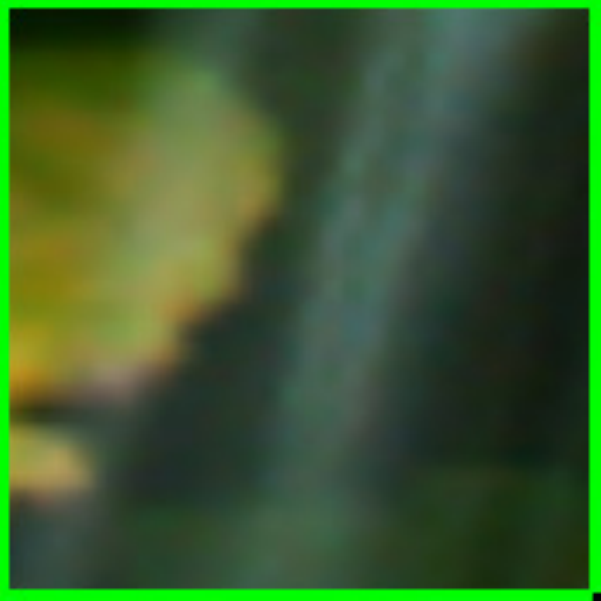} &
            \includegraphics[width=\ssubwidth\linewidth]{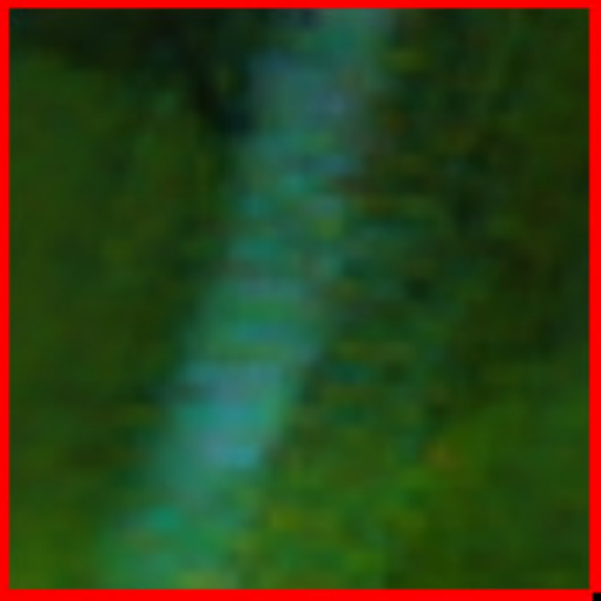} &
            \includegraphics[width=\ssubwidth\linewidth]{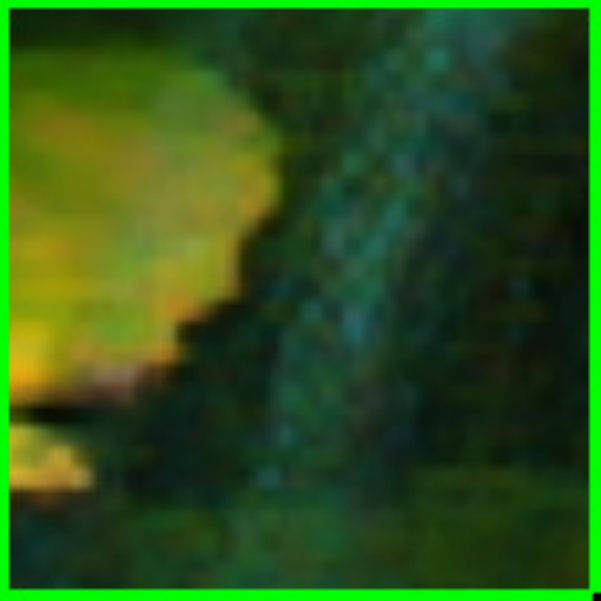} &
            \includegraphics[width=\ssubwidth\linewidth]{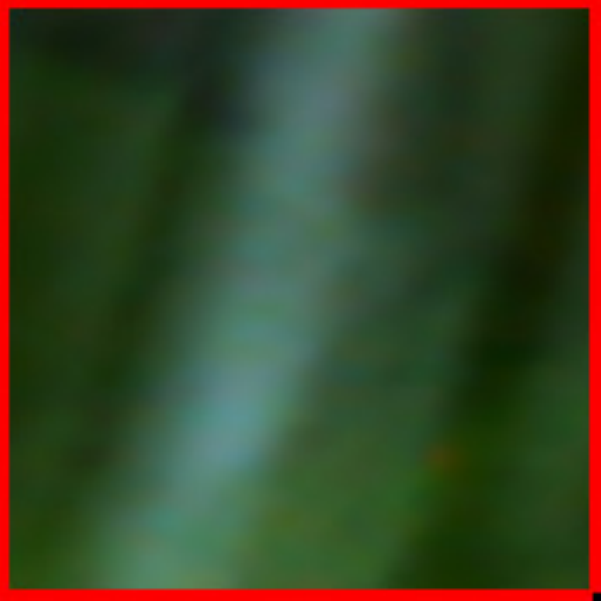} &
            \includegraphics[width=\ssubwidth\linewidth]{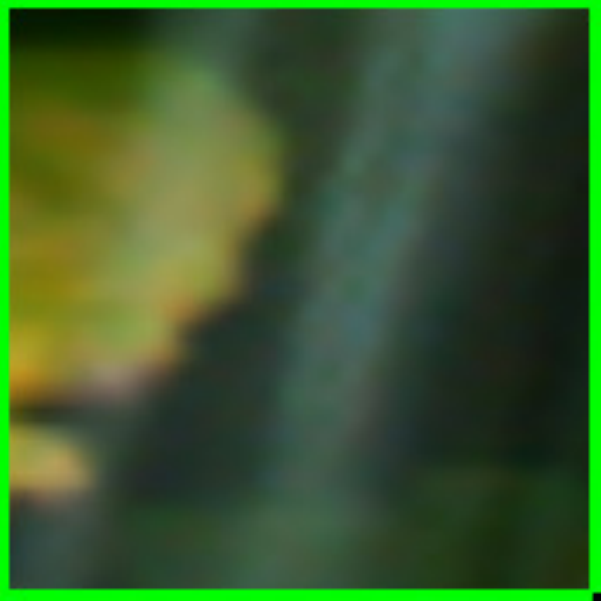} &
            \includegraphics[width=\ssubwidth\linewidth]{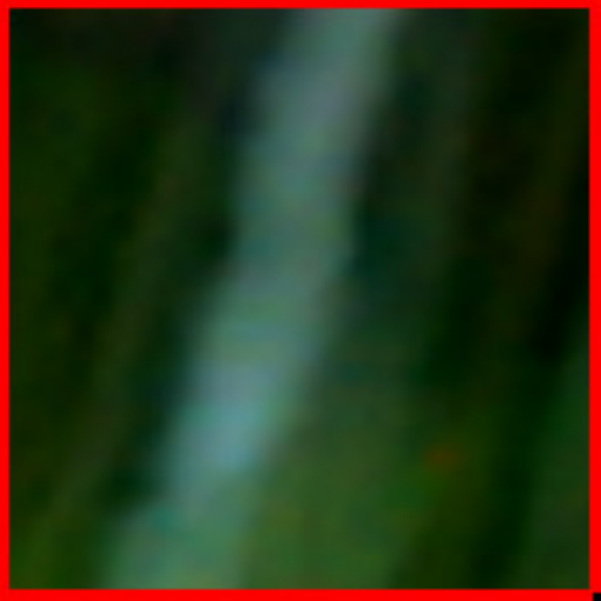} &
            \includegraphics[width=\ssubwidth\linewidth]{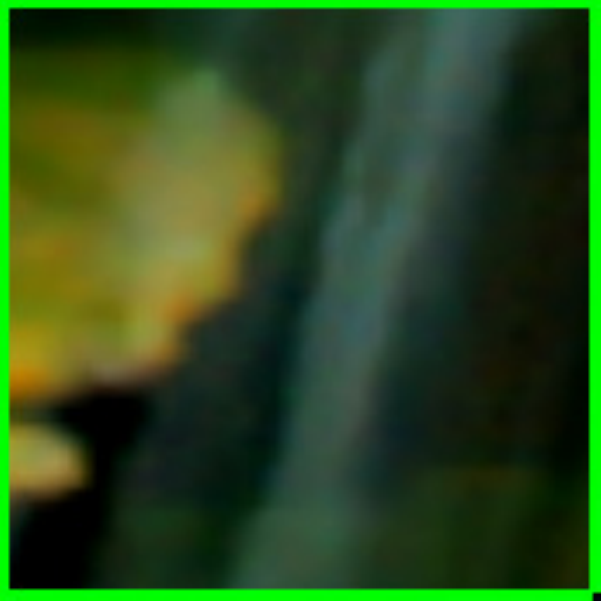} &
            \includegraphics[width=\ssubwidth\linewidth]{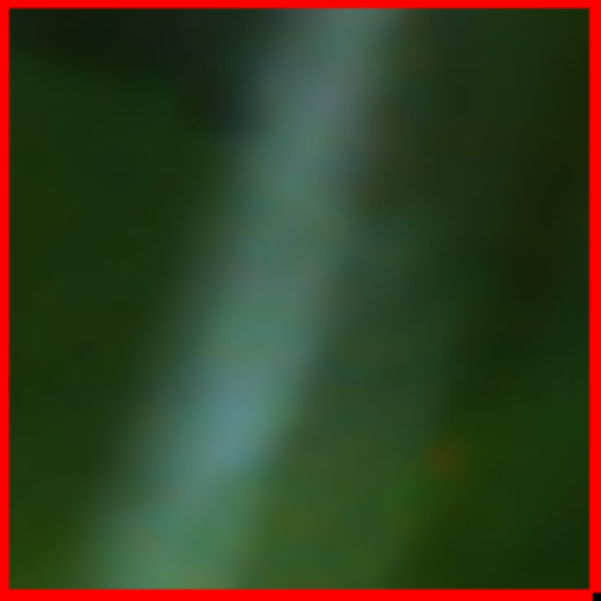} &
            \includegraphics[width=\ssubwidth\linewidth]{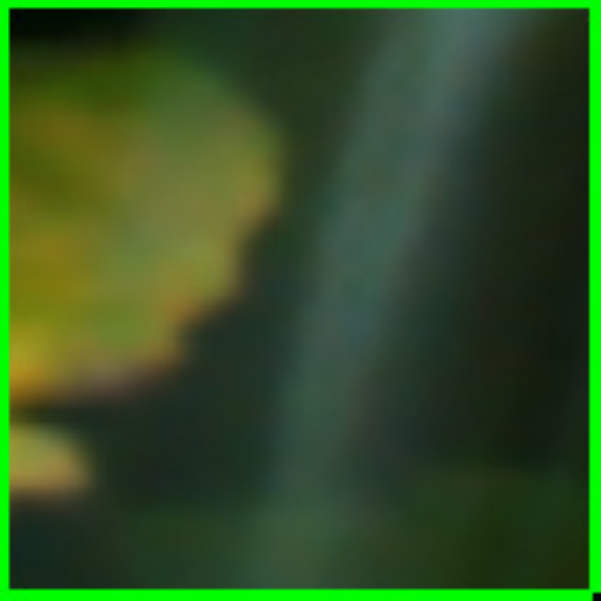} &

            \includegraphics[width=\ssubwidth\linewidth]{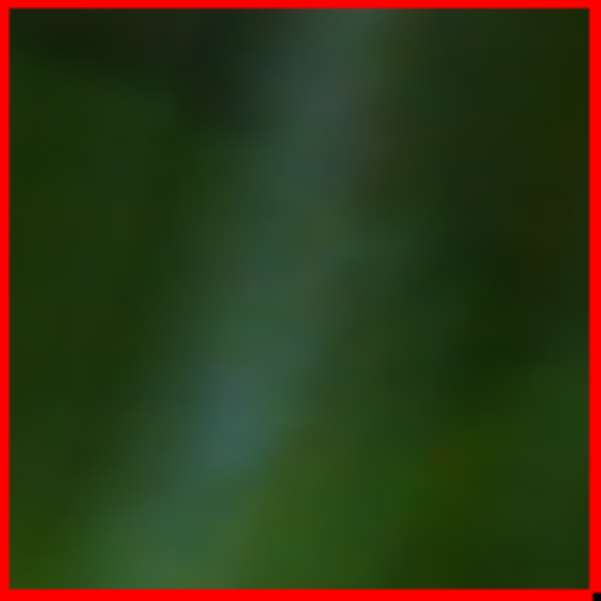} &
            \includegraphics[width=\ssubwidth\linewidth]{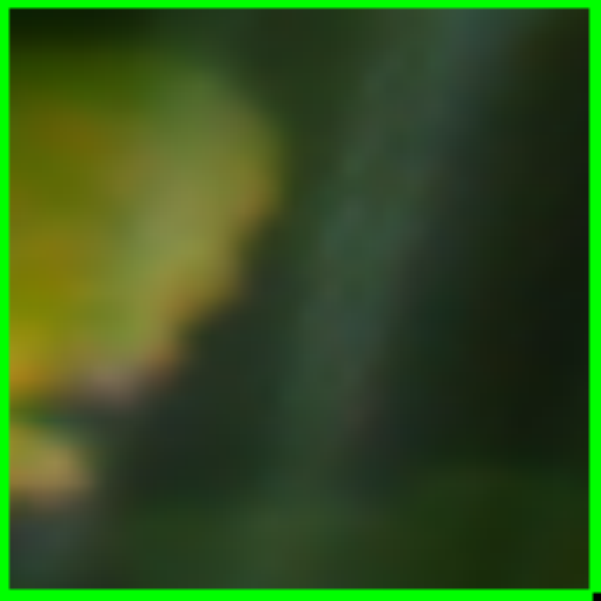} &
            \includegraphics[width=\ssubwidth\linewidth]{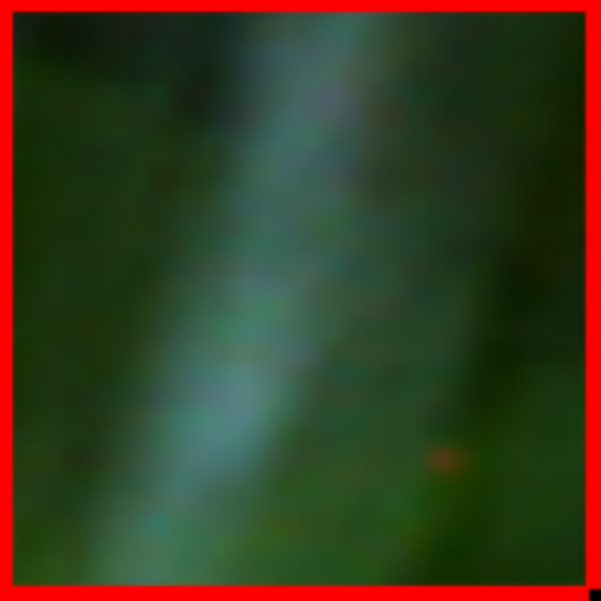} &
            \includegraphics[width=\ssubwidth\linewidth]{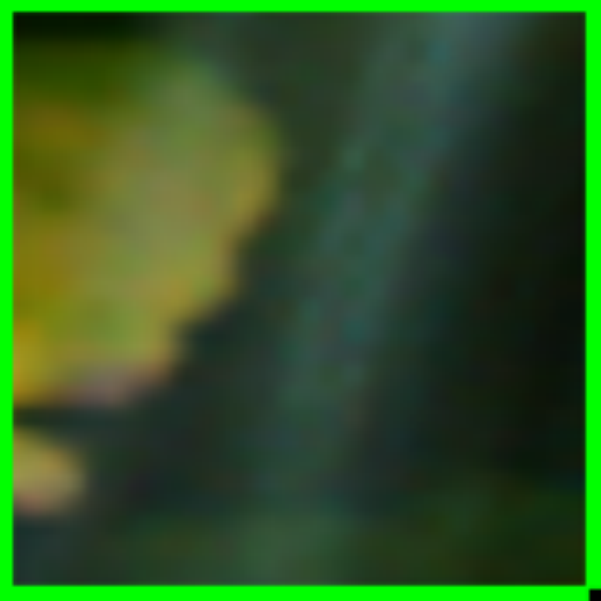} &
            \includegraphics[width=\ssubwidth\linewidth]{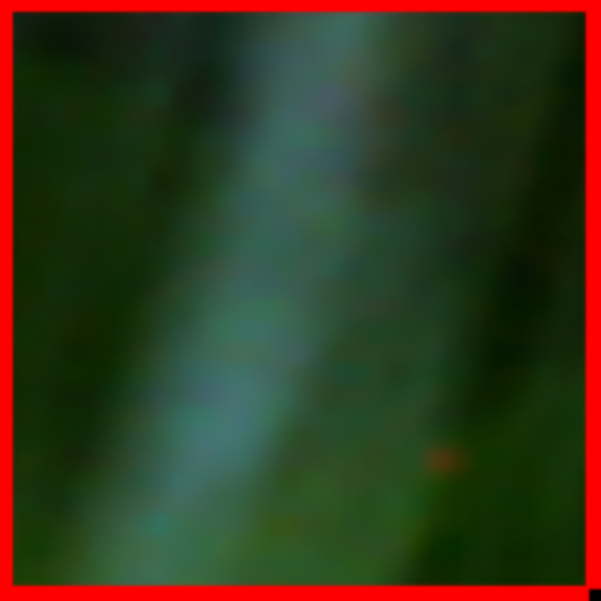} &
            \includegraphics[width=\ssubwidth\linewidth]{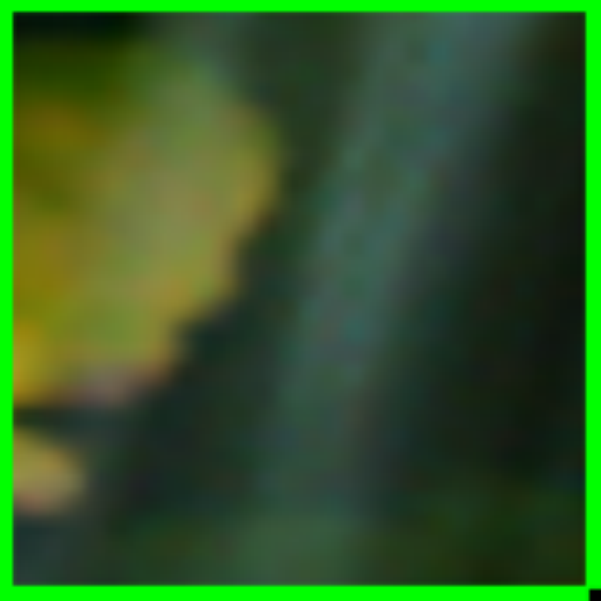} &
            \includegraphics[width=\ssubwidth\linewidth]{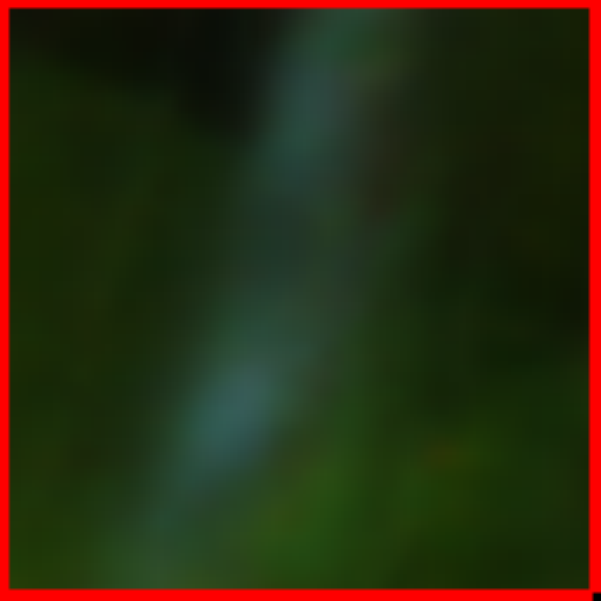} &
            \includegraphics[width=\ssubwidth\linewidth]{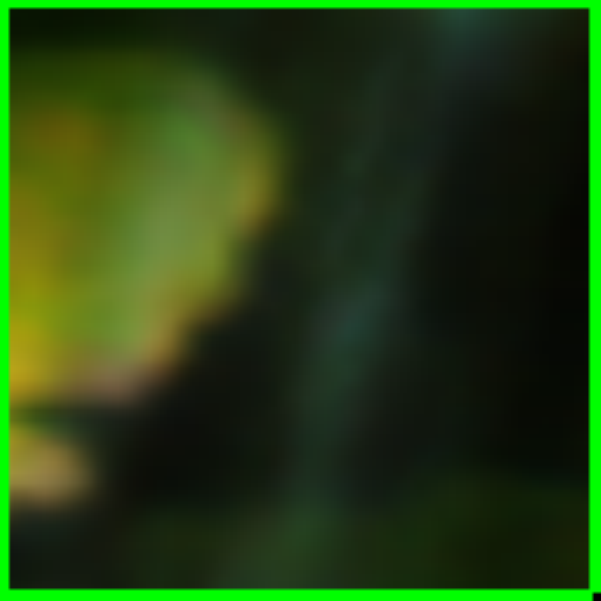} &
            \includegraphics[width=\ssubwidth\linewidth]{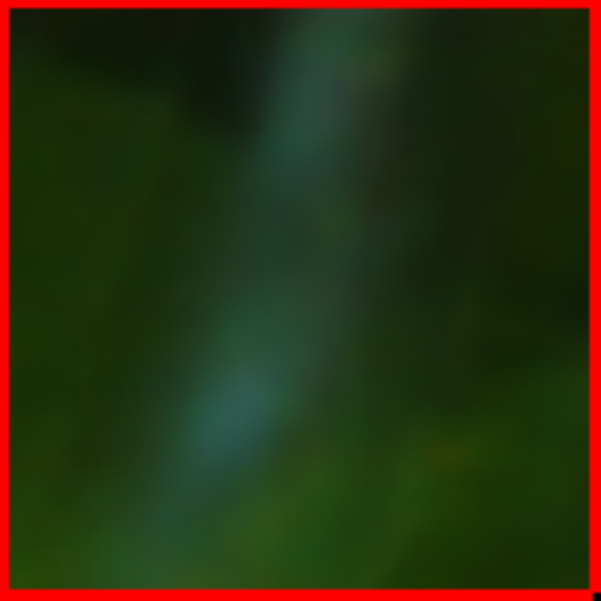} &
            \includegraphics[width=\ssubwidth\linewidth]{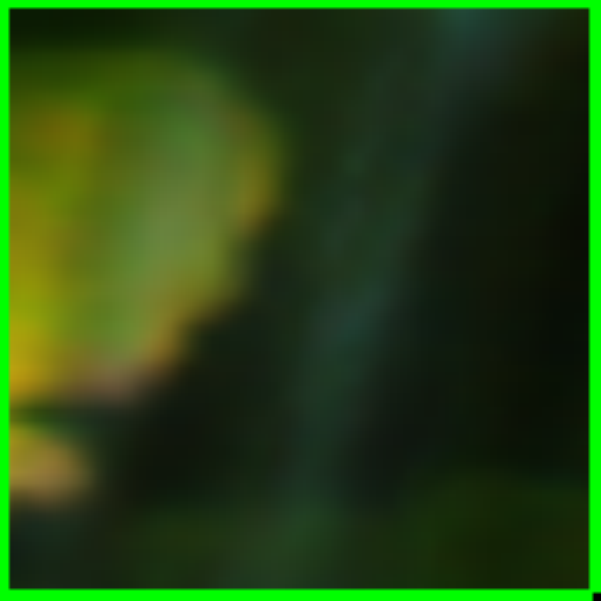} \\

            \multicolumn{2}{c}{\includegraphics[width=\subwidth\linewidth]{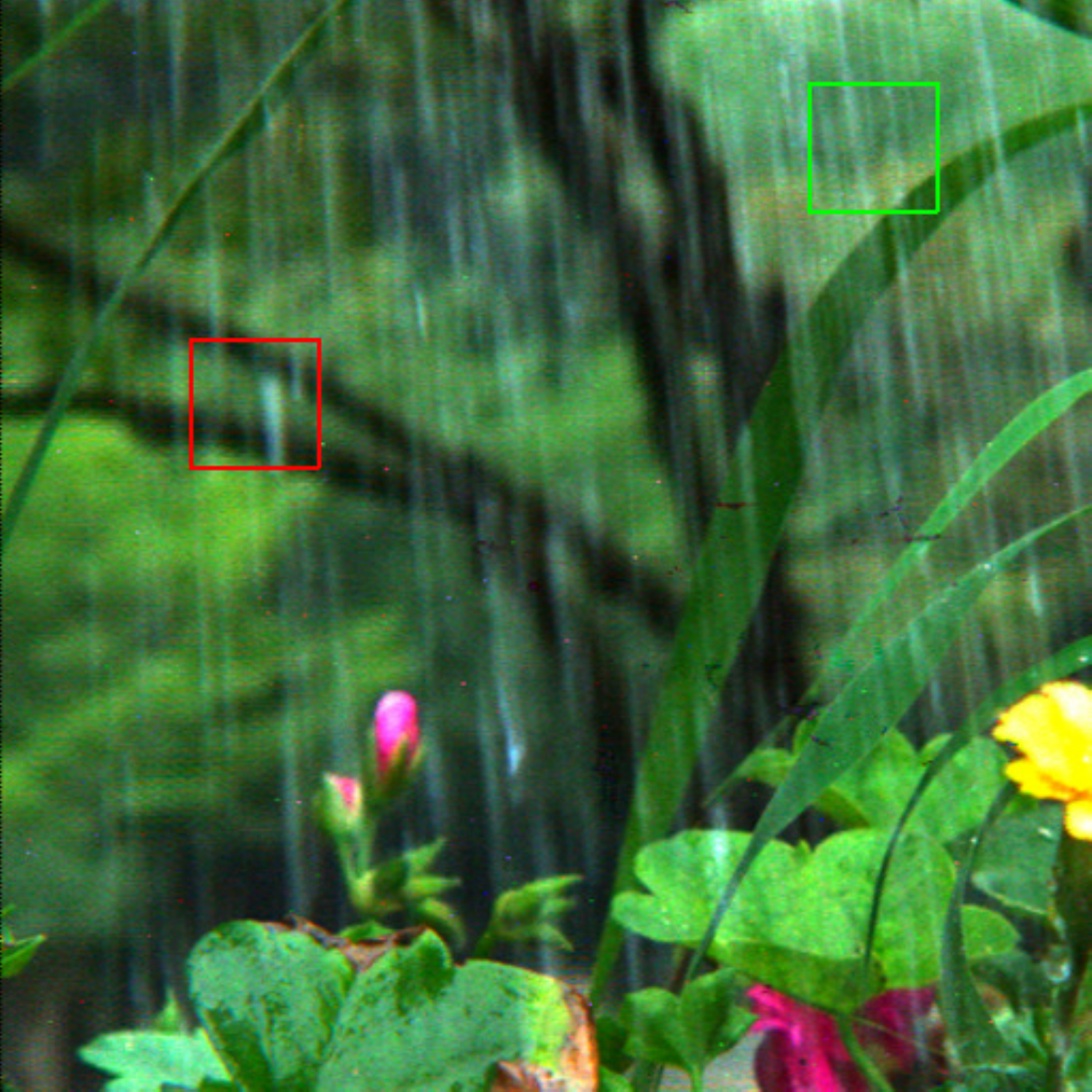}} &
            \multicolumn{2}{c}{\includegraphics[width=\subwidth\linewidth]{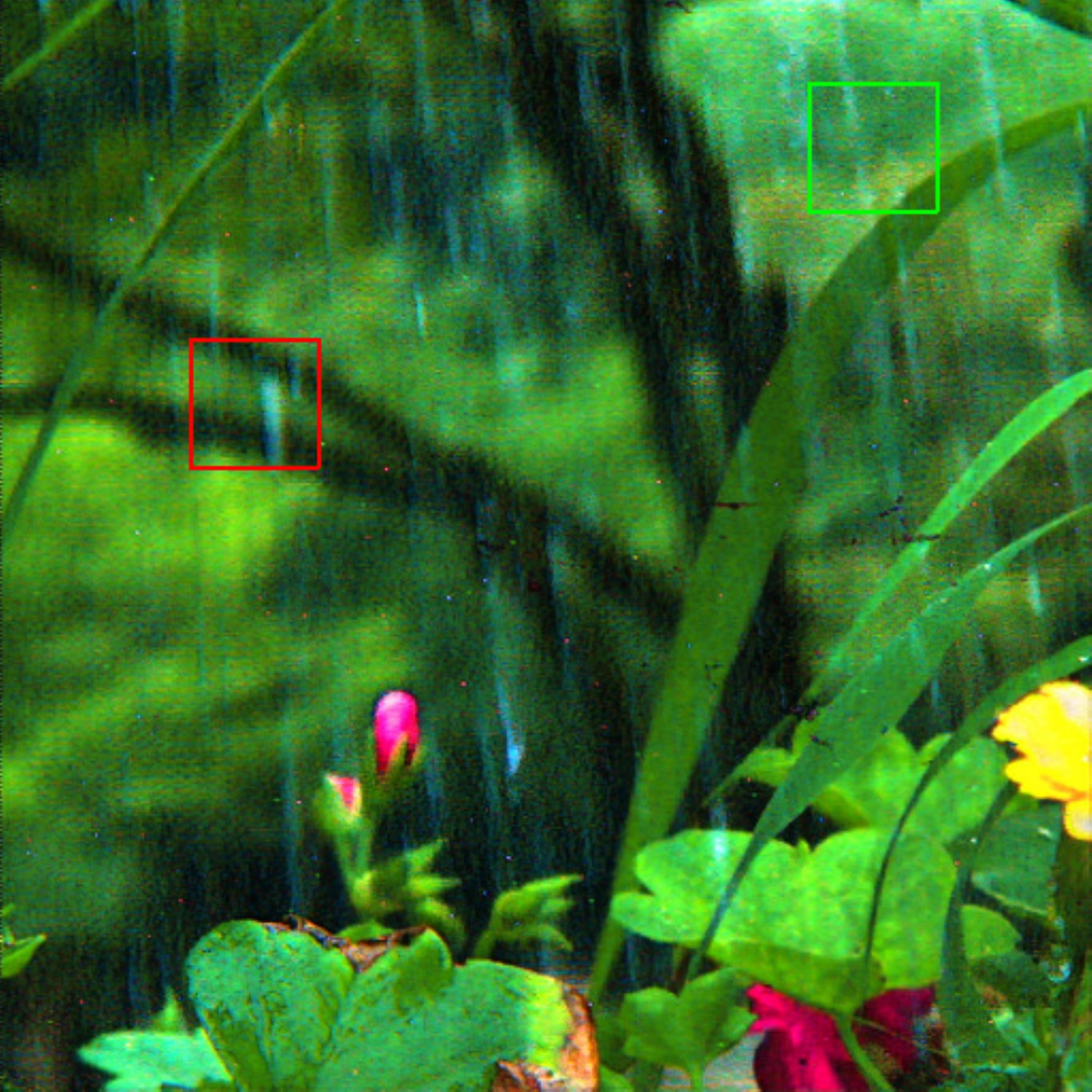}} &
            \multicolumn{2}{c}{\includegraphics[width=\subwidth\linewidth]{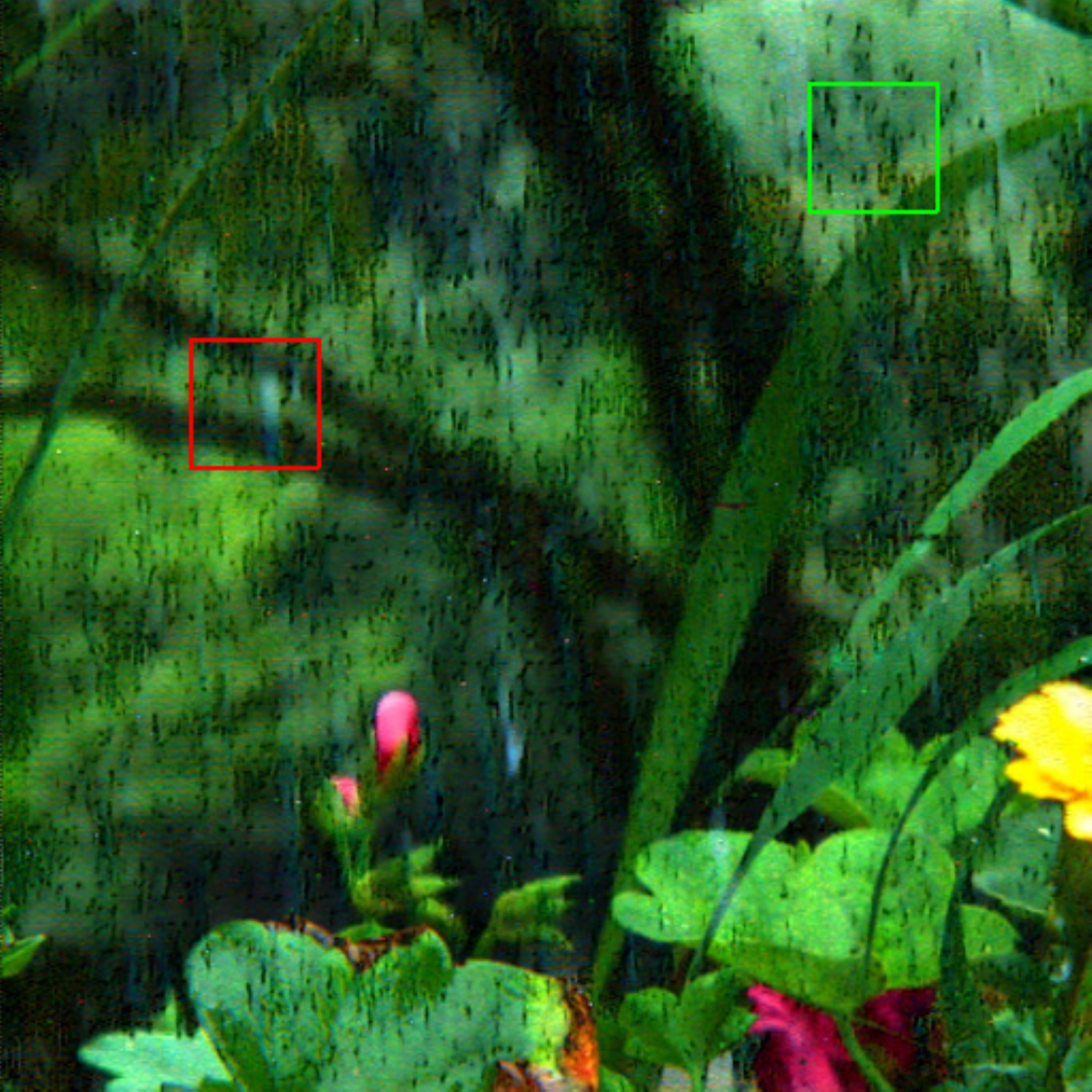}} &
            \multicolumn{2}{c}{\includegraphics[width=\subwidth\linewidth]{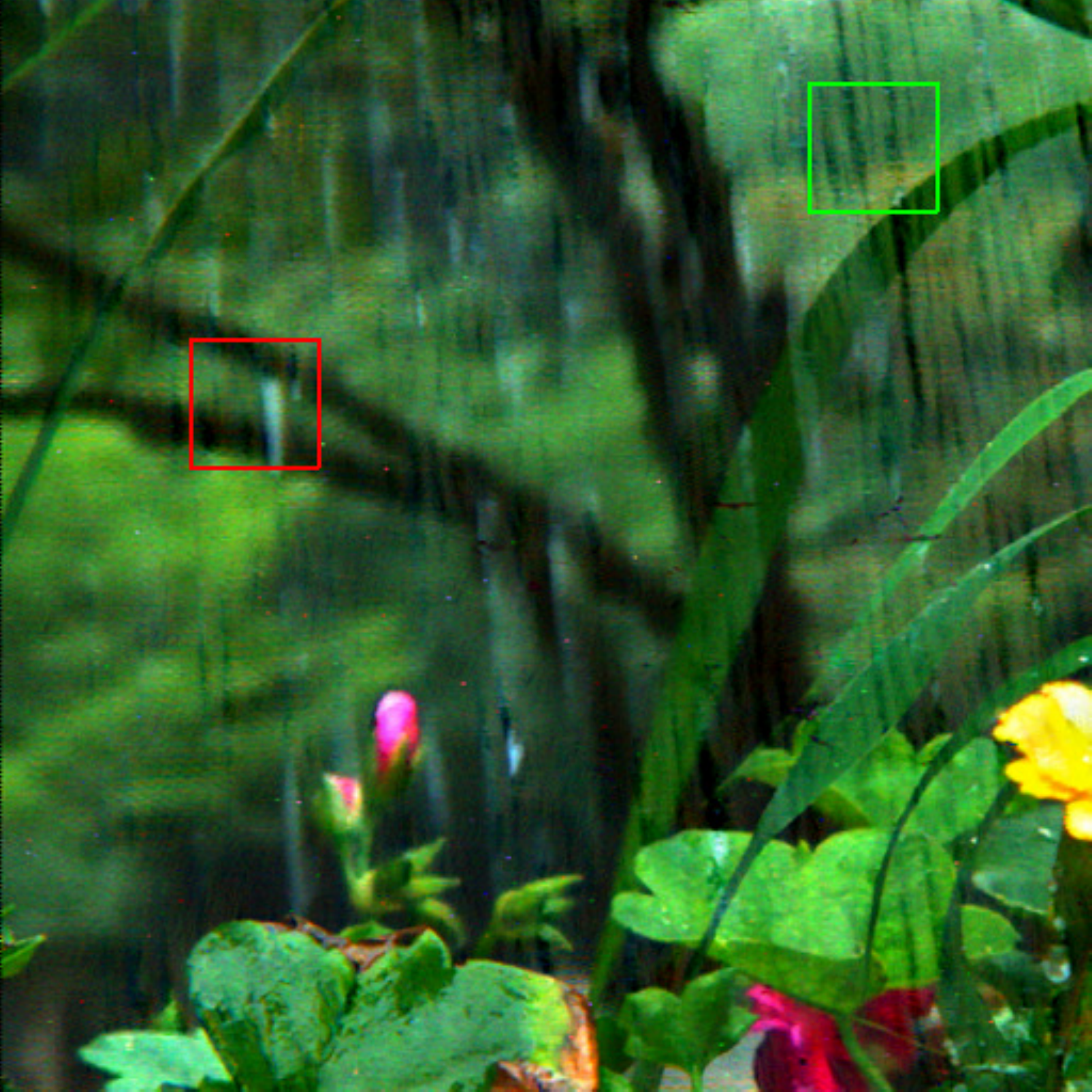}} &
            \multicolumn{2}{c}{\includegraphics[width=\subwidth\linewidth]{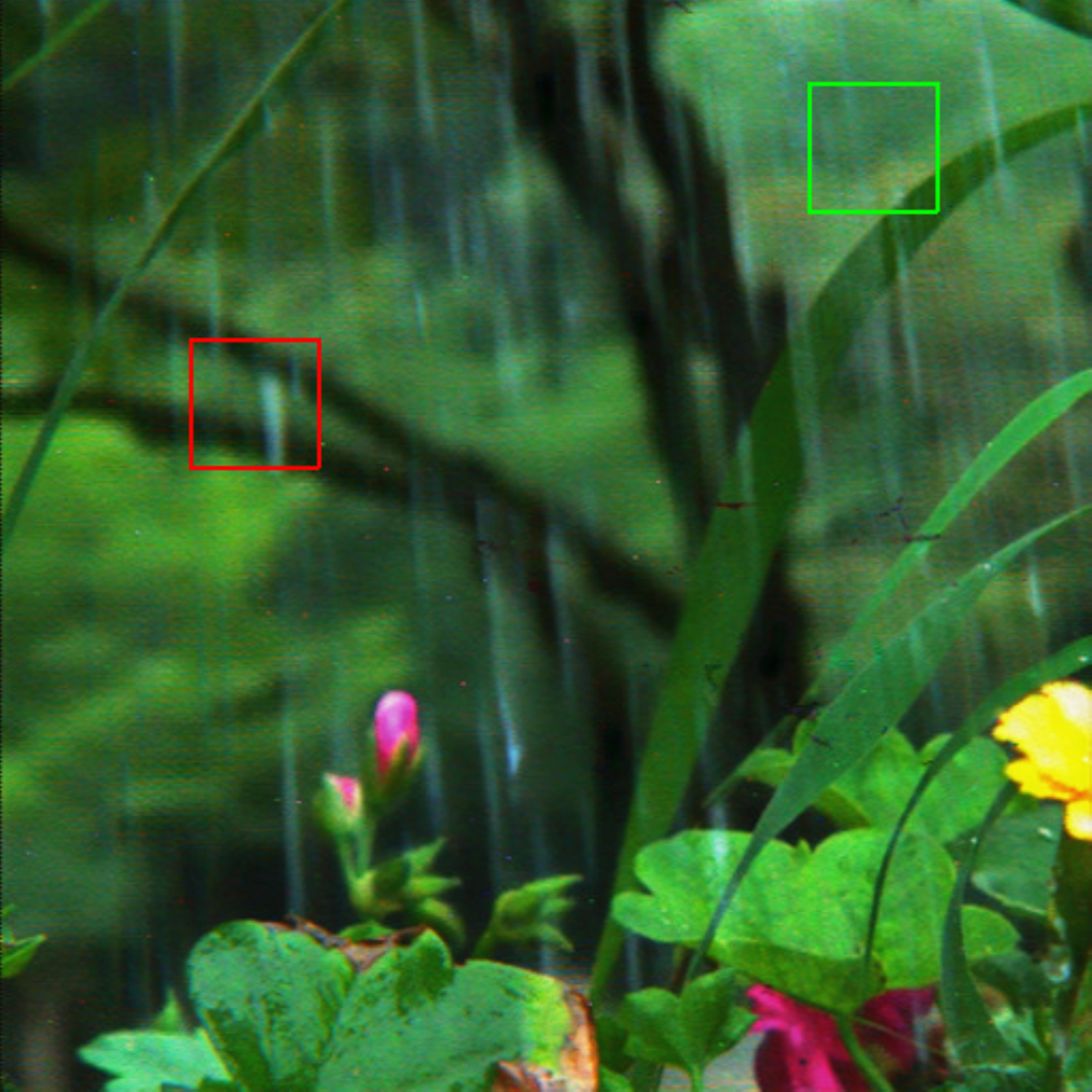}} &

            \multicolumn{2}{c}{\includegraphics[width=\subwidth\linewidth]{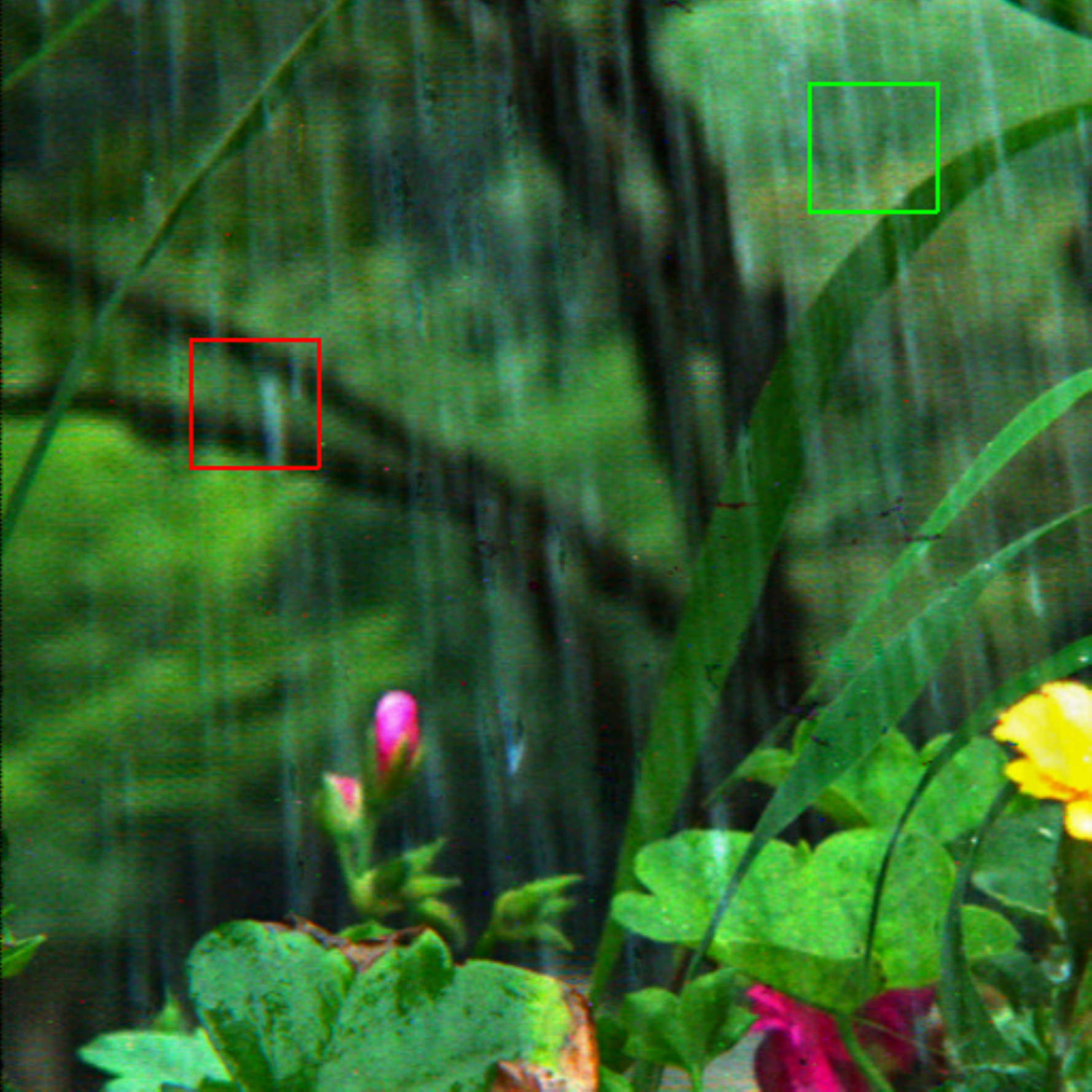}} &
            \multicolumn{2}{c}{\includegraphics[width=\subwidth\linewidth]{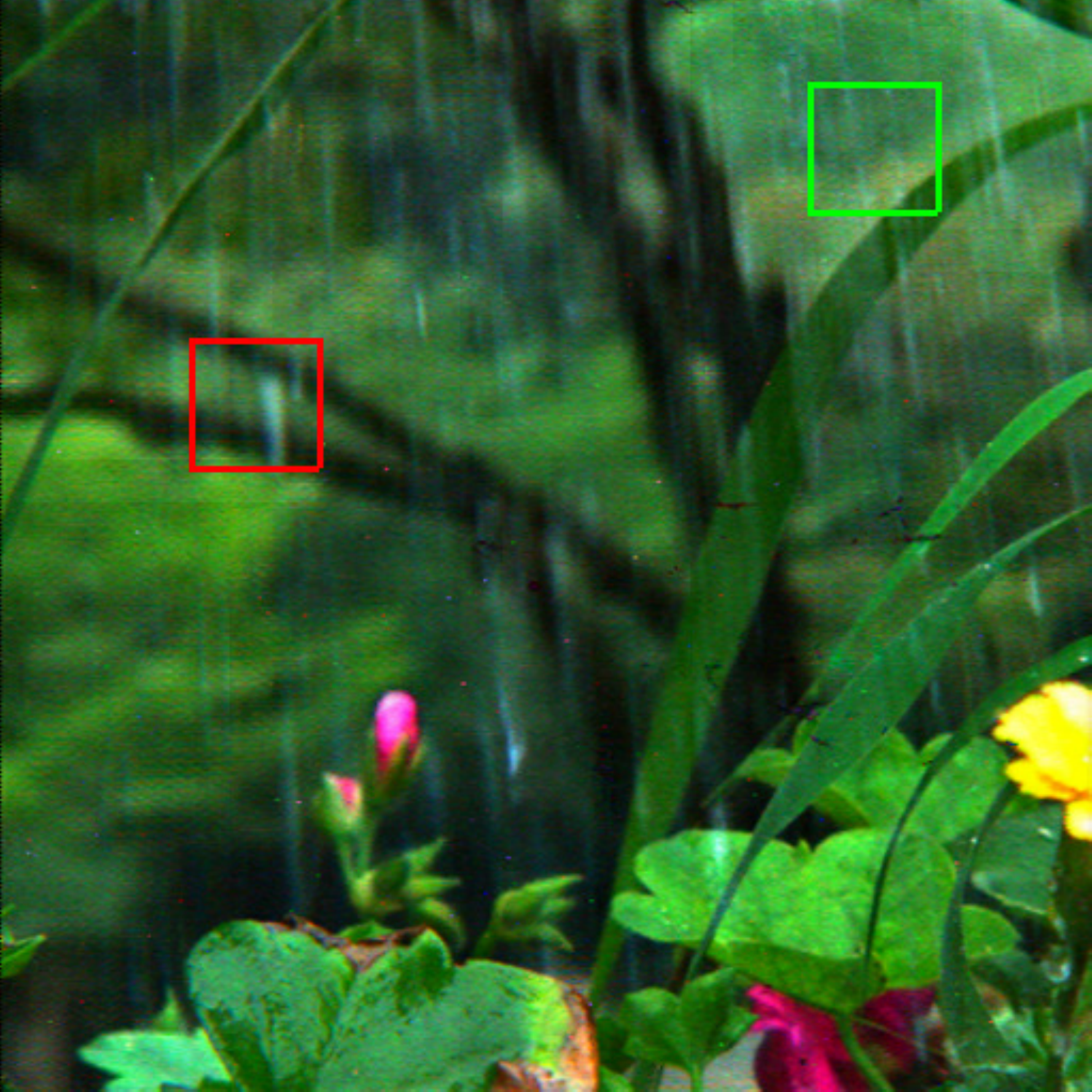}} &
            \multicolumn{2}{c}{\includegraphics[width=\subwidth\linewidth]{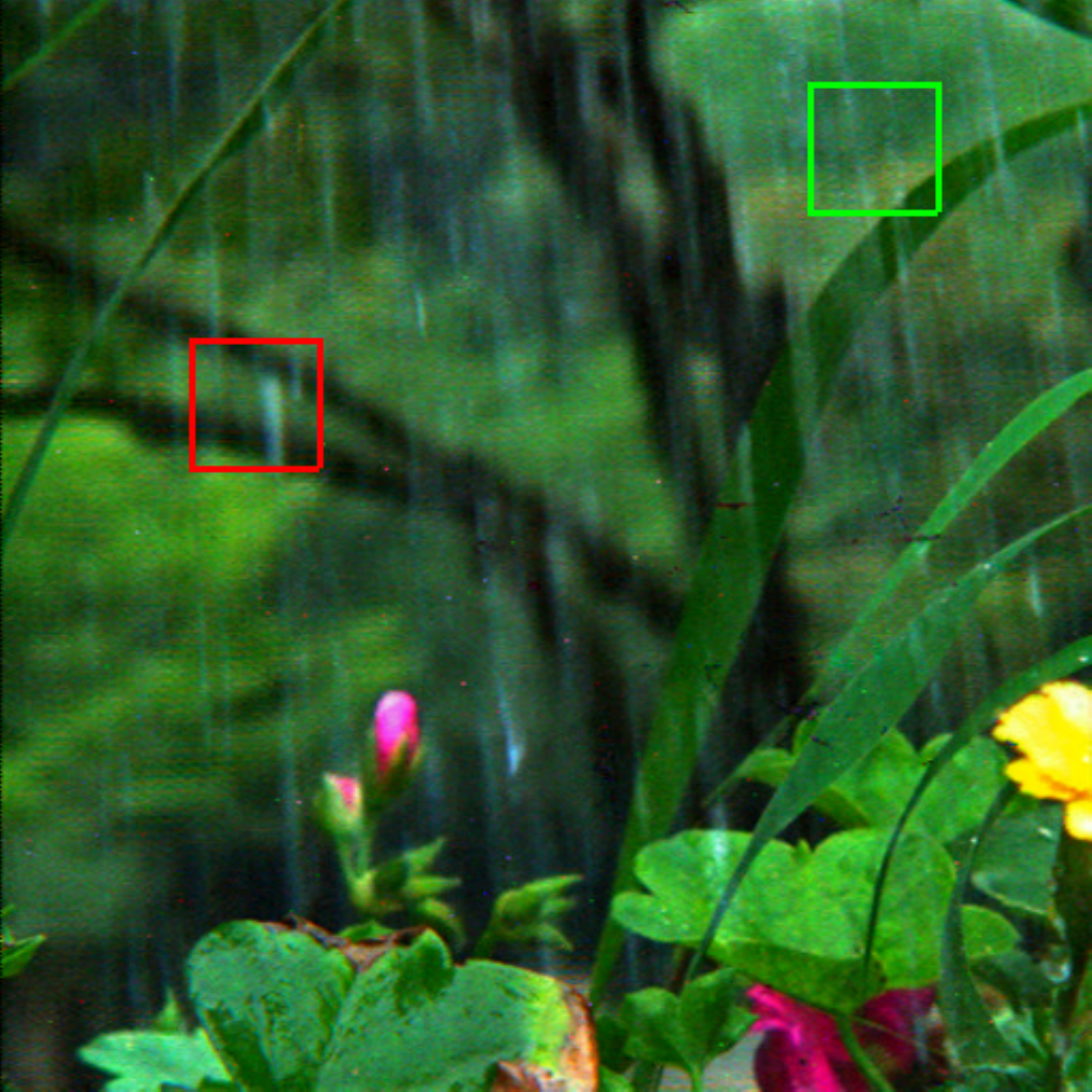}} &
            \multicolumn{2}{c}{\includegraphics[width=\subwidth\linewidth]{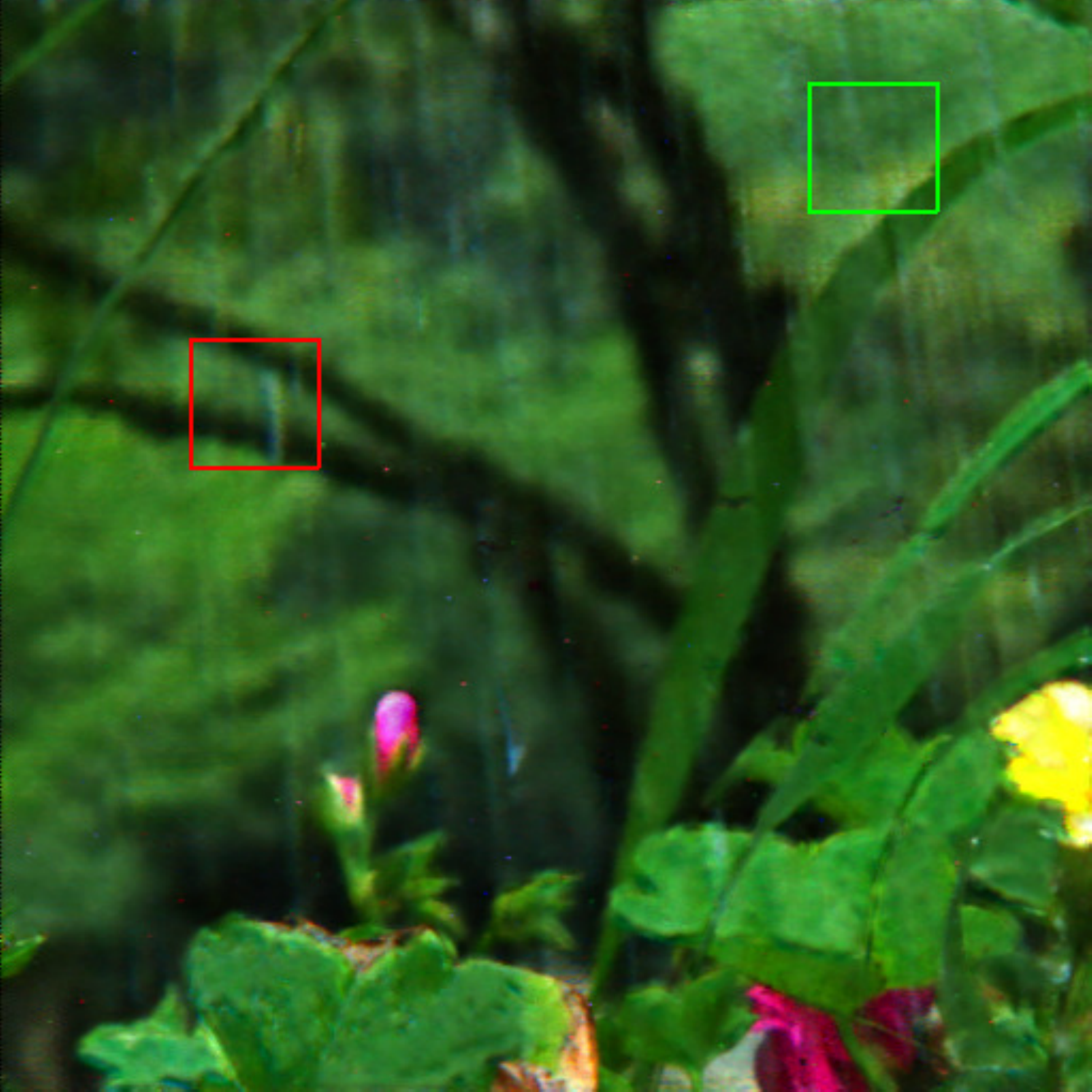}} &
            \multicolumn{2}{c}{\includegraphics[width=\subwidth\linewidth]{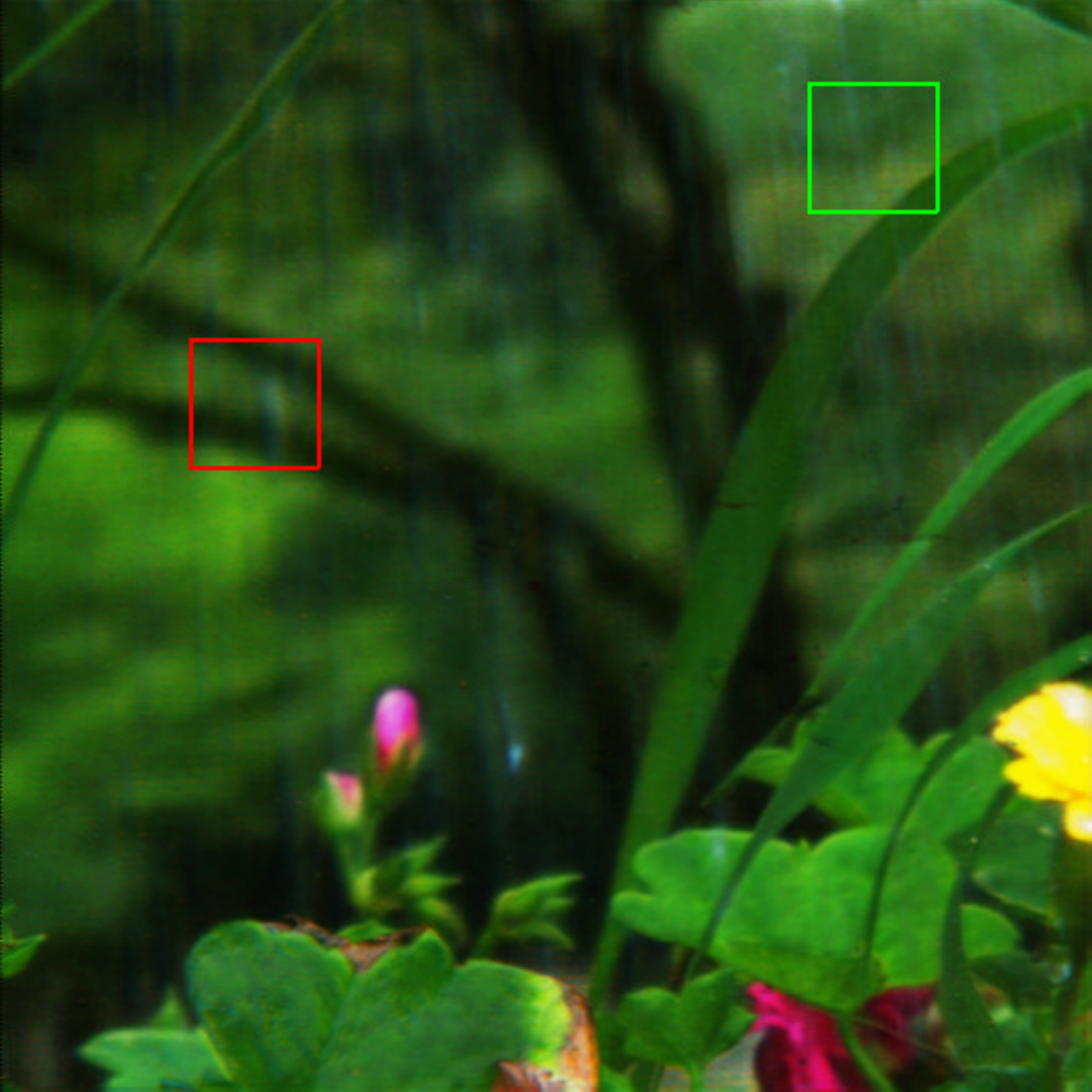}} \\

            \includegraphics[width=\ssubwidth\linewidth]{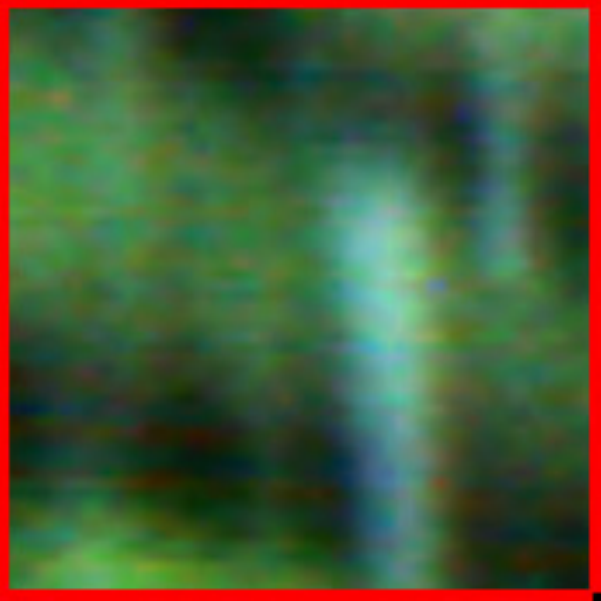} &
            \includegraphics[width=\ssubwidth\linewidth]{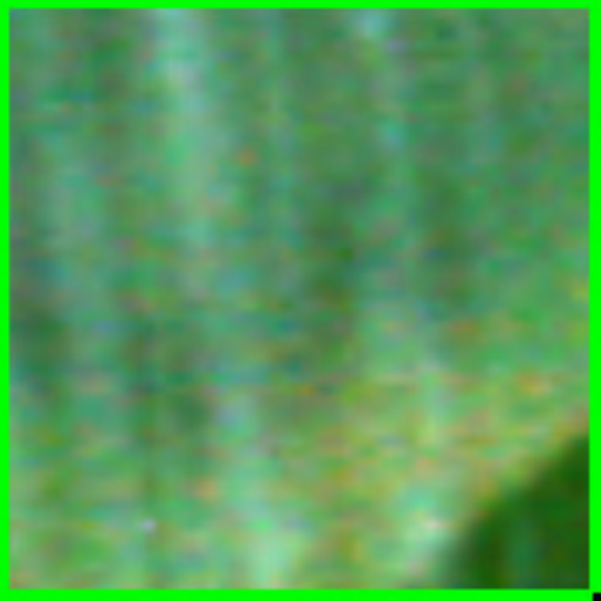} &
            \includegraphics[width=\ssubwidth\linewidth]{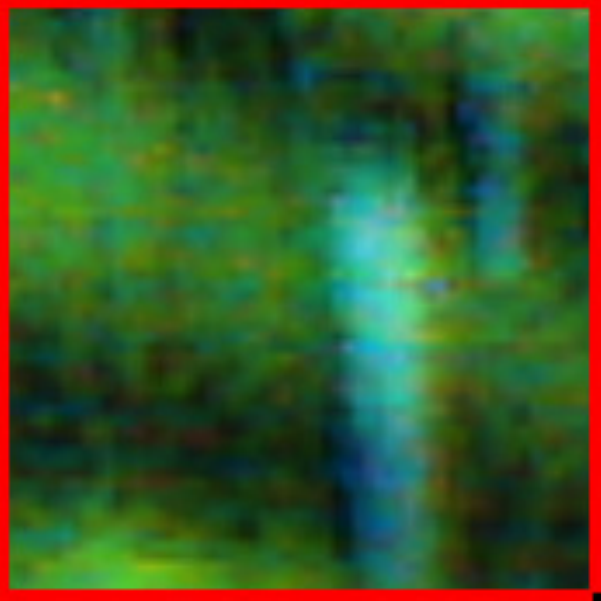} &
            \includegraphics[width=\ssubwidth\linewidth]{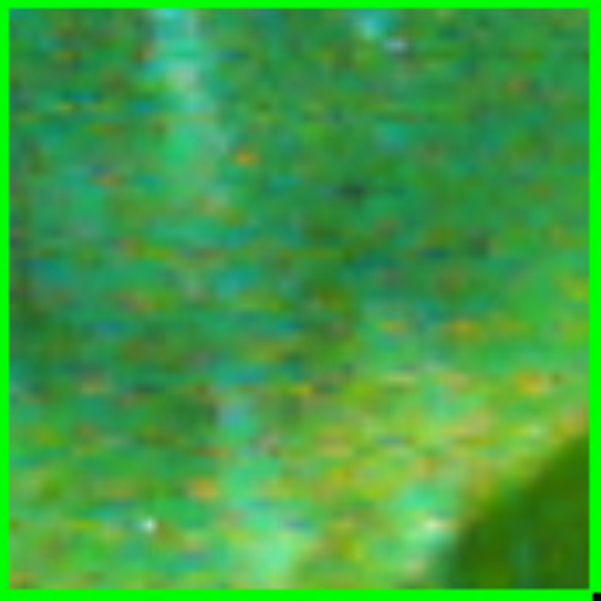} &
            \includegraphics[width=\ssubwidth\linewidth]{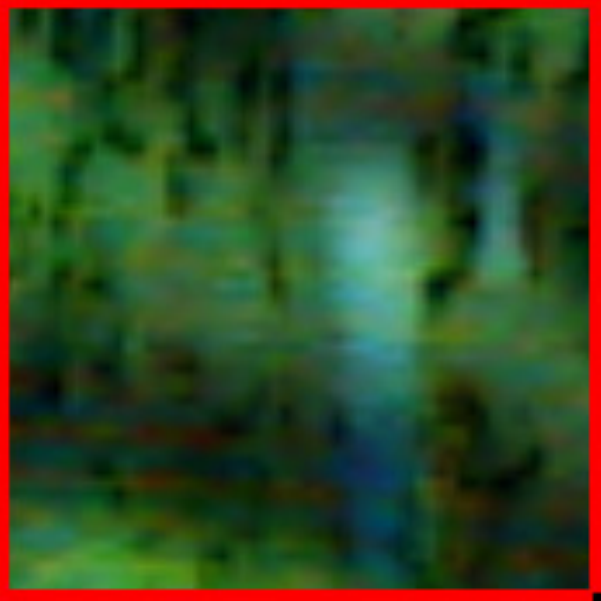} &
            \includegraphics[width=\ssubwidth\linewidth]{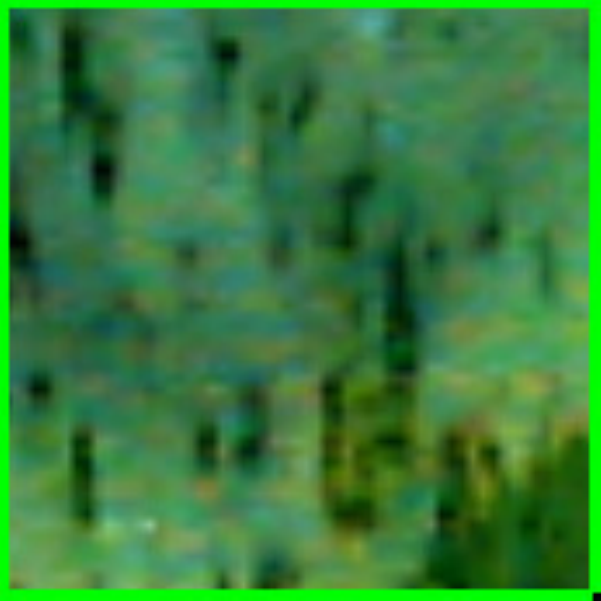} &
            \includegraphics[width=\ssubwidth\linewidth]{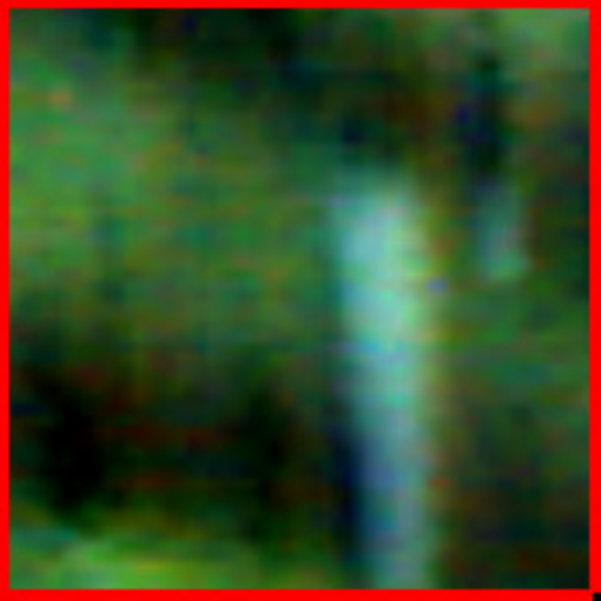} &
            \includegraphics[width=\ssubwidth\linewidth]{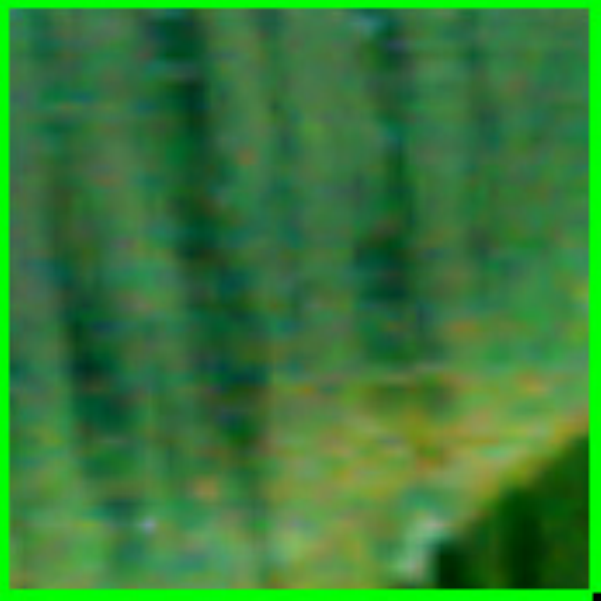} &
            \includegraphics[width=\ssubwidth\linewidth]{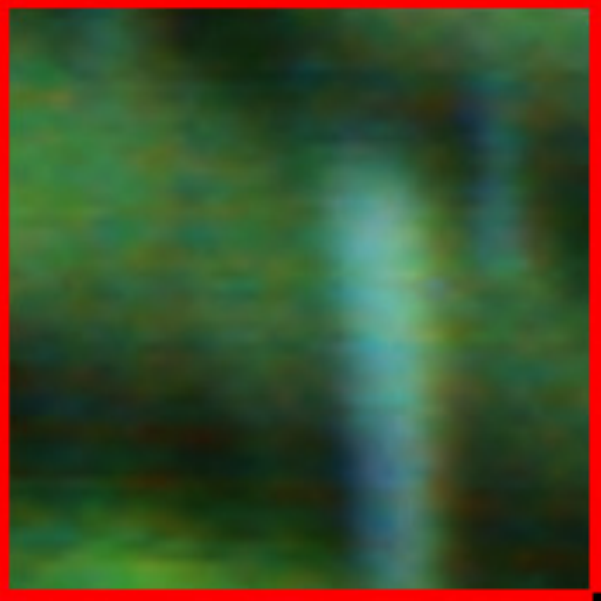} &
            \includegraphics[width=\ssubwidth\linewidth]{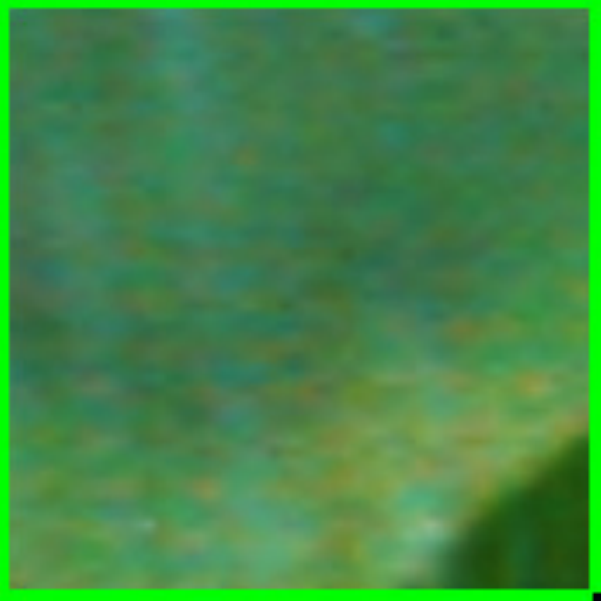} &

            \includegraphics[width=\ssubwidth\linewidth]{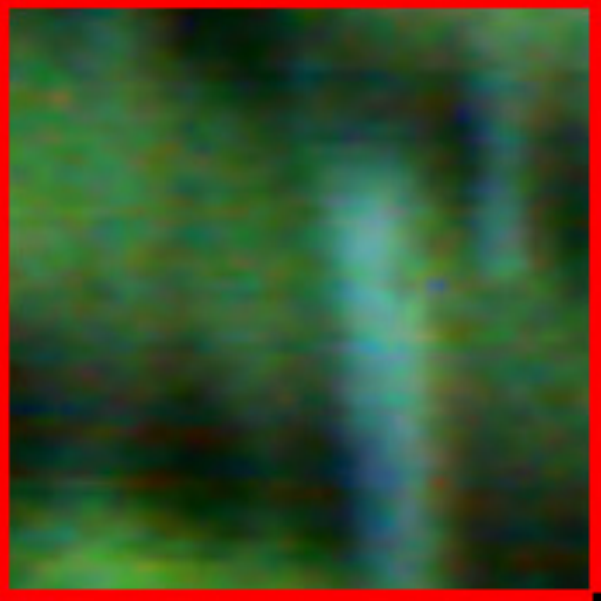} &
            \includegraphics[width=\ssubwidth\linewidth]{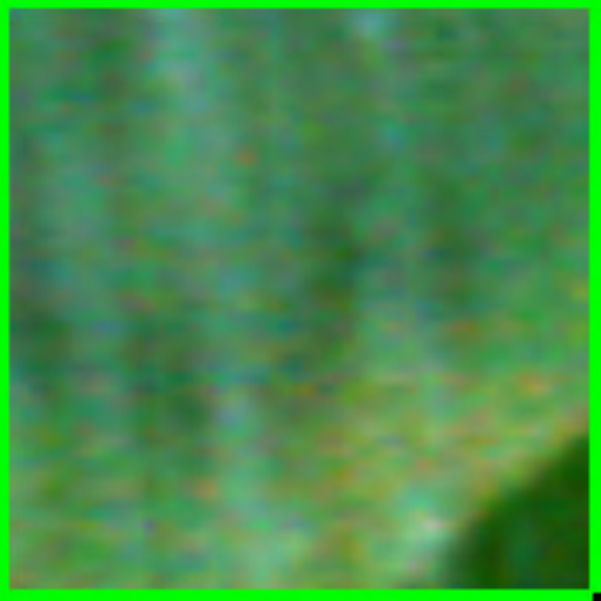} &
             \includegraphics[width=\ssubwidth\linewidth]{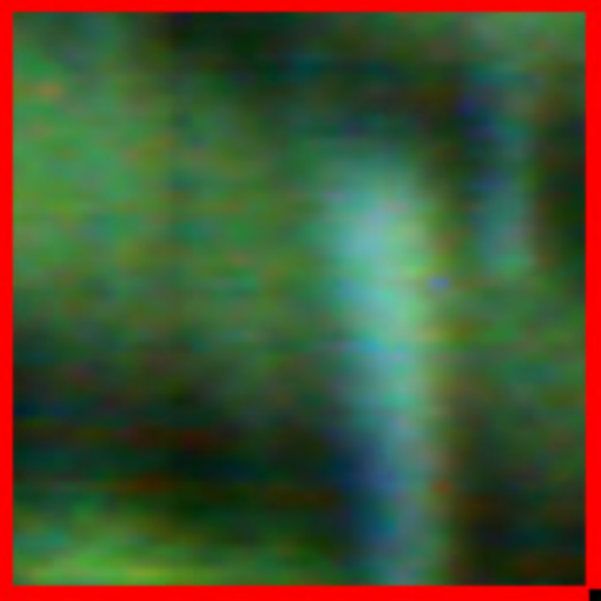} &
            \includegraphics[width=\ssubwidth\linewidth]{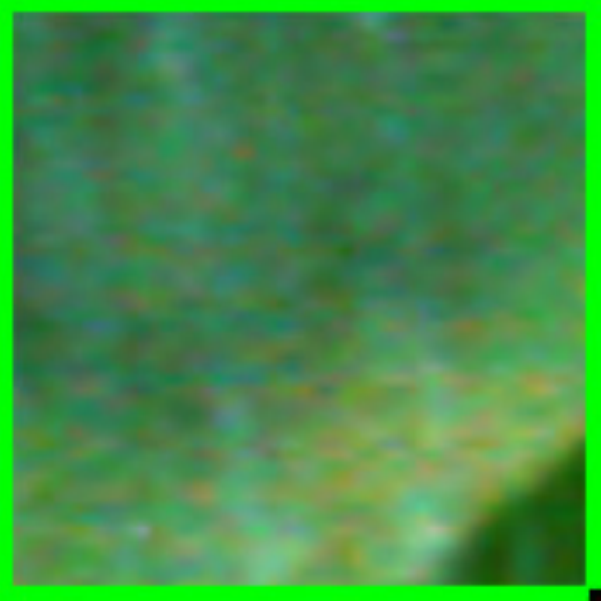} &
            \includegraphics[width=\ssubwidth\linewidth]{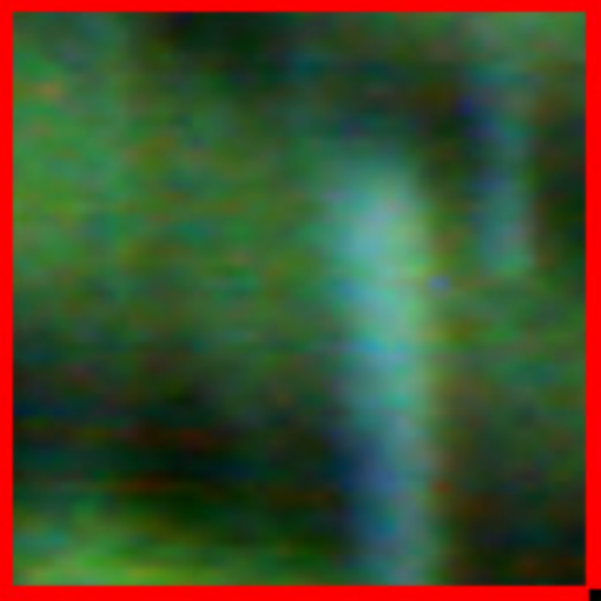} &
            \includegraphics[width=\ssubwidth\linewidth]{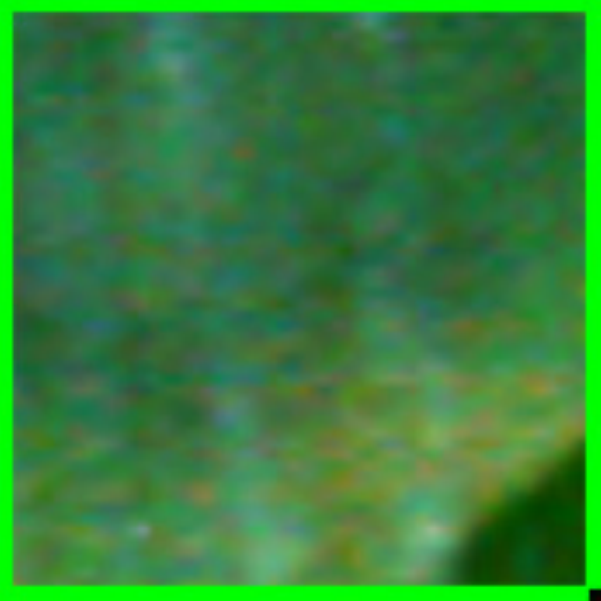} &
            \includegraphics[width=\ssubwidth\linewidth]{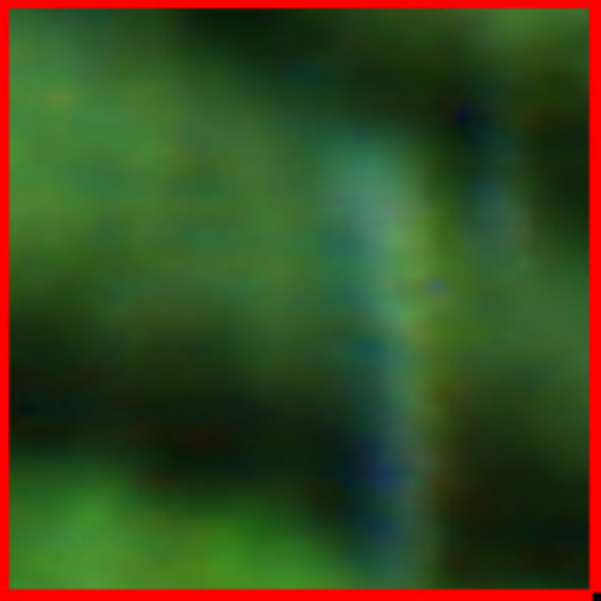} &
            \includegraphics[width=\ssubwidth\linewidth]{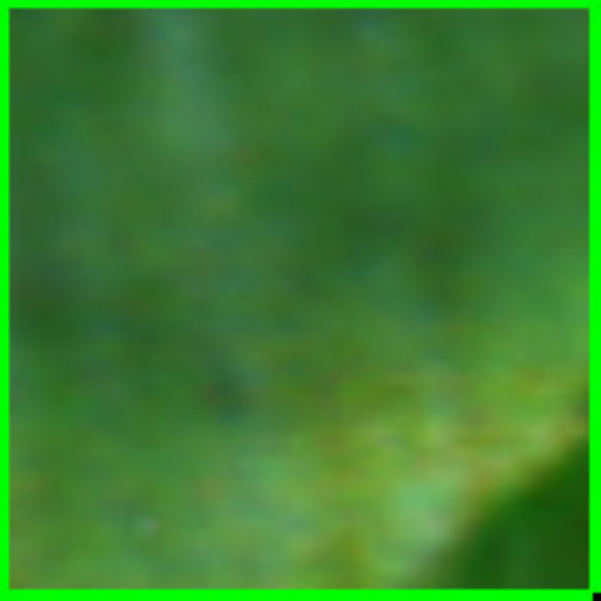} &
            \includegraphics[width=\ssubwidth\linewidth]{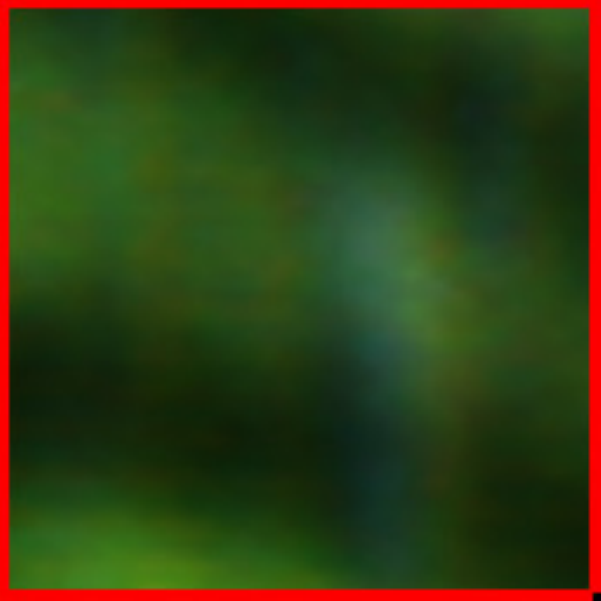} &
            \includegraphics[width=\ssubwidth\linewidth]{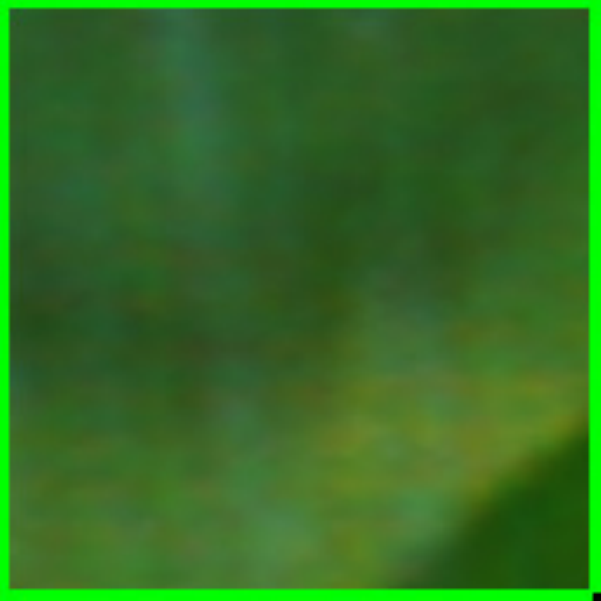} \\

            \multicolumn{2}{c}{\includegraphics[width=\subwidth\linewidth]{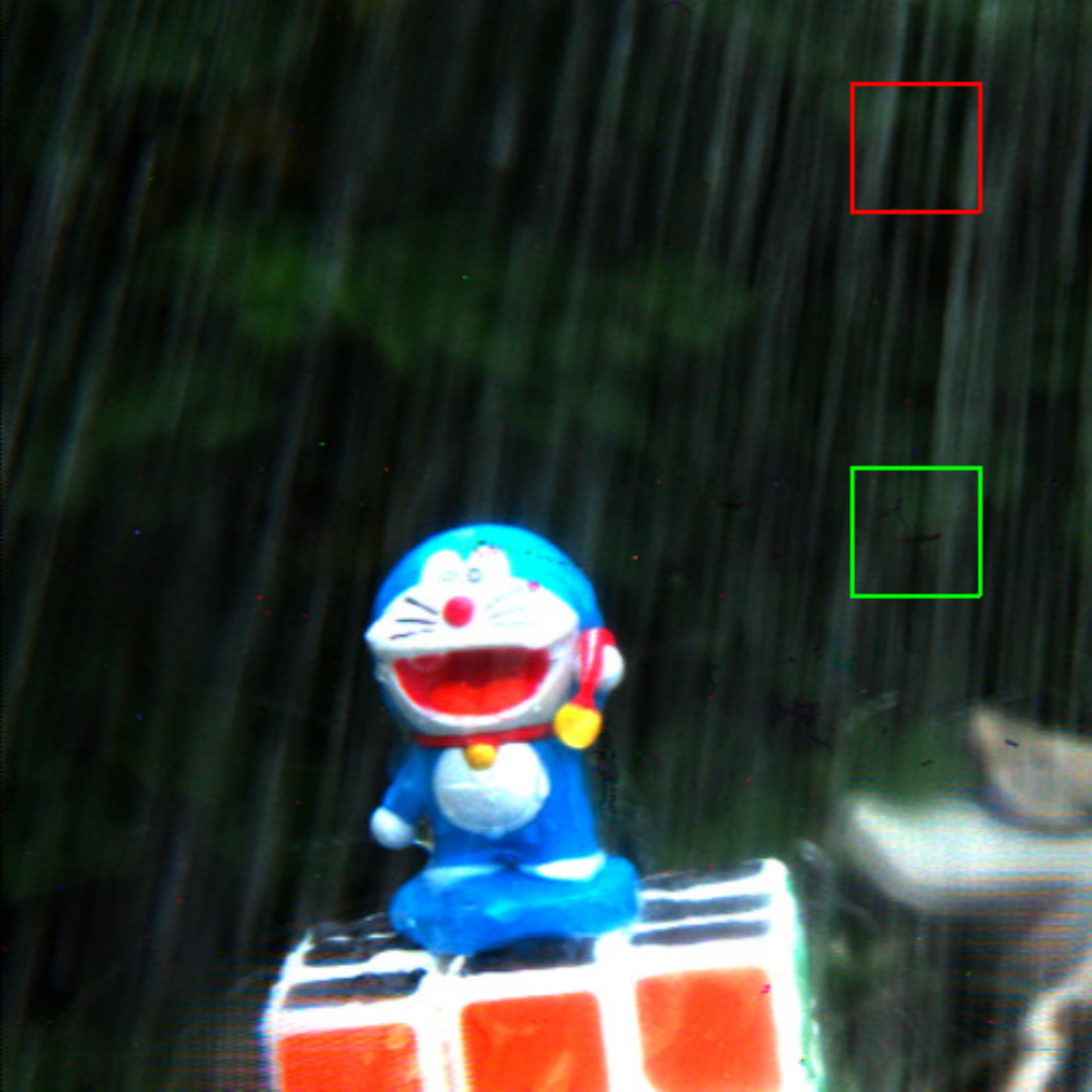}} &
            \multicolumn{2}{c}{\includegraphics[width=\subwidth\linewidth]{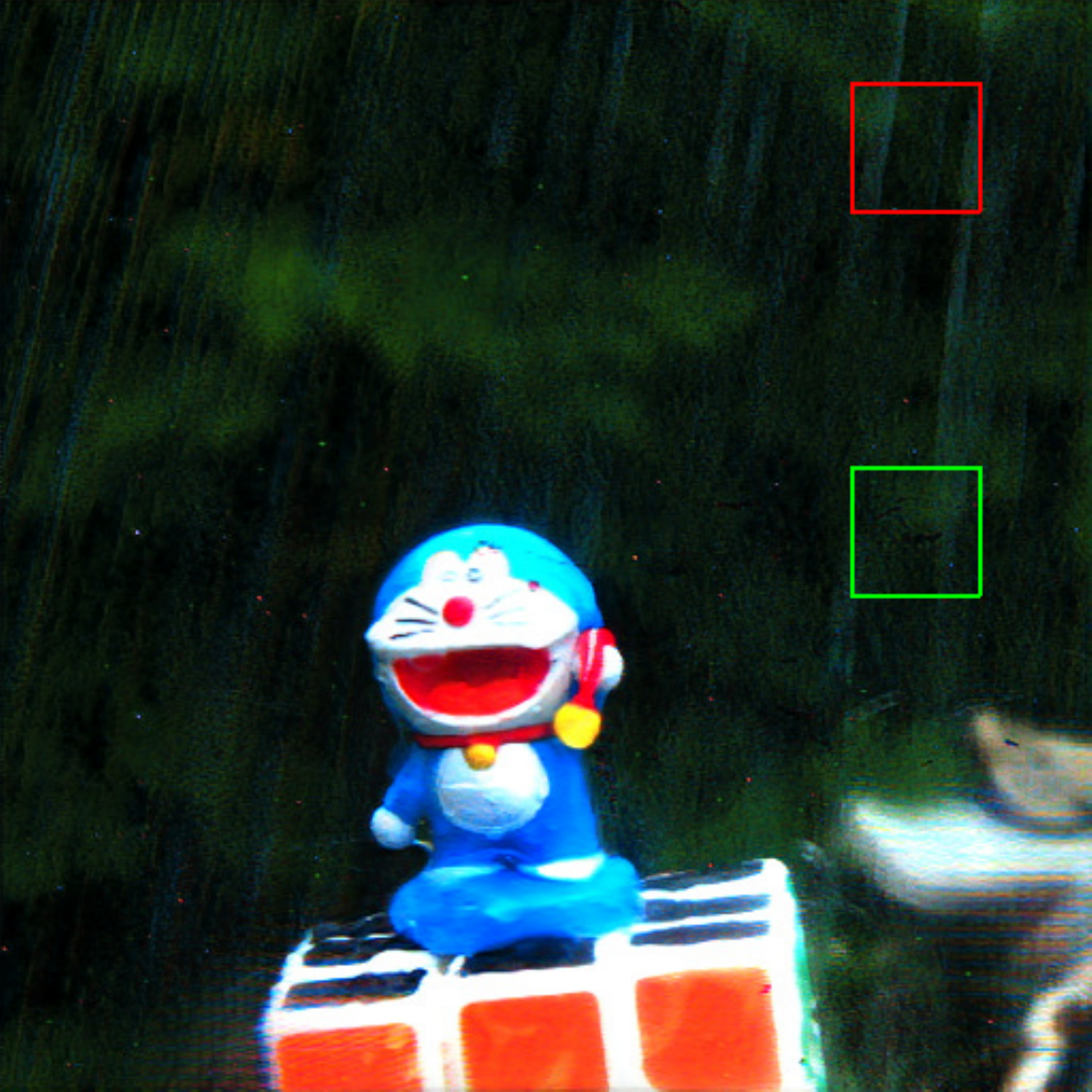}} &
            \multicolumn{2}{c}{\includegraphics[width=\subwidth\linewidth]{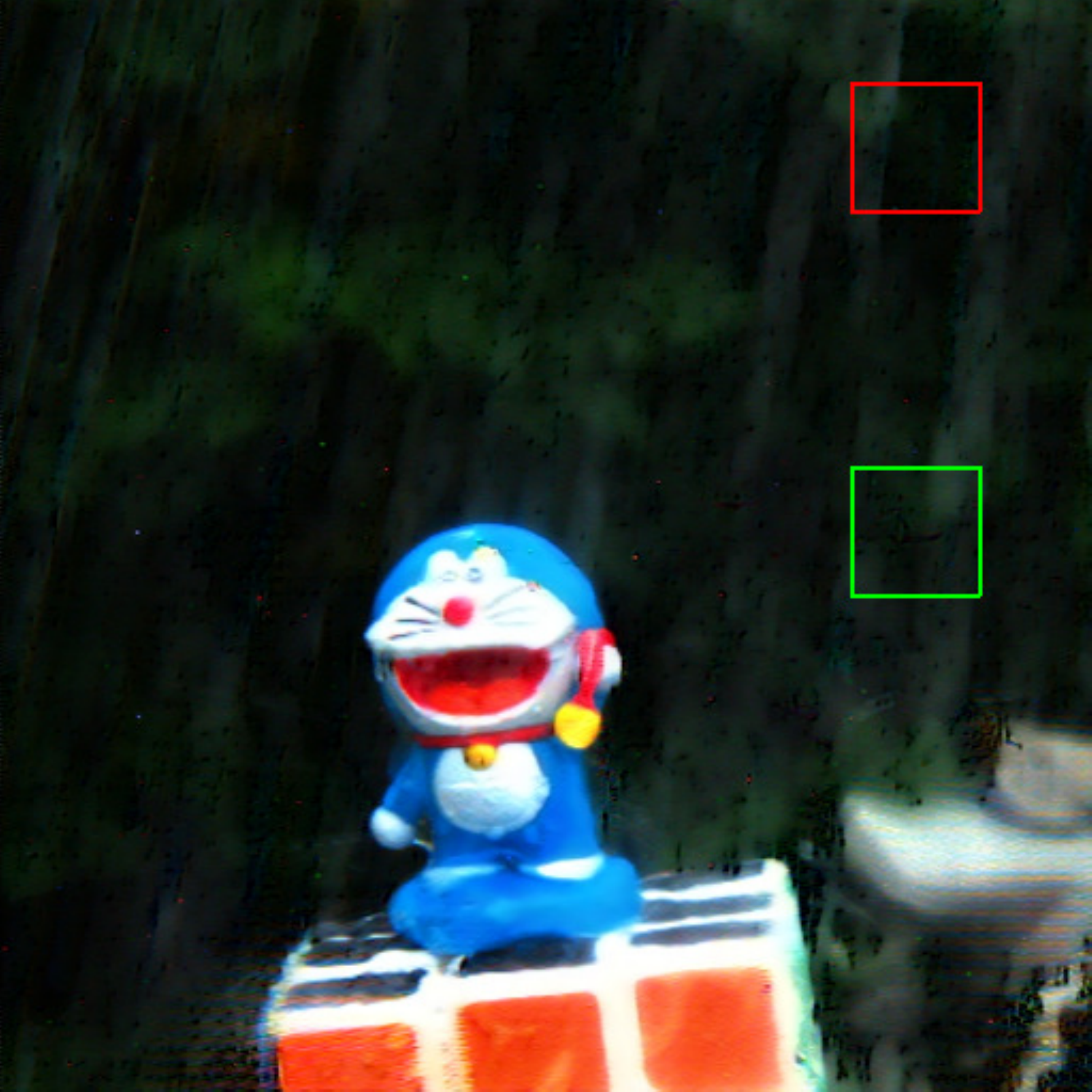}} &
            \multicolumn{2}{c}{\includegraphics[width=\subwidth\linewidth]{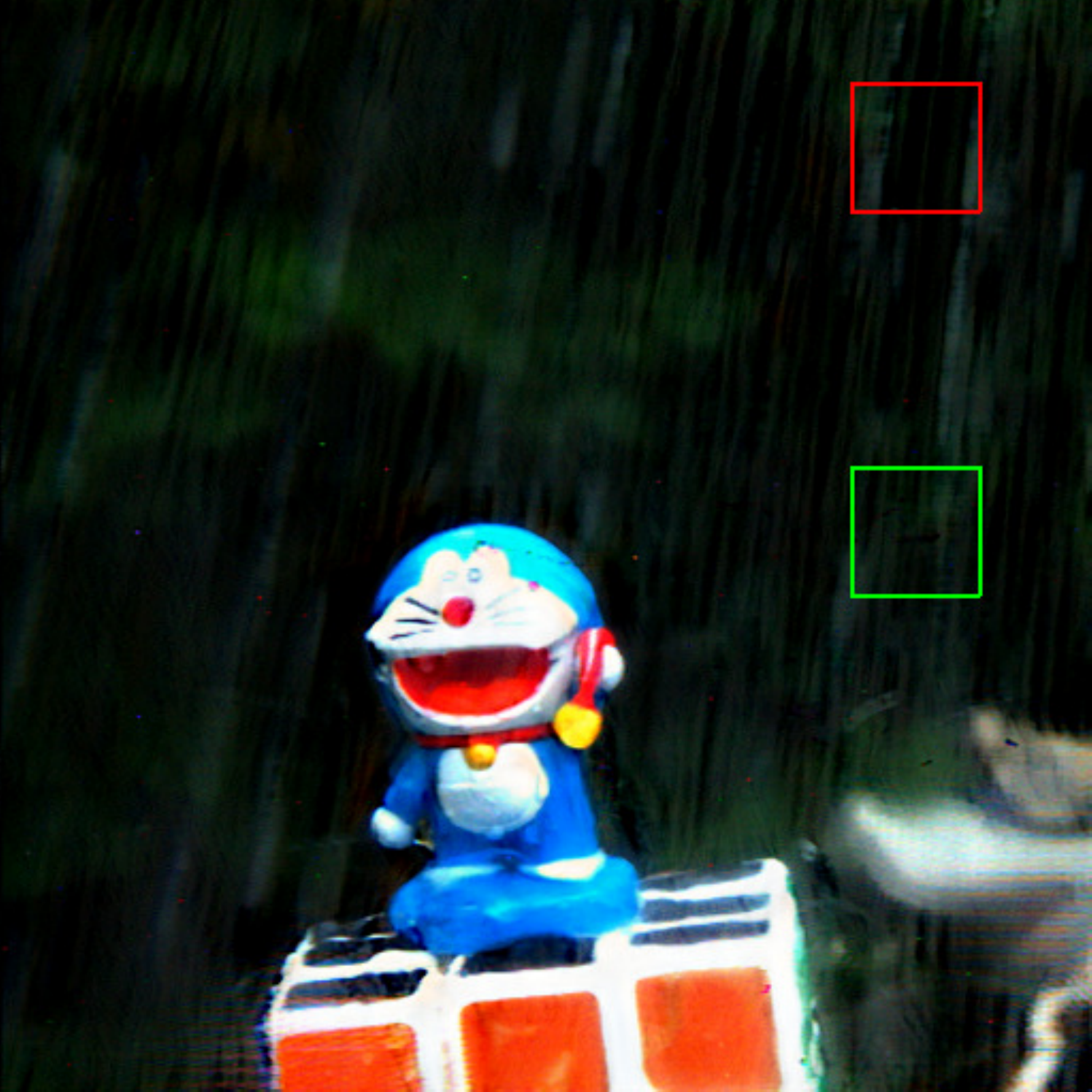}} &
            \multicolumn{2}{c}{\includegraphics[width=\subwidth\linewidth]{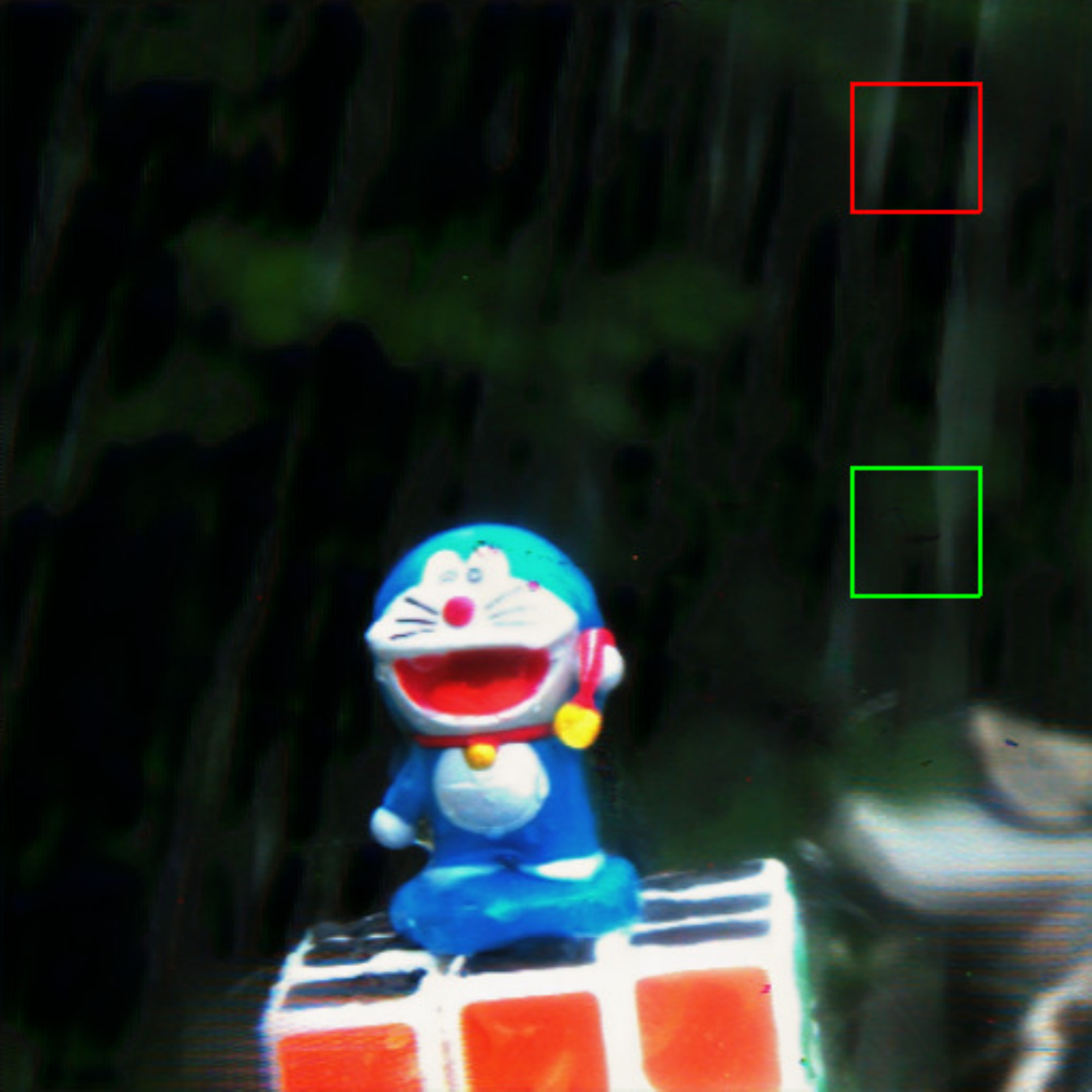}} &

            \multicolumn{2}{c}{\includegraphics[width=\subwidth\linewidth]{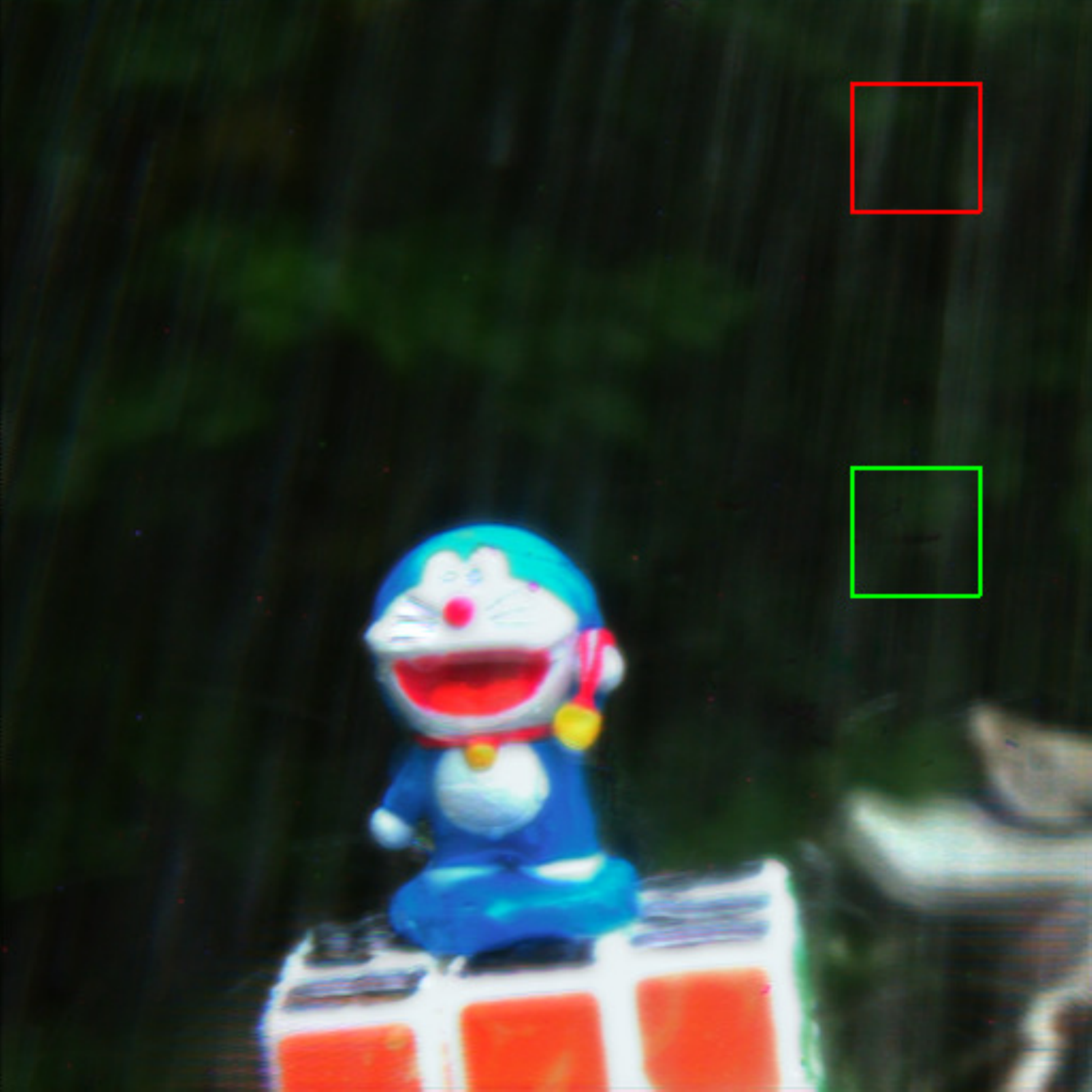}} &
            \multicolumn{2}{c}{\includegraphics[width=\subwidth\linewidth]{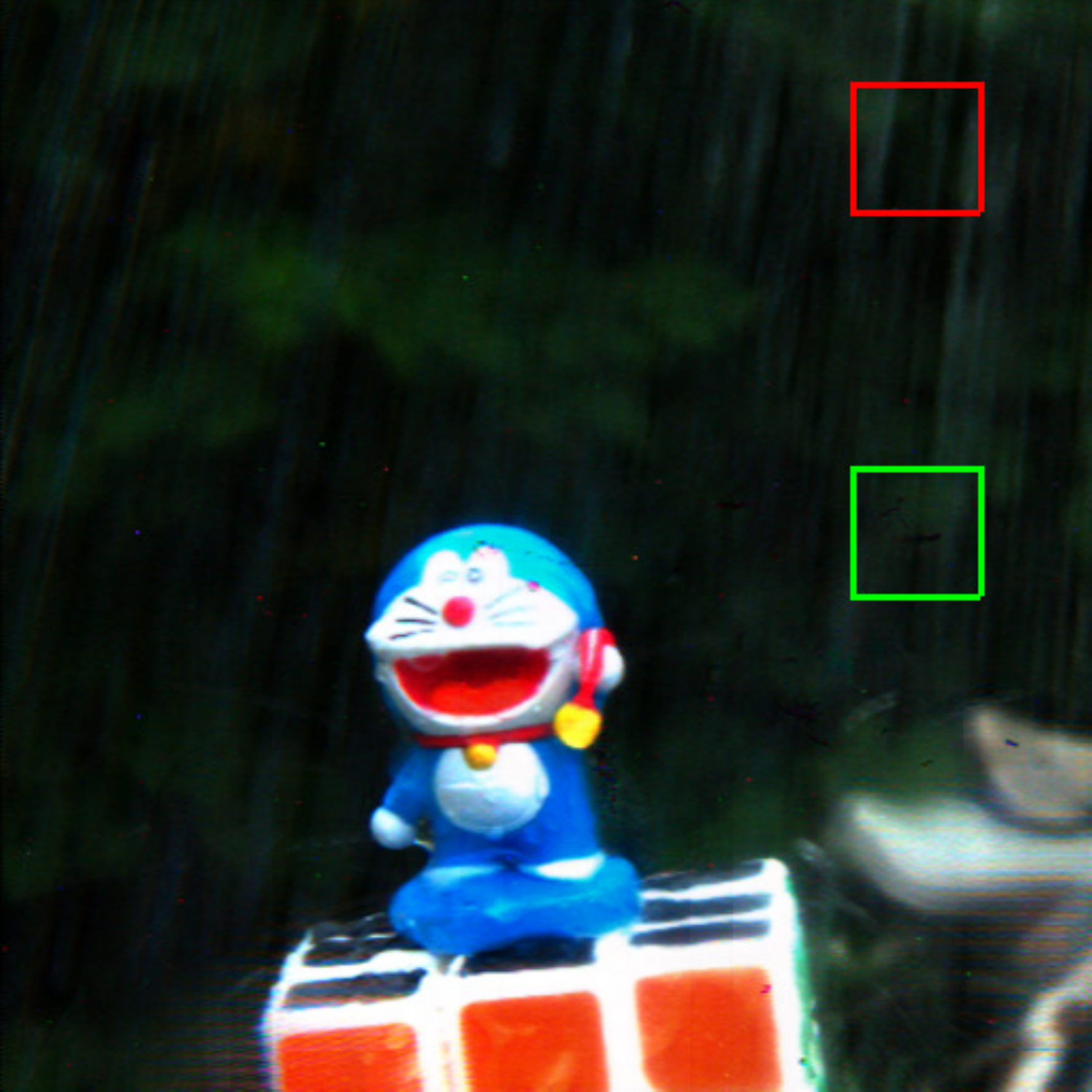}} &
            \multicolumn{2}{c}{\includegraphics[width=\subwidth\linewidth]{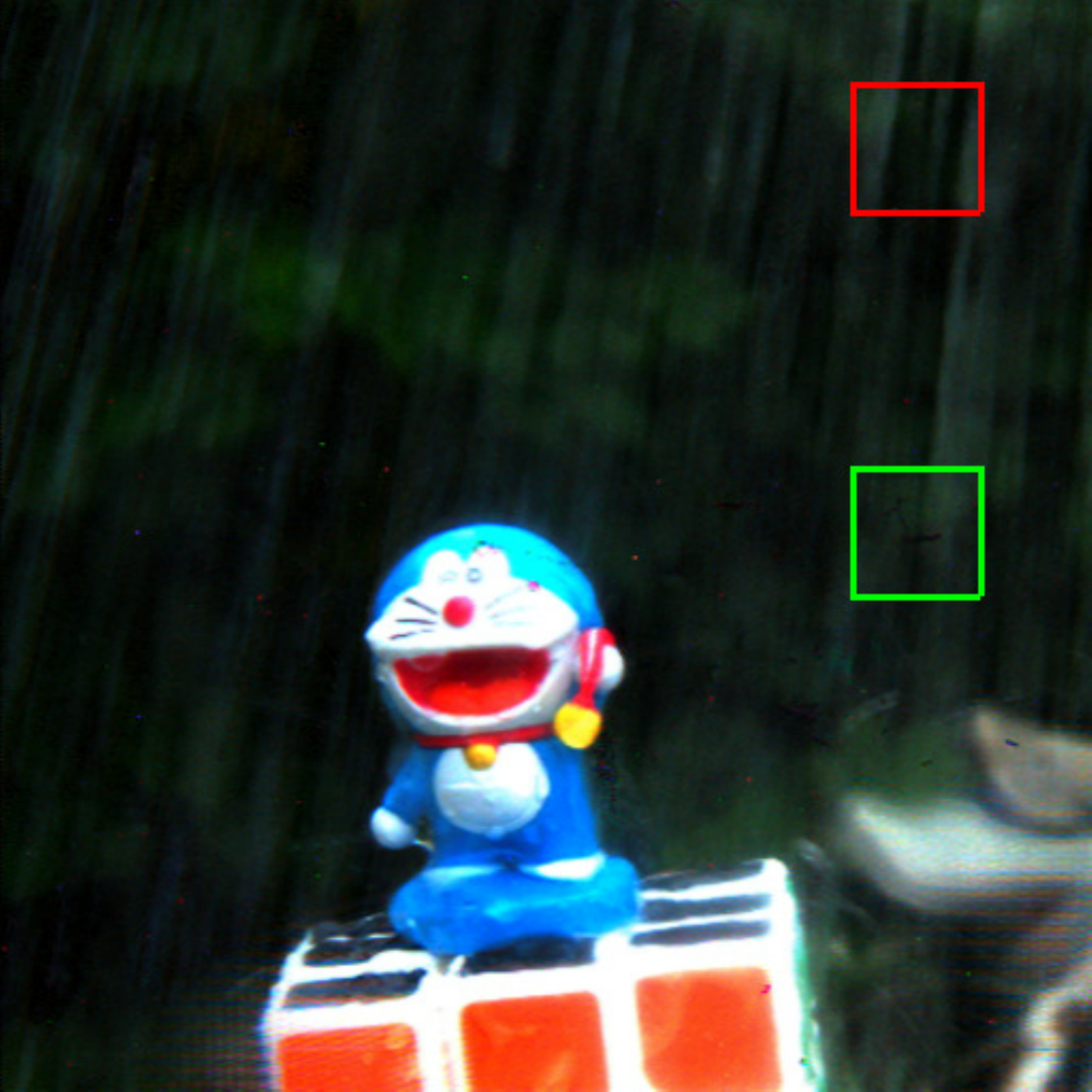}} &
            \multicolumn{2}{c}{\includegraphics[width=\subwidth\linewidth]{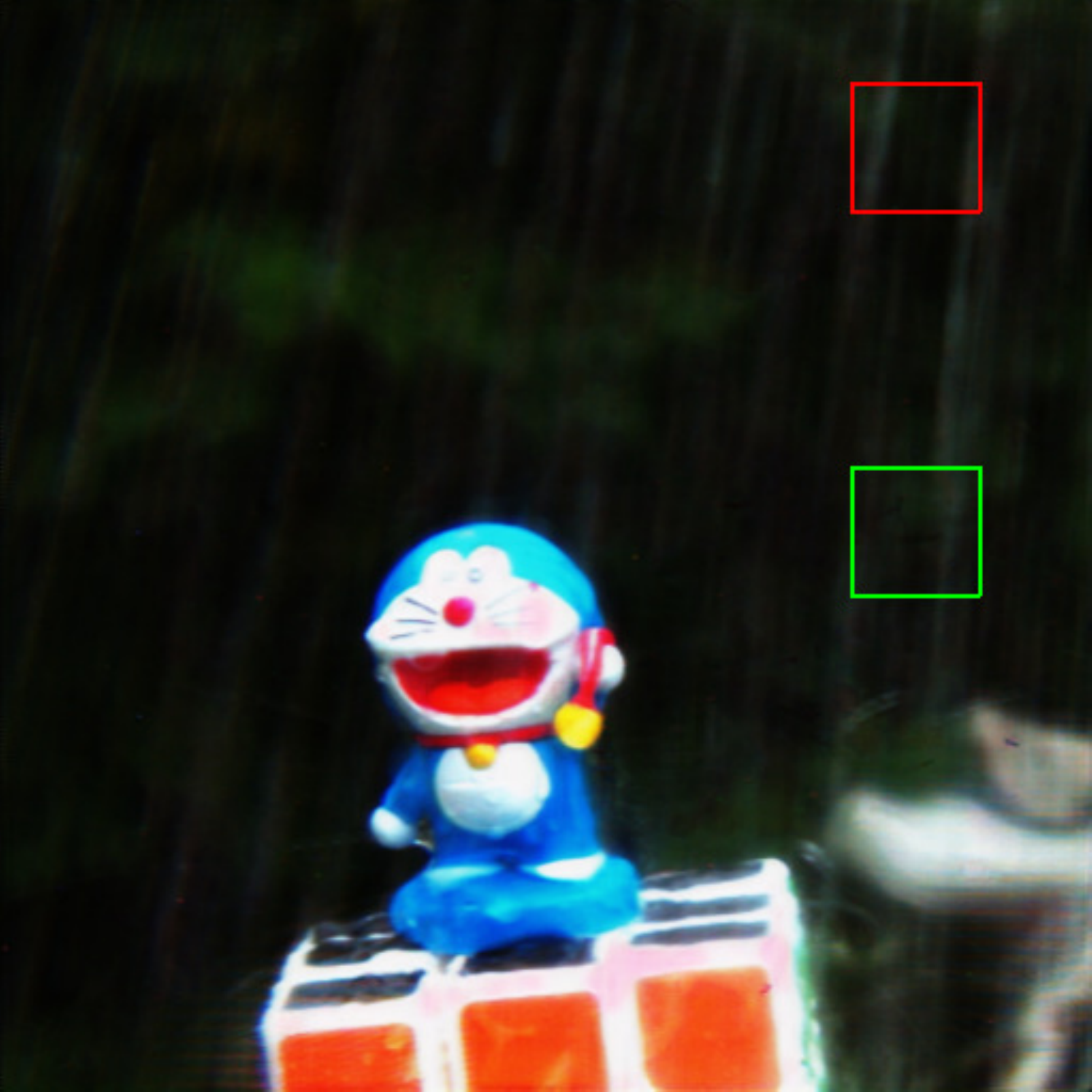}} &
            \multicolumn{2}{c}{\includegraphics[width=\subwidth\linewidth]{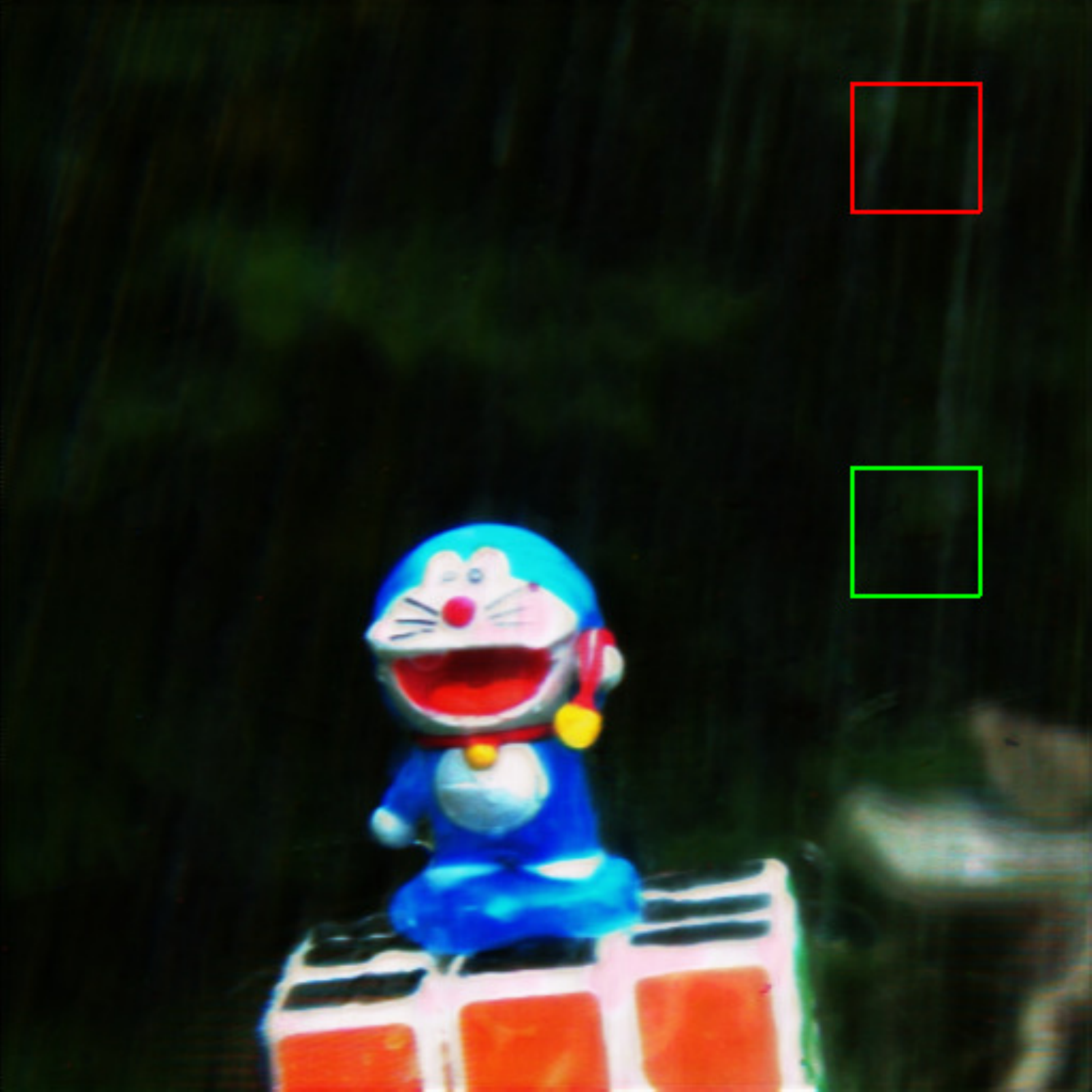}} \\

            \includegraphics[width=\ssubwidth\linewidth]{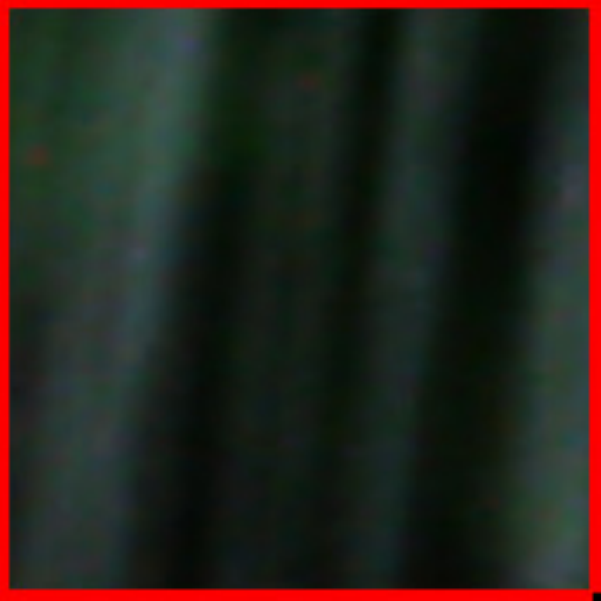} &
            \includegraphics[width=\ssubwidth\linewidth]{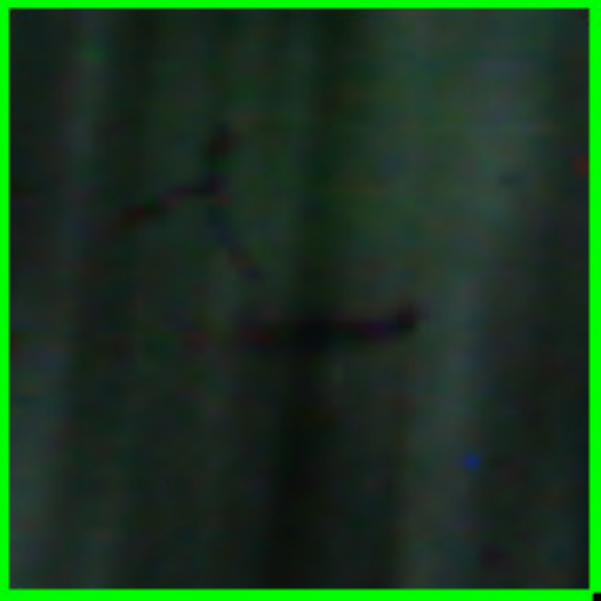} &
            \includegraphics[width=\ssubwidth\linewidth]{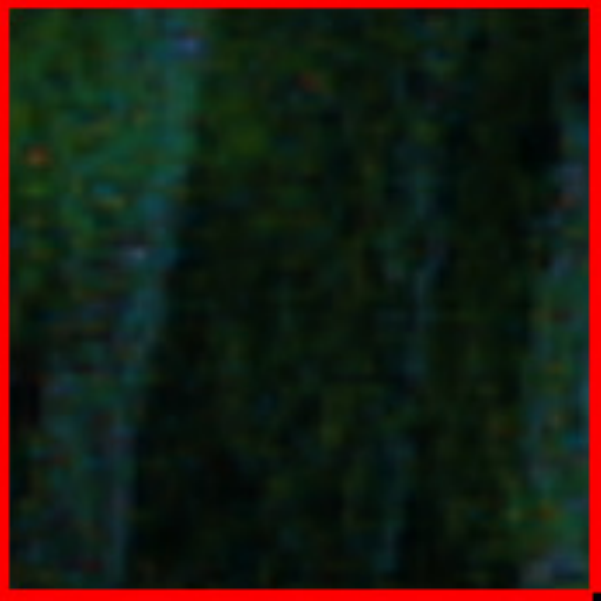} &
            \includegraphics[width=\ssubwidth\linewidth]{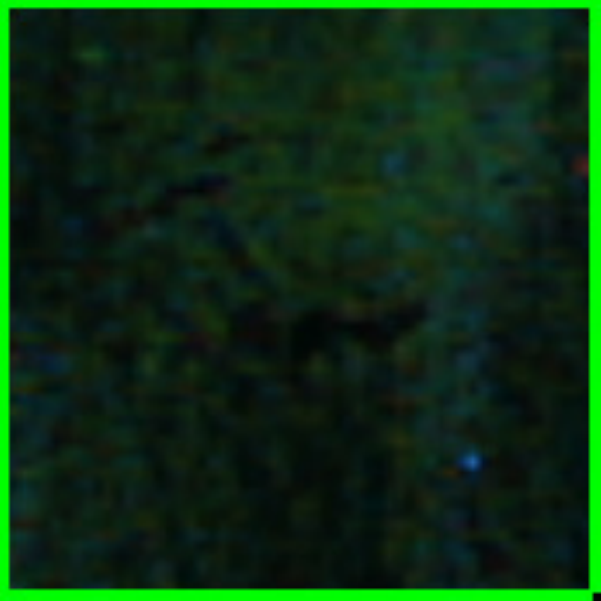} &
            \includegraphics[width=\ssubwidth\linewidth]{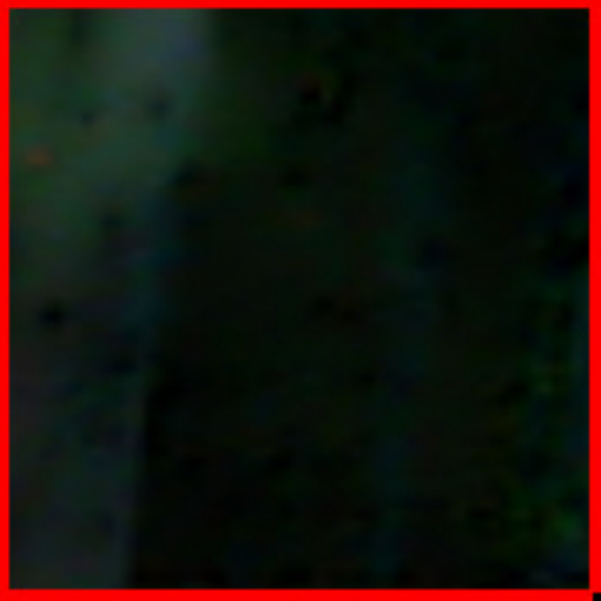} &
            \includegraphics[width=\ssubwidth\linewidth]{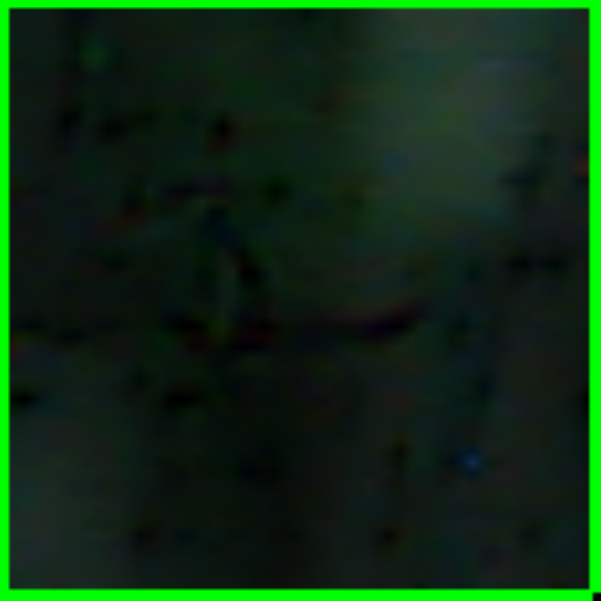} &
            \includegraphics[width=\ssubwidth\linewidth]{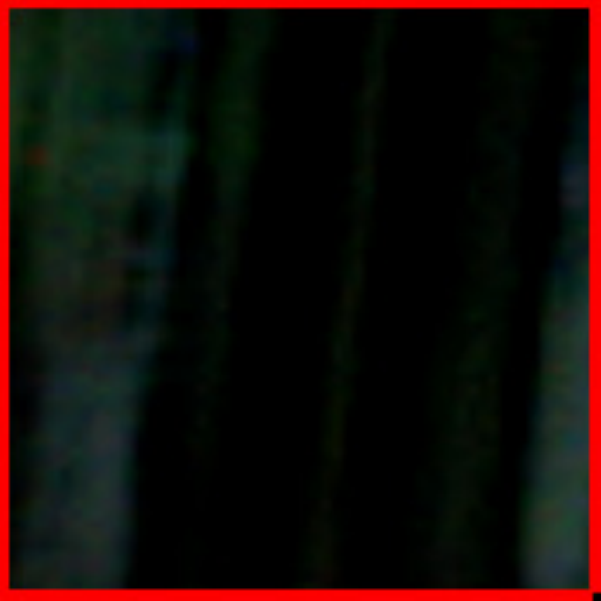} &
            \includegraphics[width=\ssubwidth\linewidth]{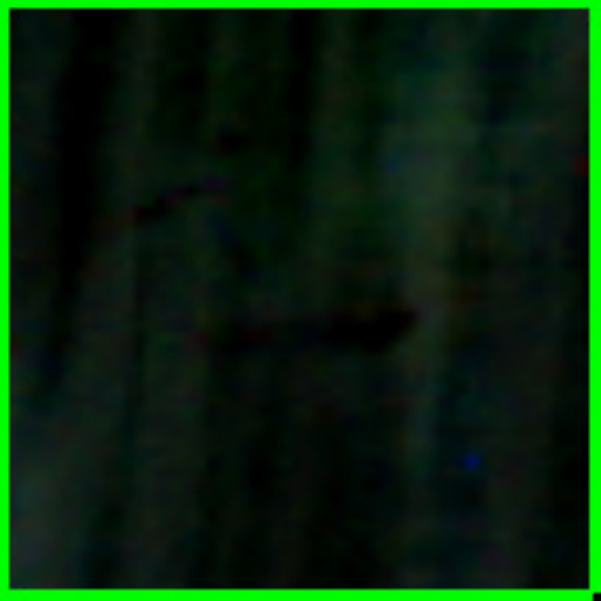} &
            \includegraphics[width=\ssubwidth\linewidth]{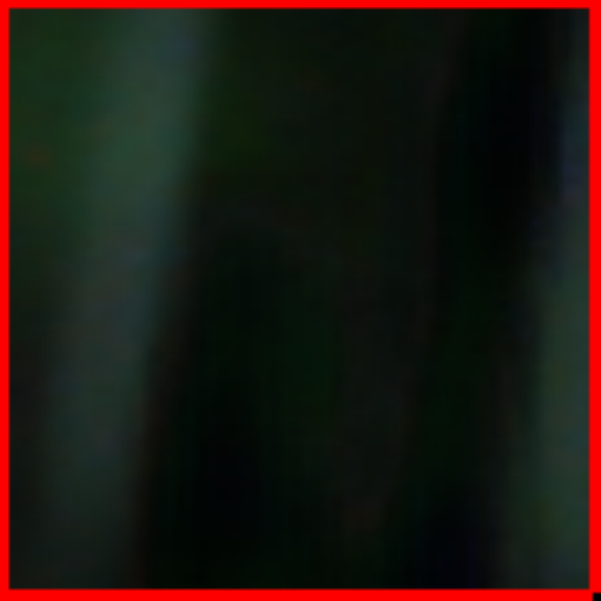} &
            \includegraphics[width=\ssubwidth\linewidth]{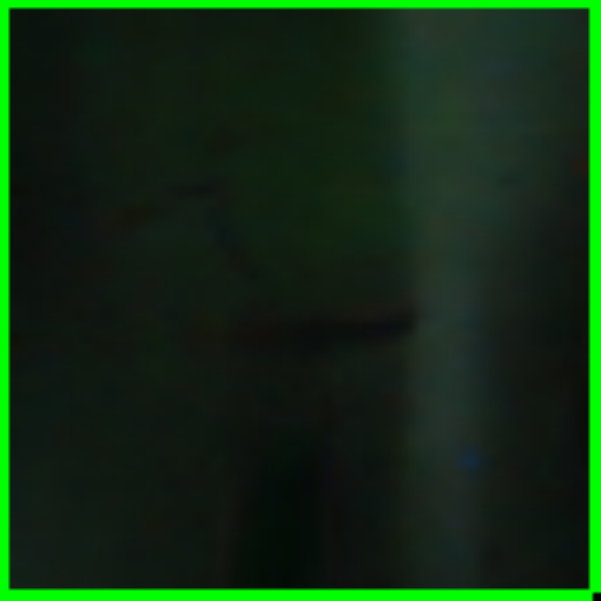} &

            \includegraphics[width=\ssubwidth\linewidth]{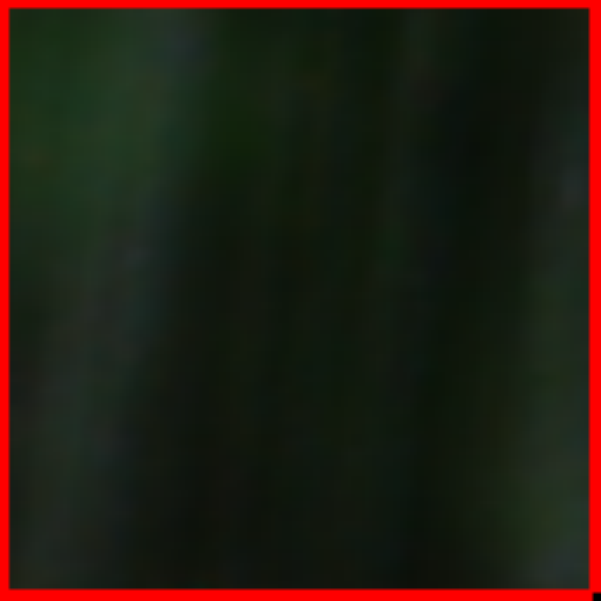} &
            \includegraphics[width=\ssubwidth\linewidth]{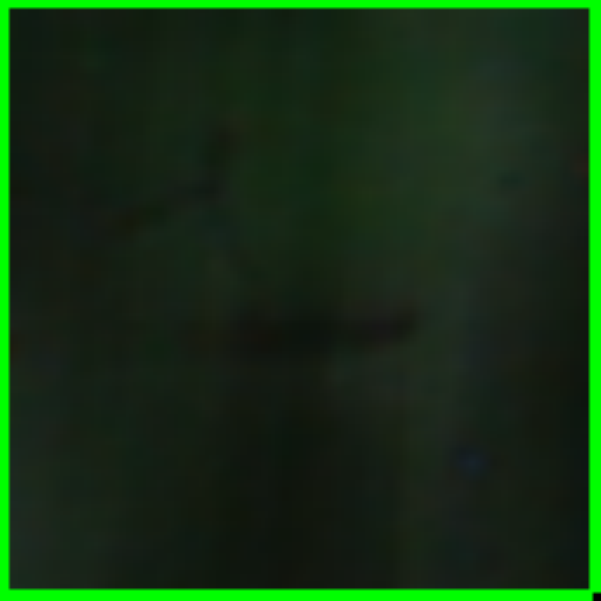} &
            \includegraphics[width=\ssubwidth\linewidth]{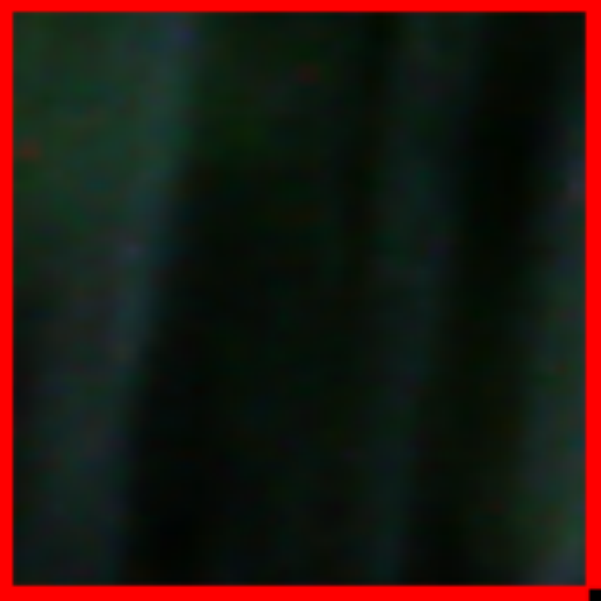} &
            \includegraphics[width=\ssubwidth\linewidth]{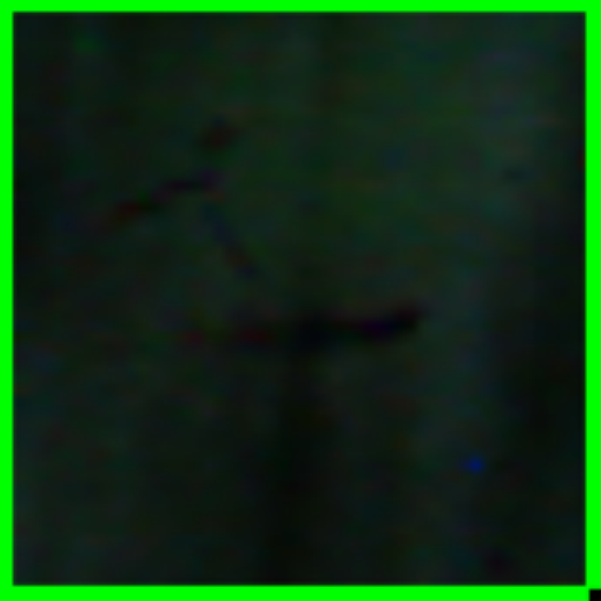} &
            \includegraphics[width=\ssubwidth\linewidth]{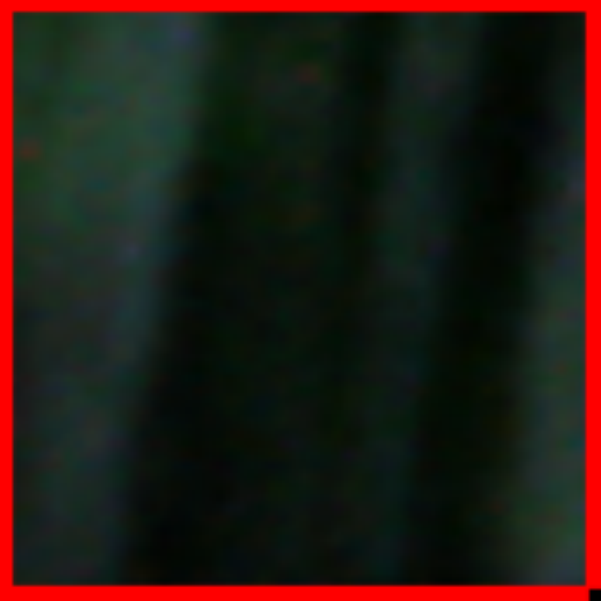} &
            \includegraphics[width=\ssubwidth\linewidth]{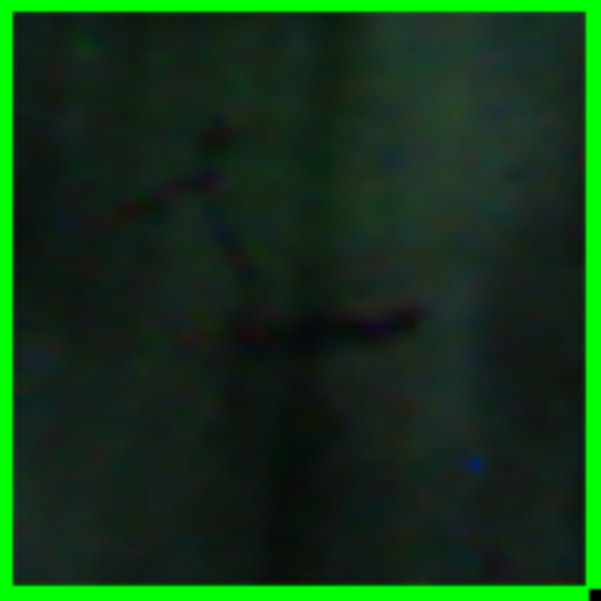} &
            \includegraphics[width=\ssubwidth\linewidth]{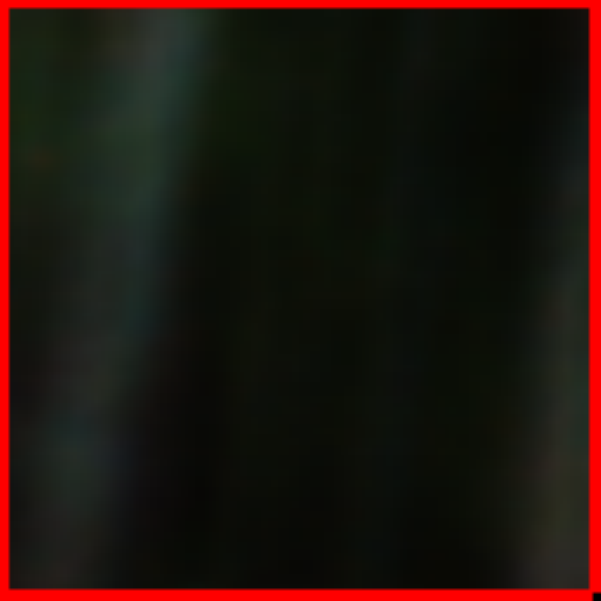} &
            \includegraphics[width=\ssubwidth\linewidth]{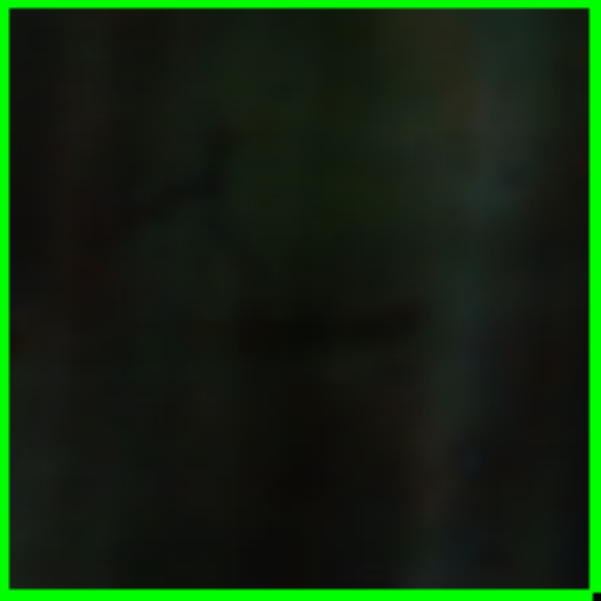} &
            \includegraphics[width=\ssubwidth\linewidth]{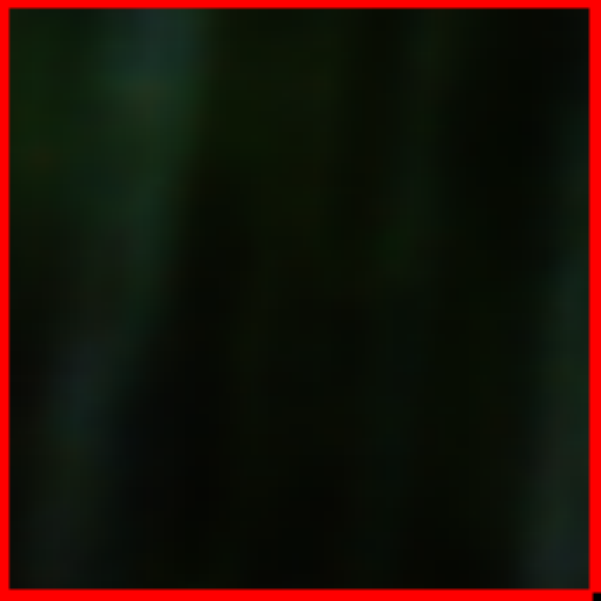} &
            \includegraphics[width=\ssubwidth\linewidth]{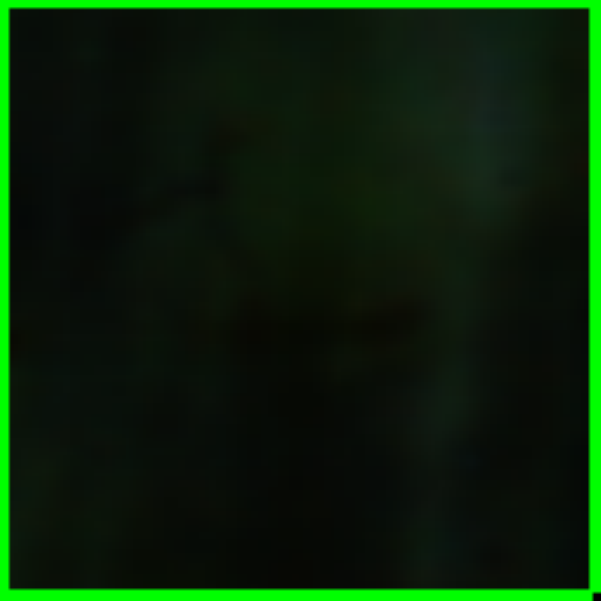} \\

            \multicolumn{2}{c}{\includegraphics[width=\subwidth\linewidth]{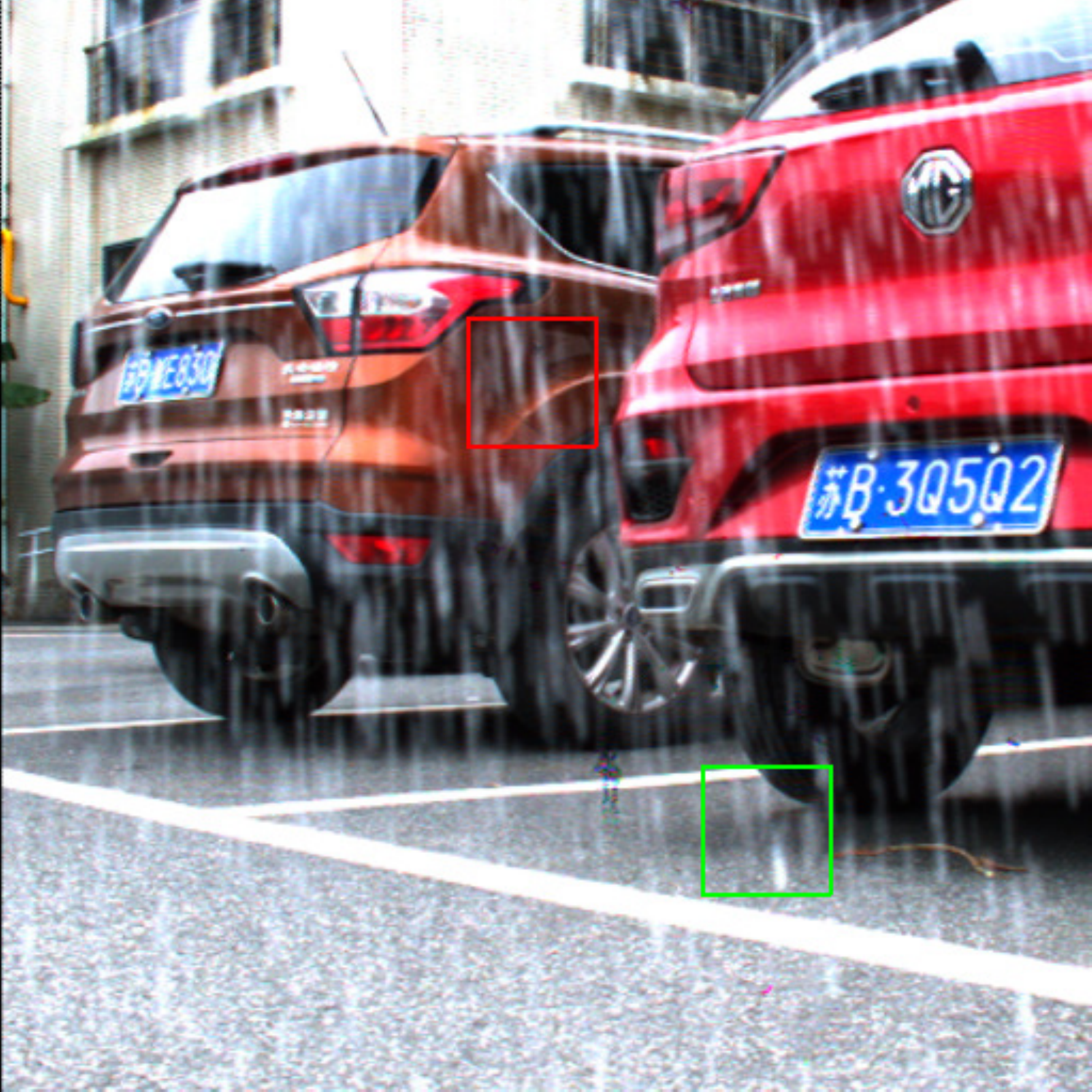}} &
            \multicolumn{2}{c}{\includegraphics[width=\subwidth\linewidth]{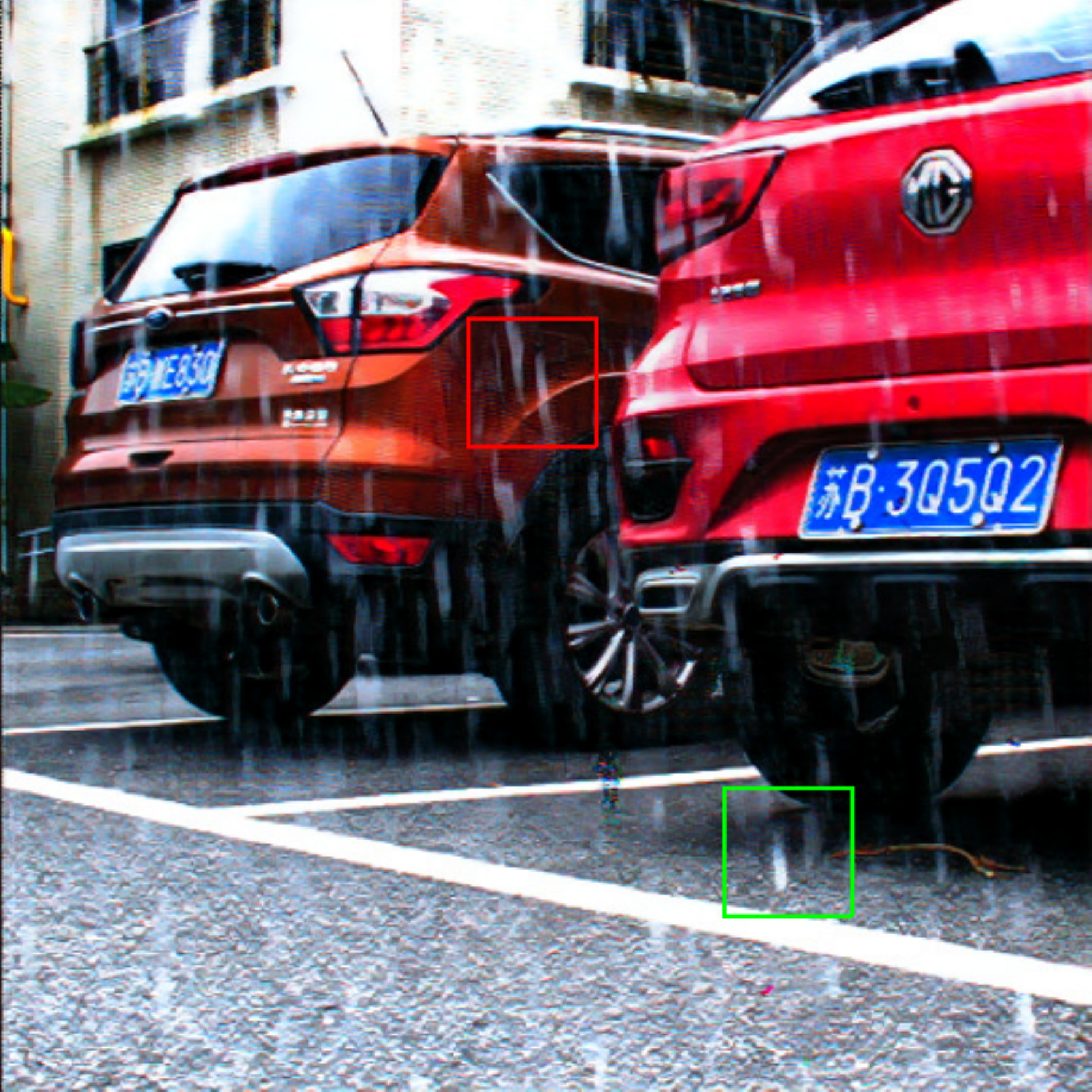}} &
            \multicolumn{2}{c}{\includegraphics[width=\subwidth\linewidth]{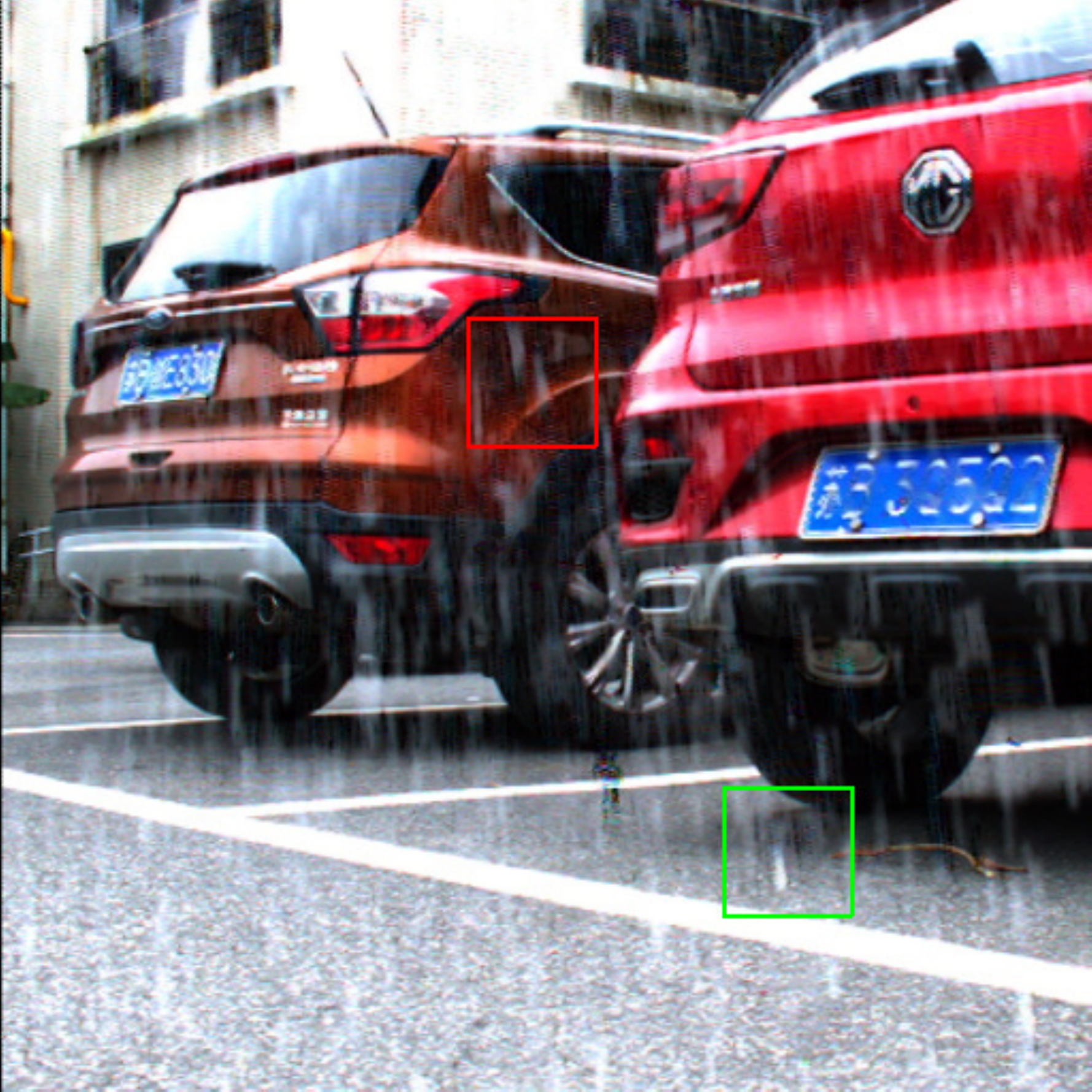}} &
            \multicolumn{2}{c}{\includegraphics[width=\subwidth\linewidth]{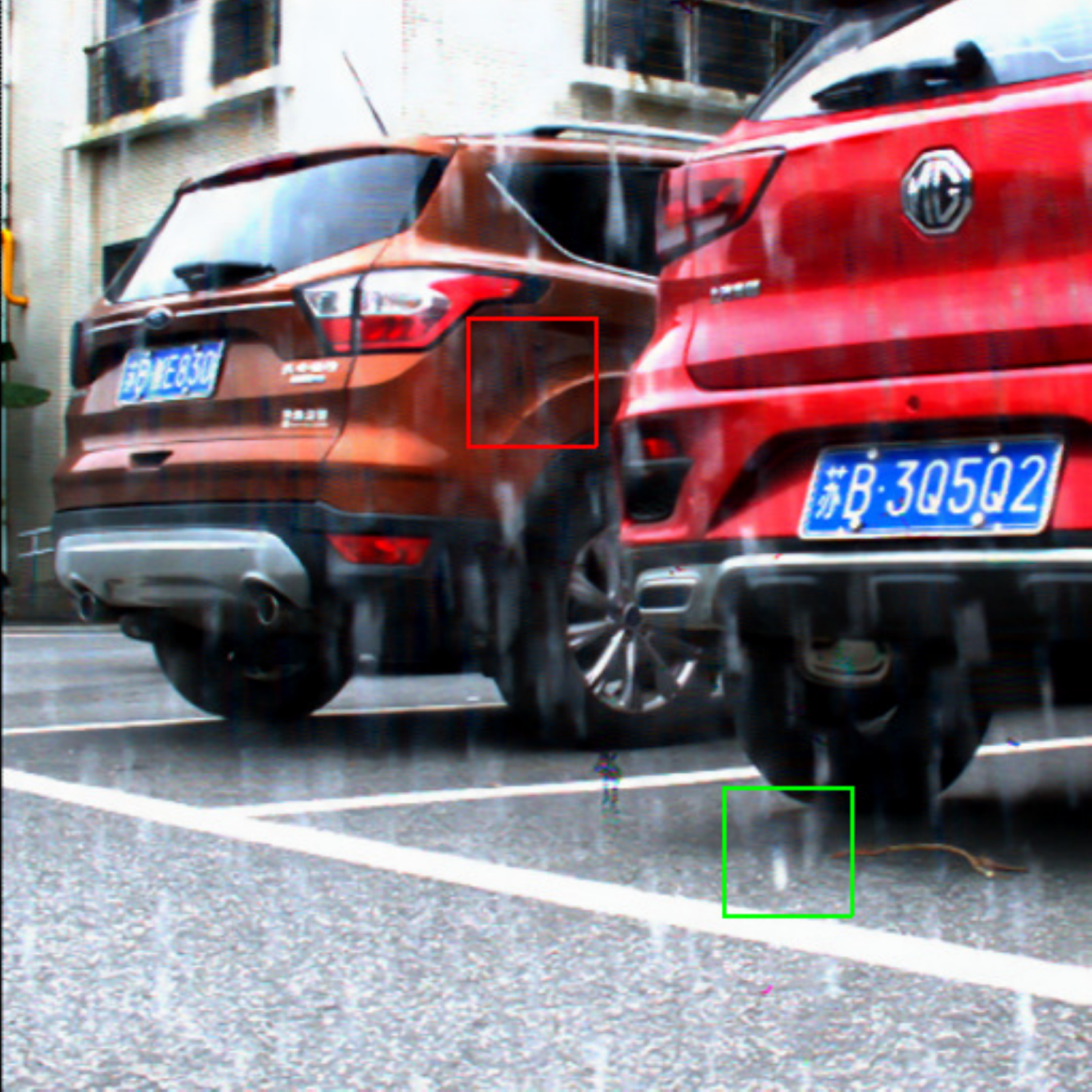}} &
            \multicolumn{2}{c}{\includegraphics[width=\subwidth\linewidth]{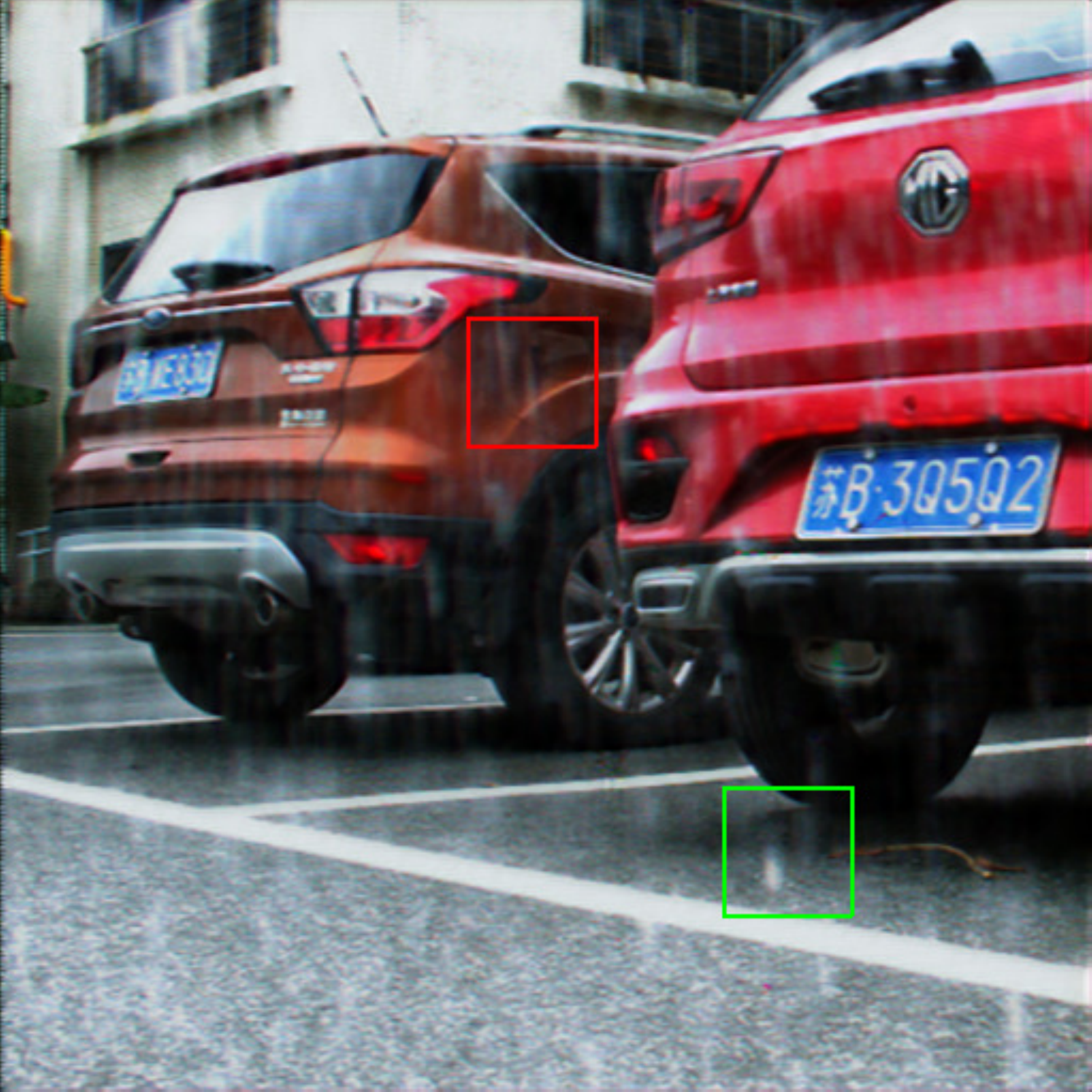}} &

            \multicolumn{2}{c}{\includegraphics[width=\subwidth\linewidth]{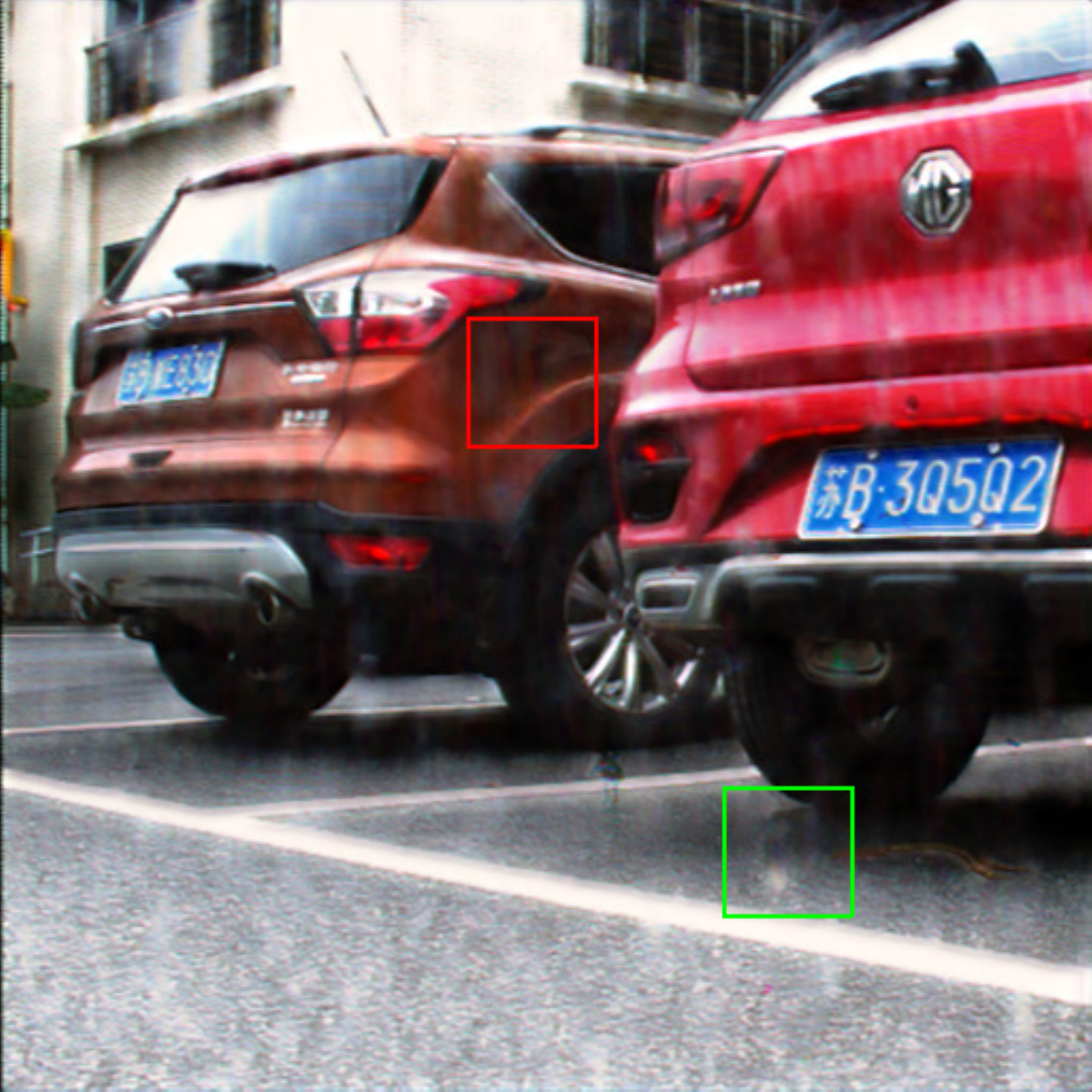}} &
            \multicolumn{2}{c}{\includegraphics[width=\subwidth\linewidth]{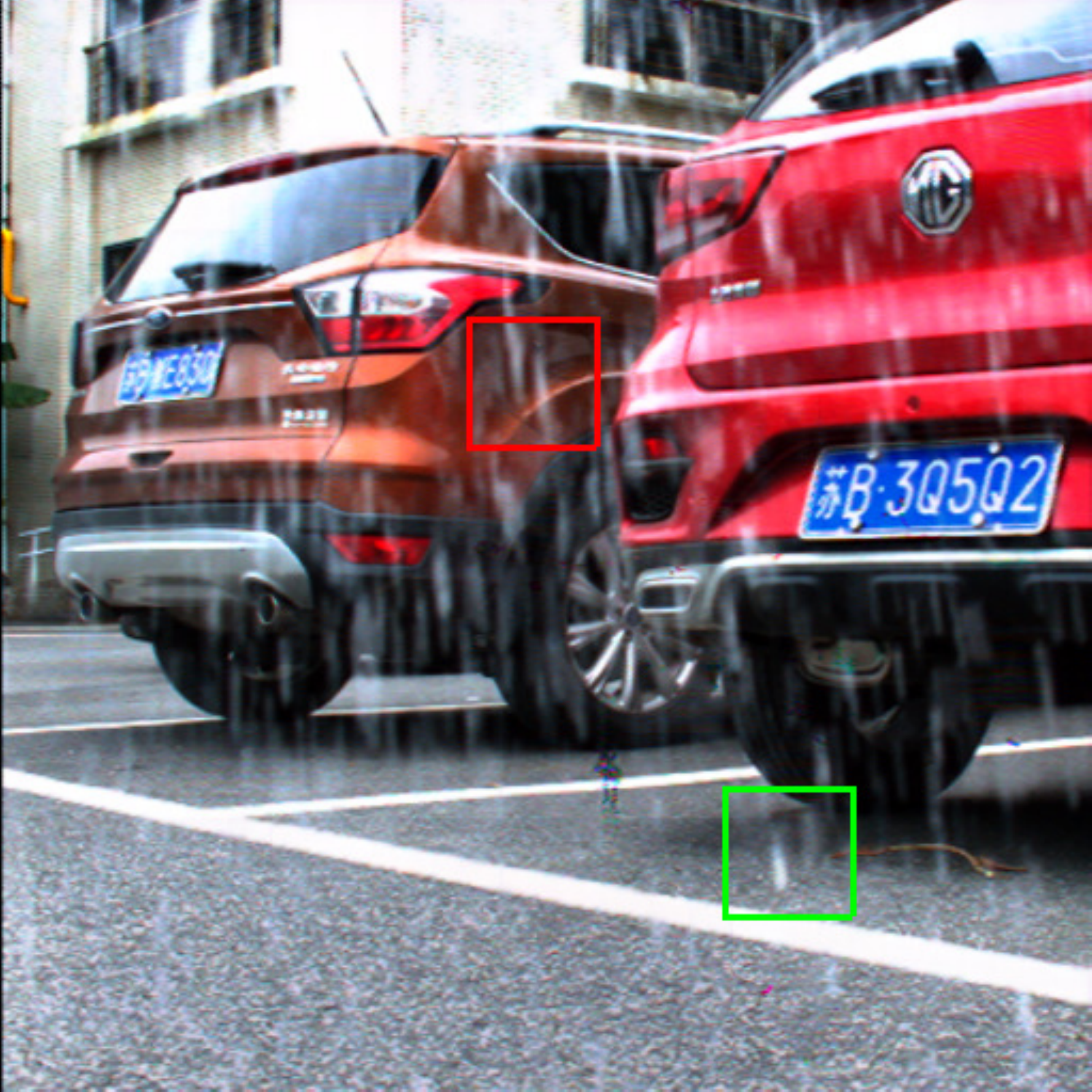}} &
            \multicolumn{2}{c}{\includegraphics[width=\subwidth\linewidth]{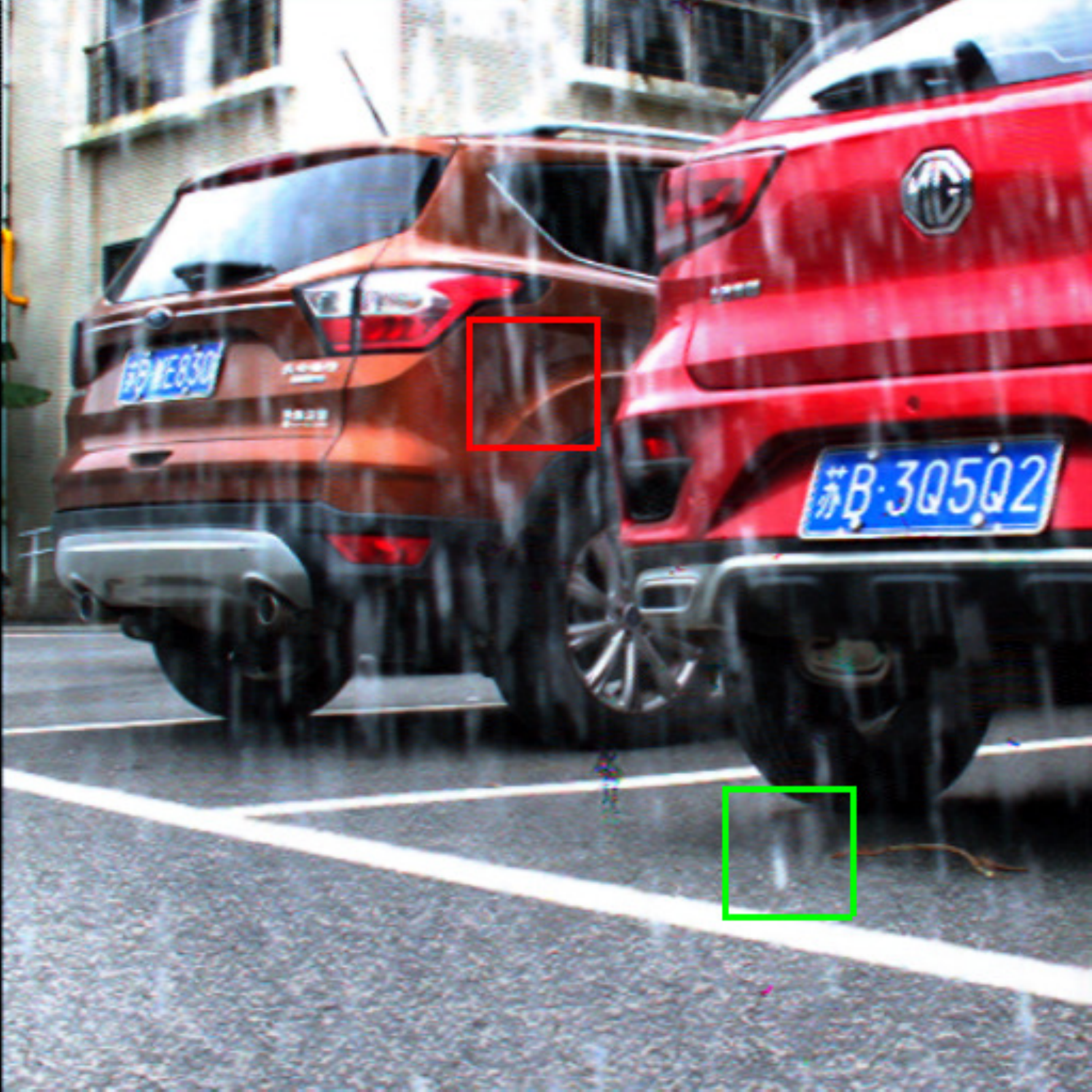}} &
            \multicolumn{2}{c}{\includegraphics[width=\subwidth\linewidth]{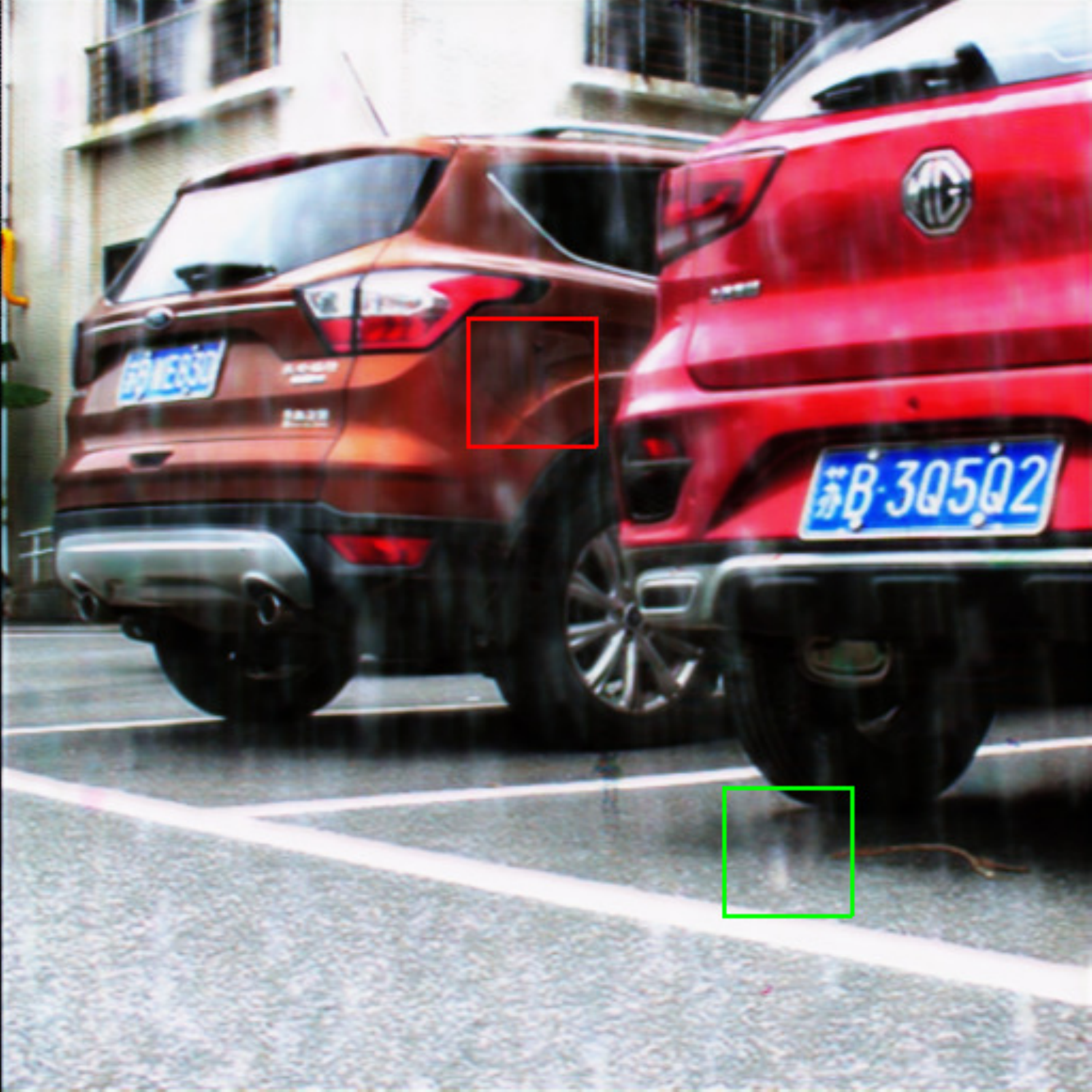}} &
            \multicolumn{2}{c}{\includegraphics[width=\subwidth\linewidth]{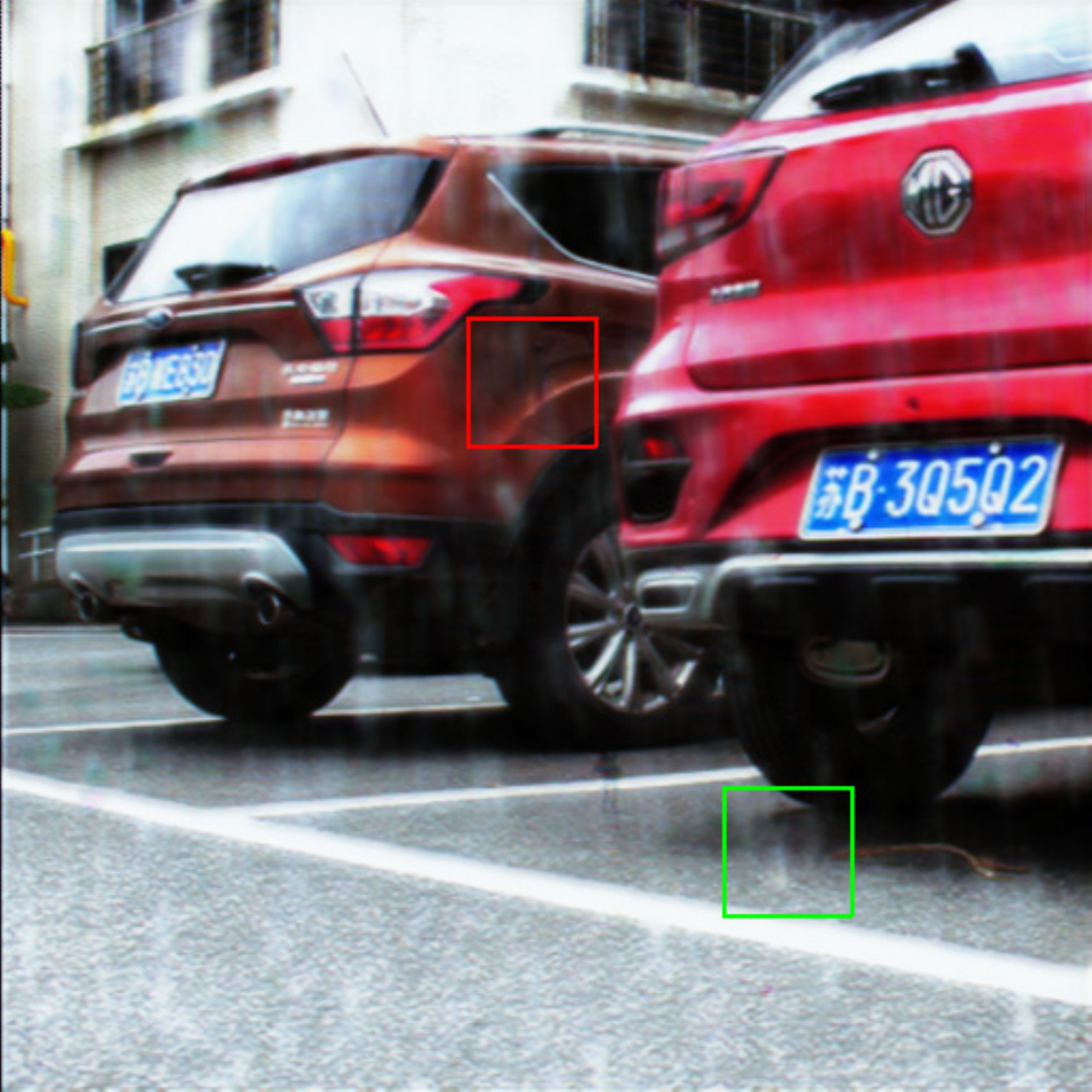}} \\

            \includegraphics[width=\ssubwidth\linewidth]{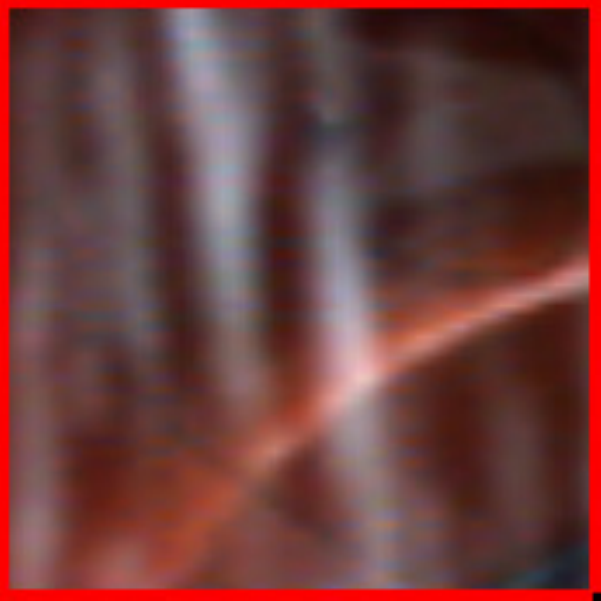} &
            \includegraphics[width=\ssubwidth\linewidth]{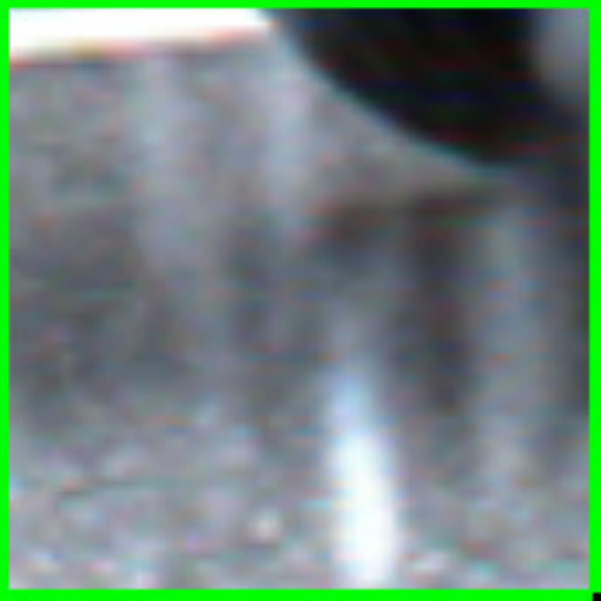} &
            \includegraphics[width=\ssubwidth\linewidth]{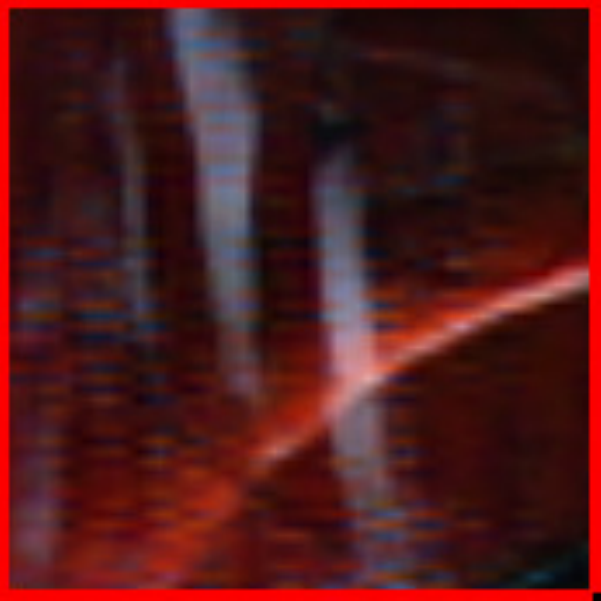} &
            \includegraphics[width=\ssubwidth\linewidth]{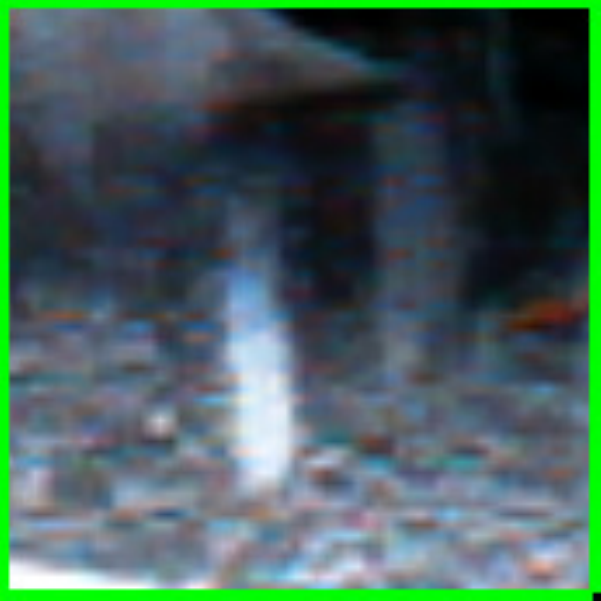} &
            \includegraphics[width=\ssubwidth\linewidth]{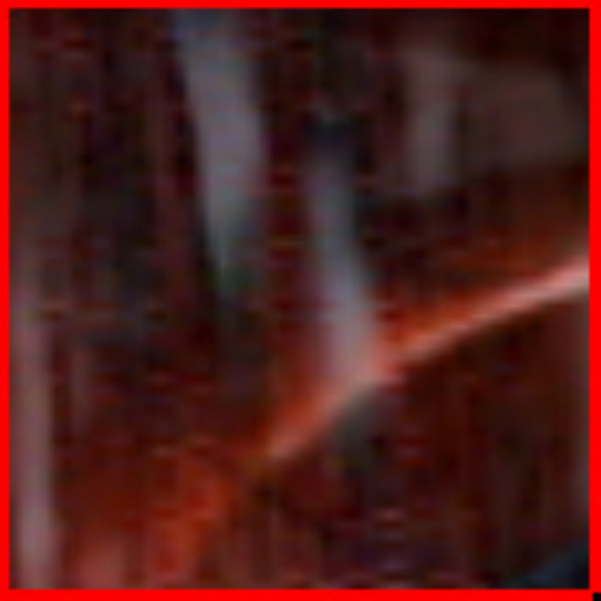} &
            \includegraphics[width=\ssubwidth\linewidth]{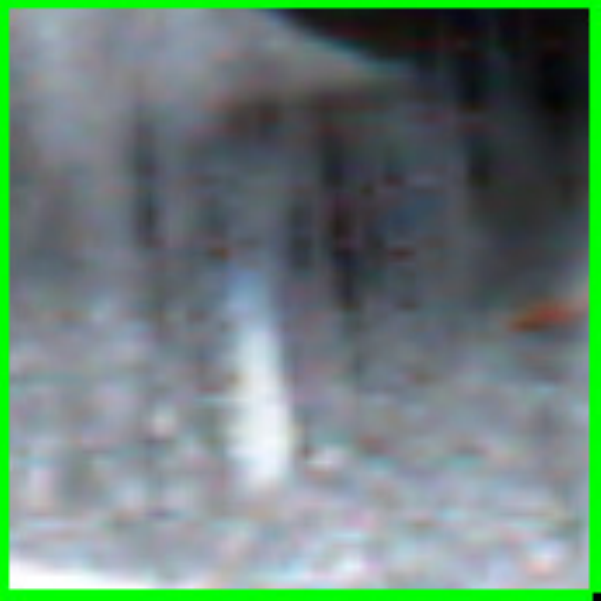} &
            \includegraphics[width=\ssubwidth\linewidth]{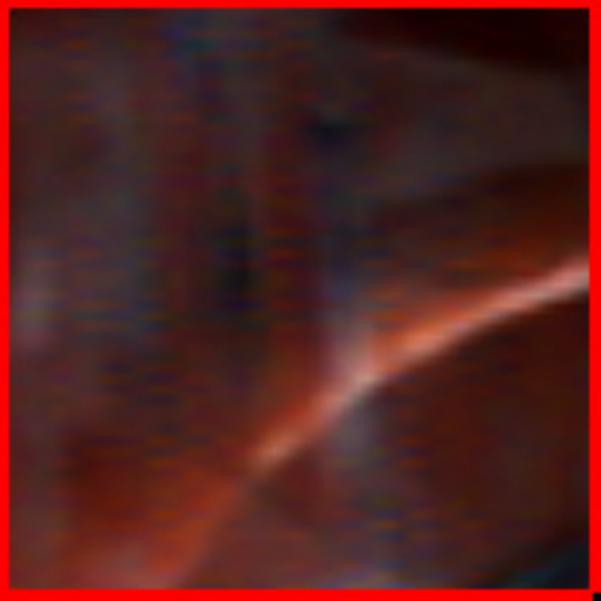} &
            \includegraphics[width=\ssubwidth\linewidth]{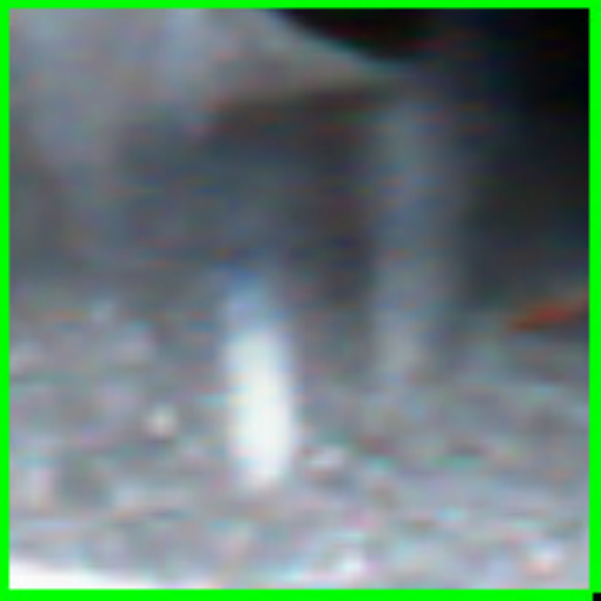} &
            \includegraphics[width=\ssubwidth\linewidth]{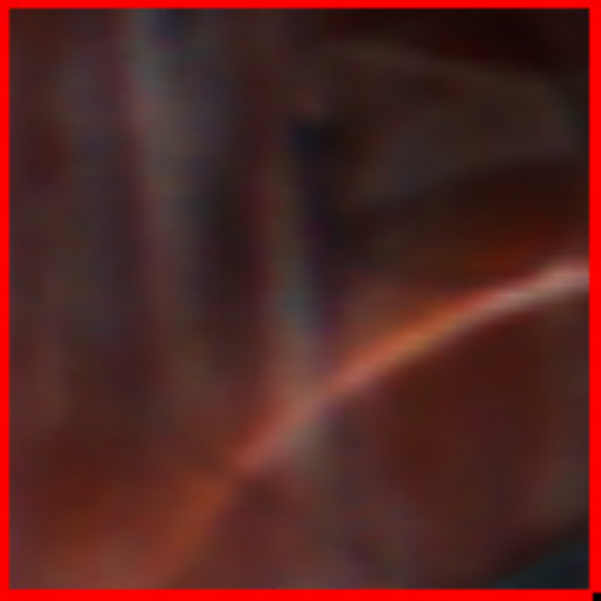} &
            \includegraphics[width=\ssubwidth\linewidth]{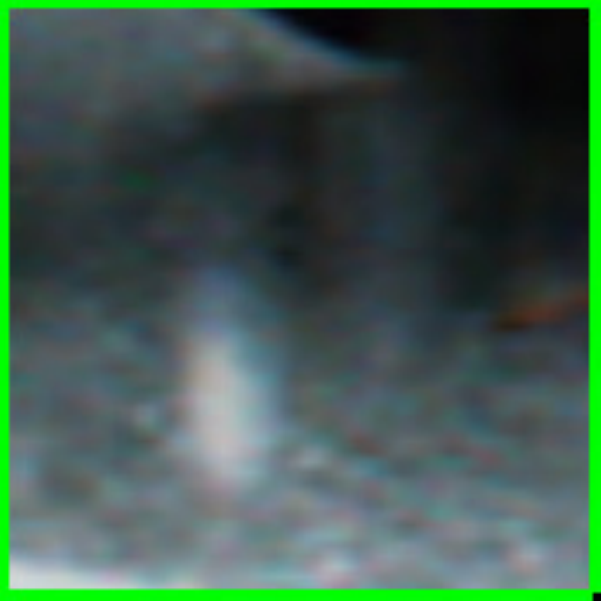} &

            \includegraphics[width=\ssubwidth\linewidth]{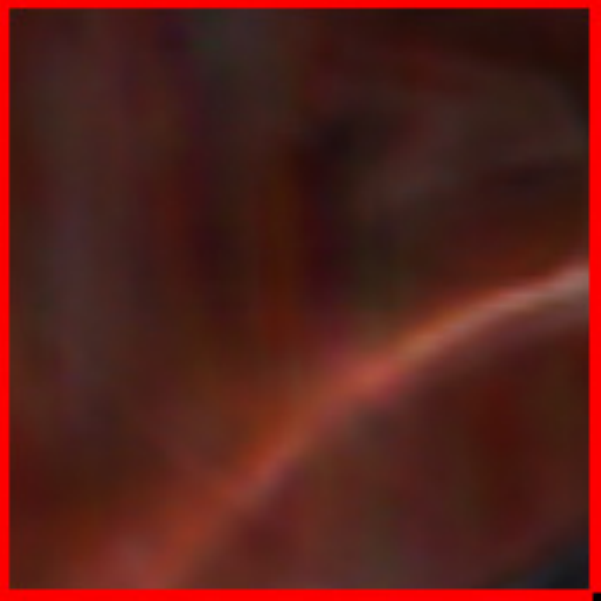} &
            \includegraphics[width=\ssubwidth\linewidth]{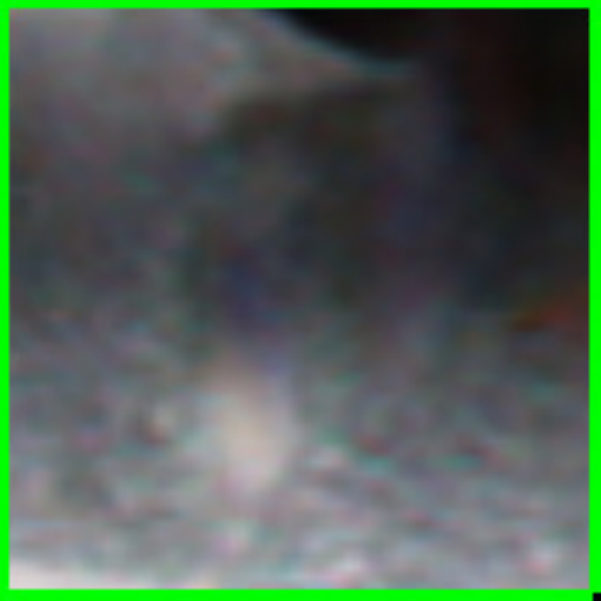} &
            \includegraphics[width=\ssubwidth\linewidth]{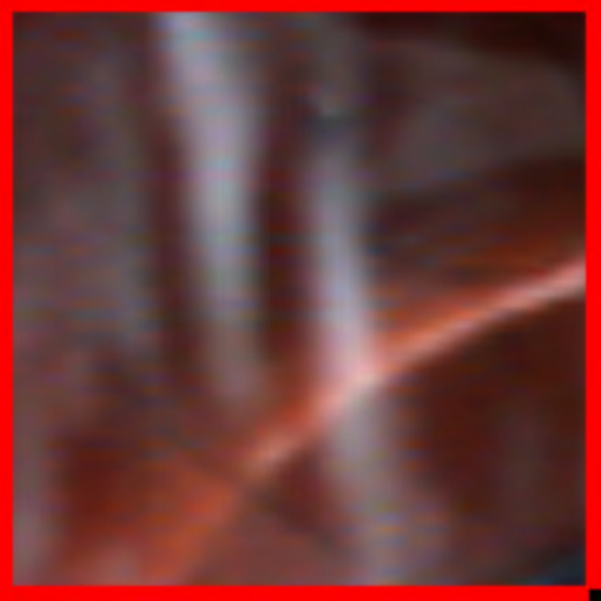} &
            \includegraphics[width=\ssubwidth\linewidth]{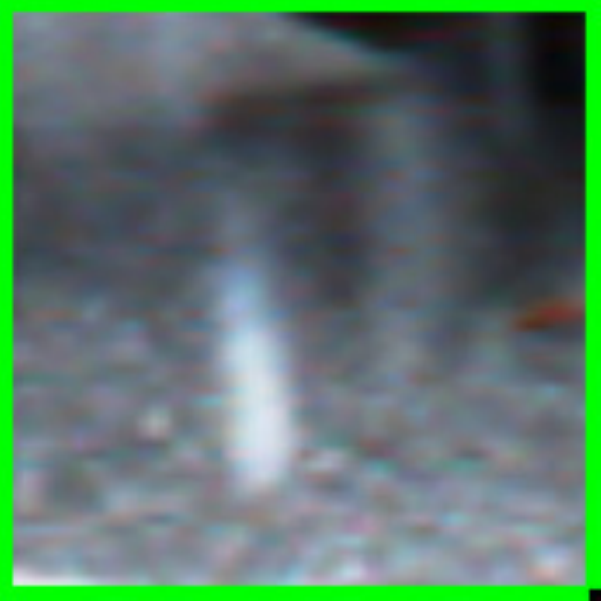} &
            \includegraphics[width=\ssubwidth\linewidth]{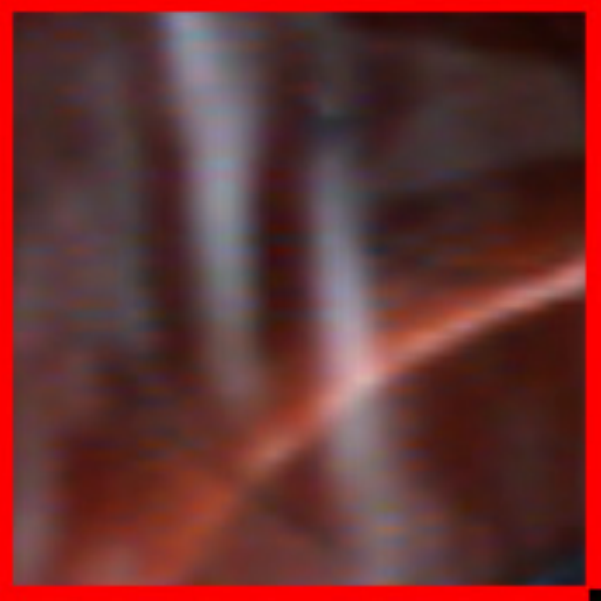} &
            \includegraphics[width=\ssubwidth\linewidth]{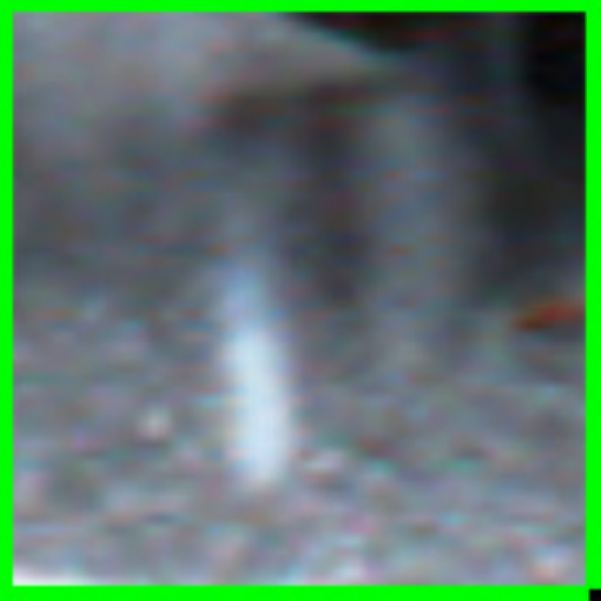} &
            \includegraphics[width=\ssubwidth\linewidth]{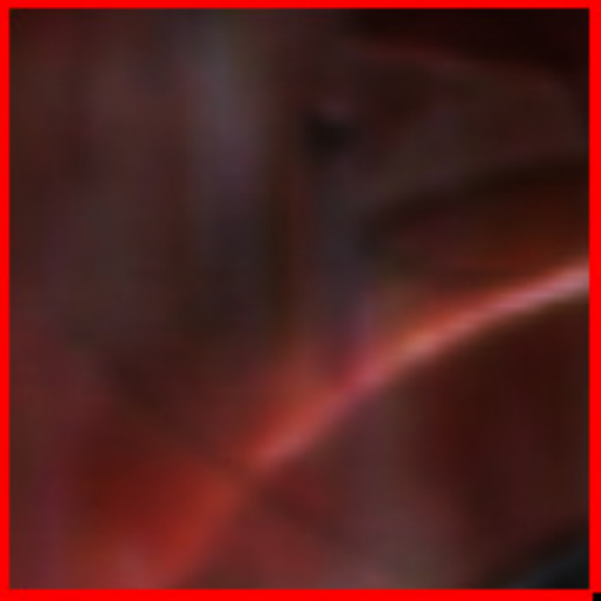} &
            \includegraphics[width=\ssubwidth\linewidth]{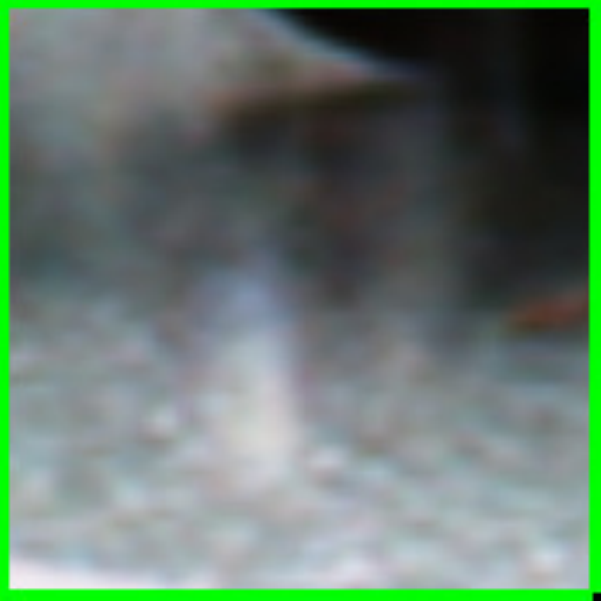} &
            \includegraphics[width=\ssubwidth\linewidth]{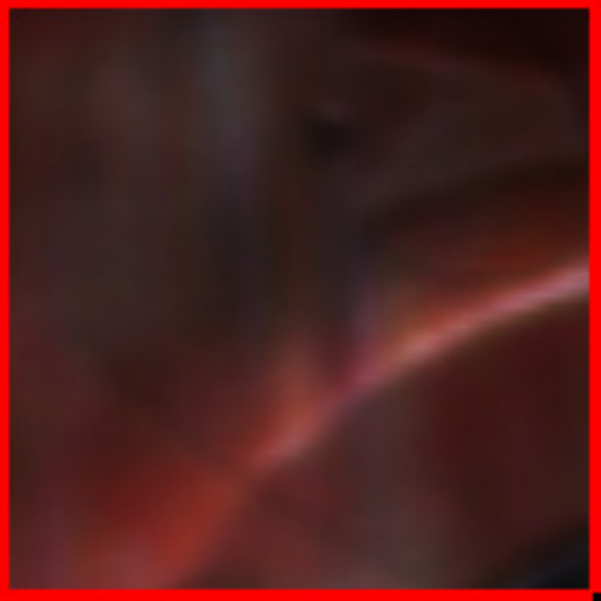} &
            \includegraphics[width=\ssubwidth\linewidth]{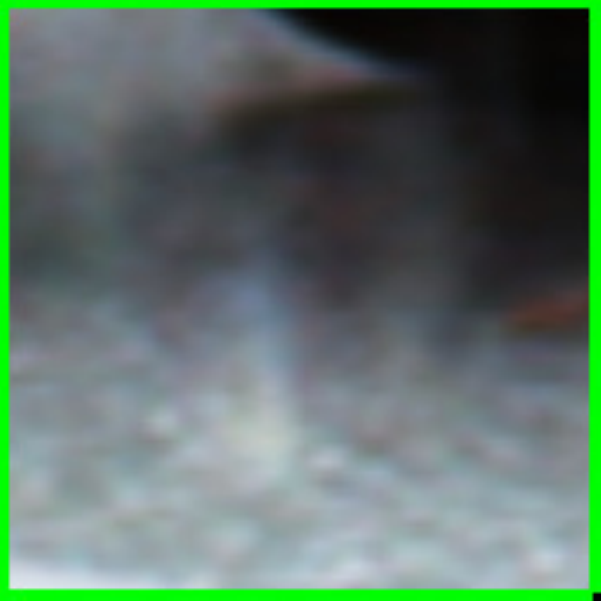} \\

            \multicolumn{2}{c}{\includegraphics[width=\subwidth\linewidth]{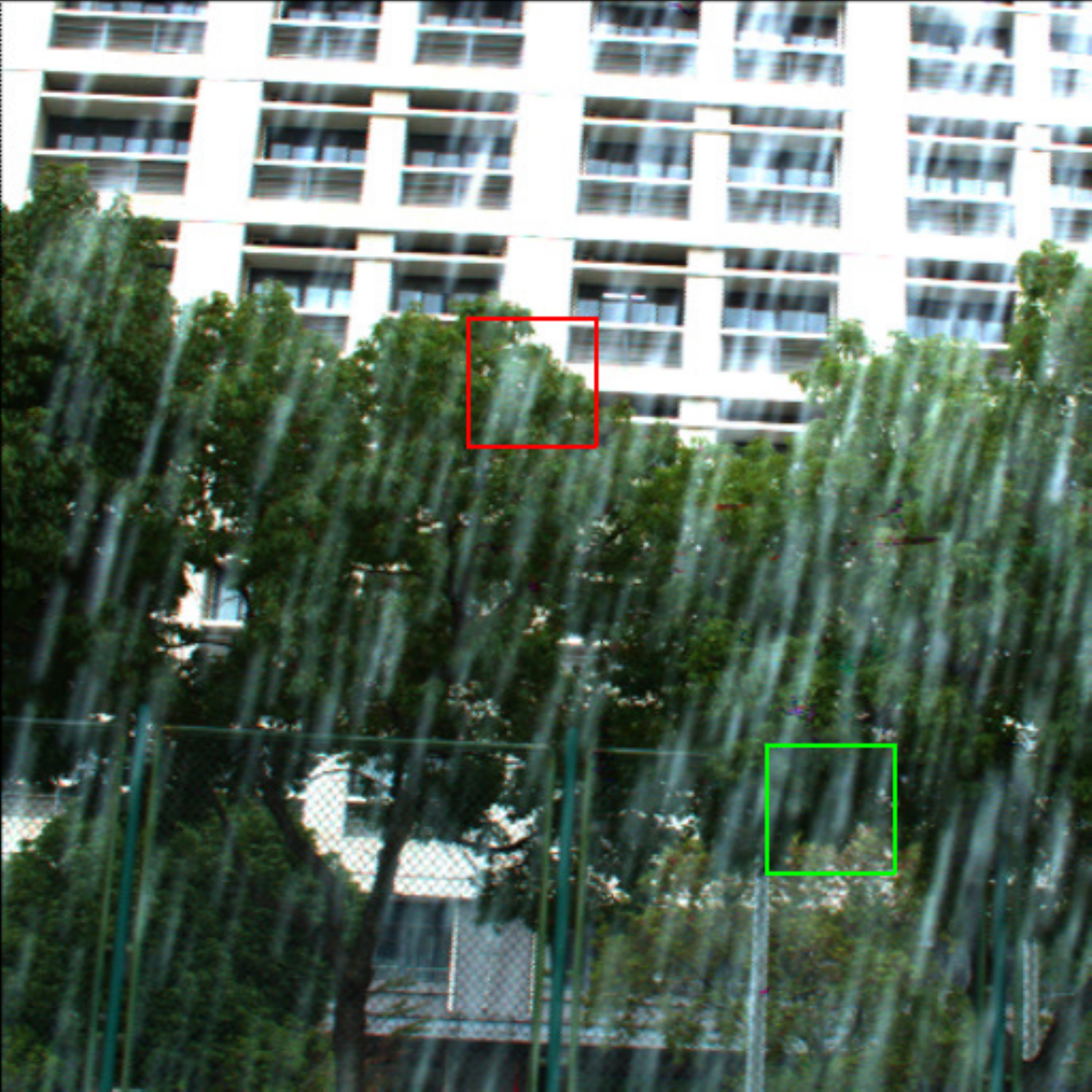}} &
            \multicolumn{2}{c}{\includegraphics[width=\subwidth\linewidth]{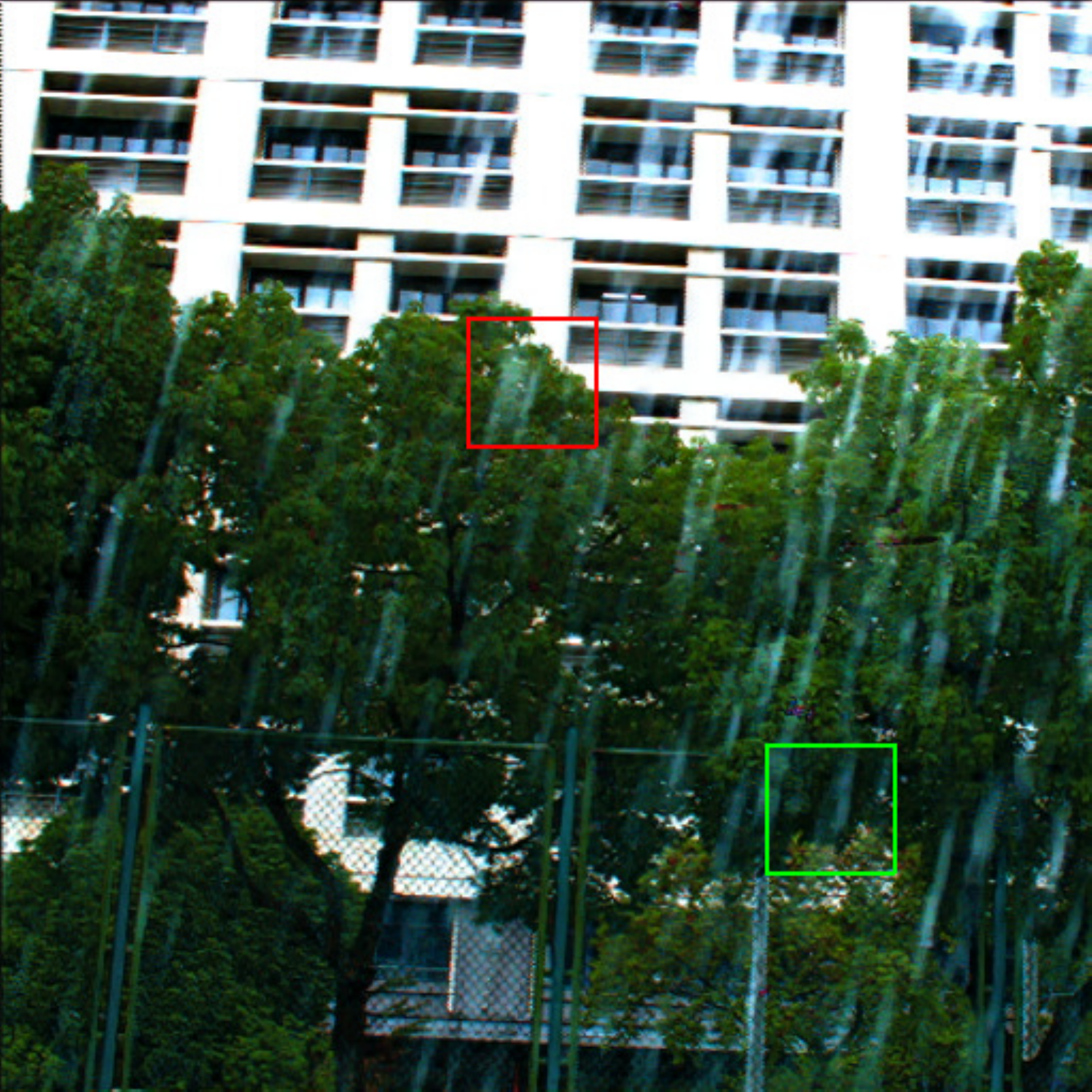}} &
            \multicolumn{2}{c}{\includegraphics[width=\subwidth\linewidth]{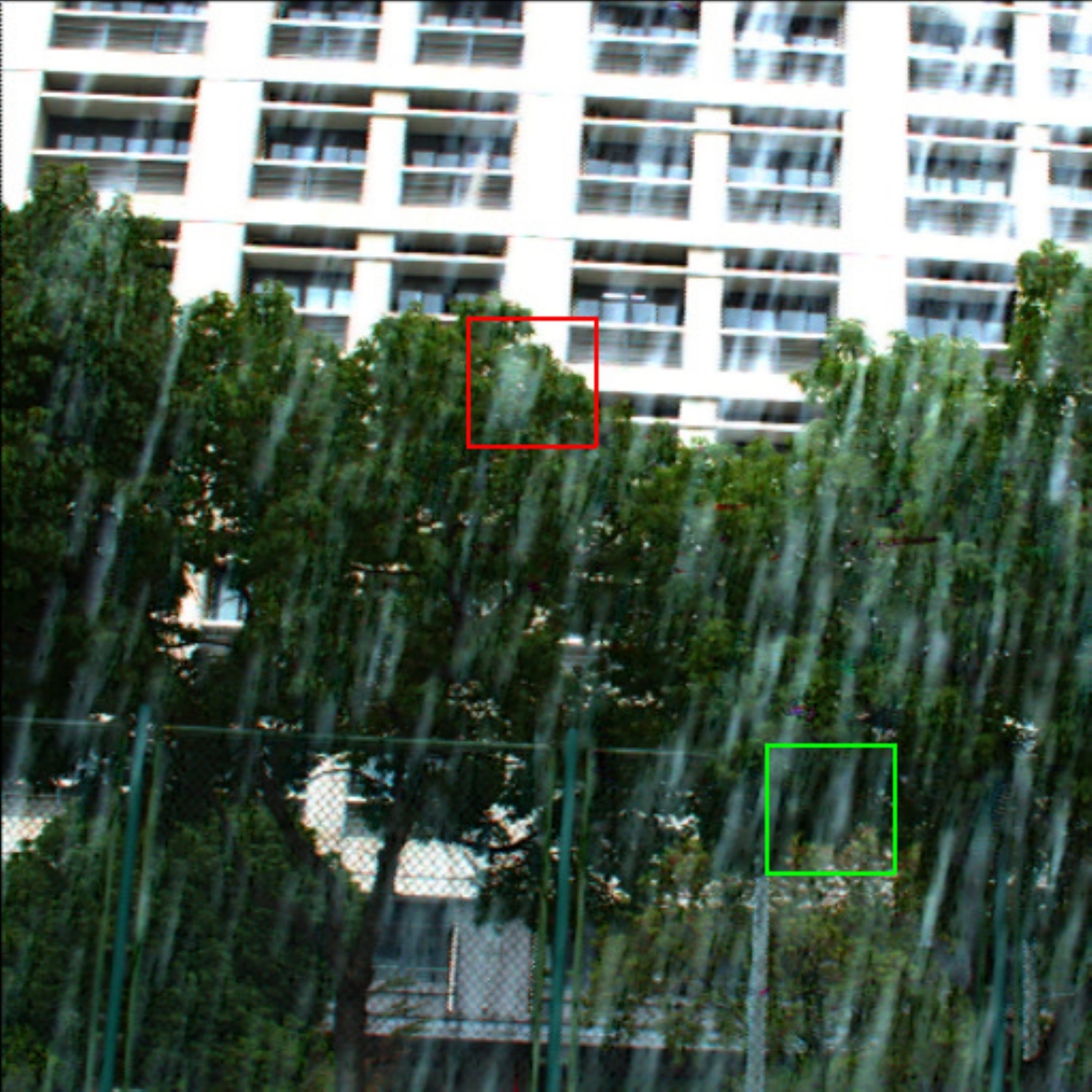}} &
            \multicolumn{2}{c}{\includegraphics[width=\subwidth\linewidth]{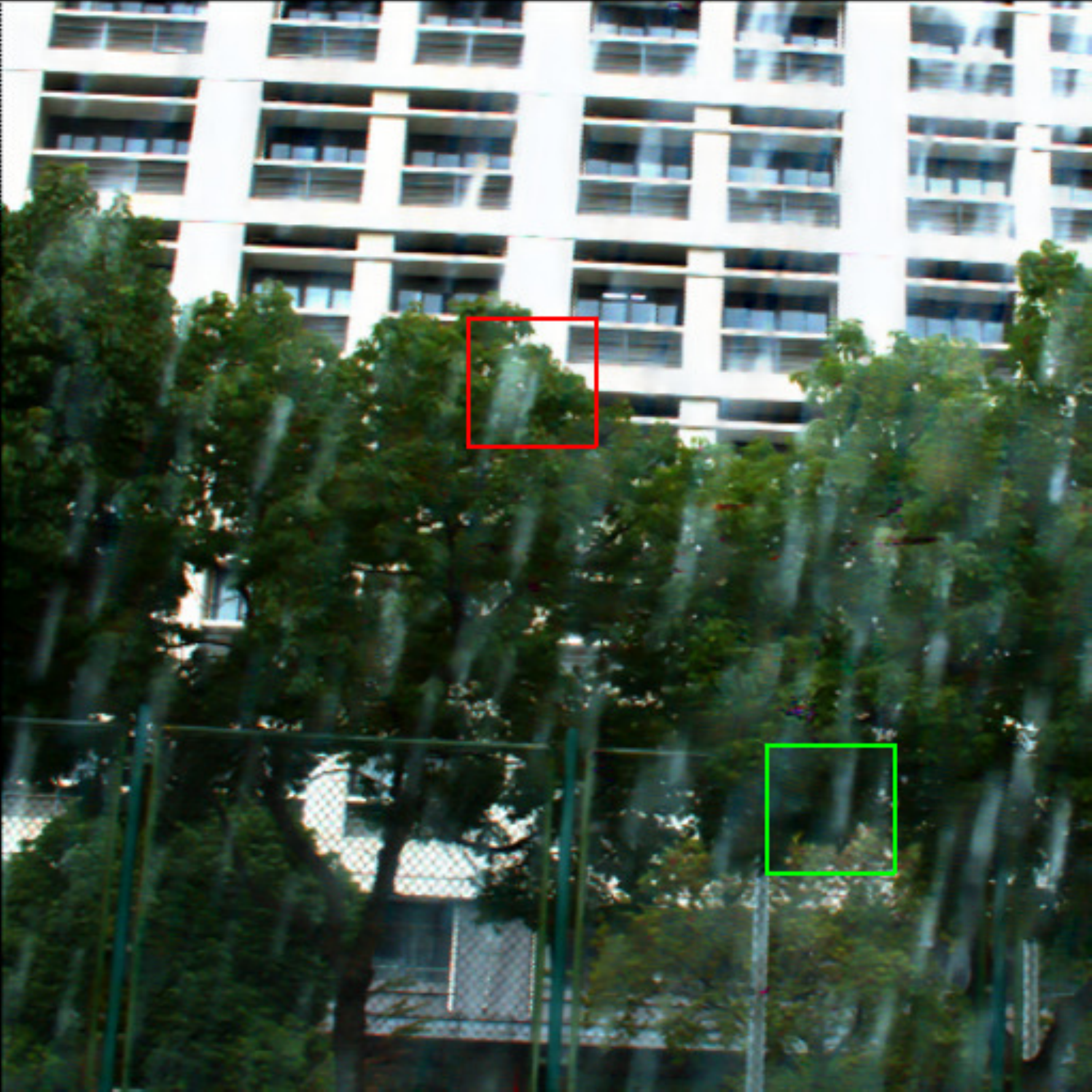}} &
            \multicolumn{2}{c}{\includegraphics[width=\subwidth\linewidth]{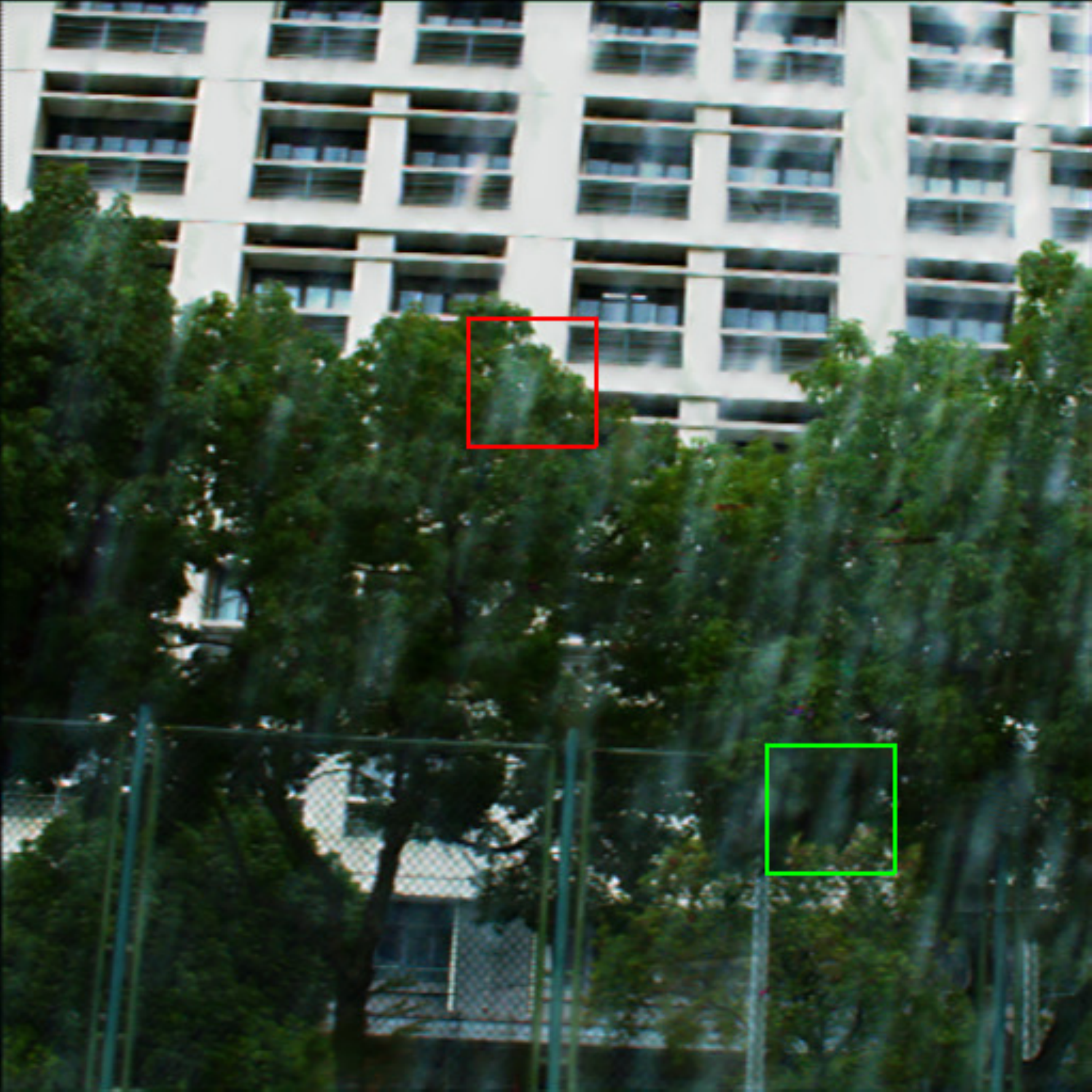}} &

            \multicolumn{2}{c}{\includegraphics[width=\subwidth\linewidth]{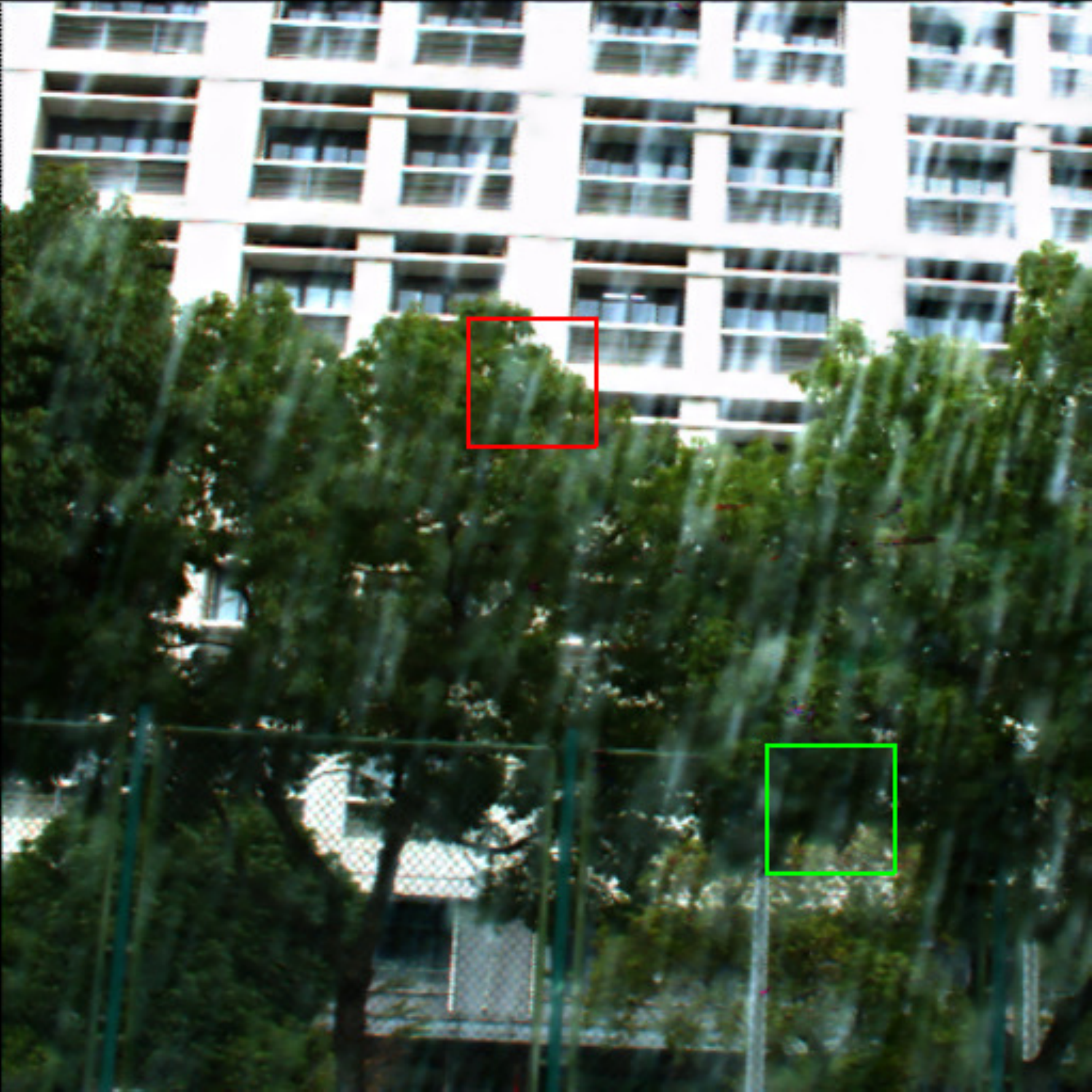}} &
            \multicolumn{2}{c}{\includegraphics[width=\subwidth\linewidth]{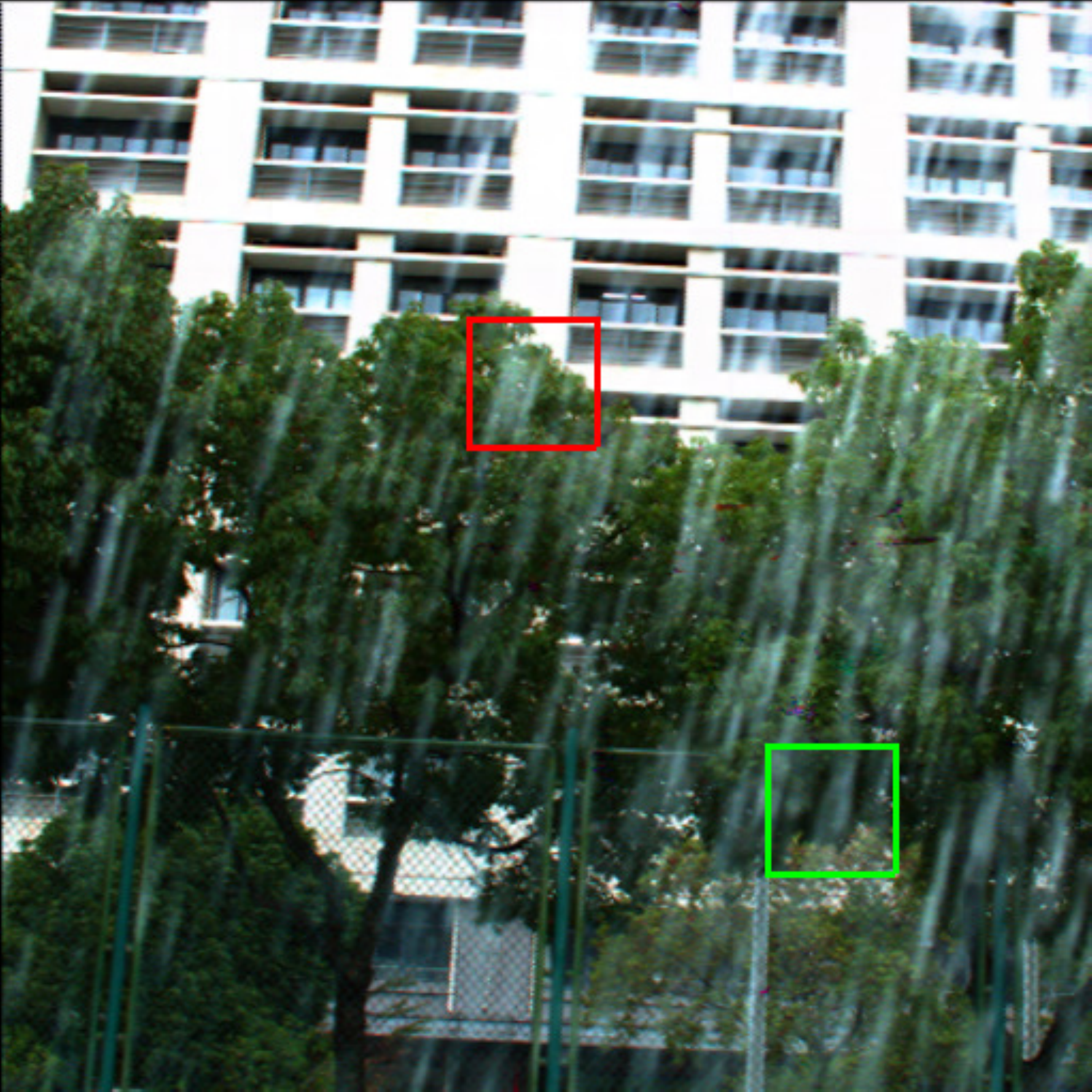}} &
            \multicolumn{2}{c}{\includegraphics[width=\subwidth\linewidth]{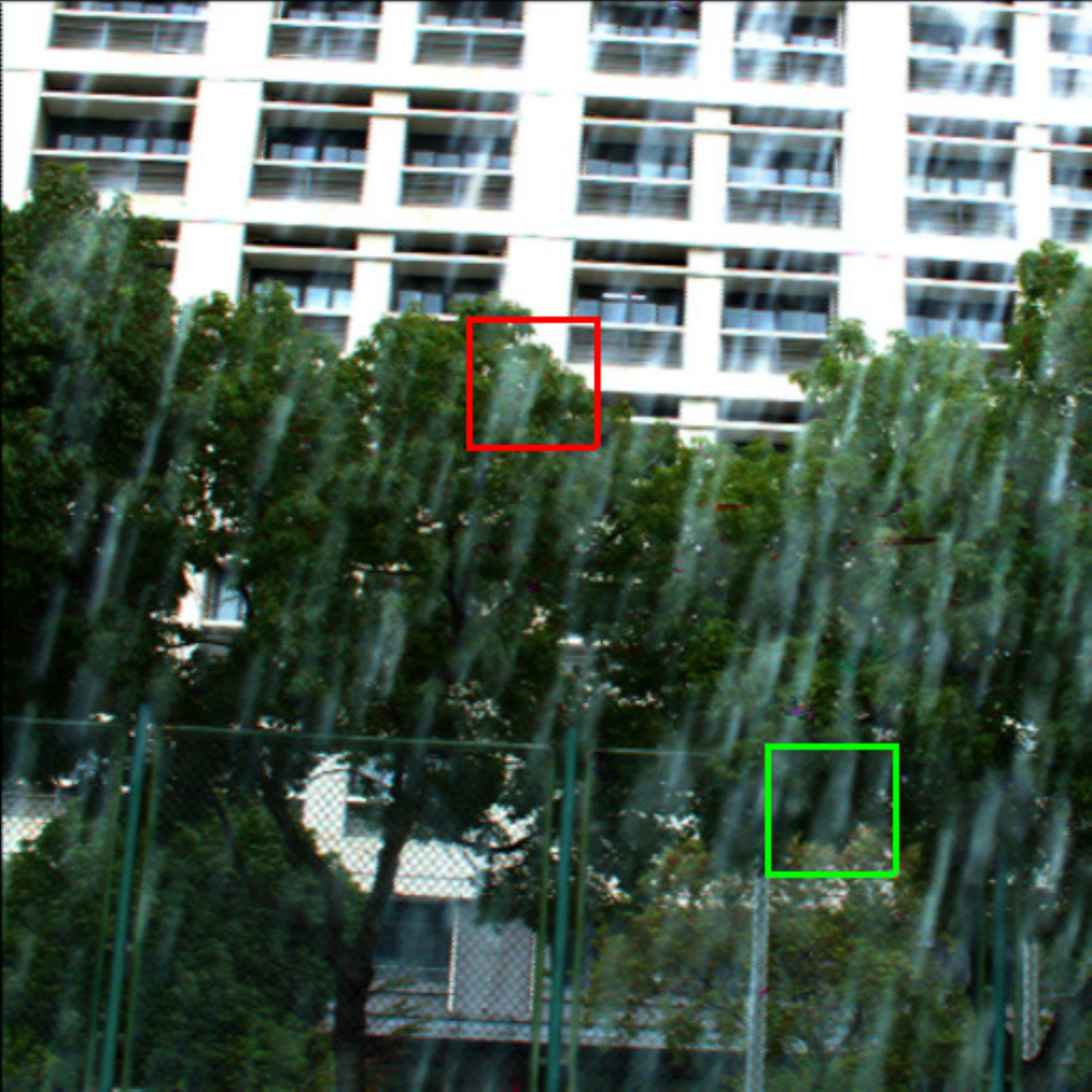}} &
            \multicolumn{2}{c}{\includegraphics[width=\subwidth\linewidth]{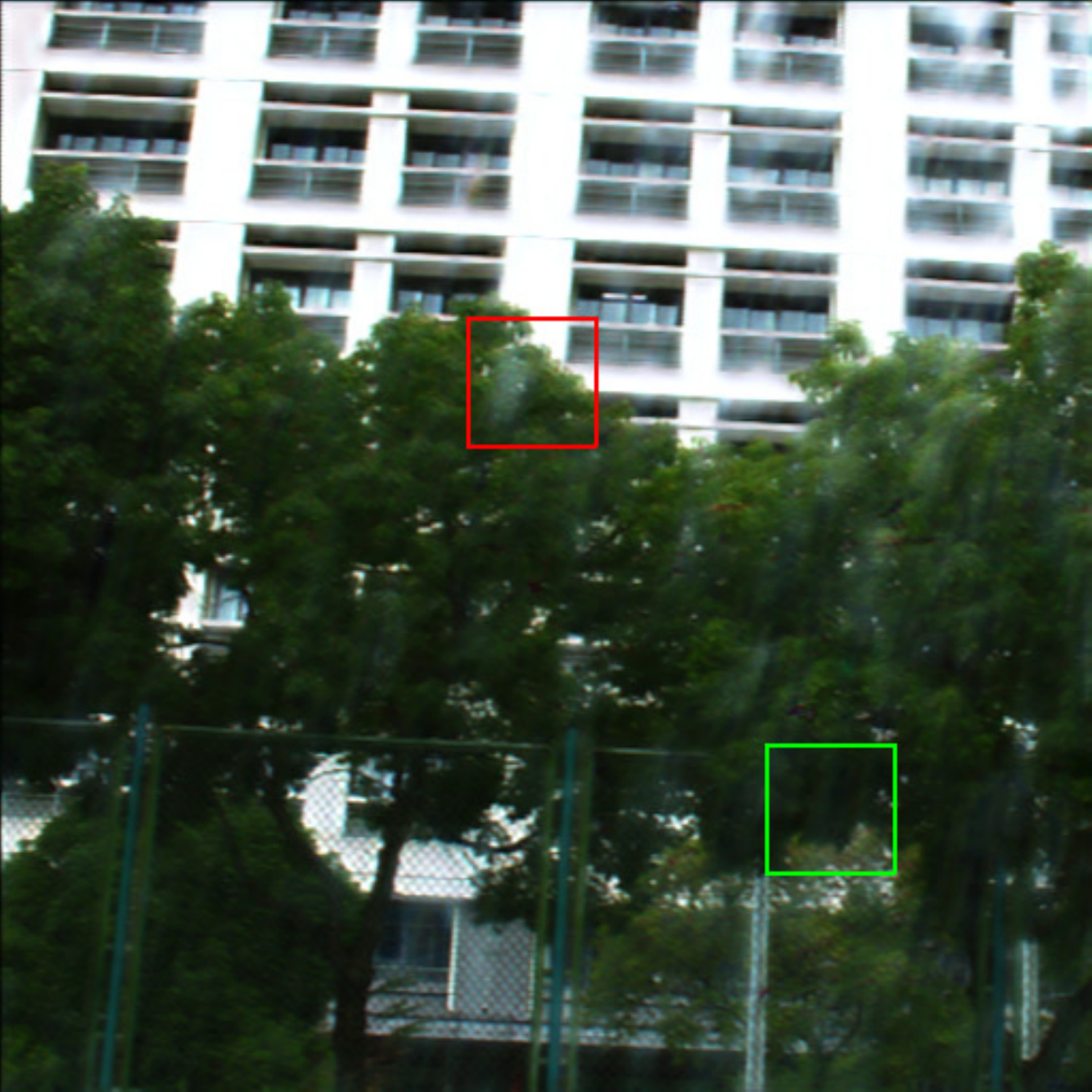}} &
            \multicolumn{2}{c}{\includegraphics[width=\subwidth\linewidth]{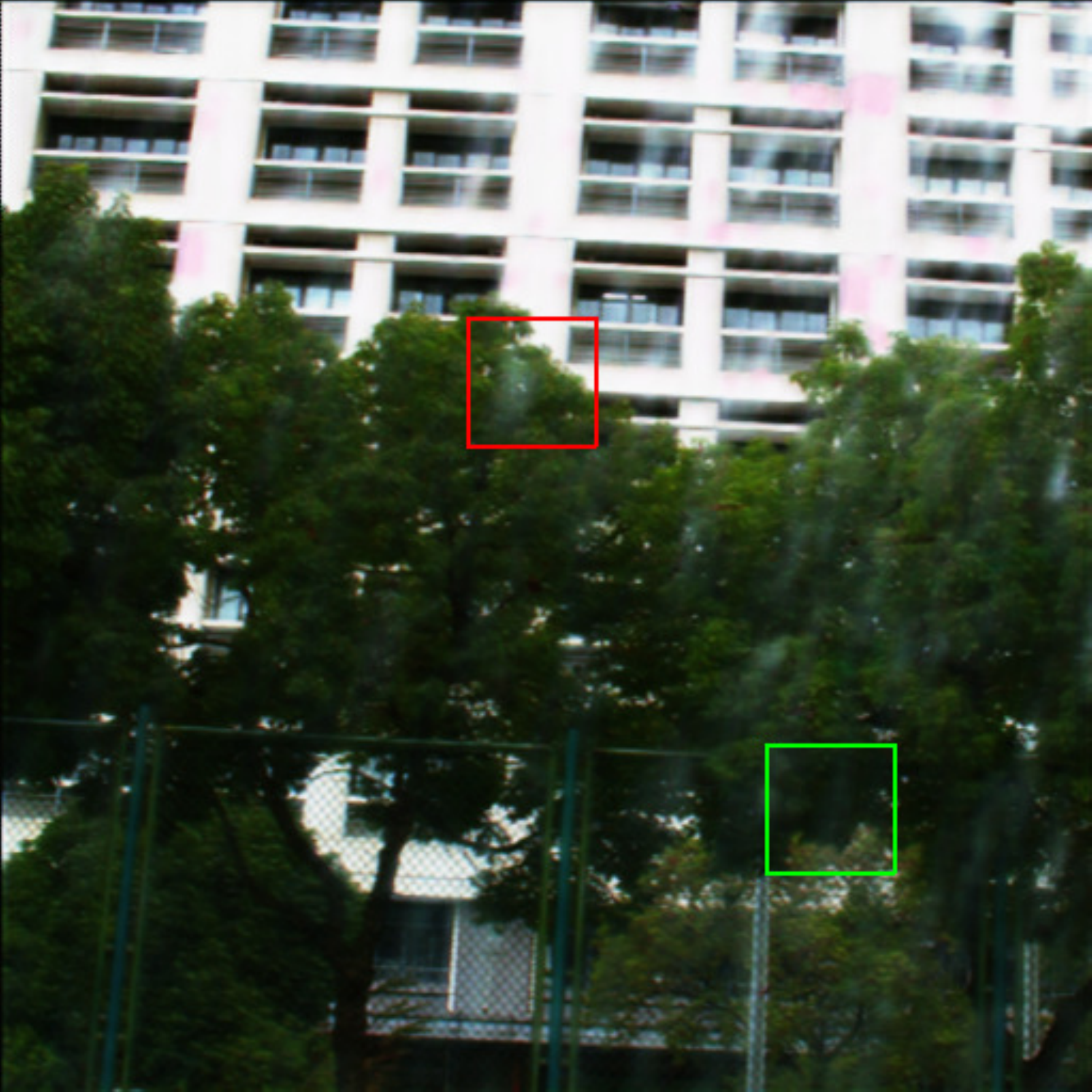}} \\

            \includegraphics[width=\ssubwidth\linewidth]{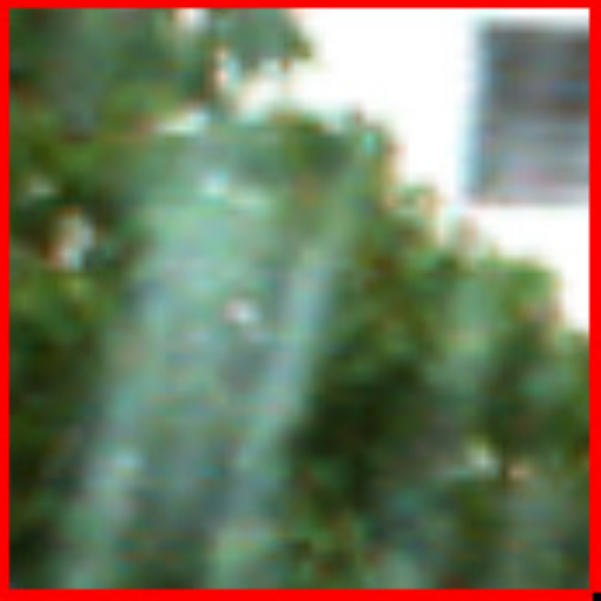} &
            \includegraphics[width=\ssubwidth\linewidth]{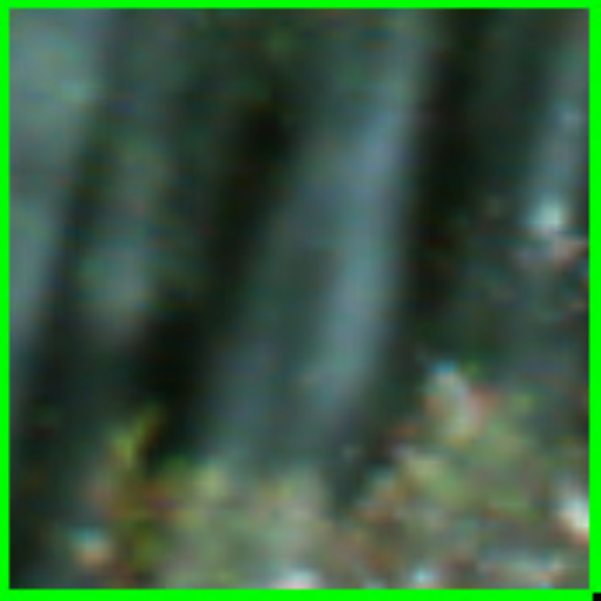} &
            \includegraphics[width=\ssubwidth\linewidth]{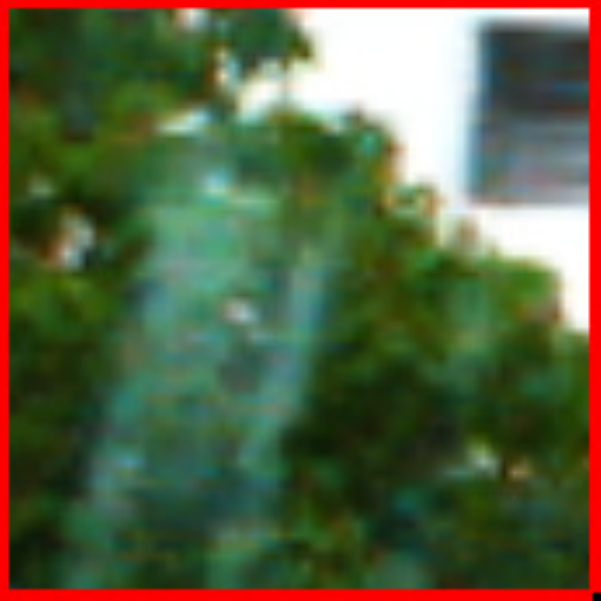} &
            \includegraphics[width=\ssubwidth\linewidth]{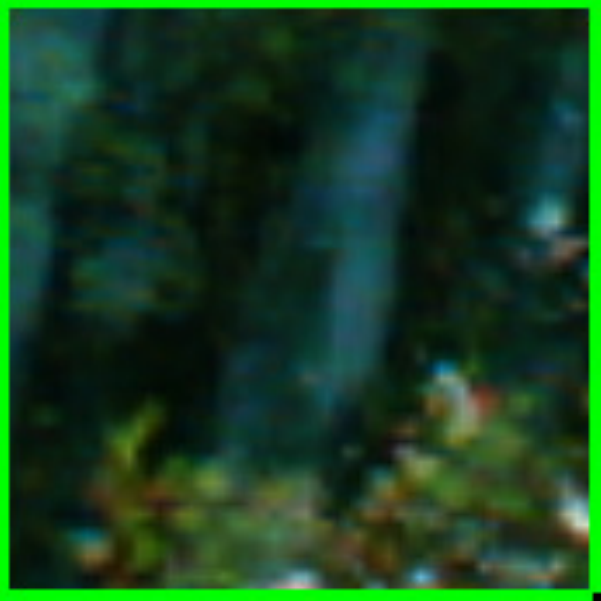} &
            \includegraphics[width=\ssubwidth\linewidth]{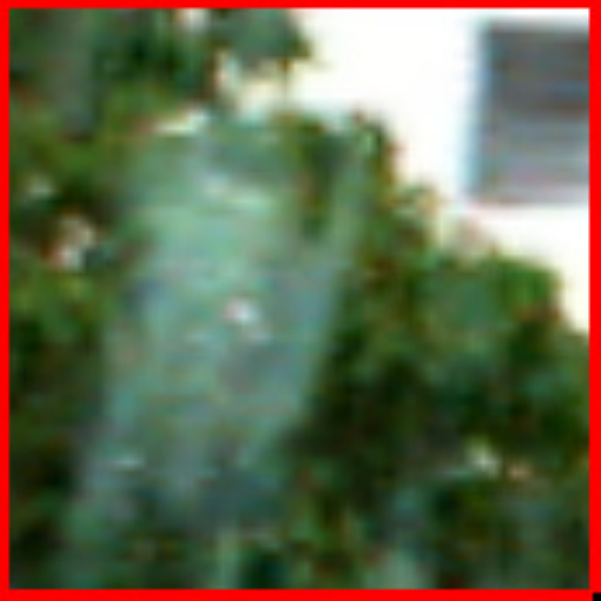} &
            \includegraphics[width=\ssubwidth\linewidth]{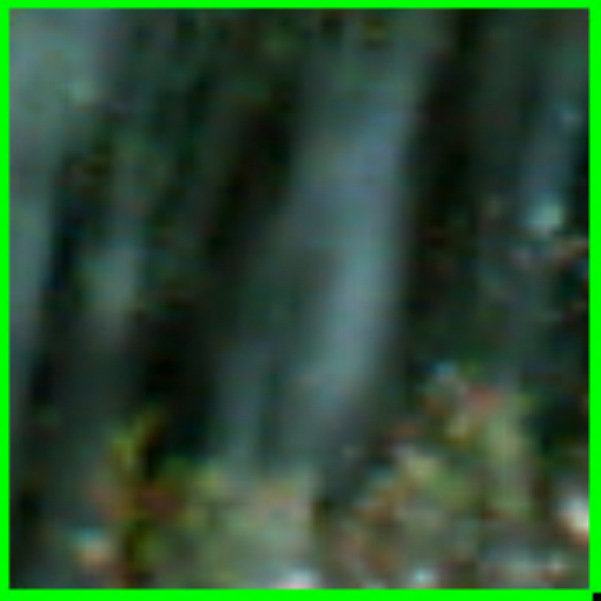} &
            \includegraphics[width=\ssubwidth\linewidth]{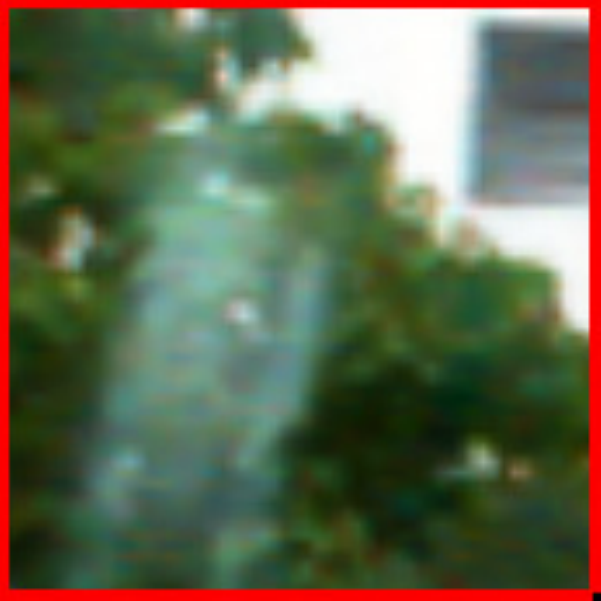} &
            \includegraphics[width=\ssubwidth\linewidth]{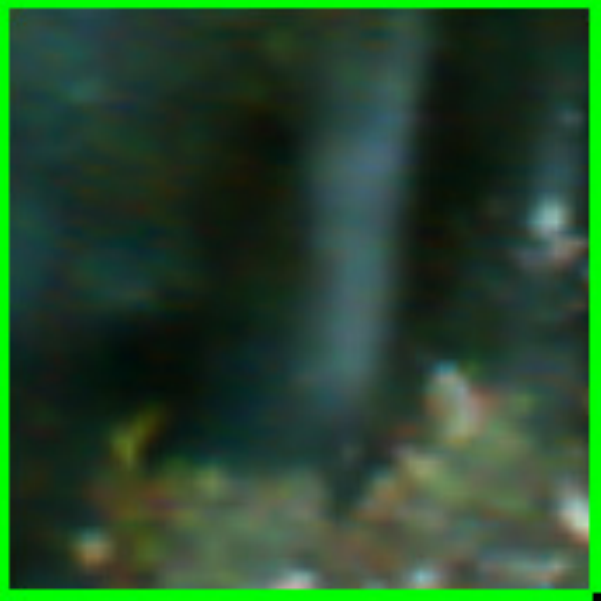} &
            \includegraphics[width=\ssubwidth\linewidth]{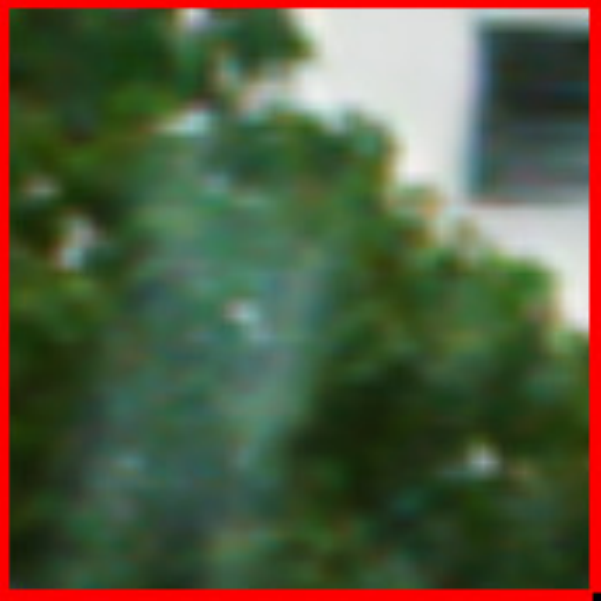} &
            \includegraphics[width=\ssubwidth\linewidth]{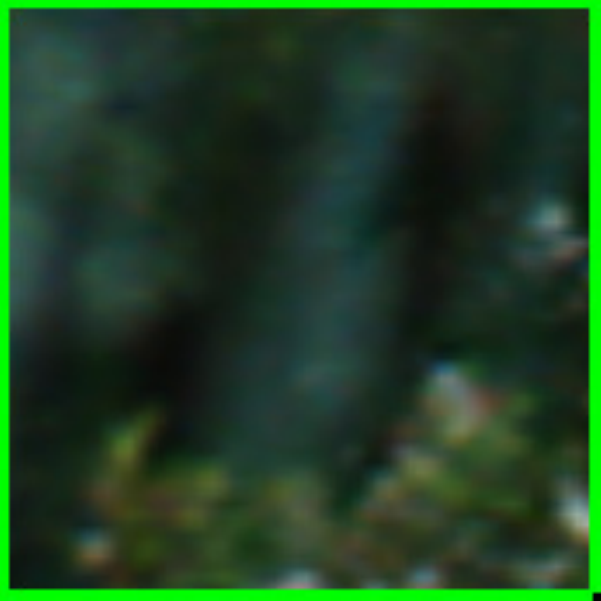} &

            \includegraphics[width=\ssubwidth\linewidth]{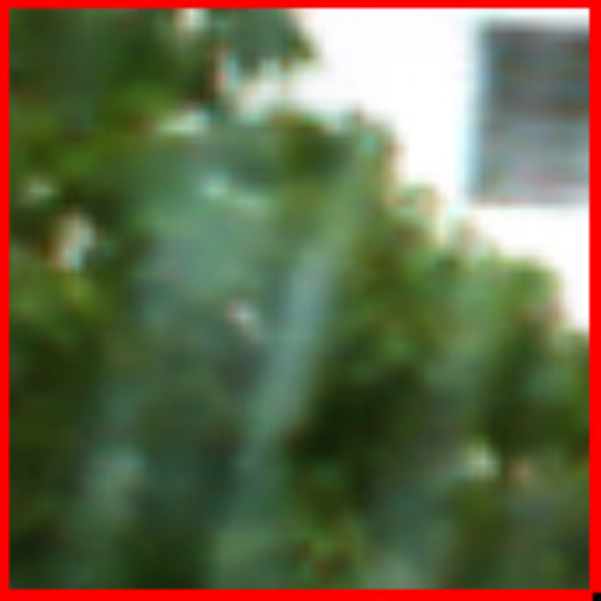} &
            \includegraphics[width=\ssubwidth\linewidth]{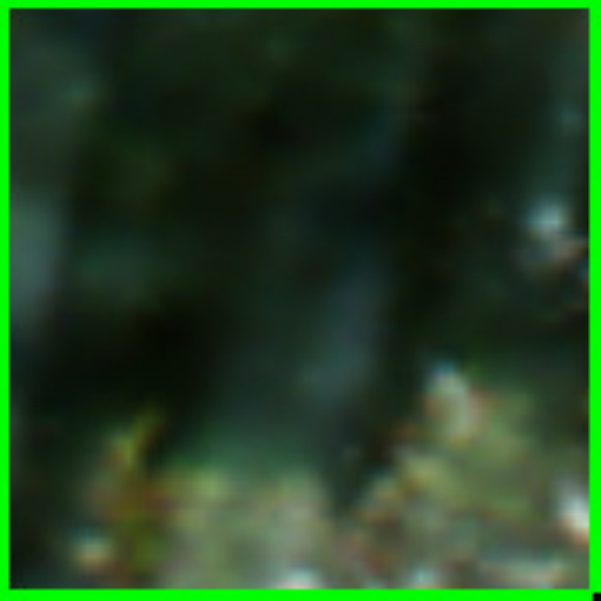} &
            \includegraphics[width=\ssubwidth\linewidth]{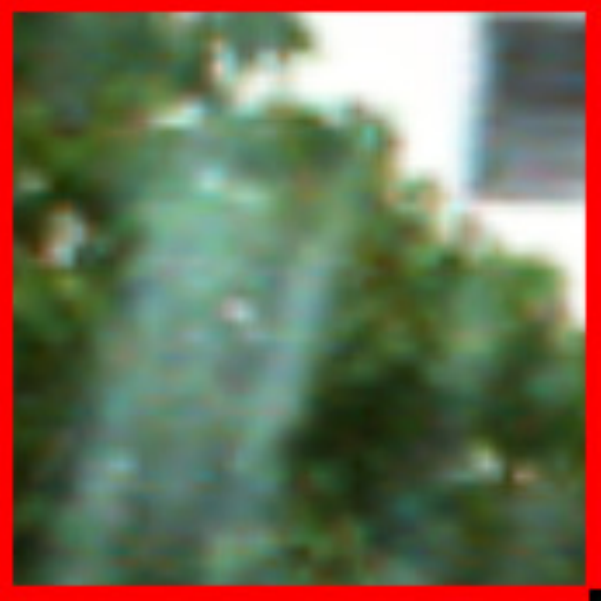} &
            \includegraphics[width=\ssubwidth\linewidth]{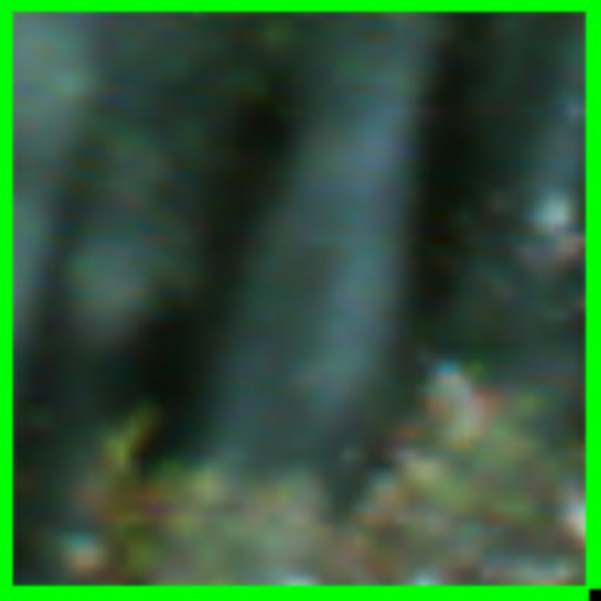} &
            \includegraphics[width=\ssubwidth\linewidth]{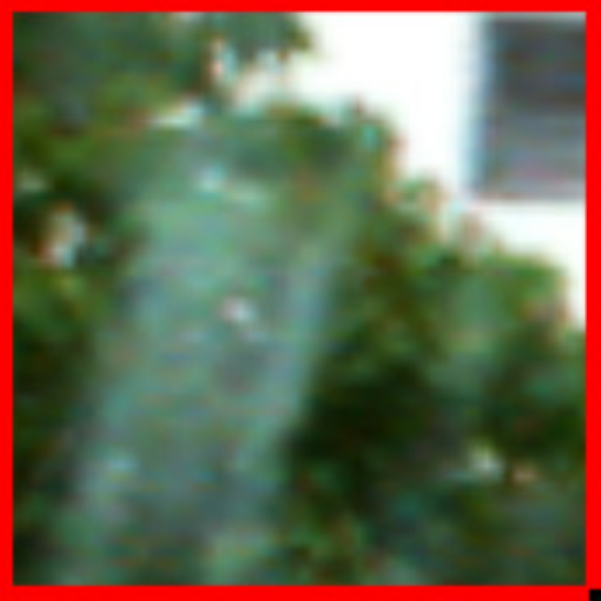} &
            \includegraphics[width=\ssubwidth\linewidth]{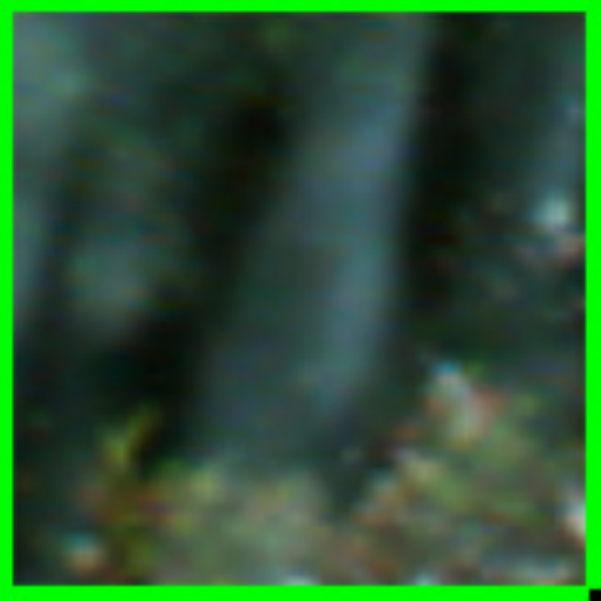} &
            \includegraphics[width=\ssubwidth\linewidth]{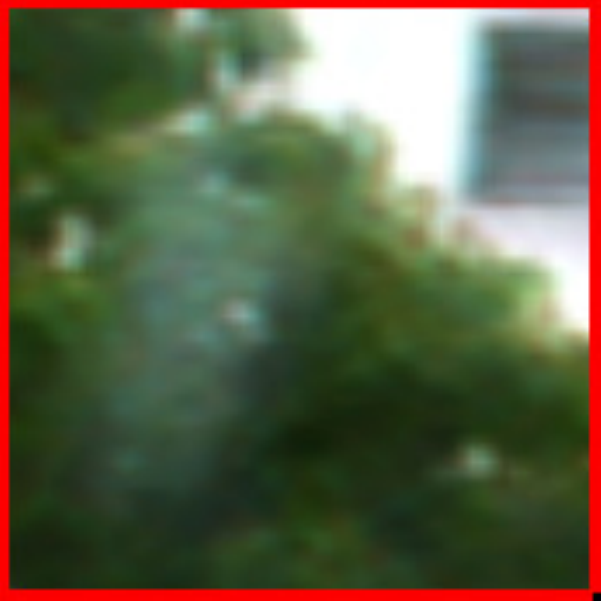} &
            \includegraphics[width=\ssubwidth\linewidth]{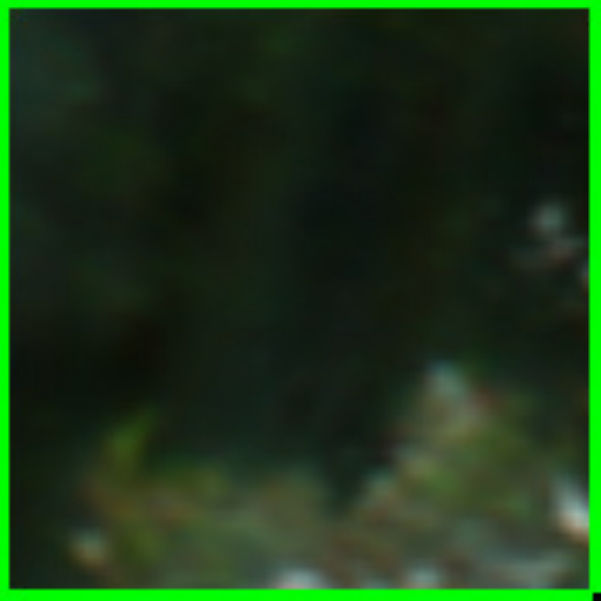} &
            \includegraphics[width=\ssubwidth\linewidth]{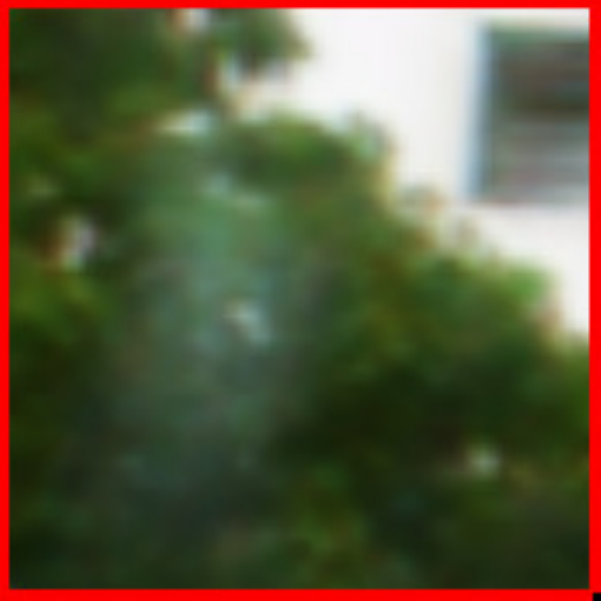} &
            \includegraphics[width=\ssubwidth\linewidth]{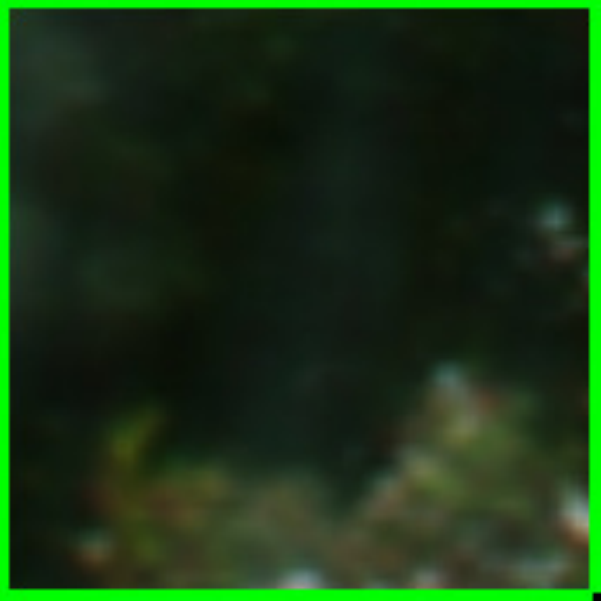} \\

            \multicolumn{2}{c}{\scriptsize{(a) Input}} &
            \multicolumn{2}{c}{\scriptsize{(b) Li~\cite{Li19a}}} &
            \multicolumn{2}{c}{\scriptsize{(c) Ren~\cite{ren20}}} &
            \multicolumn{2}{c}{\scriptsize{(d) Wei~\cite{wei19}}} &
            \multicolumn{2}{c}{\scriptsize{(e) Yasarla~\cite{yasarla20}}} &
            \multicolumn{2}{c}{\scriptsize{(h) Ding~\cite{ding21}}} &
            \multicolumn{2}{c}{\scriptsize{(f) \revised{Xiao~\cite{xiao22}}}}&
            \multicolumn{2}{c}{\scriptsize{(g) \revised{Zamir~\cite{zamir22}}}}&
            \multicolumn{2}{c}{\scriptsize{(i) Ours-GD}} &
            \multicolumn{2}{c}{\scriptsize{(j) Ours}} \\

		\end{tabular}
	\end{center}
	\vspace{-0.016\textwidth}
	\caption{Comparison of the state-of-the-art deraining methods and our method on the challenging real-world LFIs coming from RLMB. The last two columns show the de-rained sub-views obtained by our network with the global discriminator and global-local discriminator, respectively.}
	\label{fig:real result}
\end{figure*}

\renewcommand{\subwidth}{0.138}
\begin{figure*}[t]
	\renewcommand{\tabcolsep}{1.0pt}
	\renewcommand\arraystretch{0.8}
	\begin{center}
		\begin{tabular}{ccccccc}
			\includegraphics[width=\subwidth\linewidth]{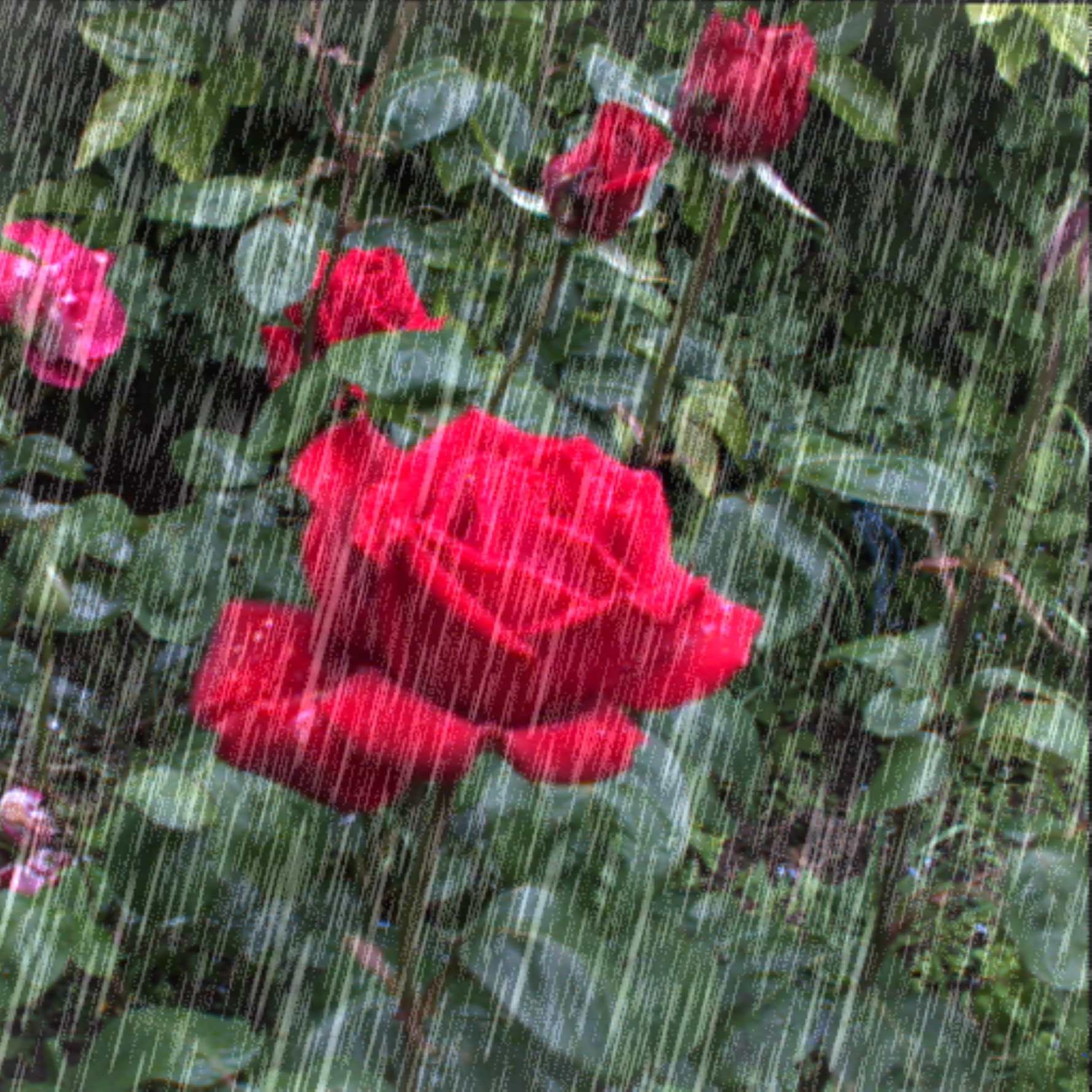} &
            \includegraphics[width=\subwidth\linewidth]{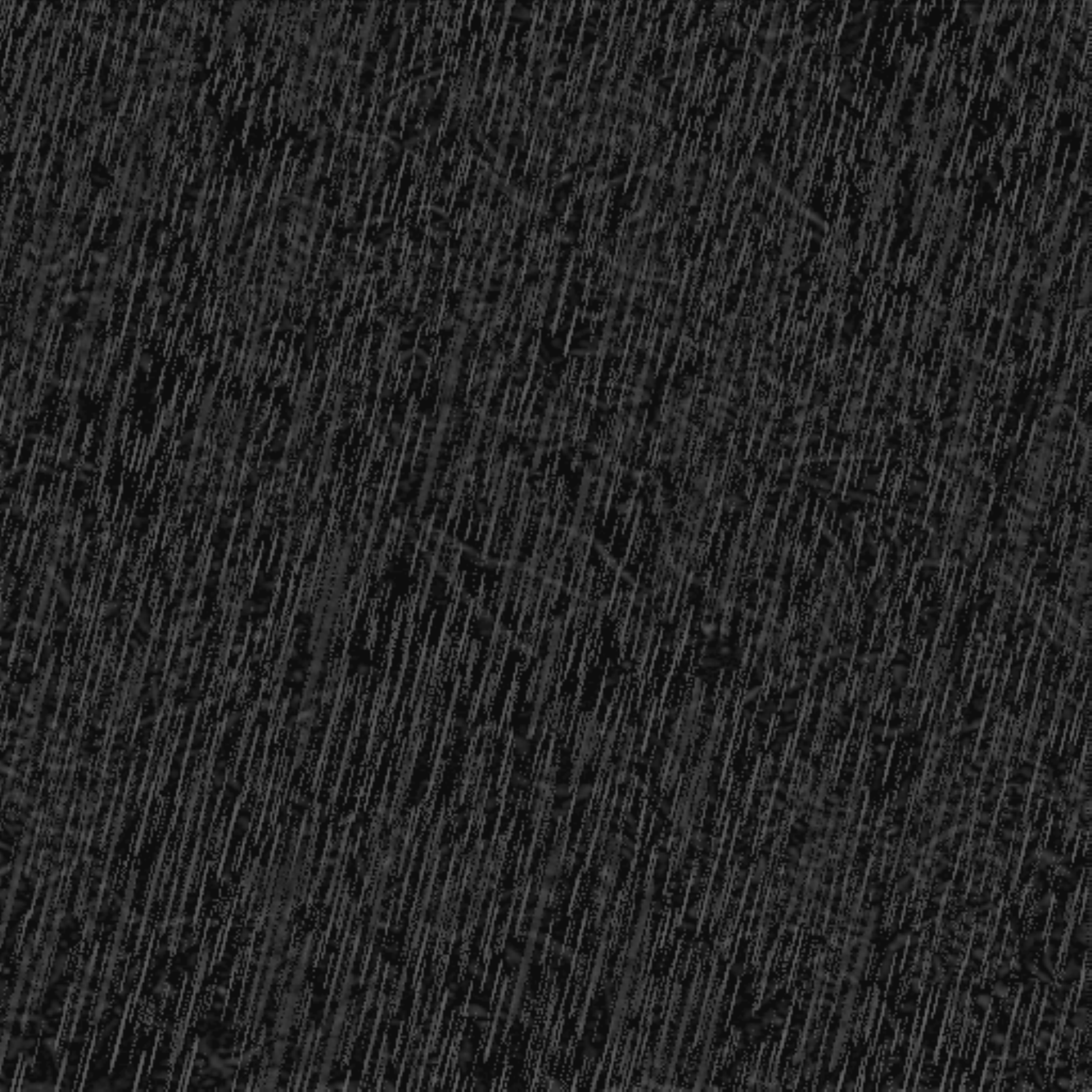} &
           	\includegraphics[width=\subwidth\linewidth]{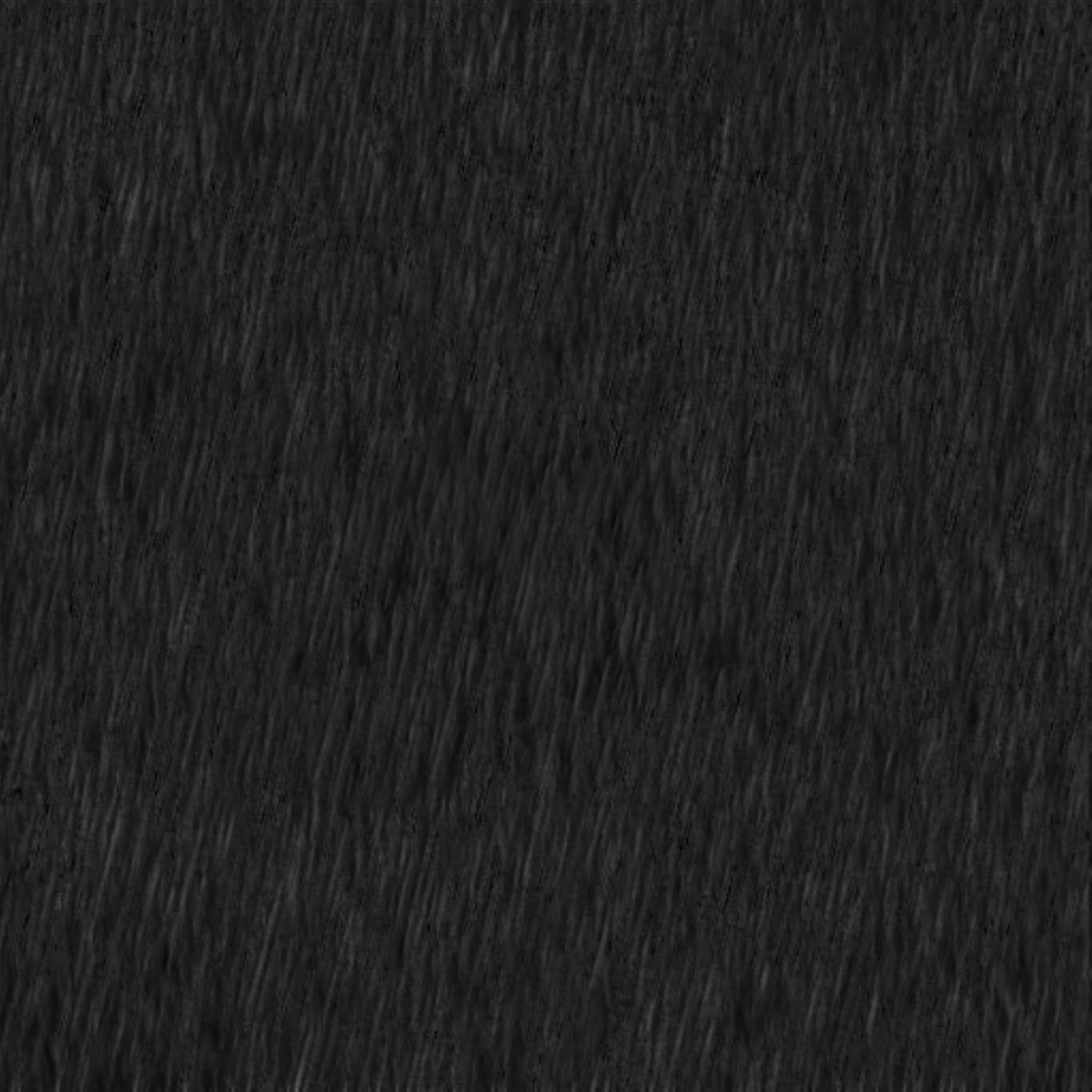} &
			\includegraphics[width=\subwidth\linewidth]{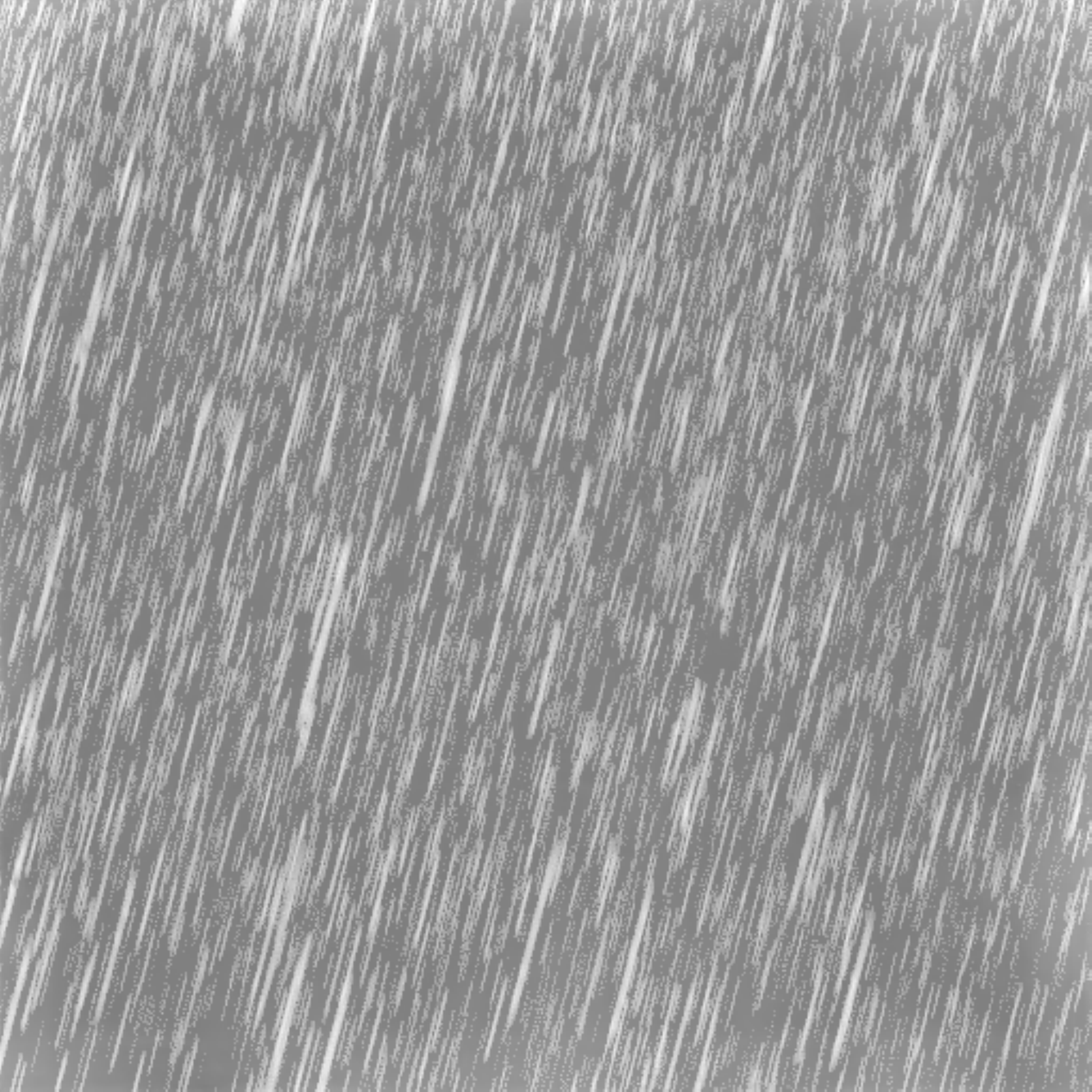} &
			\includegraphics[width=\subwidth\linewidth]{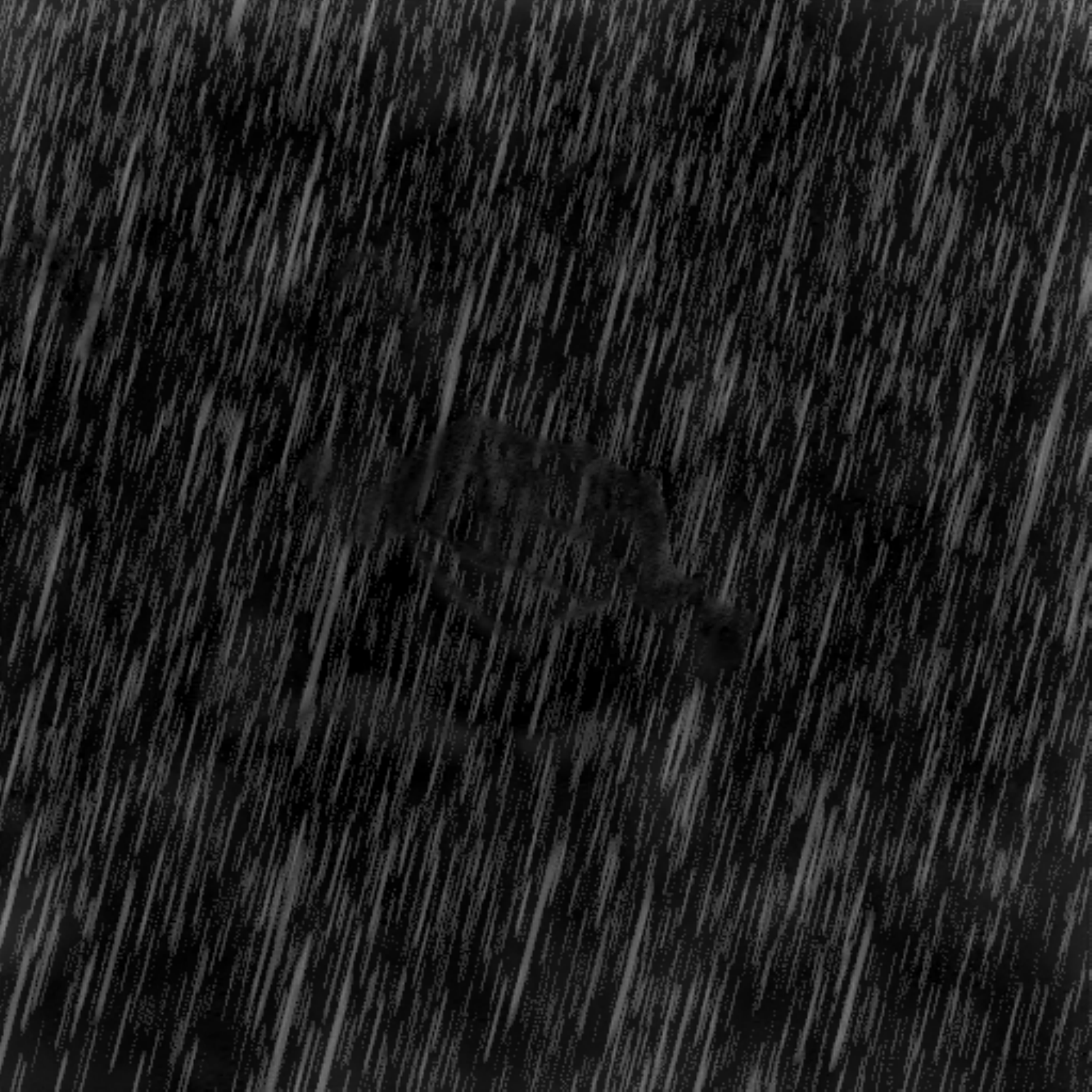} &
			\includegraphics[width=\subwidth\linewidth]{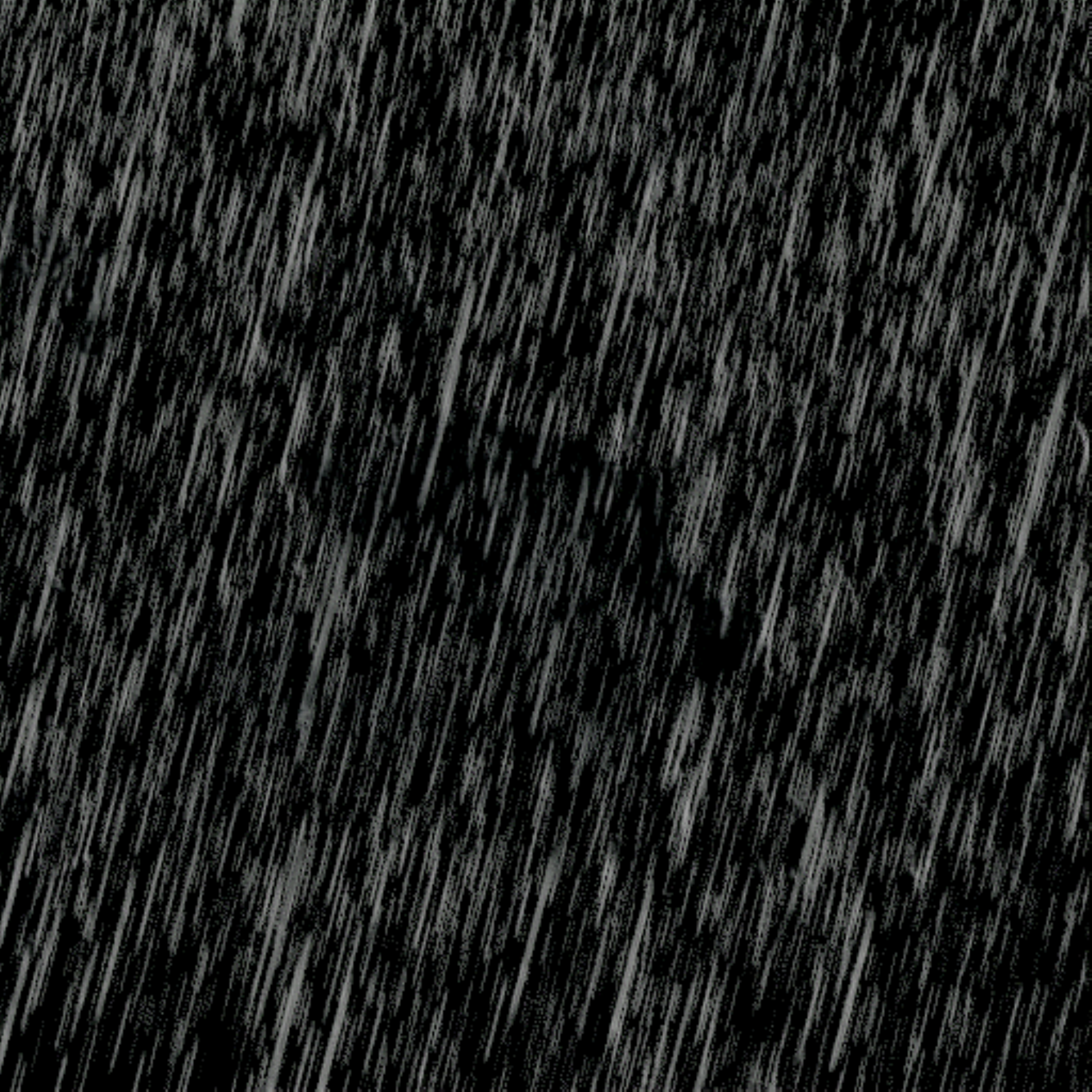} &
			\includegraphics[width=\subwidth\linewidth]{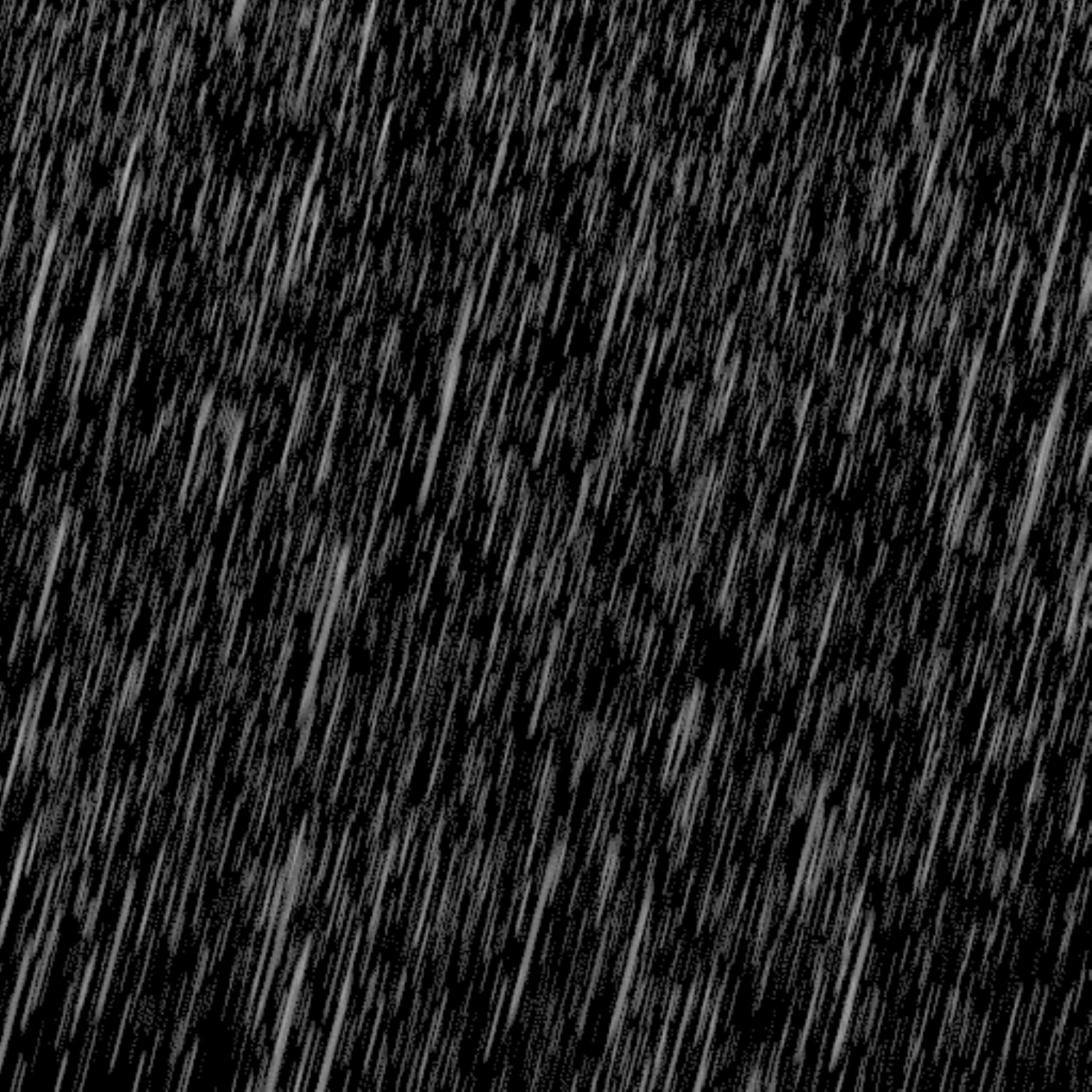} \\ %rain/116

            \scriptsize{Input}&
            \scriptsize{~\cite{li14}, $19.78/0.689$}&
            \scriptsize{~\cite{Li19a}, $15.93/0.130$}&
            \scriptsize{~\cite{jiang2020}, $6.50/0.350$}&
            \scriptsize{~\cite{ding21}, $\emp{22.03}/\emp{0.789}$}&
            \scriptsize{Ours, $\textbf{23.68/0.847}$}&
            \scriptsize{GT}\\

		\end{tabular}
	\end{center}
	\vspace{-0.02\textwidth}
	\caption{Comparison of rain streaks detected on the central sub-view of a synthetic rainy LFI by different methods.}
    \label{fig:synthetic result_rain}
    \vspace{-2mm}
\end{figure*}

\renewcommand{\subwidth}{0.138}
\begin{figure*}[t]
	\renewcommand{\tabcolsep}{1.0pt}
	\renewcommand\arraystretch{0.8}
	\begin{center}
		\begin{tabular}{ccccccc}
			\includegraphics[width=\subwidth\linewidth]{fig2/test/input/1006} &
            \includegraphics[width=\subwidth\linewidth]{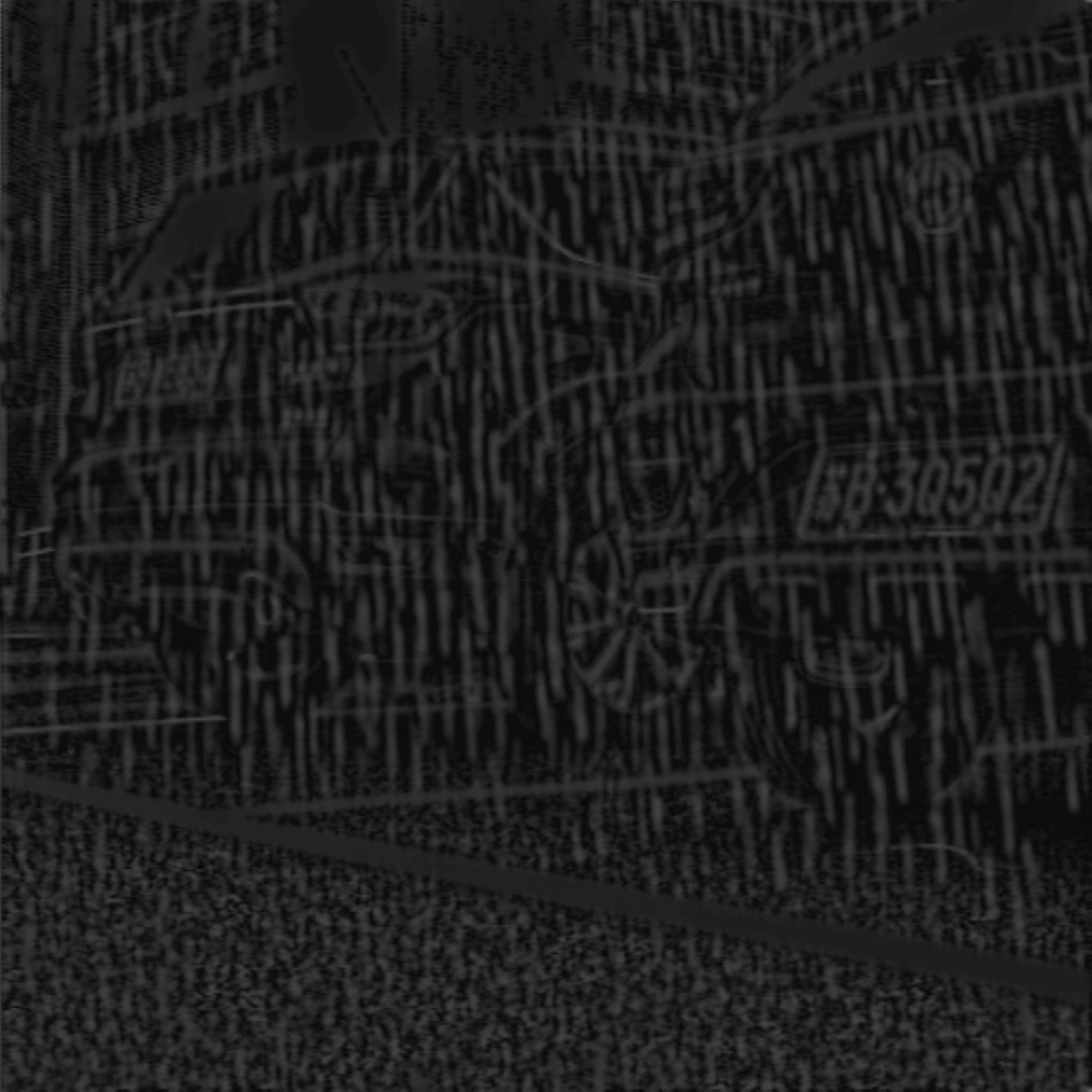} &
           	\includegraphics[width=\subwidth\linewidth]{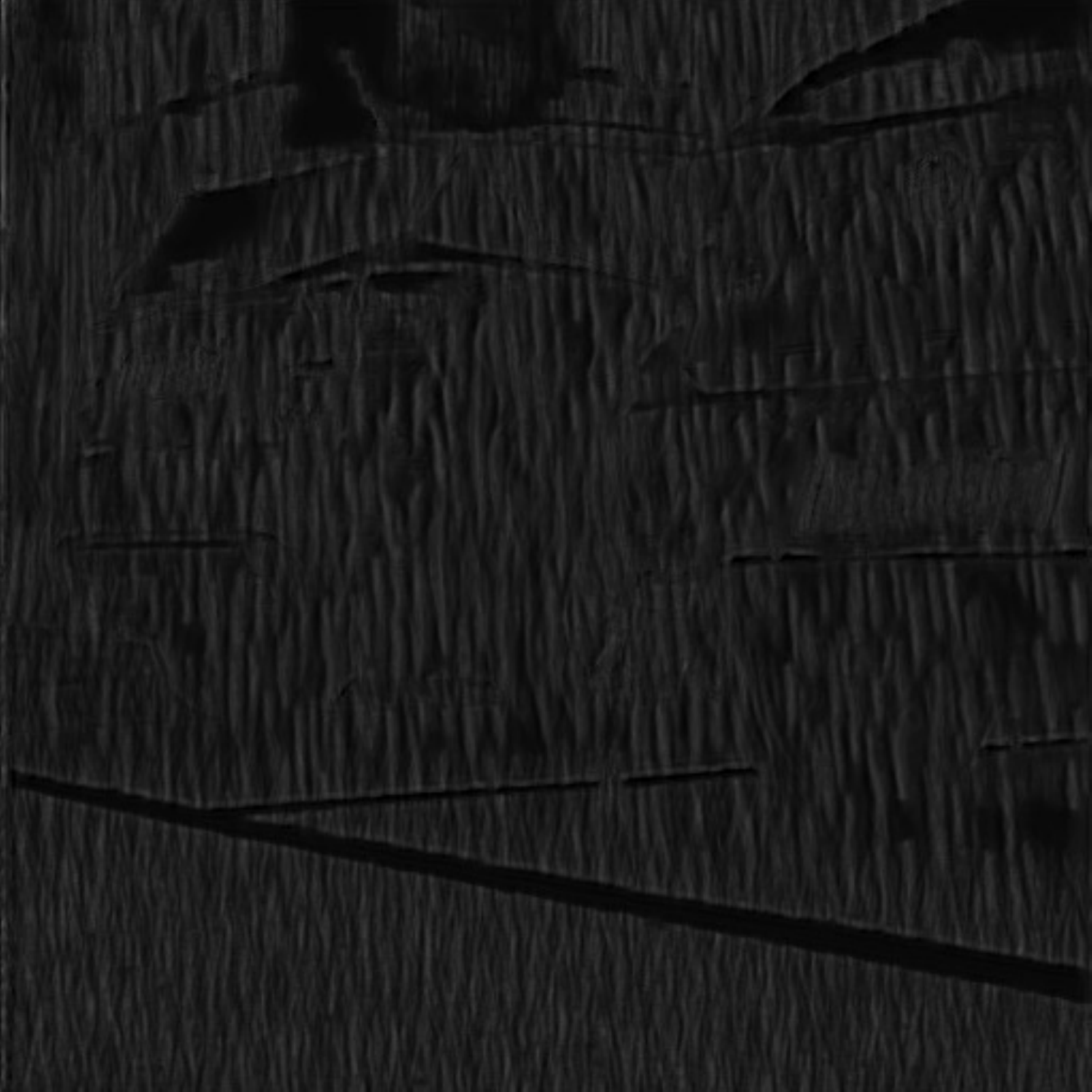} &
			\includegraphics[width=\subwidth\linewidth]{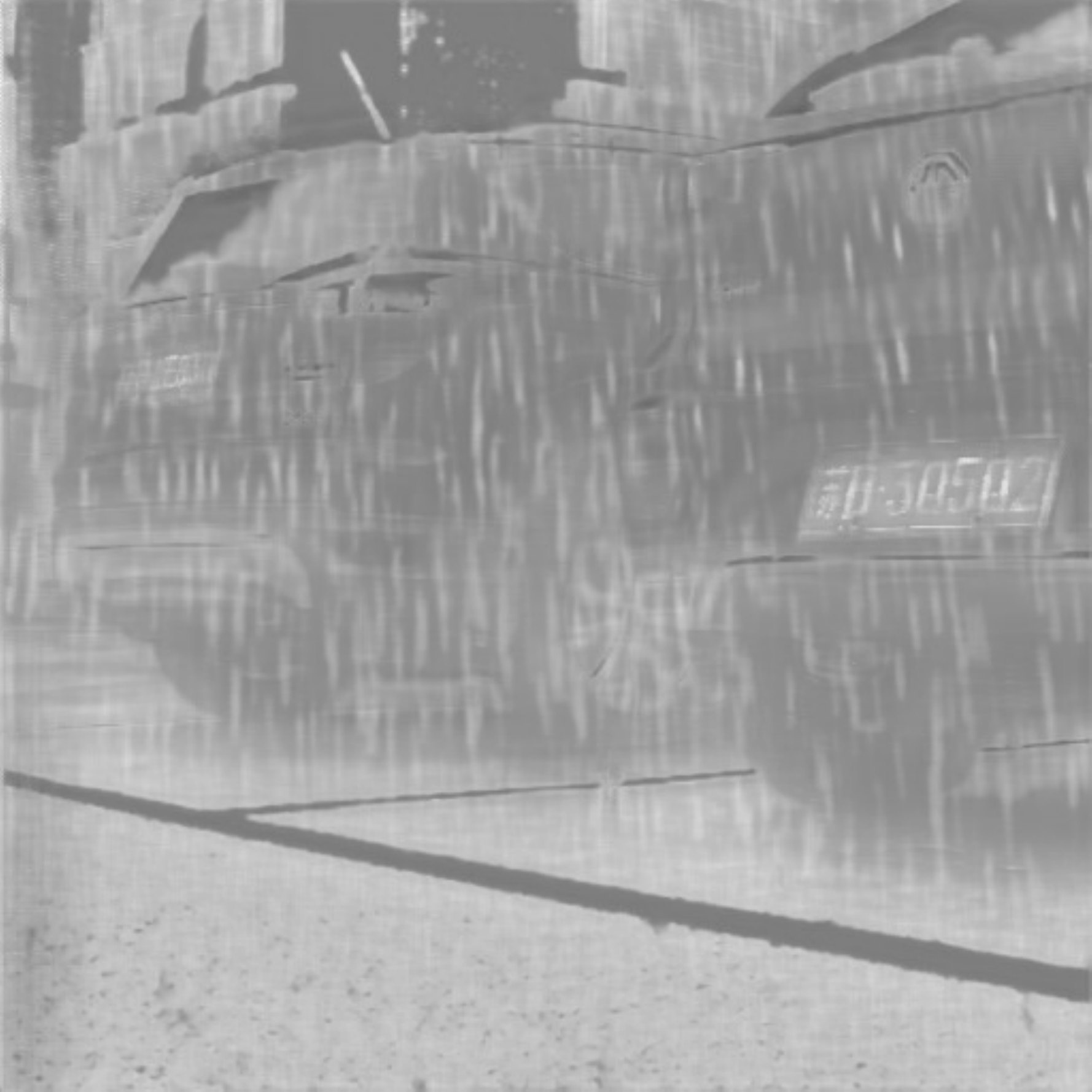} &
			\includegraphics[width=\subwidth\linewidth]{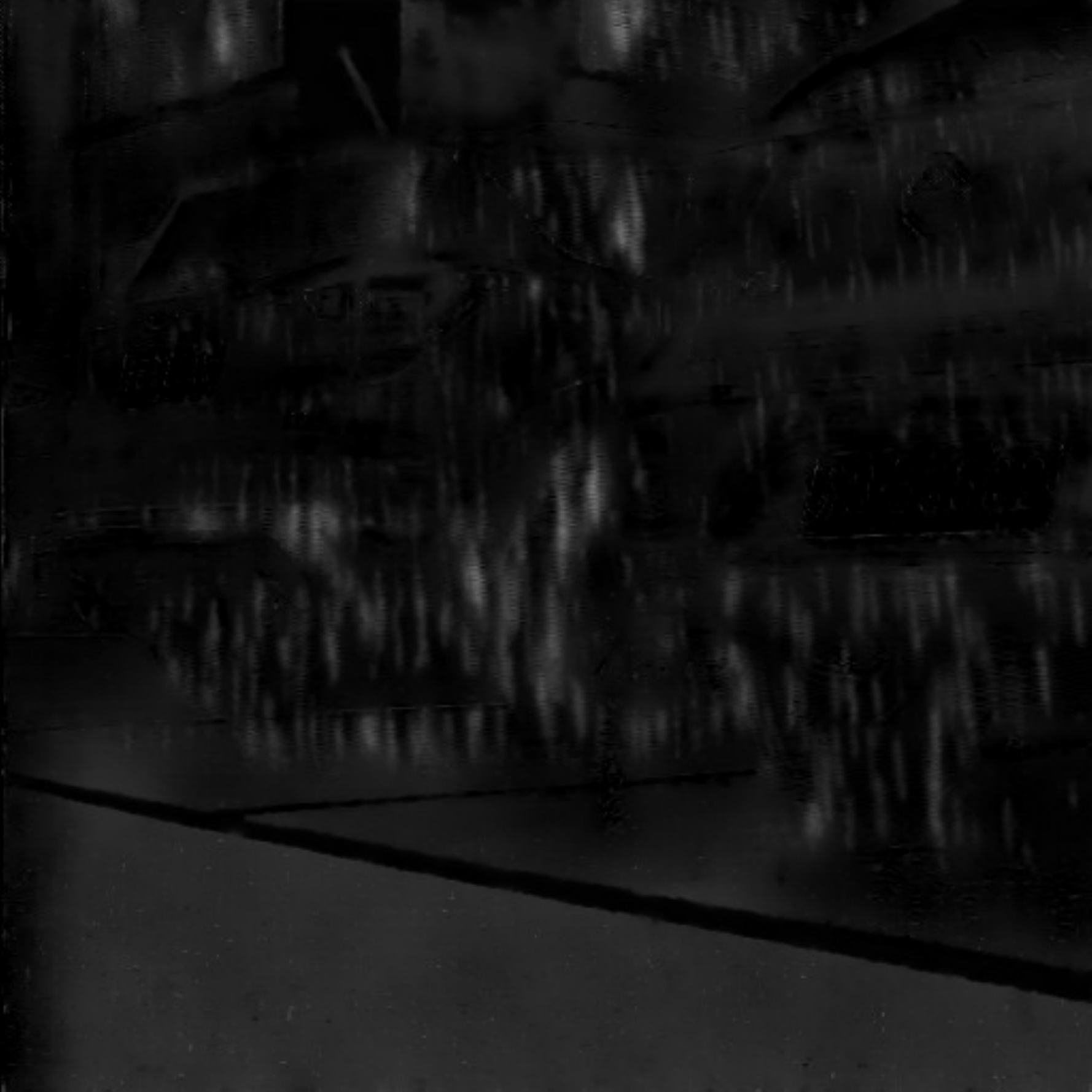} &
			\includegraphics[width=\subwidth\linewidth]{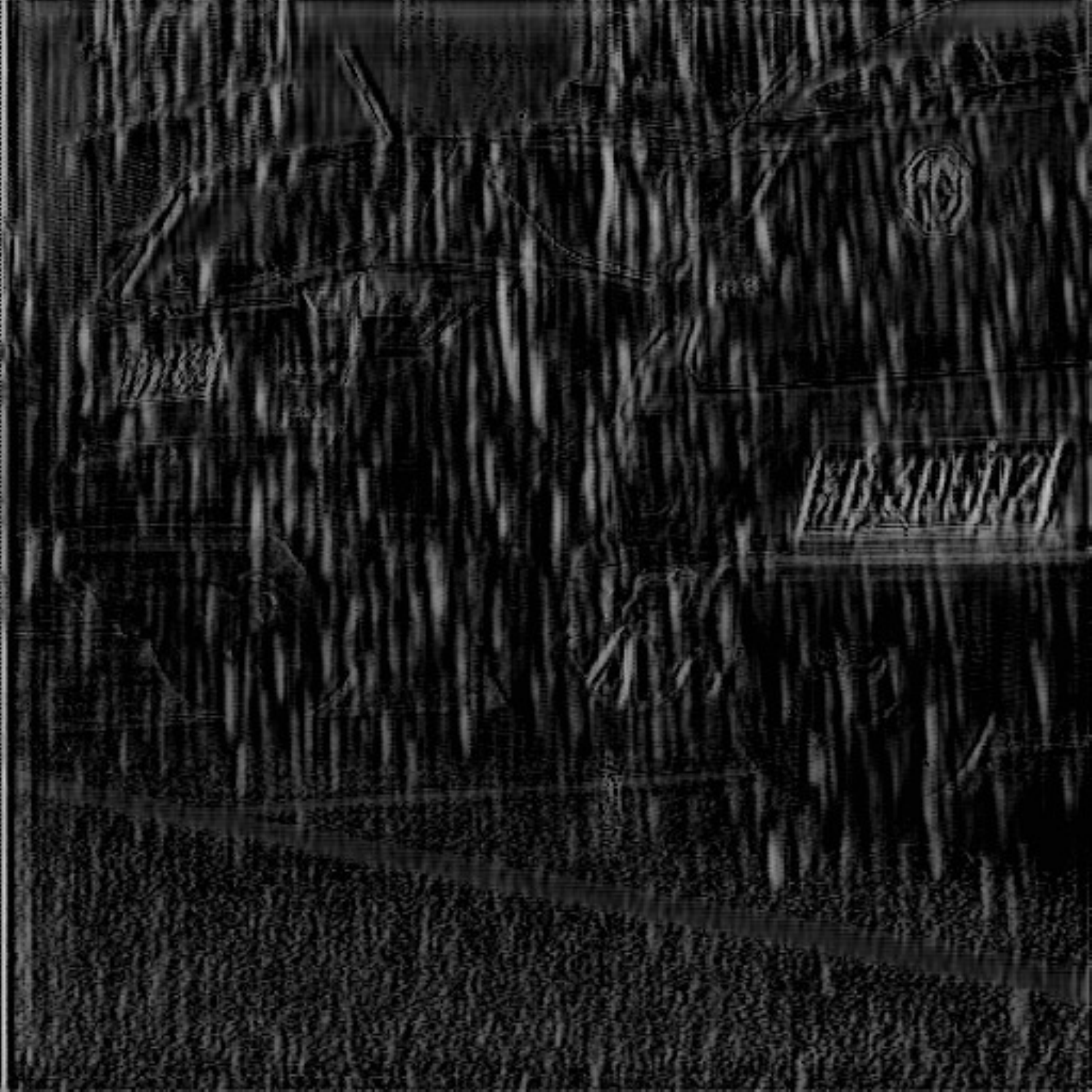} &
			\includegraphics[width=\subwidth\linewidth]{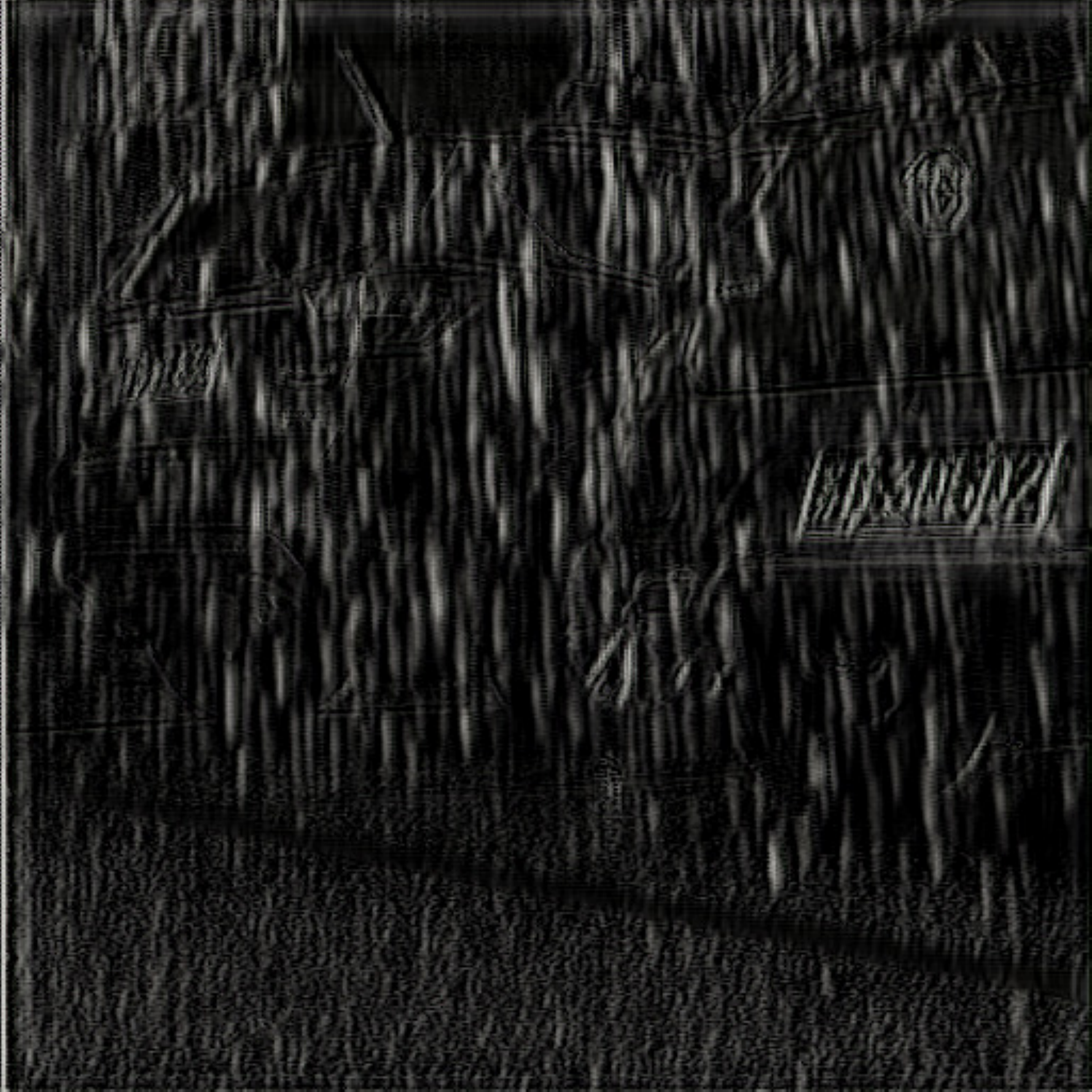} \\ %rain/116

			\includegraphics[width=\subwidth\linewidth]{fig2/test/input/1008} &
            \includegraphics[width=\subwidth\linewidth]{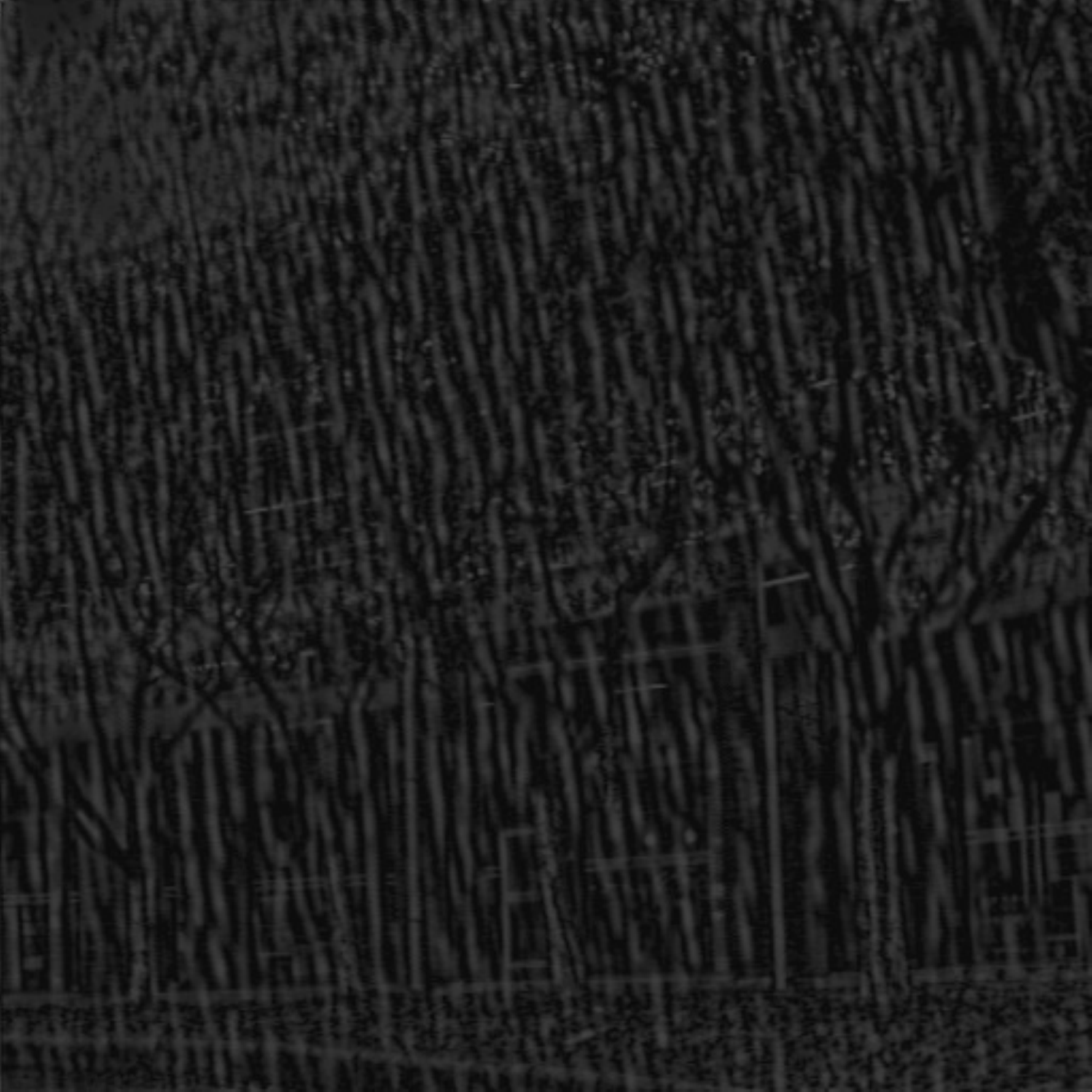} &
           	\includegraphics[width=\subwidth\linewidth]{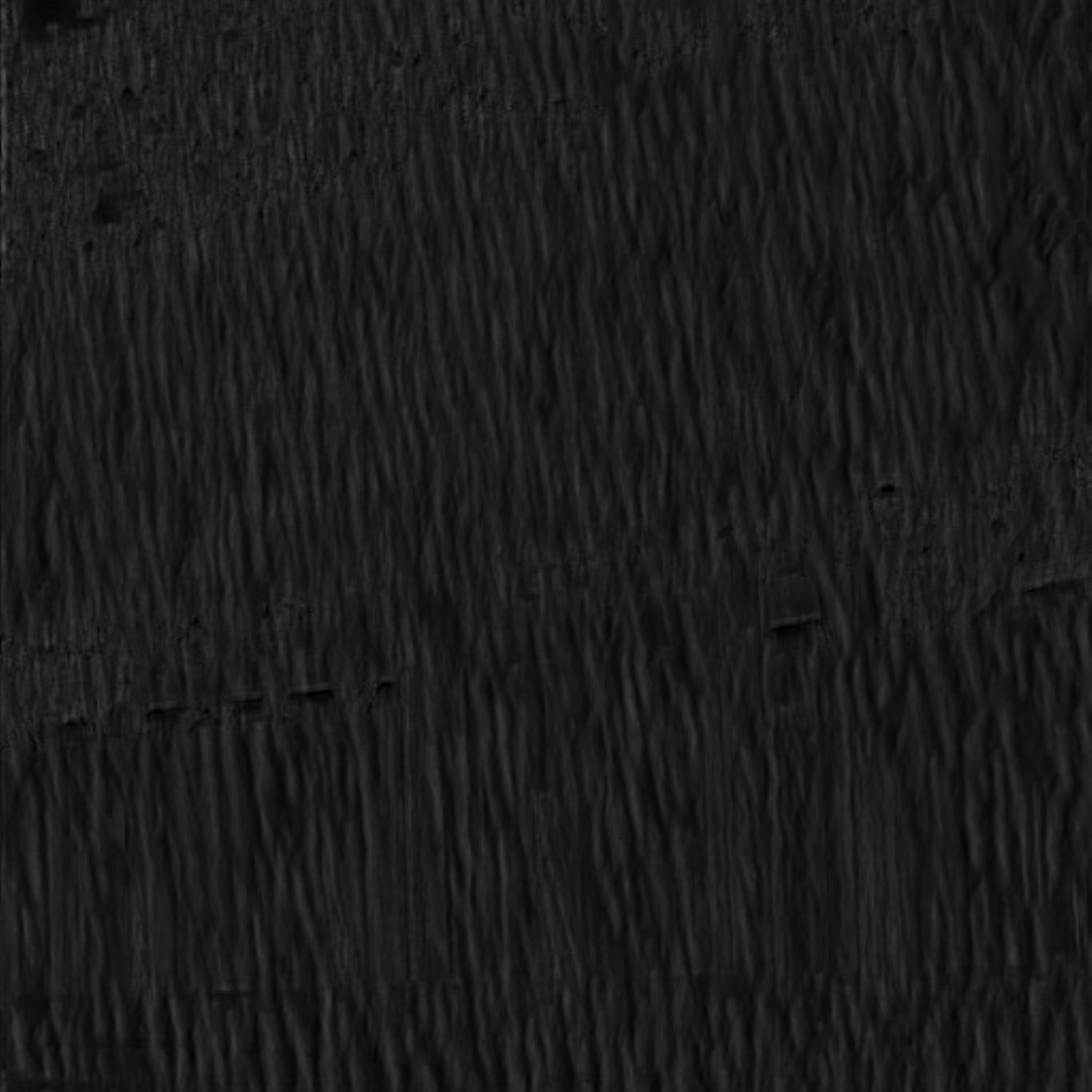} &
			\includegraphics[width=\subwidth\linewidth]{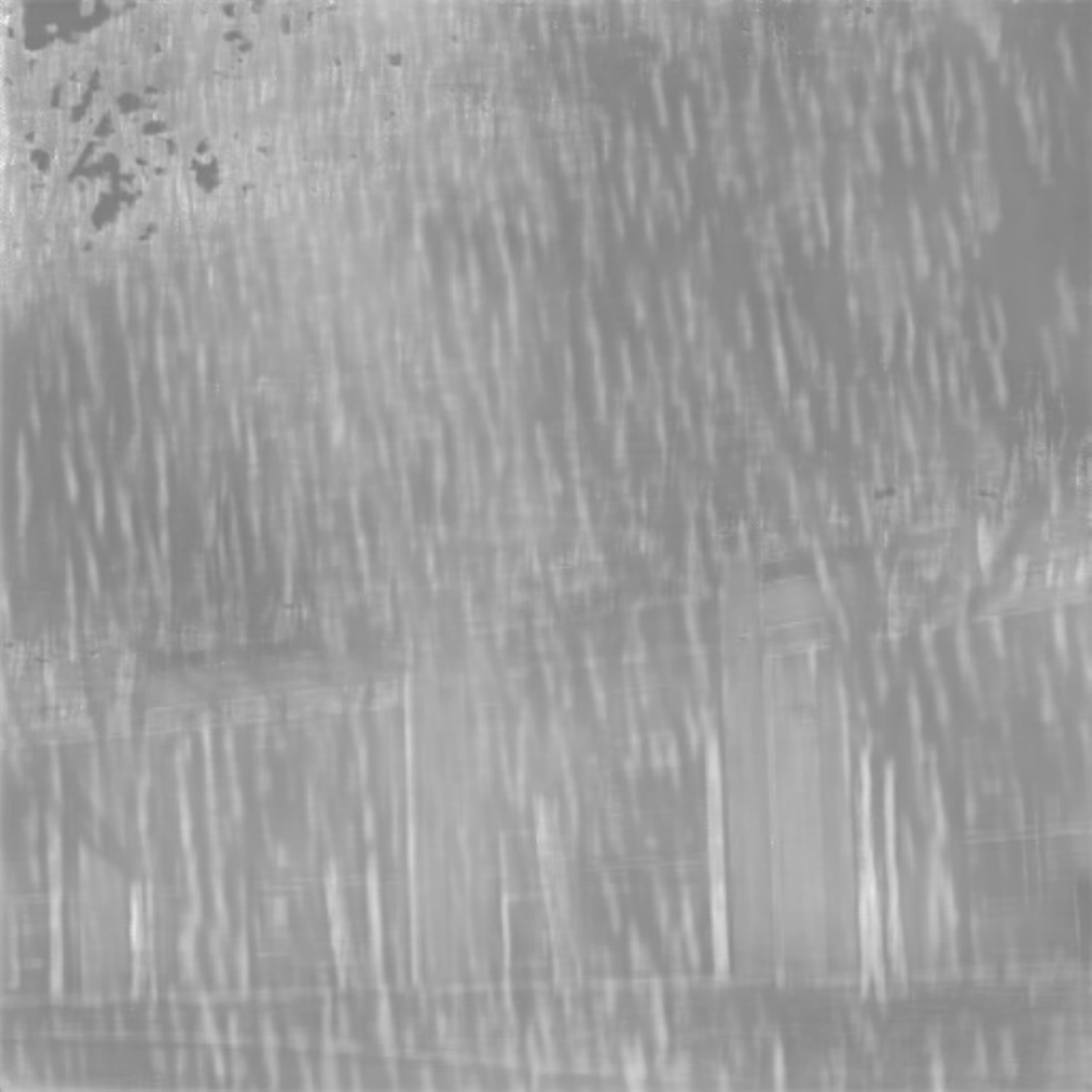} &
			\includegraphics[width=\subwidth\linewidth]{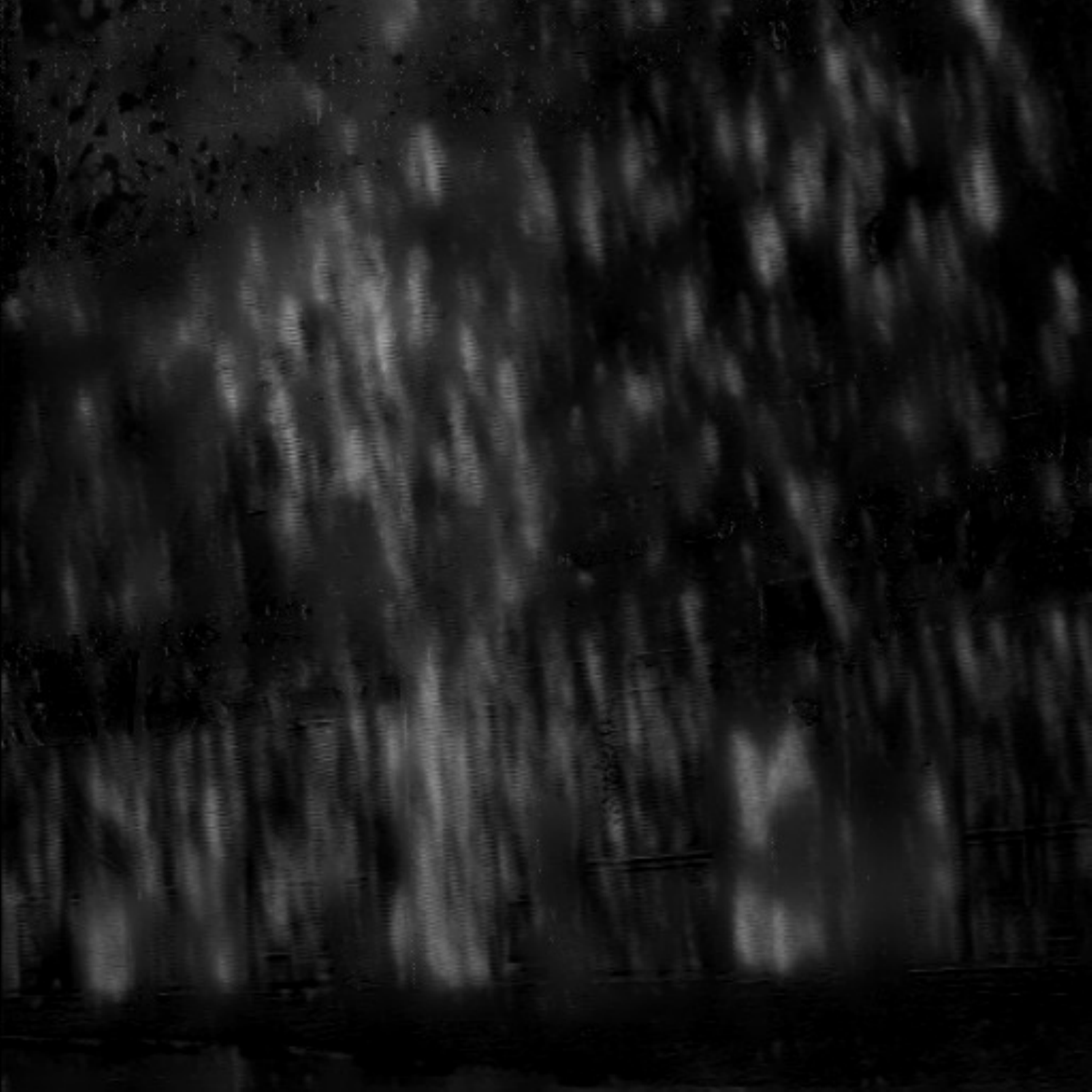} &
			\includegraphics[width=\subwidth\linewidth]{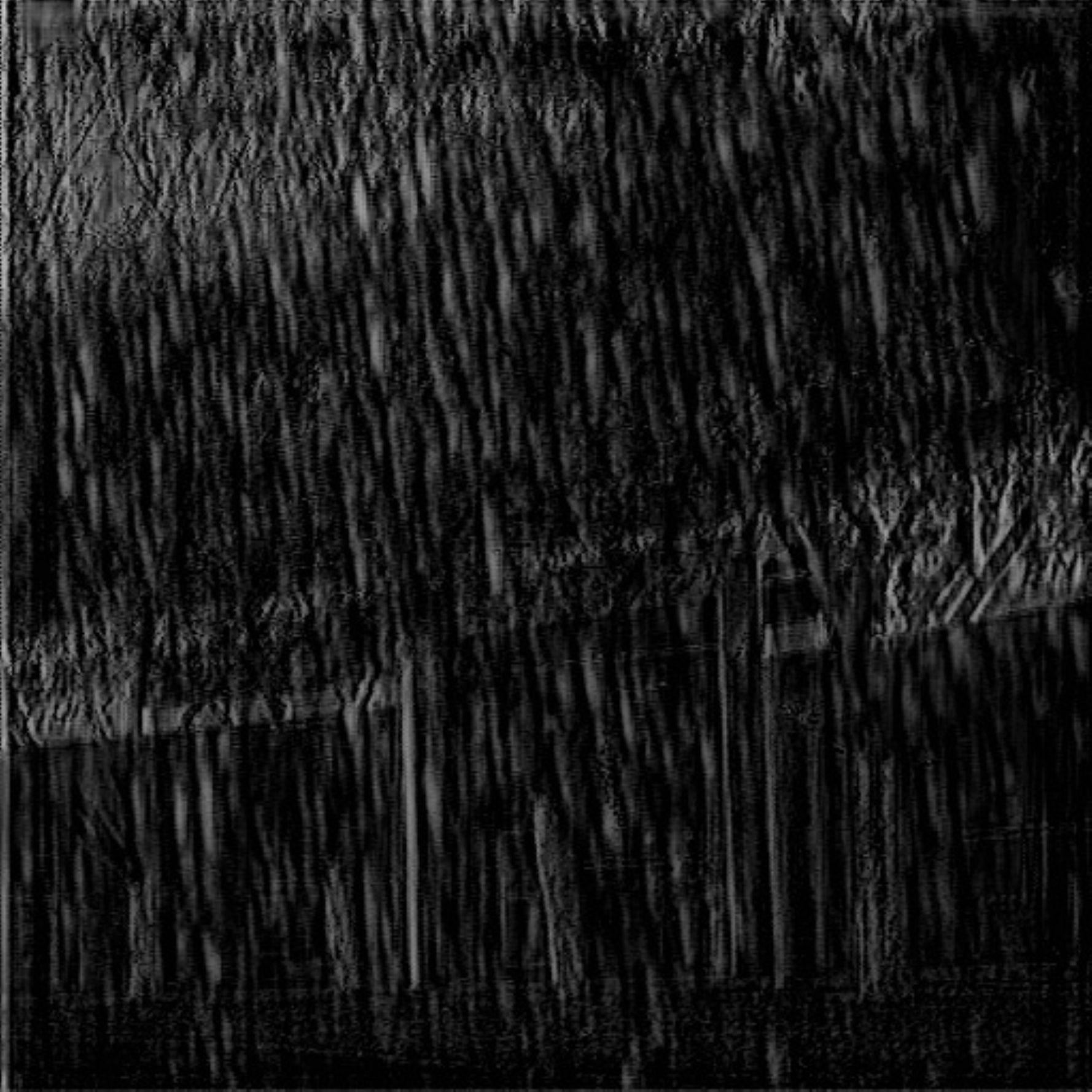} &
			\includegraphics[width=\subwidth\linewidth]{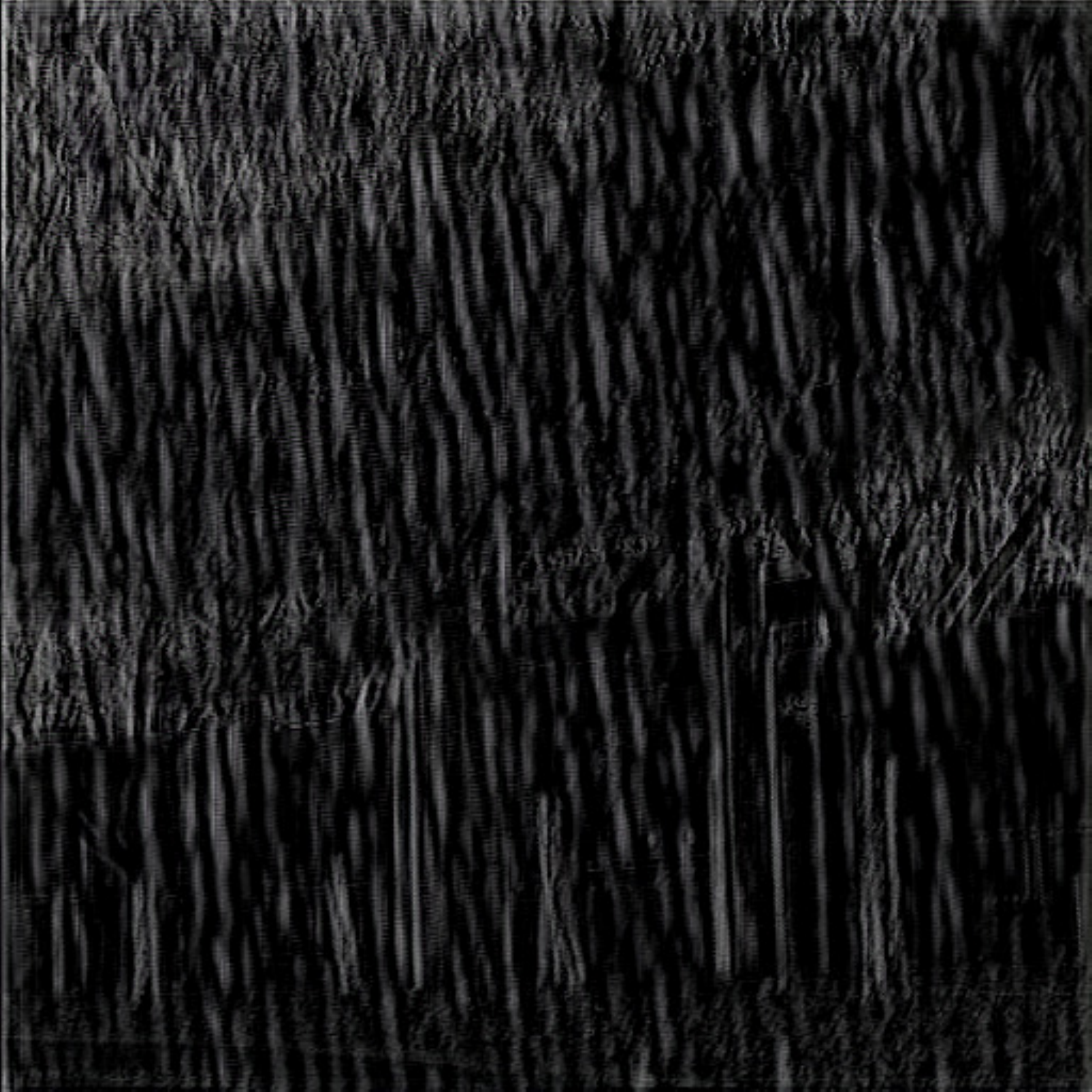} \\ %rain/116

            \scriptsize{Input}&
            \scriptsize{Li~\cite{li14}}&
            \scriptsize{Li~\cite{Li19a}}&
            \scriptsize{Wang~\cite{jiang2020}}&
            \scriptsize{Ding~\cite{ding21}}&
            \scriptsize{Ours-GD}&
            \scriptsize{Ours}\\
		\end{tabular}
	\end{center}
	\vspace{-0.02\textwidth}
	\caption{Comparison of real-world rain streaks estimated by the state-of-the-art methods~\cite{li14,Li19a,wang19b,ding21} and our method. The last two columns show the rain streaks obtained by our method with \revised{only} global discriminator and global-local discriminator, respectively.}
    \label{fig:rain streaks}
    \vspace{-2mm}
\end{figure*}

\renewcommand{\subwidth}{0.19}
\begin{figure}[h]
	\renewcommand{\tabcolsep}{1.0pt}
	\renewcommand\arraystretch{0.8}
	\begin{center}
		\begin{tabular}{ccccc}
			\includegraphics[width=\subwidth\linewidth]{fig2/test/input/117} &
           	\includegraphics[width=\subwidth\linewidth]{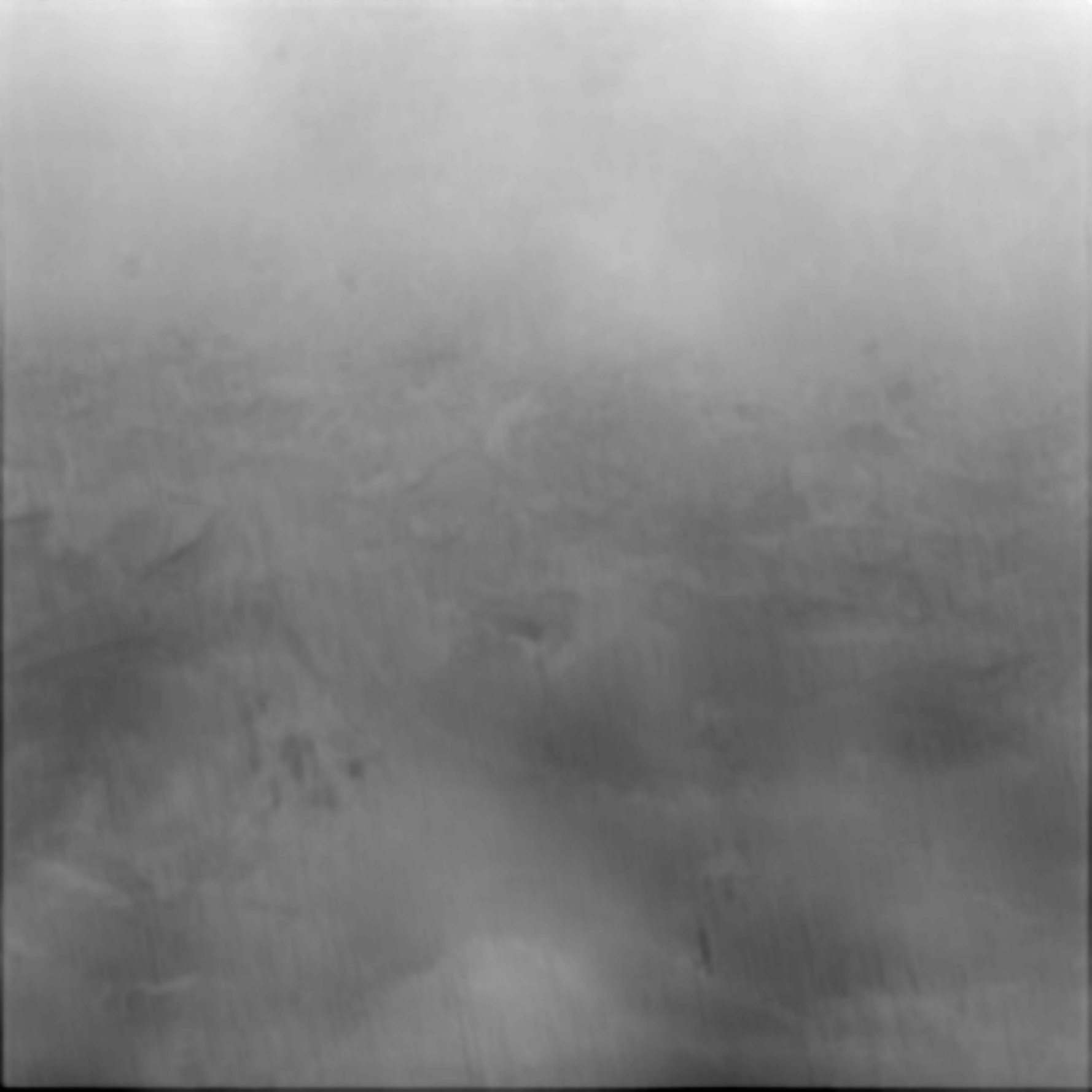} &
			\includegraphics[width=\subwidth\linewidth]{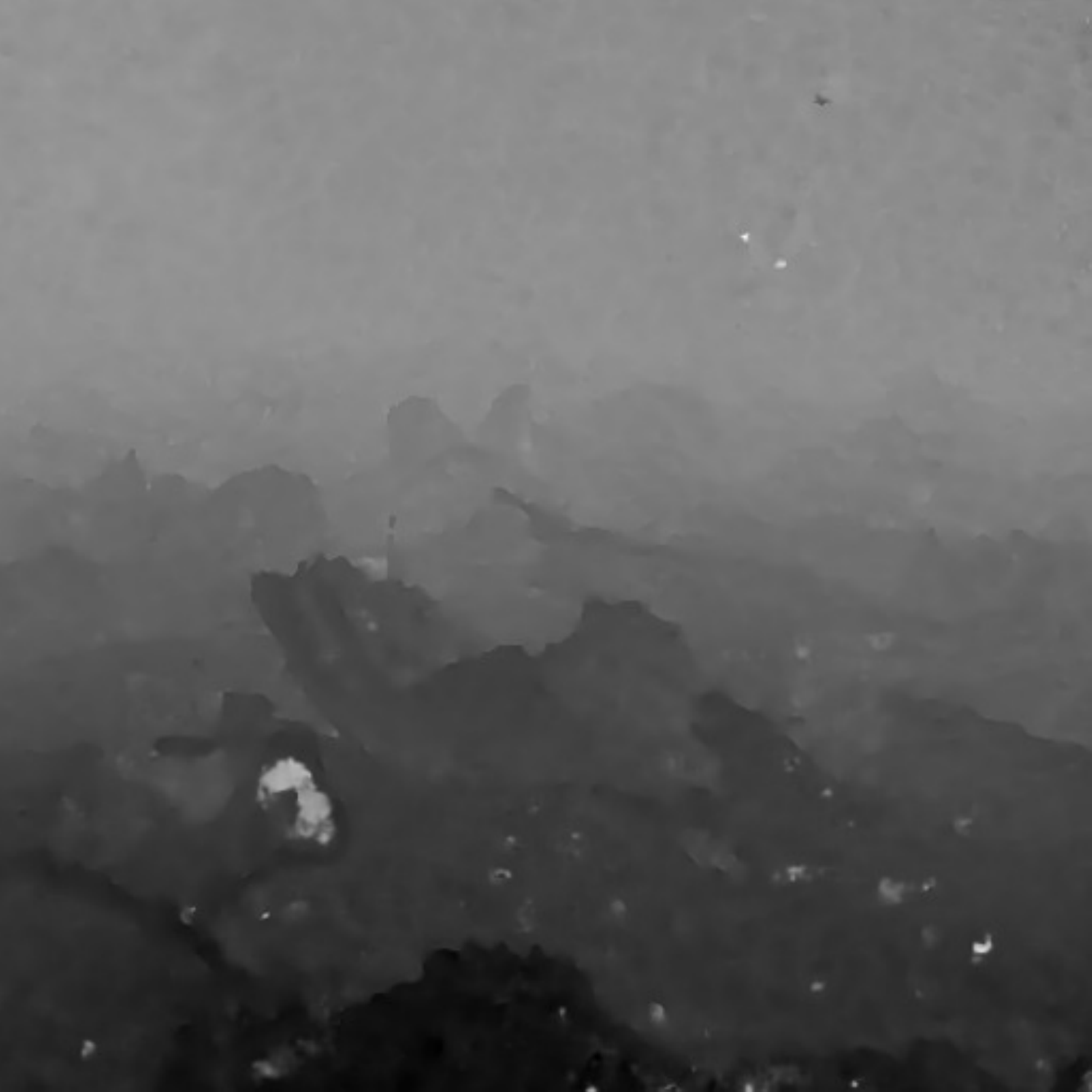} &
			\includegraphics[width=\subwidth\linewidth]{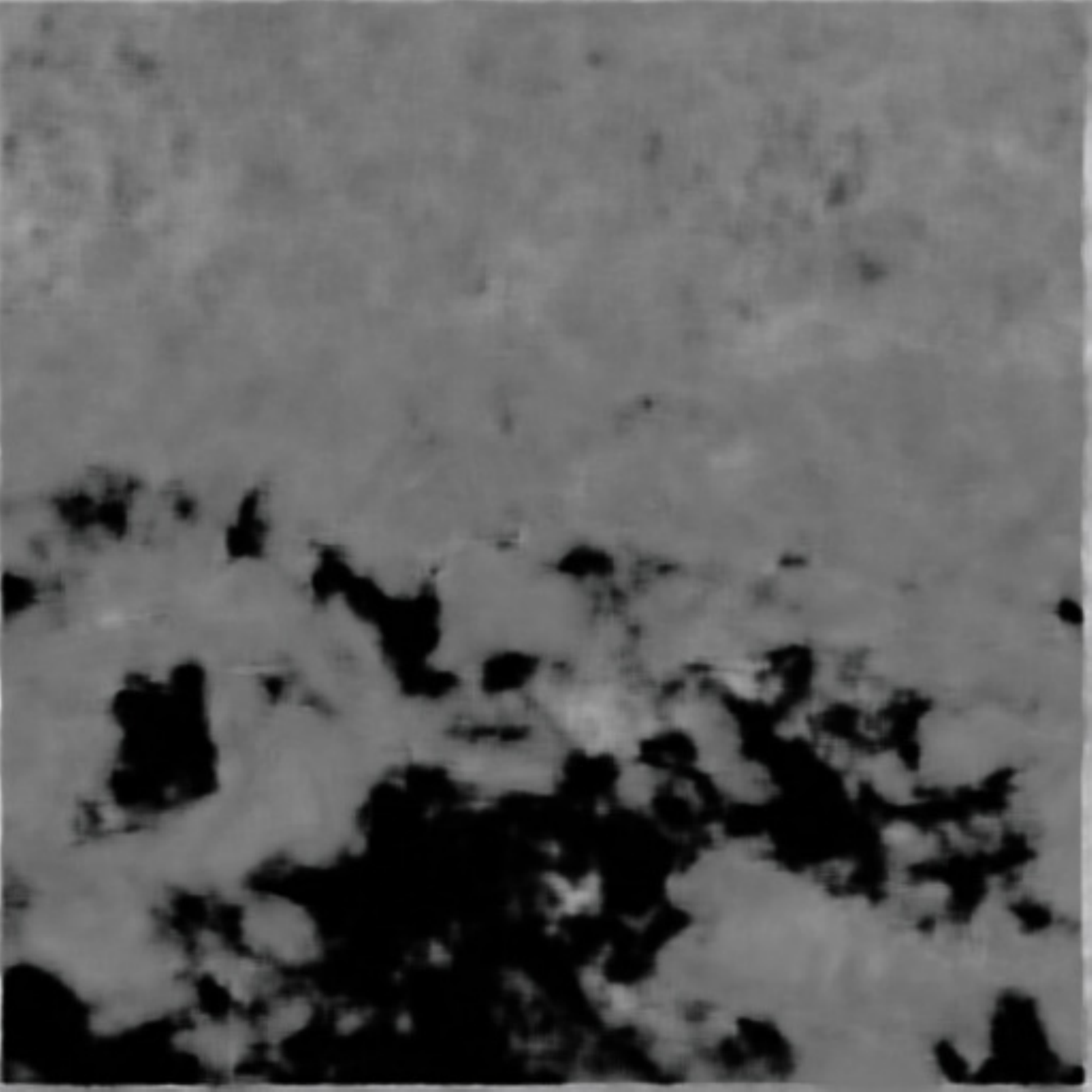} &
			\includegraphics[width=\subwidth\linewidth]{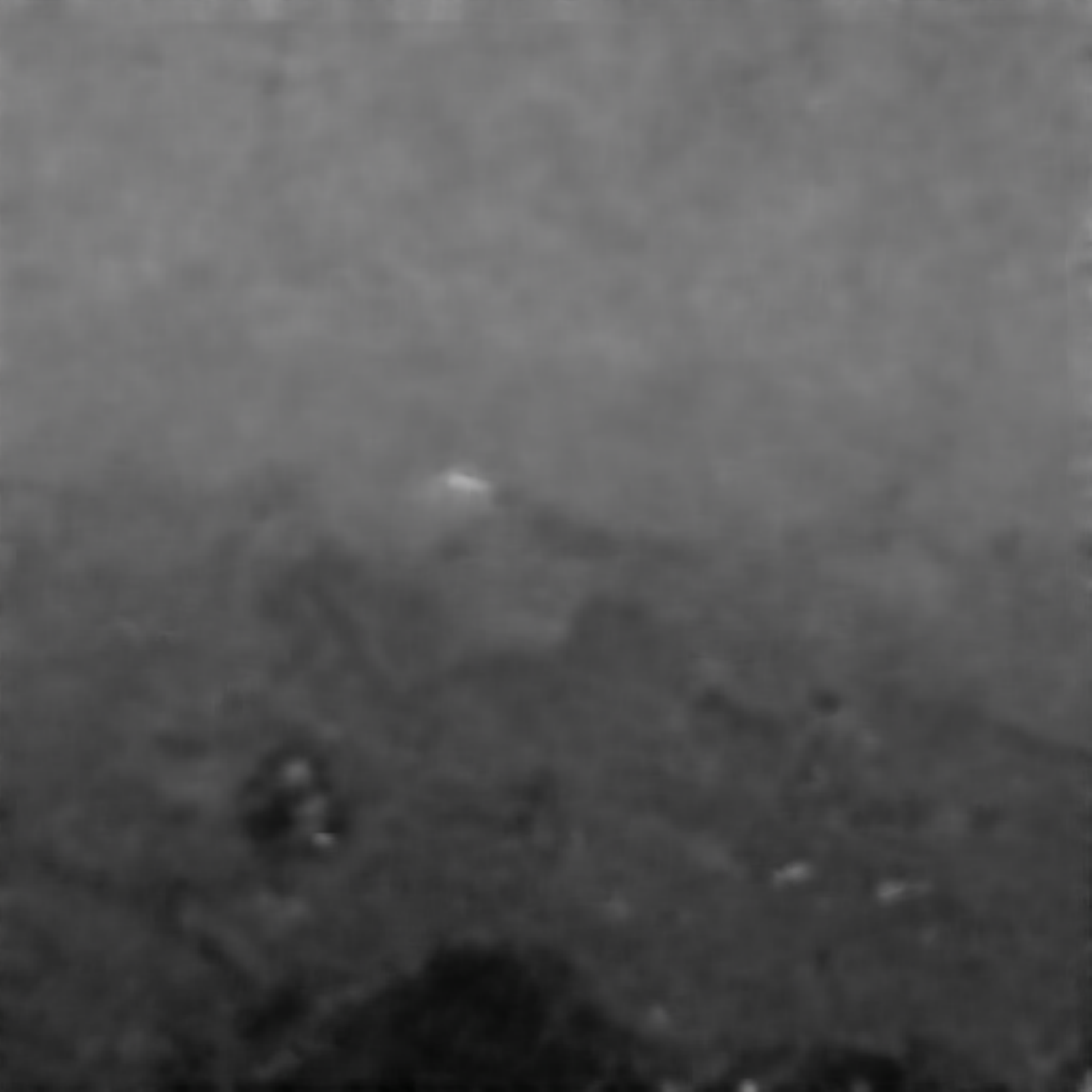} \\ %rain/116

			\includegraphics[width=\subwidth\linewidth]{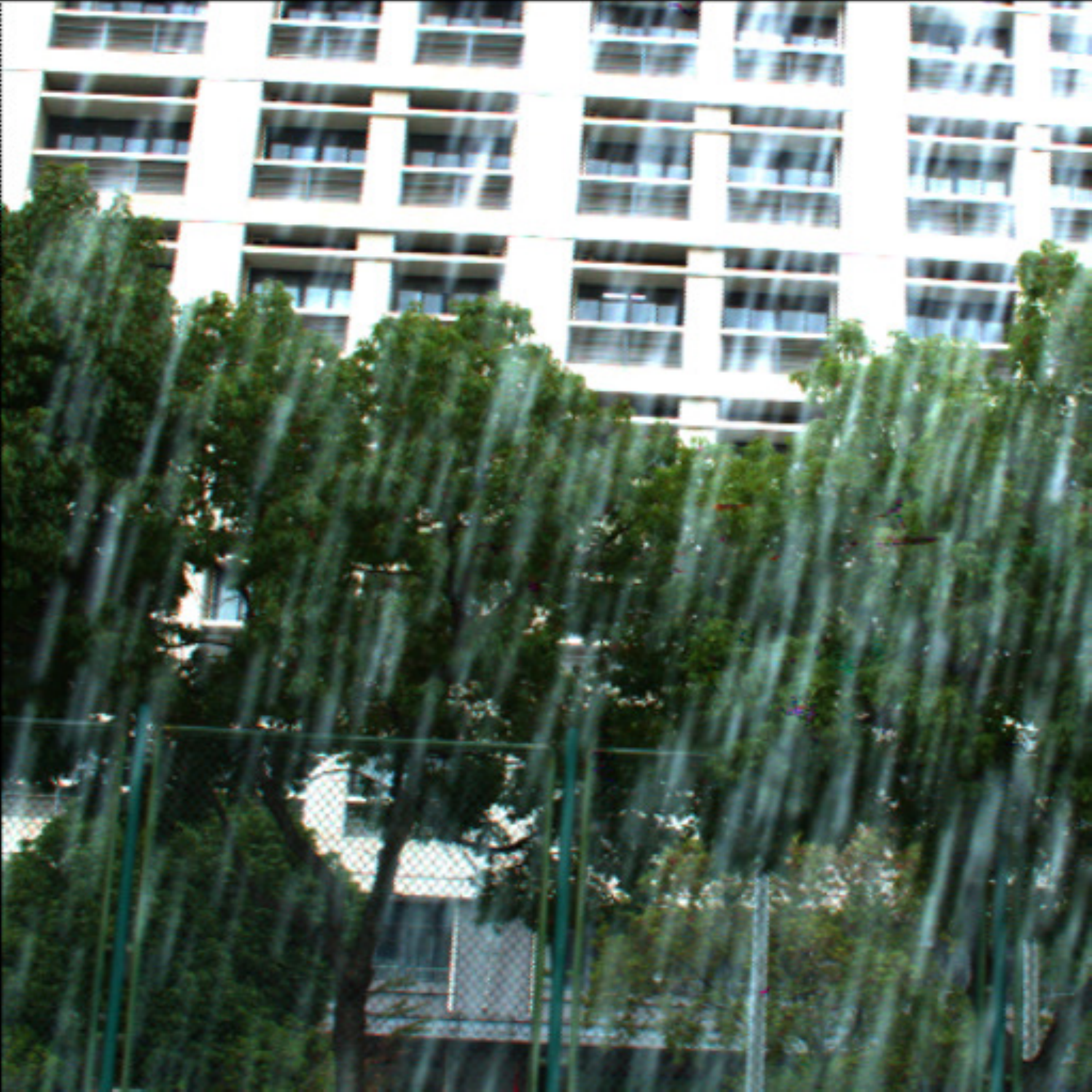} &
           	\includegraphics[width=\subwidth\linewidth]{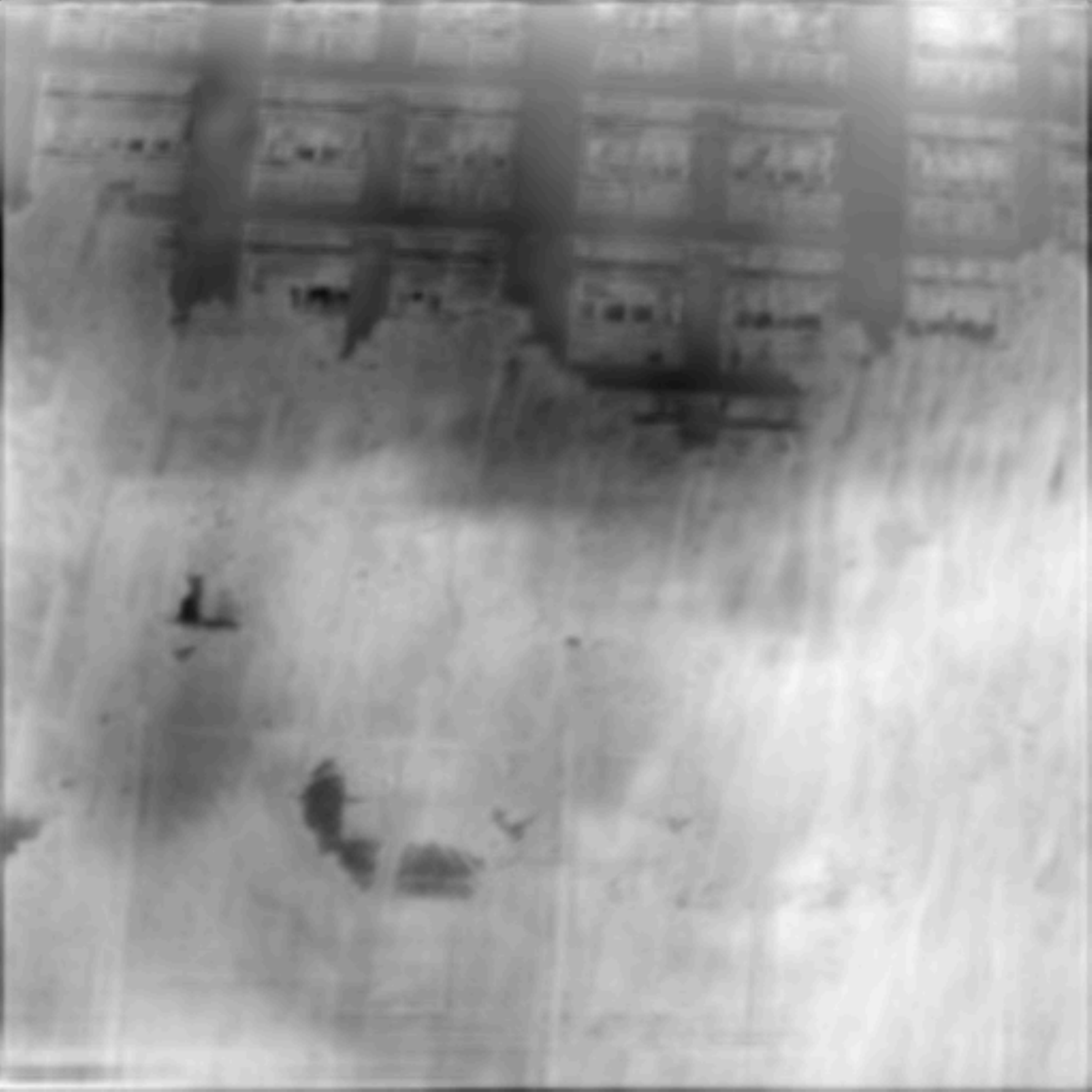} &
			\includegraphics[width=\subwidth\linewidth]{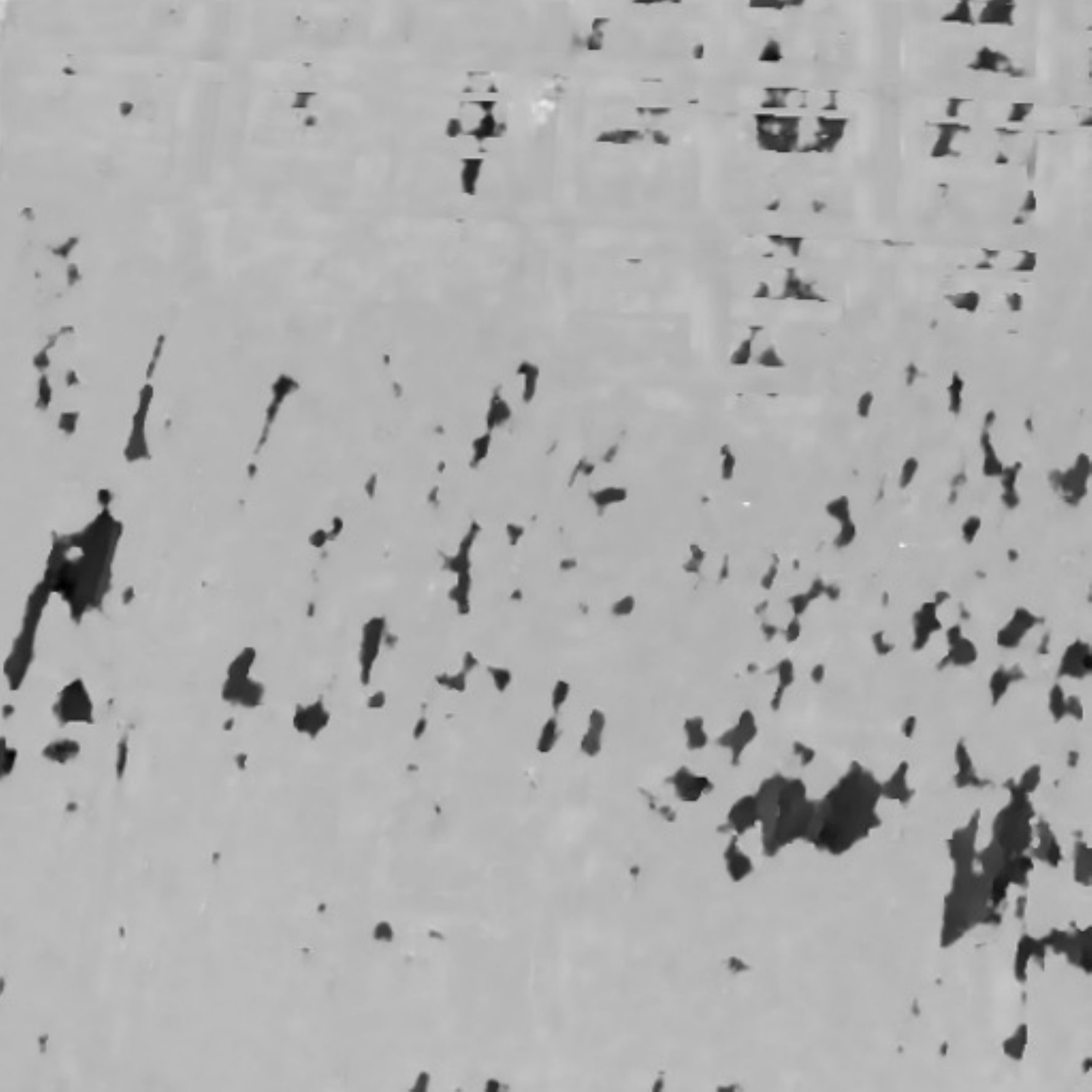} &
			\includegraphics[width=\subwidth\linewidth]{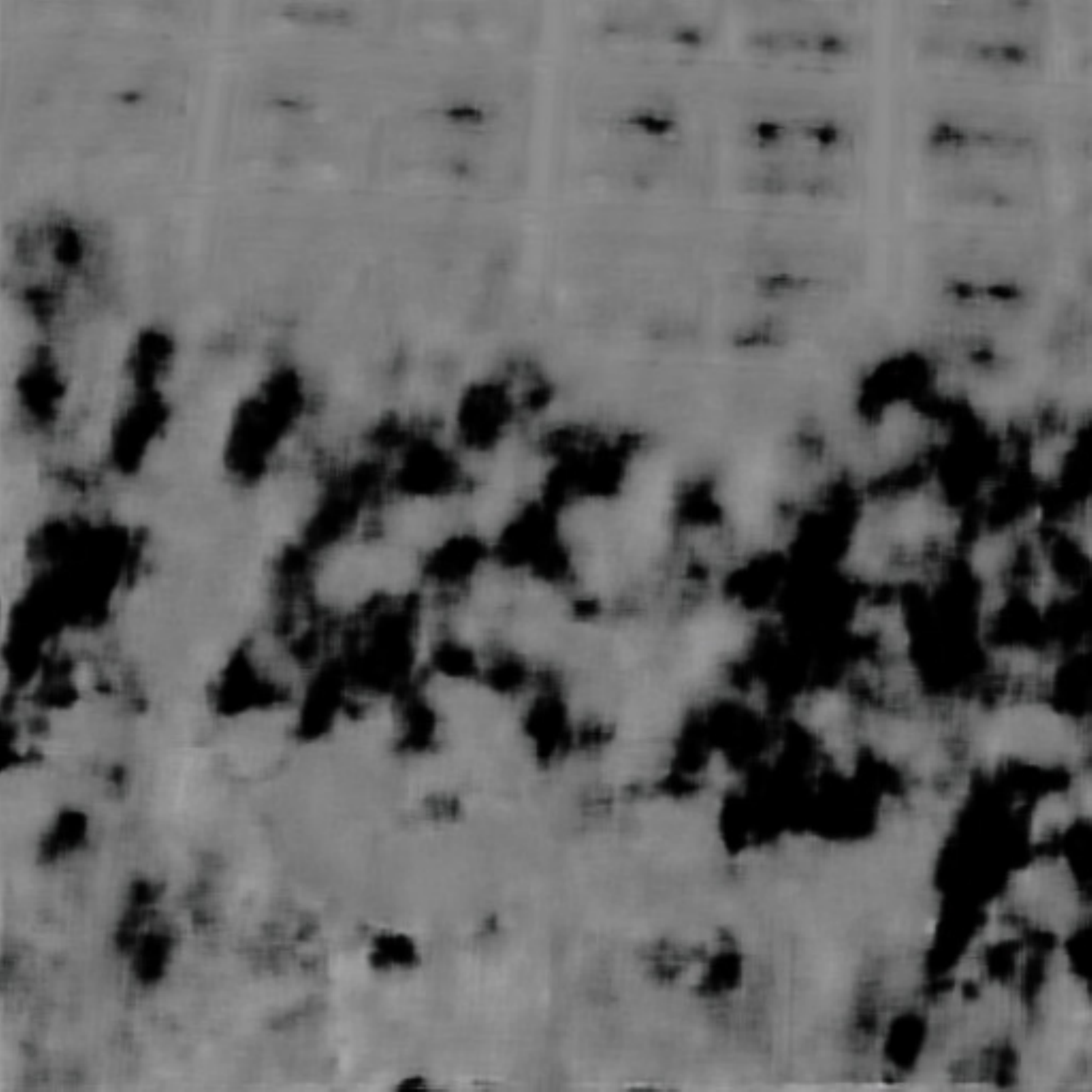} &
			\includegraphics[width=\subwidth\linewidth]{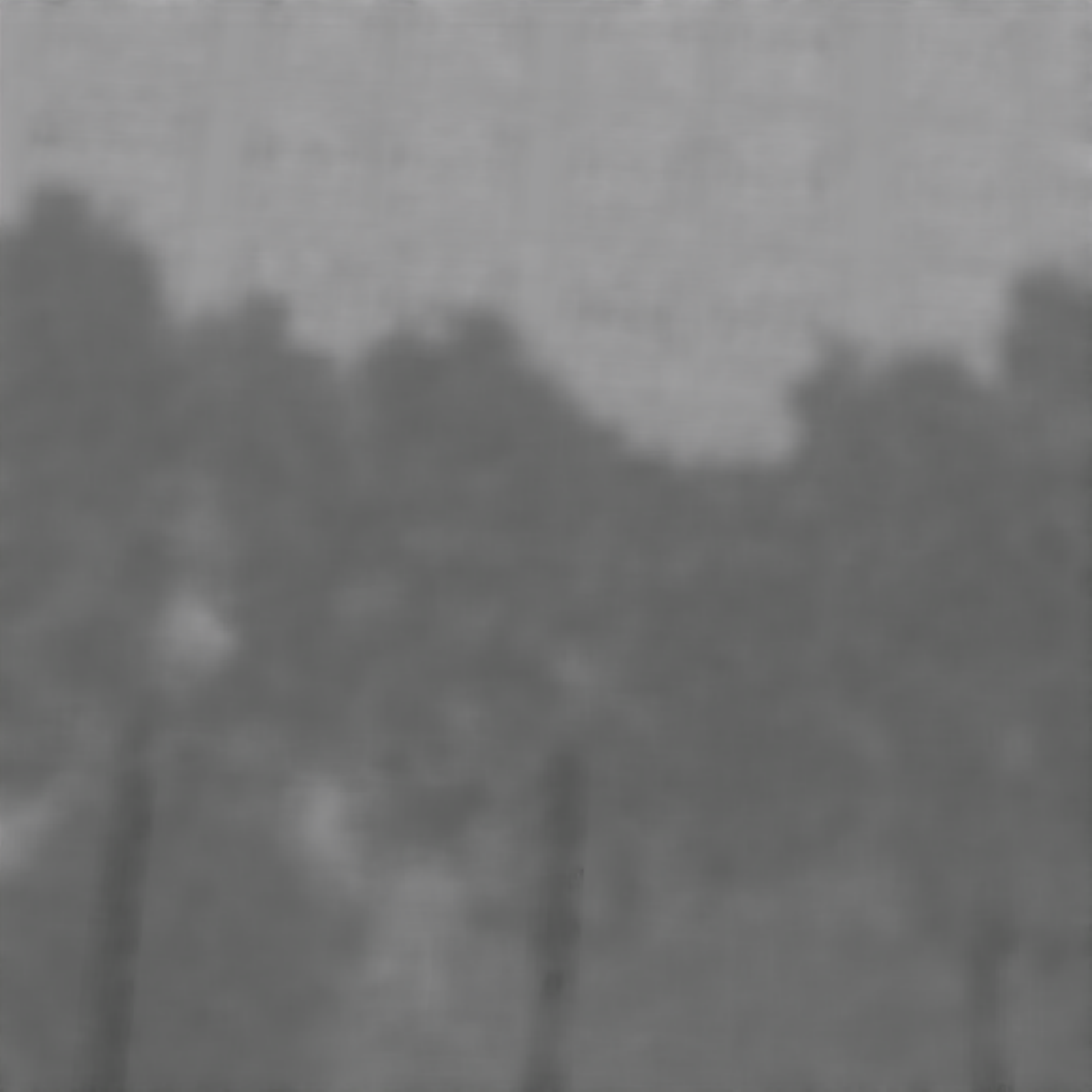} \\ %rain/116

            \scriptsize{Input}&
            \scriptsize{Li~\cite{Li19a}} &
           % \scriptsize{Jeon~\cite{jeon15}} &
            \scriptsize{Tsai~\cite{tsai20}} &
            \scriptsize{Ding~\cite{ding21}} &
            \scriptsize{Ours}\\

		\end{tabular}
	\end{center}
    	\vspace{-0.02\textwidth}
	\caption{Comparison of depth maps estimated by our method and the state-of-the-art methods~\cite{Li19a,tsai20,ding21} on a synthetic rainy LFI (first row) and a real-world rainy LFI (second row).}
    \label{fig:depth}
    \vspace{-0.02\textwidth}
\end{figure}

\section{Experimental Results and Discussion}

\subsection{Network Training Details}
\label{subsec:training_details}

Our proposed \revised{LFI} dataset, called RLMB, includes $400$ synthetic rainy LFIs and $200$ real-world rainy LFIs. It is divided into a training set of $300$ synthetic rainy LFIs and $100$ real-world rainy LFIs, and a test set of the remaining images. Each LFI contains $9\times9$ sub-views. Our \revised{proposed} 4D-MGP-SRRNet is implemented in PyTorch on a PC with two NVIDIA GeForce RTX 3090 GPUs. Adam optimization is adopted to train our network with a learning rate set to $0.0002$ for a total of $200$ epochs. We reduce the learning rate by a factor of $0.5$ at every $80$ epochs. The constant hyper-parameters in Eq.~\ref{eq:Ls}, Eq.~\ref{eq:Lr}, Eq.~\ref{eq:Lg} and Eq.~\ref{eq:Ld} are set as $\lambda_p=0.04$, $\lambda_{p,real}=0.04$, $\lambda_{GP}=0.015$, $\lambda_{p,g}=0.04$ and $\lambda_{gan}=0.01$. The $\omega$ in the MSGP is \revised{empirically} set as $0.5$.

In each iteration, the rainy LFI is first fed into MGPDNet to detect rain streaks. The residuals of the rainy LFI and the obtained rain streak map are then fed into DERNet, which directly loads the pre-trained parameters to estimate the depth map. The depth map is converted to a fog map later. Finally, the rainy LFI concatenated with the corresponding rain streaks and fog maps are fed into RNNAT for rain removal. Our MGPDNet and RNNAT are learning together with the parameters of DERNet frozen.

\subsection{Quantitative Evaluation}
\label{subsec:quantitative evaluation}

We conduct quantitative evaluation on the synthetic rainy LFIs. We compare our method with competing methods~\cite{wang19b,Li19a,wei19,jiang2020,Yang20b,ren20,jiang20,yasarla20,zamir2021,hu2021,ding21}\revised{~\cite{xiao22,zamir22}}, with their source codes re-trained on our generated rainy LFIs.

Two most widely used metrics, peak signal-to-noise ratio (PSNR) and structural similarity index (SSIM), are adopted to conduct our quantitative evaluation. The average PSNR and SSIM values calculated from all $100$ test synthetic LFIs are shown in Tab.~\ref{tab:result}, which demonstrates that the de-rained center sub-views recovered by our network with the global-local discriminator are much better (more than $1.30$db on PSNR and $0.019$ on SSIM) than those produced by other methods\revised{~\cite{wang19b,Li19a,wei19,jiang2020,Yang20b,ren20,jiang20,yasarla20,zamir2021,hu2021,ding21}}, and our network with the global-local discriminator outperforms itself with only the global discriminator. \revised{It is worth noting that the latest Transformer-based methods~\cite{xiao22,zamir22} perform comparably to our network on the test sub-set of our RLMB dataset. Specifically, on PSNR, methods~\cite{xiao22,zamir22} obtain the best and the second-best performances, while our network with/without local discriminator obtain the third-best and fourth-best performances. On SSIM metric, our network with/without local discriminator obtain the best and the third-best performances, while methods~\cite{xiao22,zamir22} obtain the second-best and fourth-best performances.}

Experimental results for a close shot and a distant shot of real-world-like LFIs are shown in Fig.~\ref{fig:synthetic result2} and~\ref{fig:synthetic result8}. Some methods~\cite{Li19a,wei19,jiang20,zamir2021} cannot effectively remove rain or haze in most scenes, and only perform better in a small part of the scenes. \cite{jiang2020} removes fog/mist very well. Although it can sometimes obtain the second-best de-rained results, these results are not good enough. Although~\cite{hu2021} obtains satisfactory effects for removing rain and fog, a lot of blur is introduced to the de-rained images, such as in Fig.~\ref{fig:synthetic result8}. Methods~\cite{wang19b,Yang20b,ren20,yasarla20} can remove most of the rain streaks, but when they encounter very tiny rain streaks, they cannot clearly remove them, as shown in Fig.~\ref{fig:synthetic result2} and~\ref{fig:synthetic result8}. The residue of fog or rain streaks and the introduction of blur will obscure the details of the background. %and hinder visual tasks such as depth/disparity estimation and target recognition.}
\revised{The LFI-based rain removal method}~\cite{ding21} always performs the second best in challenging scenes. \revised{The performances of the latest methods~\cite{xiao22,zamir22} are comparable with the LFI-based method~\cite{ding21} even our proposed 4D-MGP-SRRNet on synthetic scenes.} Our method can clearly remove various rain streaks and fog, especially in challenging scenes, as shown in Fig.~\ref{fig:synthetic result8}. In conclusion, our method is able to remove rain streaks and fog clearly from LFIs captured in various synthetic challenging scenes, whether it is close or long shots.

%We also compare our method with the state-of-the-art video rain streaks removal methods~\cite{Li18v}~\cite{Liu18v}~\cite{li21}, as shown in Fig.~\ref{fig:Video}. The results demonstrate that the methods~\cite{Li18v}~\cite{Liu18v}~\cite{li21} have limited effect on rain removal from synthetic LFIs, especially for the first synthetic three scenes and last real-world scene in Fig.~\ref{fig:Video}. In addition, none of them are able to remove fog from the exhibited rainy LFIs as our proposed method.

\revised{In addition, we compare our proposed method with several state-of-the-art video rain streak removal methods~\cite{Li18v,Liu18v,li21,zhang2022}, as shown in Fig.~\ref{fig:Video} (Fig.~\tao{19} of the \textbf{Supplemental}). All these competing methods take all sub-views of an LFI as input (frame sequence) for rain streak removal.
%Fig.~\ref{fig:Video} exhibits the comparison of our method with \revised{four} state-of-the-art video rain streaks removal methods~\cite{Li18v,Liu18v,li21},
Fig.~\ref{fig:Video} demonstrates that our proposed method considerably outperforms the video rain streak removal methods. These results show that~\cite{Li18v,Liu18v,li21} have limited effects on rain streak removal from synthetic LFIs, especially for the \revised{first} three synthetic scenes and the last real-world scene. In addition, none of them is able to remove fog/mist from the exhibited rainy LFIs as our method.} \revised{\cite{zhang2022} can clearly remove rain streaks from the synthetic scenes (first three rows), but fails to clearly remove rain streaks from real-world scenes (last two rows) like our proposed network.}
%The results demonstrate that methods~\cite{Li18v}~\cite{Liu18v}~\cite{li21} have limited effects on rain streak removal from synthetic LFIs, especially for the left three synthetic scenes and last real-world scene. In addition, none of them is able to remove fog/mist from the exhibited rainy LFIs as our proposed method.

\renewcommand{\subwidth}{0.132}
\renewcommand{\ssubwidth}{0.066}
\begin{figure*}[h]
	\renewcommand{\tabcolsep}{0.8pt}
	\renewcommand\arraystretch{0.8}
	\begin{center}
		\begin{tabular}{cccccccccccccc}

            \multicolumn{2}{c}{\includegraphics[width=\subwidth\linewidth]{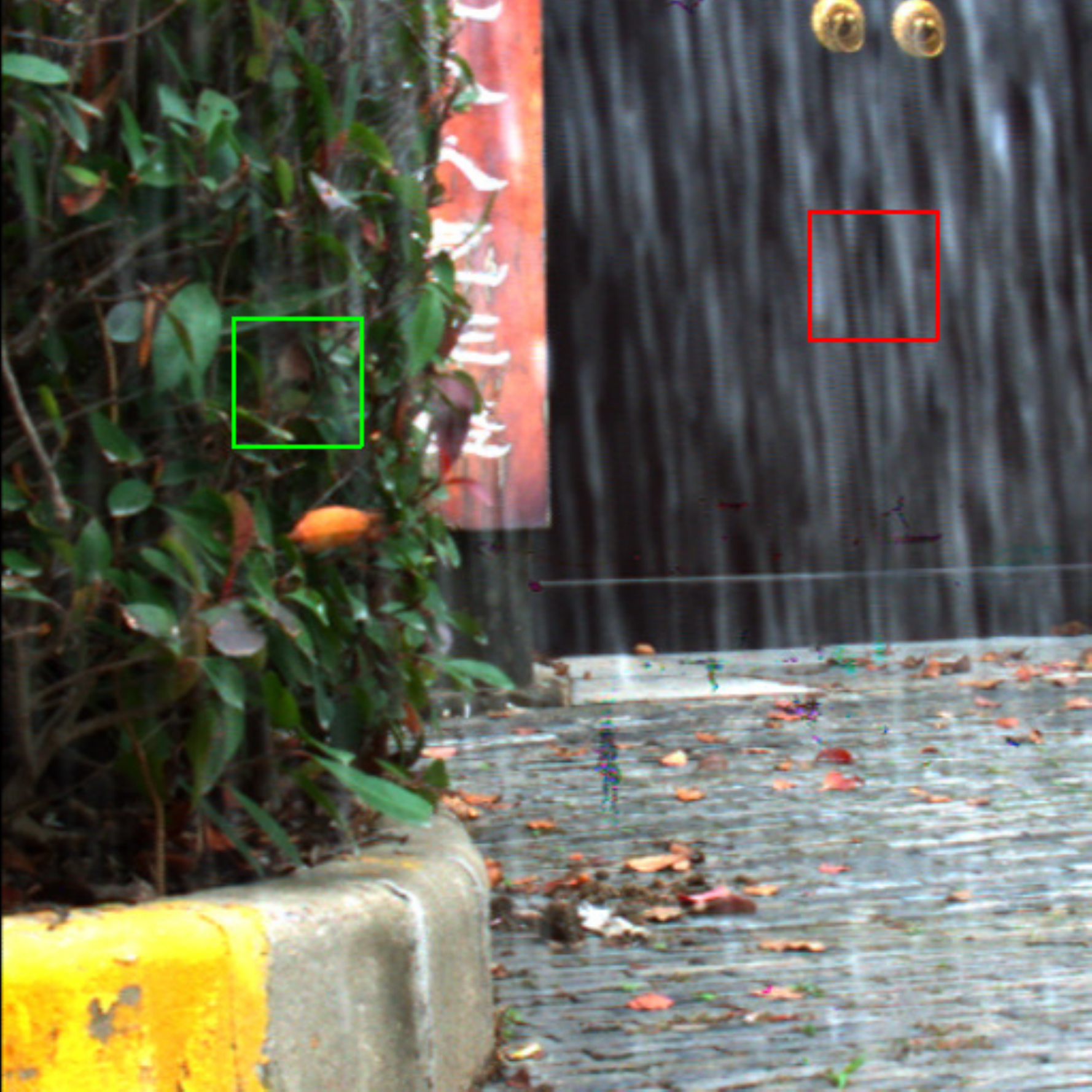}} &
            \multicolumn{2}{c}{\includegraphics[width=\subwidth\linewidth]{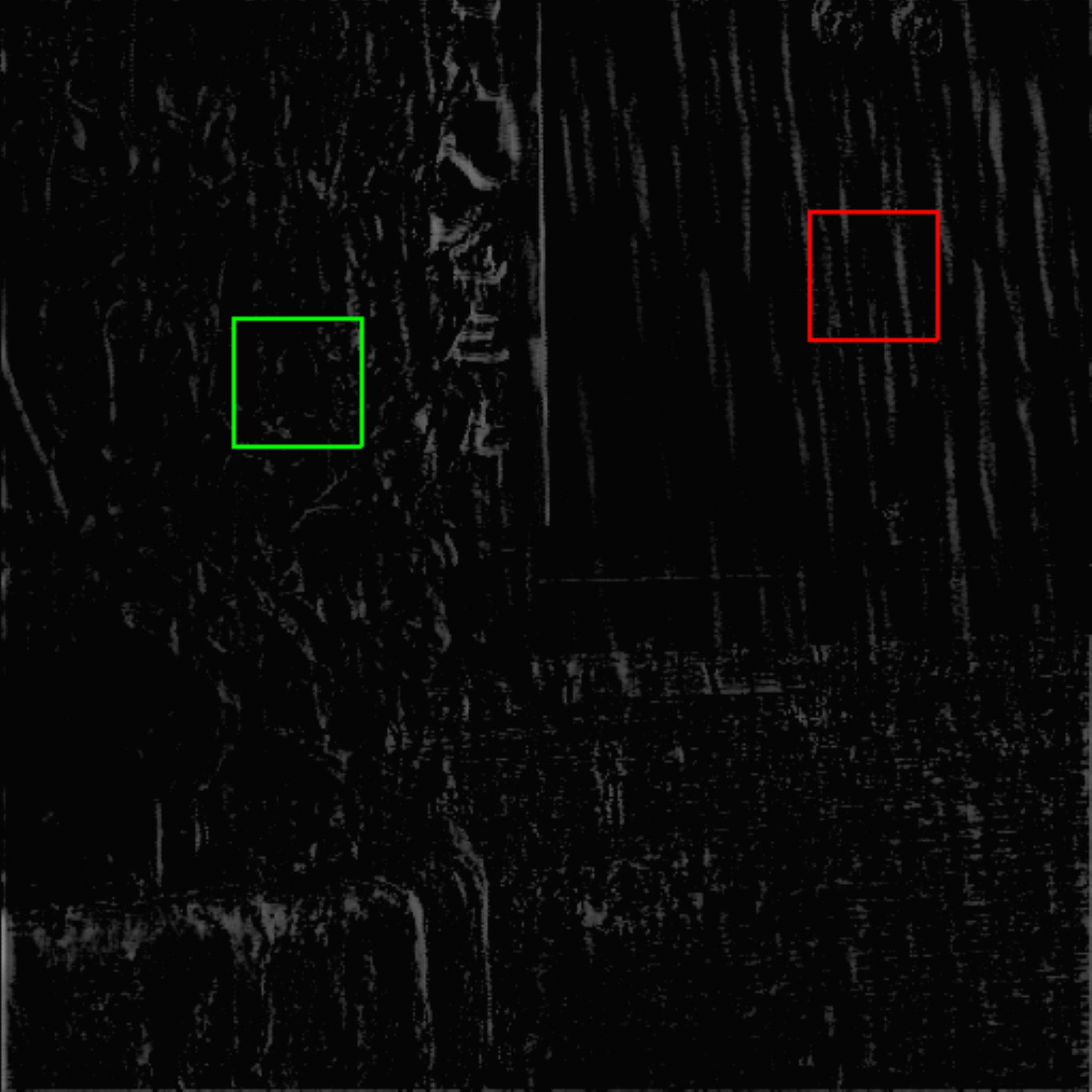}} &
            \multicolumn{2}{c}{\includegraphics[width=\subwidth\linewidth]{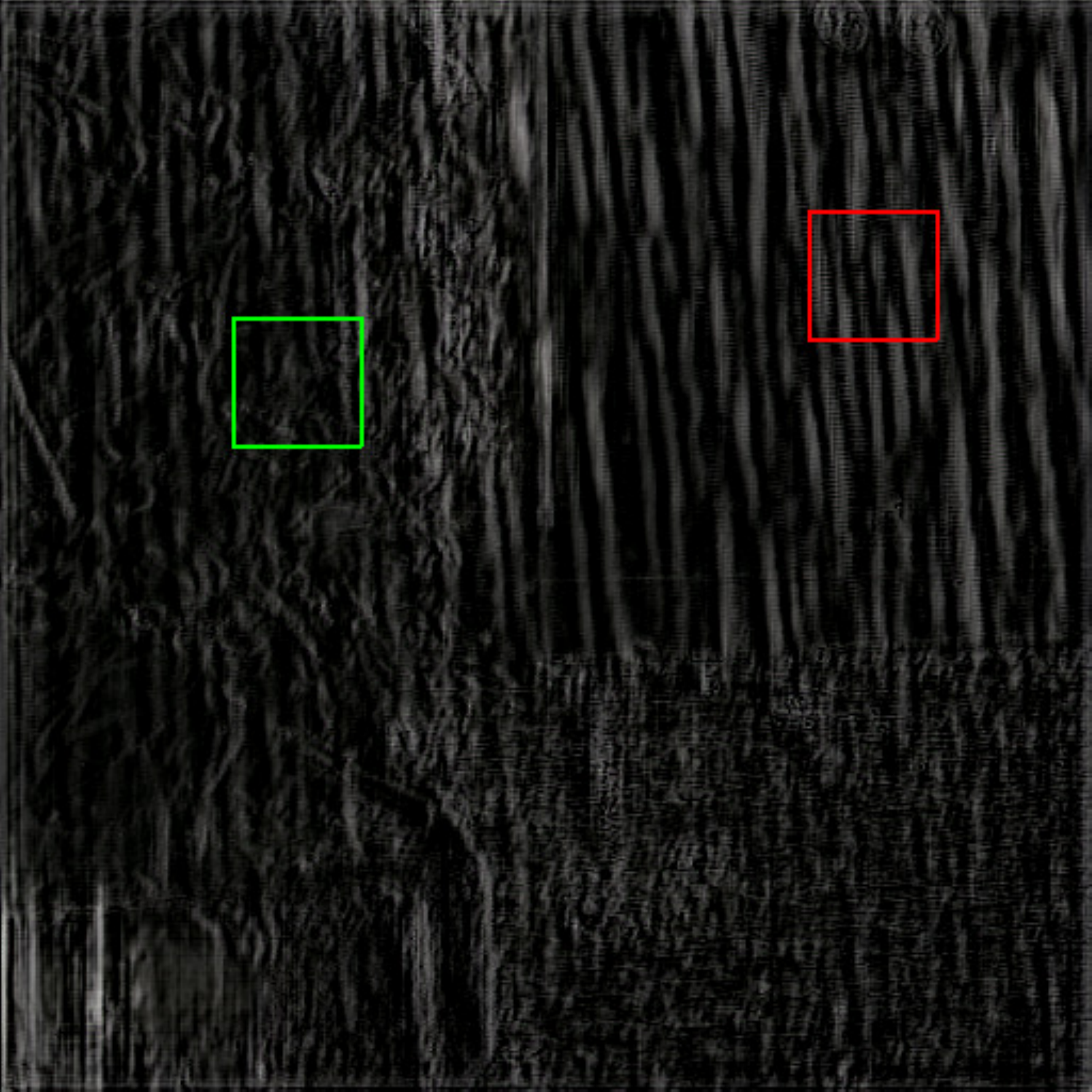}} &
            \multicolumn{2}{c}{\includegraphics[width=\subwidth\linewidth]{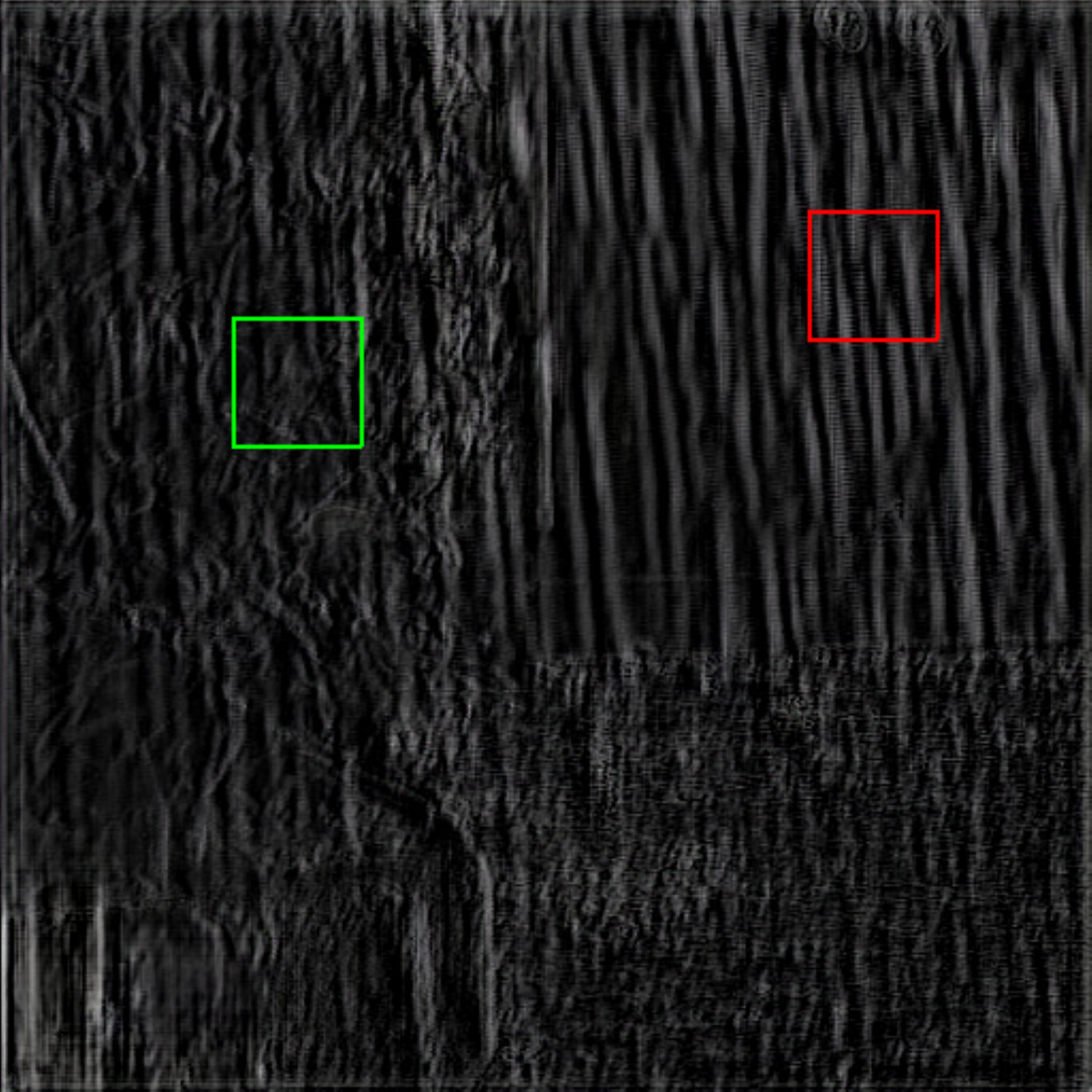}} &
            \multicolumn{2}{c}{\includegraphics[width=\subwidth\linewidth]{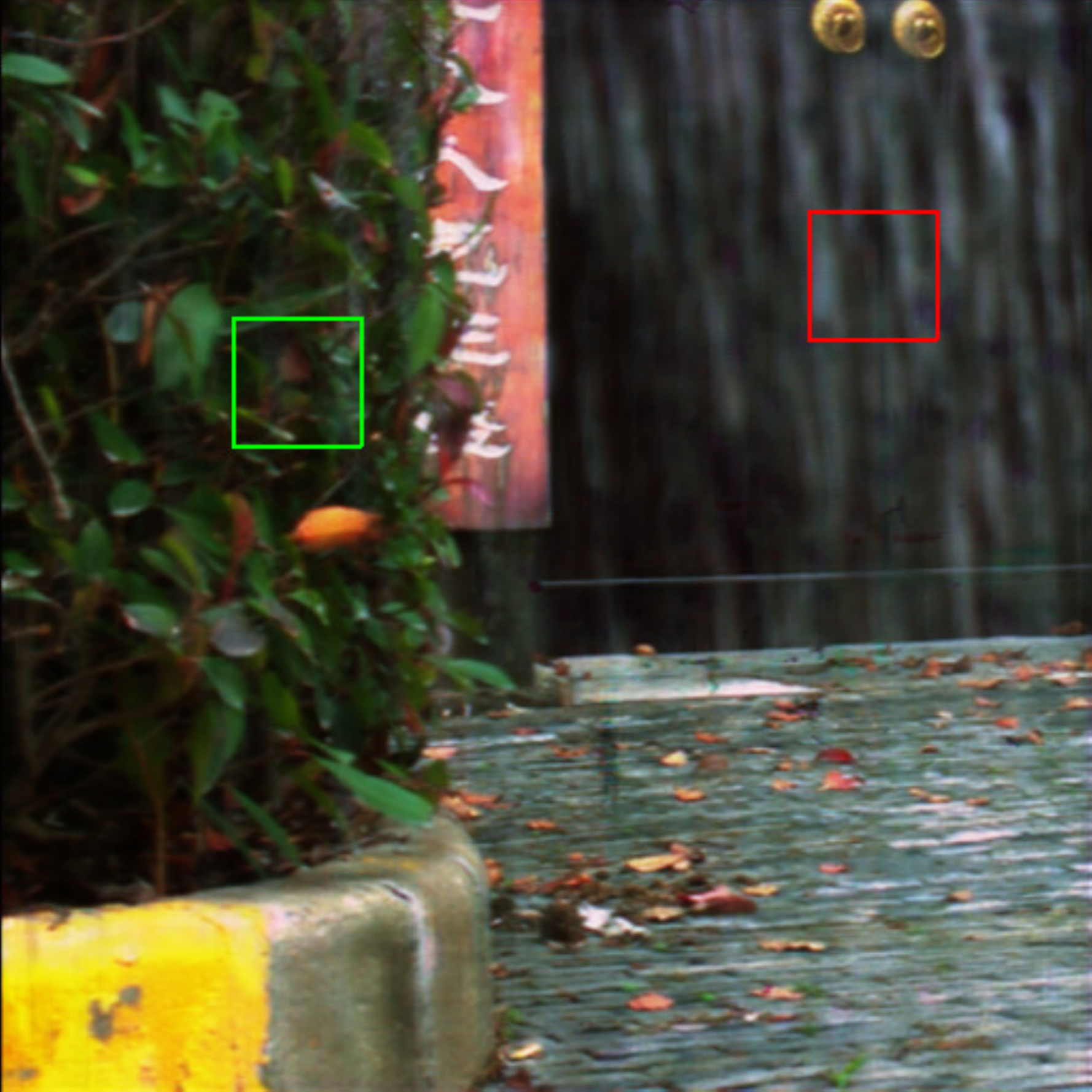}} &
            \multicolumn{2}{c}{\includegraphics[width=\subwidth\linewidth]{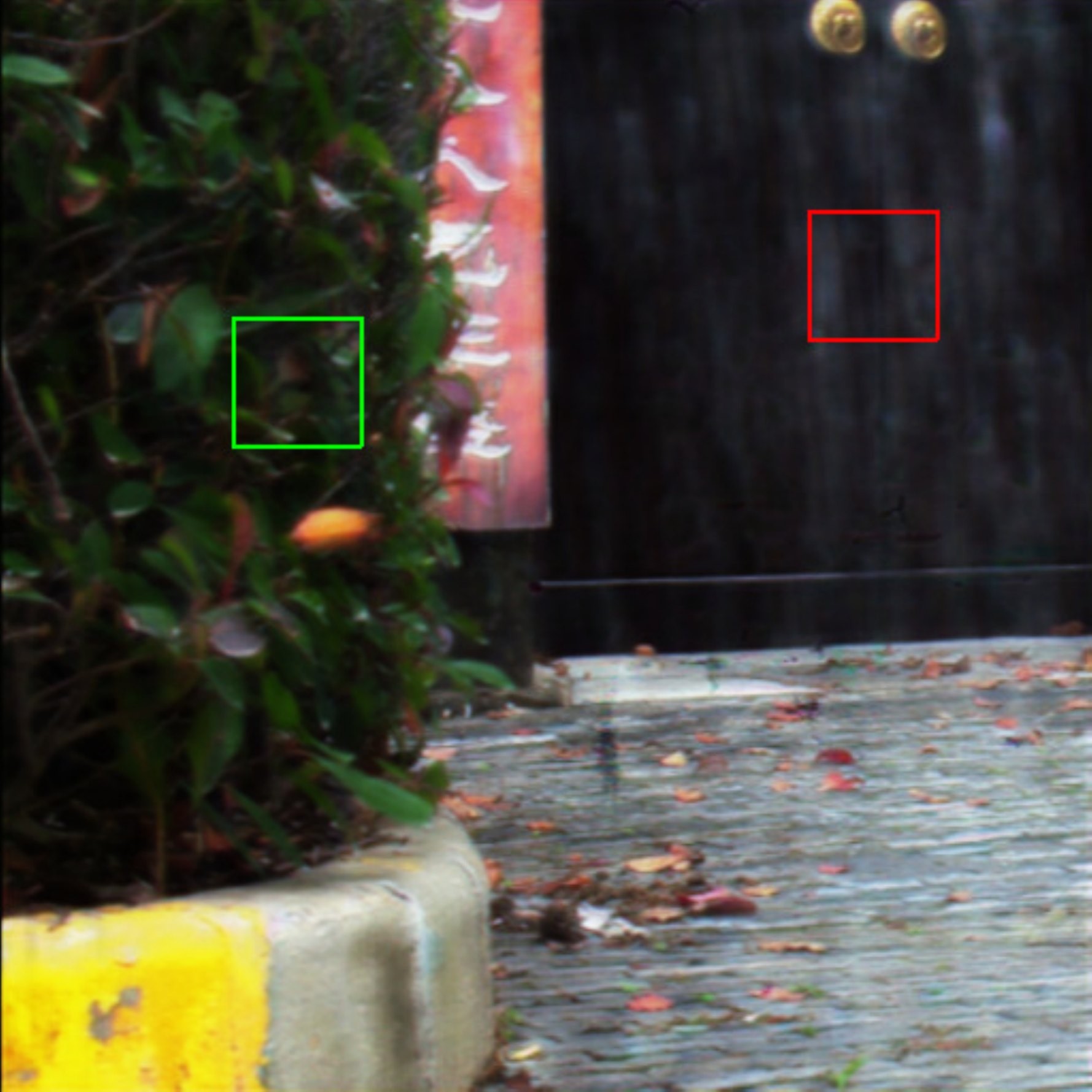}} &
            \multicolumn{2}{c}{\includegraphics[width=\subwidth\linewidth]{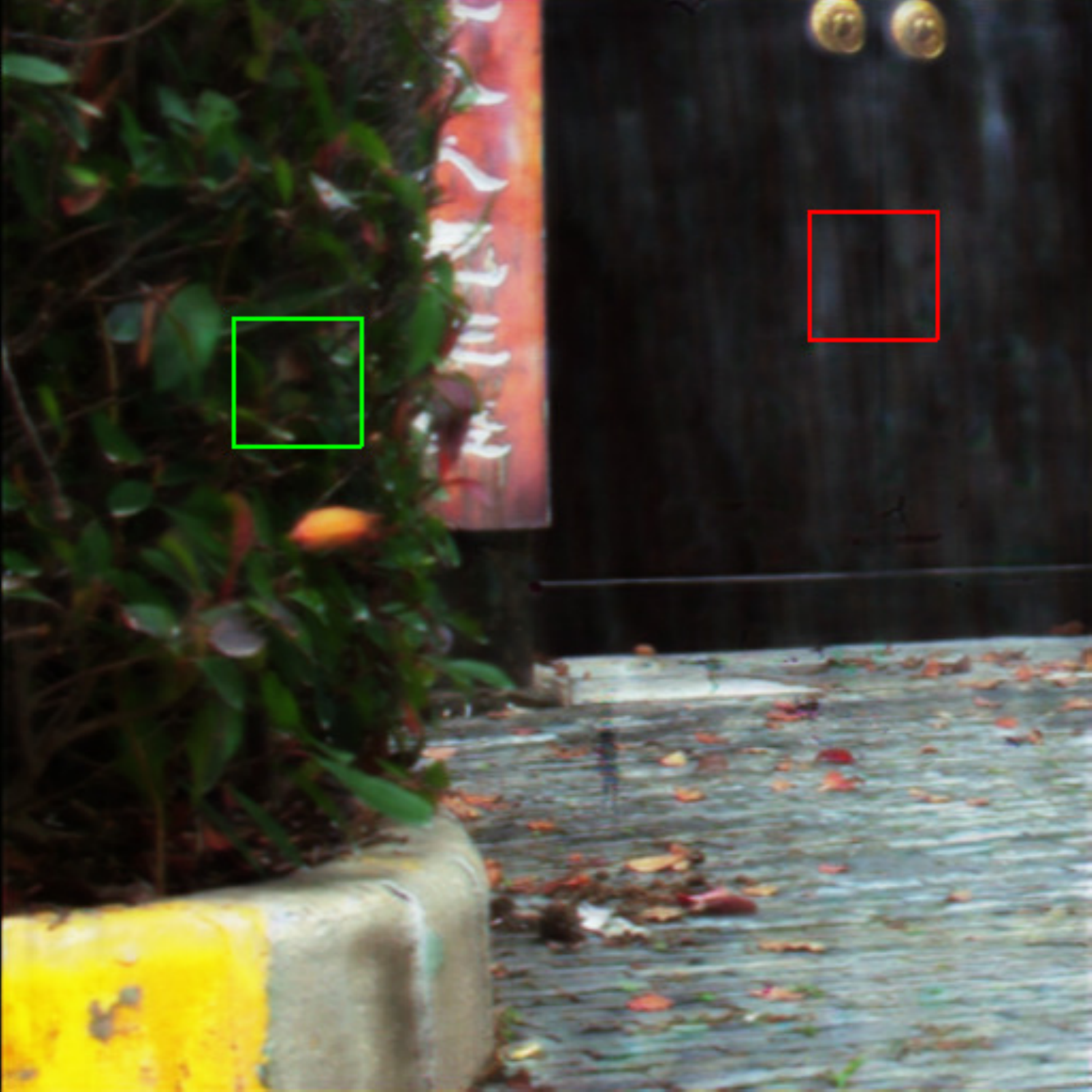}} \\

            \includegraphics[width=\ssubwidth\linewidth]{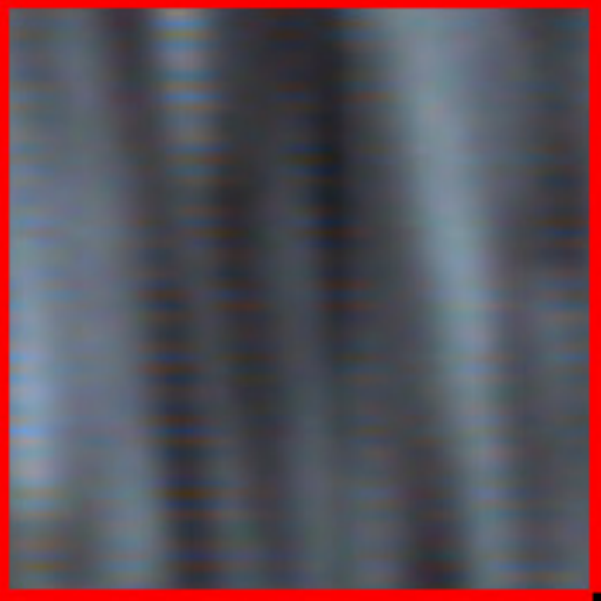} &
            \includegraphics[width=\ssubwidth\linewidth]{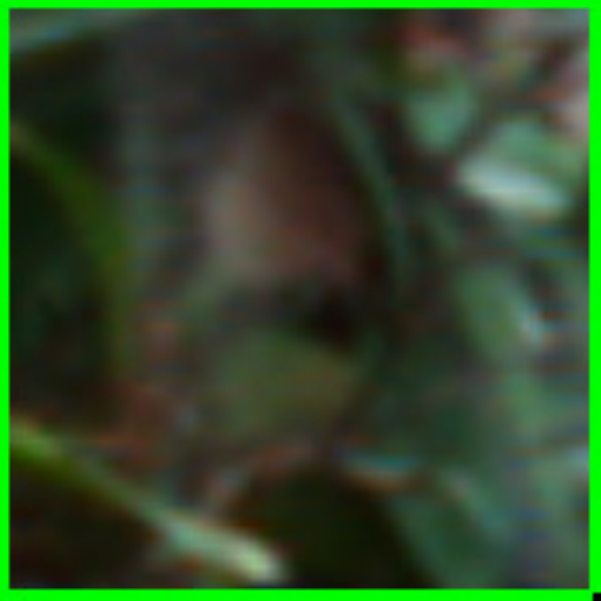} &
            \includegraphics[width=\ssubwidth\linewidth]{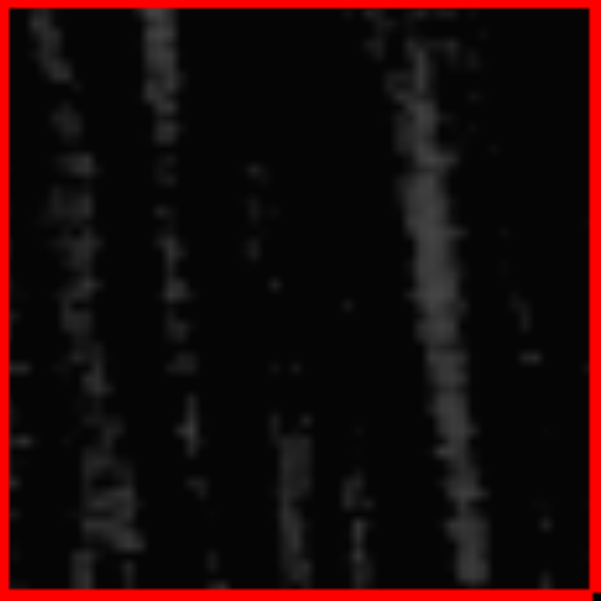} &
            \includegraphics[width=\ssubwidth\linewidth]{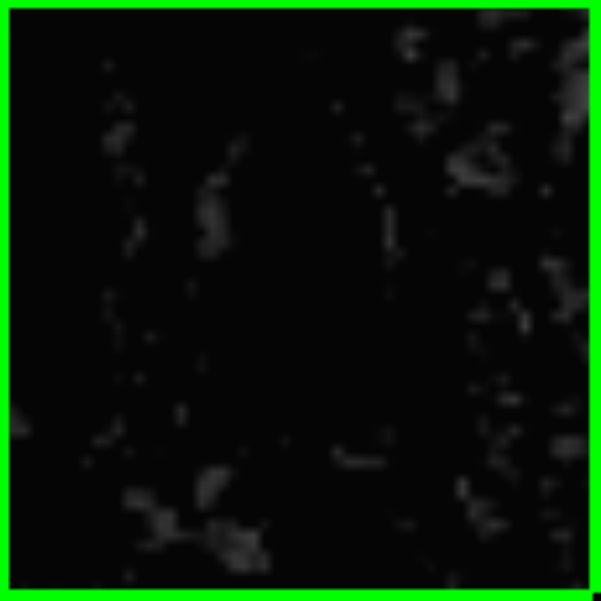} &
            \includegraphics[width=\ssubwidth\linewidth]{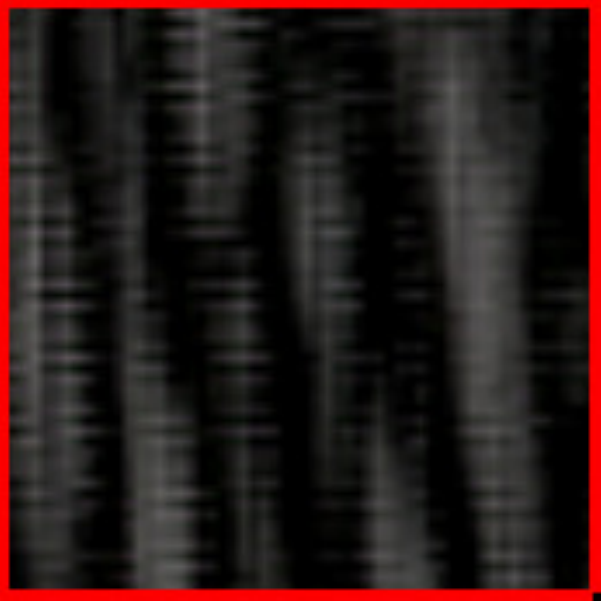} &
            \includegraphics[width=\ssubwidth\linewidth]{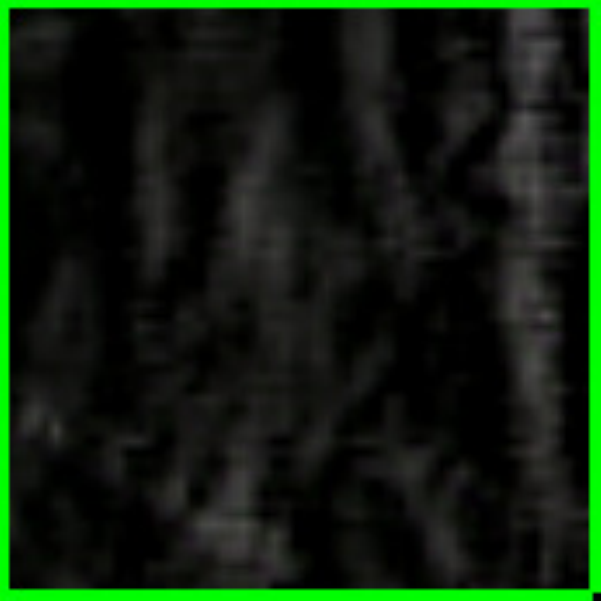} &
            \includegraphics[width=\ssubwidth\linewidth]{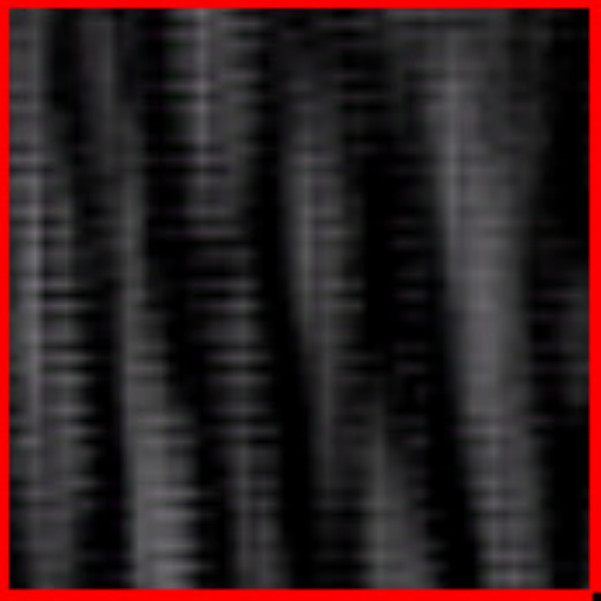} &
            \includegraphics[width=\ssubwidth\linewidth]{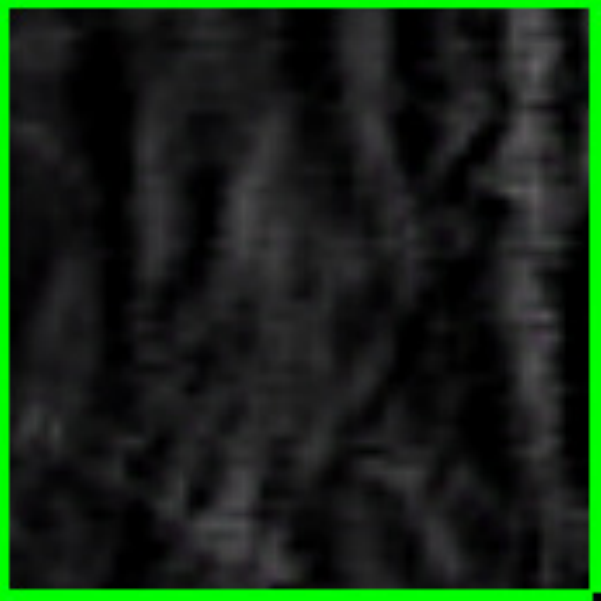} &
            \includegraphics[width=\ssubwidth\linewidth]{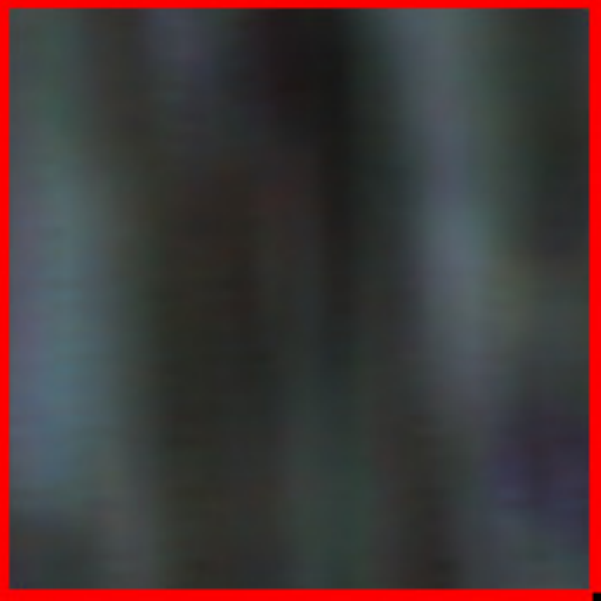} &
            \includegraphics[width=\ssubwidth\linewidth]{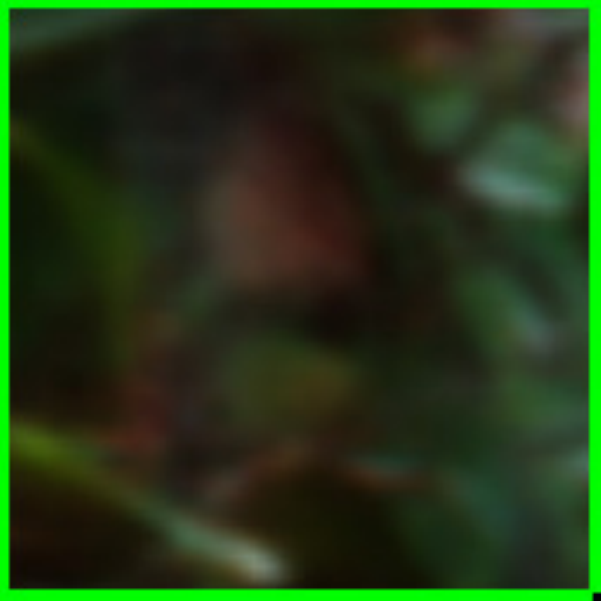} &
            \includegraphics[width=\ssubwidth\linewidth]{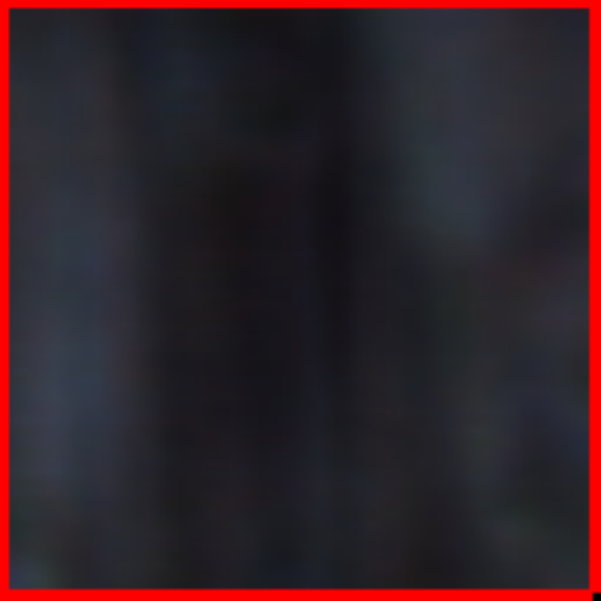} &
            \includegraphics[width=\ssubwidth\linewidth]{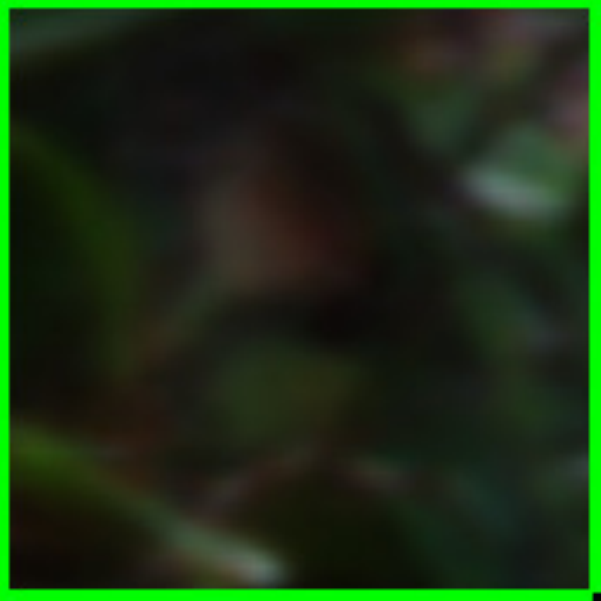} &
            \includegraphics[width=\ssubwidth\linewidth]{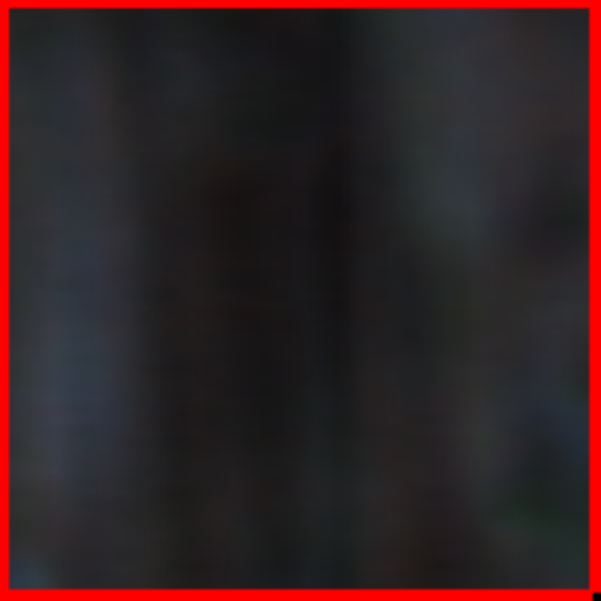} &
            \includegraphics[width=\ssubwidth\linewidth]{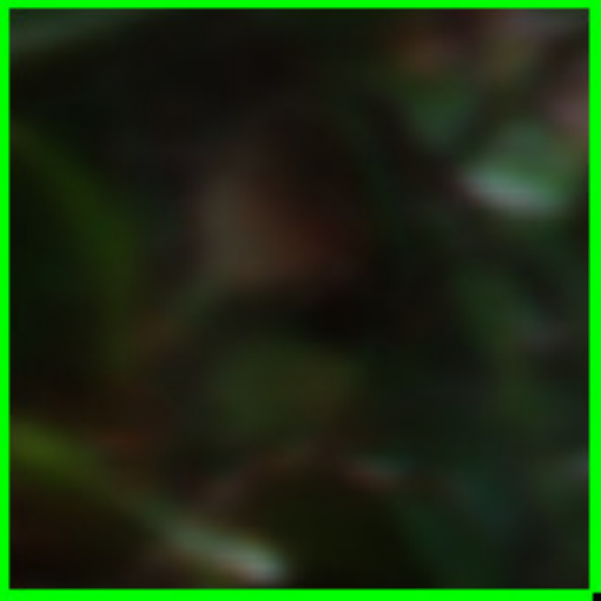}  \\

            \multicolumn{2}{c}{\scriptsize{(a) Input}}&
            \multicolumn{2}{c}{\scriptsize{(b) Rain w/o MSGP}}&
            \multicolumn{2}{c}{\scriptsize{(c) Rain w/ MGP}}&
            \multicolumn{2}{c}{\scriptsize{(d) Rain w/ MSGP}}&
            \multicolumn{2}{c}{\scriptsize{(e) Output w/o MSGP}}&
            \multicolumn{2}{c}{\scriptsize{(e) Output w/ MGP}}&
            \multicolumn{2}{c}{\scriptsize{(f) Output w/ MSGP}}\\
		\end{tabular}
	\end{center}
	\vspace{-0.016\textwidth}
	\caption{Ablation Study of our 4D-MGP-SRRNet on MSGP tested on a real-world rainy LFI.}
	\label{fig:GP}
	\vspace{-0.016\textwidth}
\end{figure*}

\renewcommand{\subwidth}{0.132}
\renewcommand{\ssubwidth}{0.066}
\begin{figure*}[h]
	\renewcommand{\tabcolsep}{0.8pt}
	\renewcommand\arraystretch{0.8}
	\begin{center}
		\begin{tabular}{cccccccccccccc}

            \multicolumn{2}{c}{\includegraphics[width=\subwidth\linewidth]{fig2/test/input/1007_mark}} &
            \multicolumn{2}{c}{\includegraphics[width=\subwidth\linewidth]{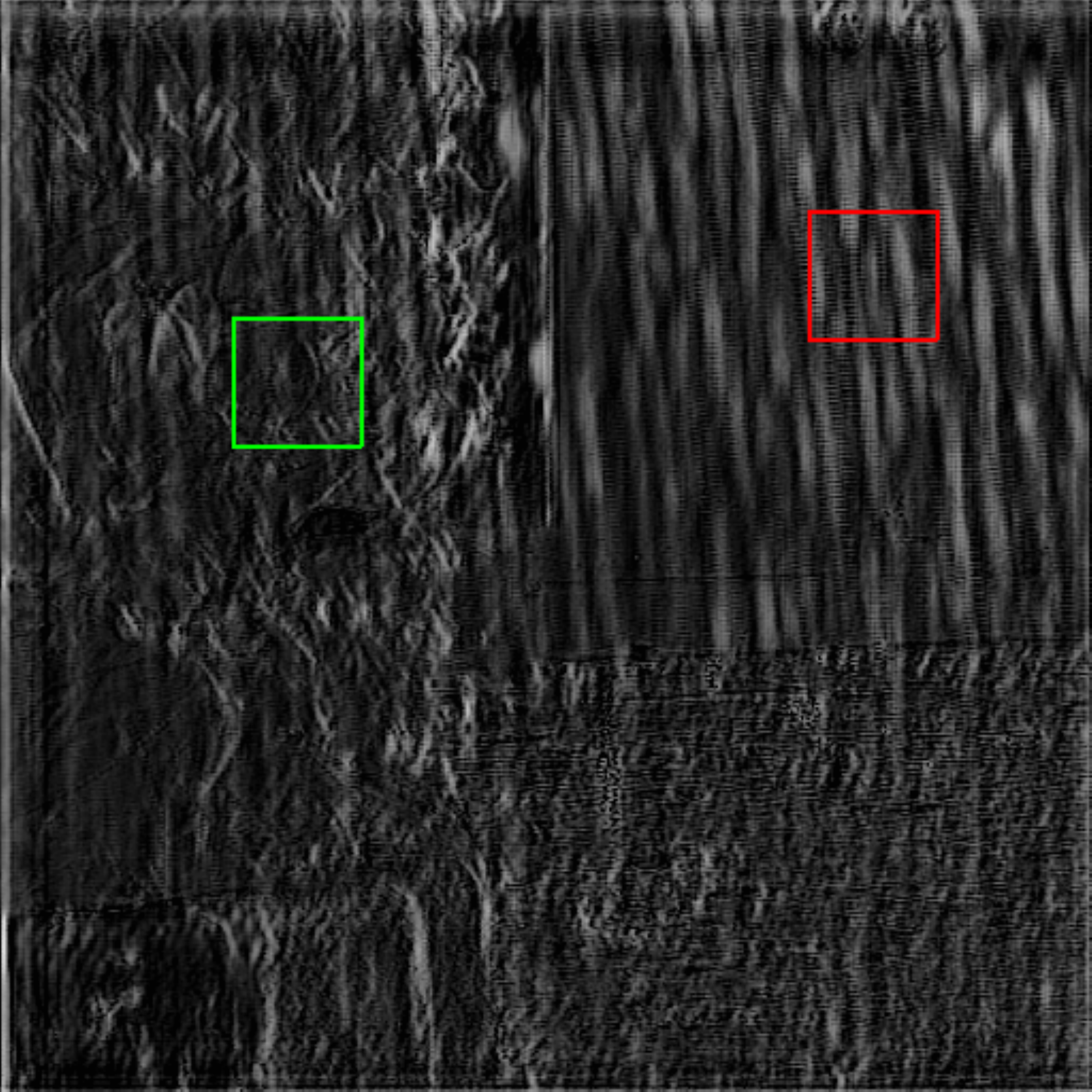}} &
            \multicolumn{2}{c}{\includegraphics[width=\subwidth\linewidth]{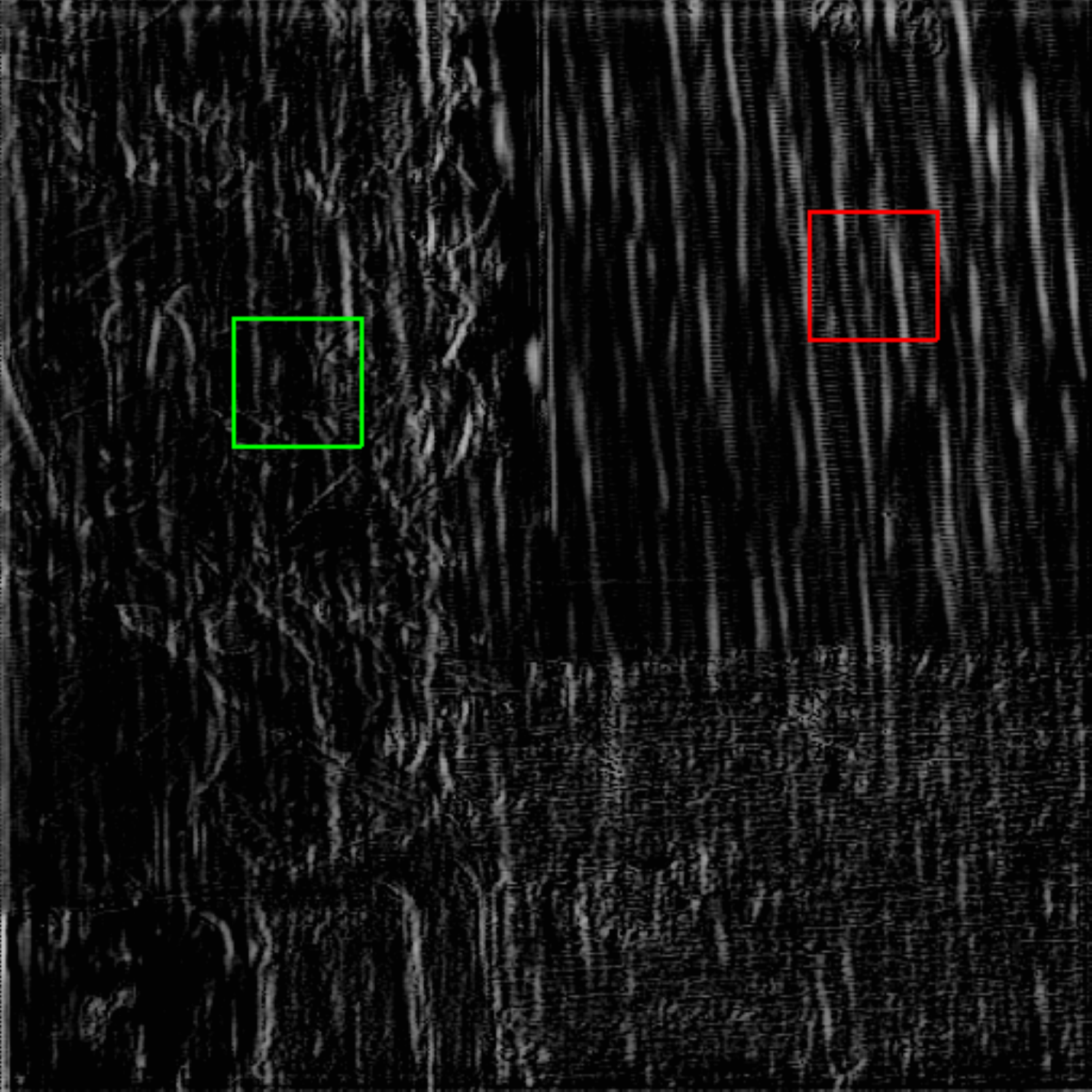}} &
            \multicolumn{2}{c}{\includegraphics[width=\subwidth\linewidth]{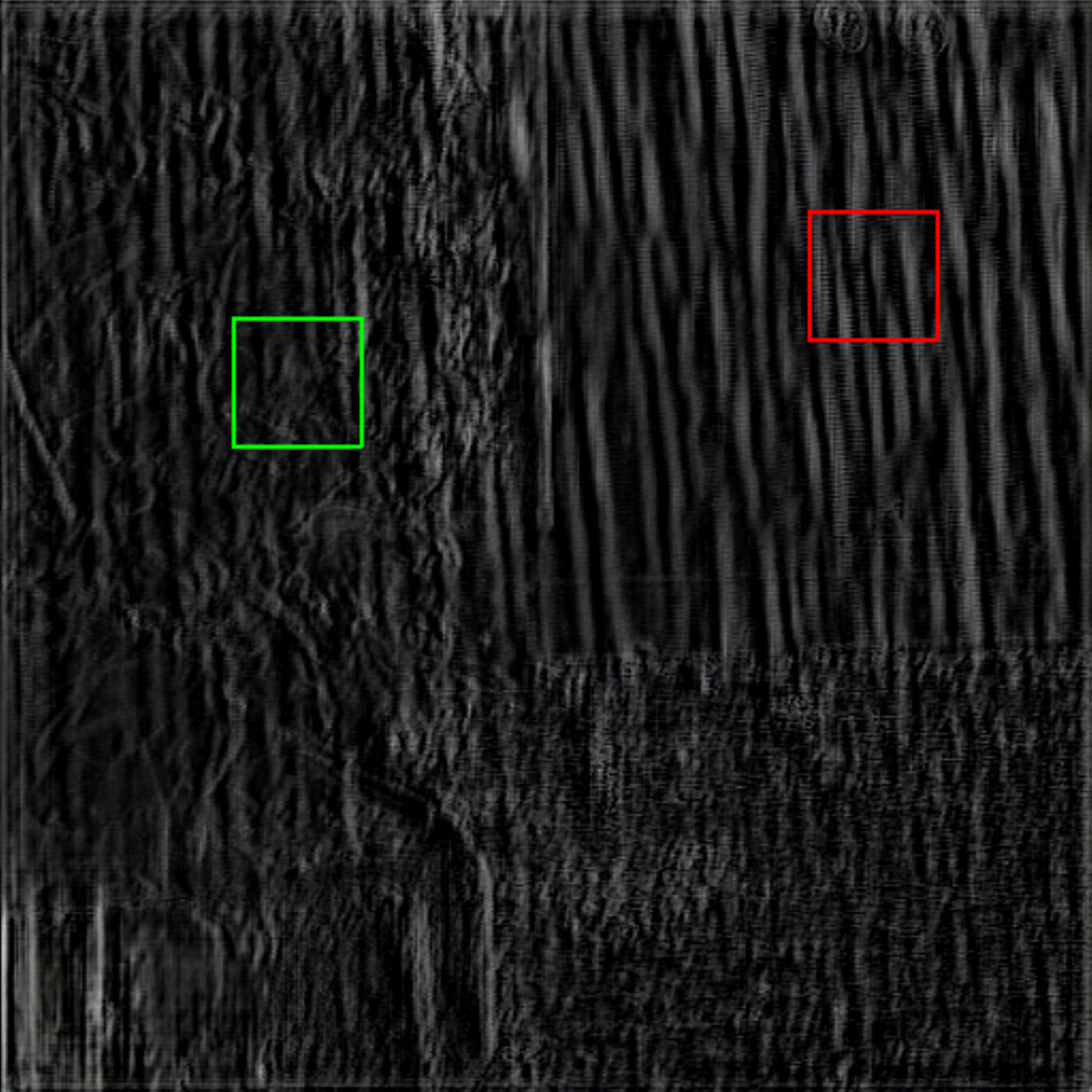}} &
            \multicolumn{2}{c}{\includegraphics[width=\subwidth\linewidth]{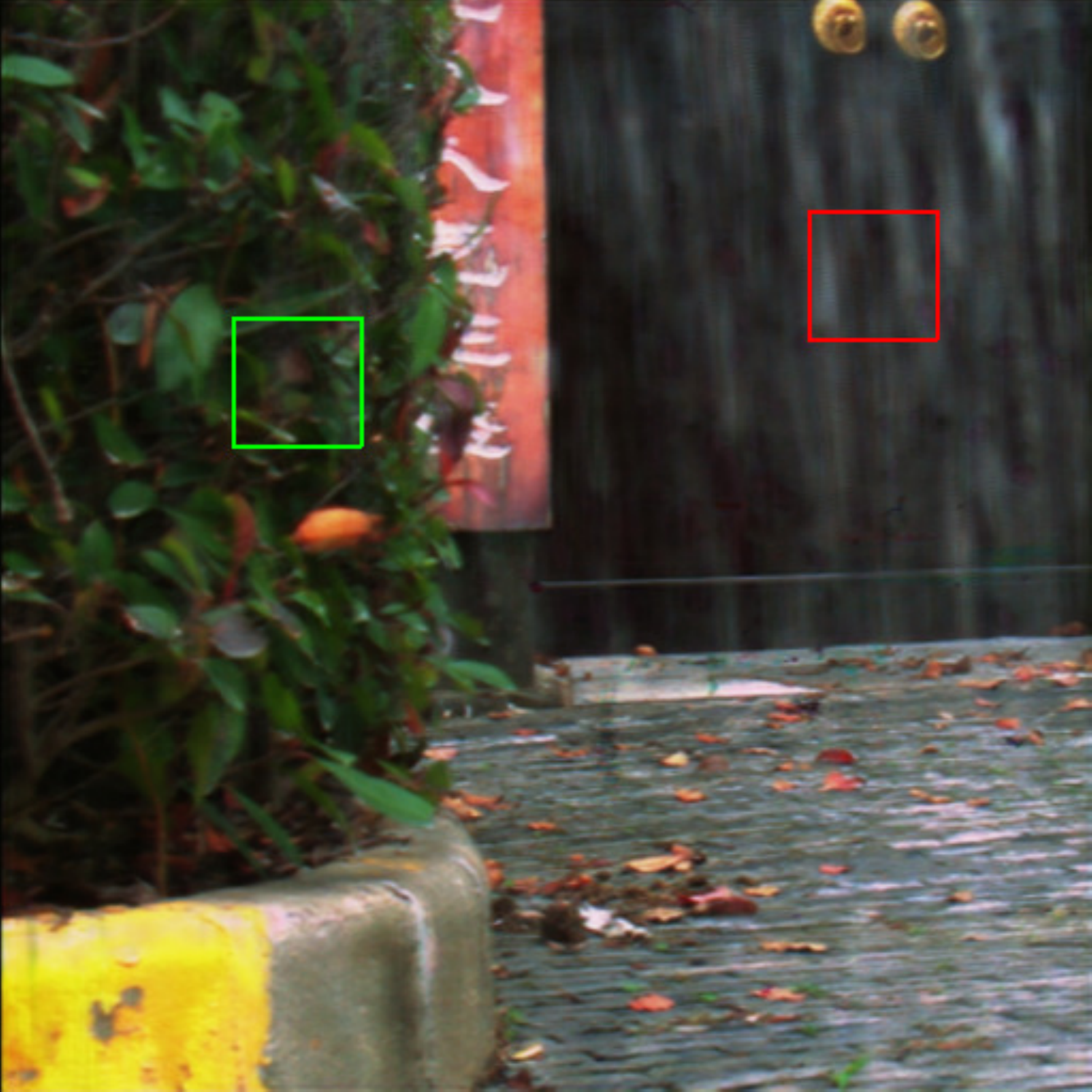}} &
            \multicolumn{2}{c}{\includegraphics[width=\subwidth\linewidth]{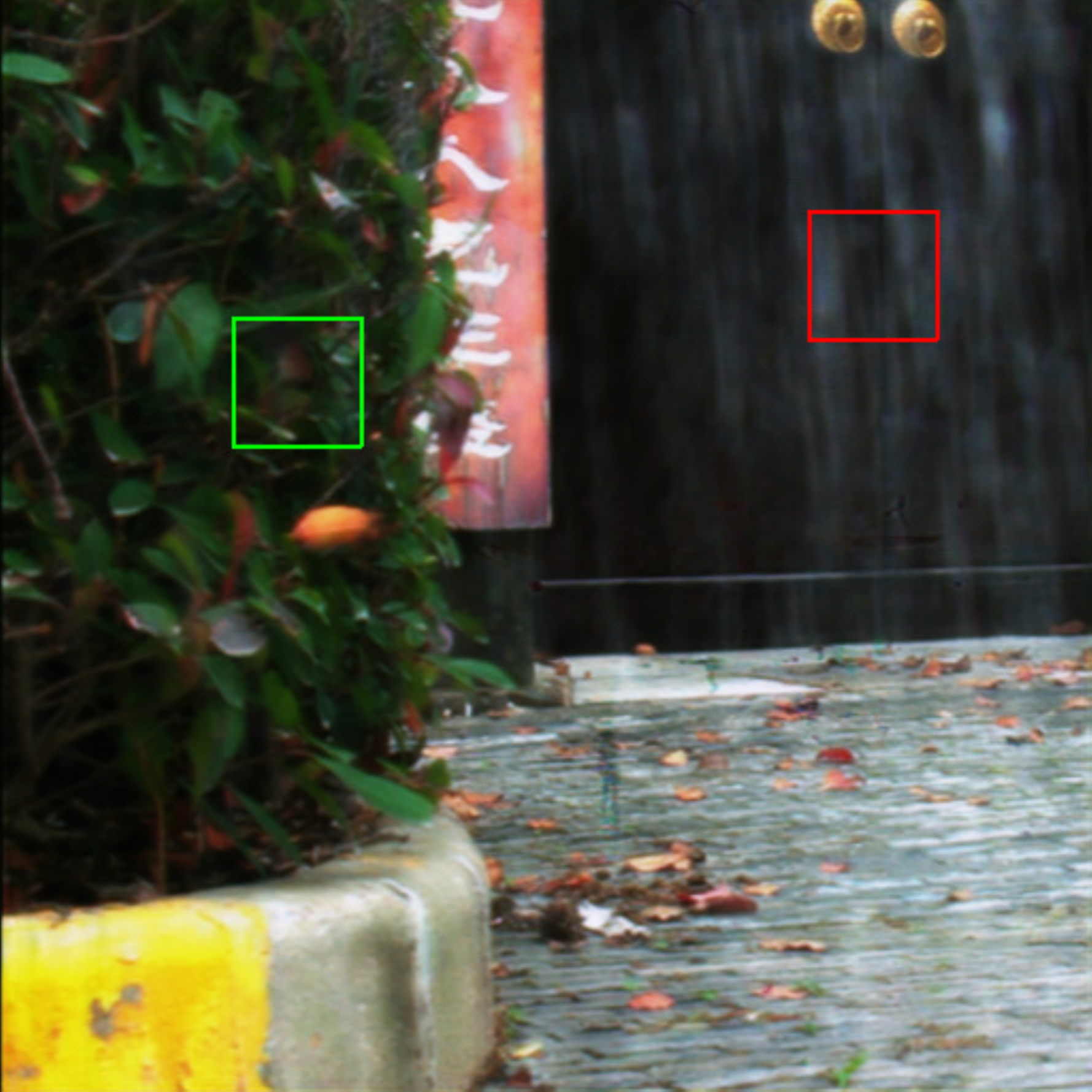}} &
            \multicolumn{2}{c}{\includegraphics[width=\subwidth\linewidth]{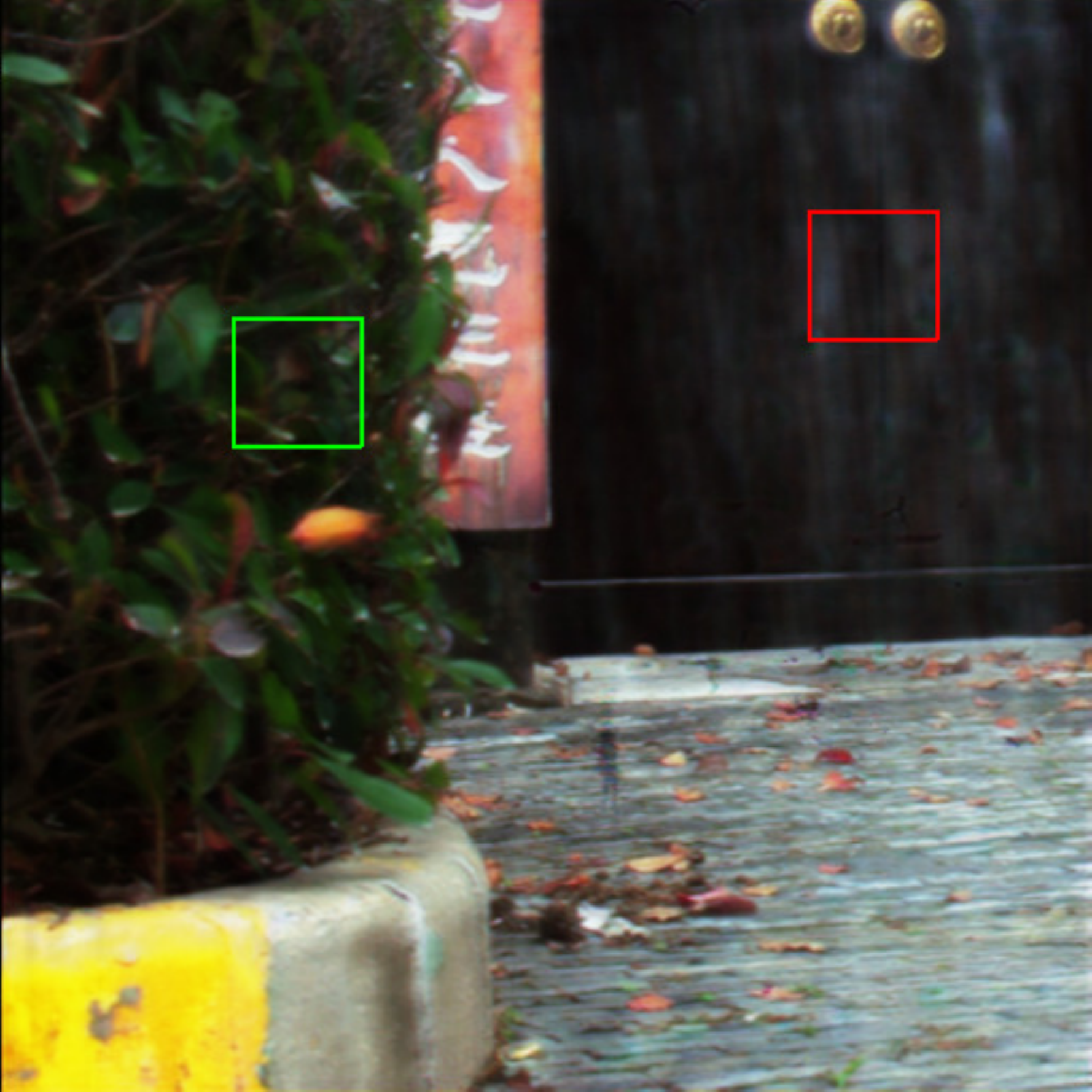}} \\

            \includegraphics[width=\ssubwidth\linewidth]{fig2/test/input/1007_mark1} &
            \includegraphics[width=\ssubwidth\linewidth]{fig2/test/input/1007_mark2} &
            \includegraphics[width=\ssubwidth\linewidth]{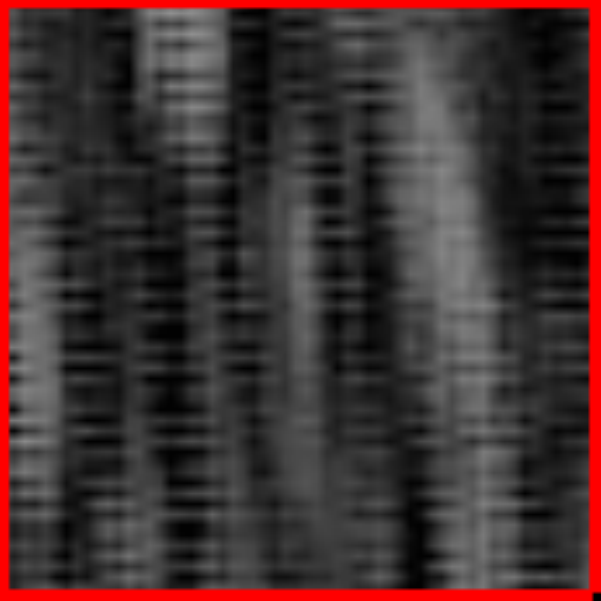} &
            \includegraphics[width=\ssubwidth\linewidth]{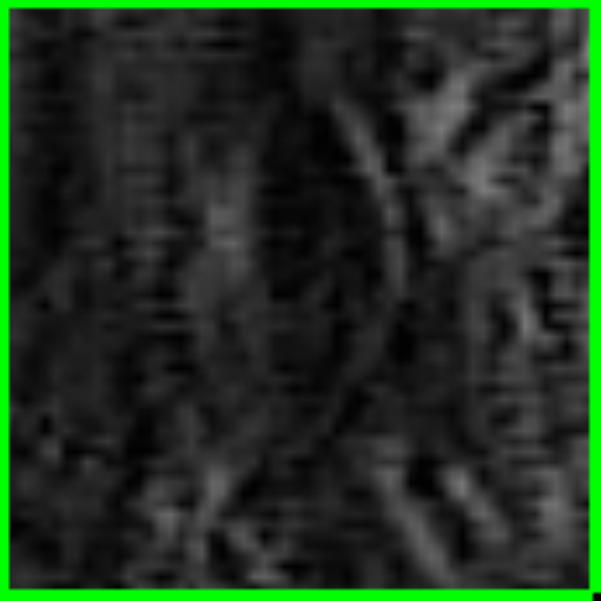} &
            \includegraphics[width=\ssubwidth\linewidth]{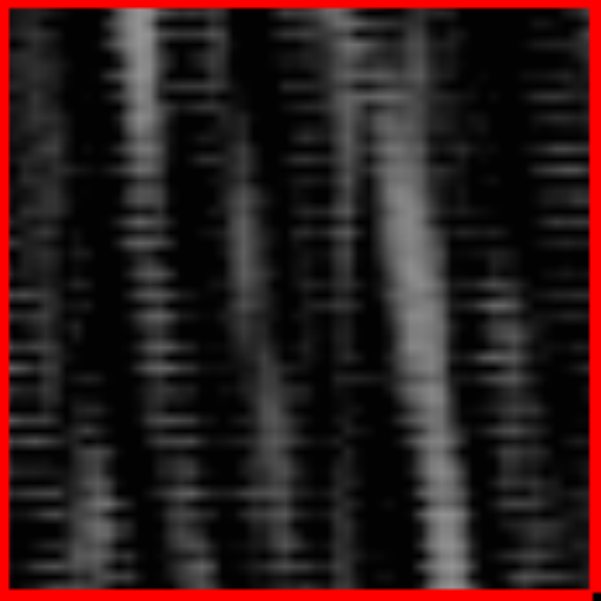} &
            \includegraphics[width=\ssubwidth\linewidth]{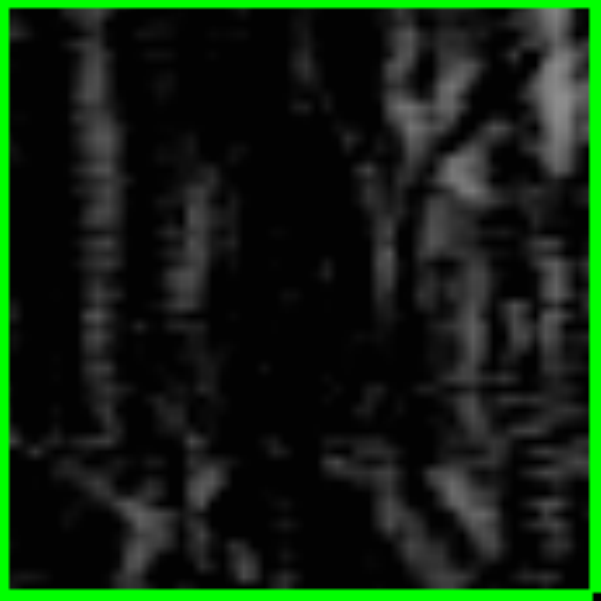} &
            \includegraphics[width=\ssubwidth\linewidth]{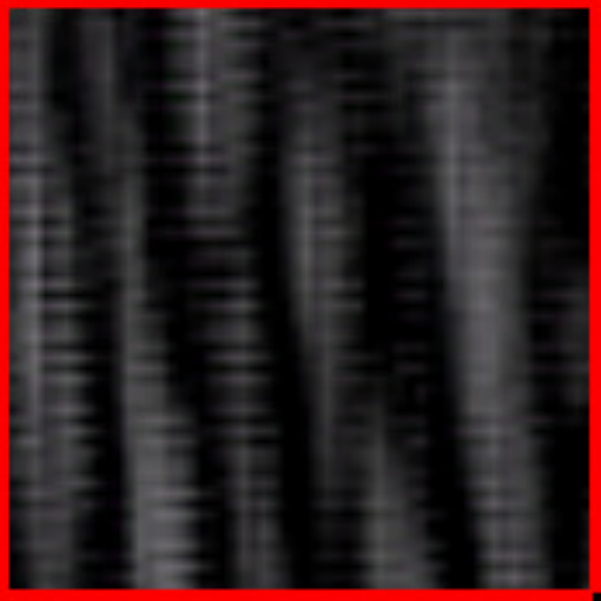} &
            \includegraphics[width=\ssubwidth\linewidth]{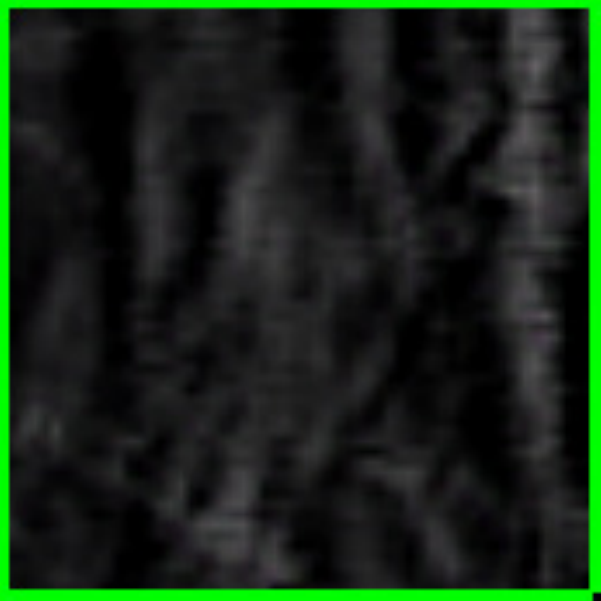} &
            \includegraphics[width=\ssubwidth\linewidth]{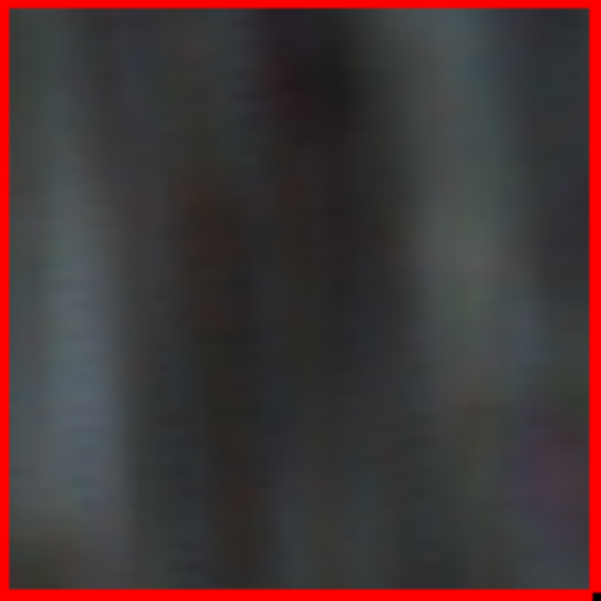} &
            \includegraphics[width=\ssubwidth\linewidth]{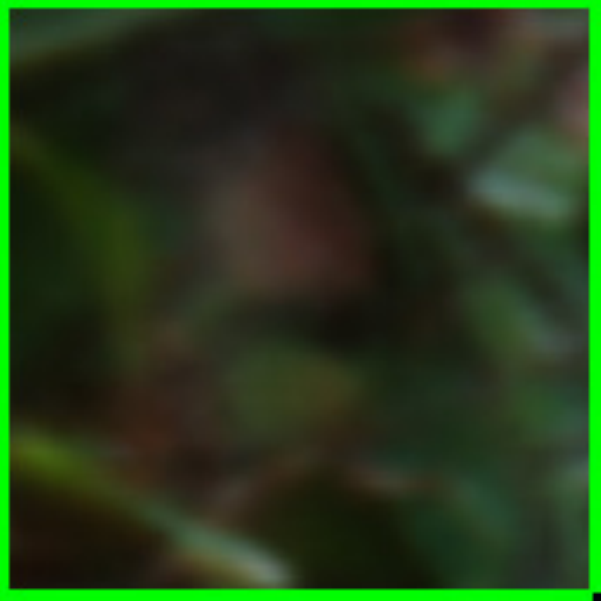} &
            \includegraphics[width=\ssubwidth\linewidth]{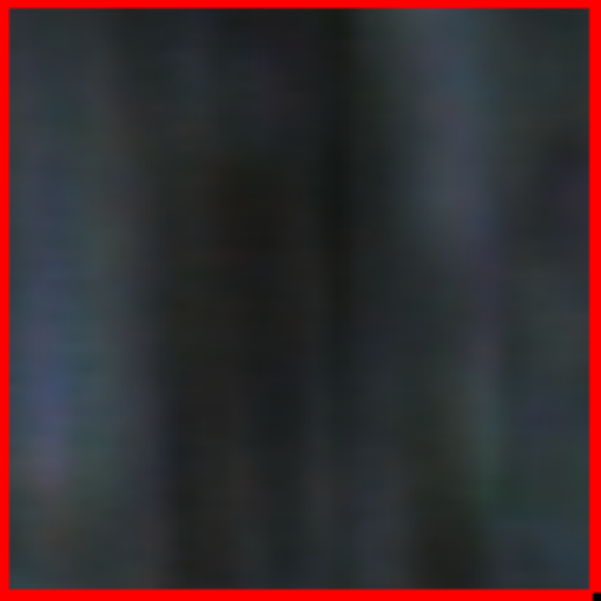} &
            \includegraphics[width=\ssubwidth\linewidth]{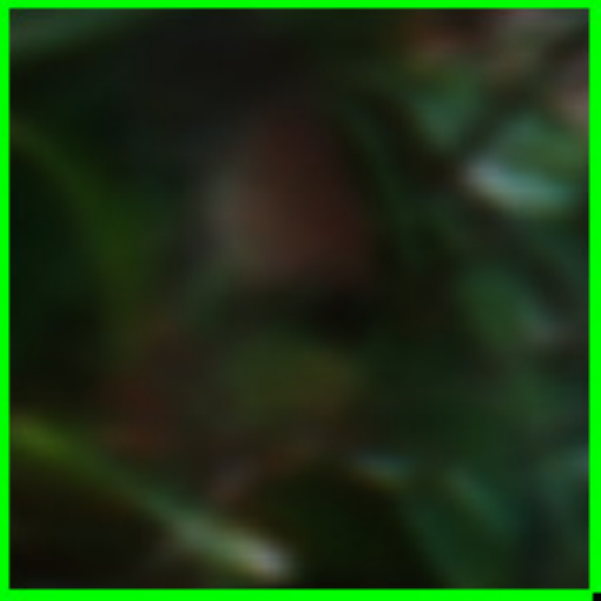} &
            \includegraphics[width=\ssubwidth\linewidth]{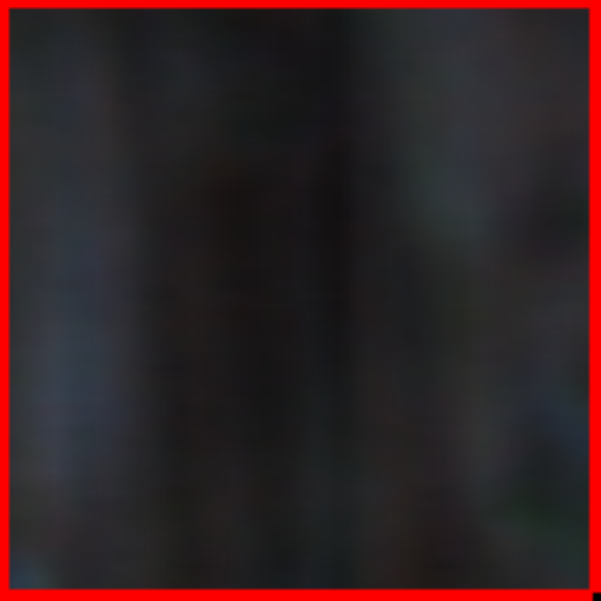} &
            \includegraphics[width=\ssubwidth\linewidth]{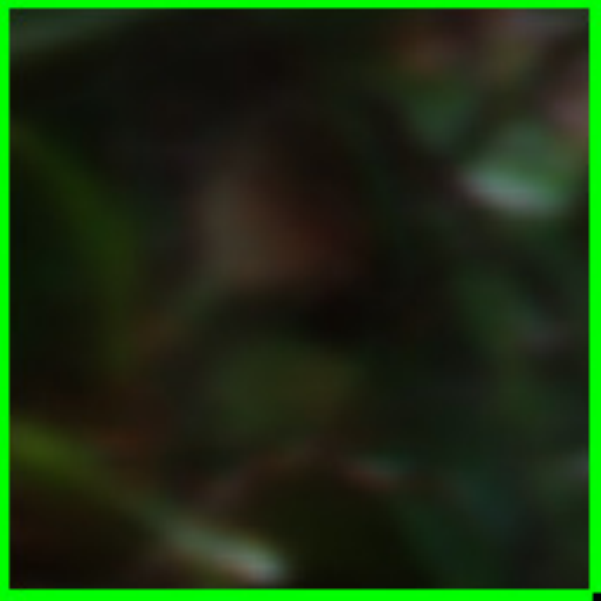}  \\

            \multicolumn{2}{c}{\scriptsize{(a) Input}}&
            \multicolumn{2}{c}{\scriptsize{(b) Rain w/ 2D}}&
            \multicolumn{2}{c}{\scriptsize{(c) Rain w/ 3D}}&
            \multicolumn{2}{c}{\scriptsize{(d) Rain w/ 4D}}&
            \multicolumn{2}{c}{\scriptsize{(e) Output w/ 2D}}&
            \multicolumn{2}{c}{\scriptsize{(e) Output w/ 3D}}&
            \multicolumn{2}{c}{\scriptsize{(f) Output w/ 4D}}\\
		\end{tabular}
	\end{center}
	\vspace{-0.016\textwidth}
	\caption{Ablation Study of our 4D-MGP-SRRNet using 2D/3D/4D convolution on a real-world rainy LFI.}
	\label{fig:4Dnet}
	\vspace{-0.016\textwidth}
\end{figure*}

\subsection{Qualitative Evaluation}
\label{subsec:qualitative_evaluation}

We conduct qualitative evaluation on the carefully collected challenging dataset containing $100$ real-world rainy LFIs. Since the process of method~\cite{Li19a} for 2D image rain streak removal is similar to that of our rain streak removal from LFIs, and method~\cite{ren20} performs very well on synthetic rainy LFIs, we especially compare our method with these two methods~\cite{Li19a},~\cite{ren20} and the semi-supervised methods~\cite{wei19,yasarla20,ding21} on several challenging real-world LFIs, as shown in Fig.~\ref{fig:real result}. The results demonstrate that our method with the global-local discriminator performs much better than the compared methods, and achieves state-of-the-art performances. Rain streaks and fog in the rain-free center sub-view restored by our method are almost completely removed. Method~\cite{Li19a} removes part of the rain streaks, but introduces a lot of noise. The performance of~\cite{ren20} is inferior to that of~\cite{wei19}. It cannot clearly remove the rain streaks and introduces a lot of noise to the de-rained images. \cite{yasarla20} performs much better and removes most of the rain streaks. \cite{ding21} removes most of the rain streaks in the first-row and third-row scenes, but it introduces blur in the de-rained result in the second-row scenes. Further, it cannot remove large and complex rain streaks with motion blur in the first-row and fourth-row scenes.
\revised{In Fig.~\ref{fig:real result}, the two latest methods~\cite{xiao22,zamir22} fail to clearly remove even obvious rain streaks, which means that their performances on challenging real-world scenes are not as good as those on synthetic scenes.}

In conclusion, results produced by these methods are not as good as the de-rained results obtained by our method, which nearly removes all complex rain streaks and without introducing any blur. It is worth noting that the de-rained images produced by our method may be a little darkened in some scenes such as the third and last two scenes of Fig.~\ref{fig:real result}, and some rain streaks may still be there, such as the enlarged patch of the first row of Fig.~\ref{fig:real result}.
%\revised{The high-quality rain-free images obtained can better assist other computer vision tasks, such as image recognition, target detection, etc.}

Fig.~\ref{fig:synthetic result_rain} shows rain streaks estimated by our method and existing methods~\cite{li14,Li19a,jiang2020,ding21}, accompanying with their corresponding PSNR/SSIM values. \cite{li14} performs poorly in the two scenes. Although the rain streaks estimated by~\cite{Li19a} seem acceptable, it misses some obvious rain streaks and fails to detect shapes/boundaries of tiny/large rain streaks. \cite{jiang2020} incorrectly takes many background details as rain streaks. \cite{ding21} obtains more accurate rain streaks on synthetic scenes, while its effect on real-world scenes is not good (Fig.~\ref{fig:rain streaks}). In contrast, our MGPDNet obtains more accurate rain streaks than the other methods. Accurate rain streak detection provides critical shape and position information of rain streaks for the subsequent rain removal network RNNAT to restore clean de-rained LFI well.
%What's more worth mentioning is that the proposed more accurate rain streak detection method can be applied to the detection of trivial, translucent objects that resemble rain streaks.}

We also compare depth maps obtained by our method and the deep learning-based methods~\cite{Li19a,tsai20,ding21}, as shown in Fig.~\ref{fig:depth}. Our method is able to estimate satisfactory depth maps for rainy LFIs, and its performances are only a little worse than the state-of-the-art LFI depth estimation method~\cite{tsai20} on the synthetic rainy LFI (the first row of Fig.~\ref{fig:depth}). In conclusion, our DERNet can obtain satisfactory depth maps for all sub-views of a rainy LFI.

\revised{More experimental results for real-world scenes can be found at our project page: \url{https://github.com/YT3DVision/4D-MGP-SRRNet}.}

\subsection{Ablation Study}
\label{subsec:ablation_study}

\textbf{MGPDNet and RNNAT:} We conduct ablation experiments to evaluate the effectiveness of MGPDNet and RNNAT in our network. Tab.~\ref{tab:AB} shows the ablation study of MGPDNet with/without non-local and MSGP, and RNNAT with/without depth maps estimated by DERNet for LFI deraining. In addition, we evaluate the performance of our network while replacing the basic block DSTB of RNNAT with DenseBlock.

Fig.~\ref{fig:GP} shows the estimated rain streaks and de-rained center-view produced by our network with or without MSGP, with simple MGP, respectively. It demonstrates the effectiveness/role of MSGP in 4D-MGP-SRRNet.%At the same time, the value of PSNR and SSIM show that the network structure we designed is the best choice.}

\setlength\tabcolsep{5pt}
\begin{table}[h]
	\centering
    \caption{Ablation study on the basic modules of our proposed network. DB means dense block. RB means residual block.}
	\label{tab:AB}
    \vspace{-0.01\textwidth}
    \begin{tabular*}{7.5cm}{ccccc|cc}
		\hline
		\multicolumn{2}{c|}{MGPDNet}&\multicolumn{3}{c|}{RNNAT}&\multicolumn{2}{c}{Metric}\\
		\hline
		Non-local & MSGP  & Depth  & DSTB & DB & PSNR & SSIM \\
		\hline
        $\surd$ & $\surd$ &$\surd$ &        & $\surd$ &    27.90    &    0.924   \\
        $\surd$ & $\surd$ &        &$\surd$ &         &    28.32    &    0.927  \\
        $\surd$ &         &$\surd$ &$\surd$ &         &    29.12    &    0.932    \\
                & $\surd$ &$\surd$ &$\surd$ &         &    29.35    &    0.950   \\
        $\surd$ & $\surd$ &$\surd$ &$\surd$ &         & \textbf{29.89} & \textbf{0.959}\\
		\hline
	\end{tabular*}
	\vspace{-0.016\textwidth}
\end{table}

\setlength\tabcolsep{2.6pt}
\begin{table}[h]
	\centering
	\caption{Comparison of our network deraining with 2D/3D/4D convolutions tested on synthetic LFIs coming from RLMB.}
	\label{tab:4Dnet}
    \vspace{-0.01\textwidth}
	\begin{tabular*}{6cm}{c|ccc}
		\hline
		 & Net-2D conv&  Net-3D conv& Net-4D conv \\
		\hline
        PSNR & 27.82 &  29.06 & \textbf{29.89} \\
        SSIM & 0.884 & 0.908 & \textbf{0.959} \\
		\hline
	\end{tabular*}
	\vspace{-0.016\textwidth}
\end{table}

\setlength\tabcolsep{1.5pt}
\begin{table*}[h]
	\centering
	\caption{Time-complexity, params-complexity and GFLOPs for deraining methods evaluated on the test set of synthetic rainy LFIs.}
	\label{tab:time-params-Gflops-complexity}
	\begin{tabular*}{\linewidth}{c|cccccccccccccc}
		\hline
		Methods & Wang~\cite{wang19b}&Li~\cite{Li19a}&Wei~\cite{wei19}&Jiang~\cite{jiang2020}&Yang~\cite{Yang20b} &Ren~\cite{ren20}&Jiang~\cite{jiang20}&Yasarla~\cite{yasarla20}
&Zamir~\cite{zamir2021}&Hu~\cite{hu2021}&Ding~\cite{ding21} &Xiao~\cite{xiao22} &Zamir~\cite{zamir22}& Ours\\
		\hline
        Ave.inf.time (s)& 1.12 & 0.35 &  1.54 & 0.30 &  \textbf{0.12} & 0.38 & 2.16 &0.28 & 0.30 &  \emp{0.18} & 0.36 & 0.85 & 0.39 &  0.48 \\
		Params(M) & 2.10 & 40.60 &  \textbf{0.07}& 0.98 & 4.17& 0.41 & \emp{0.28} &2.62& 3.64 &  4.03 & 13.24 & 16.39  & 26.10 &  14.91  \\
        GFLOPs       &  9.06 & 50.08 & \textbf{1.89} &10.07 & 68.18 & 24.56 &151.36 & 5.32  & 426.8 & \emp{4.93} &35.97 & 14.48   & 35.29 & 47.32\\
		\hline
	\end{tabular*}
\end{table*}

\textbf{2D/3D/4D Convolutions:} We conduct an experiment to evaluate the deraining effect of our network using 2D/3D/4D convolutions. The quantitative analysis is shown in Tab.~\ref{tab:4Dnet} and Fig.~\ref{fig:4Dnet}. Fig.~\ref{fig:4Dnet} shows that our 4D-MGP-SRRNet using 2D convolutions incorrectly takes a lot of background details as rain streaks and cannot clearly remove the rain streaks. Our 4D-MGP-SRRNet using 3D convolutions can only extract part of the rain streaks and the rain removal effect is also inferior to our 4D-MGP-SRRNet adopting 4D convolutions.

\revised{Tab.~\ref{tab:time-params-Gflops-complexity} reports the average tested time for each LFI coming from the test subset of our RLMB dataset, model parameters and GFLOPs of the competing methods and our network. From the second row, it can be seen that the amount of model parameters of our network is smaller than the state-of-the-art Transformer-based single image rain streak removal methods[20,21].}

\section{Conclusion}
In this paper, we have proposed a progressive network called 4D-MGP-SRRNet for detecting and removing rain streaks from LFIs. By leveraging 4D convolution, our network can make full use of all sub-views of LFIs to exploit abundant textural and structural information embedding in LFIs. We propose the MSGP module for accurate semi-supervised learning-based rain streak detection, which improves the generalization and performance of our network for real-world LFIs. More accurate depth maps are predicted from the results of rain streaks subtracted from rainy sub-views for fog estimation, and de-rained sub-views are produced by the powerful RNNAT.
We also propose a new rainy LFI dataset RLMB, which consists of both synthetic and real-world rainy LFIs. Extensive experiments on both synthetic and real-world rainy LFIs demonstrate that our proposed method has great superiority over other state-of-the-art methods. However, since it is difficult to collect abundant real-world rainy LFIs, our proposed 4D-MGP-SRRNet is learned on a relatively small size real-world LFI dataset. In addition, the computation cost for rain removal from LFIs is higher than that from regular images.

%\section{Acknowledgements}
%We would like to thank the anonymous reviewers and the associate editor for their valuable comments and suggestions on this paper. We also thank Weijiang He and Xiangjie Zhu (master students of Jiangnan University) for comparing our network with the competing methods. This work was supported by the National Natural Science Foundation of China (Grant No. 61902151), the Natural Science Foundation of Jiangsu Province, China (Grant No. BK20170197), and two Strategic Research Grants from City University of Hong Kong (Ref.: 7005674 and 7005843).

\begin{comment}
\begin{IEEEbiographynophoto}{Jane Doe}
Biography text here without a photo.
\end{IEEEbiographynophoto}

\begin{IEEEbiography}[{\includegraphics[width=1in,height=1.25in,clip,keepaspectratio]{fig1.png}}]{IEEE Publications Technology Team}
In this paragraph you can place your educational, professional background and research and other interests.\end{IEEEbiography}
\end{comment}

\bibliographystyle{ieeetr}
\bibliography{rain_removal}%

\end{document}